\documentclass[12pt,fleqn]{tsp}

\usepackage[hang,flushmargin]{footmisc}
\techscience{vol}{no}{CMES 2023}{0}
\techscience{vol}{no}{First online at CMES on 2023.03.01, DOI: \href{http://dx.doi.org/10.32604/cmes.2023.028130}{10.32604/cmes.2023.028130} - arXiv v3}{0}

\license{\includegraphics[width=16cm,height=1.5cm]{license}}
\def\thedoi{First online at CMES on 2023 Mar 01 with DOI: \href{http://dx.doi.org/10.32604/cmes.2023.028130}{10.32604/cmes.2023.028130}}
\def\thetype{REVIEW}

\graphicspath{{./image/}}% Images path
\runningtitle{Manuscript Preparation for TSP}

\usepackage{textcomp}

\usepackage{amsmath}
\usepackage{mathtools}
\usepackage{amssymb}
\usepackage{amsthm}
\usepackage{upgreek}

\usepackage{mathrsfs}

\usepackage[mathscr]{euscript}
 \let\mathscr\relax

\usepackage[scr]{rsfso}
\usepackage{pifont}

\usepackage[linesnumbered,lined,commentsnumbered]{algorithm2e}

\usepackage{accents}
\DeclareMathSymbol{\widehatsym}{\mathord}{largesymbols}{"62}
\DeclareMathSymbol{\widetildesym}{\mathord}{largesymbols}{"65}
\usepackage{stackengine}
\usepackage{scalerel}

\usepackage{bbm}

\usepackage[dvipsnames]{xcolor}

\usepackage{graphicx,float,wrapfig}
\usepackage{subcaption}

\usepackage{tabularx,booktabs}
\newcolumntype{Y}{>{\centering\arraybackslash}X}

\usepackage{multirow}

\usepackage{times}

\usepackage[margin=1cm]{caption}

\usepackage{enumitem}
\setlist[enumerate,1]{label={(\arabic*)}}

\usepackage{tikz}
\usetikzlibrary{arrows,intersections,calc}
\usepackage{pgfplots}
\pgfplotsset{compat=1.14}
\usepackage[pagebackref]{hyperref}
\hypersetup{
	colorlinks = true,
	linkbordercolor = {white},
	linkcolor=blue,
	citecolor=[RGB]{0,50,0},
	urlcolor=[RGB]{100,0,0},
}

\usepackage[normalem]{ulem}
\usepackage{soul}

\usepackage{cancel}

\usepackage{clesm_defs_1}
\newtheorem{rem}{Remark}[section]

\usepackage[yyyymmdd]{datetime}

\usepackage[]{siunitx}
\usepackage{psfrag}

\usepackage[title]{appendix}

\newcommand{\g}{a}
\newcommand{\anneal}{\mathfrak{a}}
\newcommand{\bias}{b}
\newcommand{\Bias}{B}
\newcommand{\bbias}{\boldsymbol{\bias}}

\newcommand{\biassp}[2]{\bias_{#1}^{(#2)}}
\newcommand{\bbiassp}[2]{\bbias_{#1}^{(#2)}}
\newcommand{\bbiasp}[1]{\boldsymbol \bias^{(#1)}}
\newcommand{\bBiasp}[1]{\boldsymbol \Bias^{(#1)}}

\newcommand{\class}{\mathcal C}
\newcommand{\classs}[1]{\kars \class #1}

\newcommand{\loss}{J}
\newcommand{\losst}{\widetilde{\loss}}

\newcommand{\losstr}{\losst_{reg}}

\newcommand{\decay}{\varpi}
\newcommand{\decaywt}{\mathfrak{d}}
\newcommand{\D}{L}

\newcommand{\dtb}[1]{\hbox{$#1$}\kern-1.4ex\raise+1.8ex\hbox{\bigdot}\kern+1.4ex}

\newcommand{\dtbs}[1]{\hbox{\scriptsize $#1$}\kern-1.0ex\raise+1.4ex\hbox{$\centerdot$}\kern+1.0ex}

\newcommand{\Hs}{H}
\newcommand{\bHs}{\boldsymbol{\Hs}}
\newcommand{\bHst}{\widetilde{\bHs}}
\newcommand{\Heigen}{\lambda}
\newcommand{\Heigent}{\widetilde{\Heigen}}
\newcommand{\Hevec}{v}
\newcommand{\bHevect}{\widetilde{\boldsymbol{\Hevec}}}

\newcommand{\down}{d}
\newcommand{\bd}{\boldsymbol{\down}}

\newcommand{\downt}{\widetilde{\down}}
\newcommand{\bdt}{\widetilde{\bd}}

\newcommand{\drift}{\text{\sc d}}

\newcommand{\tei}{\text{\fontfamily{pag}\selectfont t}}

\newcommand{\stdev}{\sigma}
\newcommand{\xpc}{\mathbb{E}}

\newcommand{\fluct}{\mathscr{F}}
\newcommand{\fluctS}{\mathscr{G}}
\newcommand{\covmat}{\mathfrak{C}}
\newcommand{\cmSDE}{\boldsymbol{B}}
\newcommand{\graderr}{\mathfrak{e}}
\newcommand{\graderrt}{\mathfrak{v}}
\newcommand{\friction}{\mathfrak{f}}

\newcommand{\bxt}{\bkart \x}
\newcommand{\mm}{m}
\newcommand{\mt}{\kart \mm}
\newcommand{\bmu}{\bkar \mu}
\newcommand{\bmut}{\bkart \mu}
\newcommand{\bmuh}{\widehat {\boldsymbol \mu}}
\newcommand{\f}{f}
\newcommand{\byt}{\widetilde {\boldsymbol \y}}
\newcommand{\byh}{\widehat {\boldsymbol \y}}
\newcommand{\glub}{\text{\fontfamily{pag}\selectfont s}}

\newcommand{\grad}{g}
\newcommand{\bgrad}{\boldsymbol{\grad}}
\newcommand{\bgradp}[1]{\bgrad^{(#1)}}
\newcommand{\gradt}{\widetilde{\grad}}
\newcommand{\bgradt}{\widetilde{\bgrad}}
\newcommand{\bgradtr}{\bgradt_{reg}}

\newcommand{\K}{\mathcal{K}}
\newcommand{\Ks}[1]{\K_{#1}}

\newcommand{\bfun}{\phi}
\newcommand{\bfuns}[1]{\kars \bfun #1}

\newcommand{\bfunV}{\Phi}
\newcommand{\bbfunV}{\bkar \bfunV}
\newcommand{\Gram}{\Gamma}
\newcommand{\Grams}[2]{\kars \Gram {#1 #2}}
\newcommand{\bGram}{\bkar \Gram}
\newcommand{\iGram}{\Delta}
\newcommand{\iGrams}[2]{\kars \iGram {#1 #2}}
\newcommand{\biGram}{\bkar \iGram}

\newcommand{\h}{h}
\newcommand{\bh}{\boldsymbol{\h}}

\newcommand{\x}{x}
\newcommand{\X}{X}
\newcommand{\bx}{\boldsymbol{\x}}
\newcommand{\bX}{\boldsymbol{\X}}
\newcommand{\bxp}[1]{\bx^{(#1)}}
\newcommand{\bXp}[1]{\bX^{(#1)}}
\newcommand{\xsp}[2]{\x_{#1}^{(#2)}}
\newcommand{\bxsp}[2]{\bx_{#1}^{(#2)}}

\newcommand{\learn}{\epsilon}
\newcommand{\blearn}{\boldsymbol{\learn}}

\newcommand{\bsize}{\text{\fontfamily{pag}\selectfont m}}
\newcommand{\Bsize}{\text{\fontfamily{pag}\selectfont M}}
\newcommand{\Ibb}{\mathbb{I}}

\newcommand{\Ibbps}[2]{\Ibb_{#1}^{|#2|}}
\newcommand{\M}{M}
\newcommand{\Ms}[1]{\M_{#1}}
\newcommand{\Xbb}{\mathbb{X}}
\renewcommand{\Bbb}{\mathbb{B}}
\newcommand{\Bbbp}[1]{\Bbb^{|#1|}}
\newcommand{\Bbbps}[2]{\Bbb_{#1}^{|#2|}}
\newcommand{\Ybb}{\mathbb{Y}}
\newcommand{\Tbb}{\mathbb{T}}
\newcommand{\Tbbp}[1]{\Tbb^{|#1|}}
\newcommand{\Tbbps}[2]{\Tbb_{#1}^{|#2|}}

\newcommand{\mompar}{\zeta}
\newcommand{\nesterov}{\gamma}

\newcommand{\mom}{m}
\newcommand{\bmom}{\boldsymbol{\mom}}
\newcommand{\bmomc}{\widehat{\bmom}}
\newcommand{\var}{V}
\newcommand{\bvar}{\boldsymbol{\var}}
\newcommand{\varc}{\widehat{\var}}
\newcommand{\bvarc}{\widehat{\bvar}}
\newcommand{\noise}{\mathfrak{n}}
\newcommand{\noiseSDE}{\eta}
\newcommand{\bnSDE}{\boldsymbol{\noiseSDE}}

\newcommand{\xup}{\textup{x}}
\newcommand{\bxup}{\boldsymbol{\xup}}
\newcommand{\bxupp}[1]{\bxup^{|#1|}}
\newcommand{\bexin}[1]{\bx^{|#1|}}
\newcommand{\bexout}[1]{\by^{|#1|}}
\newcommand{\bexoutt}[1]{\nfwidetilde{\by}^{|#1|}}
\newcommand{\yup}{\textup{y}}

\newcommand{\y}{y}
\newcommand{\Y}{Y}
\newcommand{\by}{\boldsymbol{\y}}

\newcommand{\bY}{\boldsymbol{\Y}}
\newcommand{\byp}[1]{\by^{(#1)}}
\newcommand{\bYp}[1]{\bY^{(#1)}}
\newcommand{\ysp}[2]{\y_{#1}^{(#2)}}
\newcommand{\bysp}[2]{\by_{#1}^{(#2)}}

\newcommand{\z}{z}
\newcommand{\Z}{Z}
\newcommand{\bz}{\boldsymbol{\z}}
\newcommand{\bZ}{\boldsymbol{\Z}}
\newcommand{\bzp}[1]{\bz^{(#1)}}
\newcommand{\bZp}[1]{\bZ^{(#1)}}
\newcommand{\zsp}[2]{\z_{#1}^{(#2)}}
\newcommand{\bzsp}[2]{\bz_{#1}^{(#2)}}

\newcommand{\param}{\theta}
\newcommand{\Param}{\Theta}
\newcommand{\bparam}{\boldsymbol{\param}}
\newcommand{\bParam}{\boldsymbol{\Param}}
\newcommand{\bparamp}[1]{\bparam^{(#1)}}
\newcommand{\bparamsp}[2]{\bparam_{#1}^{(#2)}}
\newcommand{\bParamp}[1]{\bParam^{(#1)}}
\newcommand{\paramsp}[2]{\param_{#1}^{(#2)}}
\newcommand{\bparamt}{\widetilde{\bparam}}
\newcommand{\Tparam}{P_T}

\renewcommand{\real}{\mathbb{R}}

\newcommand{\slope}{s}

\newcommand{\weight}{w}
\newcommand{\Weight}{W}
\newcommand{\bweightp}[1]{\boldsymbol \weight^{(#1)}}
\newcommand{\bweightsp}[2]{\boldsymbol \weight_{#1}^{(#2)}}
\newcommand{\bWeight}{\boldsymbol \Weight}
\newcommand{\weightsp}[2]{\weight_{#1}^{(#2)}}

\newcommand{\width}{m}
\newcommand{\widths}[1]{\width_{(#1)}}

\newcommand{\nfwidetilde}[1]{\stackon[-0.8em]{${#1}$}{\vstretch{1.5}{\hstretch{1.0}{\widetilde{\phantom{\;}}}}}}

\renewcommand{\dotprod}{{\scriptscriptstyle{\bullet}}}

\newcommand{\out}{\nfwidetilde{\y}}
\newcommand{\bout}{\nfwidetilde{\by}}

\newcommand{\expand}[1]{\overline{#1}}

\newcommand{\gravi}{g}
\newcommand{\bgrav}{\boldsymbol{\gravi}}
\newcommand{\velo}{v}
\newcommand{\bvelo}{\boldsymbol{\velo}}
\newcommand{\bvs}[1]{\bvelo_{#1}}
\newcommand{\bvM}{\bvs M}
\newcommand{\bvm}{\bvs m}
\newcommand{\bvelot}{\widetilde{\bvelo}}
\newcommand{\bvts}[1]{\bvelot_{#1}}
\newcommand{\bvtM}{\bvts M}
\newcommand{\bvtm}{\bvts m}
\newcommand{\e}{e}
\newcommand{\es}[1]{\e_{#1}}
\newcommand{\esM}{\es M}
\newcommand{\esm}{\es m}
\newcommand{\norm}{n}
\newcommand{\bnorm}{\boldsymbol{\norm}}
\newcommand{\q}{q}
\newcommand{\bq}{\boldsymbol{\q}}
\newcommand{\bqs}[1]{\bq_{#1}}
\newcommand{\bqM}{\bqs M}
\newcommand{\bqm}{\bqs m}
\newcommand{\bqt}{\widetilde{\bq}}
\newcommand{\bqts}[1]{\bqt_{#1}}
\newcommand{\bqtM}{\bqts M}
\newcommand{\bqtm}{\bqts m}
\newcommand{\diver}{\text{div}\,}

\newcommand{\jump}[1]{[\![ #1 ]\!]}
\newcommand{\T}{T}
\newcommand{\bT}{\boldsymbol{\T}}
\newcommand{\stress}{\sigma}
\newcommand{\bstress}{\boldsymbol{\stress}}
\newcommand{\domain}{\mathcal{B}}
\newcommand{\doms}[1]{\domain_{#1}}
\newcommand{\domp}[1]{\domain^{#1}}
\newcommand{\domps}[2]{\domain^{#1}_{#2}}
\newcommand{\surface}{\Gamma}
\newcommand{\surfp}[1]{\surface^{#1}}

\newcommand{\pc}{\beta}
\newcommand{\pcs}[1]{\pc_{#1}}
\newcommand{\bpc}{\boldsymbol{\pc}}

\newcommand{\abc}{}

\newcommand{\kr}[1]{#1}

\newcommand{\Bkr}[1]{{\bf{#1}}}

\newcommand{\krb}[1]{\overline{#1}}

\newcommand{\krt}[1]{\widetilde{#1}}

\newcommand{\kart}[1]{\abc \krt{#1} \abc{}}

\newcommand{\kars}[2]{\abc {\kr {#1}} _{#2}\abc{}}

\newcommand{\karp}[2]{\abc {\kr {#1}} ^{#2}\abc{}}

\newcommand{\karsp}[3]{\abc  {\kr {#1}} _{#2} ^{#3}\abc{}}

\newcommand{\bkr}[1]{{\bfg{#1}}}

\newcommand{\bkrt}[1]{{\widetilde{\bfg{#1}}}}

\newcommand{\bkrh}[1]{{\widehat{\bfg{#1}}}}

\newcommand{\bkar}[1]{\abc \bkr{#1}\abc{}}

\newcommand{\bkart}[1]{\abc \bkrt{#1}\abc{}}

\newcommand{\bkarh}[1]{\abc \bkrh{#1}\abc{}}

\newcommand{\bkars}[2]{\abc {\bkr {#1}} _{#2}\abc{}}

\newcommand{\bkarp}[2]{\abc {\bkr {#1}} ^{#2}\abc{}}

\newcommand{\bkarsp}[3]{\abc  {\bkr {#1}} _{#2} ^{#3}\abc{}}

\DeclareMathOperator{\diverg}{div}
\DeclareMathOperator{\gradient}{grad}
\DeclareMathOperator{\Var}{Var}

\newcommand{\dx}{\mathrm dx}

\newcommand{\gt}[1]{g^{[#1]}}

\newcommand{\bv}{\boldsymbol b}
\newcommand{\bvt}[1]{\boldsymbol b^{[#1]}}
\newcommand{\fvt}[1]{\fv^{[#1]}}
\newcommand{\cv}{\boldsymbol c}
\newcommand{\cvt}[1]{\cv^{[#1]}}

\newcommand{\dv}{\boldsymbol d}
\newcommand{\ev}{\boldsymbol e}

\newcommand{\fv}{\boldsymbol f}
\newcommand{\gv}{\boldsymbol g}
\newcommand{\gvt}[1]{\gv^{[#1]}}
\newcommand{\hv}{\boldsymbol h}
\newcommand{\hvt}[1]{\hv^{[#1]}}
\newcommand{\iv}{\boldsymbol i}
\newcommand{\ivt}[1]{\iv^{[#1]}}
\newcommand{\ovt}[1]{\boldsymbol o^{[#1]}}
\newcommand{\rv}{\boldsymbol r}
\newcommand{\uv}{\boldsymbol u}
\newcommand{\vv}{\boldsymbol v}
\newcommand{\xv}{\boldsymbol x}
\newcommand{\xvt}[1]{\boldsymbol x^{[#1]}}
\newcommand{\yv}{\boldsymbol y}

\newcommand{\yvt}[1]{\boldsymbol y^{[#1]}}
\newcommand{\zv}{\boldsymbol z}
\newcommand{\zvt}[1]{\boldsymbol z^{[#1]}}
\newcommand{\qv}{\boldsymbol q}
\newcommand{\qvt}[1]{\boldsymbol q^{[#1]}}

\newcommand{\thetav}{\boldsymbol \theta}
\newcommand{\muv}{\boldsymbol \mu}
\newcommand{\phiv}{\boldsymbol \phi}

\newcommand{\Wm}{\boldsymbol W}
\newcommand{\Um}{\boldsymbol U}
\newcommand{\Vm}{\boldsymbol V}

\newcommand{\Am}{\boldsymbol A}
\newcommand{\Bm}{\boldsymbol B}
\newcommand{\Cm}{\boldsymbol C}
\newcommand{\Dm}{\boldsymbol D}

\DeclareMathOperator{\softmax}{softmax}

\newcommand{\Mm}{\boldsymbol M}
\newcommand{\I}{\boldsymbol I}
\newcommand{\deltat}{\Delta t}

\newcommand{\sigmoid}{\mathfrak{s}}
\newcommand{\bvf}{\bv_f}
\newcommand{\bvi}{\bv_i}
\newcommand{\bvg}{\bv_g}
\newcommand{\bvo}{\bv_o}
\newcommand{\Wmf}{\Wm_f}
\newcommand{\Wmi}{\Wm_i}
\newcommand{\Wmg}{\Wm_g}
\newcommand{\Wmo}{\Wm_o}
\newcommand{\Umf}{\Um_f}
\newcommand{\Umi}{\Um_i}
\newcommand{\Umg}{\Um_g}
\newcommand{\Umo}{\Um_o}

\newcommand{\Xm}{\boldsymbol X}
\newcommand{\etol}{e^{\rm tol}}
\newcommand{\qopt}{q^{\rm opt}}
\DeclareMathOperator*{\argmin}{arg\,min}
\DeclareMathOperator*{\argmax}{arg\,max}

\newcommand{\cauchy}{\boldsymbol \sigma}
\newcommand{\cauchyeff}{\cauchy^\prime}
\newcommand{\pmicro}{p_m}
\newcommand{\pmacro}{p_M}
\newcommand{\fracmacro}{\psi}
\newcommand{\czero}{c_0}
\newcommand{\fluxmacro}{\qv_M}
\newcommand{\fluxmicro}{\qv_m}
\newcommand{\permeab}{\boldsymbol k}
\newcommand{\permeabmacro}{\permeab_M}
\newcommand{\permeabmicro}{\permeab_m}
\newcommand{\uvloc}{\uv_\mu}
\newcommand{\uvjump}{[\![\uv ]\!]}

\newcommand{\encoder}{\mathcal E}
\newcommand{\encoderff}{{\mathbbm e}}
\newcommand{\encoderrnn}{{\mathbbm f}}
\newcommand{\encoderhidden}[1]{\hvt{#1}}

\newcommand{\encoderrnnfwd}{\kars{\encoderrnn}{\rm fwd}}
\newcommand{\encoderrnnrev}{\kars{\encoderrnn}{\rm rev}}
\newcommand{\encoderhiddenfwd}[1]{\bkarsp{h}{\rm fwd}{[#1]}}
\newcommand{\encoderhiddenrev}[1]{\bkarsp{h}{\rm rev}{[#1]}}

\newcommand{\decoder}{\mathcal D}
\newcommand{\decoderrnn}{{\mathbbm g}}
\newcommand{\decoderff}{{\mathbbm d}}

\newcommand{\alignmentfun}{\mathfrak a}
\newcommand{\alignment}{a_{kl}}
\newcommand{\aligments}[1]{\kars{a}{#1}}
\newcommand{\decoderhidden}[1]{\boldsymbol s^{[#1]}}

\newcommand{\alignmentweights}{\alpha_{kl}}

\newcommand{\inpseq}{\Bkr x}
\newcommand{\outseq}{\Bkr y}

\newcommand{\context}{\bkr{c}}
\newcommand{\contexts}{\bkr{C}}

\newcommand{\query}{\bkr{q}}
\newcommand{\queries}{\bkr{Q}}
\newcommand{\key}{\bkr k}
\newcommand{\keys}{\bkr K}
\newcommand{\val}{\bkr v}
\newcommand{\values}{\bkr V}
\newcommand{\heads}{\bkr{\mathcal H}}
\newcommand{\headslayer}[1]{\bkarp{\mathcal H}{(#1)}}

\newcommand{\wq}{\bkarsp{W}{j}{Q}}
\newcommand{\wk}{\bkarsp{W}{j}{K}}
\newcommand{\wv}{\bkarsp{W}{j}{V}}
\newcommand{\wo}{\bkarp{W}{O}}
\newcommand{\attention}{\mathscr{A}}
\newcommand{\multiheadattention}{\attention_{\rm mh}}
\newcommand{\queriesmh}{\bkr{\mathcal Q}}
\newcommand{\keysmh}{\bkr{\mathcal K}}
\newcommand{\valuesmh}{\bkr{\mathcal V}}
\newcommand{\selfattention}{\attention_{\rm self}}

\newcommand{\selfattentionl}{\mathscr A_{\rm self}^{(\ell)}}

\newcommand{\sourceseqlayer}[1]{\bkarp{\mathcal X}{(#1)}}
\newcommand{\sourceseq}[1]{\bkr{\mathcal X}}
\newcommand{\ffnlayer}[1]{\bkr{\mathcal Y}}
\newcommand{\outputffnlayer}[1]{\bkarp{\mathcal Y}{{(#1})}}
\newcommand{\weightffnstransformera}{\bkars{W}{1}}
\newcommand{\weightffnstransformerb}{\bkars{W}{2}}
\newcommand{\weightffnstransformerat}{\bkarsp{W}{1}{T}}
\newcommand{\weightffnstransformerbt}{\bkarsp{W}{2}{T}}
\newcommand{\biasffntransformera}{\bkars{b}{1}}
\newcommand{\biasffntransformerb}{\bkars{b}{2}}
\newcommand{\outputmh}[1]{\bkr{\mathcal Z}}
\newcommand{\outputmhlayer}[1]{\bkarp{\mathcal Z}{{(#1})}}
\newcommand{\outputmhfirstlayer}[1]{\bkarsp{\mathcal Z}{1}{{(#1})}}
\newcommand{\outputmhsecondlayer}[1]{\bkarsp{\mathcal Z}{2}{{(#1})}}

\newcommand{\decoderinputfirstlayer}[1]{\bkarsp{\mathcal Y}{1}{{(#1})}}
\newcommand{\decoderinputsecondlayer}[1]{\bkarsp{\mathcal Y}{2}{{(#1})}}
\newcommand{\decoderinputthirdlayer}[1]{\bkarsp{\mathcal Y}{3}{{(#1})}}
\newcommand{\layernorm}{\mathscr N}
\newcommand{\norminp}{\bkr{\mathcal T}}
\newcommand{\ones}{\Bkr{\mathbb{I} }}

\newcommand{\transformerffn}{\mathscr F}

\newcommand{\encoderinput}[2]{\bkars{x}{#2}}
\newcommand{\transformerffnlayer}[1]{\transformerffn^{(#1)}}
\newcommand{\ffnoutput}[2]{\bkars{y}{#2}}
\newcommand{\ffninput}[2]{\bkars{z}{#2}}
\newcommand{\encoderout}[1]{\bkr{\mathcal E}}
\newcommand{\encodertransformer}[1]{\bkarp{\mathscr E}{(#1)}}
\newcommand{\encodertransformerlayer}[1]{\bkarp{\mathcal E}{(#1)}}

\newcommand{\decoderout}[1]{\bkr{\mathcal D}}

\newcommand{\decodertransformerlayer}[1]{\bkarp{\mathcal D}{(#1)}}

\newcommand{\targetseqlayer}[1]{\bkarp{\mathcal Y}{(#1)}}
\newcommand{\targetseq}[1]{\bkr{\mathcal Y}}
\newcommand{\decoderheadsfirstlayer}[1]{\bkarsp{\mathcal H}{1}{(#1)}}

\newcommand{\posenc}{\bkr{p}}
\newcommand{\reals}{\mathbb R}
\newcommand{\realsN}{\reals^{N_s}}
\newcommand{\realsn}{\reals^{n_s}}
\newcommand{\realsr}{\reals^{n_r}}
\newcommand{\spaceD}{\mathcal D}
\newcommand{\subspace}{\mathcal S}
\DeclareMathOperator{\spanop}{span}
\newcommand{\xvtilde}{\nfwidetilde \xv}

\newcommand{\xvref}{\xv_{ref}}
\newcommand{\xvred}{\widehat \xv}
\newcommand{\rvtilde}{\nfwidetilde \rv}
\newcommand{\rvred}{\widehat \rv}
\newcommand{\vvtilde}{\nfwidetilde \vv}
\newcommand{\vvhat}{\widehat \vv}
\newcommand{\Jm}{\boldsymbol J}
\newcommand{\invJm}{\Jm^\dagger}
\newcommand{\Phim}{\boldsymbol \Phi}
\newcommand{\autoenc}{\mathbf{a e}}
\newcommand{\enc}{\mathbf{en}}
\newcommand{\dec}{\mathbf{de}}
\newcommand{\xvnormal}{\xv_{normal}}
\newcommand{\xvscale}{\xv_{scale}}
\newcommand{\Zm}{\boldsymbol Z}
\newcommand{\Pmcal}{\boldsymbol{\mathcal P}}
\newcommand{\phivhat}{\widehat \phiv}
\newcommand{\phivtilde}{\nfwidetilde \phiv}
\newcommand{\Uv}{\boldsymbol U}
\newcommand{\Vv}{\boldsymbol V}
\newcommand{\Sm}{\boldsymbol S}

\usepackage[misc]{ifsym}

\title{
	Deep Learning Applied to Computational Mechanics:
	\\
	A Comprehensive Review, State of the Art, and the Classics
	\normalsize
}
\author{
	Loc Vu-Quoc,\thanks{
		$^{, \,\textrm{\Letter} \,}$Aerospace Engineering, University of Illinois at Urbana-Champaign, IL 61801, USA $\bullet$
		$^{\textrm{\Letter} \,}${vql@illinois.edu}
	}$^{\bm , \, \textrm{\Letter}}$\ 
	Alexander Humer\thanks{
		Institute of Technical Mechanics, Johannes Kepler University, A-4040 Linz, Austria $\bullet$
		{alexander.humer@jku.at}
		\newline
		\vskip -0.26cm
		{Received: 01 December 2022; Accepted and first online: 01 March 2023; Issue published: 06 June 2023.}
	}\
}

\usepackage{xcolor}
\usepackage{mdframed}
\definecolor{mygray}{gray}{0.92}

\begin{document}
\maketitle

\begin{center}
	{\it A classic never dies.}
\end{center}
\begin{quote}
	\centering
	{\it ``We must welcome the future, for it soon will be the past,
	\\
	and we must respect the past, for it was once all that was humanly possible.''}
	\\
	% \vspace{-3mm}
	\vphantom{a}\hfill George Santayana
\end{quote}
%\vspace{-3mm}
%$\hfill\text{George Santayana}$

\vskip -0.18cm
\begin{mdframed}[
	% style=mystyle,
	%	backgroundcolor=lightgray,
	backgroundcolor=mygray,
	topline=false,
	bottomline=false,
	leftline=false,
	rightline=false
	]
	
	{
		% \fontfamily{crimson}
		\fontsize{10pt}{12pt}\selectfont
		
		\noindent	
		%	{\bf ABSTRACT}
		{\bf ABSTRACT}
		\newline

% short abstract - arXiv style
\label{para:abstract-short}
\noindent
Three recent breakthroughs due to AI in arts and science serve as motivation: An award winning digital image, protein folding, fast matrix multiplication.  Many recent developments in artificial neural networks, particularly deep learning (DL), applied and relevant to computational mechanics (solid, fluids, finite-element technology) are reviewed in detail.  Both hybrid and pure machine learning (ML) methods are discussed.   Hybrid methods combine traditional PDE discretizations with ML methods either (1) to help model complex nonlinear constitutive relations, (2) to nonlinearly reduce the model order for efficient simulation (turbulence), or (3) to accelerate the simulation by predicting certain components in the traditional integration methods. Here, methods (1) and (2) relied on Long-Short-Term Memory (LSTM) architecture, with method (3) relying on convolutional neural networks.   Pure ML methods to solve (nonlinear) PDEs are represented by Physics-Informed Neural network (PINN) methods, which could be combined with attention mechanism to address discontinuous solutions.  Both LSTM and attention architectures, together with modern and generalized classic optimizers to include stochasticity for DL networks, are extensively reviewed.   Kernel machines, including Gaussian processes, are provided to sufficient depth for more advanced works such as shallow networks with infinite width.  Not only addressing experts, readers are assumed familiar with computational mechanics, but not with DL, whose concepts and applications are built up from the basics, aiming at bringing first-time learners quickly to the forefront of research.  History and limitations of AI are recounted and discussed, with particular attention at pointing out misstatements or misconceptions of the classics, even in well-known references.  Positioning and pointing control of a large-deformable beam is given as an example.

			\vspace{0.25cm}
		\noindent
		{\bf KEYWORDS}
		\newline
%		{\color{red} [NOTE: 2022.06.04 - ADD MORE RELEVANT KEYWORDS.]}
%		{\color{red} [NOTE: 2022.12.02 - Updated.  Any more keywords?]}
		\emph{Deep learning}, breakthroughs, network architectures, 
%		weights, activation functions, 
		backpropagation,
		stochastic optimization methods from classic to modern, recurrent neural networks, long short-term memory, gated recurrent unit, attention, transformer,
		kernel machines, Gaussian processes, libraries, 
		Physics-Informed Neural Networks, 
		state-of-the-art, history, limitations, challenges;
		\emph{Applications to computational mechanics};
		Finite-element matrix integration,
		improved Gauss quadrature;
		Multiscale geomechanics, fluid-filled porous media; 
%		discrete element method, finite element method;
		Fluid mechanics, turbulence, proper orthogonal decomposition; 
%		linear projection;
		\emph{Nonlinear-manifold model-order reduction},
		autoencoder,
		hyper-reduction using gappy data; 
		control of large deformable beam.
			
	}
	
\end{mdframed}

\vskip -0.18cm
\begin{mdframed}[
   % style=mystyle,
   %	backgroundcolor=lightgray,
   backgroundcolor=mygray,
   topline=false,
   bottomline=false,
   leftline=false,
   rightline=false
   ]
	
   {
	% \fontfamily{crimson}
	\fontsize{10pt}{12pt}\selectfont
		
	\noindent	
%	{\bf ABSTRACT}
	{\bf SUMMARY}
	\newline

%\begin{abstract}
\label{para:abstract}

\noindent
Three representative applications of deep learning in computational mechanics---involving numerical integration for finite element, complex constitutive model in solid mechanics, and proper orthogonal decomposition in fluid mechanics---are reviewed in detail, and used as motivation for a further in-depth review of some key technologies of deep learning, building up from the basics to the state of the art, focusing on, to the extent possible, the most recent papers that had an important impact in the field.  
%	
%deep learning and its modern practices and applications to computational mechanics---which includes solid mechanics, fluid mechanics, heat transfer, structural health monitoring, electromagnetics, etc.---are reviewed. 
%
%Attention is given to not just statics, but also dynamics (time-dependent) in forward problems, and inverse problems.

Both static and dynamic time-dependent problems are discussed.  Discrete time-dependent problems, as a sequence of data, can be modeled with recurrent neural networks, using the 1997 classic architecture such as Long Short-Term Memory (LSTM), but also the recent 2017-18 architectures such as transformer, based on the concept of attention, 
%[{\color{red} and pervasive attention, NOTE: 2022.09.11 - Please check whether we still talk about ``pervasive attention''.  ENDNOTE}] 
% AH: 2022.09.28 - no, we don't
all of which are discussed in detail.  Continuous recurrent neural networks originally developed in neuroscience to model the brain and the connection to their discrete counterparts in deep learning are also discussed in detail.

For training networks---i.e., finding optimal parameters that yield low training error and lowest validation error---both classic deterministic optimization methods (using full batch) and stochastic optimization methods (using minibatches) are reviewed in detail, and at times even derived.  Deterministic gradient descent with classical line search methods, such as Armijo's rule, were generalized to add stochasticity. Detailed pseudocodes for these methods are provided.  The classic stochastic gradient descent (SGD), with add-on tricks such as momentum, step-length decay, cyclic annealing, weight decay are presented, often with detailed derivations.

Step-length decay is shown to be equivalent to simulated annealing using stochastic differential equation equivalent to the discrete parameter update.  A consequence is to increase the minibatch size, instead of decaying the step length.
In particular, we obtain a new result for minibatch-size increase.

Highly popular adaptive step-length (learning-rate) methods are discussed in a unified manner, which covers AdaGrad, RMSProp, the ``immensely successful'' Adam and its variants, through to the recent AdamW.   

Overlooked in (or unknown to) other review papers and even well-known books on deep learning, exponential smoothing of time series, the key technique of adaptive methods, and originating from the field of forecasting dated since the 1950s, is carefully explained.

Particular attention is given to a recent criticism of adaptive methods, revealing their marginal value for generalization, compared to good old SGD with effective initial step-length tuning and decay.  The results were confirmed in three recent independent papers.

Kernel machines, including Gaussian processes, a most important class of non-parametric modeling with accurate uncertainty estimates, are introduced to sufficient details to prepare for more advanced works on networks with infinite width that constitute the 2021 breakthrough in computer science. 

Applications of deep learning in computational mechanics often aim at reducing computational cost, which naturally connects to the field of (nonlinear) model-order reduction (MOR).
We review how LSTM networks were trained to predict the rate-dependent constitutive response in multi-scale problem of porous media and how they were used as time-integrators for reduced order models (ROMs) inferred from highly-resolved direct numerical simulations of turbulent flows.
Autoencoders based on shallow networks provide effective means in nonlinear manifold-based MOR and hyper-reduction method built on top.
%{\color{red} NOTE: 2022.06.03 - Need to mention nonlinear model-order reduction using shallow neural networks, with encoder and autoencoder.}

A rare feature of the present paper is in a detailed review of some important classics to connect to the relevant concepts in modern literature, sometimes revealing misunderstanding in recent works, which was likely due to a lack of verification of the assertions made with the corresponding classics.  For example, the first artificial neural network, conceived by
Rosenblatt
%\citename{Rosenblatt.1957}
%\cite{Rosenblatt.1957}-(\citeyear{Rosenblatt.1962}),
%\cite{Rosenblatt.1957}-\cite{Rosenblatt.1962},
(1957) \cite{Rosenblatt.1957}, (1962) \cite{Rosenblatt.1962} 
had 1000 neurons, but was reported as having a single neuron.  Going beyond probabilistic analysis, Rosenblatt even built the Mark I computer to implement his 1000-neuron network.  Another example is the ``heavy ball'' method, for which everyone referred to \cite{Polyak.1964}, but who more precisely called the ``small heavy sphere'' method.  Others were quick to dismiss classical deterministic line-search methods that have been generalized to add stochasticity for network training.   Unintended misrepresentation of the classics would mislead first-time learners, and unfortunately even seasoned researchers who used second-hand information from others, without checking the original classics themselves. 

The experiments in the 1950s that discovered the rectified linear behavior in neuronal axon, modeled as a circuit with a diode, together with the use of the rectified linear activation function in neural networks in neuroscience years before being adopted for use in deep-learning networks, are reviewed.

The use of Volterra series to model the nonlinear behavior of neuron in term of input and output firing rates, leading to continuous recurrent neural networks is examined in detail.  The linear term of the Volterra series is a convolution integral that provides a theoretical foundation for the use of linear combination of inputs to a neuron, with weights and biases. 

A goal of this in-depth review is not only to provide the state of the art for computational-mechanics readers with some familiarity of deep-learning networks, but also with first-time learners in mind, by developing relevant fundamental concepts from the basics.  Moreover, for the convenience of the readers, detailed references are provided, e.g., page numbers in thick books, links to online references and open reviews where available.

%{\color{red} HERE 2020.02.06}  

%A goal of this in-depth review is for computational-mechanics practitioners (including students) who are not familiar with artificial neural networks and deep learning to acquire quickly the necessary knowledge to start applying and/or to develop these methods for computational mechanics.  
%as a result, a combination of review and tutorial is presented, with the relevant information for modern applications of deep learning appearing first, while the historical perspective of the development of artificial intelligence in general, and deep learning in particular, is relegated to a later part of the paper.

%\end{abstract}

   }

\end{mdframed}

\vspace{0.1cm}
\hrule

% \newpage
%
% 2022.09.12
% The problem with using the command \tableofcontents is that the
% header is not printed in the first page of the TOC (or page 2 
% of a CMES article).  I replace this command by the code below.
% \tableofcontents

% 2022.09.12
% new way to do the table of contents so to have the header on 
% the first page of the TOC (or 2nd page of a CMES article)
\vspace{1.5\baselineskip}

\noindent
%{\bf Table of Contents}
{\bf TABLE OF CONTENTS}

\vspace{0.5\baselineskip}

% print the TOC using the internal command \@starttoc, which has
% the special character @, and therefore the command \newcommand must
% be enclosed between two commands \makeatletter and \makeatother
\makeatletter
\newcommand*{\toccontents}{\@starttoc{toc}}
\makeatother

% reduce the \baselineskip by half for the TOC
\setlength{\baselineskip}{0.5\baselineskip}

% print the TOC
\toccontents

% after printing the TOC, double \baselineskip back to its previous value
\setlength{\baselineskip}{2.0\baselineskip}

\vspace{0.5cm}
\hrule

% general and 3 examples of application
%\newpage

\begin{figure}[h]
	\centering
	\includegraphics[width=0.95\linewidth]{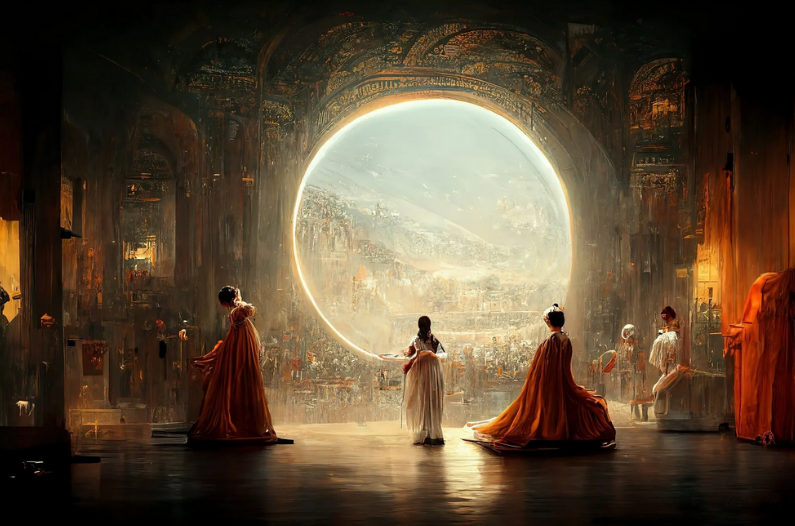}
	\caption{
		\emph{AI-generated image won contest} in the category of Digital Arts, Emerging Artists, on 2022.08.29 (Section~\ref{sc:open}).
		``Th\'e\^atre D'op\'era Spatial'' (Space Opera Theater) by ``Jason M. Allen via Midjourney'', which is ``an artificial intelligence program that turns lines of text into hyper-realistic graphics'' \cite{roose2022an}. Colorado State Fair, \href{https://web.archive.org/web/20220904163544/https://coloradostatefair.com/wp-content/uploads/2022/08/2022-Fine-Arts-First-Second-Third.pdf}{2022 Fine Arts First, Second \& Third}. 
		{\footnotesize (Permission of Jason M. Allen, CEO, \href{http://www.incarnategames.com/blog/}{Incarnate Games})}
%		{\footnotesize (Figure reproduced with permission of the author)}
		%		{\color{red} ASK PERMISSION 2019.11.25}
	}
	\label{fig:AI-art}
\end{figure}

\begin{figure}[h]
	\centering
	\includegraphics[width=0.51\linewidth]{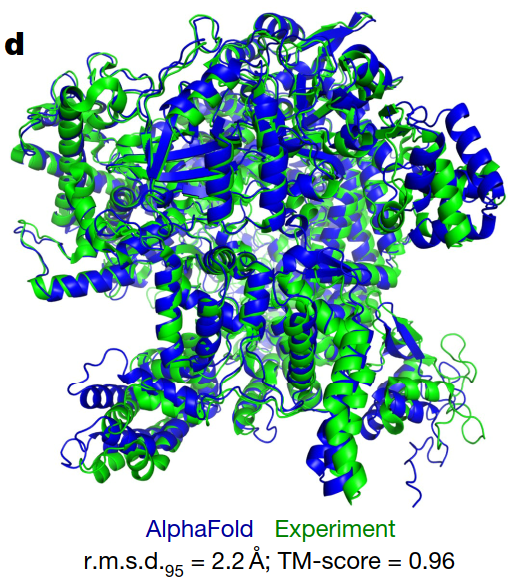}
	\includegraphics[width=0.47\linewidth]{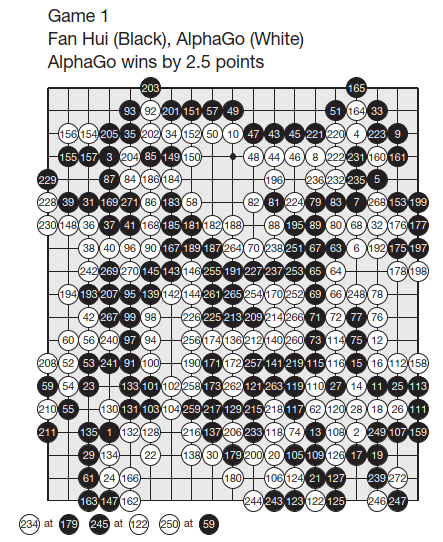}
	\caption{
		\emph{Breakthroughs in AI} (Section~\ref{sc:resurgence}).
		\emph{Left:} 
		The journal \emph{Science} 2021 Breakthough of the Year.
		Protein folded 3-D shape produced by the AI software AlphaFold compared to experiment with high accuracy \cite{jumper2021highly}. The \href{https://alphafold.ebi.ac.uk/}{AlphaFold Protein Structure Database} contains more than 200 million protein structure predictions, a holy grail sought after in the last 50 years. 
		\emph{Right:} The AI solfware AlphaGo, a runner-up in the journal \emph{Science} 2016 Breakthough of the Year, beat the European Go champion Fan Hui five games to zero in 2015
		\cite{Silver.2016}, and then went on to defeat the world Go grandmaster Lee Sedol in 2016 \cite{moyer2016how}.  
		{\footnotesize (Permission by  \href{https://www.nature.com/nature-portfolio/reprints-and-permissions}{Nature}.)}
		%		{\color{red} ASK PERMISSION 2019.11.25}
	}
	\label{fig:protein-folding-Go-game}
\end{figure}

\section{Opening remarks and organization}
\label{sc:open}
% 2022.06.04 - The comment and command below (to reset the footnote number) were from my CMES Karl Pister special issue paper.
% 2021.08.07 - using the tsp style, the footnote counter started at 3, instead of 1, and thus required setting the footnote counter to 0 at the beginning of the introduction, then everything worked.  i found this out after i got the first proof. https://tex.stackexchange.com/questions/359702/how-to-manually-reset-the-footnote-numbering-in-context
\setcounter{footnote}{0}

\emph{Breakthroughs due to AI in arts and science.}
On 2022.08.29, Figure~\ref{fig:AI-art}, an image generated by the AI software \href{https://www.midjourney.com/home/}{Midjourney}, became one of the first of its kind to win first place in an art contest.\footnote{
	See also the Midjourney \href{https://www.midjourney.com/showcase/}{Showcase}, \href{https://web.archive.org/web/20220907013425/https://www.midjourney.com/showcase/}{Internet archived on 2022.09.07}, the video \href{https://www.youtube.com/watch?v=_WfDlJY1nog}{Guide to MidJourney AI Art - How to get started FREE!} and several other Midjourney tutorial videos on Youtube.
}  The image author signed his entry to the contest as ``Jason M. Allen via Midjourney,'' indicating that the submitted digital art was not created by him in the traditional way, but under his text commands to an AI software.  Artists not using AI software---such as Midjourney, \href{https://openai.com/dall-e-2/}{DALL.E 2}, \href{https://stability.ai/blog/stable-diffusion-public-release}{Stable Diffusion}---were not happy \cite{roose2022an}.

In 2021, an AI software achieved a feat that human researchers were not able to do in the last 50 years in predicting protein structures quickly and in a large scale.  This feat was named the scientific breakthough of the year; Figure~\ref{fig:protein-folding-Go-game}, left.  In 2016, another AI software beat the world grandmaster in the Go game, which is described as the most complex game that human ever created; Figure~\ref{fig:protein-folding-Go-game}, right. 

On 2022.10.05, DeepMind published a paper on breaking a 50-year record of fast matrix multiplication by reducing the number of multiplications in multiplying two $4 \times 4$ matrices from 49 to 47 (with the traditional method requiring 64 multiplications), owing to an algorithm discovered with the help of their AI software AlphaTensor \cite{edwards2022deepmind}.\footnote{
	Their goal was of course to discover fast multiplication algorithms for matrices of arbitrarily large size.  See also ``Discovering novel algorithms with AlphaTensor,'' DeepMind, 2022.10.05, \href{https://web.archive.org/web/20221014160200/https://www.deepmind.com/blog/discovering-novel-algorithms-with-alphatensor}{Internet archive}.
}  Just barely a week later, two mathematicians announced an algorithm that required only 46 multiplications.

Since the preprint of this paper was posted on the arXiv in Dec 2022 \cite{vuquoc2022deep}, there have been considerable excitements and concerns about ChatGPT---a large language-model chatbot that can interact with humans in a conversational way---which would be incorporated into Microsoft Bing to make web ``search interesting again, after years of stagnation and stasis'' \cite{roose2023bing}, whose author wrote ``I'm going to do something I thought I'd never do: I'm switching my desktop computer's default search engine to Bing. And Google, my default source of information for my entire adult life, is going to have to fight to get me back.''   Google would release its own answer to  ChatGPT called ``Bard'' \cite{knight2023meet}.  The race is on.

\emph{Audience}.
This review paper is written by mechanics practitioners to mechanics practitioners, who may or may not be familiar with neural networks and deep learning.  
We thus assume that the readers are familiar with continuum mechanics and numerical methods such as the finite element method.  
Thus, unlike typical computer-science papers on deep learning, notation and convention of tensor analysis familiar to practitioners of mechanics are used here whenever possible.\footnote{
   Tensors are not matrices;  other concepts are summation convention on repeated indices, chain rule, and matrix index convention for natural conversion from component form to matrix (and then tensor) form.
   See Section~\ref{sc:matrix} on Matrix notation.
} 

For readers not familiar with deep learning, unlike many other review papers, this review paper is not just a summary of papers in the literature for people who already have some familiarity with this topic,\footnote{
	\label{fn:review-papers}
	See the review papers on deep learning, e.g., \cite{Schmidhuber.2015:rd0001} \cite{LeCun.2015:rd0001} \cite{Khan.2018:rd0001} \cite{Sanchez.2018:rd0001}  \vphantom{\cite{Ching.2018:rd0001}}\cite{Ching.2018:rd0001} \vphantom{\cite{Quinn.2018:rd0001}}\cite{Quinn.2018:rd0001} \cite{higham2019deep}, many of which did not provide extensive discussion on applications, particularly on computational mechanics, such as in the present review paper.
} particularly papers on deep-learning neural networks, but contains also a tutorial on this topic aiming at bringing first-time learners (including students) quickly up-to-date with modern issues and applications of deep learning, especially to computational mechanics.\footnote{
	\label{fn:confusion}
	An example of a confusing point for \emph{first-time learners} with knowledge of electrical circuits, hydraulics, or (biological) computational neuroscience \cite{Dayan.2001} would be the interpretation of the arrows in an artificial neural network such as those in Figure~\ref{fig:Oishi-network} and Figure~\ref{fig:Oishi-neuron}: Would these arrows represent real physical flows (electron flow, fluid flow, etc.)?  No, they represent function mapping (or information passing); see Section~\ref{sc:graphical-representation} on Graphical representation.
	Even a tutorial such as \vphantom{\cite{Sze.2017:rd0001}}\cite{Sze.2017:rd0001} would follow the same format as many other papers, and while alluding to the human brain in their Figure 2 (which is the equivalent of Figure~\ref{fig:Oishi-neuron} below), did not explain the meaning of the arrows.
}
As a result, this review paper is a convenient ``one-stop shopping'' that provides the necessary fundamental information, with clarification of potentially confusing points, for first-time learners to quickly acquire a general understanding of the field that would facilitate deeper study and application to computational mechanics.

\emph{Deep-learning software libraries}.
Just as there is a large number of available software in different subfields of computational mechanics, there are many excellent deep-learning libraries ready for use in applications; see Section~\ref{sc:libraries}, in which some examples of the use of these libraries in engineering applications are provided with the associated computer code.  Similar to learning  finite-element formulations versus learning how to run finite-element codes, our focus here is to discuss various algorithmic aspects of deep-learning and their applications in computational mechanics, rather than how to use deep-learning libraries in applications.  
We agree with the view that ``a solid understanding of the core principles of neural networks and deep learning'' would provide ``insights that will still be relevant years from now'' \cite{Nielsen.2015}, and that would not be obtained from just learning to run some hot libraries.

Readers already familiar with neural networks may find the presentation refreshing,\footnote{
	\label{fn:for-experts}
Particularly the top-down approach for both feedforward network (Section~\ref{sc:feedforward}) and back propagation (Section~\ref{sc:backprop}).
}
and even find new information on neural networks, depending how they used deep learning, or when they stopped working in this area due to the waning wave of connectionism and the new wave of deep learning.\footnote{
	It took five years from the publication of Rumelhart \emph{et al.} 1986 \cite{Rumelhart.1986} to the paper by Ghaboussi \emph{et al.} 1991 \cite{Ghaboussi.1991:rd0001}, in which  backpropagation (Section~\ref{sc:backprop}) was applied.  
	It took more than twenty years from the publication of Long Short-Term Memory (LSTM) units in \cite{Hochreiter.1997:rd0001} to the two recent papers \cite{Wang.2018:rd3109} and  \cite{Mohan.2018}, which are reviewed in detail here, and where recurrent neural networks (RNNs, Section~\ref{sc:recurrent})  with LSTM units (Section~\ref{sc:LSTM}) were applied, even though there were some early works on application of RNNs (without LSTM units) in civil / mechanical engineering such as \cite{Zaman.1998:rd0001} \cite{Su.1998:rd0001} \cite{Li.1999:rd0001} \cite{Waszczyszyn.2000:rd0001}.
	But already, ``fully attentional Transformer'' was proposed to render ``intricately constructed LSTM'' unnecessary \vphantom{\cite{Vaswani.2017:rd0001}}\cite{Vaswani.2017:rd0001}. 
	Most modern networks use the default rectified linear function (ReLU)---which was introduced in computational neuroscience since at least before 
	\vphantom{\cite{Hahnloser.2000:rd0002}}\cite{Hahnloser.2000:rd0002} and \cite{Dayan.2001}, and then adopted in computer science beginning with \vphantom{\cite{Jarrett.2009:rd0001}}\cite{Jarrett.2009:rd0001} and \cite{Nair.2010:rd0001}---instead of the traditional sigmoid function dated since the mid 1970s with \cite{Little.1974:rd0001}, but yet many newer activation functions continue to appear regularly, aiming at improving accuracy and efficiency over previous activation functions, e.g., 
	\cite{Ramachandran.2017:rd0001}, \cite{Wuraola.2018:rd0001}.
	In computational mechanics, by the beginning of 2019, there has not yet widespread use of ReLU activation function, even though ReLU was mentioned in \cite{Oishi.2017:rd9648}, where the sigmoid function was actually employed to obtain the results (Section~\ref{sc:integration}).
	See also Section~\ref{sc:history} on Historical perspective.
}   
If not, readers can skip these sections to go directly to the sections on applications of deep learning to computational mechanics.  

\emph{Applications of deep learning in computational mechanics}.
We select some recent papers on application of deep learning to computational mechanics to review in details in a way that readers would also understand the computational mechanics contents well enough without having to read through the original papers: 

%\newpage
\begin{itemize}

\item 
Fully-connected feedforward neural networks were employed to make element-matrix integration more efficient, while retaining the accuracy of the traditional Gauss-Legendre quadrature \cite{Oishi.2017:rd9648};\footnote{
   It would be interesting to investigate on how the adjusted integration weights using the method in \cite{Oishi.2017:rd9648} would affect the stability of an element stiffness matrix with reduced integration (even in the absence of locking) and the superconvergence of the strains / stresses at the Barlow sampling points.  
   See, e.g., \cite{Zienkiewicz.2013:rd0001}, p.~499.
   The optimal locations of these strain / stress sampling points do not depend on the integration weights, but only on the degree of the interpolation polynomials; see \cite{Barlow.1976:rd0001} \cite{Barlow.1977:rd0001}.
   ``The Gauss points corresponding to reduced integration are the Barlow points (Barlow, 1976) at which the strains are most accurately predicted if the elements are well-shaped'' \cite{Abaqus-6.14:rd0001}. 
}

\item Recurrent neural network (RNN) with Long Short-Term Memory (LSTM) units\footnote{
It is only a coincidence that (1) Hochreiter (1997), the first author in \cite{Hochreiter.1997:rd0001}, which was the original paper on the widely used and highly successful Long Short-Term Memory (LSTM) unit, is on the faculty at Johannes Kepler University (home institution of this paper's author A.H.), and that (2) Ghaboussi (1991), the first author in \cite{Ghaboussi.1991:rd0001}, who was among the first researchers to apply fully-connected feedforward neural network to constitutive behavior in solid mechanics, was on the faculty at the University of Illinois at Urbana-Champaign (home institution of author L.V.Q.). 
See also \cite{Ghaboussi.1990:rd0001}, and for early applications of neural networks in other areas of mechanics, see e.g., \cite{Chen.1989:rd0001}, \cite{Sayeh.1990:rd0001}, \cite{Chen.1991:rd0001}. 
}
was applied to multiple-scale, multi-physics problems in solid mechanics \cite{Wang.2018:rd3109}; 

\item RNNs with LSTM units were employed to obtain reduced-order model for turbulence in fluids based on the proper orthogonal decomposition (POD), a classic linear project method also known as principal components analysis (PCA) \cite{Mohan.2018}.  More recent nonlinear-manifold model-order reduction methods, incorporating encoder / decoder and hyper-reduction of dimentionality using gappy (incomplete) data, were introduced, e.g., \cite{kim.2020a} \cite{kim.2020b}. 
\end{itemize}

%{\color{red} NOTE: 2022.06.07 - Alex, pl verify the above paragraph.}
% 2022.09.30 - checked

%Other applications are given briefer review, such as inverse problem in materials design, structural health monitoring, etc.

\emph{Organization of contents}.
Our review of each of the above papers is divided into two parts.  The first part is to summarize the main results and to identify the concepts of deep learning used in the paper, expected to be new for first-time learners, for subsequent elaboration.  The second part is to explain in details how these deep-learning concepts were used to produce the results.

The results of deep-learning numerical integration \cite{Oishi.2017:rd9648} are presented in Section~\ref{sc:Oishi-summary}, where the deep-learning concepts employed are identified and listed, whereas the details of the formulation in \cite{Oishi.2017:rd9648} are discussed in Section~\ref{sc:Oishi-2}.  
Similarly, the results and additional deep-learning concepts used in a multi-scale, multi-physics problem of geomechanics \cite{Wang.2018:rd3109} are presented in Section~\ref{sc:Wang-Sun-2018}, whereas the details of this formulation are discussed in Section~\ref{sc:Wang-Sun-2018-2}.
Finally, the results and additional deep-learning concepts used in turbulent fluid simulation with proper orthogonal decomposition \cite{Mohan.2018} are presented in Section~\ref{sc:Mohan-2018}, whereas the details of this formulation, together with the nonlinear-manifold model-order reduction \cite{kim.2020a} \cite{kim.2020b}, are discussed in Section~\ref{sc:Mohan-2018-2}.

%{\color{red} NOTE: 2022.06.07 - Alex, pl verify the above paragraph.}
% AH: 2022.09.30 - checked

All of the deep-learning concepts identified from the above selected papers for in-depth are subsequently explained in detail in Sections~\ref{sc:comparison-three-fields} to \ref{sc:recurrent}, and then more in Section~\ref{sc:history} on ``Historical perspective''.

The parallelism between computational mechanics, neuroscience, and deep learning is summarized in Section~\ref{sc:comparison-three-fields}, which would put computational-mechanics first-time learners at ease, before delving into the details of deep-learning concepts.

%Following the fundamentals of artificial neural networks
Both time-independent (static) and time-dependent (dynamic) problems are discussed.  The architecture of (static, time-independent) feedforward multilayer neural networks in Section~\ref{sc:feedforward} is expounded in detail, with first-time learners in mind, without assuming prior knowledge, and where experts may find a refreshing presentation and even new information.

Backpropagation, explained in Section~\ref{sc:backprop}, is an important method to compute the gradient of the cost function relative to the network parameters for use as a descent direction to decrease the cost function for network training.  

%Details of both deterministic and stochastic optimization algorithms used in network training are extensively presented in Section~\ref{sc:training}, which would be useful for both first-time learners and experts alike since not only the classics, but also the most recent algorithms, to the extent possible, are reviewed.

For training networks---i.e., finding optimal parameters that yield low training error and lowest validation error---both classic deterministic optimization methods (using full batch) and stochastic optimization methods (using minibatches) are reviewed in detail, and at times even derived, in Section~\ref{sc:training}, which would be useful for both first-time learners and experts alike.
 
The examples used in training a network form the training set, which is complemented by the validation set (to determine when to stop the optimization iterations) and the test set (to see whether the resulting network could work on examples never seen before); see  Section~\ref{sc:training-valication-test}.

\hyperref[sc:deterministic-optimization]{Deterministic gradient descent} with classical line search methods, such as \hyperref[sc:armijo]{Armijo's rule} (Section~\ref{sc:deterministic-optimization}), were generalized to add stochasticity. Detailed pseudocodes for these methods are provided.  The classic stochastic gradient descent (\hyperref[sc:stochastic-gradient-descent]{SGD}) by Robbins \& Monro (1951) \cite{Robbins1951b} (Section~\ref{sc:stochastic-gradient-descent}, Section~\ref{sc:generic-SGD}), with add-on tricks such as \hyperref[sc:SGD-momentum]{momentum} Polyak (1964) \cite{Polyak.1964} and \hyperref[sc:SGD-momentum]{fast (accelerated) gradient} by 
%\cite{Nesterov.1983}, \cite{Nesterov.2018}, 
Nesterov 
%(\citeyear{Nesterov.1983}, \citeyear{Nesterov.2018})
(1983 \cite{Nesterov.1983}, 2018 \cite{Nesterov.2018}) 
(Section~\ref{sc:SGD-momentum}),  \hyperref[sc:step-length-decay]{step-length decay} (Section~\ref{sc:step-length-decay}), \hyperref[sc:step-length-decay]{cyclic annealing} (Section~\ref{sc:step-length-decay}), \hyperref[sc:minibatch-size-increase]{minibatch-size increase} (Section~\ref{sc:minibatch-size-increase}), \hyperref[sc:weight-decay]{weight decay} (Section~\ref{sc:weight-decay}) are presented, often with detailed derivations. 

\hyperref[sc:step-length-decay]{Step-length decay} is shown to be equivalent to simulated annealing using stochastic differential equation equivalent to the discrete parameter update.  A consequence is to increase the minibatch size, instead of decaying the step length (Section~\ref{sc:minibatch-size-increase}).
In particular, we obtain a new result for minibatch-size increase.

In Section~\ref{sc:adaptive-learning-rate-algos},
highly popular \hyperref[sc:adaptive-learning-rate-algos]{adaptive step-length} (learning-rate) methods are discussed in a unified manner in Section~\ref{sc:unified-adaptive}, followed by the first paper on \hyperref[sc:adagrad]{AdaGrad} \cite{Duchi.2011} (Section~\ref{sc:adagrad}).

Overlooked in (or unknown to) other review papers and even well-known books on deep learning, exponential smoothing of time series originating from the field of forecasting dated since the 1950s, the key technique of adaptive methods, is carefully explained in Section~\ref{sc:exponential-smoothing}.

The first adaptive methods that employed exponential smoothing were \hyperref[sc:rmsprop]{RMSProp} \cite{Tieleman.2012} (Section~\ref{sc:rmsprop}) and \hyperref[sc:adadelta]{AdaDelta} \cite{Zeiler.2012} (Section~\ref{sc:adadelta}), both introduced at about the same time, followed by the ``immensely successful'' \hyperref[sc:adam1]{Adam} (Section~\ref{sc:adam1}) and its variants (Sections~\ref{sc:amsgrad} and \ref{sc:adamx}).

Particular attention is then given to a recent criticism of adaptive methods in \vphantom{\cite{Wilson.2018}}\cite{Wilson.2018}, revealing their marginal value for generalization, compared to the good old SGD with effective initial step-length tuning and step-length decay (Section~\ref{sc:adam-criticism}).  The results were confirmed in three recent independent papers, among which is the recent  \hyperref[sc:adamw]{AdamW} adaptive method in \cite{Loshchilov.2019} (Section~\ref{sc:adamw}).   

%{\color{red} HERE, 2020.03.24}

Dynamics, sequential data, and sequence modeling are the subjects of Section~\ref{sc:recurrent}.
Discrete time-dependent problems, as a sequence of data, can be modeled with recurrent neural networks discussed in Section~\ref{sc:RNN}, using the 1997 classic architecture such as Long Short-Term Memory (LSTM) in Section~\ref{sc:LSTM}, but also the recent 2017-18 architectures such as transformer introduced 
%
% CMES style rewriting
%by
in 
\cite{Vaswani.2017:rd0001} (Section~\ref{sc:Transformer}), based on the concept of attention~\cite{Bahdanau.2015}.
% {\color{red} 2020.03.18 REFERENCE}. 
% AH: 2022.09.30 - added reference.
%, and pervasive attention by \cite{Elbayad.2018} (Section~\ref{sc:pervasive-attention}), all of which are discussed in detail.  
Continuous recurrent neural networks originally developed in neuroscience to model the brain and the connection to their discrete counterparts in deep learning are also discussed in detail, \cite{Dayan.2001} and Section~\ref{sc:dynamic-volterra-series} on ``Dynamic, time dependence, Volterra series''.

%{\color{red} HERE, 2020.03.24.  the above paragraph is not done; need REFs for ``attention''.  
%	
%2022.06.06 - Did we skip the topic ``pervasive attention''?  If yes, we need to delete the mention of it in the above paragraph.  Section~\ref{sc:pervasive-attention} is still there, incomplete, with no new writing since it was last updated.}
% 2020.09.30 - pervasive attention is skipped

The features of several popular, open-source deep-learning frameworks and libraries---such as TensorFlow, Keras, PyTorch, etc.---are summarized in Section~\ref{sc:libraries}.

As mentioned above, detailed formulations of deep learning applied to computational mechanics in \cite{Oishi.2017:rd9648} \cite{Wang.2018:rd3109} \cite{Mohan.2018} \cite{kim.2020a} \cite{kim.2020b} are reviewed in Sections~\ref{sc:Oishi-2}, \ref{sc:Wang-Sun-2018-2}, \ref{sc:Mohan-2018-2}.

%{\color{red} NOTE: 2022.06.07 - Alex, pl verify the above paragraph.}
% 2022.12.04: OK
% {\color{red} HERE, 2020.03.25}

\emph{History of AI, limitations, danger, and the classics}.
Finally, a broader historical perspective of deep learning, machine learning, and artificial intelligence is discussed in Section~\ref{sc:history}, ending with comments on the geopolitics,  limitations, and (identified-and-proven, not just speculated) danger of artificial intelligence in Section~\ref{sc:closure}.

A rare feature is in a detailed review of some important classics to connect to the relevant concepts in modern literature, sometimes revealing misunderstanding in recent works, likely due to a lack of verification of the assertions made with the corresponding classics.  For example, the first artificial neural network, conceived by
Rosenblatt
%\cite{Rosenblatt.1957}-(\citeyear{Rosenblatt.1962}),
(1957) \cite{Rosenblatt.1957}, (1962) \cite{Rosenblatt.1962}, 
had 1000 neurons, but was reported as having a single neuron (Figure~\ref{fig:network-size-time}).  Going beyond probabilistic analysis, Rosenblatt even built the Mark I computer to implement his 1000-neuron network (Figure~\ref{fig:Rosenblatt-Mark-I}, Sections~\ref{sc:linear-combo-history} and \ref{sc:static-Rosenblatt}).  Another example is the ``heavy ball'' method, for which everyone referred to Polyak (1964) \cite{Polyak.1964}, but who more precisely called the ``small heavy sphere'' method (Remark~\ref{rm:heavy-ball}).  Others were quick to dismiss classical deterministic line-search methods that have been generalized to add stochasticity for network training (Remark~\ref{rm:classic-never-dies}).   Unintended misrepresentation of the classics would mislead first-time learners, and unfortunately even seasoned researchers who used second-hand information from others, without checking the original classics themselves. 

The use of Volterra series to model the nonlinear behavior of neuron in term of input and output firing rates, leading to continuous recurrent neural networks is examined in detail.  The linear term of the Volterra series is a convolution integral that provides a theoretical foundation for the use of linear combination of inputs to a neuron, with weights and biases \cite{Dayan.2001}; see Section~\ref{sc:dynamic-volterra-series}.

The experiments in the 1950s by Furshpan et al. 
%(\citeyear{Furshpan1957}, \citeyear{Furshpan1959b}) 
%(\cite{Furshpan1957}, \cite{Furshpan1959b})
\cite{Furshpan1957} \cite{Furshpan1959b} that revealed the rectified linear behavior in neuronal axon, modeled as a circuit with a diode, together with the use of the rectified linear activation function in neural networks in neuroscience years before being adopted for use in deep learning network, are reviewed in Section~\ref{sc:ReLU-history}.

\emph{Reference hypertext links and Internet archive.}
For the convenience of the readers, whenever we refer to an online article, we provide both the link to original website, and if possible, also the link to its archived version in the Internet Archive.
For example, we included in the bibliography entry of Ref.~\cite{Gershgorn.2017:rd0001} the links to both the
\href{http://news.mit.edu/2017/explained-neural-networks-deep-learning-0414}{Original website} and the
\href{https://web.archive.org/web/20181110195900/http://news.mit.edu/2017/explained-neural-networks-deep-learning-0414}{Internet archive}.\footnote{
	While in the long run an original website may be moved or even deleted, the same website captured on the Internet Archive (also known as Web Archive or Wayback Machine) remains there permanently.
} 

\begin{figure}[h]
	\centering
	\includegraphics[width=1.0\linewidth]{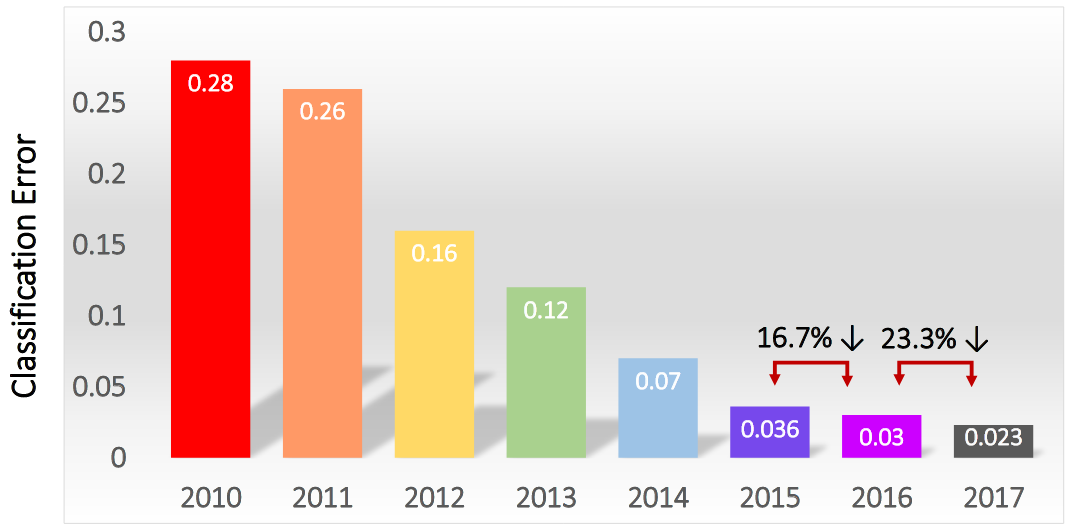}
	\caption{
		\emph{ImageNet competitions} (Section~\ref{sc:resurgence}).  Top (smallest) classification error rate versus competition year.  
		A sharp decrease in error rate in 2012 sparked a resurgence in AI interest and research \cite{LeCun.2015:rd0001}.
		By 2015, the top classification error rate surpassed human 
		classification error rate of 5.1\% 
		% \vphantom{\cite{Russakovsky.2015}}
		with Parametric Rectified Linear Unit \cite{He.2015b}; see Section~\ref{sc:parametric-ReLU} and also
		\cite{Russakovsky.2015}.
		Figure from \cite{Park.2017:rd0001}.
		%\\
		{\footnotesize (Figure reproduced with permission of the authors.)}
		% {\color{red} ASK PERMISSION}
	}
	\label{fig:ImageNet-error}
\end{figure}

\section{Deep Learning, resurgence of Artificial Intelligence}
\label{sc:resurgence}

%{\color{red} [NOTE: 2022.08.31 - filenames: AlphaFold-protein.png - 2022.08.02 - write about the breakthrough in 2021 on protein folding here.  ENDNOTE]}

In Dec 2021, the journal \emph{Science} named, as its ``2021 Breakthrough of the Year,'' the development of the AI software AlphaFold and its amazing feat of predicting a large number of protein structures \cite{beckwith2022science}. 
``For nearly 50 years, scientists have struggled to solve one of nature's most perplexing challenges---predicting the complex 3D shape a string of amino acids will twist and fold into as it becomes a fully functional protein. This year, scientists have shown that artificial intelligence (AI)-driven software can achieve this long-standing goal and predict accurate protein structures by the thousands and at a fraction of the time and cost involved with previous methods'' \cite{beckwith2022science}.

%Proteins are the building blocks of life.

The 3-D shape of a protein, obtained by folding a linear chain of amino acid, determines how this protein would interact with other molecules, and thus establishes its biological functions \cite{beckwith2022science}.
There are some 200 million proteins, the building blocks of life, in all living creatures, and 400,000 in the human body \cite{beckwith2022science}. The \href{https://alphafold.ebi.ac.uk/}{AlphaFold Protein Structure Database} already contained ``over 200 million protein structure predictions.''\footnote{
	See also AlphaFold Protein Structure Database  \href{https://web.archive.org/web/20220902132758/https://alphafold.ebi.ac.uk/}{Internet archived as of 2022.09.02}.
}
For comparison, there were only about 190 thousand protein structures obtained through experiments as of 2022.07.28 \cite{deepmind2022alphafold}.
``Some of AlphaFold's predictions were on par with very good experimental models [Figure~\ref{fig:protein-folding-Go-game}, left], and potentially precise enough to detail atomic features useful for drug design, such as the active site of an enzyme''
\cite{callaway2021deepmind}.  The influence of this software and its developers ``would be epochal.''

%https://alphafold.ebi.ac.uk/ 
%https://web.archive.org/web/20220902132758/https://alphafold.ebi.ac.uk/

On the 2019 new-year day, {\em The Guardian} \cite{Guardian.2019:rd0001} reported the most recent breakthrough in AI, published less than a month before on 2018 Dec 07 in the journal {\em Science} 
%
% CMES style rewriting
%by 
in
\vphantom{\cite{Silver.2018:rd0001}}\cite{Silver.2018:rd0001} on the development of the software AlphaZero, based on deep reinforcement learning (a combination of deep learning and reinforcement learning), that can teach itself through self-play, and then ``convincingly defeated a world champion program in the games of chess, shogi (Japanese chess), as well as Go''; see Figure~\ref{fig:protein-folding-Go-game}, right.

Go is the most complex game that mankind ever created, with more combinations of possible moves than chess, and thus the number of atoms in the observable universe.\footnote{
	The number of atoms in the observable universe is estimated at $10^{80}$. 
	For a board game such as chess and Go, the number of possible sequences of moves is $m = b^d$, with $b$ being the game breadth (or ``branching factor'', which is the ``number of legal moves per position'' or average number of moves at each turn), and $d$ the game depth (or length, also known as number of ``plies'').
	For chess, 
	$b \approx 35, \, d \approx 80$, and $m = 35^{80} \approx 10^{123}$, whereas
	For Go,
	$b \approx 250, \, d \approx 150$, and $m = 250^{150} \approx 10^{360}$.  
	See, e.g.,
	``Go and mathematics'', Wikipedia, 
	\href{https://en.wikipedia.org/w/index.php?title=Go_and_mathematics&oldid=845635147}{version 03:40, 13 June 2018};
	``Game-tree complexity'', Wikipedia,  
	\href{https://en.wikipedia.org/w/index.php?title=Game_complexity&oldid=863187839\#Game-tree_complexity}{version 07:04, 9 October 2018};
	\vphantom{\cite{Silver.2016}}
	\cite{Silver.2016}. 
}
It is ``the most challenging of classic games for artificial intelligence [AI] owing to its enormous search space and the difficulty of evaluating board positions and moves'' \cite{Silver.2016}.

This breakthrough is the crowning achievement in a string of astounding successes of deep learning (and reinforcenent learning) in taking on this difficult challenge for AI.\footnote{
	See
	\vphantom{\cite{Mnih.2015}} 
	\cite{Mnih.2015} 
	\vphantom{\cite{Silver.2016}} 
	\cite{Silver.2016}
	\vphantom{\cite{Racaniere.2017}}
	\cite{Racaniere.2017}
	\vphantom{\cite{Silver.2017:rd0001}} 
	\cite{Silver.2017:rd0001}.
	See also the film \href{https://www.alphagomovie.com/}{\em AlphaGo} (2017), ``an excellent and surprisingly touching documentary about one of the great recent triumphs of artificial intelligence, Google DeepMind's victory over the champion Go player Lee Sedol'' 
	\cite{Cellan-Jones.2017}, and  ``AlphaGo versus Lee Sedol,'' Wikipedia \href{https://en.wikipedia.org/w/index.php?title=AlphaGo_versus_Lee_Sedol&oldid=1108283463}{version 14:59, 3 September 2022}.
}
The success of this recent breakthrough prompted an AI expert to declare close the multidecade long, arduous chapter of AI research to conquer immensely-complex games such as chess, shogi, and Go, and to suggest AI researchers to consider a new generation of games to provide the next set of challenges \cite{Campbell.2018}.

In its long history, AI research went through several cycles of ups and downs, in and out of fashion, as described in \cite{Economist.2016:rd0002}, `Why artificial intelligence is enjoying a renaissance' (see also Section~\ref{sc:history} on historical perspective):

\begin{quote}
	``THE TERM ``artificial intelligence'' has been associated with hubris and disappointment since its earliest days. It was coined in a research proposal from 1956, which imagined that significant progress could be made in getting machines to ``solve kinds of problems now reserved for humans if a carefully selected group of scientists work on it together for a summer''. That proved to be rather optimistic, to say the least, and despite occasional bursts of progress and enthusiasm in the decades that followed, AI research became notorious for promising much more than it could deliver.  Researchers mostly ended up avoiding the term altogether, preferring to talk instead about ``expert systems'' or ``neural networks''. But in the past couple of years there has been a dramatic turnaround.  Suddenly AI systems are achieving impressive results in a range of tasks, and people are once again using the term without embarrassment.''
\end{quote}

The recent resurgence of enthusiasm for AI research and applications dated only since 2012 with a spectacular success of almost halving the error rate in image classification in the ImageNet competition,\footnote{
	\label{fn:imagenet}
	``ImageNet is an online database of millions of images, all labelled by hand. For any given word, such as ``balloon'' or
	``strawberry'', ImageNet contains several hundred images. The annual ImageNet contest encourages those in the field
	To compete and measure their progress in getting computers to recognise and label images automatically''
	\cite{Economist.2016:rd0001}.
	See also
	\vphantom{\cite{Russakovsky.2015}} 
	\cite{Russakovsky.2015} and 
	\cite{Gershgorn.2017:rd0001}, for a history of the development of ImageNet, which played a critical role in the resurgence of interest and research in AI by paving the way for the mentioned 2012 spectacular success in reducing the error rate in image recognition.
}
Going from 26\% down to 16\%; Figure~\ref{fig:ImageNet-error}
\vphantom{\cite{Park.2017:rd0001}} 
\cite{Park.2017:rd0001}.
In 2015, deep-learning error rate of 3.6\% was smaller than human-level error rate of 5.1\%,\footnote{
	For a report on the human image classification error rate of 5.1\%, see \cite{Dodge.2017:rd0001} and
	\vphantom{\cite{Russakovsky.2015}}
	\cite{Russakovsky.2015}, 
	Table 10.
} and then decreased by more than half to 2.3\% by 2017.

The 2012 success\footnote{
	Actually, the first success of deep learning occurred three years earlier in 2009 in speech recognition; see Section~\ref{sc:resurgence} regarding the historical perspective on the resurgence of AI.
} of a deep-learning application, which brought renewed interest in AI research out of its recurrent doldrums known as ``AI winters'',\footnote{
	See \cite{Economist.2016:rd0002}.
} is due to the following reasons:

\begin{itemize}
	
	\item 
	Availability of much larger datasets for training deep neural networks (find optimized parameters).  
	It is possible to say that without ImageNet, there would be no spectacular success in 2012, and thus no resurgence of AI. 
	Once the importance of having large datasets to develop versatile, working deep networks was realized, many more large datasets have been developed.
	See, e.g., \cite{Gershgorn.2017:rd0001}.
	
	\item 
	Emergence of more powerful computers than in the 1990s, e.g., the graphical processing unit (or GPU), ``which packs thousands of relatively simple processing cores on a single chip'' for use to process and display complex imagery, and to provide fast actions in today's video games'' \cite{Hardesty.2017:rd0001}.
	
	\item 
	Advanced software infrastructure (libraries) that facilitates faster development of deep-learning applications, e.g., TensorFlow, PyTorch, Keras, MXNet, etc.   \cite{Goodfellow.2016}, p.~25. 
	See Section~\ref{sc:libraries} on some reviews and rankings of deep-learning libraries.
			
	\item 
	Larger neural networks and better training techniques (i.e., optimizing network parameters) that were not available in the 1980s.  Today's much larger networks, which can solve once intractatable / difficult problems, are ``one of the most important trends in the history of deep learning'', but are still much smaller than the nervous system of a frog \cite{Goodfellow.2016}, p.~21; see also Section \ref{sc:depth-size}.
	A 2006 breakthrough, ushering in the dawn of a new wave of AI research and interest, has allowed for efficient training of deeper neural networks \cite{Goodfellow.2016}, p.~18.\footnote{
		\label{fn:2006-breakthough-paper}
		%
		% CMES style rewriting
		The authors of
		\cite{LeCun.2015:rd0001} cited this 2006 breakthrough paper by Hinton, Osindero \& Teh in their reference no.32 with the mention ``This paper introduced a novel and effective way of training very deep neural networks by pre-training one hidden layer at a time using the unsupervised 
		learning procedure for restricted Boltzmann machines (RBMs).'' 
		% We are not reviewing this method here as it has not been applied to computational mechanics.
		A few years later, it was found out that RBMs were not necessary to train deep networks, as it was sufficient to use rectified linear units (ReLUs) as active functions (\cite{Ford.2018}, interview with Y. Bengio); see also Section~\ref{sc:activation-functions} on active functions.  
		For this reason, we are not reviewing RBMs here.
	}
	The training of large-scale deep neural networks, which frequently involve highly nonlinear and non-convex optimization problems with many local minima, owes its success to the use of {\em stochastic-gradient} descent method first introduced in the 1950s \cite{Bottou.2018:rd0001}.   
	
	\item 
	Successful applications to difficult, complex problems that help people in their every-day lives, e.g., image recognition, speech translation, etc.

	\indent
	$\star$ In medicine, AI ``is beginning to meet (and sometimes exceed) assessments by doctors in various clinical situations. A.I. can now diagnose skin cancer like dermatologists, seizures like neurologists, and diabetic retinopathy like ophthalmologists. Algorithms are being developed to predict which patients will get diarrhea or end up in the ICU,\footnote{
		Intensive Care Unit.
	} and the FDA\footnote{
		Food and Drug Administration.
	} recently approved the first machine learning algorithm to measure how much blood flows
	through the heart---a tedious, time-consuming calculation traditionally done by cardiologists.''  Doctors lamented that they spent ``a decade in medical training learning the art of diagnosis and treatment,'' and were now easily surpassed by computers
	\cite{Khullar.2019:rd0001}. 
	``The use of artificial intelligence is proliferating in American health care---outpacing the development of government regulation. From diagnosing patients to policing drug theft in hospitals, AI has crept into nearly every facet of the health-care system, eclipsing the use of machine intelligence in other industries'' \cite{Kornfield.2020}.

	\indent
	$\star$ In micro-lending, AI has helped the Chinese company SmartFinance reduce the default rates of more than 2 millions loans per month to low single digits, a track record that makes traditional
	brick-and-mortar banks extremely jealous'' \cite{Lee2018a}.
	
	$\star$ In the popular TED talk ``How AI can save humanity'' \cite{Lee2018b}, the speaker alluded to the above-mentioned 2006 breakthrough (\cite{Goodfellow.2016}, p.~18) that marked the beginning of the ``deep learning'' wave of AI research when he said:\footnote{
		At video time 1:51.
		In less than a year, this 2018 April TED talk had more than two million views as of 2019 March. 
	} ``About 10 years ago, the grand AI discovery was made by three North American scientists,\footnote{
		See Footnote \ref{fn:2006-breakthough-paper} for the names of these three scientists.
	} and it's known as deep learning''.

	% Khullar 2019, AI could worsen health inequalities, New York Times, Jan 31.

\end{itemize}

Section~\ref{sc:history} provices a historical perspective on the development of AI, with additional details on current and future applications.

It was, however, disappointing that despite the above-mentioned exciting outcomes of AI, during the Covid-19 pandemic beginning in 2020,\footnote{
	``The World Health Organization declares COVID-19 a pandemic'' on 2020 Mar 11, \href{https://www.cdc.gov/museum/timeline/covid19.html}{CDC Museum COVID-19 Timeline}, \href{https://web.archive.org/web/20220602020133/https://www.cdc.gov/museum/timeline/covid19.html}{Internet archive 2022.06.02}.
} none of the hundreds of AI systems developed for Covid-19 diagnosis were usable for clinical applications; see Section~\ref{sc:COVID-19}.  
As of June 2022, the Tesla electric vehicle autopilot system is under increased scrutiny by the National Highway Traffic Safety Administration as there were ``16 crashes into emergency vehicles and trucks with warning signs, causing 15 injuries and one death.''\footnote{
	Krisher T., \href{https://apnews.com/article/technology-health-business-8fc617fc492847d15bf253558ed1f925}{Teslas with Autopilot a step closer to recall after wrecks}, Associated Press, 2022.06.10.
}
In addition, there are many limitations and danger in the current state-of-the-art of AI; see Section~\ref{sc:closure}.

\subsection{Handwritten equation to LaTeX code, image recognition}
\label{sc:handwriting}
An image-recognition software useful for computational mechanicists is
\href{https://mathpix.com/}{Mathpix Snip},\footnote{
	We thank Kerem Uguz for informing the senior author LVQ about Mathpix.
} which recognizes hand-written math equations, and transforms them into LaTex codes.
For example, \href{https://mathpix.com/}{Mathpix Snip} transforms the hand-written equation below by an 11-year old pupil:
\begin{figure}[H]
  \centering
  \includegraphics[width=0.5\linewidth]{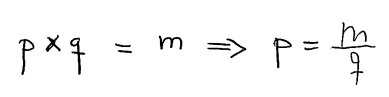}
  \caption{
  \emph{Handwritten equation 1} (Section~\ref{sc:handwriting})
  }
  \label{fig:handEq1}
\end{figure}
\noindent
into this LaTeX code ``\verb+p \times q = m \Rightarrow p = \frac { m } { q }+'' to yield the equation image:
\begin{align}
	p \times q = m \Rightarrow p = \frac { m } { q }
	\label{eq:handEq1}
\end{align}
Another example is the hand-written multiplication work below by the same pupil:
\begin{figure}[H]
  \centering
  \includegraphics[width=0.2\linewidth]{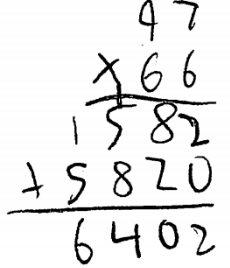}
  \caption{
  \emph{Handwritten equation 2} (Section~\ref{sc:handwriting}). Hand-written multiplication work of an eleven-year old pupil.
  }
  \label{fig:handEq5}
\end{figure}
\noindent
that \href{https://mathpix.com/}{Mathpix Snip} transformed into the equation image below:\footnote{
	\href{https://mathpix.com/}{Mathpix Snip} ``misunderstood'' that the top horizontal line was part of a fraction, and upon correction of this ``misunderstanding'' and font-size adjustment yielded the equation image shown in Eq.~(\ref{eq:handEq5b}).
}
\begin{align}
	\left. \begin{array} { r } { 97 } \\  { \times 66 } \\ \hline  { 1582 } \\ { + 5820 } \\ \hline 6402 \end{array} \right.
	\label{eq:handEq5b}
\end{align}

% \subsection{Deep learning, machine learning, artificial intelligence}
\subsection{Artificial intelligence, machine learning, deep learning}
\label{sc:AI-machine-learning}
We want to immediately clarify the meaning of the terminologies ``Artificial Intelligence'' (AI), ``Machine Learning'' (ML), and ``Deep Learning'' (DL), since their casual use could be confusing for first-time learners.

For example, 
%
% CMES style, rewriting
%\cite{Dunjko.2018:rd7724}---reviewing 
it was stated in a review of
primarily two computer-science topics called ``Neural Networks'' (NNs) and ``Support Vector Machines'' (SVMs) and a physics topic
%---stated:
that \cite{Dunjko.2018:rd7724}:\footnote{
	We are only concerned with NNs, not SVMs, in the present paper.
} 
\begin{quote}
	``The respective underlying fields of basic research---quantum 
	information versus machine learning (ML) and artificial intelligence (AI)---have their own  specific questions and challenges, which have hitherto been investigated largely independently.''
\end{quote}

Questions would immediately arise in the mind of first-time learners:
Are ML and AI two different fields, or the same fields with different names?  If one field is a subset of the other, then would it be more general to just refer to the larger set?  On the other hand, would it be more specific to just refer to the subset?

In fact, Deep Learning is a subset of methods inside a larger set of methods known as Machine Learning, which in itself is a subset of methods generally known as Artificial Intelligence.  In other words, Deep Learning is Machine Learning, which is Artificial Intelligence;
\cite{Goodfellow.2016}, p.~9.\footnote{
	\label{fn:pageNumber}
	References to books are accompanied with page numbers for specific information cited here so readers don't waste time to wade through an 800-page book to look for such information.
}  
On the other hand, Artificial Intelligence is not necessarily Machine Learning, which in itself is not necessarily Deep Learning.

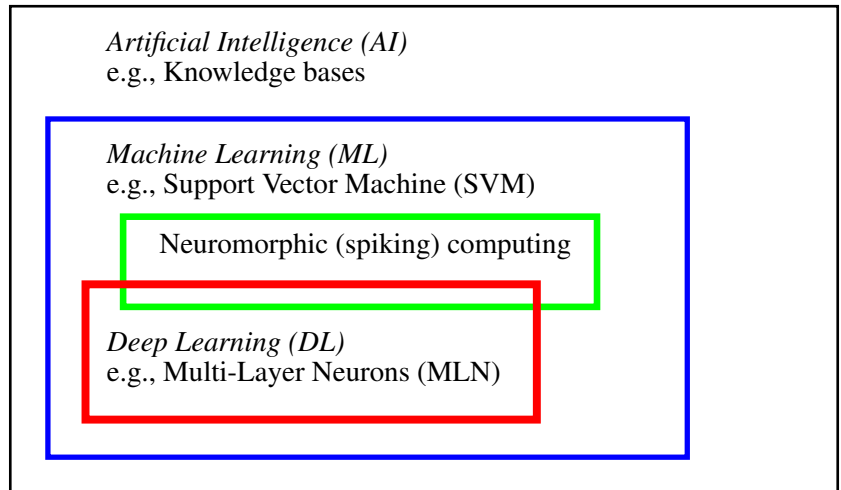
\begin{figure}[h]
  \centering
  % tikz tanh
  % https://tex.stackexchange.com/questions/176101/plotting-the-graph-of-hyperbolic-tangent
  % tikz plot function axes same scale
  % https://tex.stackexchange.com/questions/411465/same-scale-on-the-axes-but-different-lengths-of-the-axes/411472
  % tikz plot function scale figure
  % https://tex.stackexchange.com/questions/105570/how-to-plot-functions-like-x-fy-using-tikz
  % added "scale = 1.5" in the preamble of the axis environment to scale up the figure
  % added "smooth" so that the curve is smooth instead of piecewise linear, and having kinks
  
  % sigmoidal logistic function
  % need to use the following tikz libraries (add calc)
  % \usetikzlibrary{arrows,intersections,calc}
  \begin{tikzpicture}[
      thick,
      >=stealth',
      dot/.style = {
        draw,
        fill = white,
        circle,
        inner sep = 0pt,
        minimum size = 4pt
      }
    ]

	% artificial intelligence
    \draw[black, line width = 1 pt] (0,0) rectangle (11,6.5);
    \draw (1,6.0) node[label = {right:{\em Artificial Intelligence (AI)}}] {};
    \draw (1,5.6) node[label = {right:e.g., Knowledge bases}] {};
    
   	% machine learning
    \draw[blue, line width = 2 pt] (0.5,0.5) rectangle (9,5.0);
    \draw (1,4.5) node[label = {right:{\em Machine Learning (ML)}}] {};
    % \draw (1,4.1) node[label = {right:e.g., Logistic regression}] {};
    \draw (1,4.1) node[label = {right:e.g., Support Vector Machine (SVM)}] {};
    
    % in between machine learning and deep learning
    \draw[green, line width = 2.5 pt] (1.5,2.5) rectangle (7.8,3.7);
    \draw (1.7,3.3) node[label = {right:Neuromorphic (spiking) computing}] {};
    
    % deep learning
    \draw[red, line width = 3 pt] (1.0,1.0) rectangle (7,2.8);
    \draw (1,2.0) node[label = {right:{\em Deep Learning (DL)}}] {};
    \draw (1,1.6) node[label = {right:e.g., Multi-Layer Neurons (MLN)}] {};
  
  \end{tikzpicture}
  \caption{
  	\emph{Artificial intelligence and subfields} (Section~\ref{sc:AI-machine-learning}).
  	Three classes of methods---{\em Artificial Intelligence} (AI), {\em Machine Learning} (ML), and {\em Deep Learning} (DL)---and their relationship, with an example of method in each class.  A knowledge-base method is an AI method, but is neither a ML method, nor a DL method.  Support Vector Machine and spiking computing are ML methods, and thus AI methods, but not a DL method.  Multi-Layer Neural (MLN) network is a DL method, and is thus both an ML method and an AI method. 
  	See also Figure~\ref{fig:cybernetics} in
  	Appendix~\ref{app:cybernetics} on {\em Cybernetics},
%  	\ref{app:cybernetics} {\em Cybernetics},  
  	which  encompassed all of the above three classes.
  %{\color{red} [NOTE 2020.04.02: ``See also Figure~\ref{fig:cybernetics} on {\em Cybernetics} as encompassing the above three classes.'' This Figure~\ref{fig:cybernetics} had been moved to Section NOTES.  if we make cybernetics Appendix 3, then the figure would be included. ENDNOTE]}
  % {\em Cybernetics} is encompassing, and will be discussed in Section~\ref{sc:3wavesAI} on the ups and downs of AI; see Figure~\ref{fig:cybernetics}.
  % Section~\ref{sc:history} on Historical perspective.
  }
  \label{fig:AI.ML.DL}
\end{figure}

%
% CMES style, rewriting
%\cite{Dunjko.2018:rd7724} restricted their reviews 
The review in \cite{Dunjko.2018:rd7724} was restricted
to Neural Networks (which could be deep or shallow)\footnote{
	Network depth and size are discussed in Section~\ref{sc:depth}.  An example of a shallow network with one hidden layer can be found in Section~\ref{sc:ROM-hyper} on nonlinear-manifold model-order reduction applied to fluid mechanics.
} and Support Vector Machine (which is Machine Learning, but not Deep Learning); see Figure~\ref{fig:AI.ML.DL}.
Deep Learning can be thought of as multiple levels of composition, going from simpler (less abstract) concepts (or representations) to more complex (abstract) concepts (or representations).\footnote{
 See, e.g., \cite{Goodfellow.2016}, p.~5, p.~8, p.~14.
}

Based on the above relationship between AI, ML, and DL, it would be much clearer if 
%
% CMES style, rewriting
%\cite{Dunjko.2018:rd7724} had simply replaced 
the phrase ``machine learning (ML) and artificial intelligence (AI)'' in both the title of \cite{Dunjko.2018:rd7724} and the original sentence quoted above is replaced by the phrase ``machine learning (ML)'' to be more specific, since the authors mainly reviewed Multi-Layer Neural (MLN) networks (deep learning, and thus machine learning) and Support Vector Machine (machine learning).\footnote{
	\label{fn:support-vector-machine}
	For more on Support Vector Machine (SVM), see \cite{Goodfellow.2016}, p.~137.  In the early 1990s, SVM displaced neural networks with backpropagation as a better method for the machine-learning community (\cite{Ford.2018}, interview with G. Hinton). The resurgence of AI due to advances in deep learning started with the seminal paper \cite{Hinton.2006:rd0001}, in which the authors demonstrated via numerical experiments that MLN network was better than SVM in terms of error in the handwriting-recognition benchmark test using the 
	\href{http://yann.lecun.com/exdb/mnist/}{MNIST handwritten digit database},
	%\href{http://yann.lecun.com/exdb/mnist/}{http://yann.lecun.com/exdb/mnist/}, 
	which contains ``a training set of 60,000 examples, and a test set of 10,000 examples.''
	But kernel methods studied for the development of SVM have now been used in connection with networks with infinite width to understand how deep learning works; see Section~\ref{sc:kernel-machines} on ``Kernel machines'' and Section~\ref{sc:lack-understanding} on ``Lack of understanding.''
	%{\color{red}
	%	[NOTE: 2018.10.15, we may move the content of this footnote to Section~\ref{sc:history} on Historical perspective.  ENDNOTE]
	%}
}
MultiLayer Neural (MLN) network is also known as MultiLayer Perceptron (MLP).\footnote{
	\label{fn:MLP-1}
	See \cite{Goodfellow.2016}, p.~5.
}
both MLN networks and SVMs are considered as artificial intelligence, which in itself is too broad and thus not specific enough.  
% for more specific definition of AI, ML, and DL, see Section

Another reason for simplifying the title in \cite{Dunjko.2018:rd7724} is that the authors did not consider using any other AI methods, except for two specific ML methods, even though they discussed AI in the general historical context.  

The engine of neuromorphic computing, also known as spiking computing, is a hardware network built into the IBM TrueNorth chip, which contains ``1 million programmable spiking neurons and 256 million configurable synapses'',\footnote{
	The neurons are the computing units, and the synapses the memory.
	instead of grouping the computing units into a central processing unit (CPU), separated from the memory, and connect the CPU and the memory via a bus, which creates a communication bottleneck, like the brain, each neuron in the TrueNorth chip has its own synapses (local memory).} 
and consumes ``extremely low power'' \vphantom{\cite{Merolla.2014}}\cite{Merolla.2014}.   
Despite the apparent difference with the software approach of deep computing, 
%
% CMES style, rewriting
%\vphantom{\cite{Esser.2016}}\cite{Esser.2016} showed that
neuromorphic chip could implement deep-learning networks, and thus the difference was not fundamental \vphantom{\cite{Esser.2016}}\cite{Esser.2016}.  
There is thus an overlap between neuromorphic computing and deep learning, as shown in Figure~\ref{fig:AI.ML.DL}, instead of two disconnected subfields of machine learning as reported in \cite{Sze.2017:rd0001}.\footnote{
	%
	% CMES style, rewriting
%	\cite{Sze.2017:rd0001} only referred 
	In \cite{Sze.2017:rd0001}, there was only a reference to \cite{Merolla.2014}, but not to \cite{Esser.2016}.
	It is likely that the authors of \cite{Sze.2017:rd0001} were not aware of \cite{Esser.2016}, and thus 
%	were not aware that there was 
	an intersection between neuromorphic computing and deep learning.
}      

% {\color{red} HERE 2019.04.13}

\begin{figure}[h]
	\centering
	% original figure, not rotated
	% \includegraphics[width=0.5\linewidth]{figures/Oishi-network.png}
	% figure rotated clockwise by 90 deg
	\includegraphics[width=0.5\linewidth]{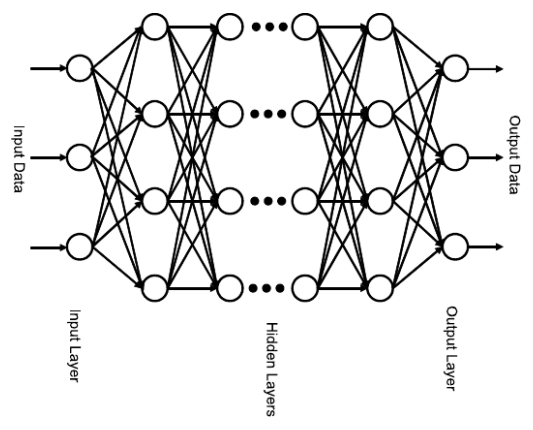}
	\caption{
		\emph{Feedforward neural network} (Section~\ref{sc:Oishi-summary}).
		A feedforward neural network in \cite{Oishi.2017:rd9648}, rotated clockwise by 90 degrees to compare to its equivalent in Figure~\ref{fig:network3b} and Figure~\ref{fig:neuron4} further below.
		All terminologies and fundamental concepts will be explained in detail in subsequent sections as \hyperref[para:concepts-1]{listed}.
		See Section~\ref{sc:top-down} for a top-down explanation and Section~\ref{sc:bottom-up} for bottom-up explanation.
		This figure of a network could be confusing to first-time learners, as already indicated in Footnote~\ref{fn:confusion}.
		%\\
		{\footnotesize (Figure reproduced with permission of the authors.)}
	}
	\label{fig:Oishi-network}
\end{figure}

\begin{figure}[h]
	\centering
	\includegraphics[width=0.5\linewidth]{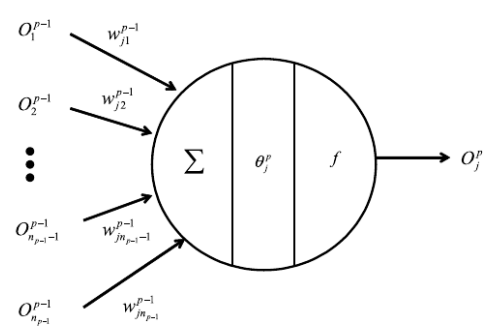}
	\caption{
		\emph{Artificial neuron} (Section~\ref{sc:Oishi-summary}).
		A neuron with its multiple inputs $O_{i}^{p-1}$ (which are outputs from the previous layer $(p-1)$, and thus the variable name ``$O$''), processing operations (multiply inputs with network weights $w_{ji}^{p-1}$, sum weighted inputs, add bias $\theta_j^p$, activation function $f$), and single output $O_j^p$ \cite{Oishi.2017:rd9648}.
		See 
%		\cite{Oishi.2017:rd9648} and   
		the equivalent Figure~\ref{fig:neuron5}, Section~\ref{sc:block-diagrams} further below.
		All terminologies and fundamental concepts will be explained in detail in subsequent sections as \hyperref[para:concepts-1]{listed}. 
		See Section~\ref{sc:feedforward} on feedforward networks,   Section~\ref{sc:top-down} on top-down explanation and Section~\ref{sc:bottom-up} on bottom-up explanation.
		This figure of a neuron could be confusing to first-time learners, as already indicated in Footnote~\ref{fn:confusion}.
		%\\
		{\footnotesize (Figure reproduced with permission of the authors.)}
	}
	\label{fig:Oishi-neuron}
\end{figure}

\begin{figure}[h]
	\centering
	\includegraphics[width=0.4\textwidth]{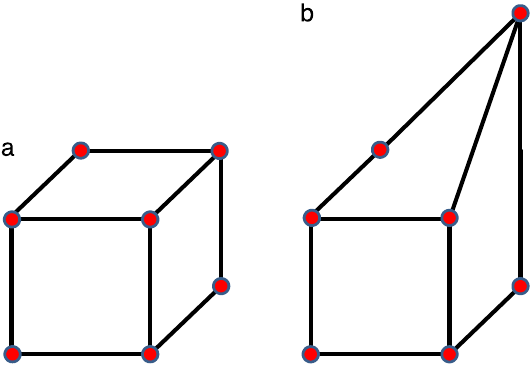}
	\caption{
		\emph{Cube and distorted cube elements}  (Section~\ref{sc:Oishi-summary}).
		Regular and distorted linear hexahedral elements 
		\cite{Oishi.2017:rd9648}.
		%\\
		{\footnotesize (Figure reproduced with permission of the authors.)}
	}
	\label{fig:Oishi-hexahedral}
\end{figure}

\subsection{Motivation, applications to mechanics}
\label{sc:motivation}

%\begin{center}
%	\label{para:past-tense}
%\framebox{
%	\parbox{0.9\linewidth}{
%		{\color{red} 
%			[NOTE Updated 2020.05.24. \emph{General principle on the use of present and past tenses.} What authors such as \cite{Oishi.2017:rd9648} \emph{particularly} did in their work (and could be done otherwise by other authors) should be written in past tense, except for what are commonly accepted as universal truths, then the present tense should be used.   ENDNOTE]
%		}
%	}
%}
%\end{center}

As motivation, we present in this section the results in three recent papers in computational mechanics, mentioned in the Opening Remarks in Section~\ref{sc:open}, and identify some deep-learning fundamental concepts (in \emph{italics}) employed in these papers, together with the corresponding sections in the present paper where these concepts are explained in detail.
First-time learners of deep learning likely find these fundamental concepts described by obscure technical jargon, whose meaning will be explained in details in the identified subsequent sections.
Experts of deep learning would understand how deep learning is applied to computational mechanics.

%{\color{red} 	
%	[NOTE: 2020.04.02.	
%	Since the figures are scattered when using the option ``[h]'',
%	in each figure caption, let's include the section number where the figure is mentioned so readers and we can jump to that section quickly.
%	I added some comments in red in this particular Section~\ref{sc:motivation}.  once you have checked, then just remove the red color, and comment out this note.
%  	ENDNOTE]  
%
%}

% 2019.12.03.
% Mark: 01-motivation.tex, questions to Oishi, Barlow points, reduced integration

\subsubsection{Enhanced numerical quadrature for finite elements}
\label{sc:Oishi-summary}

%{\color{red} 	
%	[NOTE 2020.05.03, i added adjectives to distinguish ``quadrature weights'' from ``network weights'' to avoid confusion; see also Footnote~\ref{fn:weights}. 	
%ENDNOTE]}

%\noindent
%{\color{red} [NOTE: 2020.04.06.	We should try to refer readers to the corresponding equations in subsequent sections for obscure concepts as soon as possible, as done above. ENDNOTE]}

\noindent
To integrate efficiently and accurately the element matrices in a general finite element mesh of 3-D hexahedral elements (including distorted elements),
%
% CMES style, rewriting
%\cite{Oishi.2017:rd9648} harnessed 
the power of Deep Learning was harnessed in two applications of \emph{feedforward MultiLayer Neural networks}  (MLN,\footnote{
	\label{fn:MLP-2}
	MLN is also called MultiLayer Perceptron (MLP); see Footnote~\ref{fn:MLP-1}.
}  Figures~\ref{fig:Oishi-network}-\ref{fig:Oishi-neuron}, Section~\ref{sc:feedforward}) \cite{Oishi.2017:rd9648}:
% (Figure~\ref{fig:Oishi-network}, Figure~\ref{fig:Oishi-neuron}):

\begin{enumerate}

\item 
% Method 1: 
Application 1.1:
For each element (particularly distorted elements), find the number of integration points that provides accurate integration within a given error tolerance.  
Section~\ref{sc:Oishi-1.1} contains the details.
\label{Oishi:method1}

\item 
% Method 2: 
Application 1.2:
Uniformly use $2 \times 2 \times 2$ integration points for all elements, distorted or not, and find the appropriate quadrature weights\footnote{
	\label{fn:weights}
	The \emph{quadrature} weights at integration points are not to be confused with the \emph{network} weights in a MLN network.
} ({\em different} from the traditional quadrature weights of the Gauss-Legendre method) at these integration points.
Section~\ref{sc:Oishi-1.2} contains the details.
\label{Oishi:method2}
\end{enumerate}

To \emph{train}\footnote{
	See Section~\ref{sc:training} on ``Network training, optimization methods''.
} the networks---i.e., to optimize the network parameters (weights and biases, Figure~\ref{fig:Oishi-neuron}) to minimize some \emph{loss (cost, error) function} 
%
% CMES style, rewriting
%(Sections~\ref{sc:cost-function}, \ref{sc:training})---\cite{Oishi.2017:rd9648} generated up to
(Sections~\ref{sc:cost-function}, \ref{sc:training})---up to
\num{20000} randomly distorted hexahedrals were generated by displacing nodes from a regularly shaped element  \cite{Oishi.2017:rd9648}, see Figure~\ref{fig:Oishi-hexahedral}. % and Figure~\ref{fig:training-set-hexa}. 
For each distorted shape, 
%
% CMES style, rewriting
%they determined 
the following are determined:
\ref{Oishi:method1} the minimum number of integration points required to reach a prescribed accuracy, and \ref{Oishi:method2} corrections to the quadrature weights by trying one million randomly generated sets of correction factors, among which the best one was retained.

While Application 1.1 used one \emph{fully-connected} (Section~\ref{sc:depth}) {feedforward neural network} (Section~\ref{sc:feedforward}), Application 1.2 relied on two neural networks: The first neural network was a classifier that took the element shape (18 normalized nodal coordinates) as input and estimated whether or not the numerical integration (quadrature) could be improved by adjusting the quadrature  weights for the given element (one output), i.e., the network classifier only produced two outcomes, yes or no.
If an error reduction was possible, a second neural network performed regression to predict the corrected quadrature  weights (eight outputs for $2 \times 2 \times 2$ quadrature) from the input element shape (usually distorted).

{%\color{blue} 
To train the classifier network, \num{10000} element shapes were selected from the prepared dataset of \num{20000} hexahedrals, 
%(Figure~\ref{fig:training-set-hexa})
which were divided into a \emph{training set} and a \emph{validation set} (Section~\ref{sc:training-valication-test}) of \num{5000} elements each.\footnote{
	For the definition of training set and test set, see Section~\ref{sc:training-valication-test}. Briefly, the training set is used to optimize the network parameters, while the test set is used to see how good the network with these optimized parameters can predict the targets of never-seen-before inputs. 
}

To train the second regression network, 
%
% CMES style, rewriting
%\cite{Oishi.2017:rd9648} selected 
\num{10000} element shapes were selected for which quadrature could be improved by adjusting the quadrature weights \cite{Oishi.2017:rd9648}. 

Again, the training set and the test set comprised \num{5000} elements each. 
The parameters of the neural networks (\emph{weights, biases}, Figure~\ref{fig:Oishi-neuron}, Section~\ref{sc:network-layer-details}) were optimized (trained) using a \emph{gradient descent method} (Section~\ref{sc:training}) that minimizes a \emph{loss function} (Section~\ref{sc:cost-function}), whose gradients with respect to the parameters are computed using \emph{backpropagation} (Section~\ref{sc:backprop}).
}
%\sout{Once the training-data set is prepared (Figure~\ref{fig:training-set-hexa}), the parameters of the neural networks (weights, biases) are optimized (trained) using a gradient descent method that reduces the loss function, whose gradients is computed using backpropagation.}

\begin{figure}[h]
	\centering
	\includegraphics[]{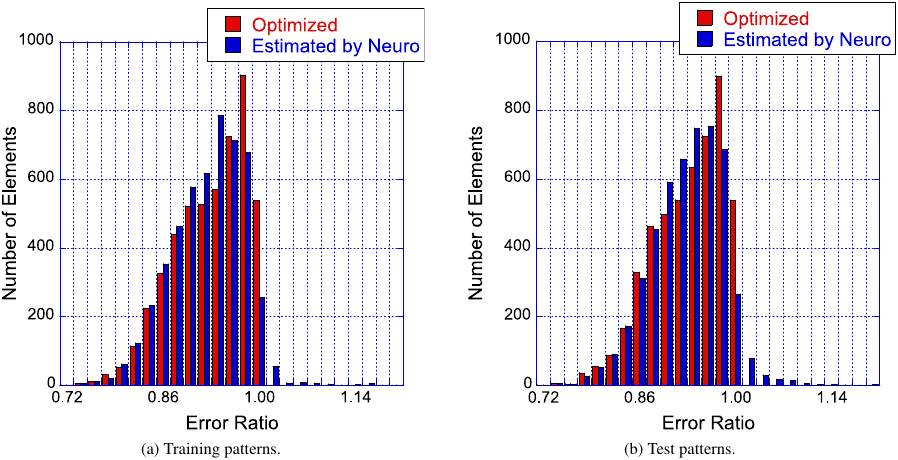}
	\caption{
		{\it Effectiveness of quadrature weight prediction}  (Section~\ref{sc:Oishi-summary}).
		%The effectiveness of the quadrature weight prediction is quantified by the error-reduction ratio $R_{\rm error}$.
		Subfigure (a): Distribution of error-reduction ratio $R_{\rm error}$ for the 5,000 elements in the training set.  
		The red bars (``Optimized'') are the error ratios obtained from the optimal weights {(found by a large number of trials-and-errors)} that were used to train the network.  The blue bars (``Estimated by Neuro'') are the error ratios obtained from the trained neural network.  
		$R_{\rm error} < 1$ indicates improved quadrature accuracy.
		As a result of using the optimal weights, there were no red bars with $R_{\rm error} > 1$.  That there were very few blue bars with $R_{\rm error} > 1$ showed that the proposed method worked in reducing the integration error in more than 97\% of the elements. 
		%see also Figure~\ref{fig:Oishi-2017-weight-correction-classification-net} in Section~\ref{sc:Oishi-1.2}.
		Subfigure (b): Error ratios for the test set with \num{5000} elements \cite{Oishi.2017:rd9648}.
		More detailed explanation is provided in Section~\ref{sc:Oishi-method-2-application}. 
		%See also Figure~\ref{fig:Oishi-2017-weight-correction-results}, Section~\ref{sc:Oishi-method-2-application}.
		%{\color{red}
		%[NOTE: 2018.12.07, the error-ratio distribution is still not clear.  take SubFigure (a), for the bar with error ratio = 1, i.e., there was no improvement, why its height was not equal to 15,000 - 3,000 = 12,000, if 3,000 was the sum of all the bar heights with error ratios below 1 ? i.e., there should be 12,000 elements in which there was no improvement.  something is still not clear.  
		%what is the sum of the blue bar heights with error ratios above 1 ?  it looks like around 10, which was very small compared to 15,000.
		%ENDNOTE]
		%{\color{blue}
		%5000 elements are used to train the network, not \num{15000}. The sum of the bars therefore does make sense.
		%}
		%}
%		\cite{Oishi.2017:rd9648}.
		%\\
		{\footnotesize (Figure reproduced with permission of the authors.)}
	}
	\label{fig:Oishi-error}
\end{figure}

The best results were obtained from a classifier with four \emph{hidden layers} (Figure~\ref{fig:Oishi-network}, Section~\ref{sc:big-picture}, Remark~\ref{rm:hidden}) with 30 \emph{neurons} (Figure~\ref{fig:Oishi-neuron}, Figure~\ref{fig:neuron5}, Section~\ref{sc:block-diagrams}) each and a regression network that had a depth of five hidden layers, where each layer was \num{50} neurons wide, Figure~\ref{fig:Oishi-network}.
The results were obtained using the \emph{logistic sigmoid function} (Figure~\ref{fig:sigmoid}) as \emph{activation function} (Section~\ref{sc:activation-functions}) due to existing software, even though the \emph{rectified linear function} (Figure~\ref{fig:ReLU}) were more efficient, but yielded comparable accuracy on a few test cases.\footnote{
	Information provided by author A. Oishi of \cite{Oishi.2017:rd9648} through a private communication to the authors on 2018 Nov 16.
}

{
To quantify the 
%
% CMES style, rewriting
%\emph{effectivity} 
effectiveness of the approach in \cite{Oishi.2017:rd9648}, 
%\cite{Oishi.2017:rd9648} introduced 
an error-reduction ratio was introduced, i.e., the quotient of the quadrature error with quadrature weights predicted by the neural network and the error obtained with the standard quadrature weights of Gauss-Legendre quadrature with $2 \times 2 \times 2$ integration points; see Eq.~(\ref{eq:Oishi-error-reduction-ratio}) in Section~\ref{sc:Oishi-1.2} with $q=2$ and ``$opt$'' stands for ``optimized'' (or ``predicted'').
When the error-reduction ratio is less than 1, the integration using the predicted quadrature weights is more accurate than that using the standard quadrature weights. 
To compute the two quadrature errors mentioned above (one for the predicted quadrature weights and one for the standard quadrature weights, both for the same $2 \times 2 \times 2$ integration points), the reference values considered as most accurate were obtained using $30 \times 30 \times 30$ integration points with the standard quadrature quadrature weights; see Eq.~(\ref{eq:oishi-error}) in Section~\ref{sc:Oishi-1.1} with $q_{max} = 30$.
%{\color{red} HERE 2020.04.01. explain how error-reduction ratio works, i.e., more accurate if error reduction ratio is less than 1.}

For most element shapes of both the training set (a) and the test set (b), each of which comprised \num{5000} elements, the blue bars in Figure~\ref{fig:Oishi-error} indicate an error ratio below one, i.e., the quadrature weight correction effectively improved the accuracy of numerical quadrature.
}
%The blue bars in Figure~\ref{fig:Oishi-error} show the ratio of the quadrature error with/without weight correction for the data used in training the network (a) and the test data by which the generalization error is quantified (b).
%

Readers familiar with Deep Learning and neural networks can go directly to Section~\ref{sc:integration}, where the details of the formulations in \cite{Oishi.2017:rd9648} are presented. 
Other sections are also of interest such as classic and state-of-the-art optimization methods in Section~\ref{sc:training}, attention and transformer unit in Section~\ref{sc:recurrent}, historical perspective in Section~\ref{sc:history}, limitations and danger of AI in Section~\ref{sc:closure}.

%{\color{red} NOTE: 2022.06.14 - Alex, pl check whether we still talk about ``transformer unit''.}
% AH: 2022.10.03 - yes, we do

Readers not familiar with Deep Learning and neural networks will find below a list of the concepts that will be explained in subsequent sections.  
To facilitate the reading, we also provide the section number (and the link to jump to) for each concept.

\vspace{2ex}
{\bf Deep-learning concepts to explain and explore:}
\label{para:concepts-1}
\begin{enumerate}
	%\label{para:concepts-1}
	
	\item 
	Feedforward neural network (Figure~\ref{fig:Oishi-network}): Figure~\ref{fig:network3b} and Figure~\ref{fig:neuron4}, Section~\ref{sc:feedforward}
	
	\item 
	Neuron (Figure~\ref{fig:Oishi-neuron}): Figure~\ref{fig:neuron5} in Section~\ref{sc:artificial-neuron} (artificial neuron), and Figure~\ref{fig:bio-neuron} in Section~\ref{sc:inspired-from-biology} (biological neuron)
	
	\item 
	Inputs, output, hidden layers, Section~\ref{sc:big-picture}
	
	\item 
	Network depth and width: Section~\ref{sc:big-picture}
	% \item how deep is ``deep'' ?
	
	\item 
	Parameters, weights, biases~\ref{sc:weigths-biases}
	
	\item 
	Activation functions: Section~\ref{sc:activation-functions}
	
	\item 
	What is ``deep'' in ``deep networks'' ? Size, architecture, Section~\ref{sc:depth}, Section~\ref{sc:architecture}
	
	\item 
	Backpropagation, computation of gradient: Section~\ref{sc:backprop}
	
	\item 
	Loss (cost, error) function, Section~\ref{sc:cost-function}
	
	\item 
	Training, optimization, stochastic gradient descent: Section~\ref{sc:training}
	
	\item 
	Training error, validation error, test (or generalization) error: Section~\ref{sc:training-valication-test}
	
\end{enumerate}
This list is continued further \hyperref[para:concepts-2]{below} in Section~\ref{sc:Wang-Sun-2018}.
Details of the formulation in \cite{Oishi.2017:rd9648} are discussed in Section~\ref{sc:Oishi-2}.

% 2019.12.10
% Mark: 01-motivation.tex, illustration of randomly distorted elements

%{\color{red} 
%	NOTE: 2019.01.08.  there was something still confusing.  Alex said there were 20,000 elements, out of which 5,000 were selected.  but what were the criteria for selection ?  it is not clear.
%}
%
%{\color{blue} 
%[NOTE: 2019.01.09.  they do not provide information about the selection process.  with the element shapes being random, I think it that does not really matter.
%ENDNOTE]
%}
%
%
%{\color{red} 
%	NOTE: 2019.01.09.  would the sum of the bars in Figure~\ref{fig:Oishi-error} be 5,000 ?   if 5,000 elements were used to train the network, then what the remaining 15,000 elements were used for ?   validation ?
%	Figure~\ref{fig:Oishi-error} seems to make the following sense: Out of 5,000 elements randomly selected and distorted, most of them show improvements in accuracy, as most bars are to the left of 1 (no improvements).
%	could you confirm ?
%}

\begin{figure}[h]
	\centering
	\includegraphics[width=0.7\linewidth]{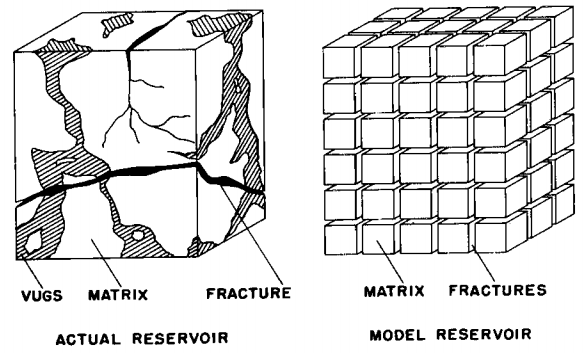}
	\caption{
		\emph{Dual-porosity single-permeability medium} (Section~\ref{sc:Wang-Sun-2018}).  \emph{Left}: Actual reservoir. Dual (or double) porosity indicates the presence of two types of porosity in naturally-fractured reservoirs (e.g., of oil): (1) Primary porosity in the matrix (e.g., voids in sands) with low permeability, within which fluid does not flow, (2) Secondary porosity due to fractures and vugs (cavities in rocks) with high (anisotropic) permeability, within which fluid flows.  Fluid exchange is permitted between the matrix and the fractures, but not between the matrix blocks (sugar cubes), of which the permeability is much smaller than in the fractures. 
		\emph{Right}: Model reservoir, idealization. The primary porosity is an array of cubes of homogeneous, isotropic material.  The secondary porosity is an ``orthogonal system of continuous, uniform fractures'', oriented along the principal axes of anisotropic permeability
		\cite{warren1963behavior}.
		{\footnotesize (Figure reproduced with permission of the publisher SPE.)}
	}
	\label{fig:dual-porosity-reservoir}
\end{figure}

\begin{figure}[h]
	\centering
	\includegraphics[width=0.7\linewidth]{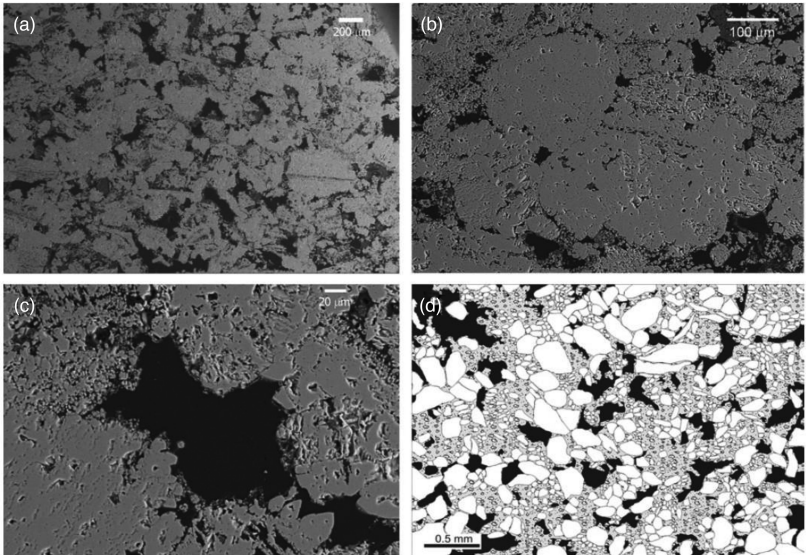}
	\caption{
		\emph{Pore structure of Majella limestone, dual porosity} (Section~\ref{sc:Wang-Sun-2018}), a carbonate rock with high total porisity at 30\%. 
		Backscattered SEM images of Majella limestone: (a)-(c) sequence of zoomed-ins; (d) zoomed-out. (a) The larger macropores (dark areas) have dimensions comparable to the grains (allochems), having an average diameter of 54 $\mu$m, with macroporosity at 11.4\%. (b) Micropores embedded in the grains and cemented regions, with microporosity at 19.6\%, which is equal to the total porosity at 30\% minus the macroporosity. (c) Numerous micropores in the periphery of a macropore. (d) Map performed manually under optical microscope  showing the partitioning of grains, matrix (mostly cement) and porosity \cite{ji2015characterization}.
		See Section~\ref{sc:Wang-Sun-2018-2} and Remark~\ref{rm:Wang-not-realistic}.  
		{\footnotesize (Figure reproduced with permission of the authors.)}
	}
	\label{fig:Majella-pore-structure}
\end{figure}

\begin{figure}[h]
	\centering
	\includegraphics[width=0.45\linewidth]{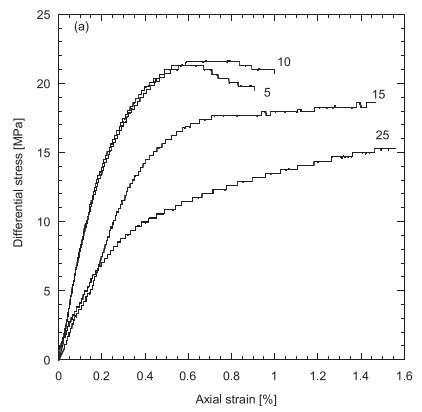}
	\includegraphics[width=0.45\linewidth]{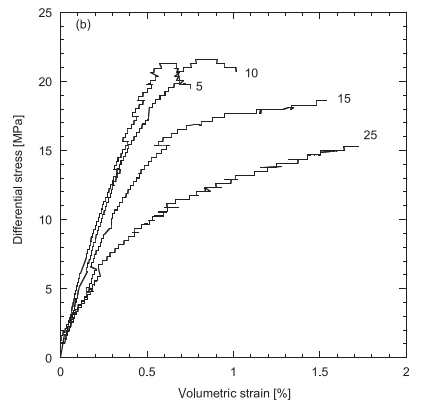}
	\caption{
		\emph{Majella limestone, nonlinear stress-strain relations} (Section~\ref{sc:Wang-Sun-2018}). 
%		Nonlinear stress-strain relation.
		Differential stress (i.e., the difference between the largest principal stress and the smallest one) vs axial strain (left) and vs volumetric strain (right) \cite{ji2015characterization}.
		See Remark~\ref{rm:Wang-field-size-simulation}, Section~\ref{sc:Wang-balance-equations}, and Remark~\ref{rm:Wang-no-nonlinear-stress-strain}, Section~\ref{sc:Wang-strong-discontinuities}.  
		{\footnotesize (Figure reproduced with permission of the authors.)}
	}
	\label{fig:Majella-stress-strain}
\end{figure}

\subsubsection{Solid mechanics, multiscale modeling}
\label{sc:Wang-Sun-2018}

One way that deep learning can be used in solid mechanics is to model complex, nonlinear constitutive behavior of materials.  In single physics, balance of linear momentum and strain-displacement relation are considered as definitions or ``universal principles'', leaving the constitutive law, or stress-strain relation, to a large number of models that have limitations, no matter how advanced \cite{Christensen.2013}.  Deep learning can help model complex constitutive behaviors in ways that traditional phenomenological models could not; see Figure~\ref{fig:Wang-2018-single-physics}.

\begin{figure}[h]
	\centering
	\includegraphics[scale=1]{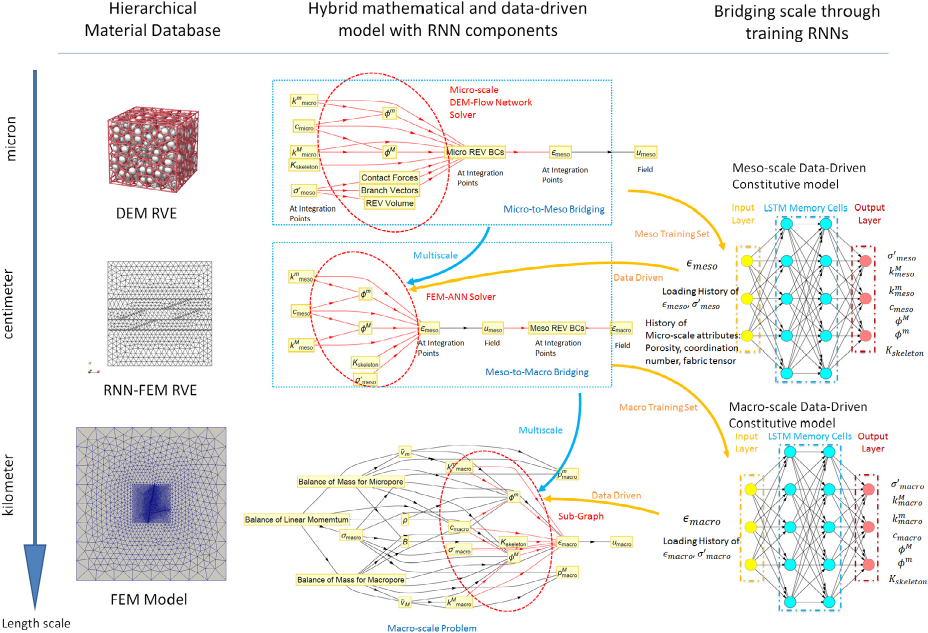}
	\caption{
		%{\color{red} 
			{\it Hierarchy of a multi-scale multi-physics poromechanics problem for fluid-infiltrating media} \cite{Wang.2018:rd3109} (Sections~\ref{sc:Wang-Sun-2018}, %\ref{sc:Wang-Sun-2018-2},
			\ref{sc:Wang-multiscale-problems}, \ref{sc:Wang-RNN-scale-bridging}, \ref{sc:Wang-optimal-RNN}, \ref{sc:Wang-strong-discontinuities}). 
			Microscale ($\mu$), mesoscale (cm), macroscale (km).
			DEM RVE = Discrete Element Method Representative Volume Element.
			RNN-FEM = Recurrent Neural Network (Section~\ref{sc:RNN}) - Finite Element Method.
			LSTM = Long Short-Term Memory (Section~\ref{sc:LSTM}).
			The mesoscale has embedded strong discontinuities equivalent to the fracture system in Figure~\ref{fig:dual-porosity-reservoir}.
			See Figure~\ref{fig:Wang-DEM-FEM-three-scales}, where the orientations of the RVEs are shown, Figure~\ref{fig:Wang-micro-RVE} for the microscale RVE (Remark~\ref{rm:Wang-micro-RVE}) and Figure~\ref{fig:Wang-mesoscale-RVE} for the mesoscale RVE (Remark~\ref{rm:Wang-meso-RVE}).
%			\cite{Wang.2018:rd3109}.
			\footnotesize (Figure reproduced with permission of the authors.)
		%}
	}
	\label{fig:Wang-RNN-FEM}
\end{figure}

%\begin{figure}[h]
%	\centering
%	\includegraphics[width=0.9\textwidth]{figures/Wang-2018-single-physics}
%	\caption{
%		\emph{Single-physics block diagram} (Section~\ref{sc:Wang-Sun-2018}).
%		Single physics is an easiest way to see the role of deep learning in modeling complex nonlinear constitutive behavior (stress-strain relation, red arrow), as first realized by \cite{Ghaboussi.1991:rd0001}, where balance of linear momentum and strain-displacement relation are definitions or accepted ``universal principles'' (black arrows). The approach is extended to multiphysics in \cite{Wang.2018:rd3109} for porous media, with the block diagram given in Figure~\ref{fig:multiphysics}.
%		%\\
%		{\footnotesize (Figure reproduced with permission of the authors.)}
%	}
%	\label{fig:Wang-2018-single-physics}
%\end{figure}

%
% CMES style, rewriting
%\cite{Wang.2018:rd3109} used 
Deep \emph{recurrent neural networks} (RNNs) (Section~\ref{sc:RNN}) was used as a scale-bridging method to efficiently simulate multiscale problems in hydromechanics,  specifically plasticity in porous media with dual porosity \emph{and} dual permeability \cite{Wang.2018:rd3109}.\footnote{
	\label{fn:dual-porosity-permeability}
	%\sout{(dual-porosity, dual-permeability poroplasticity)}.
	Porosity is the ratio of void volume over total volume.
	Permeability is a scaling factor, which when multiplied by the negative of the pressure gradient, and divided by the fluid dynamic viscosity, gives the fluid velocity in Darcy's law, Eq.~\eqref{eq:darcy-law-1}.
	%
	% CMES style, rewriting
%	\cite{Wang.2018:rd3109}, p.~340, used 
	The expression ``dual-porosity dual-permeability poromechanics problem'' used in \cite{Wang.2018:rd3109}, p.~340, could confuse first-time readers---especially those who are familiar with traditional reservoir simulation, e.g., in \vphantom{\cite{balogun2007verification}}\cite{balogun2007verification}---since dual porosity (also called ``double porosity'' in \cite{warren1963behavior}) and dual permeability are two different models of naturally-fractured porous media;
%	 \cite{ho2000dual} studied 
	these two models for radionuclide transport around nuclear waste repository were studied in \cite{ho2000dual}. 
	Further added to the confusion is that the dual-porosity model is more precisely called \emph{dual-porosity-single permeability} model, whereas the dual-permeability model is called \emph{dual-porosity dual-permeability} model \cite{datta2007streamline}, which has a different meaning than the one used in \cite{Wang.2018:rd3109}. 
	%The difference is in the rock matrix, within which the fluid cannot move globally in the dual-porosity single-permeability model, but can in the more general dual-porosity dual-permeability model, \cite{datta2007streamline}. \cite{ho2000dual} separately studied each of these two \emph{distinct} models for radionuclide transport around nuclear waste repository.  See also \vphantom{\cite{nie2012dual}}\cite{nie2012dual}, \cite{mesbah2018streamline},  \cite{lu2019new}.
	%{\color{red} [NOTE 2020.06.24. the answer appears to be YES. 2020.06.22.  the question is whether \cite{Wang.2018:rd3109} \emph{simultaneously} used both dual porosity and dual permeability in their model. ENDNOTE]}
	%On the other hand, the abstract and the list of keywords indicate that \cite{Wang.2018:rd3109} mainly focused on dual porosity, even though the Introduction was generalized to include dual permeability in the same proposed framework.
}

%\noindent
%{\color{red} 2020.07.24.  we can add more to the explanation of dual-porosity dual-permeability as in \cite{datta2007streamline}, but what we have now is sufficient for first-time readers not to get lost in new terminologies. 2020.07.03. To explain: What is dual porosity and dual permeability.  some of this material would be moved to Section~\ref{sc:Wang-Sun-2018-2}.}

%To model naturally-fractured reservoirs, \vphantom{\cite{barenblatt1960basic}}\cite{barenblatt1960basic} introduced a dual-media approach in which the fracture system and the rock matrix were ``two separate continua throughout the reservoir'', \vphantom{\cite{datta2007streamline}}\cite{datta2007streamline}, p.~295, ``but the analytical solutions to these equations are very complex and inconvenient to use'', \cite{lu2019new}.

The \emph{dual-porosity single-permeability} (DPSP) model was first introduced for use in oil-reservoir simulation \cite{warren1963behavior}, Figure~\ref{fig:dual-porosity-reservoir}, where the fracture system was the main flow path for the fluid (e.g., two phase oil-water mixture, one-phase oil-solvent mixture). Fluid exchange is permitted between the rock matrix and the fracture system, but not between the matrix blocks. In the DPSP model, the fracture system and the rock matrix, each has its own porosity, with values not differing from each other by a large factor.  On the contrary, the permeability of the fracture system is much larger than that in the rock matrix, and thus the system is considered as having only a single permeability. When the permeability of the fracture system and that of the rock matrix do not differ by a large factor, then both permeabilities are included in the more general \emph{dual-porosity dual-permeability} (DPDP) model \cite{datta2007streamline}.

%\noindent
%{\color{red} 2020.07.03. To explain: What is Majella limestone, why study it, why use it as motivation for dual porosity.}

Since 60\% of the world's oil reserve and 40\% of the world's gas reserve are held in carbonate rocks, there has been a clear interest in developing an understanding of the mechanical behavior of carbonate rocks such as limestones, having from lowest porosity (Solenhofen at 3\%) to high porosity (e.g., Majella at 30\%).  Chalk (Lixhe) is a carbonate rock with highest porosity at 42.8\%.
Carbonate rock reservoirs are also considered to store carbon dioxide and nuclear waste \vphantom{\cite{CROIZE2013181}}\cite{CROIZE2013181} \cite{ho2000dual}.

In oil-reservoir simulations in which the primary interest is the flow of oil, water, and solvent, the porosity (and pore size) within each domain (rock matrix or fracture system) is treated as constant and homogeneous \cite{datta2007streamline} \vphantom{\cite{lu2019new}}\cite{lu2019new}.\footnote{
	See, e.g., \cite{datta2007streamline}, p.~295, Chap.~9 on ``Advanced Topics: Fluid Flow in Fractured Reservoirs and Compositional Simulation''.
}
On the other hand, under mechanical stress, the pore size would change, cracks and other defects would close, leading to a change in the porosity in carbonate rocks.
Indeed,
``at small stresses, experimental mechanical deformation of carbonate rock is usually characterized by a non-linear stress-strain relationship, interpreted to be related to the closure of cracks, pores, and other defects. The non-linear stress-strain relationship can be related to the amount of cracks and various type of pores''  \vphantom{\cite{CROIZE2013181}}\cite{CROIZE2013181}, p.~202.  Once the pores and cracks are closed, the stress-strain relation becomes linear, at different stress stages, depending on the initial porosity and the geometry of the pore space \cite{CROIZE2013181}.
% 
%``In Tavel, Indiana, Majella, Solnhofen limestones the influence of the porosity type on the development of mechanical failure was studied. Pore collapse was found to initiate at larger pores (Zhu et al., 2010). For rocks containing both macro- and microporosity, macropores determine the localization of fractures.''
%\vphantom{\cite{CROIZE2013181}}\cite{CROIZE2013181}, p.~202.

%\noindent
%hello \vphantom{\cite{baud2009compaction}}\cite{baud2009compaction} \cite{vinciguerra2009rock}

Moreover, pores have different sizes, and can be classified into different pore sub-systems. For the Majella limestone in Figure~\ref{fig:Majella-pore-structure} with total porosity at 30\%, its pore space can be partitioned into two subsystems (and thus dual porosity), the macropores with macroporosity at 11.4\% and the micropores with microporosity at 19.6\%.  Thus the meaning of dual-porosity as used in \cite{Wang.2018:rd3109} is different from that in oil-reservoir simulation.
Also characteristic of porous rocks such as the Majella limestone is the non-linear stress-strain relation observed in experiments, Figure~\ref{fig:Majella-stress-strain}, due the changing size, and collapse, of the pores.

Likewise, the meaning of ``dual permeability'' is different in \cite{Wang.2018:rd3109} in the sense that ``one does not seek to obtain a single effective permeability for the entire pore space''.  Even though it was not explicitly spelled out,\footnote{
	At least at the beginning of Section 2 in \cite{Wang.2018:rd3109}.
} it appears that each of the two pore sub-systems would have its own permeability, and that fluid is allowed to exchange between the two pore sub-systems, similar to the fluid exchange between the rock matrix and the fracture system in the DPSP and DPDP models in oil-reservoir simulation \cite{datta2007streamline}.

%\noindent
%{\color{red} [NOTE: 2020.06.21. Done, footnote added, plus pointed out writing problem in \cite{Wang.2018:rd3109}.  2020.04.12. first-time learners would likely not understand terminologies like ``dual-porosity, dual-permeability poroplasticity''; perhaps a footnote to explain these terminologies would be welcome.]}

In the problem investigated in \cite{Wang.2018:rd3109}, the presence of localized discontinuities demands three scales---microscale ($\mu$), mesoscale (cm), macroscale (km)---to be considered in the modeling, see Figure~\ref{fig:Wang-RNN-FEM}.
Classical approaches to consistently integrate microstructural properties into macroscopic constitutive laws relied on hierarchical simulation models and homogenization methods (e.g., discrete element method (DEM)--FEM coupling, FEM$^2$). 
If more than two scales were to be considered, the computational complexity would become prohibitively, if not intractably, large.
%

%\noindent
%{\color{red}
%	NOTE: 2020.06.25.  i moved this alternative description of Figure~\ref{fig:Wang-RNN-FEM} to Section~\ref{sc:Wang-Sun-2018-2}. 2020.04.07.  i initially incorporated the text below into the caption of Figure~\ref{fig:Wang-RNN-FEM}, but Latex cannot process it, so i moved this text out of the caption, to use it somewhere else.
%}

\begin{figure}[h]
	\centering
	\includegraphics[scale=0.4]{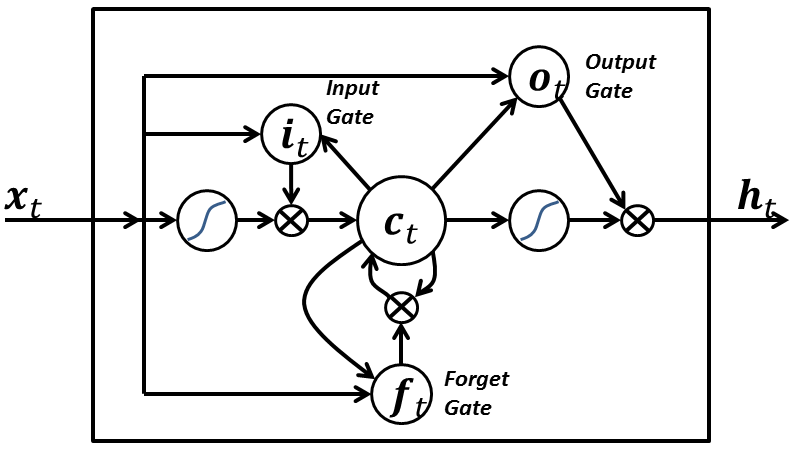}
	\caption[test]{
		%\sout{Schematic representation of an} 
		\emph{LSTM variant with ``peephole'' connections}, block diagram (Sections~\ref{sc:Wang-Sun-2018}, \ref{sc:LSTM}).\protect\footnotemark{} %\footnote{see, e.g.,~\cite{Gers.200}} 
		%{\color{red} [NOTE: 2020.05.11.  nice way to put footnote in a caption; i did not know this trick.  should we explain the concept of ``peephole'' in Section~\ref{sc:LSTM}, where ``peephole'' was not yet included ? ENDNOTE] }
		Unlike the original LSTM unit (see Section~\ref{sc:LSTM}), both the input gate and the forget gate in an LSTM unit with peephole connections receive the cell state as input.
		%  TODO: this is not a conventional LSTM cell but a so-called ``peephole'' LSTM. I suppose Wang and Sun were not aware of that, but I will double check.
		The above figure from Wikipedia,~\href{https://commons.wikimedia.org/w/index.php?title=File:Long_Short_Term_Memory.png&oldid=174444321}{\rm version 22:56, 4 October 2015}, is identical to Figure~10 in \cite{Wang.2018:rd3109}, whose authors erroneously used this figure without mentioning the source, but where the original LSTM unit \emph{without} ``peepholes'' was actually used, and with the detailed block diagram in Figure~\ref{fig:our-lstm_cell}, Section~\ref{sc:LSTM}.
		See also 
		Figure~\ref{fig:Olah-lstm_chain} and
%		Figure~\ref{fig:Mohan-RNN-LSTM}
		Figure~\ref{fig:Mohan-LSTM-BiLSTM} 
		for the original LSTM unit applied to fluid mechanics.
		%\sout{who used TensorFlow framework version 1 around 2017 or 2018 that might not have the LSTM variant with peephole connections implemented}. 
		(\href{https://creativecommons.org/licenses/by-sa/4.0/deed.en}{CC-BY-SA 4.0})
	}
	\label{fig:Wang-LSTM_cell}
\end{figure}
\footnotetext{ 
	The LSTM variant with peephole connections is not the original LSTM cell (Section~\ref{sc:LSTM}); see, e.g, \cite{Gers.2000}.  The equations describing the LSTM unit in \cite{Wang.2018:rd3109}, 
	%
	% CMES style, rewriting
%	who 
	whose authors
	never mentioned the word ``peephole'', correspond to the original LSTM without peepholes.  It was likely a mistake to use this figure in \cite{Wang.2018:rd3109}.  
}

Instead of coupling multiple simulation models online,
%
% CMES style, rewriting
% \cite{Wang.2018:rd3109} proposed to link 
two (adjacent) scales were linked by a neural network that was trained offline using data generated by simulations on the smaller scale \cite{Wang.2018:rd3109}. 
The trained network subsequently served as a surrogate model in online simulations on the larger scale.
With three scales being considered, two recurrent neural networks (RNNs) with Long Short-Term Memory (LSTM) units were employed consecutively:  
\begin{enumerate}
\item {\it Mesoscale RNN with LSTM units:} 
On the microscopic scale, a representative volume element (RVE) was an assembly of discrete-element particles, subjected to large variety of representative loading paths to generate training data for the supervised learning of the mesoscale RNN with LSTM units, a neural network that
%
% CMES style, rewriting 
%\cite{Wang.2018:rd3109} 
was referred to as ``Mesoscale data-driven constitutive model'' \cite{Wang.2018:rd3109} (Figure~\ref{fig:Wang-RNN-FEM}).
Homogenizing the results of DEM-flow model provided constitutive equations for the traction-separation law and the evolution of anisotropic permeabilities in damaged regions.
\item {\it Macroscale RNN with LSTM units:} 
The mesoscale RVE (middle row in Figure~\ref{fig:Wang-RNN-FEM}), in turn, was a finite-element model of a porous material with embedded strong discontinuities equivalent to the fracture system in oil-reservoir simulation in Figure~\ref{fig:dual-porosity-reservoir}. 
The host matrix of the RVE was represented by an isotropic linearly elastic solid. 
In localized fracture zones within, the traction-separation law and the hydraulic response were provided by the mesoscale RNN with LSTM units developed above.
Training data for the macroscale RNN with LSTM units---a network 
%
% CMES style, rewriting
%which \cite{Wang.2018:rd3109} 
referred to as ``Macroscale data-driven constitutive model'' \cite{Wang.2018:rd3109}---is generated by computing the (homogenized) response of the mesoscale RVE to various loadings.
In macroscopic simulations, the mesoscale RNN with LSTM units provided the constitutive response at a sealing fault that represented a strong discontinuity. 
\end{enumerate}

Path-dependence is a common characteristic feature of the constitutive models that are often realized as neural networks; see, e.g., \cite{Ghaboussi.1991:rd0001}.
For this reason, 
%
% CMES style, rewriting
%\cite{Wang.2018:rd3109} employed
it was decided to employ 
RNN with LSTM units, which could mimick internal variables and corresponding evolution equations that were intrinsic to path-dependent material behavior \cite{Wang.2018:rd3109}.
%The LSTM neural network is trained with time histories
These authors chose to use a neural network that had a depth of two hidden layers with 80 LSTM units per layer, and that had proved to be a good compromise of performance and training efforts. 
After each hidden layer, a dropout layer with a dropout rate 0.2 were introduced to reduce overfitting on noisy data, but yielded minor effects, as reported in \cite{Wang.2018:rd3109}. %{\color{red} [NOTE: 2020.07.25. we need to define ``dropout'' layer and ``dropout'' rate.]}
The output layer was a fully-connected layer with a logistic sigmoid as activation function.
%The sigmoid function is used as activation function of the output layer.
%
%\begin{figure}[H]
%  \centering
%  \includegraphics[scale=1]{./figures/Goodfellow-LSTM.pdf}
%  \caption{
%  {\color{blue}
%  ASK PERMISSION!
%  }
%  }
%  \label{fig:LSTM_cell}
%\end{figure}
%

%

\begin{figure}[h]
	\centering
	\includegraphics[width=0.25\linewidth]{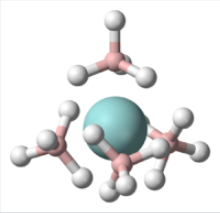}
	\includegraphics[width=0.65\linewidth]{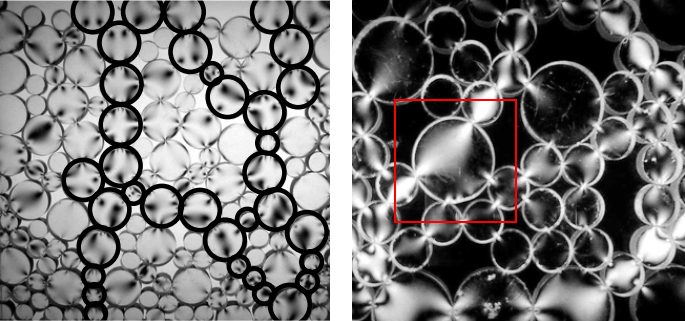}
	\caption{
		\emph{Coordination number $CN$} (Section~\ref{sc:Wang-Sun-2018}, \ref{sc:Wang-microstructure-data}). (a) Chemistry. Number of bonds to the central atom. Uranium borohydride U(BH$_4$)$_4$ has $CN = 12$ hydrogen bonds to uranium. 
		(b, c) Photoelastic discs showing number of contact points (coordination number) on a particle.  
		%Skeletal force distribution. 
		(b) Random packing and force chains, different force directions along principal chains and in secondary particles. 
		(c) Arches around large pores, precarious stability around pores. The coordination number for the large disc (particle) in red square is 5, but only 4 of those had non-zero contact forces based on the bright areas showing stress action. 
		Figures~(b, c) also provide a visualization of the ``flow'' of the contact normals, and thus the fabric tensor \cite{santamarina2003soil}.
		See also Figure~\ref{fig:Wang-prediction-LSTM-microstructure}.
%		\cite{santamarina2003soil}. 
		\footnotesize (Figure reproduced with permission of the author.)
	}
	\label{fig:Wang-coordination-number}
	
\end{figure}

\begin{figure}[h]
	\centering
	\includegraphics[width=0.9\linewidth]{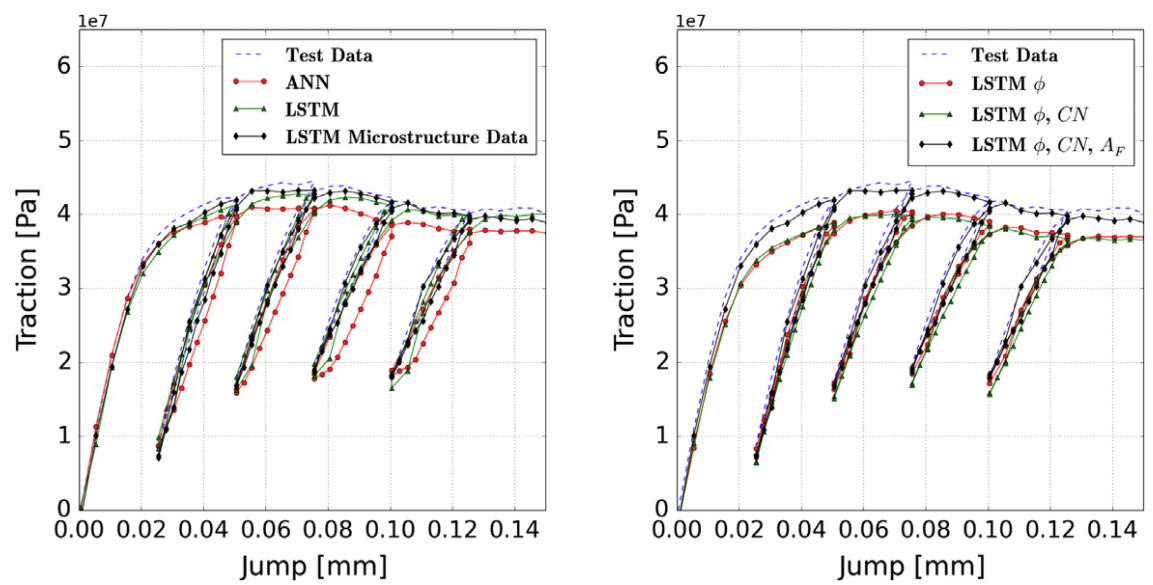}
	\caption{
		\emph{Network with LSTM and microstructure data} (porosity $\phi$, coordination number $CN = N_c$, Figure~\ref{fig:Wang-coordination-number}, fabric tensor $\boldsymbol{F} = \boldsymbol{A}_F \cdot \boldsymbol{A}_F$, Eq.~\eqref{eq:Wang-fabric-tensor}) (Section~\ref{sc:Wang-Sun-2018}, \ref{sc:Wang-microstructure-data}). 
		Simple shear test using Discrete Element Method to provide network training data under loading-unloading conditions.  
		ANN = Artificial Neural Network with no LSTM units.
		While network with LSTM units and ($\phi$, $CN$, $\boldsymbol{A}_F$) improved the predicted traction, compared to network with LSTM units and only ($\phi$) or ($\phi$, $CN$); the latter two networks produced predicted traction that was worse compared to network with LSTM alone, indicating the important role of the fabric tensor, which contained directional data that were absent in scalar fields like ($\phi$, $CN$)
		\cite{Wang.2018:rd3109}. 
		\footnotesize (Figure reproduced with permission of the author.)
	}
	\label{fig:Wang-prediction-LSTM-microstructure}
	
\end{figure}

%\noindent
%{\color{red} [NOTE:  2020.07.25.  See my notes in  \href{https://docs.google.com/document/d/1yHJNp_vCEZSLUz5rn5ls02yGXd6JRTQfiM6aEYsQ5RE/edit}{Deep-learning computational mechanics B - doc}.  it seems that TensorFlow version 1 did have LSTM with peepholes, but not Keras and TensorFlow.  2020.05.12. TensorFlow version 1, released in 20XX, did not have the LSTM variant with peephole connections implemented. ENDNOTE] }

%\noindent
%{\color{red} HERE, 2020.07.25}

An important observation is that including micro-structural data---the porosity $\phi$, the coordination number $CN$ (number of contact points, Figure~\ref{fig:Wang-coordination-number}), the fabric tensor (defined based on the normals at the contact points, Eq.~\eqref{eq:Wang-fabric-tensor} in Section~\ref{sc:Wang-Sun-2018-2}; Figure~\ref{fig:Wang-coordination-number} provides a visualization)---as network inputs significantly improved the prediction capability of the neural network.
Such improvement is not surprising since
soil fabric---described by scalars (porosity, coordination number, particle size) and vectors (fabric tensors, particle orientation, branch vectors)---exerts great influence on soil behavior \vphantom{\cite{alam2018study}}\cite{alam2018study}.  
Coordination number\footnote{
	The \href{https://en.wikipedia.org/w/index.php?title=Coordination_number&oldid=970032729}{coordination number} (Wikipedia version 20:43, 28 July 2020)  is a concept originated from chemistry, signifying the number of bonds from the surrounding atoms to a central atom.  In Figure~\ref{fig:Wang-coordination-number} (a), the \href{https://en.wikipedia.org/w/index.php?title=Uranium_borohydride&oldid=887379908}{uranium borohydride} U(BH$_4$)$_4$ complex (Wikipedia version 08:38, 12 March 2019) has 12 hydrogen atoms bonded to the central uranium atom.
} has been used to predict soil particle breakage \vphantom{\cite{karatza2019effect}}\cite{karatza2019effect}, morphology and crushability \cite{alam2018study}, and in a study of internally-unstable soil involving a mixture of coarse and fine particles \vphantom{\cite{shire2014fabric}}\cite{shire2014fabric}.
Fabric tensors, with theoretical foundation developed in \cite{kanatani1984distribution}, provide a mean to represent directional data such as normals at contact points, even though other types of directional data have been proposed to develop fabric tensors \cite{fu2015relationship}.
To model anisotropic behavior of granular materials, 
%
% CMES style, rewriting
%\vphantom{\cite{hu2019constitutive}}\cite{hu2019constitutive} incorporated 
contact-normal fabric tensor was incorporated in an isotropic constitutive law. 

Figure~\ref{fig:Wang-prediction-LSTM-microstructure} illustrates the importance of incorporating microstructure data, particularly the fabric tensor, in network training to improve prediction accuracy.

%

%\begin{figure}[H]
%  \centering
%  \includegraphics[scale=1]{./figures/Wang-results_mesoscale_rnn.pdf}
%  \caption{
%  Comparison of normal (a) and tangential (b) tractions obtained from the meso-scale RVE under displacement loading where the microscopic scale is represented by a DEM model (solid blue line) and the meso-scale LSTM network (dotted red line), respectively.
%  Numbers indicate the sequence of loading--unloading steps.
%  {\color{blue}
%  ASK PERMISSION!
%  }
%  }
%  \label{fig:Wang-results_mesoscale_rnn}
%\end{figure}
%
%\begin{figure}[H]
%  \centering
%  \includegraphics[scale=1]{./figures/Wang-results_macroscale_rnn.pdf}
%  \caption{
%  Comparison of the normal traction (Tn) due to prescribed displacements (Un).
%  Blue lines indicate training and test data from selected training (TR1--TR3) and test sets (TE1--TE3) obtained from the meso-scale FEM--LSTM model, where numbers indicate the sequence of loading--unloading steps.
%  Red lines represent the corresponding predictions of the trained macro-scale RNN.
%  The mean squared error (MSE) is used as loss function to quantify the performance of the RNN.
%  {\color{blue}
%  ASK PERMISSION!
%  }
%  }
%  \label{fig:Wang-results_macroscale_rnn}
%\end{figure}

\vspace{2ex}
{\bf Deep-learning concepts to explain and explore:} (continued from \hyperref[para:concepts-1]{above} in Section~\ref{sc:Oishi-summary})
\label{para:concepts-2}
\begin{enumerate}[start=12]
	
	\item
	Recurrent neural network (RNN), 
	% 2022.06.15
	% no need to cite Section 7, but only 7.1
%	Sections~\ref{sc:recurrent}, 
	Section~\ref{sc:RNN}
	
	\item 
	Long Short-Term Memory (LSTM), Section~\ref{sc:LSTM}
	%\item Dropout layer
	
%	{\color{red} NOTE: 2022.06.14 - we need to mention ``attention'' and ``transformer unit'' (if we write about this topic) in relation to LSTM, since we did further above toward the end of Section~\ref{sc:Oishi-summary}.}
% AG: 2022.10.03
	\item 
	Attention and Transformer, Section~\ref{sc:Transformer}
	
	\item
	Dropout layer and dropout rate,\footnote{
		Briefly, dropout means to drop or to remove non-output units (neurons) from a base network, thus creating an ensemble of sub-networks (or models) to be trained for each example, and can also be considered as a way to add noise to inputs, particularly of hidden layers, to train the base network, thus making it more robust, since neural networks were known to be not robust to noise.  Adding noise is also equivalent to increasing the size of the dataset for training, \cite{Goodfellow.2016}, p.~233, Section 7.4 on ``Dataset augmentation''. 
	} which had minor effects in the particular work repoorted in \cite{Wang.2018:rd3109}, and thus will not be covered here.  See \cite{Goodfellow.2016}, p.~251, Section 7.12.

\end{enumerate}
Details of the formulation in \cite{Wang.2018:rd3109} are discussed in Section~\ref{sc:Wang-Sun-2018-2}.

\subsubsection{Fluid mechanics, reduced-order model for turbulence}
\label{sc:Mohan-2018}

The accurate simulation of turbulence in fluid flows ranks among the most demanding tasks in computational mechanics. 
Owing to both the spatial and the temporal resolution, transient analysis of turbulence by means of high-fidelity methods such as Large Eddy Simulation (LES) or direct numerical simulation (DNS) involves millions of unknowns even for simple domains.

\begin{figure}[h]
	\centering
	\includegraphics[width=0.9\linewidth]{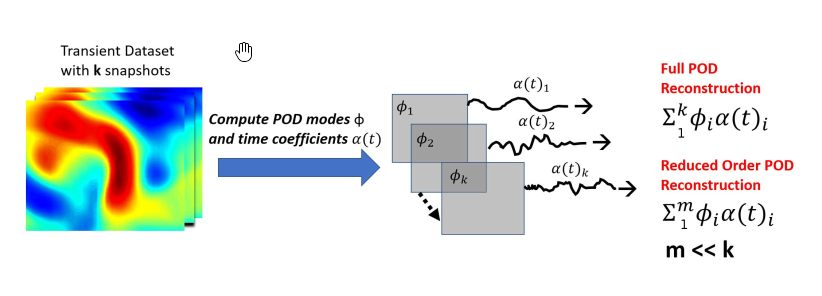}
	\caption{ 
		\emph{Reduced-order POD basis} (Sections~\ref{sc:Mohan-2018}, \ref{sc:Mohan-POD}).
		For each dataset (also Figure~\ref{fig:Mohan-2D-datasets}), which contained $k$ snapshots, the full POD reconstruction of the flow-field dynamical quantity $u (\bx, t)$, where $\bx$ is a point in the 3-D flow field, consists of all $k$ basis functions $\phi_i (\bx)$, with $i = 1, \ldots, k$, using Eq.~\eqref{eq:Mohan-POD-combination}; see also Eq.~(\ref{eq:Mohan-POD}). 
		%such that $u (\bx, t) \approx \sum_{i=1}^{i=k}\phi_i (\bx) \alpha_i (t)$.
		Typically, $k$ is large; a reduced-order POD basis consists of selecting $m \ll k$ basis functions for the reconstruction, with the smallest error possible. 
		See Figure~\ref{fig:Mohan-LSTM-BiLSTM-ROM} for the use of deep-learning networks to predict $\alpha_i(t+t^\prime)$, with $t^\prime > 0$, given $\alpha_i(t)$
		\cite{Mohan.2018}. 
		\footnotesize (Figure reproduced with permission of the author.)
	}
	\label{fig:Mohan-reduced-order-POD-basis}
\end{figure}

%\begin{figure}[h]
%	\centering
%	\includegraphics[width=0.45\linewidth]{figures/Mohan-2018-RNN-LSTM-units}
%	\includegraphics[width=0.45\linewidth]{figures/Mohan-2018-LSTM-unit}
%	\caption{ 
%		\emph{RNN with traditional LSTM units.} 
%		\emph{Left:} Recurrent neural network (RNN) as a chain of Long Short-Term Memory (LSTM) units (cells; see also Figure~\ref{fig:Olah-lstm_chain}). 
%		\emph{Right:} Details of an LSTM unit (See also Figure~\ref{fig:Wang-LSTM_cell} in Section~\ref{sc:Wang-Sun-2018} for LSTM with peepholes and Figure~\ref{fig:our-lstm_cell} in Section~\ref{sc:LSTM} for more details on the \emph{folded} original LSTM unit).
%		\cite{Mohan.2018}. 
%		\footnotesize (Figure reproduced with permission of the author.)
%	}
%	\label{fig:Mohan-RNN-LSTM}
%\end{figure}

%{\color{red} HERE 2020.09.13}

%\begin{figure}[h]
%	\centering
%	\includegraphics[scale=1]{./figures/Wang-results_macroscale_differential_stress.pdf}
%	\caption{
%		{\color{red} 2020.05.11. this figure is not referred to anywhere in the text in this latex file.  also we already have the permission.}
%		{\color{blue}
%			ASK PERMISSION!
%		}
%	}
%	\label{fig:Wang-results_macroscale_differential_stress}
%\end{figure}

To simulate complex geometries over larger time periods, reduced-order models (ROMs) that can capture the key features of turbulent flows within a low-dimensional approximation space need to be resorted to. 
Proper Orthogonal Decomposition (POD) is a common data-driven approach to construct an orthogonal basis $\{ \phi_1 (\bx) , \phi_2 (\bx), \ldots \phi_\infty (\bx) \}$ from high-resolution data obtained from high-fidelity models or measurements, in which $\bx$ is a point in a 3-D fluid domain $\domain$; see Section~\ref{sc:Mohan-POD}.   A flow dynamic quantity $u (\bx , t)$, such as a component of the flow velocity field, can be projected on the POD basis by separation of variables as (Figure~\ref{fig:Mohan-reduced-order-POD-basis})
\begin{align}
	u (\bx, t) 
	= 
	\sum_{i=1}^{i=\infty} \phi_i (\bx) \alpha_i (t)
	\approx 
	\sum_{i=1}^{i=k} \phi_i (\bx) \alpha_i (t)
	\ , \text{ with } k < \infty
	\ ,
	\label{eq:Mohan-POD-combination}
\end{align}
where 
$k$ is a finite number, which could be large, and
$\alpha_i(t)$ a time-dependent coefficient for $\phi_i (\bx)$.  The computation would be more efficient if a much smaller subset with, say, $m \ll k$, POD basis functions, 
\begin{align}
	u (\bx, t) 
	\approx 
	\sum_{j=1}^{j=m} \phi_{i_j} (\bx) \alpha_{i_j} (t)
	\ , 
	\text{ with } m \ll k
	\text{ and } i_j \in \{ 1 , \ldots , k\}
	\ ,
	\label{eq:Mohan-reduced-POD-basis}
\end{align}
where $\{ i_j , j=1, \ldots, m \}$ is a subset of indices in the set $\{1 , \ldots, k\}$, and such that the approximation in Eq.~\eqref{eq:Mohan-reduced-POD-basis} is
with minimum error compared to the approximation in Eq.~(\ref{eq:Mohan-POD-combination})$_2$ using the much larger set of $k$ POD basis functions. 

In a Galerkin-Project (GP) approach to reduced-order model, a small subset of dominant modes form a basis onto which high-dimensional differential equations are projected to obtain a set of lower-dimensional differential equations for cost-efficient computational analysis.
\begin{figure}[h]
	\centering
	\includegraphics[scale=1]{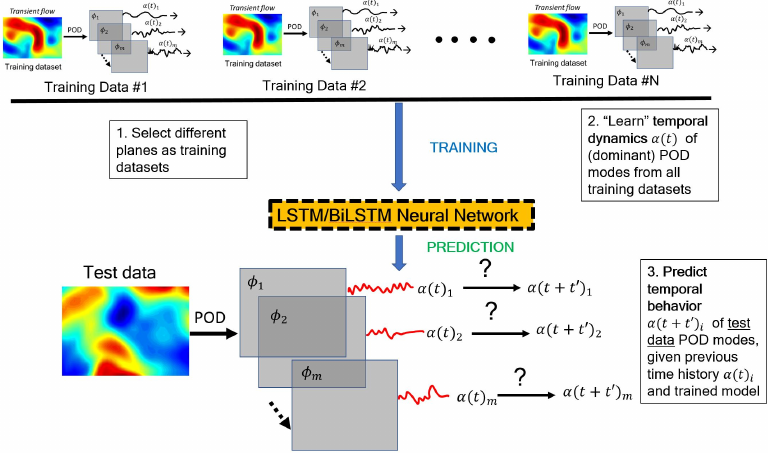}
	\caption{
		\emph{Deep-learning LSTM/BiLSTM Reduced Order Model} (Sections~\ref{sc:Mohan-2018}, \ref{sc:Mohan-reduced-POD}).
		See Figure~\ref{fig:Mohan-reduced-order-POD-basis} for POD reduced-order basis.
		The time-dependent coefficients $\alpha_i(t)$ of the dominant POD modes $\phi_i$, with $i = 1, \ldots, m$, from each training dataset were used to train LSTM (or BiLSTM, Figure~\ref{fig:Mohan-LSTM-BiLSTM}) neural networks to predict $\alpha_i (t + t^\prime)$, with $t^\prime > 0$, given $\alpha_i(t)$ of the test datasets.  
%		See Figure~\ref{fig:Mohan-LSTM-BiLSTM}.
		\cite{Mohan.2018}. 
		\footnotesize (Figure reproduced with permission of the author.)
		%		{\color{blue} I'm not sure whether we want to include that or not... let's decide later on.}
	}
\label{fig:Mohan-LSTM-BiLSTM-ROM}
\end{figure}

Instead of using GP, 
%
% CMES style, rewriting
%\cite{Mohan.2018} used 
RNNs (Recurrent Neural Networks) were used in \cite{Mohan.2018} to predict the evolution of fluid flows, specifically the coefficients of the dominant POD modes, rather than solving differential equations.
For this purpose, their LSTM-ROM (Long Short-Term Memory - Reduced Order Model) approach combined concepts of ROM based on POD with deep-learning neural networks using either the original LSTM units, 
%Figure~\ref{fig:Mohan-RNN-LSTM} 
%Figure~\ref{fig:our-lstm_cell} 
Figure~\ref{fig:Mohan-LSTM-BiLSTM} (left)
\cite{Hochreiter.1997:rd0001}, or the bidirectional LSTM (BiLSTM), 
Figure~\ref{fig:Mohan-LSTM-BiLSTM} (right)
{\cite{Graves2005}}, the internal states of which were well-suited for the modeling of dynamical systems. 

%{\color{red} HERE 2020.09.13}

%It is understood that training data plays a crucial role in training neural networks.
To obtain training/testing data, which were crucial to train/test neural networks,
%
% CMES style, rewriting 
%\cite{Mohan.2018} used 
the data from transient 3-D Direct Navier-Stokes (DNS) simulations of two physical problems, as provided by the Johns Hopkins turbulence database \vphantom{\cite{graham2016web}}\cite{graham2016web} were used \cite{Mohan.2018}:
%\footnote{
%	\vphantom{\cite{Graham2016}}\cite{Graham2016}
%} 
(1) The \emph{Forced Isotropic Turbulence} (ISO) and (2) The \emph{Magnetohydrodynamic Turbulence} (MHD).

% from Alex, missing biblio details; use mine
%, \vphantom{\cite{Graham2016}}\cite{Graham2016}

To generate training data for LSTM/BiLSTM networks, the 3-D turbulent fluid flow domain of each physical problem was decomposed into five equidistant 2-D planes (slices), with one additional equidistant 2-D plane served to generate testing data (Section~\ref{sc:Mohan-2018-2}, Figure~\ref{fig:Mohan-2D-datasets}, Remark~\ref{rm:Mohan-reduced-order-POD}). 
%
%{\color{blue} [NOTE: 2012.12.11. what is meant by ``U-velocity field''? END NOTE] }
%
For the same subregion in each of those 2-D planes, POD was applied on the $k$ snapshots of the velocity field ($k = 5,023$ for ISO, $k = 1,024$ for MHD, Section~\ref{sc:Mohan-POD}), and out of $k$ POD modes $\{\phi_i(\bx), \ i=1, \ldots, k\}$, the five ($m=5 \ll k$) most dominant POD modes $\{\phi_i(\bx), \ i=1, \ldots, m\}$ representative of the flow dynamics (Figure~\ref{fig:Mohan-reduced-order-POD-basis}) were retained to form a reduced-order basis onto which the velocity field was projected.
%The reduced basis formed by the POD modes is used to project the velocity field of each time-step of each 2-D plane onto a corresponding reduced, five-dimensional sub-space. 
%The projected solution (``temporal coefficients'') represent evolutions of (modal) participation factors, which, in turn, are decomposed into signals with a length of 20 time-steps using a moving window. 
%To train the neural network, the first ten time-steps of each sample are the input data. 
%The remaining ten time-steps, which are referred to as predictive horizon, represent the output signal, i.e., the signal to be predicted by the neural network for a given input.
%The projected solution (``temporal coefficients'') represent evolutions of (modal) participation factors, which, in turn, are decomposed into signals with a specified length using a moving window. 
The coefficient $\alpha_i(t)$ of the POD mode $\phi_i(\bx)$ represented the evolution of the participation of that mode in the velocity field, and was decomposed into thousands of small samples using a moving window. 
The first half of each sample was used as input signal to an LSTM network, whereas the second half of the sample was used as output signal for supervised training of the network.
%The second half, whose length represents the so-called predictive horizon, represents the output signal, i.e., the signal to be predicted by the neural network for a given input signal.
%
%
% CMES style, rewriting
%\cite{Mohan.2018} proposed 
Two different methods were proposed \cite{Mohan.2018}:
%{\color{red} HERE 2020.10.25}

% \vspace{0.5cm}
\begin{enumerate}
\item \emph{Multiple-network method:} Use a RNN for each coefficient of the dominant POD modes 
%to separately predict 
%trajectories of participation vectors 
%the coefficient $\alpha_i (t+t^\prime)$, for $t^\prime > 0$ and $i = 1,\ldots,5$, given $\alpha_i(t)$, see Figure~\ref{fig:Mohan-nn-separate}. 
%Hyperparameters (layer width, learning rate, batch size) are tuned for the most dominant POD mode and reused for training the other neural networks. 

\item \emph{Single-network method:} Use a single RNN for all coefficients of the dominant POD modes
%to predict all coefficients $\{\alpha_i (t+t^\prime) \ , t^\prime > 0 \ , i = 1,\ldots,5\}$ at once, given $\{\alpha_i (t) \ , i = 1,\ldots,5\}$, see Figure~\ref{fig:Mohan-nn-unified}.
\end{enumerate}
%
%The idea of the unified approach is to better capture inter-modal interactions that are relevant to describe, e.g., energy transfer from larger to smaller scales.
%Vortices that spread over multiple POD modes also support the unified approach, which does not artificially constrain flow features to separate POD modes.
For both methods, variants with the original LSTM units or the BiLSTM units were implemented.
Each of the employed RNN had a single hidden layer.
%
%\begin{itemize}
%\item \sout{overall topic: turbulence in fluids}
%\item \sout{CFD: large eddy simulation (LES), direct numerical simulation (DNS) $\rightarrow$ very expensive} (huge data)
%\item reduced order models (ROM)
%	\begin{itemize}
%	\item \sout{model key dynamics}/coherent features (?)
%	\item efficient means for data compression for LES/DNS datasets
%	\item \sout{idea: extract key features in flow-field (high-dimensional $\rightarrow$ low-dimensional)}
%	\item \sout{common: POD + Galerkin projection (GP)}
%	\item non-Galkerin: e.g. deep learning
%	\end{itemize}
%\item data: 2 datasets from John Hopkins turbulence database (3-D DNS Navier-Stokes simulations); single Reynolds number
%	\begin{itemize}
%	\item 2-D planes extracted from 3-D (five equidistant planes for training, one for testing)
%	\item POD: five modes
%	\item moving window (split time histories): \SI{0.1}{\second} input, \SI{0.1}{\second} output
%	\end{itemize}
%\item approach + methodology:
%	\begin{itemize}
%	\item LSTM + bidirectional LSTM (BiLSTM)
%	\end{itemize}
%\end{itemize}
%

%{\color{red} 
%	[NOTE: 2020.03.13.  you may not want to use the adjective ``Exemplary'' below, since ``exemplary'' means ``excellent''; it is sufficient to write ``Examples of the prediction capabilities...'']
%}

\begin{figure}[h]
	\centering
	\begin{subfigure}[b]{0.48\textwidth}
		\centering
		\includegraphics[clip=true, trim=0 5pt 0 3.5pt, width=\linewidth]{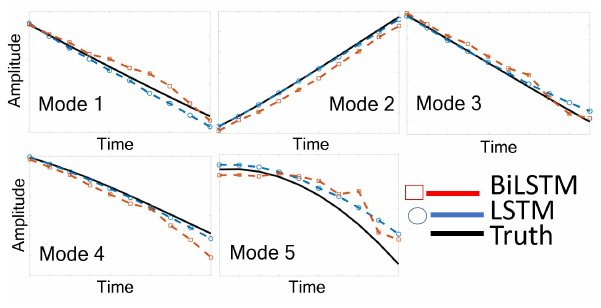}
		\caption{Isotropic turbulence (ISO)
		}\label{fig:Mohan-result_isotropic}
	\end{subfigure}
	\ 
	\begin{subfigure}[b]{0.48\textwidth}
		\centering
		\includegraphics[clip=true, trim=4pt 4pt 2pt 6pt, width=\linewidth]{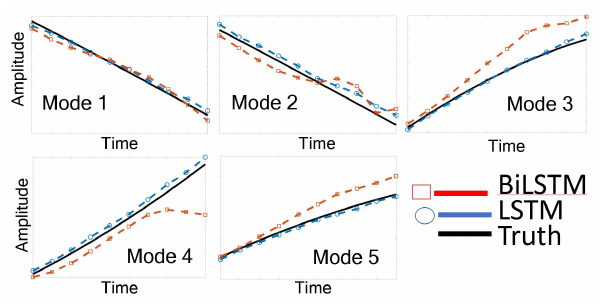}
		\caption{Magnetohydrodynamics (MHD)
		}\label{fig:Mohan-result_mhd}
	\end{subfigure}
	\caption{
		\emph{Prediction results and errors for LSTM/BiLSTM networks} (Sections~\ref{sc:Mohan-2018}, \ref{sc:Mohan-results}). 
		Coefficients $\alpha_i(t)$, for $i=1,\ldots,5$, of dominant POD modes.
		LSTM had smaller errors compared to BiLSTM for both physical problems (ISO and MHD).
	}
	\label{fig:Mohan-results}
\end{figure}

Demonstrative results for the prediction capabilities of both the original LSTM and the BiLSTM networks are illustrated in Figure~\ref{fig:Mohan-results}.
Contrary to the authors' expectation, networks with the original LSTM units performed better than those using BiLSTM units in both physical problems of isotropic turbulence (ISO) (Figure~\ref{fig:Mohan-result_isotropic}) and magnetohydrodyanmics (MHD) (Figure~\ref{fig:Mohan-result_mhd}) \cite{Mohan.2018}.

Details of the formulation in \cite{Mohan.2018} are discussed in Section~\ref{sc:Mohan-2018-2}.

% 2019.12.07, commented out this subsection
% \subsubsection{Other applications}
% inverse problems
% 2018 Nature article
% heat transfer
% structural health monitoring
% electromagnetics
% quantum mechanics
%\cite{Dunjko.2018:rd7724}

% big picture, solving PDE and brain modeling

%\section{Big picture, solving PDE and brain neural networks}
%\section{Engineering continuum and the brain, modeling}
%\section{Computational mechanics, computational neuroscience, deep learning}
\section{Computational mechanics, neuroscience, deep learning}
\label{sc:comparison-three-fields}

Table~\ref{tb:compare-mecha-neuro-DL} below presents a rough comparison that shows the parallelism in the modeling steps in three fields: Computational mechanics, neuroscience, and deep learning, which heavily indebted to neuroscience until it reached more mature state, and then took on its own development.

We assume that readers are familiar with the concepts listed in the second column on  ``Computational mechanics'', and briefly explain some key concepts in the third column on ``Neuroscience'' to connect to the fourth column ``Deep learning'', which is explained in detail in subsequent sections.

See Section~\ref{sc:linear-combo-history} for more details on the theoretical foundation based on Volterra series for the spatial and temporal combinations of inputs, weights, and biases, widely used in artificial neural networks or multilayer perceptrons.

\begin{figure}[h]
	\centering
	\includegraphics[width=0.8\linewidth]{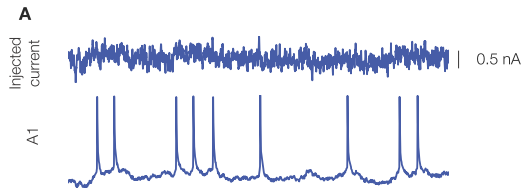}
	\caption{
		\emph{Biological neuron response to stimulus, experimental result} (Section~\ref{sc:comparison-three-fields}). An oscillating current was injected into the neuron (top), and neuron spiking response was recorded (below) 
		\cite{Cyrille.2011}, 
		with permission from 
		\href{https://doi.org/10.3389/fnins.2011.00009}{the authors and Frontiers Media SA}.  A spike, also called an action potential, is an electrical potential pulse across the cell membrane that lasts about 100 milivolts over 1 miliseconds.  ``Neurons represent and transmit information by firing sequences of spikes in various temporal patterns'' \cite{Dayan.2001}, pp.~3-4.
	}
	\label{fig:spike-train}
\end{figure}
Neuron spiking response such as shown in Figure~\ref{fig:spike-train} can be modelled accurately using a model such as ``Integrate-and-Fire''. The firing-rate response $r(\cdot)$ of a biological neuron to a stimulus $s(\cdot)$ is described by a convolution integral\footnote{
	From here on, if Eq.~(\ref{eq:firing-rate}) is found a bit abstract at first reading, first-time learners could skip the remaining of this short Section~\ref{sc:comparison-three-fields} to begin reading Section~\ref{sc:feedforward}, and come back later after reading through subsequent sections, particularly Section~\ref{sc:linear-combo-history}, to have an overview of the connection among seemingly separate topics.
}
\begin{align}
	r(t) = r_0 + \int\limits_{\tau=-\infty}^{\tau=t} \K \, (t - \tau) \, s(\tau) \, d \tau
	\ ,
	\label{eq:firing-rate} 
\end{align}
where $r_0$ is the background firing rate at zero stimulus, $\K (\cdot)$ is the synaptic
kernel, and  $s(\cdot)$ the stimulus; see, e.g., \cite{Dayan.2001}, p.~46, Eq.~(2.1).\footnote{
	Eq.~(\ref{eq:firing-rate}) is a reformulated version of Eq.~(2.1) in
	\cite{Dayan.2001}, p.~46, and is similar to Eqs.~(7.1)-(7.2) in 
	\cite{Dayan.2001}, p.~233, Chapter ``7 Network Models''.
}.  The stimulus $s(\cdot)$ in Eq.~(\ref{eq:firing-rate}) is a train (sequence in time) of spikes described by
\begin{align}
	s(\tau) = \sum_{i} \delta(\tau - t_i)
	\label{eq:spike-train}
\end{align}
where $\delta(\cdot)$ is the Dirac delta.  Eq.~(\ref{eq:firing-rate}) then describes the firing rate $r(t)$ at time $t$ as the collective memory effect of all spikes, going from the current time $\tau = t$ back far in the past with $\tau = -\infty$, with the weight for the spike $s(\tau)$ at time $\tau$ provided by the value of the synaptic kernel $\K(t - \tau)$ at the same time $\tau$.

It will be seen in Section~\ref{sc:dynamic-volterra-series} on ``Dynamics, time dependence, Volterra series'' that the convolution integral in Eq.~(\ref{eq:firing-rate}) corresponds to the linear part of the Volterra series of nonlinear response of a biological neuron in terms of the stimulus, Eq.~(\ref{eq:linear-volterra-series}), which in turn provides the theoretical foundation for taking the linear combination of inputs, weights, and biases for an artificial neuron in a multilayer neural networks, as represented by Eq.~(\ref{eq:linearComboInputsBias}).

The Integrate-and-Fire model for biological neuron provides a motivation for the use of the rectified linear units (ReLU) as activation function in multilayer neural networks (or perceptrons); see Figure~\ref{fig:neuron-firing}.

Eq.~(\ref{eq:firing-rate}) is also related to the exponential smoothing technique used in forecasting and applied to stochastic optimization methods to train multilayer neural networks; see Section~\ref{sc:exponential-smoothing} on ``Forecasting time series, exponential smoothing''. 
%in Section~\ref{sc:adaptive-learning-rate-algos} on adaptive learning-rate algorithms.  

\begin{table}
	\centering
	\caption{\emph{Top-down (rough) comparison of modeling steps in three fields} (Section~\ref{sc:comparison-three-fields}): Computational mechanics, neuroscience, and deep learning}
	\begin{tabularx}{1.0\textwidth}{c *{3}{Y}}
		\toprule[3pt]
		Study object		
		& \multicolumn{1}{c}{Engineering continuum}  
		& \multicolumn{1}{c}{The Brain} 
		%& \multicolumn{1}{c}{Play Go game}
		& \multicolumn{1}{c}{Image recognition} 
		\\
		\cmidrule(lr){1-1} \cmidrule(l){2-2} \cmidrule(l){3-3} \cmidrule(l){4-4}
		Field		
		& \multicolumn{1}{c}{Computational mechanics}  
		& \multicolumn{1}{c}{Computational neuroscience} 
		& \multicolumn{1}{c}{Deep learning} 
		\\
		\cmidrule(lr){1-1} \cmidrule(l){2-2} \cmidrule(l){3-3} \cmidrule(l){4-4}
		Modeling 
		& Partial Differential Equations 
		& Biological neural networks 
		& Artificial neural networks
		\\
		\\
		Inputs
		& Forces (solids), velocities (fluids)
		& Firing rate as stimulus
		& An image to classify
		\\
		\midrule[2pt]
		\vspace{3mm}
		1  
		& Weak form, finite-element mesh, order of interpolation 
		& Network architectures,{\newline} two layers (input, output), several neurons per layer  
		& Network architectures,{\newline} many layers (input, hidden, output), very high number of neurons and parameters
		\\
		\vspace{3mm}
		2  
		& Elements 
		& Neurons, dendrites, synapses, axons 
		& Processing units{\newline}  (neurons, perceptrons)		
		\\
		\vspace{3mm}
		3  
		& Nonlinear force-displacement and stress-strain ($\stress$-$\epsilon$) relations 
		& Firing model, spiking model, firing rate vs input current (FI) relation, continuous stimulus and response, Volterra series, kernels of increasing orders 
		& ---
		\\
		\vspace{3mm}
		4
		& Linearized force-displacement and stress-strain relations (Hooke's law)
		& Linear term in Volterra series, synaptic kernel $\K_1 (\tau)$ of order 1,  continuous temporal weight
		& Many hidden layers (discrete weights and biases)
		\\
		\vspace{3mm}
		5
		& ---
		& Linear combination of inputs, with input weights $\boldsymbol{w}$
		& Linear combination of inputs plus biases, input weights $\boldsymbol{w}$
		\\
		\vspace{3mm}
		6
		& ---
		& Static nonlinearity
		& Activation function
		\\
		\bottomrule[2pt]		
		\\
		Outputs
		& Displacements (solids), velocities (fluids)
		& Firing rate as response
		& Image classified (car, frog, human)
		\\
	\end{tabularx}
	\label{tb:compare-mecha-neuro-DL}
\end{table}

% feedforward network

\section{Statics, feedforward networks}
\label{sc:feedforward}

We examine in detail the forward propagation in feedforward networks, in which the function mappings flow\footnote{
	There is no physical flow here, only function mappings.
} only one forward direction, from input to output.

% \section{Data-driven computing, applications}
\subsection{Two concept presentations}

There are two ways to present the concept of deep-learning neural networks: The top-down approach versus the bottom-up approach.

\subsubsection{Top-down approach}
\label{sc:top-down}
The {\em top-down} approach starts by giving up-front the mathematical big picture of what a neural network is, with the big-picture (high level) graphical representation, then gradually goes down to the detailed specifics of a processing unit (often referred to as an {\em artificial neuron}) and its low-level graphical representation.  A definite advantage of this top-down approach is that readers new to the field immediately have the big picture in mind, before going down to the nitty-gritty details, and thus tend not to get lost.
An excellent reference for the top-down approach is \cite{Goodfellow.2016}, and there are not many such references.

Specifically, for a multilayer feedforward network, by top-down, we mean starting from a general description in Eq.~(\ref{eq:compositions}) and going down to the detailed construct of a neuron through a weighted sum with bias in Eq.~(\ref{eq:linearComboInputsBias}) and then a nonlinear activation function in Eq.~(\ref{eq:activationFunction}).

In terms of block diagrams, we begin our \emph{top-down} descent from the big picture of the overall multilayer neural network with $L$ layers in Figure~\ref{fig:network3b}, through Figure~\ref{fig:network2} for a typical layer $(\ell)$ and Figure~\ref{fig:neuron4} for the lower-level details of layer $(\ell)$, then down to the most basic level, a neuron in Figure~\ref{fig:neuron5} as one row in layer $(\ell)$ in Figure~\ref{fig:neuron4}, the equivalent figure of Figure~\ref{fig:Oishi-neuron}, which in turn was the starting point in \cite{Oishi.2017:rd9648}.

\subsubsection{Bottom-up approach}
\label{sc:bottom-up}

The {\em bottom-up} approach typically starts with a biological neuron 
(see Figure~\ref{fig:bio-neuron} in Section~\ref{sc:inspired-from-biology} below), then introduces an {\em artificial neuron} that looks similar to the biological neuron (compare Figure~\ref{fig:Oishi-neuron} to Figure~\ref{fig:bio-neuron}), with multiple inputs and a single output, which becomes an input to each of a multitude of other artificial neurons; see, e.g., \cite{Ghaboussi.1991:rd0001} \cite{Nielsen.2015} \cite{Oishi.2017:rd9648} \cite{Sze.2017:rd0001}.\footnote{
	Figure2 in\cite{Sze.2017:rd0001} is essentially the same as Figure~\ref{fig:Oishi-neuron}.
}
Even though Figure~\ref{fig:Oishi-network}, which preceded Figure~\ref{fig:Oishi-neuron} in \cite{Oishi.2017:rd9648}, showed a network, but the information content is not the same as Figure~\ref{fig:network3b}.

Unfamiliar readers when looking at the graphical representation of an artificial neural network (see, e.g., Figure~\ref{fig:Oishi-network}) could be misled in thinking in terms of electrical (or fluid-flow) networks, in which Kirchhoff's law applies at the junction where the output is split into different directions to go toward other artificial neurons.  The big picture is not clear at the outset, and could be confusing to readers new to the field, who would take some time to understand; see also Footnote~\ref{fn:confusion}.
By contrast, Figure~\ref{fig:network3b} clearly shows a multilevel function composition, assuming that first-time learners are familiar with this basic mathematical concept.

% \subsection{Mathematical representation}
\subsection{Matrix notation}
\label{sc:matrix}

In mechanics and physics, tensors are intrinsic geometrical objects, which can be represented by infinitely many matrices of components, depending on the coordinate systems.\footnote{
	See, e.g., \cite{Brillouin.1964:rd0001}, \cite{Misner.1973:rd0001}.
} vectors are tensors of order one (1).  
For this reason, we do {\em not} use neither the name ``vector'' for a column matrix, nor the name ``tensor'' for an array with more than two indices.\footnote{
	See, e.g., \cite{Goodfellow.2016}, p.~31, where a ``vector'' is a column matrix, and a ``tensor'' is an array with coefficients (elements) having more than two indices, e.g., $A_{ijk}$.  
	It is important to know the terminologies used in computer-science literature.
} All arrays are matrices.

The matrix notation used here can follow either (1) the Matlab / Octave code syntax, or (2) the more compact component convention for tensors in mechanics.

Using Matlab / Octave code syntax, the inputs to a network (to be defined soon) are gathered in an $n \times 1$ column matrix $\boldsymbol x$ of real numbers, and the ({\em target} or labeled) outputs\footnote{
	\label{fn:predicted-output}
	The inputs $\boldsymbol x$ and the {\em target} (or labeled) outputs $\boldsymbol y$ are the data used to train the network, which produces the predicted (or approximated) output denoted by $\nfwidetilde{\by}$ with an overhead tilde reminiscent of the approximation symbol $\approx$.
	See also Footnote~\ref{fn:predicted-value-hat}.
} in an $m \times 1$ column matrix $\boldsymbol y$
\begin{align}
  \boldsymbol x 
  = [ x_1 , \cdots , x_n ]^T 
  = [ x_1 ; \cdots ; x_n ] 
  = 
  \begin{bmatrix}
  x_1
  \\
  \vdots
  \\
  x_n
  \end{bmatrix}
  \in {\mathbb R}^{n \times 1}
  \ , \quad
  \boldsymbol y = [ y_1 , \cdots , y_m ]^T
  \in {\mathbb R}^{m \times 1}
  \label{eq:matrices_x_y}
\end{align}
where the commas are separators for matrix elements in a row, and the semicolons are separators for rows.  For matrix transpose, we stick to the standard notation using the superscript ``$T$'' for written documents, instead of the prime ``$\prime$'' as used in matlab / octave code.
In addition, the prime ``$\prime$'' is more customarily used to denote derivative in handwritten and in typeset equations.

Using the component convention for tensors in mechanics,\footnote{
See, e.g., 
\cite{Brillouin.1964:rd0001} 
\cite{Malvern.1969:rd0001} \cite{Marsden.1994:rd0001}.
}
The coefficients of a $n \times m$ matrix shown below
\begin{align}
\left[ A_{ij} \right] = \left[ A^i_j \right] \in {\mathbb R}^{n \times m}
\label{eq:matrixIndices}
\end{align}
are arranged according to the following convention for the free indices $i$ and $j$, which are automatically expanded to their respective full range, i.e., $i = 1, \ldots , n$ and $j = 1, \ldots m$ when the variable $A_{ij} = A^i_j$ are enclosed in square brackets: 
\begin{enumerate}
	
	\item 
	In case both indices are subscripts, then the left subscript (index $i$ of $A_{ij}$ in Eq.~(\ref{eq:matrixIndices})) denotes the row index, whereas the right subscript (index $j$ of $A_{ij}$ in Eq.~(\ref{eq:matrixIndices})) denotes the column index.  
	
	\item 
	In case one index is a superscript, and the other index is a subscript, then the superscript (upper index $i$ of $A^i_j$ in Eq.~(\ref{eq:matrixIndices})) is the row index, and the subscript (lower index $j$ of $A^i_j$ in Eq.~(\ref{eq:matrixIndices})) is the column index.\footnote{
	See, e.g., \cite{Vu-Quoc.1995:rd2053}, Footnote 11.  For example, $A_{32} = A^3_2$ is the coefficient in row $3$ and column $2$.  
	%, or \cite{Vu-Quoc.1995:rd9392}, footnote 12.
	}

\end{enumerate}

With this convention (lower index designates column index, while upper index designates row index), the coefficients of array $\boldsymbol x$ in Eq.~(\ref{eq:matrices_x_y}) can be presented either in row form (with lower index) or in column form (with upper index) as follows:
\begin{align}
	\left[ x_i \right] = \left[ x_1 , x_2 , \cdots , x_n \right] \in {\mathbb R}^{1 \times n}
	\ , \quad
	\left[ x^i \right] = 
	\begin{bmatrix}
	x^1
	\\
	x^2
	\\
	\vdots
	\\
	x^n
	\end{bmatrix}
	\in {\mathbb R}^{n \times 1}
	\ , \quad
	\text{ with }
	x_i = x^i
	\ , \forall i = 1 , \cdots , n
	\label{eq:index_i}
\end{align}
Instead of automatically associating any matrix variable such as $\boldsymbol x$ to the column matrix of its components, the matrix dimensions are clearly indicated as in Eq.~(\ref{eq:matrices_x_y}) and Eq.~(\ref{eq:index_i}), i.e., by specifying the values $m$ (number of rows) and $n$ (number of columns) of its containing space ${\mathbb{R}^{m \times n}}$.

Consider the Jacobian matrix
\begin{align}
\frac{\partial \boldsymbol y}{\partial \boldsymbol x} = \left[ \frac{\partial y_i}{\partial x_j} \right] \in {\mathbb R}^{m \times n} \text{ and let }  A^i_j := \frac{\partial y_i}{\partial x_j}
\label{eq:jacobian}
\end{align}
where $\boldsymbol y$ and $\boldsymbol x$ are column matrices shown in Eq.~(\ref{eq:matrices_x_y}). 
Then the coefficients of this Jacobian matrix are arranged with the upper index $i$ being the row index, and the lower index $j$ being the column index.\footnote{
For example, the coefficient $A^3_2 = \frac{\partial y_3}{\partial x_2}$ is in row $3$ and column $2$. 
The Jacobian matrix in this convention is the transpose of that used in \cite{Zienkiewicz.2013:rd0001}, p.~175.
} 
This convention is natural when converting a chain rule in component form into matrix form, i.e., consider the composition of matrix functions
\begin{align}
\boldsymbol z (\boldsymbol y (\boldsymbol x (\boldsymbol \theta)))
\ , 
\text{ with }
\boldsymbol \theta \in \mathbb{R}^p
\ , \quad
\boldsymbol x : {\mathbb R}^p \rightarrow {\mathbb R}^n 
\ , \quad
\boldsymbol y : {\mathbb R}^n \rightarrow {\mathbb R}^m
\ , \quad
\boldsymbol z : {\mathbb R}^m \rightarrow {\mathbb R}^l
\end{align}
where implicitly
\begin{align}
{\mathbb R}^p \equiv {\mathbb R}^{p \times 1}
\ , \quad
{\mathbb R}^n \equiv {\mathbb R}^{n \times 1}
\ , \quad
{\mathbb R}^m \equiv {\mathbb R}^{m \times 1}
\ , \quad
{\mathbb R}^l \equiv {\mathbb R}^{l \times 1}
\end{align}
are the spaces of column matrices, 
Then using the chain rule
\begin{align}
\frac{\partial z_i}{\partial \theta_j}
=
\frac{\partial z_i}{\partial y_r} 
\frac{\partial y_r}{\partial x_s} 
\frac{\partial x_s}{\partial \theta_j}
\end{align}
where the summation convention on the repeated indices $r$ and $s$ is implied.
Then the Jacobian matrix $\frac{\partial \boldsymbol z}{\partial \boldsymbol \theta}$ can be obtained directly as a product of Jacobian matrices from the chain rule just by putting square brackets around each factor:
\begin{align}
\frac{
\partial
\boldsymbol z
(
	\boldsymbol y
    (
    	\boldsymbol x (\boldsymbol \theta)
    )
)
}{\partial \boldsymbol \theta}
=
\left[
\frac{\partial z_i}{\partial \theta_j}
\right]
=
\left[
\frac{\partial z_i}{\partial y_r} 
\right]
\left[
\frac{\partial y_r}{\partial x_s} 
\right]
\left[
\frac{\partial x_s}{\partial \theta_j}
\right]
\end{align}

Consider the scalar function $E : {\mathbb R}^m \rightarrow {\mathbb R}$ that maps the column matrix $\boldsymbol y \in {\mathbb R}^m$ into a scalar, then the components of the gradient of $E$ with respect to $\boldsymbol y$ are arranged in a row matrix defined as follows:
\begin{align}
\left[ \frac{\partial E}{\partial y_j} \right]_{1 \times m}
=
\left[ \frac{\partial E}{\partial y_1} , \cdots , \frac{\partial E}{\partial y_m} \right]
=
\nabla_{\boldsymbol y}^T E
\in \mathbb{R}^{1 \times m}
\end{align}
with $\nabla_{\boldsymbol y}^T E$ being the transpose of the $m \times 1$ column matrix $\nabla _{\boldsymbol y} E$ containing these same components.\footnote{
	%
	% CMES style, rewriting
%\cite{Goodfellow.2016}, p.~82, refer to 
In \cite{Goodfellow.2016}, the column matrix (which is called ``vector'') $\nabla _{\boldsymbol y} E$ is referred to as the gradient of $E$.
Later on in the present paper, $E$ will be called the error or ``loss'' function, $\boldsymbol y$ the outputs of a neural network, and the gradient of $E$ with respect to $\boldsymbol y$ is the first step in the ``backpropagation'' algorithm in Section \ref{sc:backprop} to find the gradient of $E$ with respect to the network parameters collected in the matrix $\boldsymbol \theta$ for an optimization descent direction to minimize $E$.
}

Now consider this particular scalar function below:\footnote{
Soon, it will be seen in Eq.~(\ref{eq:linearComboInputsBias}) that the function $z$ is a linear combination of the network inputs $\boldsymbol y$, which are outputs coming from the previous network layer, with $\boldsymbol w$ being the weights.  
An advantage of defining $\boldsymbol w$ as a row matrix, instead of a column matrix like $\boldsymbol y$, is to de-clutter the equations in dispensing of (1) the superscript $T$ designating the transpose as in $\boldsymbol w^T \boldsymbol y$, or (2) the dot product symbol as in $\boldsymbol w \dotprod \boldsymbol y$, leaving space for other indices, such as in Eq.~(\ref{eq:linearComboInputsBias}).
}
\begin{align}
	Z = \boldsymbol w \boldsymbol y
	\text{ with }
	\boldsymbol w \in \mathbb{R}^{1 \times n} \text{ and } \boldsymbol y \in \mathbb{R}^{n \times 1}
	\label{eq:z=wy}
\end{align}
Then the gradients of $z$ are\footnote{
The gradients of $z$ will be used in the backpropagation algorithm in Section~\ref{sc:backprop} to obtain the gradient of the error (or loss) function $E$ to find the optimal weights that minimize $E$.
}
\begin{align}
	\frac{\partial z}{\partial w_j} = y_j 
	\Longrightarrow 
	\left[ \frac{\partial z}{\partial w_j} \right] = \left[ y_j \right] = \boldsymbol y^T
	\in \mathbb{R}^{1 \times n}
	\text{ and }
	\left[ \frac{\partial z}{\partial y_j} \right] = \left[ w_j \right] = \boldsymbol w
	\in \mathbb{R}^{1 \times n}
	\label{eq:dz}
\end{align}

%\subsection{Big picture of entire network}
\subsection{Big picture, composition of concepts}
\label{sc:big-picture}

A fully-connected feedforward network is a chain of successive applications of functions $f^{(\ell)}$ with $\ell = 1, \ldots , \D$, one after another---with $\D$ being the number of ``layers'' or the {\it depth} of the network---on the input $\boldsymbol x$ to produce the predicted output $\bout$ for the target output $\by$:
\begin{align}
	%\boldsymbol y
	\bout 
    =
    f (\boldsymbol x)
    & 
    =
    ( f^{(\D)} \circ f^{(\D-1)} \circ \cdots \circ f^{(\ell)} \circ \cdots \circ f^{(2)} \circ f^{(1)} ) (\boldsymbol x)
    \nonumber
    \\
    &
    =
    f^{(\D)} 
	( 
    	f^{(\D-1)} 
        ( 
        	\cdots 
            ( 
            	f^{(\ell)} 
                ( 
                	\cdots 
                    ( 
                    	f^{(2)} 
                        ( 
                        	f^{(1)} (\boldsymbol x) 
                        ) 
                    )
                	\cdots
                )
            )
            \cdots
        ) 
    )
    \label{eq:compositions}
\end{align}
or breaking Eq.~(\ref{eq:compositions}) down, step by step, from inputs to outputs:\footnote{
	To alleviate the notation, the predicted output $\byp{\ell}$ from layer $(\ell)$ is indicated by the superscript ${(\ell)}$, without the tilde.
	The output $\byp{L}$ from the last layer $(L)$ is the network predicted output $\bout$.  
}
\begin{align}
	&
    \boldsymbol y^{(0)}
    =
    \boldsymbol x^{(1)} 
    = 
    \boldsymbol x \text{ (inputs) }
    \nonumber
    \\
    &
    \boldsymbol y^{(\ell)}
    = 
    f^{(\ell)} (\boldsymbol x^{(\ell)})
    =
    \boldsymbol x^{(\ell + 1)}
    \text{ with }
    \ell = 1 , \cdots , \D-1
    \nonumber
    \\
    &
    \boldsymbol y^{(\D)}
    = 
    f^{(\D)} (\boldsymbol x^{(\D)})
    = 
    \bout  \text{ (predicted outputs) }
    \label{eq:input-predicted-output}
\end{align}
\begin{rem}
	{\rm 
		The notation $\byp{\ell}$, for $\ell = 0, \cdots , L$ in
		Eq.~(\ref{eq:input-predicted-output}) will be useful to develop a concise
		formulation for the computation of the gradient of the cost function
		relative to the parameters by backpropagation for use in training (finding
		optimal parameters); see Eqs.~(\ref{eq:gradient-1})-(\ref{eq:gradient-2}) in Section~\ref{sc:backprop} on Backpropagation.
		$\hfill\blacksquare$
	}
\end{rem}

The quantities associated with layer $(\ell)$ in a network are indicated with the superscript $(\ell)$, so that the inputs to layer $(\ell)$ as gathered in the $m_{(\ell-1)} \times 1$ column matrix
\begin{align}
	\boldsymbol x^{(\ell)} = [ x_1^{(\ell)} , \cdots , x_{m_{(\ell-1)}}^{(\ell)} ]^T 
    = 
    \boldsymbol y^{(\ell - 1)}
    \in {\mathbb R}^{m_{(\ell-1)} \times 1}
\end{align}
are the {\em predicted} outputs from the previous layer $(\ell - 1)$, gathered in the matrix $\boldsymbol y^{(\ell - 1)}$,
With $m_{(\ell-1)}$ being the {\em width} of layer $(\ell-1)$.
Similarly,
The outputs of layer $(\ell)$ as gathered in the $m_{(\ell)} \times 1$ matrix
\begin{align}
	\boldsymbol y^{(\ell)} = [ y_1^{(\ell)} , \cdots , y_{m_\ell}^{(\ell)} ]^T = \boldsymbol x^{(\ell + 1)}
    \in {\mathbb R}^{m_{(\ell)} \times 1}
\end{align}
are the inputs to the subsequent layer $(\ell + 1)$, gathered in the matrix $\boldsymbol x^{(\ell + 1)}$, with $m_{(\ell)}$ being the {\em width} of layer $(\ell)$.

\begin{rem}
	\label{rm:hidden}
	{\rm
	The output for layer $(\ell)$, denoted by $\byp{\ell}$, can also be written as $\bh^{(\ell)}$, where ``$h$'' is mnemonic for ``hidden'', since the inner layers between the input layer $(1)$ and the output layer $(L)$ are considered as being ``hidden''.
	Both notations $\byp{\ell}$ and $\bh^{(\ell)}$ are equivalent
	\begin{align}
		\by^{(\ell)}
		\equiv
		\bh^{(\ell)}
		\label{eq:equiv-y-h}
	\end{align}
	and can be used interchangeably.
	In the current Section~\ref{sc:feedforward} on ``Static, feedforward networks'', the notation $\byp{\ell}$ is used, whereas in Section~\ref{sc:recurrent} on ``Dynamics, sequential data, sequence modeling'', the notation $\bh^{[n]}$ is used to designate the output of the ``hidden cell'' at state $[n]$ in a recurrent neural network, keeping in mind the equivalence in Eq.~(\ref{eq:equiv-y-h-cell}) in Remark~\ref{rm:hidden-cell}.  
	Whenever necessary, readers are reminded of the equivalence in Eq.~(\ref{eq:equiv-y-h}) to avoid possible confusion when reading deep-learning literature.
	}
	\hfill$\blacksquare$ 
\end{rem}

The above chain in Eq.~(\ref{eq:compositions})---see also Eq.~(\ref{eq:network}) and Figure~\ref{fig:network3b}---is referred to as ``multiple levels of composition'' that characterizes modern deep learning, which no longer attempts to mimic the working of the brain from the neuroscientific perspective.\footnote{
   See \cite{Goodfellow.2016}, p.~14, p.163.
}   
Besides, a complete understanding of how the brain functions is still far remote.\footnote{
   %
   % CMES style, rewriting
%   \cite{Schmidhuber.2015:rd0001}---in 
   In the review paper \cite{Schmidhuber.2015:rd0001} addressing to computer-science experts, and dense with acronyms and jargon ``foreign'' to first-time learners, the authors mentioned ``It is ironic that artificial NNs [neural networks] (ANNs) can help to better understand biological NNs (BNNs)'', and cited a 2012 paper that won an ``image segmentation'' contest in helping to construct a 3-D model of the ``brain's neurons and dendrites'' from ``electron microscopy images of stacks of thin slices of animal brains''.
}

\subsubsection{Graphical representation, block diagrams}
\label{sc:graphical-representation}

\begin{figure}[h]
  \centering
  %
  % 2022.12.17
  % remove ".eps" for arXiv
  % \includegraphics[width=0.5\linewidth]{figures/network1.eps}
  \includegraphics[width=0.5\linewidth]{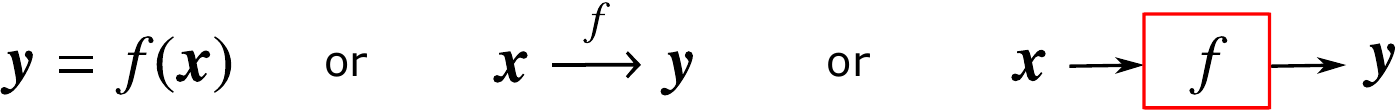}
  \caption{\emph{Function mapping, graphical representation} (Section~\ref{sc:graphical-representation}): $n$ inputs in $\boldsymbol x \in {\mathbb R}^{n \times 1}$ ($n \times 1$ column matrix of real numbers) are fed into function $f$ to produce $m$ outputs in $ \boldsymbol y \in {\mathbb R}^{m \times 1}$.}
  \label{fig:network1}
\end{figure}

A function can be graphically represented as in Figure~\ref{fig:network1}.
\noindent

\noindent
The multiple levels of compositions in Eq.~\eqref{eq:compositions} can then be represented by
\begin{align}
    \underbrace{
      \boldsymbol x
      =
      \boldsymbol y^{(0)} 
    }_{\text{Input}}
    \quad
    \underbrace{
      \stackrel{f^{(1)}}{\longrightarrow}
      % \overset{A}{B}
      \;
      \boldsymbol y^{(1)}
      \;
      \stackrel{f^{(2)}}{\longrightarrow}
      \;
      \cdots
      \;
      % \stackrel{f^{(\ell-1)}}{\longrightarrow} 
      % \;
      \boldsymbol y^{(\ell-1)}
      \;
      \stackrel{f^{(\ell)}}{\longrightarrow}
      \;
      \boldsymbol y^{(\ell)}
      \;
      \cdots
      \;
      \stackrel{f^{(\D-1)}}{\longrightarrow} 
      \;
      \boldsymbol y^{(\D-1)}
      \;
      \stackrel{f^{(\D)}}{\longrightarrow}
    }_{\text{Network as multilevel composition of functions}}
    \quad
    \underbrace{
      \boldsymbol y^{(\D)}
      =
      \bout 
    }_{\text{Output}}
    \label{eq:network}
\end{align}
revealing the structure of the {\em feedforward network} as a multilevel composition of functions (or chain-based network) in which the output $\boldsymbol y^{(\ell-1)}$ of the previous layer $(\ell-1)$ serves as the input for the current layer $(\ell)$, to be processed by the function $f^{(\ell)}$ to produce the output $\boldsymbol y^{(\ell)}$.  the input $\boldsymbol x = \boldsymbol y^{(0)}$ for the input layer $(1)$ is the input for the entire network.  The output $\boldsymbol y^{(L)} = \bout$ of the (last) layer $(L)$ is the predicted output for the entire network.
\begin{figure}[h]
  \centering
  %
  % 2022.12.17
  % add "-eps-converted-to.pdf" for arXiv
  % \includegraphics[width=0.8\linewidth]{figures/network3b}
  \includegraphics[width=0.8\linewidth]{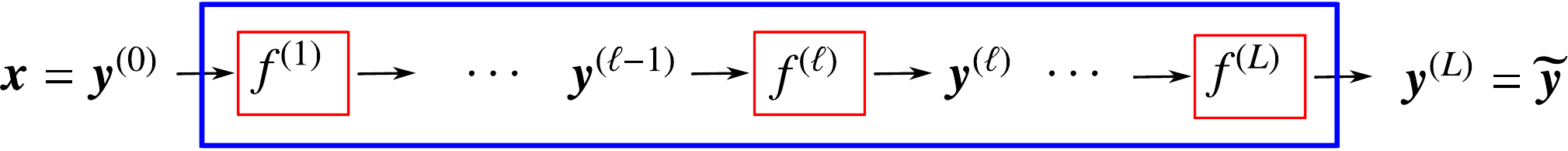}
  %
  % 2019.04.16, for some reasons, i could not use the macros \bout, or \nfwidetilde{\by},
  % inside the caption, but they worked outside the caption. 
  % so i just used \widetilde{\by}; it worked.
  \caption{
  	\emph{Feedforward network} (Sections~\ref{sc:graphical-representation}, \ref{sc:artificial-neuron}):
  	Multilevel composition in feedforward network with $L$ layers represented as a sequential application of functions $f^{(\ell)}$, with $\ell = 1, \cdots, L$, to $n$ inputs gathered in $\boldsymbol x = \boldsymbol y^{(0)} \in {\mathbb R}^{n \times 1}$ ($n \times 1$ column matrix of real numbers) to produce $m$ outputs gathered in $\boldsymbol y^{(L)} = \widetilde{\by} \in \mathbb{R}^{m \times 1}$.
  	This figure is a higher-level block diagram that corresponds to the lower-level neural network in Figure~\ref{fig:Oishi-network} or in Figure~\ref{fig:neuron4}.
  }
  \label{fig:network3b}
\end{figure}

\begin{rem}
	\label{rm:layer-definitions}
	Layer definitions, action layers, state layers.
	{\rm 
		In Eq.~\eqref{eq:network} and in Figure~\ref{fig:network3b}, an \emph{action layer} is defined by the action, i.e., the function $f^{(\ell)}$, on the inputs $\byp{\ell-1}$ to produce the outputs $\byp{\ell}$. There are $L$ action layers.
		A \emph{state layer} is a collection of inputs or outputs, i.e., $\byp{\ell}, \ell = 0, \ldots, L$, each describes a state of the system, thence the number of state layers is $L+1$, and the number of hidden (state) layers (excluding the input layer $\byp{0}$ and the output layer $\byp{L}$) is $(L-1)$.  For an illustration of  state layers, see \cite{Goodfellow.2016}, p.~6, Figure1.2.
		See also Remark~\ref{rm:Wang-layer-number}.  From here on,  ``hidden layer'' means ``hidden \emph{state} layer'', agreeing with the terminology in \cite{Goodfellow.2016}.
		%See also Section~\ref{sc:depth-size} on ``What is ``deep'' in ``deep networks'' ? Size, architecture'' and Section~\ref{sc:depth} on ``Depth, size''.
		See also Remark~\ref{rm:depth-definitions} on depth definitions in Section~\ref{sc:depth} on ``Depth, size''.
	}
	$\hfill\blacksquare$
\end{rem}

%\subsection{Processing unit, artificial neuron}
\subsection{Network layer, detailed construct}
\label{sc:network-layer-details}
\subsubsection{Linear combination of inputs and biases}
\label{sc:weigths-biases}

First, an affine transformation on the inputs (see Eq.~\eqref{eq:linearComboInputsBias}) is carried out, in which the coefficients of the inputs are called the weights, and the constants are called the biases.
The output $\boldsymbol y^{(\ell-1)}$ of layer $(\ell-1)$ is the input to layer $(\ell)$
\begin{align}
	\boldsymbol y^{(\ell-1)}
    =
    [ y^{(\ell-1)}_{1} , \cdots , y^{(\ell-1)}_{m_{(\ell-1)}} ]^T
    \in
    \mathbb{R}^{m_{(\ell-1)} \times 1}
    \ .
\end{align}
The column matrix $\boldsymbol z^{(\ell)}$ 
\begin{align}
	\boldsymbol z^{(\ell)}
    =
    [ z^{(\ell)}_{1} , \cdots , z^{(\ell)}_{m_{(\ell)}} ]^T
    \in
    \mathbb{R}^{m_{(\ell)} \times 1}
\end{align}
is a linear combination of the inputs in $\boldsymbol y^{(\ell-1)}$ plus the biases (i.e., an affine transformation)\footnote{
	See Eq.~(\ref{eq:linear-volterra-series}) for the continuous temporal summation, counterpart of the discrete spatial summation in Eq.~(\ref{eq:linearComboInputsBias}).
}
\begin{align}
\boxit{
	\boldsymbol z^{(\ell)}
    =
    \boldsymbol W^{(\ell)} \boldsymbol y^{(\ell-1)}
    +
    \boldsymbol b^{(\ell)}
    \text{ such that }
    z^{(\ell)}_i
    =
    \boldsymbol w^{(\ell)}_i \boldsymbol y^{(\ell-1)}
    +
    b^{(\ell)}_i
    \ , \text{ for }
    i = 1, \ldots , m_{(\ell)}
    \ ,
}
    \label{eq:linearComboInputsBias}
\end{align}
\noindent
where the $m_{(\ell)} \times m_{(\ell-1)}$ matrix $\boldsymbol W$ contains the weights\footnote{
	It should be noted that the use of both $\boldsymbol W$ and $\boldsymbol W^T$ in \cite{Goodfellow.2016} in equations equivalent to Eq.~(\ref{eq:linearComboInputsBias}) is confusing.  
	For example, on p.~205, in Section 6.5.4 on backpropagation for fully-connected feedforward network, Algorithm 6.3, an equation that uses $\boldsymbol W$ in the same manner as Eq.~(\ref{eq:linearComboInputsBias}) is 
	$
	\bkarp a {(k)}
	=
	\bkarp b {(k)}
	+
	\bkarp W {(k)}
	\bkarp h {(k)}
	$,
	whereas on p.~191, in Section 6.4, Architecture Design, Eq.~(6.40) uses $\boldsymbol W^T$ and reads as
	$
	\bkarp h {(1)}
	=
	G^{(1)}
	(
	\bkarp W {(1)T}
	\bkar x
	+
	\bkarp b {(1)} 
	)
	$, which is similar to Eq.~(6.36) on p.~187.
	On the other hand, both 
	$\boldsymbol W$ and $\boldsymbol W^T$ appear on the same p.~190 in the expressions
	$
	\cos
	(
	\bkar W \bkar x + \bkar b
	)
	$
	and
	$
	\bkar h = g(\bkarp W T \bkar x + \bkar b)
	$.
	Here, we stick to a single definition of $\bkarp W {(\ell)}$ as defined in Eq.~(\ref{eq:defineW}) and used in Eq.~(\ref{eq:linearComboInputsBias}).
}
\begin{align}
	\boldsymbol w^{(\ell)}_i
    =
    [ w^{(\ell)}_{i 1} , \cdots , w^{(\ell)}_{i m_{(\ell-1)}} ]
    \in
    \mathbb{R}^{m_{(\ell-1)} \times 1}
    \ , \quad
    \boldsymbol W^{(\ell)} = [ \boldsymbol w^{(\ell)}_1 ; \cdots ; \boldsymbol w^{(\ell)}_{m_{(\ell)}} ]
    =
    \begin{bmatrix}
    \boldsymbol w^{(\ell)}_1
    \\
    \vdots
    \\
    \boldsymbol w^{(\ell)}_{m_{(\ell)}}
    \end{bmatrix}
    \in
    \mathbb{R}^{m_{(\ell)} \times m_{(\ell-1)}}
    \ ,
    \label{eq:defineW}
\end{align}
and the $m_{(\ell)} \times 1$ column matrix $\boldsymbol b^{(\ell)}$ the biases:\footnote{
	Eq.~(\ref{eq:linearComboInputsBias}) is a linear (additive) combination of inputs with possibly non-zero biases.
	%
	% CMES style, rewriting
%	\cite{Schmidhuber.2015:rd0001} mentioned 
	An additive combination of inputs with zero bias, and a ``multiplicative'' combination of inputs of the form $z^{(\ell)}_i = \prod_{k=1}^{k=\widths{\ell}} \weightsp{ik}{\ell} \ysp{k}{\ell}$ with zero bias, were mentioned in \cite{Schmidhuber.2015:rd0001}.
	%
	% CMES style, rewriting
%	\cite{Werbos.1988} 
	In \cite{Werbos.1988}, the author went even further to propose the general case in which $\byp{\ell} = \mathcal{F}^{(\ell)} ( \byp{k} , \,  \text{ with }k < \ell )$, where $\mathcal{F}^{(\ell)}_i$ is {\em any} differentiable function.
	But it is not clear whether any of these more complex functions of the inputs were used in practice, as we have not seen any such use, e.g., in \cite{Nielsen.2015} \cite{Goodfellow.2016}, and many other articles, including review articles such as \cite{LeCun.2015:rd0001} \cite{Sze.2017:rd0001}.
	On the other hand, the additive combination has a clear theoretical foundation as the linear-order approximation to the Volterra series Eq.~(\ref{eq:volterra-series}); see Eq.~(\ref{eq:linear-volterra-series}) and also \cite{Dayan.2001}.  
}
\begin{align}
	\boldsymbol z^{(\ell)}
    =
    [ z^{(\ell)}_{1} , \cdots , z^{(\ell)}_{m_{(\ell)}} ]^T
    \in
    \mathbb{R}^{m_{(\ell)} \times 1}
    \ , \quad
    \boldsymbol b^{(\ell)}
    =
    [ b^{(\ell)}_{1} , \cdots , b^{(\ell)}_{m_{(\ell)}} ]^T
    \in
    \mathbb{R}^{m_{(\ell)} \times 1}
    \ .
\end{align}
Both the weights and the biases are collectively known as the network parameters, defined in the following matrices for layer $(\ell)$:
\begin{align}
	\bparamsp i \ell
	=
	\left[ \paramsp {ij} \ell \right]
	=
	\left[ \paramsp {i1} \ell , \ldots , \paramsp {i m_{(\ell-1)}} \ell \ | \ \paramsp {i (m_{(\ell-1)} + 1)} \ell \right]
	&
	=
	\left[ \weightsp {i1} \ell , \ldots , \weightsp {i m_{(\ell-1)}} \ell \ | \ \biassp i \ell \right]
	\nonumber
	\\
	&
	=
	\left[ \bweightsp i \ell \ | \ \biassp i \ell \right]
	\in \mathbb{R}^{1 \times [m_{(\ell-1)} + 1]}
\end{align}
\begin{align}
	\bParamp \ell
	=
	\left[ 
	\paramsp {ij} \ell
	\right]
	=
	\begin{bmatrix}
	\bparamsp {1} \ell
	\\
	\vdots
	\\
	\bparamsp {m_{(\ell)}} \ell
	\end{bmatrix}
	=
	\left[
	\left.
	\boldsymbol W^{(\ell)} 
	\ \right| \ 
	\boldsymbol b^{(\ell)}
	\right]
	\in \mathbb{R}^{m_{(\ell)} \times [m_{(\ell-1)} + 1]}
	\label{eq:parameters}
\end{align}
For simplicity and convenience, the set of all parameters in the network is denoted by $\bparam$, and the set of all parameters in layer $(\ell)$ by $\bparamp \ell$:\footnote{
	For the convenience in further reading, wherever possible, we use the same notation as in \cite{Goodfellow.2016}, p.xix.
	}
\begin{align}
	\bparam 
	=
	\{ \bparamp 1 ,\cdots, \bparamp \ell , \cdots , \bparamp L \}
	= 
	\{ \bParamp 1 ,\cdots, \bParamp \ell , \cdots , \bParamp L \}
	\, , \text{ such that }
	\bparamp \ell \equiv \bParamp \ell
	\label{eq:theta}
\end{align}
Note that the set $\bparam$ in Eq.~(\ref{eq:theta}) is not a matrix, but a set of matrices, since the number of rows $m_{(\ell)}$ for a layer $(\ell)$ may vary for different values of $\ell$, even though in practice, the widths of the layers in a fully connected feed-forward network may generally be chosen to be the same.

Similar to the definition of the parameter matrix $\bparamp \ell$ in Eq.~(\ref{eq:parameters}), which includes the biases $\bbiasp \ell$, it is convenient for use later in elucidating the backpropagation method in Section~\ref{sc:backprop} (and Section~\ref{sc:gradient} in particular) to expand the matrix $\boldsymbol{y}^{(\ell-1)}$ in Eq.~(\ref{eq:linearComboInputsBias}) into the matrix $\expand{\by}^{(\ell-1)}$ (with an overbar) as follows:
\begin{align}
	\boldsymbol z^{(\ell)}
	=
	\boldsymbol W^{(\ell)} \boldsymbol y^{(\ell-1)}
	+
	\boldsymbol b^{(\ell)}
	\equiv
	\left[
	\left.
	\boldsymbol W^{(\ell)} 
	\ \right| \ 
	\boldsymbol b^{(\ell)}
	\right]
	\begin{bmatrix}
		\boldsymbol y^{(\ell-1)}
		\\
		1
	\end{bmatrix}
	=:
	\bparamp \ell
	\ 
	\expand{\by}^{(\ell-1)}
	\ ,
	\label{eq:expanded-output-1}
\end{align}
with
\begin{align}
	\bparamp \ell
	:=
	\left[
	\left.
	\boldsymbol W^{(\ell)} 
	\ \right| \ 
	\boldsymbol b^{(\ell)}
	\right]
	\in \mathbb{R}^{m_{(\ell)} \times [m_{(\ell-1)} + 1]}
	\ , 
	\text{ and }
	\expand{\by}^{(\ell-1)}
	:=
	\begin{bmatrix}
	\boldsymbol y^{(\ell-1)}
	\\
	1
	\end{bmatrix}
	\in \mathbb{R}^{[m_{(\ell-1)} + 1] \times 1}
	\ .
	\label{eq:expanded-output-2}
\end{align}
The total number of parameters of a fully-connected feedforward network is then
\begin{align}
	\Tparam
	:=
	\sum_{\ell = 1}^{\ell = L} m_{(\ell)} \times [m_{(\ell-1)} + 1]
	\ .
	\label{eq:totalParams}
\end{align}
But why using a linear (additive) combination (or superposition) of inputs with weights, plus biases, as expressed in Eq.~(\ref{eq:linearComboInputsBias}) ? See Section~\ref{sc:linear-combo-history}.  

\subsubsection{Activation functions}
\label{sc:activation-functions}

An activation function $\g: {\mathbb R} \rightarrow {\mathbb R}$, which is a nonlinear real-valued function, is used to decide when the information in its argument is relevant for a neuron to activate.  
In other words, an activation function filters out information deemed insignificant, and is applied {\em element-wise} to the matrix $\boldsymbol z^{(\ell)}$ in Eq.~(\ref{eq:linearComboInputsBias}), obtained as a linear combination of the inputs plus the biases:
\begin{align}
\boxit{
	\boldsymbol y^{(\ell)}
    =
    \g (\boldsymbol z^{(\ell)})
    \text{ such that } y^{(\ell)}_i = \g (z^{(\ell)}_i)
}
    \label{eq:activationFunction}
\end{align}
Without the activation function, the neural network is simply a linear regression, and cannot learn and perform complex tasks, such as image classification, language translation, guiding a driver-less car, etc. 
See Figure~\ref{fig:neuron2} for the block diagram of a one-layer network.
 
An example is a linear one-layer network, without activation function, being unable to represent the seemingly simple XOR (exclusive-or) function, which brought down the first wave of AI (cybernetics), and that is described in Section~\ref{sc:XORfunction}.   

\begin{figure}[h]
  % tikz examples
  % http://www.texample.net/tikz/examples/linear-regression/
  % need to use the following tikz libraries
  % \usetikzlibrary{arrows,intersections}
  
  % rectified linear unit and its 1st and 2nd derivatives
  \begin{tikzpicture}[
      thick,
      >=stealth',
      dot/.style = {
        draw,
        fill = white,
        circle,
        inner sep = 0pt,
        minimum size = 4pt
      }
    ]
   
    % naming convention
    % pt (x,y) : absolute coordinates (x,y), relative to first system of axes 
    
    % plot 1, rectified linear unit (ReLU)

    % label function as rectified linear function
    \draw (0,3.0) node[label = {Rectified linear function}] {};

	% horizontal x axis, from pt (-2,0) to pt (3,0), labeled as "z"
    \draw[->] (-2,0) -- (3,0) coordinate[label = {below:$z$}];

    % vertical y axis, from pt (0,-0.3) to pt (0,2.5)
    \draw[->] (0,-0.3) -- (0,2.5);
    % label y axis as "$g(z) = \max(0,z)$" above the tip of y axis
    \draw (0,2.3) node[label = {$\g(z) = \max(0,z)$}] {};
   
    % name pt (0,0) as "origin1"
    \coordinate (origin1) at (0,0);
    
    % horizontal thick red line, at y = 0, for x < 0, from pt (-2,0) to "origin1"
    % \draw [line width = 3pt, red]     (-2,0) -- (0,0);
    \draw [line width = 3pt, red]     (-2,0) -- (origin1);
    % inclined line at 45 deg with x axis, from "origin1" to pt (2.5,2.5)
    \draw [line width = 3pt, red]     (origin1) -- (2.5,2.5);
   
   	% draw black dot at "origin1"
    \filldraw (origin1) circle (2pt);
   
    % put label "0" above left of "origin1"
    \draw (origin1) node[label = {above left:$0$}] {};
       
    % -----------------------------------------
    % plot 2, 1st derivative of ReLU, Heaviside function
    
    % label function as Heaviside function
    \draw (7,2.7) node[label = {Rectified linear function}] {};
    
    % horizontal x axis
    \draw[->] (4,0) -- (9,0) coordinate[label = {below:$z$}];
    
    % vertical y axis
    \draw[->] (6,-0.3) -- (6,2.5) coordinate[label = {right:$\g^\prime (z) = H(z)$}];

    % naming coordinate (0,0) as "origin2"
    \coordinate (origin2) at (6,0);

    % horizontal thick red line, at y = 0, for x < 0, 
    % from x = -2 [pt (8,0) or (origin)] to x = 0 [pt (10,1)]
    \draw [line width = 3pt, red]     (4,0) -- (origin2);

    % horizontal thick red line, at y = 1, for x >= 0, from pt (10,1) to pt (14,1)
    \draw [line width = 3pt, red]     (6,1) -- (9,1);
    
    % put open circle at (10,1), and number 1 at above left of coordinate (0,1)
    \draw (6,1) node[dot, label = {above left:$1$}] {};

    % open circle at (origin2), and number 0 at above left of "origin2" 
    \draw (origin2) node[dot, label = {above left:$0$}] {};
       
    % -----------------------------------------
    % plot 3, 2nd derivative of ReLU, Dirac delta function

    % label function as Direc delta function
    \draw (13,2.7) node[label = {Dirac delta}] {};

    % horizontal x axis
    \draw[->] (10,0) -- (14,0) coordinate[label = {below:$z$}];
    
    % vertical y axis
    \draw[->] (12,-0.3) -- (12,2.5) coordinate[label = {right:$\g^{\prime\prime} (z) = \delta(z)$}];

    % naming coordinate (0,0) as "origin2"
    \coordinate (origin3) at (12,0);	
    
    % Dirac delta
    \draw[line width = 3pt, red, ->] (12,0) -- (12,1.5);

    % open circle at (origin3), and number 0 at above left of "origin3" 
    \draw (origin3) node[dot, label = {above left:$0$}] {};

  \end{tikzpicture}
  \caption{
  	\emph{Activation function} (Section~\ref{sc:activation-functions}): Rectified linear function  and its derivatives.   
  	See also Section~\ref{sc:parametric-ReLU} and Figure~\ref{fig:parametric-relu} for Parametric ReLU that helped surpass human level performance in ImageNet competition for the first time in 2015, Figure~\ref{fig:ImageNet-error} \cite{He.2015b}.
  	See also Figure~\ref{fig:halfwave} for a halfwave rectifier.
  }
  \label{fig:ReLU}
\end{figure}
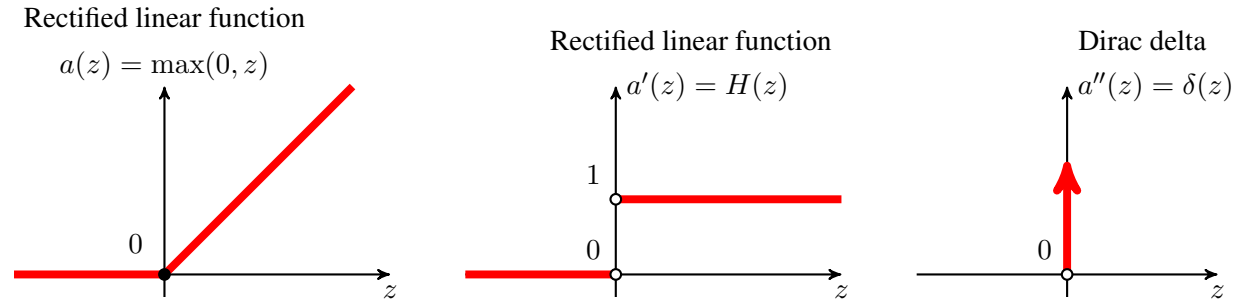

{\bf Rectified linear units (ReLU).}
Nowadays, for the choice of activation function $\g(\cdot)$,
Most modern {\em large} deep-learning networks use the default,\footnote{
	``In modern neural networks,
	The default recommendation is to use the rectified linear unit, or ReLU,'' \cite{Goodfellow.2016}, p.~168.
} well-proven 
{\em rectified linear function} (more often known as the ``positive part'' function) defined as\footnote{
	The notation $z^+$ for positive part function is used in the mathematics literature, e.g., ``Positive and negative parts'', Wikipedia,  \href{https://en.wikipedia.org/w/index.php?title=Positive_and_negative_parts&oldid=830205996}{version 12:11, 13 March 2018}, and less frequently in the computer-science literature, e.g., \cite{Jarrett.2009:rd0001}.
	% Renze,  
	% \href{http://mathworld.wolfram.com/PositivePart.html}{``Positive part''}, 
	% MathWorld--A Wolfram Web Resource, created by Eric W. Weisstein; 
	The notation $[z]_+$ is found in the neuroscience literature, e.g., \vphantom{\cite{Hahnloser.2000:rd0002}} \cite{Hahnloser.2000:rd0002} \cite{Dayan.2001}, p.~63.
	The notation $\max(0,z)$ is more widely used in the computer-science literature, e.g., \cite{Nair.2010:rd0001} \cite{Glorot.2011:rd0001}, \cite{Goodfellow.2016}.
} 
\begin{align}
	\g(z) = z^+ = [z]_+ = \max(0,z)
	=
	\begin{cases}
	0 & \text{ for } z \le 0
	\\	z & \text{ for } 0 < z
	\end{cases}
	\label{eq:ReLU}
\end{align}
and depicted in Figure~\ref{fig:ReLU}, for which the processing unit is called the {\em rectified linear unit} (ReLU),\footnote{
	A similar relation can be applied to define the Leaky ReLU in Eq.~(\ref{eq:leaky-relu}).
} which was demonstrated to be superior to other activation functions in many problems.\footnote{
	%
	% CMES style, rewriting
%	\cite{Goodfellow.2016}, p.~15, cited 
    In \cite{Goodfellow.2016}, p.~15, the authors cited the original papers \cite{Jarrett.2009:rd0001} and \cite{Nair.2010:rd0001}, where ReLU was introduced in the context of image / object recognition, and \cite{Glorot.2011:rd0001},
	where the superiority of ReLU over hyperbolic-tangent units and sigmoidal units was demonstrated.
} 
Therefore, in this section, we discuss in detail the rectified linear function, with careful explanation and motivation.
It is important to note that ReLU is superior for {\em large} network size, and may have about the same, or less, accuracy than the older logistic sigmoid function for ``very small'' networks, while requiring less computational efforts.\footnote{
	See, e.g., \cite{Goodfellow.2016}, p.~219, and Section~\ref{sc:active-function-history} on the history of active functions.
	See also  
	Section~\ref{sc:depth-size} for a discussion of network size.
	The reason for less computational effort with ReLU is due to (1) it being an identity map for positive argument, (2) zero for negative argument, and (3) its first derivative being the 
	Step (Heaviside) function as shown in Figure~\ref{fig:ReLU}, and explained in Section~\ref{sc:backprop} on Backpropagation.
}  

\begin{figure}[h]
	% tikz examples
	% http://www.texample.net/tikz/examples/linear-regression/
	% need to use the following tikz libraries
	% \usetikzlibrary{arrows,intersections}
	
	% rectified linear unit and its 1st and 2nd derivatives
	\begin{tikzpicture}[
	thick,
	>=stealth',
	dot/.style = {
		draw,
		fill = white,
		circle,
		inner sep = 0pt,
		minimum size = 4pt
	}
	]

	% naming convention
	% pt (x,y) : absolute coordinates (x,y), relative to first system of axes 
	
	% -----------------------------------------
	% plot 1, I-V plot for ideal diode
	
	% label function as ideal diode
	\draw (0,2.5) node[label = {Ideal diode}] {};
	
	% horizontal x axis, from pt (-2,0) to pt (2,0), labeled as "V" (voltage)
	\draw[->] (-2,0) -- (2,0) coordinate[label = {below:$V$}];
	
	% vertical y axis, from pt (0,-0.3) to pt (0,2.5)
	\draw[->] (0,-0.3) -- (0,2.5);
	% label y axis as "I" (current) above the tip of y axis
	\draw (0.4,2.0) node[label = {$I$}] {};
	
	% name pt (0,0) as "origin1"
	\coordinate (origin1) at (0,0);
	
	% horizontal thick red line, at y = 0, for x < 0, from pt (-2,0) to "origin1"
	% \draw [line width = 3pt, red]     (-2,0) -- (0,0);
	\draw [line width = 3pt, red]     (-2,0) -- (origin1);
	% vertical line from "origin1" to pt (0,2.0)
	\draw [line width = 3pt, red]     (origin1) -- (0,2.0);
	
	% draw black dot at "origin1"
	\filldraw (origin1) circle (2pt);
	
	% put label "0" above left of "origin1"
	\draw (origin1) node[label = {above left:$0$}] {};
	
	% -----------------------------------------
	% plot 2, I-V plot for resistance
	
	% label function as Resistance
	\draw (5.0,2.5) node[label = {Resistance}] {};
	
	% horizontal x axis
	\draw[->] (3,0) -- (8,0) coordinate[label = {below:$V$}];
	
	% vertical y axis
	%\draw[->] (6,-0.3) -- (6,2.5) coordinate[label = {right:$I = V / R$}];
	\draw[->] (5,-0.3) -- (5,2.5);
	\draw (6.0,1.9) node[label = {$I = V / R$}] {};
	% \draw (6.0,1.4) node[label = {$R = 2$}] {};
	
	% naming coordinate (0,0) as "origin2"
	\coordinate (origin2) at (5,0);
	
	% thick red line from pt (-1+5,-1/2) to pt (2+5,1), for I = V / 2
	% \draw [line width = 3pt, red]     (4,-1/2) -- (7,1);
	% thick red line from pt (-1+5,-1/2) to pt (3+5,1.5), for I = V / 2
	\draw [line width = 3pt, red]     (4,-1/2) -- (8,1.5);
	
	% draw black dot at "origin2"
	\filldraw (origin2) circle (2pt);
	
	% -----------------------------------------
	% plot 3, ReLU function max(0 , z / R)
	
	% label function as Heaviside function
	\draw (12,2.7) node[label = {{\em Scaled} rectified linear function}] {};
	
	% horizontal x axis
	\draw[->] (9,0) -- (14,0) coordinate[label = {below:$z$}];
	
	% vertical y axis
	\draw[->] (11,-0.3) -- (11,2.5) coordinate[label = {right:$a(z)=max(0 , z/R)$}];
	
	% naming coordinate (0,0) as "origin2"
	\coordinate (origin3) at (11,0);	
		
	% horizontal thick red line, at y = 0, for x < 0, from pt (-2,0) to "origin1"
	% \draw [line width = 3pt, red]     (-2,0) -- (0,0);
	\draw [line width = 3pt, red]     (9,0) -- (origin3);
	
	% thick red line from origin2 to pt (3+11,1.5), for I = V / 2
	\draw [line width = 3pt, red]     (origin3) -- (14,1.5);
	
	% open circle at (origin3), and number 0 at above left of "origin3" 
	\draw (origin3) node[label = {above left:$0$}] {};
	
	% draw black dot at "origin2"
	\filldraw (origin3) circle (2pt);
		
	\end{tikzpicture}
	\caption{
		\emph{Current $I$ versus voltage $V$} (Section~\ref{sc:activation-functions}): Ideal diode, resistance, {\em scaled} rectified linear function as activation (transfer) function for the ideal diode and resistance in series. 
		(Figure plotted with $R=2$.)
		See also Figure~\ref{fig:halfwave} for a halfwave rectifier.
	}
	\label{fig:diode-ReLU}
\end{figure}
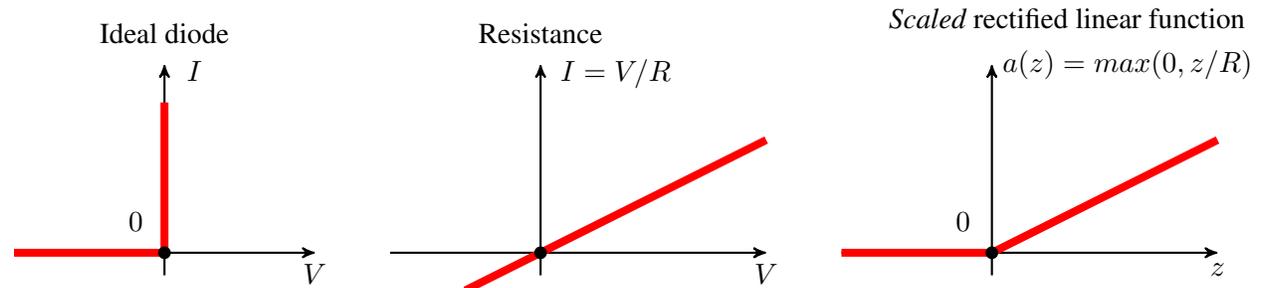

\begin{figure}[h]
  \centering
  %
  % 2022.12.17
  % remove ".eps" for arXiv
  % \includegraphics[width=0.9\linewidth]{figures/halfwave.eps}
  \includegraphics[width=0.9\linewidth]{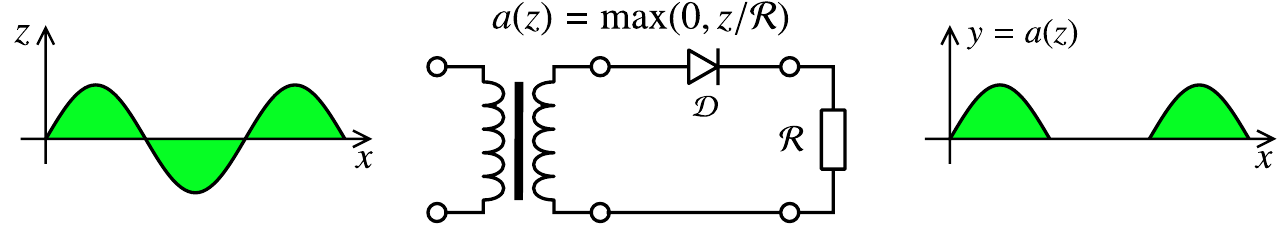}
  \caption{
  {\em Halfwave rectifier circuit} (Section~\ref{sc:activation-functions}), with a primary alternative current $z$ going in as input (left), passing through a transformer to lower the voltage amplitude, with the secondary alternative current out of the transformer being put through a closed circuit with an ideal diode $\mathcal{D}$ and a resistor $\mathcal{R}$ in series, resulting in a halfwave output current, which can be grossly approximated by the {\em scaled} rectified linear function $y \approx a(z) = \max (0, z / \mathcal{R})$ (right) as shown in Figure~\ref{fig:diode-ReLU}, with scaling factor $1/\mathcal{R}$.  The rectified linear unit in Figure~\ref{fig:ReLU} corresponds to the case with $\mathcal{R} = 1$.  
  For a more accurate Shockley diode model,
  The relation between current $I$ and voltage $V$ for this circuit is given in Figure~\ref{fig:I-V-halfwave}.
  Figure based on source in Wikipedia, \href{https://commons.wikimedia.org/w/index.php?title=File:Halfwave.rectifier.en.svg&oldid=145584715}{\rm version 01:49, 7 January 2015}. 
	}
  \label{fig:halfwave}
\end{figure}

To transform an alternative current into a direct current, the first step is to rectify the alternative current by eliminating its negative parts, and thus 
The meaning of the adjective ``rectified'' in {\em rectified linear unit} (ReLU).
Figure~\ref{fig:diode-ReLU} shows the current-voltage relation for an ideal diode, for a resistance, which is in series with the diode, and for the resulting ReLU function that rectifies an alternative current as input into the halfwave rectifier circuit in 
Figure~\ref{fig:halfwave}, resulting in a halfwave current as output.  

Mathematically, a periodic function remains periodic after passing through a (nonlinear) rectifier (active function):
\begin{align}
	z(x + T) = z(x) \Longrightarrow y(x+T) = a(z(x+T)) = a(z(x)) = y(x)
	\label{eq:periodic}
\end{align}
where $T$ in Eq.~\eqref{eq:periodic} is the period of the input current $z$.

\begin{figure}[h]
	\centering
	\includegraphics[width=0.9\textwidth]{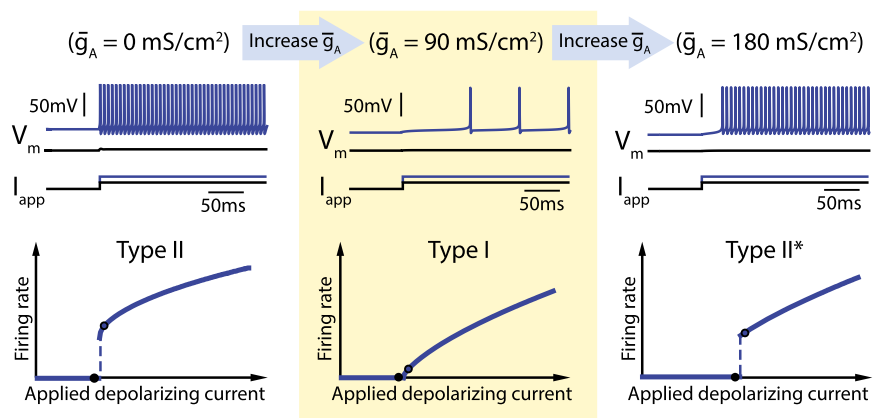}
	\caption{
		\emph{FI curves} (Sections~\ref{sc:activation-functions}, \ref{sc:dynamic-volterra-series}).
		Firing rate frequency (F) versus applied depolarizing current (I), thus FI curves.  Three types of FI curves.  The time histories of voltage $V_m$ provide a visualization of the spikes, current threshold, and spike firing rates.  The applied (input) current $I_{app}$ in increased gradually until it passes a current threshold, then the neuron begins to fire.  Two input current levels (two black dots on FI curves at the bottom) near the current threshold are shown, with one just below the threshold (black-line time history for $I_{app}$) and and one just above the threshold (blue line).  two corresponding histories of voltage $V_m$ (flat black line, and blue line with spikes) are also shown.  Type I displays a continuous increase in firing frequency from zero to higher values when the current continues to increase past the current threshold.  Type II displays a discontinuity in firing frequency, with a sudden jump from zero to a finite frequency, when the current passes the threshold.  At low concentration $\bar g_A$ of potassium, the neuron exhibits Type-II FI curve, then transitions to Type-I FI curve as $\bar g_A$ is increased, and returns to Type-II$^\star$ for higher concentration $\bar g_A$.  see \cite{Drion.2015}.  the {\em scaled} rectified linear unit (scaled ReLU, Figure~\ref{fig:diode-ReLU} and Figure~\ref{fig:halfwave}) can be viewed as approximating Type-I FI curve, see also Figure~\ref{fig:neuron-firing} and Eq.~(\ref{eq:firing-rate-constant-current}) where the FI curve is used in biological neuron firing-rate models.  \href{https://www.pnas.org/page/about/rights-permissions}{Permission of NAS}.
	}
	\label{fig:firing-rate}
\end{figure}

Biological neurons encode and transmit information over long distance by generating (firing) electrical pulses called action potentials or spikes with a wide range of frequencies \cite{Dayan.2001}, p.~1; see Figure~\ref{fig:firing-rate}.  ``To reliably encode a wide range of signals, neurons need to achieve a broad range of firing frequencies and to move smoothly between low and high firing rates'' \vphantom{\cite{Drion.2015}}\cite{Drion.2015}.  From the neuroscientific standpoint, the rectified linear function could be motivated as an idealization of the ``Type I'' relation between the firing rate (F) of a biological neuron and the input current (I), called the FI curve.  Figure~\ref{fig:firing-rate} describes three types of FI curves, with Type I in the middle subfigure, where there is a continuous increase in the firing rate with increase in input current.

\begin{figure}
	\centering
	\begin{subfigure}[b]{0.48\textwidth}
	\includegraphics[width=\linewidth]{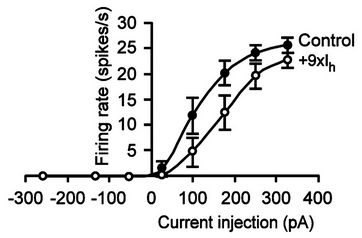}
	\caption{
		Experiment. Neuron firing rate (F) vs input current (I), FI curve. Hyperpolarization-activated cation current $I_h$ shifts FI curve to the right \cite{Welie.2004}. Copyright 2004 NAS, with 
		\href{https://www.pnas.org/page/about/rights-permissions}{permission of NAS}. 
		% {\color{red} ASK PERMISSION}
	} 
	\end{subfigure}
	\ 
	\begin{subfigure}[b]{0.48\textwidth}
		\includegraphics[width=\linewidth]{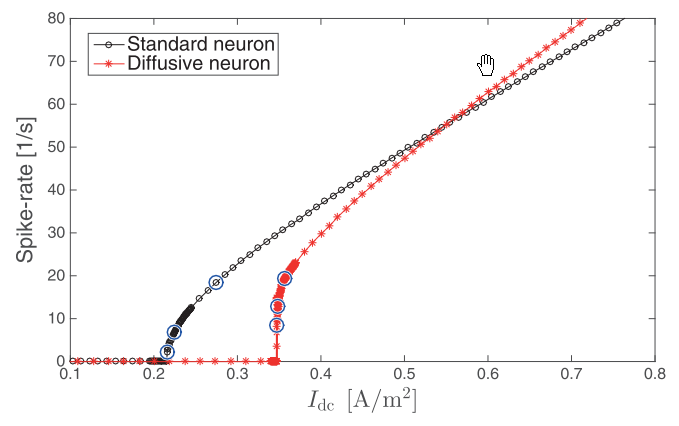}
		\caption{
			Modeling. FI curve for a Type-I neuron (black, Wilson equations
			Eqs.~(\ref{eq:Wilson-eq-1})-(\ref{eq:Wilson-eq-2})), and FI curve resulting from an aggregation of a large population of $10^4$ to $10^5$ of these Type-I neurons giving rise to a Type-II response (red, diffusive Wilson equations) \cite{Steyn-Ross.2016:rd0001}. 
			%Firing rate of Type I neuron (black, Wilson equations) and of Type II neuron (red, diffusive Wilson equations) vs. Input current.  \cite{Steyn-Ross.2016:rd0001},
			\href{https://creativecommons.org/licenses/by/3.0/}{CC-BY-3.0} license.
		}. 
	\end{subfigure}
	\hfill
	\begin{subfigure}[b]{0.48\textwidth}
		\includegraphics[width=\linewidth]{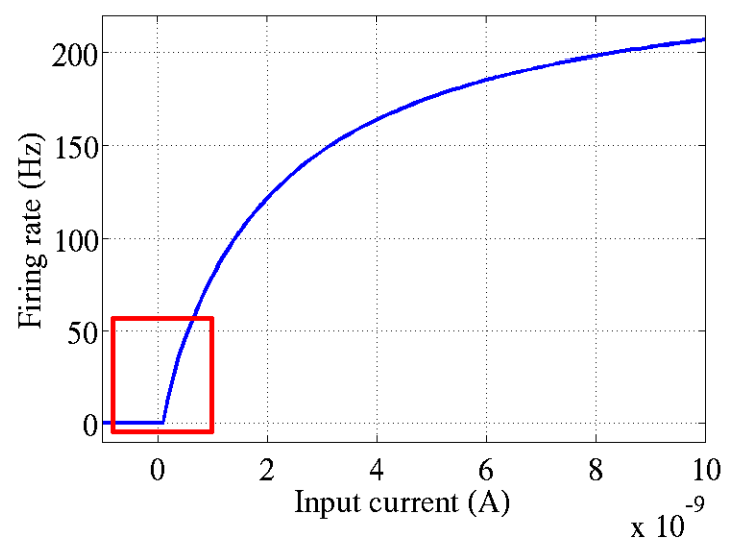}
		\caption{
			 Modeling. FI curve using the Integrate-and-Fire model; see
			 \cite{Dayan.2001}, p.~164, Eq.~(5.11).  The figure is from \cite{Glorot.2011:rd0001}. 
			 \\
			 {\footnotesize (Figure reproduced with permission of the authors.)}
			 % {\color{red} ASK PERMISSION}
		} 
	\end{subfigure}
	\ 
	\begin{subfigure}[b]{0.48\textwidth}
		\includegraphics[width=\linewidth]{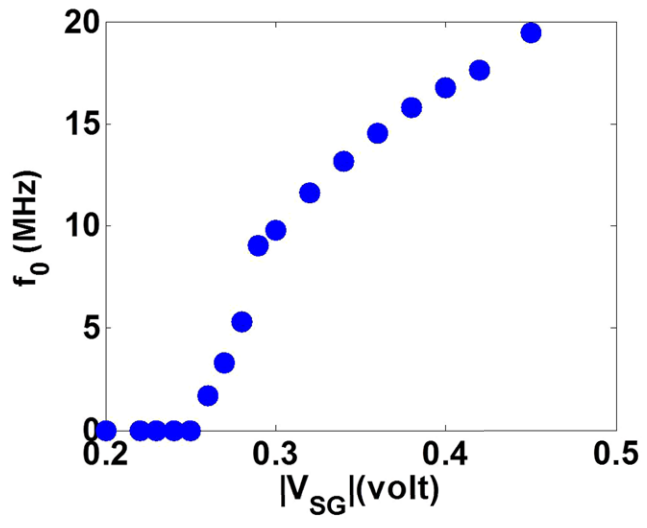}
		\caption{
			Hardware. Electronic-neuron firing rate (F)  vs. Input voltage (V).  Very high firing rates up to $20$ MHz, roughly $10^5$ to $10^6$ times faster than firing rate of biological neurons \cite{Dutta.2017:rd0001}, 
			\href{https://creativecommons.org/licenses/by/4.0/}{CC-BY-4.0} license.
			% {\color{red} ASK PERMISSION}
		} 
	\end{subfigure}
	\caption{
		\emph{FI or FV curves} (Sections~\ref{sc:comparison-three-fields}, \ref{sc:activation-functions}, \ref{sc:dynamic-volterra-series}).
		Neuron firing rate (F) versus input current (I) (FI curves, a,b,c) or voltage (V).  The Integrate-and-Fire model in SubFigure~(c) can be used to replace the sigmoid function to fit the experimental data points in SubFigure~(a).  The ReLU function in Figure~\ref{fig:ReLU} can be used to approximate the region of the FI curve just beyond the current or voltage
		thresholds, as indicated in the red rectangle in SubFigure~(c).  
		Despite the advanced mathematics employed to produce the Type-II FI curve of a large number of Type-I neurons, as shown in SubFigure~(b), it is not clear whether a similar result would be obtained if the single neuron displays a behavior as in Figure~\ref{fig:firing-rate}, with transition from Type II to Type I to Type II$^\star$ in a single neuron.  
		See Section~\ref{sc:dynamic-volterra-series} on ``Dynamic, time dependence, Volterra series'' for more discussion on 
		%
		% CMES style, rewriting
%		\cite{Wilson.1999}'s 
		Wilson's equations  Eqs.~(\ref{eq:Wilson-eq-1})-(\ref{eq:Wilson-eq-2}) \cite{Wilson.1999}.  On the other hand for deep-learning networks, the above results are more than sufficient to motivate the use of ReLU, which has deep roots in neuroscience.
	}
	\label{fig:neuron-firing}
\end{figure}
\begin{figure}[h]
	\centering
	\begin{subfigure}[b]{0.48\textwidth}
  %
  % 2022.12.17
  % remove ".eps" for arXiv
		% \includegraphics[width=\linewidth]{figures/diode-resistance-1.eps}
		\includegraphics[width=\linewidth]{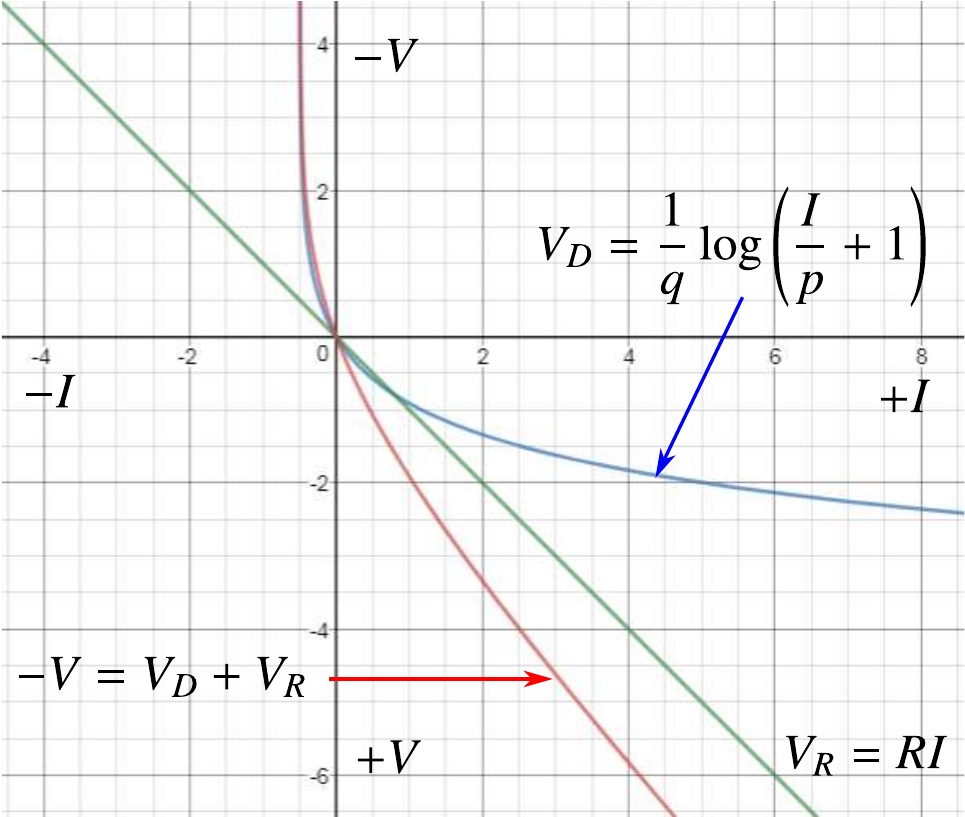}
		\caption{
			{\em Blue}: Voltage $V_D$ across diode $\mathcal{D}$ versus current $I$.
			{\em Green}: Voltage $V_R$ across resistance $\mathcal{R}$ versus current $I$.
			{\em Red}:   
			Voltage $-V$ across $\mathcal{D}$ and $\mathcal{R}$ in series versus $I$.
		}
	\end{subfigure}
	\ 
	\begin{subfigure}[b]{0.48\textwidth}
  %
  % 2022.12.17
  % remove ".eps" for arXiv
		% \includegraphics[width=\linewidth]{figures/diode-resistance-2.eps}
		\includegraphics[width=\linewidth]{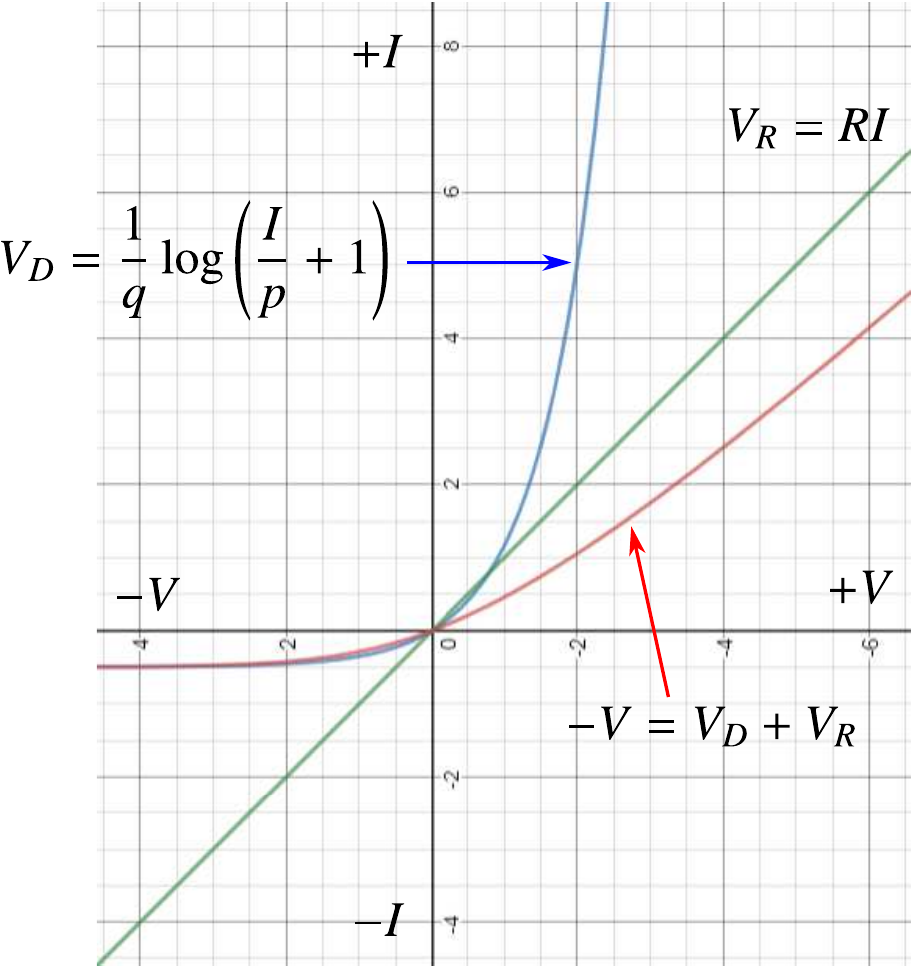}
		\caption{The same SubFigure~(a) rotated by +90 deg to show current $+I$ versus voltage $+V$, as a graphical solution for the nonlinear Eq.~\eqref{eq:diode-resistance}.}
	\end{subfigure}
	\caption{
		\emph{Halfwave rectifier} (Sections~\ref{sc:activation-functions}, \ref{sc:relu}).
		Current $I$ versus voltage $V$ [red line in SubFigure~(b)] in the halfwave rectifier circuit of Figure~\ref{fig:halfwave}, for which the ReLU function in Figure~\ref{fig:ReLU} is a gross approximation.  SubFigure~(a) was plotted with $p = 0.5$, $q = -1.2$, $R = 1$.  See also Figure~\ref{fig:crayfish-rectifier-experiment} for the synaptic response of crayfish similar to the red line in SubFigure~(b). 
	}
	\label{fig:I-V-halfwave}
\end{figure}

The Shockley equation for a current $I$ going through a diode $\mathcal{D}$, in terms of the voltage $V_D$ across the diode, is given in mathematical form as:
\begin{align}
	I = p \left[ e^{q V_D} - 1 \right] \Longrightarrow \displaystyle V_D = \frac{1}{q} \log \left( \frac{I}{p} + 1 \right) 
	\ .
	\label{eq:shockley}
\end{align}
With the voltage across the resistance being $V_R = RI$, the voltage across the diode and the resistance in series is then
\begin{align}
	-V = V_D + V_R = \displaystyle V_D = \frac{1}{q} \log \left( \frac{I}{p} + 1 \right) + RI
	\ ,
	\label{eq:diode-resistance}
\end{align}
which is plotted in Figure~\ref{fig:I-V-halfwave}.
The rectified linear function could be seen from Figure~\ref{fig:I-V-halfwave} as a very rough approximation of the current-voltage relation in a halfwave rectifier circuit in Figure~\ref{fig:halfwave}, in which a diode and a resistance are in series.  
In the Shockley model, the diode is leaky in the sense that there is a small amount of current flow when the polarity is reversed, unlike the case of an ideal diode or ReLU (Figure~\ref{fig:ReLU}), and is better modeled by the Leaky ReLU activation function, in which there is a small positive (instead of just flat zero) slope for negative $z$:
\begin{align}
	a(z) = 
	\max( 0.01 z , z)
	=
	\begin{cases}
	0.01 \, z & \text{ for } z \le 0
	\\
	z         & \text{ for } 0 < z
	\end{cases}
	\label{eq:leaky-relu}
\end{align}

\begin{figure}[h]
	\centering
	% 2019.09.05
	% see file psfrag.tex for the tikz figure, which is replaced here by the screenshot to accelerate the compilation
	\includegraphics[width=0.7\linewidth]{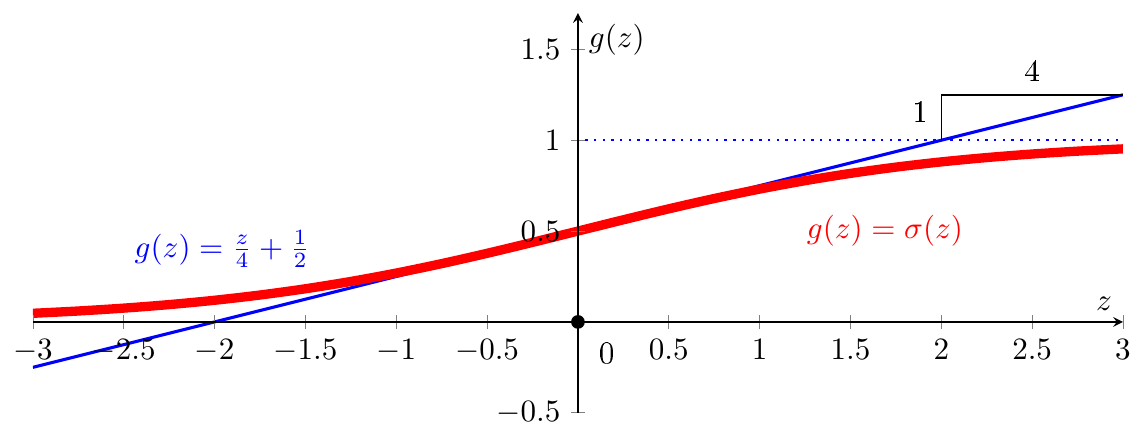}
	\caption{
%		Activation function: 
		\emph{Logistic sigmoid function} (Sections~\ref{sc:activation-functions}, \ref{sc:classification},
		\ref{sc:logistic-sigmoid}, \ref{sc:new-active-functions}): $\sigmoid (z) = [ 1 + \exp(-z) ]^{-1} = [ \tanh(z / 2) + 1 ] / 2$ (red), with the tangent at the origin $z = 0$ (blue).
		See also Remark~\ref{rm:softmax} and Figure~\ref{fig:sigmoid-z-minus-z} on the softmax function.
	}
	\label{fig:sigmoid}
\end{figure}

\begin{figure}[h]
	\centering
	% 2019.09.05
	% see file psfrag.tex for the tikz figure, which is replaced here by the screenshot to accelerate the compilation
\includegraphics[width=0.7\linewidth]{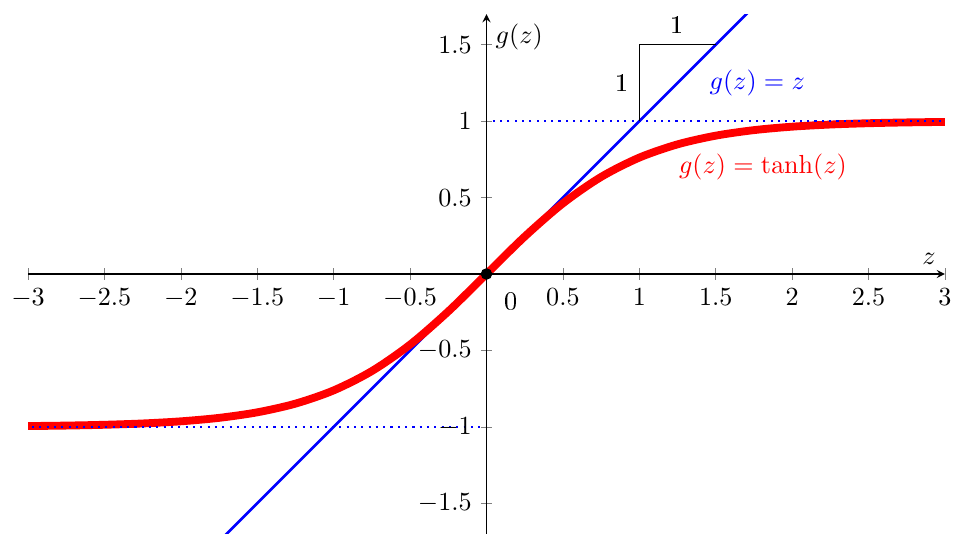}
	\caption{
%		Activation function: 
		\emph{Hyperbolic tangent function} (Section~\ref{sc:activation-functions}):
		$g(z) = \tanh (z) = 2 \sigmoid (2z) - 1$ (red) and its tangent $g(z) = z$ at the coordinate origin (blue), showing that this activation function is {\em identity} for small signals.}
	\label{fig:tanh}
\end{figure}

Prior to the introduction of ReLU, which had been long widely used in neuroscience as activation function prior to 2011,\footnote{
	See, e.g., \cite{Dayan.2001}, p.~14, where ReLU was called the ``half-wave rectification operation'', the meaning of which is explained above in Figure~\ref{fig:halfwave}.
	The logistic sigmoid function (Figure~\ref{fig:sigmoid}) was also used in neuroscience since the 1950s.
} the state-of-the-art for deep-learning activation function was the hyperbolic tangent (Figure~\ref{fig:tanh}), which performed better than the widely used, and much older, sigmoid function\footnote{
	See Section~\ref{sc:sigmoid-history} for a history of the sigmoid function, which dated back at least to 1974 in neuroscience. 
} (Figure~\ref{fig:sigmoid}); see \cite{Glorot.2011:rd0001}, in which it was reported that
\begin{quote}
	``While logistic sigmoid neurons are more biologically plausible than hyperbolic tangent neurons, the latter work better for training multilayer neural networks. Rectifying neurons are an even better model of biological neurons and yield equal or better performance than hyperbolic tangent networks in spite of the hard non-linearity and non-differentiability at zero.''
\end{quote}

The hard non-linearity of ReLU is localized at zero, but otherwise ReLU is a very simple function---identity map for positive argument, zero for negative argument---making it highly efficient for computation.

Also, due to errors in numerical computation, it is rare to hit exactly zero, where there is a hard non-linearity in ReLU: 
\begin{quote}
	``In the case of $g(z) = \max({0, z})$, the left derivative at $z = 0$ is $0$, and the right derivative is $1$.  Software implementations of neural network training usually return one of the one-sided derivatives rather than reporting that the derivative is undefined or raising an error. This may be heuristically justified by observing that gradient-based optimization on a digital computer is subject to numerical error anyway.  When a function is asked to evaluate $g(0)$, it is very unlikely that the underlying value truly was $0$. Instead, it was likely to be some small value that was rounded to 0.'' \cite{Goodfellow.2016}, p.~186.  
	%pdf p.~209
\end{quote}

Thus, in addition to the ability to train deep networks, another advantage of using ReLU is the high efficiency in computing both the layer outputs and the gradients for use in optimizing the parameters (weights and biases) to lower cost or loss, i.e., training; see Section~\ref{sc:training} on Training, and in particular Section~\ref{sc:stochastic-gradient-descent} on Stochastic Gradient Descent.

The activation function ReLU approximates closer to how biological neurons work than other activation functions (e.g., logistic sigmoid, tanh, etc.), as it was established through experiments some sixty years ago, and have been used in neuroscience long (at least ten years) before being adopted in deep learning in 2011.  Its use in deep learning is a clear influence from neuroscience; see Section~\ref{sc:active-function-history} on the history of activation functions, and Section~\ref{sc:ReLU-history} on the history of the rectified linear function.  

Deep-learning networks using ReLU mimic biological neural networks in the brain through a trade-off between two competing properties \cite{Glorot.2011:rd0001}:
\begin{enumerate}
	
	\item
	\emph{Sparsity}.
	Only 1\% to 4\% of brain neurons are active at any one point in time.   Sparsity saves brain energy.  In deep networks, ``rectifying non-linearity gives rise to real zeros of activations and thus truly sparse representations.'' Sparsity provides representation robustness in that the non-zero features\footnote{
		See the definition of image ``predicate'' or image ``feature'' in Section~\ref{sc:static-Rosenblatt}, and in particular Footnote~\ref{fn:features}.
	} would have small changes for small changes of the data. 
	
	\item 
	\emph{Distributivity}.
	Each feature of the data is represented distributively by many inputs, and each input is involved in distributively representing many features.  Distributed representation is a key concept dated since the revival of connectionism with \cite{Rosenblatt.1958} \cite{Block1962a} and others; see Section~\ref{sc:static-Rosenblatt}.
	
\end{enumerate}

\subsubsection{Graphical representation, block diagrams}
\label{sc:block-diagrams}

\begin{figure}[h]
  \centering
  %
  % 2022.12.17
  % remove ".eps" for arXiv
  % \includegraphics[width=0.40\linewidth]{figures/neuron2b.eps}
  \includegraphics[width=0.40\linewidth]{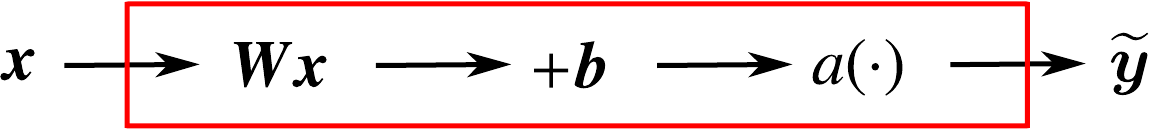}
  \caption{
  	\emph{One-layer network} (Section~\ref{sc:block-diagrams}) representing the relation between the predicted output $\widetilde{\boldsymbol y}$ and the input $\boldsymbol x$, i.e.,  $\widetilde{\boldsymbol y} = f ({\boldsymbol x}) = \g ({\boldsymbol W} {\boldsymbol x} + {\boldsymbol b}) = \g (\boldsymbol{z})$, with the weighted sum $\boldsymbol{z} := {\boldsymbol W} {\boldsymbol x} + {\boldsymbol b}$; see Eq.~(\ref{eq:linearComboInputsBias}) and Eq.~(\ref{eq:activationFunction}) with $\ell = 1$.
  	For a lower-level details of this one layer, see Figure~\ref{fig:neuron1}.
  }
  \label{fig:neuron2}
\end{figure}

\begin{figure}[h]
  \centering
  %
  % 2022.12.17
  % remove ".eps" for arXiv
  % \includegraphics[width=0.5\linewidth]{figures/neuron1b.eps}
  \includegraphics[width=0.5\linewidth]{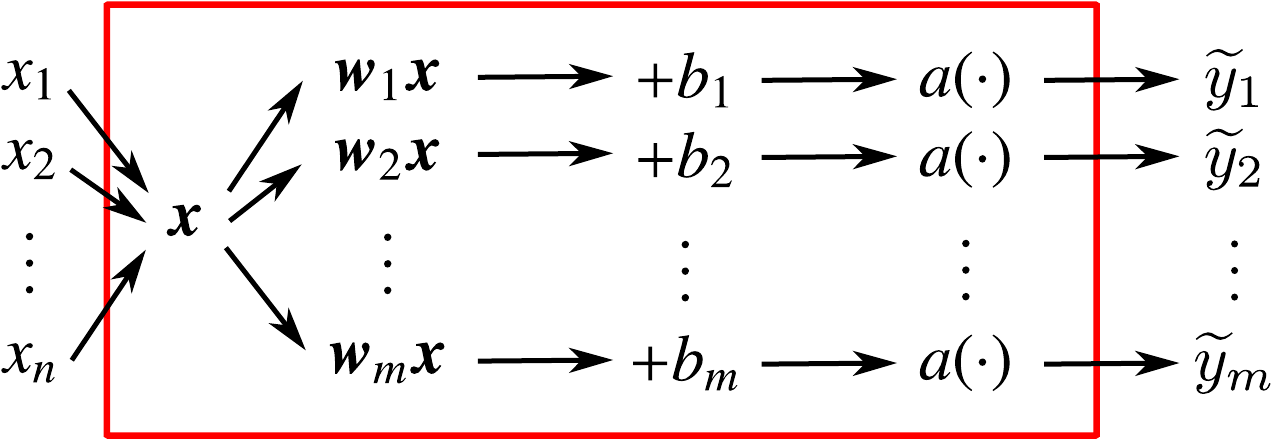}
  \caption{
  	\emph{One-layer network} (Section~\ref{sc:block-diagrams}) 
  	in Figure~\ref{fig:neuron2}: Lower level details, with $m$ processing units (rows or neurons), inputs $\boldsymbol x = [ x_1, x_2, \ldots, x_n ]^T$ and predicted outputs $\widetilde{\boldsymbol y} = [ \widetilde{y}_1, \widetilde{y}_2, \ldots, \widetilde{y}_m ]^T$.
  }
  \label{fig:neuron1}
\end{figure}

The block diagram for a one-layer network is given in Figure~\ref{fig:neuron2}, with more details in terms of the number of inputs and of outputs given in Figure~\ref{fig:neuron1}.

\begin{figure}[h]
  \centering
  %
  % 2022.12.17
  % remove ".eps" for arXiv
  % \includegraphics[width=0.6\linewidth]{figures/network2.eps}
  \includegraphics[width=0.6\linewidth]{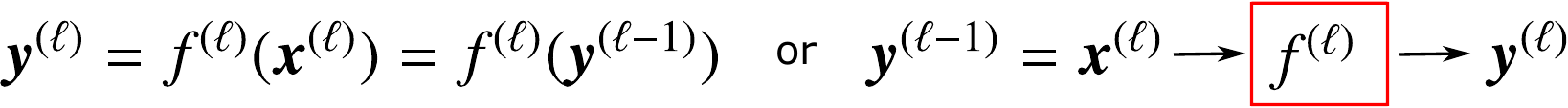}
  \caption{
  	\emph{Input-to-output mapping} (Sections~\ref{sc:block-diagrams}, \ref{sc:artificial-neuron}):
  	Layer $(\ell)$ in network with $L$ layers in Figure~\ref{fig:network3b}, input-to-output mapping $f^{(\ell)}$ for layer $(\ell)$.
  }
  \label{fig:network2}
\end{figure}

\begin{figure}[h]
  \centering
  %
  % 2022.12.17
  % remove ".eps" for arXiv
  % \includegraphics[width=0.65\linewidth]{figures/neuron4b.eps}
  \includegraphics[width=0.65\linewidth]{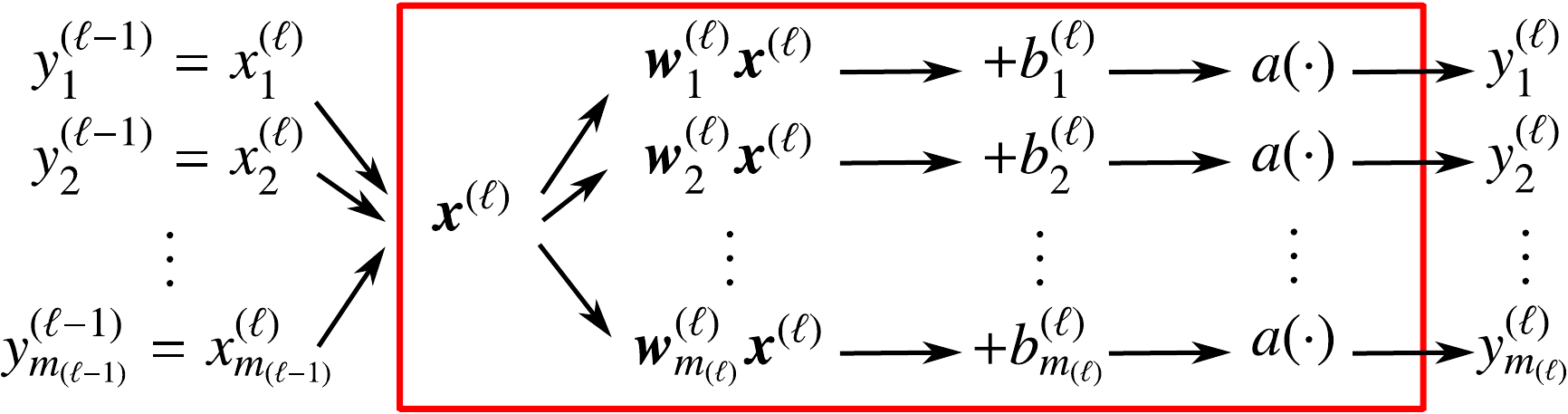}
  \caption{
  	\emph{Low-level details of layer} $(\ell)$ (Sections~\ref{sc:block-diagrams}, \ref{sc:artificial-neuron}) of the multilayer neural network in Figure~\ref{fig:network3b}, with $m_\ell$ as the number of processing units (rows or neurons), and thus the width of this layer, representing the layer processing (input-to-output) function $f^{(\ell)}$ in Figure~\ref{fig:network2}.
  %The number of processing units $m_\ell$ is the width of this layer.
  }
  \label{fig:neuron4}
\end{figure}

For a multilayer neural network with $L$ layers, with input-output relation shown in Figure~\ref{fig:network2}, the detailed components are given in Figure~\ref{fig:neuron4}, which generalizes Figure~\ref{fig:neuron1} to layer $(\ell)$. 


\subsubsection{Artificial neuron}
\label{sc:artificial-neuron}

\begin{figure}[h]
  \centering
  %
  % 2022.12.17
  % remove ".eps" for arXiv
  % \includegraphics[width=0.50\linewidth]{figures/neuron5b.eps}
  \includegraphics[width=0.50\linewidth]{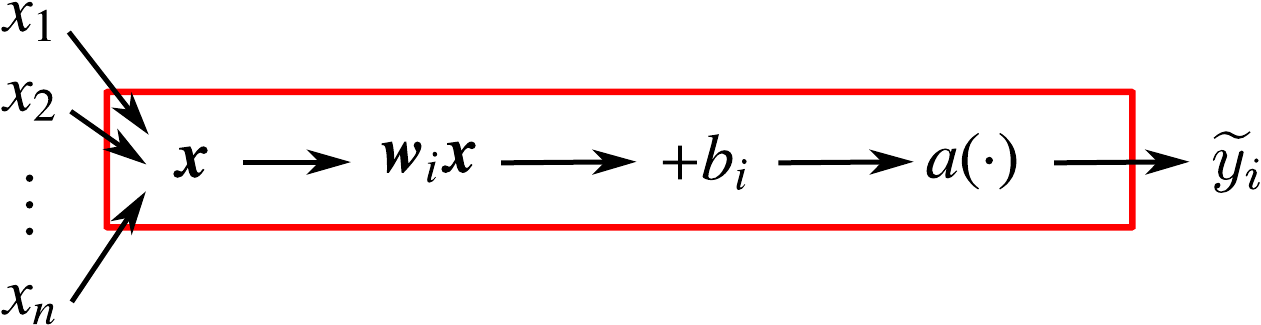}
  \caption{
  	\emph{Artificial neuron} (Sections~\ref{sc:Oishi-summary}, \ref{sc:artificial-neuron}, \ref{sc:inspired-from-biology}), row $i$ in layer $(\ell)$ in Figure~\ref{fig:neuron4}, representing the multiple-inputs-to-single-output relation $\widetilde{y}_i = \g( \boldsymbol w_i \boldsymbol x + b_i) = \g ( \sum_{i=1}^n w_{ij} x_j + b_i ) $ with $\boldsymbol x  = [x_1 , x_2, \cdots, x_n ]^T$ and $\boldsymbol w_i = [ w_{i1} , w_{i2} , \cdots , w_{i n} ]$.  This block diagram is the exact equivalent of Figure~\ref{fig:Oishi-neuron}, Section~\ref{sc:Oishi-summary}, and in \cite{Oishi.2017:rd9648}.
  	See Figure~\ref{fig:bio-neuron} for the corresponding biological neuron in Section~\ref{sc:inspired-from-biology} on ``Early inspiration from biological neurons''.
  }
  \label{fig:neuron5}
\end{figure}

\begin{figure}[H]
	\centering
	%
	% 2022.12.17
	% add "-eps-converted-to.pdf" for arXiv
	% \includegraphics[width=0.3\linewidth]{figures/network5a-2}
	\includegraphics[width=0.3\linewidth]{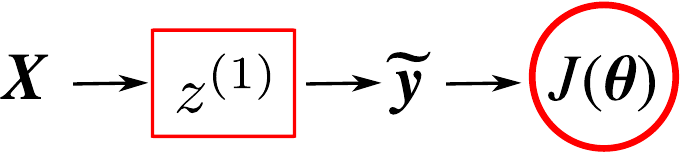}
	\caption{
		\emph{Representing XOR function} (Sections~\ref{sc:XORfunction}, \ref{sc:linear-combo-history}). This one-layer network (which is not the Rosenblatt perceptron in Figure~\ref{fig:perceptron}) cannot perform this task.  For each input matrix $\bx_{i}$ in the design matrix $\boldsymbol{X} = [ \bx_{1} , \ldots , \bx_{4} ] \in \real^{2 \times 4}$, with $i=1,\ldots,4$ (see Table~\ref{tb:XOR-values}), the linear unit (neuron) $z^{(1)} (\bx_{i}) = \boldsymbol{w} \bx_{i} + b \in \real$ in Eq.~(\ref{eq:XOR-linear-1}) predict a value $\widetilde{y}_i = z^{(1)} (\bx_{i})$ as output, which is collected in the output matrix $\widetilde{\boldsymbol{y}} = [\widetilde{y}_1 , \ldots , \widetilde{y}_4] \in \real^{1 \times 4}$.  The MSE cost function $J(\boldsymbol{\theta})$ in Eq.~(\ref{eq:MSE}) is used in a gradient descent to find the parameters $\boldsymbol{\theta} = [\boldsymbol{w} , b]$.  The result is a constant function, $\widetilde{y}_i = \frac12$, for $i=1,\ldots,4$, which cannot represent the XOR function.
	}
	\label{fig:XOR-one-layer}
\end{figure}

And finally, we now complete our \emph{top-down} descent from the big picture of the overall multilayer neural network with $L$ layers in Figure~\ref{fig:network3b}, through Figure~\ref{fig:network2} for a typical layer $(\ell)$ and Figure~\ref{fig:neuron4} for the lower-level details of layer $(\ell)$, then down to the most basic level, a neuron in Figure~\ref{fig:neuron5} as one row in layer $(\ell)$ in Figure~\ref{fig:neuron4}.

\subsection{Representing XOR function with two-layer network}
\label{sc:XORfunction}

The XOR (exclusive-or) function played an important role in bringing down the first wave of AI, known as the cybernetics wave (\cite{Goodfellow.2016}, p.~14) since 
%
% CMES style, rewriting
%\cite{Minsky.1969} showed 
it was shown in \cite{Minsky.1969}
that Rosenblatt's perceptron (1958 \cite{Rosenblatt.1958}, 1962 \cite{Block1962a} \cite{Rosenblatt.1962}) could not represent the XOR function, defined in Table~\ref{tb:XOR-values}:
% \cite{Minsky.1988} 
\begin{table}[h]
	\caption{
		\emph{Exclusive-or (XOR) function} (Section~\ref{sc:XORfunction}) produces the {\em True} value only if two arguments are different.  The symbol $\oplus$ (``Oh-plus'') denotes the XOR operator.  A concrete example of the XOR function would be that there is one and only one of two poker player would be the winner, and there is no tie possible, i.e., both players cannot win, and both cannot lose.
		% race runners 
		% runners 
	}
	\centering
	% https://tex.stackexchange.com/questions/31672/column-and-row-padding-in-tables
	{\renewcommand{\arraystretch}{1.5}
		\begin{tabular}{|c|c|c|}
			\hline
			% $i$ & $\boldsymbol x_i = (x_{i1} , x_{i2})$ & $y_i = f (\boldsymbol x_i) = x_{i1} \oplus x_{i2}$ (XOR)
			$j$ & $\boldsymbol x_j = [ x_{1j} , x_{2j} ]^T$ & $y_i = f (\boldsymbol x_j) = x_{1j} \oplus x_{2j}$ (XOR)
			% \rule{0pt}{0.3em}
			%\\[0.3em]
			\\
			\hline
			\hline
			% 1 & (0,0) & 0
			1 & $[0,0]^T$ & 0
			\\
			2 & $[1,0]^T$ & 1
			\\
			3 & $[0,1]^T$ & 1
			\\
			4 & $[1,1]^T$ & 0
			\\
			\hline
		\end{tabular}
	}
	\label{tb:XOR-values}
\end{table}

\noindent
The dataset or design matrix\footnote{
	See \cite{Goodfellow.2016}, p.~103.
} $\boldsymbol{X}$ is the collection of the coordinates of all four points in Table~\ref{tb:XOR-values}:
\begin{align}
	\boldsymbol{X}
	=
	[ \bx_{1} , \ldots , \bx_{4} ]
	=
	\left[
		\begin{array}{c c c c}
			0 & 1 & 0 & 1
			\\
			0 & 0 & 1 & 1
		\end{array}
	\right]
	\in \real^{2 \times 4}
	\ , \text{ with }
	\bx_{i} 
	\in \real^{2 \times 1}
	\label{eq:design-matrix-1}
\end{align}
\noindent
An approximation (or prediction) for the XOR function $y = f (\boldsymbol x)$ with $\boldsymbol \theta$ parameters is denoted by $\nfwidetilde{y} = \nfwidetilde{f} (\boldsymbol x , \boldsymbol \theta)$, with mean squared error (MSE) being:
\begin{align}
	J(\boldsymbol \theta) 
	= 
	\frac{1}{4} 
	\sum_{i=1}^4 
	\left[ 
	 	\nfwidetilde{f} (\boldsymbol x_i , \boldsymbol \theta) - f (\boldsymbol x_i) 
	\right]^2
	=
	\frac{1}{4} 
	\sum_{i=1}^4 
	\left[ \nfwidetilde{y}_i - y_i \right]^2
	\label{eq:MSE}
\end{align}

%\newtheorem{assu}{Assumption}

%\begin{assu}
%hello
%\end{assu}
\noindent
We begin with a one-layer network to show that it cannot represent the XOR function,\footnote{
	 This one-layer network is not the Rosenblatt perceptron in Figure~\ref{fig:perceptron} due to the absence of the Heaviside function as activation function, and thus Section~\ref{sc:XOR-one-layer} is not the proof that the Rosenblatt perceptron cannot represent the XOR function.  For such proof, see \cite{Minsky.1969}. 
} then move on to a two-layer network, which can. 

\subsubsection{One-layer network}
\label{sc:XOR-one-layer}

Consider the following one-layer network,\footnote{
	See \cite{Goodfellow.2016}, p.~167.
} in which the output $\nfwidetilde{y}$ is a linear combination of the coordinates $(x_1 , x_2)$ as inputs:
\begin{align}
	\nfwidetilde{y}
	=
	f
	(\boldsymbol x , \boldsymbol \theta) 
	=
	f^{(1)} (\boldsymbol x)
	=
	\bweightsp{1}{1} \boldsymbol x + \biassp{1}{1}
	=
	\boldsymbol w \boldsymbol x + b
	=
	% \sum_{i=1}^2 w_i \cdot x_i + b
	w_1 x_1 + w_2 x_2 + b
	\ ,
	\label{eq:XOR-linear-1}
\end{align}
with the following matrices
\begin{align}
	\bweightsp{1}{1}
	=
	\boldsymbol{w}
	=
	\left[ w_1 , w_2 \right]
	\in \real^{1 \times 2}
	\ , \quad
	\bx 
	= 
	\left[ x_1 , x_2 \right]^T
	\in \real^{2 \times 1}
	\ , \quad
	\biassp{1}{1} = b 
	\in \real^{1 \times 1}
\end{align}
\begin{align}
	\boldsymbol \theta
	=
	\left[
	\theta_1 , \theta_2 , \theta_3
	\right]^T
	=
	\left[
	w_1 , w_2 , b
	\right]^T
\end{align}
since 
%
% CMES style, rewriting
%\cite{Goodfellow.2016}, p.~14, wrote:
it is written in \cite{Goodfellow.2016}, p.~14:
\begin{quote}
	``Model based on the $f(\boldsymbol x, \boldsymbol{w}) = \sum_i w_i x_i$ used by the perceptron and ADALINE are called linear models.  Linear models have many limitations... Most famously, they cannot learn the XOR function... Critics who observed these flaws in linear models caused a backlash against biologically inspired learning in general (Minsky and Papert, 1969). This was the first major dip in the popularity of neural networks.''
\end{quote}
First-time learners, who have not seen the definition of 
%
% CMES style, rewriting
%\cite{Rosenblatt.1958}'s 
Rosenblatt's (1958) perceptron \cite{Rosenblatt.1958}, could confuse Eq.~(\ref{eq:XOR-linear-1}) as the perceptron---which was not a linear model, but more importantly the Rosenblatt perceptron was a network with many neurons\footnote{
	See Section~\ref{sc:linear-combo-history} on the history of the linear combination (weighted sum) of inputs with biases.}---because Eq.~(\ref{eq:XOR-linear-1}) is only a linear unit (a single neuron), and does not have an (nonlinear) activation function.  
A neuron in the Rosenblatt perceptron is Eq.~(\ref{eq:perceptron}) in Section~\ref{sc:linear-combo-history}, with the Heaviside (nonlinear step) function as activation function; see Figure~\ref{fig:perceptron}.
The MSE cost function in Eq.~(\ref{eq:MSE}) becomes
\begin{align}
  J (\boldsymbol \theta) 
  = 
  \frac{1}{4} 
  \sum_{i=1}^4 
  \left[ 
  b^2 + (1 - w_1 - b)^2 + (1 - w_2 - b)^2 + (w_1 + w_2 + b)^2
  \right]^2
  \label{eq:XOR-cost-1}
\end{align}
Setting the gradient of the cost function in Eq.~(\ref{eq:XOR-cost-1}) to zero and solving the resulting equations, we obtain the weights and the bias:
\begin{align}
	\nabla_{\boldsymbol \theta} J (\boldsymbol \theta) 
    =
    \left[
      \frac{\partial J}{\theta_1} , 
      \frac{\partial J}{\theta_2} , 
      \frac{\partial J}{\theta_3}
    \right]^T
	=
    \left[
      \frac{\partial J}{\partial w_1} , 
      \frac{\partial J}{\partial w_2} , 
      \frac{\partial J}{\partial b}
    \right]^T
\end{align}
\begin{align}
	% \label{eq:normalEqs}
	\left.
	\begin{array}{l}
	\displaystyle
	\frac{\partial J}{\partial w_1} = 0 \Longrightarrow 2 w_1 + w_2 + 2b = 1
	\vspace{2mm}
	\\
	\displaystyle
	\frac{\partial J}{\partial w_2} = 0 \Longrightarrow w_1 + 2 w_2 + 2b = 1
	\vspace{2mm}
	\\
	\displaystyle
	\frac{\partial J}{\partial b} = 0 \Longrightarrow   w_1 + w_2 + 2b = 1
	\end{array}  
	\right\rbrace
	\Longrightarrow
	w_1 = w_2 = 0 , \quad b = \frac12
	\ ,
	\label{eq:XOR-params-1}
\end{align}
from which the predicted output $\nfwidetilde{y}$ in Eq.~(\ref{eq:XOR-linear-1}) is a constant for any points in the dataset (or design matrix) $\boldsymbol{X} = [ \bx_{1} , \ldots , \bx_{4}]$:
\begin{align}
	\nfwidetilde{y}_i = f (\bx_{i} , \bparam) = \frac12 
	\ , \text{ for } i = 1, \ldots, 4
\end{align}
and thus this one-layer network cannot represent the XOR function.
Eqs.~(\ref{eq:XOR-params-1}) are called the ``normal'' equations.\footnote{
	In least-square linear regression, the normal equations are often presented in matrix form, starting from the errors (or residuals) at the data points, gathered in the matrix 
	$
	\boldsymbol e 
	= \boldsymbol y - \boldsymbol X \boldsymbol \theta
	$.
	To minimize the squared of the errors represented by
	$
	\parallel \boldsymbol e \parallel^2 
	$, consider a perturbation
	$
	\boldsymbol \theta_\epsilon = \boldsymbol \theta + \epsilon \gamma
	$ and
	$
	\boldsymbol e_\epsilon 
	= \boldsymbol y - \boldsymbol X \boldsymbol \theta_\epsilon 
	$, then set  the directional derivative of 
	$
	\parallel \boldsymbol e \parallel^2 
	$ to zero, i.e.,
	$
	\left.
	\frac{d}{d \epsilon}
	\parallel \boldsymbol e_\epsilon \parallel^2 
	\right|_{\epsilon=0}
	=
	\boldsymbol X^T (\boldsymbol d - \boldsymbol X \boldsymbol \theta)
	= 0
	$, which is the ``normal equation'' in matrix form, since the error matrix
	$
	\boldsymbol e
	$ is required to be ``normal'' (orthogonal) to the span of 
	$
	\boldsymbol X
	$.
	For the above XOR function with four data points, the relevant matrices are (using the Matlab / Octave notation)
	$
	\boldsymbol e = [ e_1 , e_2 , e_3 , e_4 ]^T
	$,
	$
	\boldsymbol y = [0, 1, 1, 0]^T
	$, and
	$
	\boldsymbol X = [[0,0,1]; [1,0,1]; [0,1,1]; [1,1,1]]
	$, which also lead to 
	% Eq.~(\ref{eq:normalEqs}).
	Eq.~(\ref{eq:XOR-params-1}).
	See, e.g., \cite{Herzberger.1949} 
	\cite{Weisstein.normalEq} and
	\cite{Goodfellow.2016}, p.~106.
}

\subsubsection{Two-layer network}

\begin{figure}[H]
	\centering
	%
	% 2022.12.17
	% add "-eps-converted-to.pdf" for arXiv
	% \includegraphics[width=0.3\linewidth]{figures/network5b}
	\includegraphics[width=0.3\linewidth]{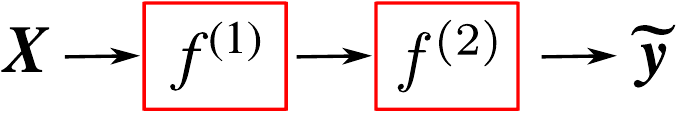}
	\\[1em]
	%
	% 2022.12.17
	% add "-eps-converted-to.pdf" for arXiv
	% \includegraphics[width=0.6\linewidth]{figures/network5c}
	\includegraphics[width=0.6\linewidth]{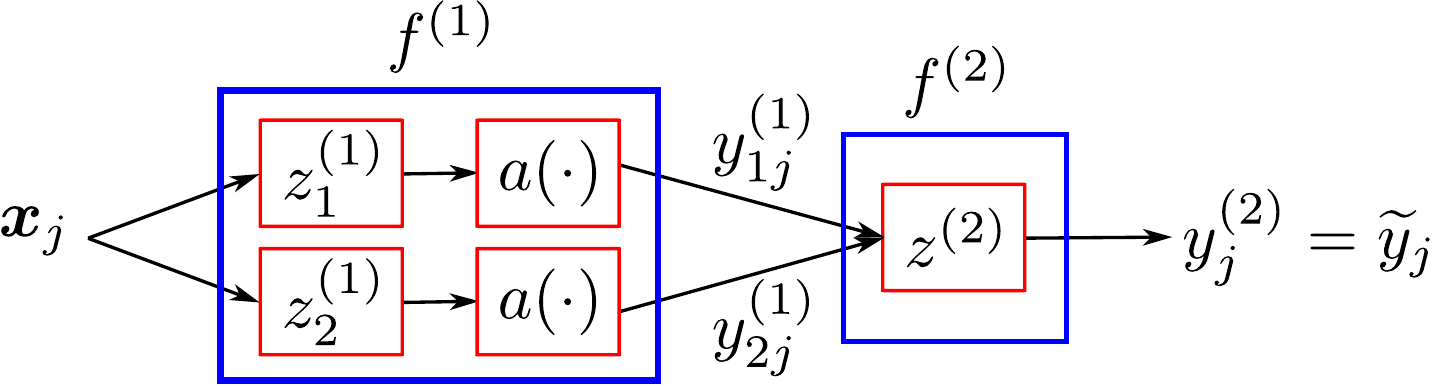}
	\caption{
		\emph{Representing XOR function} (Sections~\ref{sc:XORfunction}). 
		This two-layer network can perform this task.  The four points in the design matrix $\boldsymbol{X} = [ \bx_{1} , \ldots , \bx_{4} ] \in \real^{2 \times 4}$ (see Table~\ref{tb:XOR-values}) are converted into three points that are linearly separable by the two nonlinear units (neurons or rows) of Layer (1), i.e., $\bYp{1} = [ \bysp{1}{1} , \ldots , \bysp{1}{4} ] = f^{(1)} (\bXp{1}) = a(\bZp{1}) = \bXp{2}  \in \real^{2 \times 4}$, with $\bZp{1} = \bweightp{1} \bXp{1} + \bBiasp{1}  \in \real^{2 \times 4}$, as in Eq.~(\ref{eq:XOR-outputs-layer-1b}), and $a(\cdot)$ a nonlinear activation function.
		Layer (2) consists of a single linear unit (neuron or row) with three parameters, i.e., $\widetilde{\by} = [\widetilde{y}_1 , \ldots , \widetilde{y}_4] = f^{(2)} (\bXp{2}) = \bweightsp{1}{2}\bXp{2} + \biassp{1}{2}  \in \real^{1 \times 4}$.
		The three non-aligned points in $\bXp{2}$ offer three equations to solve for the three parameters $\bparamp{2} = [\bweightsp{1}{2} , \biassp{1}{2}]  \in \real^{1 \times 3}$; see Eq.~(\ref{eq:XOR-params-Eqs}).
	}
	\label{fig:XOR-two-layer}
\end{figure}

\begin{figure}[H]
	\centering
	%
	% 2022.12.17
	% add "-eps-converted-to.pdf" for arXiv
	% \includegraphics[width=0.7\linewidth]{figures/XOR-1}
	\includegraphics[width=0.7\linewidth]{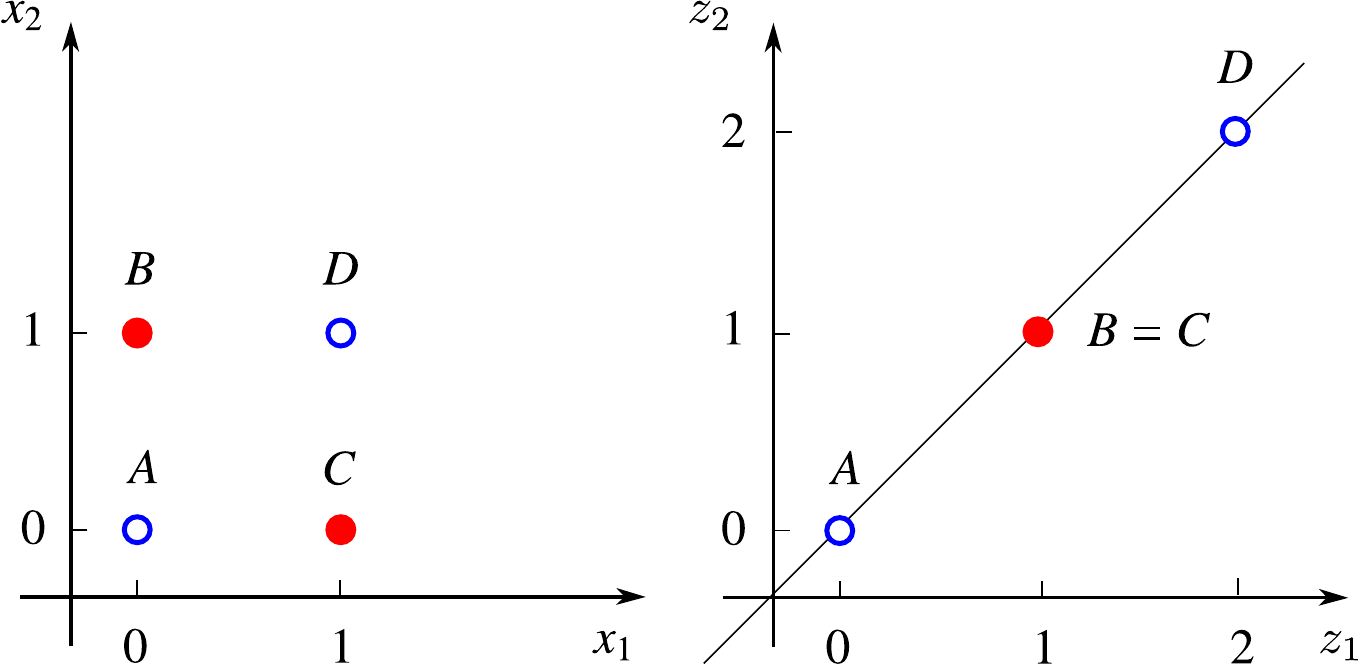}
	\caption{
		\emph{Two-layer network for XOR representation} (Sections~\ref{sc:XORfunction}).
		{\em Left}: XOR function, with 
		$A = \bxsp{1}{1} = [0, 0]^T$, $B = \bxsp{2}{1} = [0, 1]^T$, 
		$C = \bxsp{3}{1} = [1, 0]^T$, $D = \bxsp{4}{1} = [1, 1]^T$; see Eq.~(\ref{eq:XOR-Z-X-layer-1}).
		The XOR value for the solid red dots is 1, and for the open blue dots 0.
		{\em Right}: Images of points $A , B , C , D$ in the $z$-plane due {\em only} to the first term of 
		%
		% wrong equation number
		% Eq.~(\ref{eq:XOR-outputs-layer-1b}),
		Eq.~\eqref{eq:Z-layer-1}, 
		i.e., $\bweightp{1} \bXp{1}$, which is shown in Eq.~(\ref{eq:XOR-layer-1-first-term}).
		See also Figure~\ref{fig:XOR-two-layer-2}.
	}
	\label{fig:XOR-two-layer-1}
\end{figure}

\begin{figure}[H]
	\centering
	%
	% 2022.12.17
	% add "-eps-converted-to.pdf" for arXiv
	% \includegraphics[width=0.7\linewidth]{figures/XOR-2}
	\includegraphics[width=0.7\linewidth]{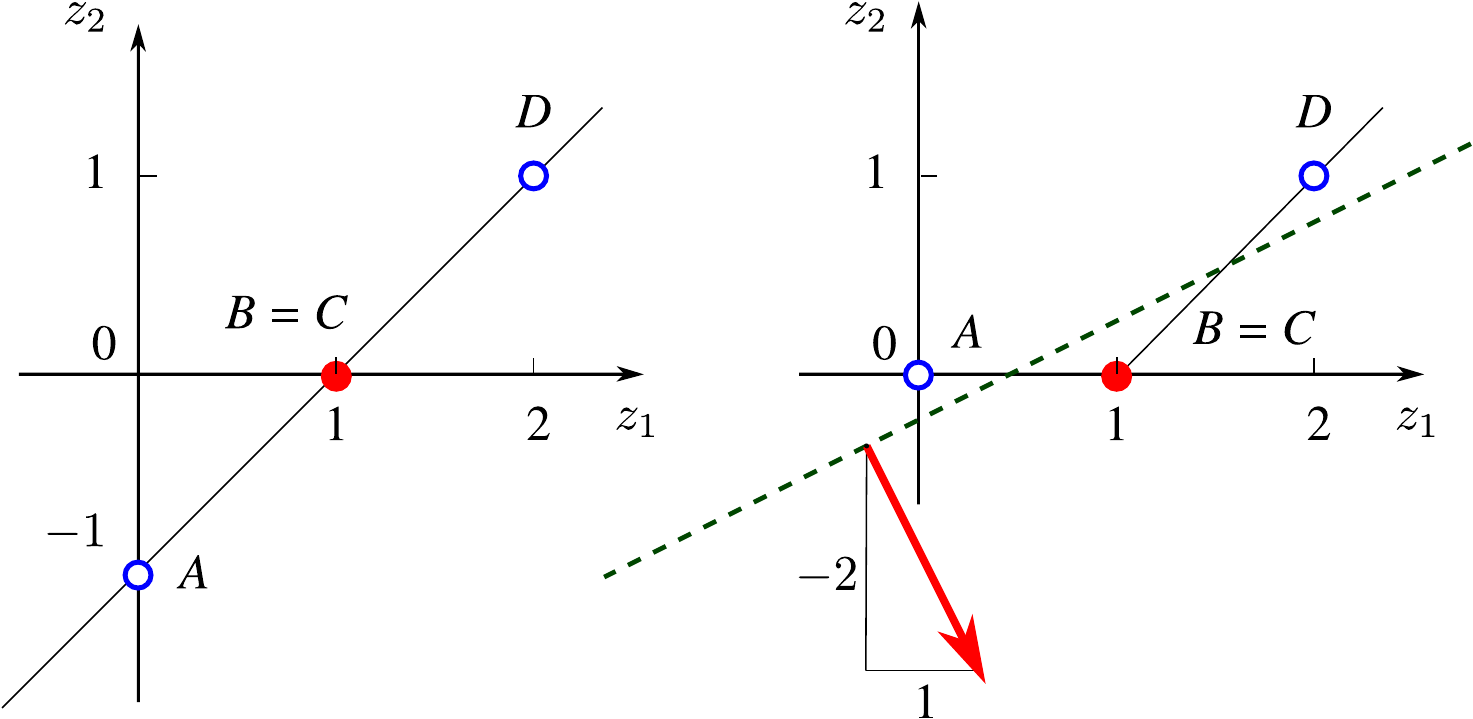}
	\caption{
		\emph{Two-layer network for XOR representation} (Sections~\ref{sc:XORfunction}).
		{\em Left}: Images of points $A , B , C , D$ of $\bZp{1}$ in Eq.~(\ref{eq:bZ1}), obtained after a translation by adding the bias $\bbiasp{1} = [0 , -1]^T$ in Eq.~(\ref{eq:XOR-weight-bias-layer-1}) to the same points $A , B , C , D$ in the right subfigure of Figure~\ref{fig:XOR-two-layer-1}.
		The XOR value for the solid red dots is 1, and for the open blue dots 0.
		{\em Right}: Images of points $A , B , C , D$ after applying the ReLU activation function, which moves point $A$ to the origin; see Eq.~(\ref{eq:XOR-outputs-layer-1}). the points $A$, $B$ $(=C)$, $D$ are no longer aligned, and thus linearly separable by the green dotted line, whose normal vector has the components $[1 , -2]^T$, which are the weights shown in Table~\ref{tb:XOR-params}.
	}
	\label{fig:XOR-two-layer-2}
\end{figure}

The four points in Table~\ref{tb:XOR-values} are not linearly separable, i.e., there is no straight line that separates these four points such that the value of the XOR function is zero for two points on one side of the line, and one for the two points on the other side of the line.  One layer could not represent the XOR function, as shown above.
%
% CMES style rewriting
Rosenblatt (1958)
\cite{Rosenblatt.1958} wrote:
\begin{quote}
	``It has, in fact, been widely conceded by psychologists that  there  is little  point  in trying  to `disprove' any  of the  major learning theories  in use today, since by  extension, or a change in parameters,  they have  all  proved  capable  of  adapting to any specific  empirical data.  In considering this approach, one is reminded of  a remark attributed to Kistiakowsky, that {\em`given  seven  parameters,  I could fit an elephant}.' ''
\end{quote}
So we now add a second layer, and thus more parameters in the hope to be able to represent the XOR function, as shown in Figure~\ref{fig:XOR-two-layer}.\footnote{
	Our presentation is more detailed and more general than in \cite{Goodfellow.2016}, pp.~167-171, where there was no intuitive explanation of how the numbers were obtained, and where only the activation function ReLU was used. 
}
{\bf Layer (1):} six parameters (4 weights, 2 biases), plus a (nonlinear) activation function.  The purpose is to change coordinates to move the four input points of the XOR function into three points, such that the two points with XOR value equal 1 are coalesced into a single point, and such that these three points are aligned on a straight line.   Since these three points remain not linearly separable, the activation function then moves these three points out of alignment, and thus linearly separable. 
\begin{align}
	\bzsp{i}{1}
	=
	\bweightsp{i}{1}
	\bx_{i}
	+
	\bbiassp{i}{1}
	=
	\bweightp{1}
	\bx_{i}
	+
	\bbiasp{1}
	\ , \text{ for }
	i = 1, \ldots, 4
\end{align}
\begin{align}
	\bweightsp{i}{1}
	=
	\bweightp{1}
	=
	\begin{bmatrix}
    	1 & 1
    	\\
    	1 & 1
	\end{bmatrix}
	\ , \quad
	\bbiassp{i}{1}
	=
	\bbiasp{1}
	=
	\begin{bmatrix}
		0
		\\
		-1
	\end{bmatrix}
	\ , \text{ for }
	i = 1, \ldots, 4
	\label{eq:XOR-weight-bias-layer-1}
\end{align}
\begin{align}
	\bZp{1}
	=
	\left[
		\bzsp{1}{1}
		, \ldots ,
		\bzsp{4}{1}
	\right]
	\in \real^{2 \times 4}
	\ , \quad
	\bXp{1}
	=
	\left[
		\bxsp{1}{1}
		, \ldots ,
		\bxsp{4}{1}
	\right]
	=
	\left[
		\begin{array}{c c c c}
			0 & 1 & 0 & 1
			\\
			0 & 0 & 1 & 1
		\end{array}
	\right]
	\in \real^{2 \times 4}
	\label{eq:XOR-Z-X-layer-1}
\end{align}
\begin{align}
	\bBiasp{1}
	=
	\left[
		\bbiassp{1}{1}
		, \ldots ,
		\bbiassp{4}{1}
	\right]
	=
	\left[
	\begin{array}{r r r r}
		0  & 0  & 0  & 0
		\\
		-1 & -1 & -1 & -1
	\end{array}
	\right]
	\in \real^{2 \times 4}
	\label{eq:XOR-Bias}
\end{align}
\begin{align}
	\bZp{1}
	=
	\bweightp{1}
	\bXp{1}
	+
	\bBiasp{1}
	\in \real^{2 \times 4}
	\ .
	\label{eq:Z-layer-1}
\end{align}
To map the two points $\bx_2 = [1 , 0]^T$ and $\bx_3 = [0 , 1]^T$, at which the XOR value is 1, into a single point, the two rows of $\bweightp{1}$ are selected to be identically $[1 , 1]$ as shown in Eq.~(\ref{eq:XOR-weight-bias-layer-1}).  The first term in Eq.~(\ref{eq:Z-layer-1}) yields three points aligned along the bisector in the first quadrant (i.e., the line $z_2 = z_1$ in the z-plane), with all positive coordinates, Figure~\ref{fig:XOR-two-layer-1}:
\begin{align}
	\bweightp{1} \bXp{1}
	=
	\left[
		\begin{array}{c c c c}
			0 & 1 & 1 & 2
			\\
			0 & 1 & 1 & 2
		\end{array}
	\right]
	\in \real^{2 \times 4}
	\ .
	\label{eq:XOR-layer-1-first-term}
\end{align}
For activation functions such as ReLu or Heaviside\footnote{
	In general, the Heaviside function is not used as activation function since its gradient is zero, and thus would not work for gradient descent.  But for this XOR problem {\em without} using gradient descent, the Heaviside function offers a workable solution as the rectified linear function.
} to have any effect, the above three points are next translated in the negative $z_2$ direction using the biases in Eq.~(\ref{eq:XOR-Bias}), so that Eq.~(\ref{eq:Z-layer-1}) yields:
\begin{align}
	\bZp{1}
	=
	\left[
	\begin{array}{c c c c}
		0 & 1 & 1 & 2
		\\
		-1 & 0 & 0 & 1
	\end{array}
	\right]
	\in \real^{2 \times 4}
	\ ,
	\label{eq:bZ1}
\end{align}
and thus
\begin{align}	
	\bYp{1}
	=
	\left[
		\bysp{1}{1} , \ldots , \bysp{4}{1}
	\right]
	=
	a(\bZp{1})
	=
	\left[
		\begin{array}{c c c c}
			0 & 1 & 1 & 2
			\\
			0 & 0 & 0 & 1
		\end{array}
	\right]
	=
	\bXp{2}
	=
	\left[
		\bxsp{1}{2} , \ldots , \bxsp{4}{2}
	\right]
	\in \real^{2 \times 4}
	\ ,
	\hfill
	\label{eq:XOR-outputs-layer-1}
\end{align}
For general activation function $a(\cdot)$, the outputs of Layer (1) are:
\begin{align}
	\bYp{1} = \bXp{2} = a(\bZp{1})
	=
	\left[
		\begin{array}{c c c c}
			a(0) & a(1) & a(1) & a(2)
			\\
			a(-1) & a(0) & a(0) & a(1)
		\end{array}
	\right]
	\in \real^{2 \times 4}
	\ .
	\label{eq:XOR-outputs-layer-1b}
\end{align}

{\bf Layer (2):} three parameters (2 weights, 1 bias), no activation function.  Eq.~(\ref{eq:XOR-linear-2}) for this layer is identical to Eq.~(\ref{eq:XOR-linear-1}) for the one-layer network above, with the output $\byp{1}$ of Layer (1) as input $\bxp{2} = \byp{1}$, as shown in Eq.~(\ref{eq:XOR-outputs-layer-1}):  
\begin{align}
	\nfwidetilde{y}_j
	=
	f
	(\bxsp{j}{2}, \bparamp{2}) 
	=
	f^{(2)} (\bxsp{j}{2})
	=
	\bweightsp{1}{2} \bxsp{j}{2} + \biassp{1}{2}
	% =
	% \bweightp{2} \byp{1} + \biassp{1}{2}
	=
	\boldsymbol w \bxsp{j}{2} + b
	=
	% \sum_{i=1}^2 w_i \cdot x_i + b
	w_1 \xsp{1j}{2} + w_2 \xsp{2j}{2} + b
	\ ,
	\label{eq:XOR-linear-2}
\end{align}
with three distinct points in Eq.~(\ref{eq:XOR-outputs-layer-1}), because $\bxsp{2}{2} = \bxsp{3}{2} = [1 , 0]^T$, to solve for these three parameters:
\begin{align}
	\bparamp{2}
	=
	[ \bweightsp{1}{2} , \biassp{1}{2} ]
	=
	[ \weight_{1} , \weight_{2} , b]
	\label{eq:XOR-params-2}
\end{align}  
We have three equations:
\begin{align}
	\begin{bmatrix}
		\nfwidetilde{y}_1 \\ \nfwidetilde{y}_2 \\ \nfwidetilde{y}_4
	\end{bmatrix}
	=
	\begin{bmatrix}
		a(0) & a(-1) & 1
		\\
		a(1) & a(0) & 1
		\\
		a(2) & a(1) & 1
	\end{bmatrix}
	\begin{bmatrix}
		w_1 \\ w_2 \\ b
	\end{bmatrix}
	=
	\begin{bmatrix}
		\y_{1} \\ \y_{2} \\ \y_{4}
	\end{bmatrix}
	=
	\begin{bmatrix}
		0 \\ 1 \\ 0
	\end{bmatrix}
	\ ,
	\label{eq:XOR-params-Eqs}
\end{align} 
for which the exact analytical solution for the parameters $\bparamp{2}$ is easy to obtain, but the expressions are rather lengthy.  Hence, here we only give the numerical solution for $\bparamp{2}$ in the case of the logistic sigmoid function in Table~\ref{tb:XOR-params}.
\begin{table}[H]
	\caption{
		\emph{Two-layer network for XOR representation} (Section~\ref{sc:XORfunction}).
		Values of parameters $\bparamp{2}$ in Eq.~(\ref{eq:XOR-params-2}).
		The results are exact for ReLU and Heaviside, but rounded for sigmoid due to the irrational Euler's number $e$.
		See Figure~\ref{fig:XOR-two-layer-2}.
	}
	\centering
	% https://tex.stackexchange.com/questions/31672/column-and-row-padding-in-tables
	{\renewcommand{\arraystretch}{1.5}
		\begin{tabular}{|c|c|}
			\hline
			Activation function & Parameters $\bparamp{2}$
			\\
			\hline
			\hline
			ReLU & $[1 , -2 , \phantom{-} 0 ]^T$
			\\
			Heaviside & $[1 , -1 , \phantom{-} 0 ]^T$
			\\
			Sigmoid & $[24.5942 , -20.2663 , -6.8466 ]^T$
			\\
			\hline
		\end{tabular}
	}
	\label{tb:XOR-params}
\end{table}

We conjecture that any (nonlinear) function $a(\cdot)$ in the zoo of activation functions listed, e.g., in ``Activation function'', Wikipedia,  \href{https://en.wikipedia.org/w/index.php?title=Activation_function&oldid=897708534}{version 21:00, 18 May 2019} or in \cite{Ramachandran.2017:rd0001} (see Figure~\ref{fig:swish}), would move the three points in $\bZp{1}$ in Eq.~(\ref{eq:bZ1}) out of alignment, and thus provide the corresponding unique solution $\bparamp{2}$ for Eq.~(\ref{eq:XOR-params-Eqs}).  
\begin{rem}
	\label{rm:flying-elephant}
	Number of parameters.
	{\rm	
		In 1953, Physicist Freeman Dyson (Princeton Institute of Advanced Study) once consulted with Nobel Laureate Enrico Fermi about a new mathematical model for a difficult physics problem that Dyson and his students had just developed.  Fermi asked Dyson how many parameters they had.  ``Four'', Dyson replied.   Fermi then gave his now famous comment ``I remember my friend Johnny von Neumann used to say, with four parameters I can fit an elephant, and with five I can make him wiggle his trunk'' \cite{Dyson.2004}.
		
		But it was only more than sixty years later that physicists were able to plot an elephant in 2-D using a model with four complex numbers as parameters \vphantom{\cite{Mayer.2010}}\cite{Mayer.2010}.  
		
		With nine parameters, the elephant can be made to walk (representing the XOR function), and with a billion parameters, it may even perform some acrobatic maneuver in 3-D; see Section~\ref{sc:depth-size} on depth of multilayer networks. 
	}
	$\hfill\blacksquare$
\end{rem}

% \subsection{Deep-learning, how deep is ``deep'' ?}
\subsection{What is ``deep'' in ``deep networks'' ? Size, architecture}
\label{sc:depth-size}

%\subsubsection{How deep and wide could a ``deep'' network be ?}
%\subsubsection{Depth and width of practical deep networks}
% \subsubsection{How deep is ``deep'' ? size, architecture}
% \subsubsection{How ``deep'' and how ``wide'' a network could be ?}
%\subsubsection{What is ``deep'' in ``deep networks'' ? size, architecture}
%\subsubsection{Network depth, size, architecture}

\subsubsection{Depth, size}
\label{sc:depth}

The concept of network depth turns out to be more complex than initially thought.
While for a {\em fully-connected} feedforward neural network (in which all outputs of a layer are connected to a neuron in the following layer), depth could be considered as the number of layers, there is in general no consensus on the accepted definition of depth. 
%
% CMES style, rewriting
%\cite{Goodfellow.2016}, p.~8, 
It was stated in \cite{Goodfellow.2016}, p.~8, that:\footnote{
	There are two viewpoints on the definition of depth, one based on the computational graph, and one based on the conceptual graph. From the computational-graph viewpoint, depth is the number of sequential instructions that must be executed in an architecture.  From the conceptual-graph viewpoint, depth is the number of concept levels, going from simple concepts to more complex concepts. 
	See also \cite{Goodfellow.2016}, p.~163, for the depth of fully-connected feedforward networks as the ``length of the chain'' in Eq.~(\ref{eq:compositions}) and Eq.~(\ref{eq:network}), which is the number of layers.
}
\begin{quote}
	``There is no single correct value for the depth of an architecture,\footnote{
		There are several different network architectures.  {\em Convolutional neural networks} (CNN) use sparse connections, have achieved great success in image recognition, and contributed to the burst of interest in deep learning since winning the ImageNet competion in 2012 by almost halving the image classification error rate; see 
%		\cite{LeCun.2015:rd0001} \cite{Schmidhuber.2015:rd0001} \cite{Economist.2016:rd0001}.
		\cite{LeCun.2015:rd0001, Schmidhuber.2015:rd0001, Economist.2016:rd0001}.
		{\em Recurrent neural networks} (RNN) are used to process a sequence of inputs to a system with changing states as in a dynamical system, to be discussed in Section~\ref{sc:recurrent}. there are other networks with skip connections, in which information flows from layer $(\ell)$ to layer $(\ell+2)$, skipping layer $(\ell+1)$; see \cite{Goodfellow.2016}, p.~196.
	} 
	just as there is no single correct value for the length of a computer Program. Nor is there a consensus about how much depth a model requires to Qualify as `deep.' ''
\end{quote}

For example, keeping the number of layers the same, then the ``depth'' of a {\em sparsely-connected} feedforward network (in which not all outputs of a layer are connected to a neuron in the following layer) should be smaller than the ``depth'' of a {\em fully-connected} feedforward network.

%
% CMES style, rewriting
%\cite{Schmidhuber.2015:rd0001} echoed 
The lack of consensus on the boundary between ``shallow'' and ``deep'' networks is echoed in \cite{Schmidhuber.2015:rd0001}:
\begin{quote}
	``At which problem depth does {\em Shallow Learning} end, and {\em Deep Learning} begin? Discussions with DL experts have not yet yielded a conclusive response to this question. Instead of committing myself to a precise answer, let me just define for the purposes of this overview: problems of depth $>$ 10 require {\em Very Deep Learning}.''
\end{quote}

\begin{rem}
	\label{rm:depth-definitions}
	Action depth, state depth.
	{\rm
		In view of Remark~\ref{rm:layer-definitions}, which type of layer (action or state) were they talking about in the above quotation?  We define here \emph{action depth} as the number of action layers, and \emph{state depth} as the number of state layers.  The abstract network in Figure~\ref{fig:network3b} has action depth $\D$ and state depth $(\D+1)$, with $(\D-1)$ as the number of hidden (state) layers.
	}
$\hfill\blacksquare$
\end{rem}

%
% CMES style, rewriting
%\cite{Oishi.2017:rd9648} attributed to 
The review paper \cite{LeCun.2015:rd0001} was attributed in \cite{Oishi.2017:rd9648} for stating that ``training neural networks with more than three hidden layers is called deep learning'', implying that a network is considered ``deep'' if its number of hidden (state) layers 
$(\D - 1) > 3$.
%
% CMES style, rewriting
%In their own work, \cite{Oishi.2017:rd9648} 
In the work reported in \cite{Oishi.2017:rd9648}, the authors
used networks with number of hidden (state) layers $(\D - 1)$ varying from one to five, and with a constant hidden (state) layer width of $m_{(\ell)} = 50$, for all hidden (state) layers $\ell = 1, \ldots, 5$; see Table 1 in 
%
% CMES style, rewriting
%their paper, 
\cite{Oishi.2017:rd9648},
reproduced in Figure~\ref{fig:Oishi-tab1} in Section~\ref{sc:Oishi-method-1-training}.

%{\color{red} [NOTE 2020.08.22. introduce ``action layer'' versus ``state layer''. ENDNOTE]}

\begin{figure}[h]
  \centering
  \includegraphics[width=0.8\linewidth]{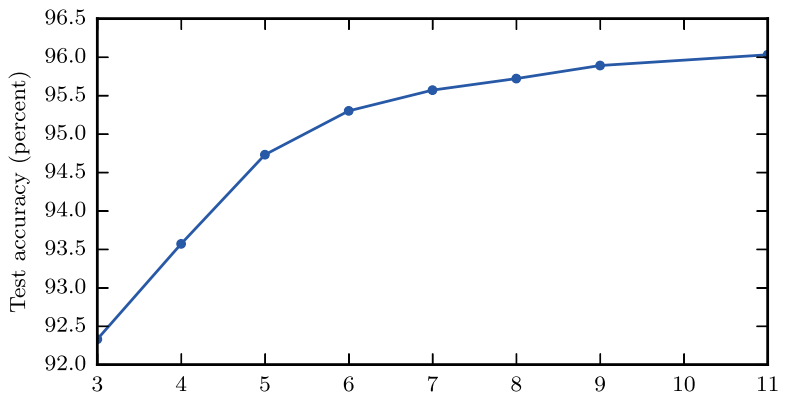}
  \caption{
  \emph{Test accuracy versus network depth} (Section~\ref{sc:depth}), showing that test accuracy for this example increases monotonically with the network depth (number of layers).
  \cite{Goodfellow.2016}, p.~196.
  %\\
  %{\small (Figure reproduced with permission of the authors.)} 
  {\footnotesize (Figure reproduced with permission of the authors.)}
  }
  \label{fig:Accuracy-Depth}
\end{figure}

%
% CMES style, rewriting
%\cite{Goodfellow.2016}, p.~196, provided 
An example of recognizing multidigit numbers in photographs of addresses, in which the test accuracy increased (or test error decreased) with increasing depth, is provided in \cite{Goodfellow.2016}, p.~196; see Figure~\ref{fig:Accuracy-Depth}.

But it is not clear where 
%
% CMES style, rewriting
%\cite{LeCun.2015:rd0001} actually 
in \cite{LeCun.2015:rd0001} that it was actually said that a network is ``deep'' if the number of hidden (state) layers is greater than three.
% defined the ``depth of a network''. 
%
% CMES style, rewriting
%These authors  
An example in image recognition having more than three layers was, however, given in \cite{LeCun.2015:rd0001} (emphases are ours):
\begin{quote}
	``An image, for example, comes in the form of an array of pixel values, and the learned features in the {\em first layer} of representation typically represent the presence or absence of edges at particular orientations and locations in the image. The {\em second layer} typically detects motifs by spotting particular arrangements of edges, regardless of small variations in the edge positions. The {\em third layer} may assemble motifs into larger combinations that correspond to parts of familiar objects, and {\em subsequent layers} would detect objects as combinations of these parts.''
\end{quote}
But the above was not a criterion for a network to be considered as ``deep''.
% definition of network depth.
%
% CMES style, rewriting
%\cite{LeCun.2015:rd0001} 
It was further noted on the number of the model parameters (weights and biases) and the size of the training dataset for a ``typical deep-learning system'' as follows \cite{LeCun.2015:rd0001} (emphases are ours):
\begin{quote}
	`` In a typical {\em deep-learning} system, there may be 
	{\em hundreds of millions} of these adjustable {\em weights}, and {\em hundreds of millions} of labelled examples with which to train the machine.''
\end{quote}
See Remark~\ref{rm:depth-of-RNN} on recurrent neural networks (RNNs) as equivalent to ``very deep feedforward networks''.
%
% CMES style, rewriting
%\cite{LeCun.2015:rd0001} also provided 
Another example was also provided in \cite{LeCun.2015:rd0001}:
\begin{quote}
	``Recent ConvNet [convolutional neural network, or CNN]\footnote{
		\label{fn:ConVNet}
		A special type of deep network that went out of favor, then now back in favor, among the computer-vision and machine-learning communities after the spectacular success that ConvNet garnered at the 2012 ImageNet competition; see \cite{LeCun.2015:rd0001} \cite{Economist.2016:rd0001} \cite{Economist.2016:rd0002}.  Since we are reviewing in detail some specific applications of deep networks to computational mechanics, {we will not review ConvNet here}, but focus on MultiLayer Neural (MLN)---also known as MultiLayer Perceptron (MLP)---networks.} 
	architectures have 10 to 20 layers of ReLUs [rectified linear units], hundreds of millions of weights, and billions of connections between Units.''\footnote{
		A network processing ``unit'' is also called a ``neuron''.}
\end{quote}

%{\color{red} NOTE: 2022.06.21 - Alex, do we review in detail CNN?  If yes, we need to rewrite Footnote~\ref{fn:ConVNet} related to ``ConvNet'' in the quote above.}
%{\color{blue} NOTE: 2022.10.03 - not yet, but we should have one, I suppose, given the importance of ConvNets.}

%
% CMES style, rewriting
%In 2015, \cite{Hsu.2015:rd0001} reported perhaps the largest neural network at the time, 
A neural network with 160 billion parameters was perhaps the largest in 2015 \cite{Hsu.2015:rd0001}:
\begin{quotation}
	``Digital Reasoning, a cognitive computing company based in Franklin, Tenn., recently announced that it has trained a neural network consisting of 160 billion parameters---more than 10 times larger than previous neural networks.
	
	The Digital Reasoning neural network easily surpassed previous records held by Google's 11.2-billion parameter system and Lawrence Livermore National Laboratory's 15-billion parameter system.''
\end{quotation}

As mentioned above, for general network architectures (other than feedforward networks), not only that there is no consensus on the definition of depth, there is also no consensus on how much depth a network must have to qualify as being ``deep''; see \cite{Goodfellow.2016}, p.~8, who offered the following intentionally vague definition:
\begin{quote}
	``Deep learning can be safely regarded as the study of models that involve a greater amount of composition of either learned functions or learned concepts than traditional machine learning does.''
\end{quote}

\begin{figure}[h]
	\centering
	\includegraphics[width=1.0\linewidth]{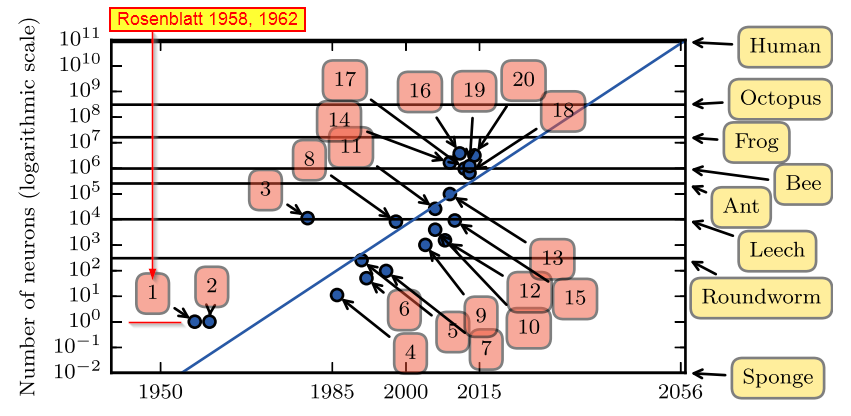}
	\caption{
		\emph{Increasing network size over time} (Section~\ref{sc:depth}, \ref{sc:linear-combo-history}).  All networks before 2015 had their number of neurons smaller than that of a frog at $1.6 \times 10^7$, and still far below that in a human brain at $8.6 \times 10^{10}$; see ``List of animals by number of neurons'', Wikipedia, 
		\href{https://en.wikipedia.org/w/index.php?title=List_of_animals_by_number_of_neurons&oldid=896223835}{version 02:46, 9 May 2019}.
		% See 
		%
		% CMES style, rewriting
		In \cite{Goodfellow.2016}, p.~23, it was estimated that neural network size would double every 2.4 years (a clear parallel to Moore's law, which stated that the number of transistors on integrated circuits doubled every 2 years).
		%
		% CMES style, rewriting
%		\cite{Goodfellow.2016}, p.~23, cited 
		It was mentioned in \cite{Goodfellow.2016}, p.~23, that Network 1 by %\cite{Rosenblatt.1958} 
		Rosenblatt 
%		(\citeyear{Rosenblatt.1958}, \citeyear{Rosenblatt.1962}) 
		(1958 \cite{Rosenblatt.1958}, 1962 \cite{Rosenblatt.1962})
		as having one neuron  (see figure above), which was incorrect, since Rosenblatt (1957) \cite{Rosenblatt.1957} conceived a network with 1000 neurons, and even built the Mark I computer to run this network; see Section~\ref{sc:linear-combo-history} and Figure~\ref{fig:Rosenblatt-Mark-I}. 	
		%\\
		%{\small (Figure reproduced with permission of the authors.)} 
		{\footnotesize (Figure reproduced with permission of the authors.)}
	}
	\label{fig:network-size-time}
\end{figure}

Figure~\ref{fig:network-size-time} depicts the increase in the number of neurons in neural networks over time, from 1958 (Network 1 by Rosenblatt (1958) \cite{Rosenblatt.1958} in Figure~\ref{fig:network-size-time} with one neuron, which was an error in \cite{Goodfellow.2016}, as discussed in Section~\ref{sc:linear-combo-history}) to 2014 (Network 20 GoogleNet with more than one million neurons), which was still far below the more than ten million biological neurons in a frog.

\subsubsection{Architecture}
\label{sc:architecture}

The architecture of a network is the number of layers (depth), the layer width (number of neurons per layer), and the connection among the neurons.\footnote{
	See \cite{Goodfellow.2016}, p.~166.
} 
We have seen the architecture of fully-connected feedforward neural networks above; see Figure~\ref{fig:network3b} and Figure~\ref{fig:neuron4}.

One example of an architecture different from that fully-connected feedforward networks is convolutional neural networks, which are based on the convolutional integral (see Eq.~(\ref{eq:linear-volterra-series}) in Section~\ref{sc:dynamic-volterra-series} on ``Dynamic, time dependence, Volterra series''), and which had proven to be successful long before deep-learning networks:
\begin{quote}
	``Convolutional networks were also some of the first neural networks to solve important commercial applications and remain at the forefront of commercial applications of deep learning today.   By the end of the 1990s, this system deployed by NEC was reading over 10 percent of all the checks in the United States. Later, several OCR and handwriting recognition systems based on convolutional nets were deployed by Microsoft.''
	\cite{Goodfellow.2016}, p.~360.
\end{quote}
\begin{quote}
	``Fully-connected networks were believed not to work well.  It may be that the primary barriers to the success of neural networks were psychological (practitioners did not expect neural networks to work, so they did not make a serious effort to use neural networks).
	Whatever the case, it is fortunate that convolutional networks performed well decades ago. In many ways, they carried the torch for the rest of deep learning and paved the way to the acceptance of neural networks in general.''
	\cite{Goodfellow.2016}, p.~361.
\end{quote}

\begin{figure}[h]
	\centering
	\includegraphics[width=1.0\linewidth]{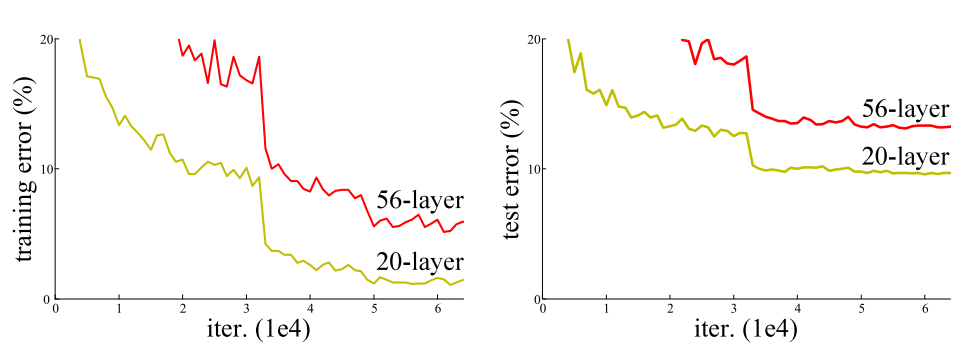}
	\caption{
		\emph{Training/test error vs. iterations, depth}
		(Sections~\ref{sc:architecture}, \ref{sc:training}).
		The training error and test error of deep fully-connected networks increased when the number of layers (depth) increased \cite{He.2015:rd0001}.
		%
		% CMES style, rewriting
%		From \cite{He.2015:rd0001}.
		%\\
		%{\small (Figure reproduced with permission of the authors.)} 
		{\footnotesize (Figure reproduced with permission of the authors.)}
	}
	\label{fig:deep-network-errors}
\end{figure}

Here, we present a more recent and successful network architecture different from the fully-connected feedforward network.
%
% CMES style, rewriting
%\cite{He.2015:rd0001} introduced 
Residual network was introduced in \cite{He.2015:rd0001} to address the problem of vanishing gradient that plagued ``very deep'' networks with as few as 16 layers during training (see Section~\ref{sc:backprop} on Backpropagation) and the problem of increased training error and test error with increased network depth as shown in Figure~\ref{fig:deep-network-errors}.

\begin{rem}
	\label{rm:training-test-error}
	Training error, test (generalization) error.
	{\rm 
		Using a set of data, called training data, to find the parameters that minimize the loss function (i.e., doing the training) provides the training error, which is the least square error between the predicted outputs and the training data.  Then running the optimally trained model on a different set of data, which was not been used for the training, called test data, provides the test error, also known as generalization error.  More details can be found in Section~\ref{sc:training}, %Section~\ref{sc:training-test-validation}, 
		and in \cite{Goodfellow.2016}, p.~107.
		$\hfill\blacksquare$ 
	}
\end{rem}

\begin{figure}[h]
  \centering
  \includegraphics[width=0.35\linewidth]{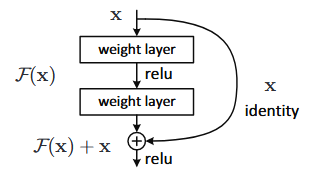}
  \caption{
  \emph{Residual network} (Sections~\ref{sc:architecture}, \ref{sc:training}), basic building block having two layers with the rectified linear activation function (ReLU), for which the input is $\bx$, the output is $\mathcal{H} (\bx) = \mathcal{F} (\bx) + \bx$, where the internal mapping function $\mathcal{F} (\bx) = \mathcal{H} (\bx) - \bx$ is called the residual.  Chaining this building block one after another forms a deep residual network; see Figure~\ref{fig:resNet-full} \cite{He.2015:rd0001}. 
  %
  % CMES style, rewriting
%  Figure from \cite{He.2015:rd0001}.
  \\
  %{\small (Figure reproduced with permission of the authors.)} 
  {\footnotesize (Figure reproduced with permission of the authors.)}
  }
  \label{fig:resNet-basic}
\end{figure}

\begin{figure}[h]
  \centering
  \includegraphics[width=1.0\linewidth]{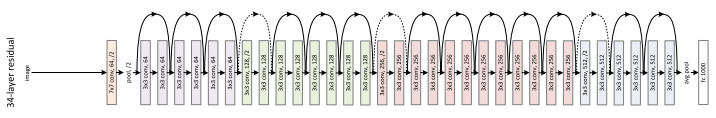}
  \caption{
  \emph{Full residual network} (Sections~\ref{sc:architecture}, \ref{sc:training}) with 34 layers, made up from 16 building blocks with two layers each (Figure~\ref{fig:resNet-basic}), together with an input and an output layer.  This residual network has a total of 3.6 billion floating-point operations (FLOPs with fused multiply-add operations), which could be considered as the network ``computational depth'' \cite{He.2015:rd0001}.
  % 
  % CMES style, rewriting
%  Figure from \cite{He.2015:rd0001}.
%  \\
  %{\small (Figure reproduced with permission of the authors.)} 
  {\footnotesize (Figure reproduced with permission of the authors.)}
  }
  \label{fig:resNet-full}
\end{figure}

The basic building block of residual network is shown in Figure~\ref{fig:resNet-basic}, and a full residual network in Figure~\ref{fig:resNet-full}.
The rationale for residual networks was that, if the identity map were optimal, it would be easier for the optimization (training) process to drive the residual $\mathcal{F} (\bx)$ down to zero than to fit the identity map with a bunch of nonlinear layers; see \cite{He.2015:rd0001}, where it was mentioned that deep residual networks won 1st places in several image recognition competitions.   

\begin{rem}
	\label{rm:residual-network}
	{\rm 
		The identity map that jumps over a number of layers in the residual network building block in Figure~\ref{fig:resNet-basic} and in the full residual network in Figure~\ref{fig:resNet-full} is based on a concept close to that for the path of the cell state $\boldsymbol{c}^{[k]}$ in the Long Short-Term Memory (LSTM) unit for recurrent neural networks (RNN), as described in Figure~\ref{fig:our-lstm_cell} in Section~\ref{sc:LSTM}.
		$\hfill\blacksquare$
	}
\end{rem}

%
% CMES style, rewriting
%\cite{Huang.2016:rd0001} proposed 
A deep residual network with more than 1,200 layers was proposed in \cite{Huang.2016:rd0001}.
%
% CMES style, rewriting 
%\cite{Zagoruyko.2017} present 
A wide residual-network architecture that outperformed deep and thin networks was proposed in \cite{Zagoruyko.2017}:  
``For instance, [their] wide 16-layer deep network has the same accuracy as a 1000-layer thin deep network and a comparable number of parameters, although being several times faster to train.''

It is still not clear why some architecture worked well, while others did not:
\begin{quote}
	``The design of hidden units is an extremely active area of research and does not yet have many definitive guiding theoretical principles.''
	\cite{Goodfellow.2016}, p.~186.
\end{quote}

% backpropagation method

\section{Backpropagation}
\label{sc:backprop}

Backpropagation, sometimes abbreviated as ``backprop'', was a child  
%with many fathers, 
of whom many could claim to be the father,
and is used to compute the gradient of the cost function with respect to the parameters (weights and biases); see Section~\ref{sc:backprop-history} for a history of backpropagation.   This gradient is then subsequently used in an optimization process, usually the Stochastic Gradient Descent method, to find the parameters that minimize the cost or loss function.

\subsection{Cost (loss, error) function}
\label{sc:cost-function}

Two types of cost function are discussed here: (1) the mean squared error (MSE), and (2) the maximum likelihood (probability cost).\footnote{
	For other types of loss function, see, e.g., (1) Section ``Loss functions'' in ``torch.nn --- PyTorch Master Documentation'' (\href{https://pytorch.org/docs/stable/nn.html}{Original website}, \href{https://web.archive.org/web/20191031180518/https://pytorch.org/docs/stable/nn.html}{Internet archive}), and (2) Jah 2019, A Brief Overview of Loss Functions in Pytorch (\href{https://medium.com/udacity-pytorch-challengers/a-brief-overview-of-loss-functions-in-pytorch-c0ddb78068f7}{Original website}, \href{https://web.archive.org/web/20191118171435/https://medium.com/udacity-pytorch-challengers/a-brief-overview-of-loss-functions-in-pytorch-c0ddb78068f7}{Internet archive}).
}

\subsubsection{Mean squared error}
\label{sc:mean-squared-error}
% \cite{Goodfellow.2016}, p.~172 (pdf p.~195), Section 6.2.1, Cost functions.
For a given input $\bx$ (a single example) and target output $\by \in \real^m$, the squared error (SE) of the predicted output $\bout \in \real^m$ for use in least squared error problem is defined as half the squared error:
\begin{align}
	J (\bparam)
	=
	\frac12 \, \text{SE}	
	= 
	\frac{1}{2} \parallel \by - \bout \parallel^2
	=
	\frac{1}{2} \sum_{k=1}^{k=m} \left( \y_k - \out_k \right)^2
	\ . 
	\label{eq:half-MSE}
\end{align}
The factor $\frac12$ is for the convenience of avoiding to carry the factor $2$ when taking the gradient of the cost (or loss) function $J$.\footnote{
	\label{fn:predicted-value-hat}
	There is an inconsistent use of notation in \cite{Goodfellow.2016} that could cause confusion, e.g., in \cite{Goodfellow.2016}, Chap.~5, p.~104, Eq.~(5.4), the notation $\widehat{\by}$ (with the hat) was defined as the network outputs, i.e., predicted values, with $\by$ (without the hat) as target values, whereas later in Chap.~6, p.~163, the notation $\by$ (without the hat) was used for the network outputs.
	Also, 
	%
	% CMES style, rewriting
%	\cite{Goodfellow.2016}, p.~105, defined 
	in \cite{Goodfellow.2016}, p.~105, 
	the cost function was defined as the mean squared error, without the factor $\frac12$.
	See also Footnote~\ref{fn:predicted-output}.
}
%\cite{Goodfellow.2016}, p.~182: principle of maximum likelihood

While the components $y_k$ on the output matrix $\by$ cannot be \emph{independent and identically distributed} (\emph{i.i.d.}), since $\by$ must represent a recognizable pattern (e.g., an image), in the case of training with $\bsize$ examples as inputs:\footnote{
	\label{fn:output-examples-size-notation}
	In our notation, $m$ is the dimension of the output array $\by$, whereas $\bsize$ (in a different font) is here the number of examples in Eq.~(\ref{eq:training-examples}), and
	later represents the minibatch size in Eqs.~(\ref{eq:minibatch-1})-(\ref{eq:minibatch-3}).  The size of the whole training set, called the ``full batch'' (Footnote~\ref{fn:full-batch-minibatch}), is denoted by $\Bsize$ (Footnote~\ref{fn:imagenet-size-notation}).
}
\begin{align}
	\mathbb{X}
	= 
	\{ \bexin{1} , \cdots , \bexin{\bsize}  \}
	\text{ and }
	\mathbb{Y}
	=
	\{ \bexout{1} , \cdots , \bexout{\bsize}  \}
	\label{eq:training-examples}
\end{align}
where $\mathbb{X}$ is the set of $\bsize$ examples, and $\mathbb{Y}$ the set of the corresponding outputs, the examples $\bexin{k}$ can be \emph{i.i.d.}, and the half MSE cost function for these outputs is half the expectation of the SE:
\begin{align}
	J (\bparam)
	=
	\frac12 \, \text{MSE}	
	=
	\frac12 \, \mathbb{E}(\{ SE^{|k|} , k= 1, \cdots, \bsize \}) 
	=
	\frac{1}{2 \bsize} \sum_{k=1}^{k=\bsize} \parallel \bexout{k} - \bout^{|k|} \parallel^2
	\ .
	\label{eq:half-MSE-2}
\end{align}

\subsubsection{Maximum likelihood (probability cost)}
\label{sc:maximum-likelihood}
Many (if not most) modern networks employed a probability cost function based in the principle of maximum likelihood, which has the form of negative log-likelihood, describing the cross-entropy between the training data with probability distribution $\hat{p}_{data}$ and the model with probability distribution $p_{model}$ (\cite{Goodfellow.2016}, p.~173):
% https://en.wikipedia.org/wiki/List_of_mathematical_symbols_by_subject
\begin{align}
	J (\bparam)
	= 
	-
	\mathbb{E}_{{\textup{\bf x}}, {\textup{\bf y}} \sim \hat{p}_{data}} 
	\log p_{model} (\by \, | \, \bx ; \bparam )
	\ ,
	\label{eq:likelihood-cost}
\end{align}
where $\mathbb{E}$ is the expectation; $\textup{\bf x}$ and $\textup{\bf y}$ are random variables for training data with distribution $\hat{p}_{data}$; the inputs $\bx$ and the target outputs $\by$ are values of $\textup{\bf x}$ and $\textup{\bf y}$, respectively, and $p_{model}$ is the conditional probability of the distribution of the target outputs $\by$ given the inputs $\bx$ and the parameters $\bparam$, with the predicted outputs $\nfwidetilde{\by}$ given by the model $f$ (neural network), having as arguments the inputs $\bx$ and the parameters $\bparam$:
\begin{align}
	\bout = f (\bx , \bparam)
	\Leftrightarrow
	\out_k = f_k (\bx , \bparam)
	\label{eq:predicted-outputs}
\end{align}
The expectations of a function $g(\xup = x)$ of a random variable $\xup$, having a probability distribution $P(\xup = x)$ for the discrete case, and the probability distribution density $p(\xup = x)$ for the continuous case, are respectively\footnote{
	\label{fn:expectation-notation}
	The simplified notation $\langle \cdot \rangle$ for expectation $\xpc (\cdot)$, with implied probability distribution, is used in Section~\ref{sc:step-length-decay} on step-length decay and simulated annealing (Remark~\ref{rm:annealing}, Section~\ref{sc:minibatch-size-increase}) as an add-on improvement to the stochastic gradient descent algorithm.
}
\begin{align}
	\mathbb{E}_{{\textup{\bf x}} \sim {P}} \, g(x) = \sum_x g(x) P(x) 
	= 
	\langle g(x) \rangle
	\ , \quad
	\mathbb{E}_{{\textup{\bf x}} \sim {p}} \, g(x) = \int g(x) p(x) dx
	=
	\langle g(x) \rangle
	\ .
	\label{eq:expectation}
\end{align}

\begin{rem}
	Information content, Shannon entropy, maximum likelihood.
	{\rm
		The expression in Eq.~(\ref{eq:likelihood-cost})---with the minus sign and the log function---can be abstract to readers not familiar with the probability concept of maximum likelihood, which is related to the concepts of information content and Shannon entropy.  First, an event $x$ with low probability (e.g., an asteroid will hit the Earth tomorrow) would have higher information content than an event with high probability (e.g., the sun will rise tomorrow morning).  since the probability of $x$, i.e., $P(x)$, is between 0 and 1, the negative of the logarithm of $P(x)$, i.e.,  
		\begin{align}
			I(x) = - \log P(x)
			\ ,
			\label{eq:information}
		\end{align}
		called the information content of $x$, would have large values near zero, and small values near 1. In addition, the probability of two independent events to occur is the product of the probabilities of these events, e.g., the probability of having two heads in two coin tosses is
		\begin{align}
			P(\textup{x} = \text{head}, \textup{y} = \text{head}) 
			= 
			P (\text{head}) \times P (\text{head})
			=
			\frac12 \times \frac12 = \frac14
			\label{eq:two-coin-tosses}
		\end{align}
		The product (chain) rule of conditional probabilities consists of expressing a joint probability of several random variables $\{ \bxupp{1} , \bxupp{2} , \cdots , \bxupp{n} \}$ as the product\footnote{
			See, e.g., \cite{Goodfellow.2016}, p.~57.
			The notation $\bxupp{k}$ (with vertical bars enclosing the superscript $k$) is used to designate example $k$ in the set $\mathbb{X}$ of examples in Eq.~(\ref{eq:param-min-2b}), instead of the notation $\bxp{k}$ (with parentheses), since the parentheses were already used to surround the layer number $k$, as in Figure~\ref{fig:neuron4}.
		}
		\begin{align}
			P(\bxupp{1} , \cdots , \bxupp{n})
			=
			P(\bxupp{1}) \prod_{k=2}^{k=n} P(\bxupp{k} | \bxupp{1} , \cdots , \bxupp{k-1})
			\label{eq:product-rule} 
		\end{align}
		The logarithm of the products in Eq.~(\ref{eq:two-coin-tosses}) and Eq.~(\ref{eq:product-rule}) is the sum of the factor probabilities, and provides another reason to use the logarithm in the expression for information content in Eq.~(\ref{eq:information}): Independent events have additive information.  Concretely, the information content of two  asteroids independently hitting the Earth should double that of one asteroid hitting the Earth.  
		
		The parameters $\widetilde{\bparam}$ that minimize the probability cost $J(\bparam)$ in Eq.~(\ref{eq:likelihood-cost}) can be expressed as\footnote{
			A tilde is put on top of $\bparam$ to indicate that the matrix $\widetilde{\bparam}$ contains the estimated values of the parameters (weights and biases), called the estimates, not the true parameters.  Recall from Footnote~\ref{fn:predicted-value-hat} that \cite{Goodfellow.2016} used an overhead ``hat'' ($\hat{\cdot}$) to indicate predicted value; see \cite{Goodfellow.2016}, p.~120, where $\bparam$ is defined as the true parameters, and $\hat{\bparam}$ the predicted (or estimated) parameters.
		}
		\begin{align}
			\widetilde{\bparam}
			&
			=
			-
			\arg\min_{\bparam} 
			\mathbb{E}_{{\textup{\bf x}}, {\textup{\bf y}} \sim \hat{p}_{data}} 
			\log p_{model} (\by \, | \, \bx ; \bparam )
			=
			\arg\max_{\bparam}
			\frac{1}{\bsize}
			\sum_{k=1}^{k=\bsize}
			\log p_{model} (\bexout{k} \, | \, \bexin{k} ; \bparam )
			\label{eq:param-min-1}
			\\
			&
			=
			\frac{1}{\bsize}
			\arg\max_{\bparam} \log \prod_{k=1}^{k=\bsize}
			P_{model} (\bexout{k} \, | \, \bexin{k} ; \bparam )
			=
			\arg\max_{\bparam} p_{model} ( \mathbb{Y} \, | \, \mathbb{X} ; \bparam)
			\label{eq:param-min-2a}
			\\
			&
			\text{with }
			\mathbb{X}
			= 
			\{ \bexin{1} , \cdots , \bexin{\bsize}  \}
			\text{ and }
			\mathbb{Y}
			=
			\{ \bexout{1} , \cdots , \bexout{\bsize}  \}
			\ ,
			\label{eq:param-min-2b}
		\end{align}
		where $\mathbb{X}$ is the set of $\bsize$ examples that are \emph{independent and identically distributed} (\emph{i.i.d.}), and $\mathbb{Y}$ the set of the corresponding outputs.
		The final form in Eq.~(\ref{eq:param-min-2a}), i.e.,
		\begin{align}
			\widetilde{\bparam}
			=
			\arg\max_{\bparam} p_{model} ( \mathbb{Y} \, | \, \mathbb{X} ; \bparam)
			\label{eq:principle-max-likelihood}
		\end{align}
		is called the {\em Principle of Maximum Likelihood}, in which the model parameters are optimized to maximize the likelihood to reproduce the empirical data.\footnote{
			See, e.g., \cite{Goodfellow.2016}, p.~128.
		}
		$\hfill\blacksquare$
	}
\end{rem}

\begin{rem}
	\label{rm:MSE-likelihood}
	Relation between Mean Squared Error and Maximum Likelihood.
	{\rm
		The MSE is a particular case of the Maximum Likelihood.
		Consider having $\bsize$ examples $\mathbb{X} = \{ \bexin{1} , \cdots , \bexin{\bsize} \}$ that are independent and identically distributed (\emph{i.i.d.}), as in Eq.~(\ref{eq:training-examples}).
		If the model probability $p_{model} (\bexout{k} \, | \, \bexin{k} ; \bparam )$ has a normal distribution, with the predicted output 
		\begin{align}
			\bexoutt{k} = f(\bexin{k} , \bparam)
			\label{eq:output-example-k}
		\end{align}
		as in Eq.~(\ref{eq:predicted-outputs}), predicting the mean of this normal distribution,\footnote{
			The normal (Gaussian) distribution of scalar random variable $x$, mean $\mu$, and variance $\sigma^2$ is written as
			${\mathcal N} (x ; \mu , \sigma^2) = (2 \pi \sigma^2)^{-1/2} \exp [- (x - \mu)^2 / (2 \sigma^2)]$; see, e.g., \cite{bishop2006pattern}, p.~24.
		} then
		\begin{align}
			p_{model} (\bexout{k} \, | \, \bexin{k} ; \bparam ) 
			= 
			\mathcal{N} (\bexout{k} ; f (\bexin{k} , \bparam) , \stdev^2)
			=
			\mathcal{N} (\bexout{k} ; \bexoutt{k} , \stdev^2)
			=
			\frac{1}{[2 \pi \stdev^2]^{\frac12}}
			\exp \left[ - \frac{ \parallel \bexout{k} - \bexoutt{k} \parallel^2 }{2
			\stdev^2} \right]
			\ ,
			\label{eq:p-model-normal-distribution} 
		\end{align}
		with $\stdev$ designating the standard deviation,
		i.e., the error between the target output $\bexout{k}$ and the predicted output $\bexoutt{k}$ is normally distributed.
		By taking the negative of the logarithm of $p_{model} (\bexout{k} \, | \, \bexin{k} ; \bparam )$, we have
		\begin{align}
			\log p_{model} (\bexout{k} \, | \, \bexin{k} ; \bparam )
			=
			\frac12 \log (2 \pi \stdev^2) - \frac{ \parallel \bexout{k} -
			\bexoutt{k} \parallel^2 }{2 \stdev^2}
			\ .
			\label{eq:log-p-model}
		\end{align}
		Then summing Eq.~(\ref{eq:log-p-model}) over all examples $k = 1, \cdots, \bsize$ as in the last expression in Eq.~(\ref{eq:param-min-1}) yields
		\begin{align}
			\sum_{k=1}^{k=\bsize}
			\log p_{model} (\bexout{k} \, | \, \bexin{k} ; \bparam )
			=
			\frac{\bsize}{2} \log (2 \pi \stdev^2) 
			- 
			\sum_{k=1}^{k=\bsize}
			\frac{ \parallel \bexout{k} - \bexoutt{k} \parallel^2 }{2 \stdev^2}
			\ , 
		\end{align}
		and thus the minimizer $\widetilde{\bparam}$ in Eq.~(\ref{eq:param-min-1}) can be written as
		
%		{
%		\color{blue} NOTE 2019.10.3: that should be an $\widetilde{\bparam}$ rather than $\hat{\bparam}$, right?
%		}
		\begin{align}
			\bparamt
			=
			-
			\arg\max_{\bparam} 
			\sum_{k=1}^{k=\bsize}
			\frac{ \parallel \bexout{k} - \bexoutt{k} \parallel^2 }{2 \stdev^2}
			=
			\arg\min_{\bparam}
			\sum_{k=1}^{k=\bsize}
			\frac{ \parallel \bexout{k} - \bexoutt{k} \parallel^2 }{2 \stdev^2}
			=
			\arg\min_{\bparam} J(\bparam)
			\ ,
			\label{eq:max-likelihood-MSE}
		\end{align}
		where the MSE cost function $J(\bparam)$ was defined in Eq.~(\ref{eq:half-MSE-2}), noting that constants such as $m$ or $2 m$ do not affect the value of the minimizer $\widetilde{\bparam}$.
		
		Thus finding the minimizer of the maximum likelihood cost function in Eq.~(\ref{eq:likelihood-cost}) is the same as finding the minimizer of the MSE in Eq.~(\ref{eq:half-MSE}); see also \cite{Goodfellow.2016}, p.~130.
		$\hfill\blacksquare$
	}
\end{rem}

Remark~\ref{rm:MSE-likelihood} justifies the use of Mean Squared Error as  \emph{a} Maximum Likelihood estimator.\footnote{
	See, e.g., \cite{Goodfellow.2016}, p.~130.
}
For the purpose of this review paper, it is sufficient to use the MSE cost function in Eq.~(\ref{eq:MSE}) to develop the backpropagation procedure.

\subsubsection{Classification loss function}
\label{sc:classification}

In classification tasks---such as used 
%
% CMES style, rewriting
%by 
in
\cite{Oishi.2017:rd9648}, Section~\ref{sc:Oishi-1.1}, and Footnote~\ref{fn:oishi-classification}---a neural network is trained to predict which of $k$ different classes (categories) an input $\xv$ belongs to. 
%In classification tasks, a neural network is trained to predict the probability distribution (probability mass function) over a discrete variable of $n$ different classes, see~\cite{Goodfellow.2016}, p.~175, Section 6.2.2, ``Output Units''.
%
The most simple classification problem only has two classes ($k=2$), which can be represented by the values $\{0,1\}$ of a single binary variable $y$. 
The probability distribution of such single boolean-valued variable is called \emph{Bernoulli distribution}.\footnote{
	See, e.g., \cite{bishop2006pattern}, p.~68.
}
The Bernoulli distribution is characterized by a single parameter $p(y = 1 \, | \, \xv)$, i.e., the conditional probability of $\xv$ belonging to the class $y=1$.
To perform binary classification, a neural network is therefore trained to estimate the conditional probability distribution $\nfwidetilde y = p(y = 1 \, | \, \xv)$ using the principle of maximum likelihood (see Section~\ref{sc:maximum-likelihood}, Eq.~\eqref{eq:likelihood-cost}):
\begin{equation}
\begin{aligned}
%J ( \thetav ) = - \log P(y = 1 \, | \, \xv) = - \log \sigmoid (z^{(L)}).
J ( \thetav ) &= - \mathbb{E}_{{\textup{\bf x}}, {\textup{\bf y}} \sim \hat{p}_{data}} \log p_{model} (y \, | \, \bx ; \bparam ) \\
&= - \frac 1 \bsize \sum_{k=1}^{k=\bsize} \left\{ y^{|k|} \log p_{model} (y^{|k|} = 1 \, | \, \bx^{|k|} ; \bparam ) + (1 - y^{|k|})  \log ( 1 -  p_{model} (y^{|k|} = 1 \, | \, \bx^{|k|} ; \bparam ) ) \right\} .
\end{aligned}
\end{equation}
The output of the neural network is supposed to represent the probability $p_{model} (y = 1 \, | \, \bx ; \bparam )$, i.e., a real-valued number in the interval $[0,1]$.
A linear output layer $\nfwidetilde y = y^{(L)} = z^{(L)} = \boldsymbol W^{(L)} \yv^{(L-1)} + b^{(L)}$ does not meet this constraint in general. 
To squash the output of the linear layer into the range of $[0,1]$, the logistic sigmoid function $\sigmoid$ (see Figure~\ref{fig:sigmoid}) can be added to the linear output unit to render $z^{(L)}$ a probability 
\begin{equation}
\nfwidetilde y = y^{(L)} = a (z^{(L)}) , \qquad 
a (z^{(L)}) = \sigmoid (z^{(L)}) .
\end{equation}

In case more than two categories occur in a classification problem, a neural network is trained to estimate the probability distribution over the discrete number ($k > 2$) of classes.
Such distribution is referred to as \emph{multinoulli} or \emph{categorial} distribution, which is parameterized by the conditional probabilities $p_i = p(y = i \, | \, \xv) \in [0, 1], i = 1, \ldots, k$ of an input $\xv$ belonging to the $i$-th category. The output of the neural network accordingly is a $k$-dimensional vector $\nfwidetilde \by \in \mathbb R^{k \times 1}$, where $\nfwidetilde y_i = p(y = i \, | \, \xv)$.
In addition to the requirement of each component $\nfwidetilde y_i$ being in the range $[0, 1]$, we must also guarantee that all components sum up to 1 to satisfy the definition of a probability distribution.

\begin{figure}[h]
	\centering
	% 2022.11.02
	% see file psfrag.tex for the tikz figure, which is replaced here by the screenshot to accelerate the compilation
	\includegraphics[width=0.8\linewidth]{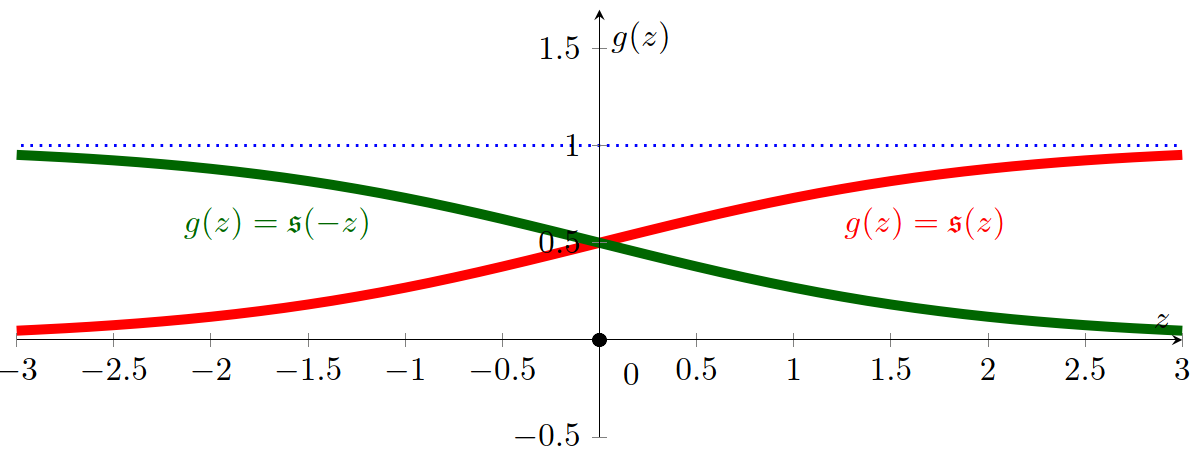}
	\caption{
		%		Activation function: 
		\emph{Sofmax function for two classes, logistic sigmoid} (Section~\ref{sc:classification}, \ref{sc:logistic-sigmoid}):
		$\sigmoid (z) = [ 1 + \exp(-z) ]^{-1}$ and $\sigmoid (-z) = [ 1 + \exp(z) ]^{-1}$, such that $\sigmoid (z) + \sigmoid (-z) = 17$.
		See also Figure~\ref{fig:sigmoid}.
	}
	\label{fig:sigmoid-z-minus-z}
\end{figure}

For this purpose, the idea of \emph{exponentiation and normalization}, which can be expressed as a change of variable in the logistic sigmoid function $\sigmoid$ (Figure~\ref{fig:sigmoid}, Section~\ref{sc:logistic-sigmoid}), as in the following example \cite{Goodfellow.2016}, p.~177:
\begin{align}
	&
	p(y) = \sigmoid [(2y - 1) z] = \frac{1}{1 + \exp [- (2y - 1) z] }\ , \text{ with } y \in \{0, 1\} \text{ and } z = \text{constant} \ ,
	\label{eq:normalized-exponential-1}
	\\
	&
	p(0) + p(1) = \sigmoid (-z) + \sigmoid (z) = \frac{1}{1 + \exp(z)} + \frac{1}{1 + \exp(-z)} = 1
	\ ,
	\label{eq:normalized-exponential-2}
\end{align}
and is generalized then to vector-valued outputs; see also Figure~\ref{fig:sigmoid-z-minus-z}.
%{\color{red} NOTE: 2022.10.29 - need reference here.  At least, mention  \cite{Goodfellow.2016}, p.~177 (pdf p.200). Please check.   ENDNOTE}

The \emph{softmax function} converts the vector formed by a linear unit $\zv^{(L)} = \boldsymbol W^{(L)} \yv^{(L-1)} + \bv^{(L)} \in \real^k$ into the vector of probabilities $\nfwidetilde \yv$ by means of
\begin{equation}
	\label{eq:softmax}
\nfwidetilde y_i = (\softmax \zv^{(L)})_i = \frac{\exp z_i^{(L)}}{\sum_{j=1}^{j=k}  \exp z_j^{(L)}} \ ,
%\nfwidetilde y_i = a( \zv^{(L)} ) , \qquad
%a( \zv^{(L)} )_i = (\softmax \zv^{(L)})_i = \frac{\exp z_i^{(L)}}{\sum_{j=1}^{j=k}  \exp z_j^{(L)}} .
\end{equation}
and is a smoothed version of the max function \cite{bishop2006pattern}, p.~198.\footnote{
	See also \cite{Goodfellow.2016}, p.~179 and p.~78, where the softmax function is used to stabilize against the underflow and overflow problem in numerical computation.
}

\begin{rem}
	\label{rm:softmax}
	Softmax function from Bayes' theorem.
	{\rm
		For a classification with multiple classes $\{ \classs{k}, k = 1, \ldots , K \}$, particularized to the case of two classes with $K = 2$, the probability for class $\classs{1}$, given the input column matrix $\bx$, is obtained from Bayes' theorem\footnote{
			Since the probability of $x$ and $y$ is $p(x,y) = p(y,x)$, and since $p(x,y) = p(x|y) p(y)$ (which is the product rule), where $p(x|y)$ is the probability of $x$ given $y$, we have $p(x|y) p(y) = p(y|x) p(x)$, and thus $p(y|x) = p(x|y) p(y) / p(x)$.  The sum rule is $p(x) = \sum_y p(x,y)$.  See, e.g., \cite{bishop2006pattern}, p.~15.  
			The right-hand side of the second equation in Eq.~\eqref{eq:sigmoid-Bayes-1}$_2$ makes common sense in terms of the predator-prey problem, in which $p(\bx , \classs{1})$ would be the percentage of predator in the total predator-prey population, and $p(\bx, \classs{2})$ the percentage of prey, as the self-proclaimed ``best mathematician of France'' Laplace said  ``probability theory is nothing but common sense reduced to calculation'' \cite{bishop2006pattern}, p.~24.
		}  as follows (\cite{bishop2006pattern}, p.~197):
		\begin{align}
			&
			p(\classs{1} | \bx) 
			= 
			\frac{p(\bx | \classs{1})  p(\classs{1})}{ p(\bx) }
			= 
			\frac{p(\bx , \classs{1})}{ p(\bx , \classs{1}) + p(\bx , \classs{2}) }
			= 
			\frac{\exp (z)}{\exp (z) + 1}
			= 
			\frac{1}{1 + \exp (-z)}
			= \sigmoid (z)
			\ , 
			\label{eq:sigmoid-Bayes-1}
			\\
			&
			\text{ with } z := \ln \frac{p(\bx , \classs{1})}{p(\bx , \classs{2})} 
			\Rightarrow
			\frac{p(\bx , \classs{1})}{p(\bx , \classs{2})} = \exp (z)
			\ ,
			\label{eq:sigmoid-Bayes-2}
		\end{align}
		where the product rule was applied to the numerator of Eq.~\eqref{eq:sigmoid-Bayes-1}$_2$, the sum rule to the denominator, and $\sigmoid$ the logistic sigmoid.
		Likewise,
		\begin{align}
			&
			p(\classs{2} | \bx) 
			= 
			\frac{p(\bx , \classs{2})}{ p(\bx , \classs{1}) + p(\bx , \classs{2}) }
			=
			\frac{1}{\exp (z) + 1} = \sigmoid (-z)
			\\
			&
			p(\classs{1} | \bx) + p(\classs{2} | \bx) = \sigmoid (z) + \sigmoid (-z) = 1
			\ ,
		\end{align}
		as in Eq.~\eqref{eq:normalized-exponential-2}, and $\sigmoid$ is also called a normalized exponential or softmax function for $K=2$.  Using the same procedure, for $K > 2$, the softmax function (version 1) can be written as\footnote{
			See also \cite{bishop2006pattern}, p.~115, version 1 of  softmax function, i.e., $\mu_k = \sum \exp (\eta_k) / [1 + \sum_j \exp \eta_j]$ and $\sum_k \mu_k \le 1$, had ``1'' as a summand in the denominator, similar to Eq.~\eqref{eq:softmax-v1} while version 2 did not, similar to Eq.~\eqref{eq:softmax-v2} \cite{bishop2006pattern}, p.~198, and was the same as Eq.~\eqref{eq:softmax}.
		}
		\begin{align}
			p (\classs{i} | \bx) =
			[{1 + \sum_{j = 1 , j \ne i}^K \exp (- z_{ij})}]^{-1}
			\ , \text{ with }
			z_{ij} := \ln \frac{p (\bx , \classs{i})}{p (\bx , \classs{j})}
			\ .
			\label{eq:softmax-v1}
		\end{align}
		Using a different definition, the softmax function (version 2) can be written as
		\begin{align}
			p (\classs{i} | \bx) = 
			\frac{\exp (z_i)}{\sum_{j=1}^K \exp (z_j)} 
			\ , \text{ with }
			z_i := \ln p(\bx , \classs{i})
			\ ,
			\label{eq:softmax-v2}
		\end{align}
		which is the same as Eq.~\eqref{eq:softmax}. 
	}
	$\hfill\blacksquare$
\end{rem}

\subsection{Gradient of cost function by backpropagation}
\label{sc:gradient}
The gradient of a cost function $J(\bparam)$ with respect to the parameters $\bparam$ is obtained using the chain rule of differentiation, and backpropagation is an efficient way to compute the chain rule.
In the forward propagation, the computation (or function composition) moves from the first layer $(1)$ to the last layer $(L)$; in the backpropagation, the computation moves in reverse order, from the last layer $(L)$ to the first layer $(1)$. 

\begin{rem}
	\label{rm:backprop}
	{\rm 
		We focus our attention on developing backpropagation for fully-connected networks, for which
		%
		% CMES style, rewriting 
%		\cite{Goodfellow.2016} did not provide 
		an explicit derivation was not provided in \cite{Goodfellow.2016}, but would help clarify the pseudocode.\footnote{
			See \cite{Goodfellow.2016}, Section 6.5.4, p.~206, Algorithm 6.4.
		}  
		The approach in \cite{Goodfellow.2016} was based on computational graph, which would not be familiar to first-time learners from computational mechanics, albeit more general in that it was applicable to networks with more general architecture, such as those with skipped connections, which require keeping track of parent and child processing units for constructing the path of backpropagation.
		See also Appendix~\ref{app:backprop-pseudocode} where the backprop Algorithm~\ref{algo:backprop} below is rewritten in a different form to explain the equivalent Algorithm 6.4 in \cite{Goodfellow.2016}, p.~206. 
	}
	$\hfill\blacksquare$
\end{rem}

\begin{figure}[h]
	\centering
	%
	% 2022.12.17
	% add "-eps-converted-to.pdf" for arXiv
	% \includegraphics[width=0.45\linewidth]{figures/network6a}
	\includegraphics[width=0.45\linewidth]{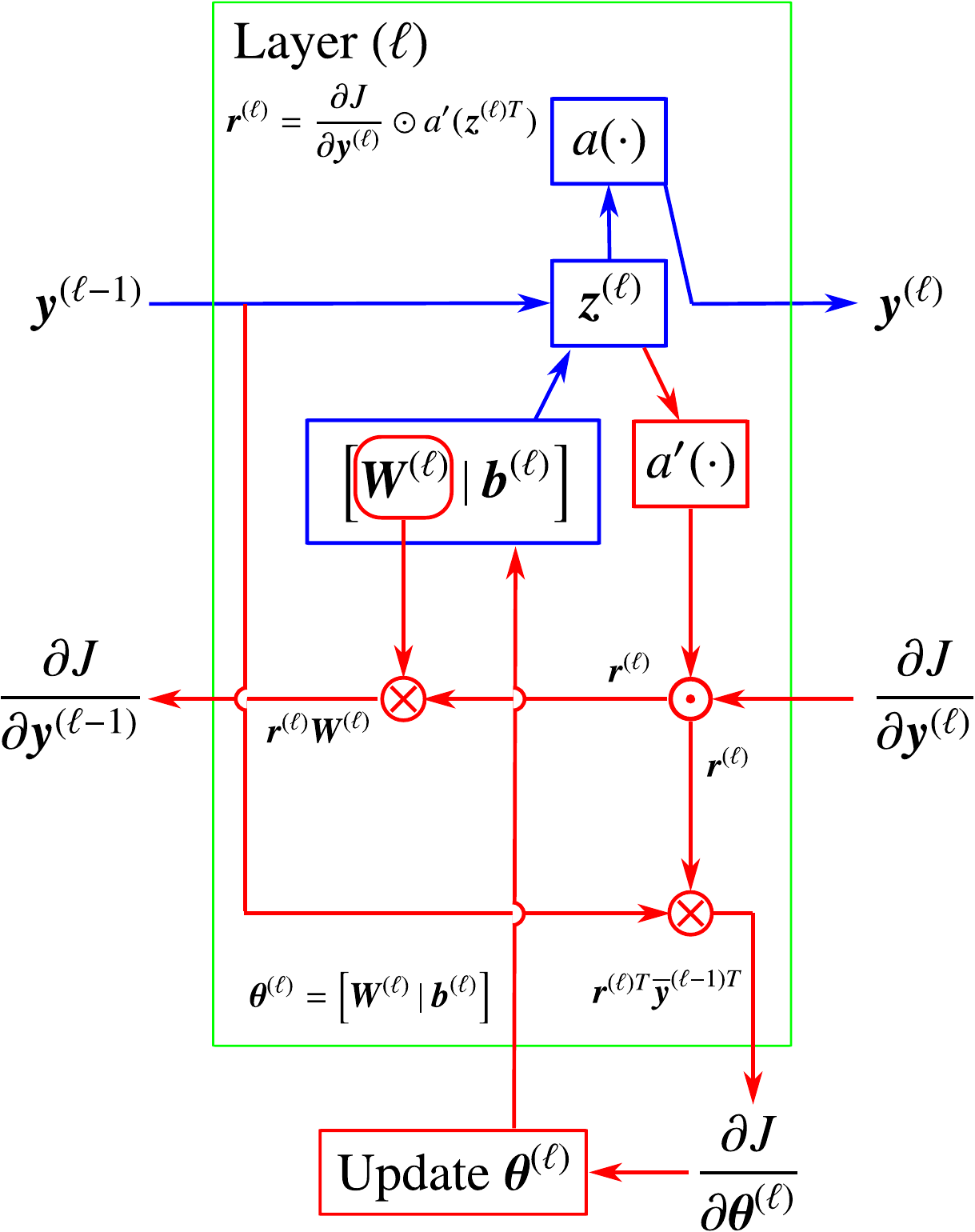}
	\caption{
		\emph{Backpropagation building block, typical layer $(\ell)$} (Section~\ref{sc:gradient}, Algorithm~\ref{algo:backprop}, Appendix~\ref{app:backprop-pseudocode}).
		The forward propagation path is shown in blue, with the backpropagation path in red.  The update of the parameters $\bparamp{\ell}$ in layer $(\ell)$ is done as soon as the gradient $\partial J / \partial \bparamp{\ell}$ is available using a gradient descent algorithm.  The row matrix $\boldsymbol{r}^{(\ell)} = \partial J / \partial \bzp{\ell}$ in Eq.~(\ref{eq:gradient-common}) can be computed once for use to evaluate both the gradient $\partial J / \partial \bparamp{\ell}$ in Eq.~(\ref{eq:gradient-7}) and the gradient $\partial J / \partial \byp{\ell-1}$ in Eq.~(\ref{eq:gradient-y(l-1)-4}), then discarded to free up memory.  See pseudocode in Algorithm~\ref{algo:backprop}.
	}.
	\label{fig:backprop-1}
\end{figure}

It is convenient to recall here some equations developed earlier (keeping the same equation numbers) for the computation of the gradient $\partial J / \partial \bparamp{\ell}$ of the cost function $J(\bparam)$ with respect to the parameters $\bparamp{\ell}$ in layer $(\ell)$, going backward from the last layer $\ell = L , \cdots , 1$.

\noindent
$\bullet$ Cost function $J(\theta)$:
\begin{align}
	J (\bparam)
	=
	\frac12 \, \text{MSE}	
	= 
	\frac{1}{2 \bsize} \parallel \by - \bout \parallel^2
	=
	\frac{1}{2 \bsize} \sum_{k=1}^{k=\bsize} \left( \y_k - \out_k \right)^2
	\ , 
	\tag{\ref{eq:half-MSE}}
\end{align}
$\bullet$ Inputs $\bx = \byp{0} \in \real^{\widths{0} \times 1}$ with $\widths{0} = n$ and predicted outputs $\byp{L} = \bout \in \real^{\widths{L} \times 1}$ with $\widths{0} = m$:
\begin{align}
	&
	\boldsymbol y^{(0)}
	=
	\boldsymbol x^{(1)} 
	= 
	\boldsymbol x \text{ (inputs) }
	\nonumber
	\\
	&
	\boldsymbol y^{(\ell)}
	= 
	f^{(\ell)} (\boldsymbol x^{(\ell)})
	=
	\boldsymbol x^{(\ell + 1)}
	\text{ with }
	\ell = 1 , \ldots , \D-1
	\nonumber
	\\
	&
	\boldsymbol y^{(\D)}
	= 
	f^{(\D)} (\boldsymbol x^{(\D)})
	= 
	\bout  \text{ (predicted outputs) }
	\tag{\ref{eq:input-predicted-output}}
\end{align}
$\bullet$ Weighted sum of inputs and biases $\boldsymbol z^{(\ell)} \in \real^{\widths{\ell} \times 1}$:
\begin{align}
	%\boxit{
		\boldsymbol z^{(\ell)}
		=
		\boldsymbol W^{(\ell)} \boldsymbol y^{(\ell-1)}
		+
		\boldsymbol b^{(\ell)}
		\text{ such that }
		z^{(\ell)}_i
		=
		\boldsymbol w^{(\ell)}_i \boldsymbol y^{(\ell-1)}
		+
		b^{(\ell)}_i
		\ , \text{ for }
		i = 1, \ldots , m_{(\ell)}
		\ ,
	%}
	\tag{\ref{eq:linearComboInputsBias}}
\end{align}

\begin{figure}[h]
	\centering
	%
	% 2022.12.17
	% add "-eps-converted-to.pdf" for arXiv
	% \includegraphics[width=1.0\linewidth]{figures/network6b}
	\includegraphics[width=1.0\linewidth]{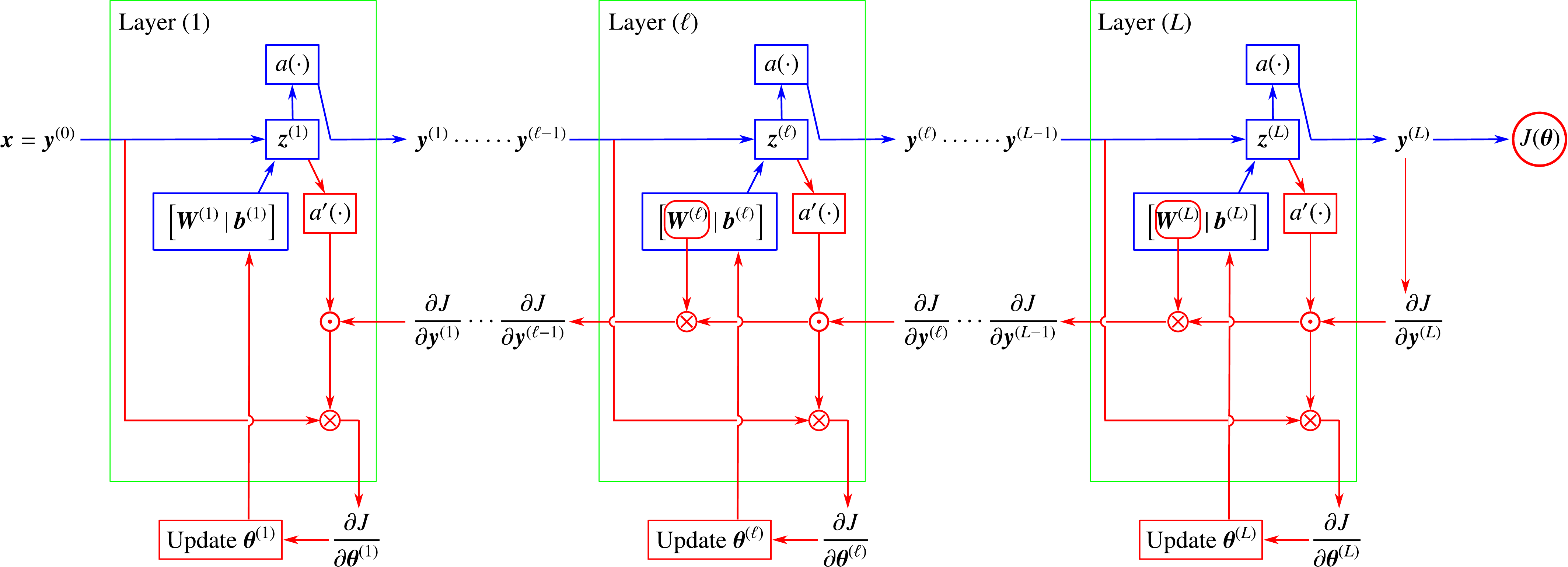}
	\caption{
		\emph{Backpropagation in fully-connected network} (Section~\ref{sc:gradient}, \ref{sc:vanish-grad}, Algorithm~\ref{algo:backprop}, Appendix~\ref{app:backprop-pseudocode}). 
		Starting from the predicted output $\widetilde{\by} = \byp{L}$
		In the last layer $(L)$ at the end of any forward propagation (blue arrows), and going backward (red arrows) to the first layer with $\ell = L , \cdots, 1$, and along the way at layer $(\ell)$, compute the gradient of the cost function $J$ relative the the parameters $\bparamp{\ell}$ to update those parameters in a gradient descent, then compute the gradient of $J$ relative to the outputs $\byp{\ell-1}$ of the lower-level layer $(\ell-1)$ to continue the backpropagation.
		See pseudocode in Algorithm~\ref{algo:backprop}. 
		For a particular example of the above general case, see Figure~\ref{fig:backprop-NN4}.
	}
	\label{fig:backprop-2}
\end{figure}

\noindent
$\bullet$ Network parameters $\bparam$ and layer parameters $\bparamp{\ell} \in \real^{\widths{\ell} \times [\widths{\ell-1} + 1] }$: 
\begin{align}
	\bParamp \ell
	=
	\left[ 
	\paramsp {ij} \ell
	\right]
	=
	\begin{bmatrix}
	\bparamsp {1} \ell
	\\
	\vdots
	\\
	\bparamsp {m_{(\ell)}} \ell
	\end{bmatrix}
	=
	\left[
	\left.
	\boldsymbol W^{(\ell)} 
	\ \right| \ 
	\boldsymbol b^{(\ell)}
	\right]
	\in \mathbb{R}^{m_{(\ell)} \times [m_{(\ell-1)} + 1]}
	\tag{\ref{eq:parameters}}
\end{align}
\begin{align}
	\bparam 
	=
	\{ \bparamp 1 ,\cdots, \bparamp \ell , \cdots , \bparamp L \}
	= 
	\{ \bParamp 1 ,\cdots, \bParamp \ell , \cdots , \bParamp L \}
	\, , \text{ such that }
	\bparamp \ell \equiv \bParamp \ell
	\tag{\ref{eq:theta}}
\end{align}
$\bullet$ Expanded layer outputs $\expand{\by}^{(\ell-1)} \in \real^{[\widths{\ell-1} + 1] \times 1}$:
\begin{align}
	\boldsymbol z^{(\ell)}
	=
	\boldsymbol W^{(\ell)} \boldsymbol y^{(\ell-1)}
	+
	\boldsymbol b^{(\ell)}
	\equiv
	\left[
	\left.
	\boldsymbol W^{(\ell)} 
	\ \right| \ 
	\boldsymbol b^{(\ell)}
	\right]
	\begin{bmatrix}
	\boldsymbol y^{(\ell-1)}
	\\
	1
	\end{bmatrix}
	=:
	\bparamp \ell
	\ 
	\expand{\by}^{(\ell-1)}
	\ ,
	\tag{\ref{eq:expanded-output-1}}
\end{align}
$\bullet$ Activation function $\g(\cdot)$:
\begin{align}
	%\boxit{
		\boldsymbol y^{(\ell)}
		=
		\g (\boldsymbol z^{(\ell)})
		\text{ such that } y^{(\ell)}_i = \g (z^{(\ell)}_i)
	%}
	\tag{\ref{eq:activationFunction}}
\end{align}

The gradient of the cost function $J(\bparam)$ with respect to the parameters $\bparamp{\ell}$ in layer $(\ell)$, for $\ell = L , \cdots, 1$, is simply:
\begin{align}
	\frac{\partial J (\bparam)}{\partial \bparamp{\ell}} 
	&
	= 
	\frac{\partial J (\bparam)}{\partial \byp{\ell}}
	\frac{\partial \byp{\ell}}{\partial \bparamp{\ell}} 
	\Leftrightarrow
	\frac{\partial J}{\partial \paramsp{ij}{\ell}}
	=
	\sum_{k=1}^{\widths{\ell}} 
	\frac{\partial J}{\partial \ysp{k}{\ell}}
	\frac{\partial \ysp{k}{\ell}}{\partial \paramsp{ij}{\ell}}
	\label{eq:gradient-1}
	\\
	\frac{\partial J (\bparam)}{\partial \bparamp{\ell}}
	&
	=
	\left[ \frac{\partial J}{\partial \paramsp{ij}{\ell}} \right]
	\in \real^{\widths{\ell} \times [\widths{\ell-1} + 1]}
	\ , \quad
	\frac{\partial J (\bparam)}{\partial \byp{\ell}} 
	=
	\left[ \frac{\partial J}{\partial \ysp{k}{\ell}} \right]
	\in \real^{1 \times \widths{\ell}}
	\text{ (row)}
	\label{eq:gradient-2}
\end{align}

\begin{algorithm}[h]
	{\bf Backpropagation pseudocode}
	\\
	\KwData{
		% Layer outputs $\byp{\ell}$, for $\ell=L, \cdots , 1$
		\\
		$\bullet$ Input into network $\bx = \byp{0} \in \real^{\widths{0} \times 1}$
		\\
		$\bullet$ Learning rate $\epsilon$; see Section~\ref{sc:learning-rate} on deterministic optimization  
		\\
		$\bullet$ Results from \emph{any} forward propagation: 
		\\
		\hspace{5mm} $\star$ Network parameters $\bparam = \{\bparamp{1} , \cdots , \bparamp{L} \}$ (all layers)
		\\
		\hspace{5mm} $\star$ Layer weighted inputs and biases $\bzp{\ell}$, for $\ell = 1, \cdots, L$
		\\
		\hspace{5mm} $\star$ Layer outputs $\byp{\ell}$, for $\ell = 1, \cdots, L$
	}
	\KwResult{
		Updated network parameters $\bparam$ to reduce cost function $J$.
	}
	\vphantom{Blank line}
	{\bf Initialize:} 
	\\
	$\bullet$ Gradient $\partial J / \partial \byp{L}$ relative to predicted output $\bout = \byp{L}$ of last layer $L$ using Eq.~(\ref{eq:d-cost-output-L}) 
	\\
	$\bullet$ Set layer counter $\ell$ to last layer $L$, i.e., $\ell \leftarrow L$
	\;
	\vphantom{Blank line}
	\While{$\ell > 0$, {\rm for current layer} $(\ell)$,}{
		Compute gradient $\boldsymbol{r}^{(\ell)} = \partial J / \partial \bzp{\ell}$ of cost $J$ relative to weighted sum $\bzp{\ell}$ using Eq.~(\ref{eq:gradient-common})
		\;
		Compute gradient $\bgradp{\ell} = \partial J / \partial \bparamp{\ell}$ of cost $J$ relative to layer parameters $\bparamp{\ell}$ using Eq.~(\ref{eq:gradient-7})
		\;
		Update layer parameter $\bparamp{\ell}$ to decrease cost $J$ using gradient descent Eq.~(\ref{eq:update-params}), Section~\ref{sc:deterministic-optimization} \label{lst:line:update-params-backprop}
		\;
		Compute gradient $\partial J / \partial \byp{\ell-1}$ of cost $J$ relative to outputs $\byp{\ell-1}$ using Eq.~(\ref{eq:gradient-y(l-1)-4})
		\;
		Decrement layer counter by one: $\ell \leftarrow \ell - 1$
		\;
		Propagate to lower-level layer $(\ell-1)$
		\;
	}
	\vphantom{Blank line}
	\caption{
		\emph{Backpropagation pseudocode} (Section~\ref{sc:gradient}).
		Compute the gradient of cost function $J$ relative to the parameters $\bparamp{\ell}$, with $\ell = L , \cdots , 1$, by backpropagation in one step of a gradient-descent algorithm to find parameters that decrease the cost function.  The focus here is backpropagation, not the overall optimization.  So the algorithm starts at the end of \emph{any} forward propagation  to begin backpropagation, from the last layer $(L)$ back to the first layer $(1)$, to update the parameters to decrease the cost function.  See also the block diagrams in Figure~\ref{fig:backprop-1} and Figure~\ref{fig:backprop-2}, and
		Appendix~\ref{app:backprop-pseudocode} where an alternative backprop Algorithm~\ref{algo:backprop-2} is used to explain the equivalent Algorithm 6.4 in \cite{Goodfellow.2016}, p.~206.
	}
	\label{algo:backprop}
\end{algorithm}

The above equations are valid for the last layer $\ell = L$, since since the predicted output $\bout$ is the same as the output of the last layer $(L)$, i.e., $\bout \equiv \byp{L}$ by Eq.~(\ref{eq:input-predicted-output}).  Similarly, these equations are also valid for the first layer $(1)$ since the input for layer $(1)$ is $\bxp{1} = \bx = \byp{0}$.  using Eq.~(\ref{eq:activationFunction}), we obtain (no sum on $k$)
\begin{align}
	\frac{\partial \ysp{k}{\ell}}{\partial \paramsp{ij}{\ell}}
	= 	
	a^\prime (\bzsp{k}{\ell})
	\frac{\partial \zsp{k}{\ell}}{\partial \paramsp{ij}{\ell}}
	=
	\sum_p a^\prime (\bzsp{k}{\ell})
	\frac{\partial \paramsp{kp}{\ell}}{\partial \paramsp{ij}{\ell}}
	\expand{\y}_p^{(\ell-1)}
	=
	\sum_p a^\prime (\bzsp{k}{\ell})
	\delta_{ki} \delta_{pj}
	\expand{\y}_p^{(\ell-1)}
	=
	a^\prime (\bzsp{k}{\ell})
	\delta_{ki}
	\expand{\y}_j^{(\ell-1)}
	%\text{ (no sum on } k)
	\label{eq:gradient-3}
\end{align}
Using Eq.~(\ref{eq:gradient-3}) in Eq.~(\ref{eq:gradient-1}) leads to the expressions for the gradient, both in component form (left) and in matrix form (right):
\begin{align}
	\frac{\partial J}{\partial \paramsp{ij}{\ell}}
	&
	=
	\frac{\partial J}{\partial \ysp{i}{\ell}} a^\prime (\bzsp{i}{\ell})
	\expand{\y}_j^{(\ell-1)}
	\  \text{(no sum on $i$)}
	\Leftrightarrow
	\left[ \frac{\partial J}{\partial \paramsp{ij}{\ell}} \right]
	=
	\left[
		\left[ \frac{\partial J}{\partial \ysp{i}{\ell}} \right]^T
		\odot
		\left[ a^\prime (\bzsp{i}{\ell}) \right]
	\right]
	\left[ \expand{\y}_j^{(\ell-1)} \right]^T
	\ ,
	\label{eq:gradient-4}
\end{align}
where $\odot$ is the elementwise multiplication, known as the Hadamard operator, defined as follows:
\begin{align}
	\left[ p_i \right]
	\ , 
	\left[ q_i \right]
	\in
	\real^{m \times 1}
	\Rightarrow
	\left[ p_i \right]
	\odot
	\left[ q_i \right]
	=
	\left[ (p_i q_i) \right]
	=
	\left[ p_1 q_1 , \cdots , p_m q_m \right]^T
	\in
	\real^{m \times 1}
	\text{ (no sum on $i$)}
	\ ,
	\label{eq:hadamard}
\end{align}
and
\begin{align}
	&
	\left[ \frac{\partial J}{\partial \ysp{i}{\ell}} \right]
	\in \real^{1 \times \widths{\ell}}
	\text{ (row)}
	\ , \ 
	\bzp{\ell} 
	= \left[\bzsp{k}{\ell}\right]
	\in 
	\real^{\widths{\ell} \times 1}
	\text{ (column)}
	\Rightarrow
	a^\prime (\bzsp{i}{\ell})
	\in 
	\real^{\widths{\ell} \times 1}
	\text{ (column)}
	\\
	&
	%\Rightarrow
	\left[
		\left[ \frac{\partial J}{\partial \ysp{i}{\ell}} \right]^T
		\odot
		\left[ a^\prime (\bzsp{i}{\ell}) \right]
	\right]
	\in 
	\real^{\widths{\ell} \times 1}
	\text{ (column, no sum on $i$)}
	\ ,
	\text{ and}
	\left[ \expand{\y}_j^{(\ell-1)} \right]
	\in
	\real^{[\widths{\ell - 1} + 1] \times 1}
	\text{ (column)}
	\label{eq:gradient-5}
	\\
	&
	\Rightarrow
	\left[ \frac{\partial J}{\partial \paramsp{ij}{\ell}} \right]
	=
	\left[
		\left[ \frac{\partial J}{\partial \ysp{i}{\ell}} \right]^T
		\odot
		\left[ a^\prime (\bzsp{i}{\ell}) \right]
	\right]
	\left[ \expand{\y}_j^{(\ell-1)} \right]^T
	\in\real^{\widths{\ell} \times [\widths{\ell - 1} + 1] }
	\ ,
	\label{eq:gradient-6}
\end{align}
which then agrees with the matrix dimension in the first expression for ${\partial J} / {\partial \bparamp{\ell}}$ in Eq.~(\ref{eq:gradient-2}).  For the last layer $\ell = L$, all terms on the right-hand side of Eq.~(\ref{eq:gradient-6}) are available for the computation of the gradient $\partial J / \partial \paramsp{ij}{L}$ since 
\begin{align}
	\byp{L} = \bout
	\Rightarrow
	\frac{\partial J}{\partial \byp{L}}
	=
	\frac{\partial J}{\partial \bout}
	=
	\frac{1}{\bsize}
	\sum_{k=1}^{k=\bsize}
	\left( \bout^{|k|} - \bexout{k} \right)
	\in \real^{1 \times \widths{L}}
	\text{ (row)}   
	\label{eq:d-cost-output-L}
\end{align}
with $m$ being the number of examples and $\widths{L}$ the width of layer $(L)$, is the mean error from the expression of the cost function in Eq.~(\ref{eq:half-MSE-2}), with $\zsp{i}{L}$ and $\expand{\y}_j^{(L-1)}$ already computed in the forward propagation.  To compute the gradient of the cost function with respect to the parameters $\bparamp{L-1}$ in layer $(L-1)$, we need the derivative $\partial J / \partial \ysp{i}{L-1}$, per Eq.~(\ref{eq:gradient-6}).
Thus, in general, the derivative of cost function $J$ with respect to the output matrix $\byp{\ell-1}$ of layer $(\ell-1)$, i.e., $\partial J / \partial \ysp{i}{\ell-1}$, can be expressed in terms of the previously computed derivative $\partial J / \partial \ysp{i}{\ell}$ and other quantities for layer $(\ell)$ as follows:  
\begin{align}
	&
	\frac{\partial J}{\partial \ysp{i}{\ell-1}}
	=
	\sum_k
	\frac{\partial J}{\partial \ysp{k}{\ell}}
	\frac{\partial \ysp{k}{\ell}}{\partial \ysp{i}{\ell-1}}
	=
	\sum_k
	\frac{\partial J}{\partial \ysp{k}{\ell}}
	a^\prime (\zsp{k}{\ell}) \weightsp{ki}{\ell}
	=
	\sum_k
	\frac{\partial J}{\partial \zsp{k}{\ell}}
	\frac{\partial \zsp{k}{\ell}}{\partial \ysp{i}{\ell-1}}
	\\
	&
	\left[ \frac{\partial J}{\partial y_i^{(\ell-1)}} \right]
	=
	\left[ 
		\left[ \frac{\partial J}{\partial y_k^{(\ell)}} \right] 
		\odot
		\left[ a^\prime ( z_k^{(\ell)} ) \right]^T 
	\right]
	\left[ \weightsp{ki}{\ell}\right]
	\in \real^{1 \times \widths{\ell-1}}
	\text{ (no sum on $k$)}
	\label{eq:gradient-y(l-1)-1}
	\\
	&
	\left[ \frac{\partial J}{\partial y_i^{(\ell-1)}} \right] 
	\in \real^{1 \times \widths{\ell-1}} 
	\ , \quad
	\left[ \frac{\partial J}{\partial y_k^{(\ell)}} \right]
	\in \real^{1 \times \widths{\ell}}
	\text{ (row)}
	\ , \quad
	\left[ a^\prime ( z_k^{(\ell)} ) \right]
	\in \real^{\widths{\ell} \times 1}
	\text{ (column)}
	\label{eq:gradient-y(l-1)-2}
	\\
	&
	\left[ \frac{\partial J}{\partial y_k^{(\ell)}} \right] 
	\odot
	\left[ a^\prime ( z_k^{(\ell)} ) \right]^T
	=
	\frac{\partial J}{\partial \zsp{k}{\ell}}
	\in \real^{1 \times \widths{\ell}}
	\text{ (no sum on $k$)}
	\ , \quad
	\left[ \weightsp{ki}{\ell}\right]
	=
	\frac{\partial \zsp{k}{\ell}}{\partial \ysp{i}{\ell-1}}
	\in \real^{\widths{\ell} \times \widths{\ell-1}}
	\ .
	\label{eq:gradient-y(l-1)-3}
\end{align}
Comparing Eq.~(\ref{eq:gradient-y(l-1)-1}) and Eq.~(\ref{eq:gradient-6}), when backpropagation reaches layer $(\ell)$, the same row matrix
\begin{align}
	\boldsymbol r^{(\ell)}
	:=
	\frac{\partial J}{\partial \bzp{\ell}}
	=
	\left[
	\left[ \frac{\partial J}{\partial \ysp{i}{\ell}} \right]
	\odot
	\left[ a^\prime (\bzsp{i}{\ell}) \right]^T
	\right]
	=
	\frac{\partial J}{\partial \byp{\ell}}
	\odot 
	\g^\prime ( \bz^{(\ell) T} )
	\in 
	\real^{1 \times \widths{\ell}}
	\text{ (row)}
	\label{eq:gradient-common}
\end{align}
is only needed to be computed once for use to compute both the gradient of the cost $J$ relative to the parameters $\bparamp{\ell}$  [see Eq.~(\ref{eq:gradient-6}) and Figure~\ref{fig:backprop-1}]
\begin{align}
	\frac{\partial J}{\partial \bparamp{\ell}}
	=
	\boldsymbol{r}^{(\ell) T}
	\expand{\by}^{(\ell-1) T}
	\in
	\real^{\widths{\ell} \times [\widths{\ell-1}  + 1]}
	\label{eq:gradient-7}
\end{align} 
and the gradient of the cost $J$ relative to the outputs $\byp{\ell-1}$ of layer $(\ell-1)$ [see Eq.~(\ref{eq:gradient-y(l-1)-1})  and Figure~\ref{fig:backprop-1}]
\begin{align}
	\frac{\partial J}{\partial \byp{\ell-1}}
	=
	\boldsymbol{r}^{{\ell}}
	\bWeight^{(\ell)}
	\in
	\real^{1 \times \widths{\ell -1}}
	\ .
	\label{eq:gradient-y(l-1)-4}
\end{align}

The block diagram for backpropagation at layer $(\ell)$---as described in Eq.~(\ref{eq:gradient-common}), Eq.~(\ref{eq:gradient-7}), Eq.~(\ref{eq:gradient-y(l-1)-4})---is given in Figure~\ref{fig:backprop-1}, and for a fully-connected network in Figure~\ref{fig:backprop-2}, with pseudocode given in Algorithm~\ref{algo:backprop}.

\subsection{Vanishing and exploding gradients}
\label{sc:vanish-grad}

To demonstrate the vanishing gradient problem,
%
% CMES style, rewriting
%\cite{Nielsen.2015} employed 
a network is used in \cite{Nielsen.2015}, having an input layer containing 784 neurons, corresponding to the $28 \times 28 = 784$ pixels in the input image, four hidden layers, with each hidden layer containing 30 neurons, and an output layer containing 10 neurons, corresponding to the 10 possible classifications for the MNIST digits ('0', '1', '2', ... , '9').  
A key ingredient is the use of the sigmoid function as active function; see Figure~\ref{fig:sigmoid}. 
 
We note immediately that the vanishing / exploding gradient problem can be resolved using the rectified linear function (ReLu, Figure~\ref{fig:ReLU}) as active function in combination with ``normalized initialization''\footnote{
	See \cite{Goodfellow.2016}, p.~295.
} and ``intermediate normalization layers'', which are mentioned in \cite{He.2015:rd0001}, and which we will not discuss here.

\begin{figure}[h]
	\centering
	\includegraphics[width=0.7\linewidth]{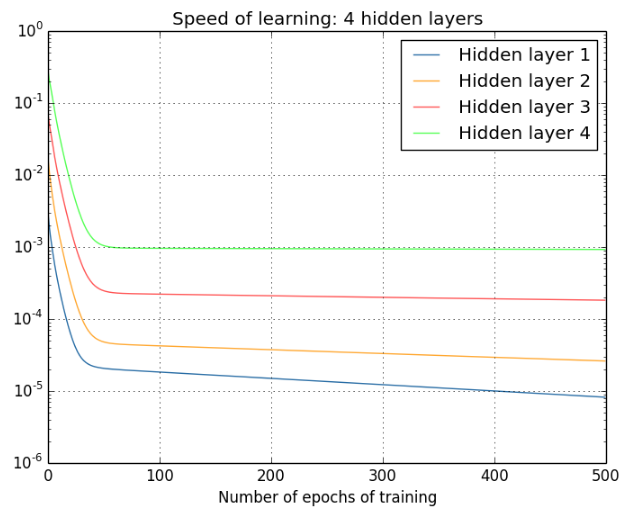}
	\caption{
		\emph{Vanishing gradient problem} (Section~\ref{sc:vanish-grad}).  Speed of learning of earlier layers is much slower than that of later layers.  Here, after $400$ epochs of training, the speed of learning of Layer (1) at $10^{-5}$ (blue line) is 100 times slower than that of Layer (4) at $10^{-3}$ (green line);
		\cite{Nielsen.2015}, Chapter 5, `Why are deep neural networks hard to train ?'
		\href{https://creativecommons.org/licenses/by-nc/3.0/deed.en_GB}{(CC BY-NC 3.0)}.
	}
	\label{fig:vanish-grad-a}
\end{figure}

The speed of learning of a hidden layer $(\ell)$ in Figure~\ref{fig:vanish-grad-a} is defined as the norm of the gradient $\bgradp \ell$ of the cost function $J(\bparam)$ with respect to the parameters $\bparamp \ell$ in the hidden layer $(\ell)$:
\begin{align}
	\parallel
	\bgradp \ell 
	\parallel
	= 
	\left|\kern-1.0pt\left|
	\frac{\partial J}{\partial \bparamp{\ell}}  
	\right|\kern-1.0pt\right|
\end{align}
The speed of learning in each of the four layers as a function of the number of epochs\footnote{
	An epoch is when all examples in the dataset had been used in a training session of the optimization process.  For a formal definition of ``epoch'', see Section~\ref{sc:generic-SGD} on stochastic gradient descent (SGD) and Footnote~\ref{fn:epoch}.
} of training drops down quickly after less than 50 training epochs, then plateaued out, as depicted in Figure~\ref{fig:vanish-grad-a}, where the speed of learning of layer (1) was 100 times less than that of layer (4) after 400 training epochs.

\begin{figure}[H]
	\centering
	%
	% 2022.12.17
	% add "-eps-converted-to.pdf" for arXiv
	% \includegraphics[width=0.9\linewidth]{figures/network4c}
	\includegraphics[width=0.9\linewidth]{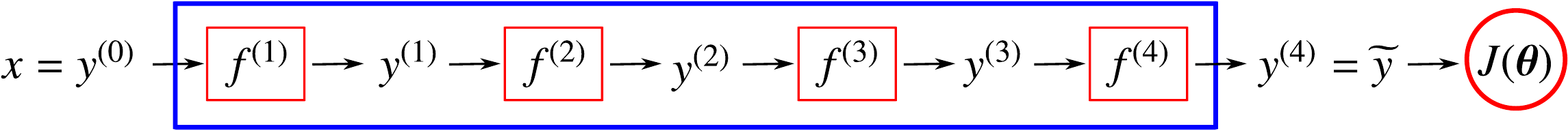}
	\caption{
		\emph{Neural network with four layers} (Section~\ref{sc:vanish-grad}), one neuron per layer, scalar input $x$, scalar output $y$, cost function $J(\boldsymbol \theta) = \frac12 (y - \widetilde{y})^2$, with $\widetilde{y} = y^{(4)}$ being the target output and also the output of layer $(4)$, such that $f^{(\ell)} (y^{(\ell - 1)}) = \g (z^{(\ell)})$, with $\g (\cdot)$ being the active function, $z^{(\ell)} = w^{(\ell)} y^{(\ell - 1)} + b^{(\ell)}$, for $\ell = 1,\ldots,4$, and the network parameters are $\boldsymbol \theta = [ w_1, \ldots , w_4 , b_1 , \ldots , b_4 ]$.
		The detailed block diagram is in Figure~\ref{fig:backprop-NN4}.
	}
	\label{fig:NN-4-layers}
\end{figure}

\begin{figure}[h]
	\centering
	%
	% 2022.12.17
	% add "-eps-converted-to.pdf" for arXiv
	% \includegraphics[width=1.0\linewidth]{figures/network4f}
	\includegraphics[width=1.0\linewidth]{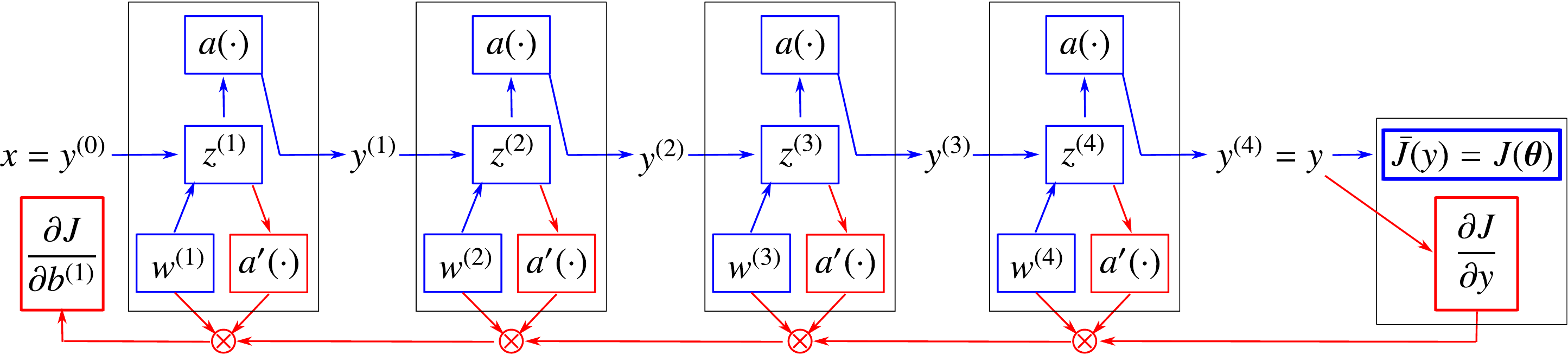}
	\caption{
		\emph{Neural network with four layers} in Figure~\ref{fig:NN-4-layers} (Section~\ref{sc:vanish-grad}). Detailed block diagram.
		Forward propagation (blue arrows) and backpropagation (red arrows).
		In the forward propagation wave, at each layer $(\ell)$, the product $a^\prime w^{(\ell)}$ is computed and stored, awaiting for the chain-rule derivative to arrive at this layer to multiply.  
		The cost function $J(\theta)$ is computed together with its derivative $\partial J / \partial y$, which is the backprop starting point, from which, when following the backpropagation red arrow, the order of the factors are as in Eq.~(\ref{eq:dJ-db1}), until the derivative $\partial J / \partial b^{(1)}$ is reached at the head of the backprop red arrow. 
		(Only the weights are shown, not the biases, which are not needed in the back propagation, to save space.)
		The speed of learning is slowed down significantly in early layers due to vanishing gradient, as shown in Figure~\ref{fig:vanish-grad-a}.
		See also the more general case in Figure~\ref{fig:backprop-2}.
	}
	\label{fig:backprop-NN4}
\end{figure}

To understand the reason for the quick and significant decrease in the speed of learning, consider a network with four layers, having one scalar input $x$ with target scalar output $y$, and predicted scalar output $\out$, as shown in Figure~\ref{fig:NN-4-layers}, where each layer has one neuron.\footnote{See also \cite{Nielsen.2015}.}
The cost function and its derivative are
\begin{align}
	J(\bparam) = \frac12 (y - \out)^2 
	\, , \quad 
	\frac{\partial J}{\partial y} = y - \out
\end{align}
The neuron in layer $(\ell)$ accepts the scalar input $y^{(\ell - 1)}$ to produce the scalar output $y^{(\ell)}$ according to
\begin{align}
	f^{(\ell)} (y^{(\ell - 1)}) = \g (z^{(\ell)}) \, , 
	\text{ with } 
	z^{(\ell)} = w^{(\ell)} y^{(\ell - 1)} + b^{(\ell)}
	\ .
\end{align}
As an example of computing the gradient, the derivative of the cost function $J$ with respect to the bias $b^{(1)}$ of layer $(1)$ is given by
\begin{align}
	\frac{\partial J}{\partial b^{(1)}}
	=
	(y - \out)
	[a^\prime ( z^{(4)} ) w^{(4)}]
	[a^\prime ( z^{(3)} ) w^{(3)}]
	[a^\prime ( z^{(2)} ) w^{(2)}]
	[a^\prime ( z^{(1)} ) w^{(1)}]
	\label{eq:dJ-db1}
\end{align}
The back propagation procedure to compute the gradient $\partial J / \partial b^{(1)}$ in Eq.~(\ref{eq:dJ-db1}) is depicted in Figure~\ref{fig:backprop-NN4}, which is a particular case of the more general Figure~\ref{fig:backprop-2}.

Whether the gradient $\frac{\partial J}{\partial b^{(1)}}$ in Eq.~(\ref{eq:dJ-db1}) vanishes or explodes depends on the magnitude of its factors
\begin{align}
	| a^\prime ( z^{(\ell)} ) w^{(\ell)} | < 1 , \  \forall \ell 
	& \Rightarrow \text{ Vanishing gradient}
	\\
	| a^\prime ( z^{(\ell)} ) w^{(\ell)} | > 1 , \ \forall \ell 
	& \Rightarrow \text{ Exploding gradient}
\end{align}
In other mixed cases, the problem of vanishing or exploding gradient could be alleviated by the changing of the magnitude $| a^\prime ( z^{(\ell)} ) w^{(\ell)} |$, above 1 and below 1, from layer to layer.

\begin{rem}
	\label{rm:vanish-gradient-MLP}
	{\rm 
		While the vanishing gradient problem for multilayer networks (static case) may be alleviated by weights that vary from layer to layer (the mixed cases mentioned above), this problem is especially critical in the case of Recurrent Neural Networks, since the weights stay constant for all state numbers (or ``time'') in a sequence of data.   
		See Remark~\ref{rm:explain-short-term} on ``short-term memory'' in Section~\ref{sc:LSTM} on Long Short-Term Memory.
		In back-propagation through the states in a sequence of data, from the last state back to the first state, the same weight keeps being multiplied by itself.   Hence, when a weight is less than 1, successive powers of its magnitude eventually decrease to zero when progressing back the first state.
	}
		$\hfill\blacksquare$
\end{rem}

\subsubsection{Logistic sigmoid and hyperbolic tangent}
\label{sc:logistic-sigmoid}
\begin{figure}[h]
	\centering
	\includegraphics[width=0.8\linewidth]{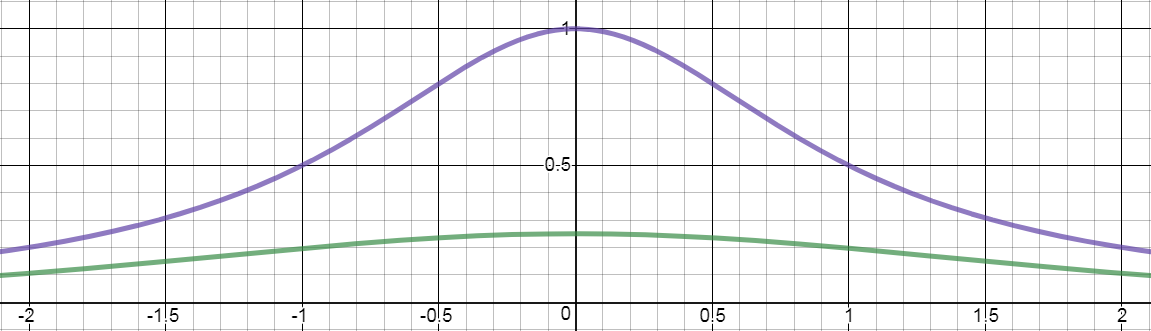}
	\caption{
		\emph{Sigmoid and hyperbolic tangent functions, derivative}
		(Section~\ref{sc:logistic-sigmoid}). The derivative of sigmoid function
		($\sigmoid^\prime (z) = \sigmoid(z) [1 - \sigmoid(z)]$, green line) is less than 1 everywhere, whereas the derivative of the hyperbolic tangent ($\tanh^\prime (z) = (1 + z^2)^{-1}$, purple line) is less than 1 everywhere, except at the abscissa $z = 0$, where it is equal to 1.
	}
	\label{fig:derivative-sigmoid-tanh}
\end{figure}

The first derivatives of the sigmoid function and hyperbolic tangent function depicted in Figure~\ref{fig:sigmoid} (also in Remark~\ref{rm:softmax} on the softmax function and Figure~\ref{fig:sigmoid-z-minus-z}) are given below:
\begin{align}
	a(z) = \sigmoid (z) = \frac{1}{1 + \exp(-z)} 
	& \Longrightarrow
	a^\prime (z) = \sigmoid^\prime (z) = \sigmoid(z) [1 - \sigmoid(z)]
	\in (0,1)
	\label{eq:logistic-sigmoid}
	\\
	a(z) = \tanh (z) 
	& \Longrightarrow
	a^\prime (z) = \tanh^\prime (z) = \frac{1}{1 + z^2}
	\in (0,1]
	\label{eq:hyperbolic-tangent}
\end{align}
and are less than 1 in magnitude (everywhere for the sigmoid function, and almost everywhere for the hyperbolic tangent tanh function), except at $z = 0$, where the derivative of the tanh function is equal to 1; Figure~\ref{fig:derivative-sigmoid-tanh}.
Successive multiplications of these derivatives will result in smaller and smaller values along the back propagation path.
If the weights $w^{(\ell)}$ in Eq.~(\ref{eq:dJ-db1}) are also smaller than 1, then the gradient $\partial J / \partial b^{(1)}$ will tend toward 0, i.e., vanish.
The problem is further exacerbated in deeper networks with increasing number of layers, and thus increasing number of factors less than 1 (i.e., $| a^\prime (z^{(\ell)}) w^{(\ell)} )| < 1$).
We have encountered the vanishing-gradient problem.

\begin{figure}[h]
	\centering
	\includegraphics[width=0.7\linewidth]{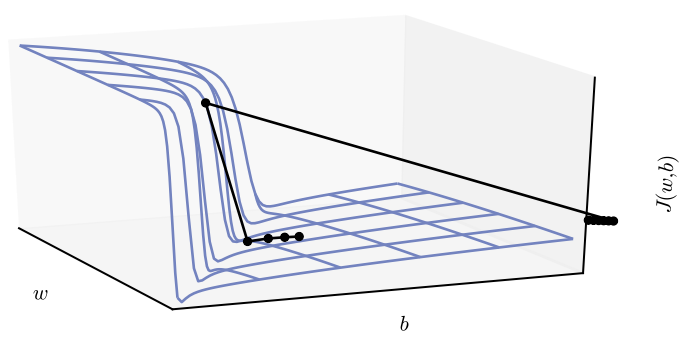}
	\caption{
		\emph{Cost-function cliff} (Section~\ref{sc:logistic-sigmoid}).
		A cliff, or a sharp drop in the cost function.  
		The parameter space is represented by a weight $w$ and a bias $b$.
		The slope at the brink of the cliff leads to large-magnitude gradients, which when multiplied with each other several times along the back propagation path would result in an exploding gradient problem. 
		\cite{Goodfellow.2016}, p.~281. 
 		% \\
		%{\small (Figure reproduced with permission of the authors.)} 
		{\footnotesize (Figure reproduced with permission of the authors.)}
	}
	\label{fig:Goodfellow-cliff}
\end{figure}

The exploding gradient problem is opposite to the vanishing gradient problem, and occurs when the gradient has its magnitude increases in subsequent multiplications, particularly at a ``cliff'', which is a sharp drop in the cost function in the parameter space.\footnote{See \cite{Goodfellow.2016}, p.~281}  
The gradient at the brink of a cliff (Figure~\ref{fig:Goodfellow-cliff}) leads to large-magnitude gradients, which when multiplied with each other several times along the back propagation path would result in an exploding gradient problem.

\subsubsection{Rectified linear function (ReLU)}
\label{sc:relu}

The rectified linear function  depicted in Figure~\ref{fig:ReLU} with its derivative (Heaviside function) equal to 1 for any input greater than zero, would resolve the vanishing-gradient problem, as 
%
% CMES style, rewriting
%\cite{Glorot.2011:rd0001} wrote: 
it is written in \cite{Glorot.2011:rd0001}:

\begin{quote}
	``For a given input only a subset of neurons are active. Computation is linear on this subset ... Because of this linearity, gradients flow well on the active paths of neurons (there is no gradient vanishing effect due to activation non-linearities of sigmoid or tanh units), and mathematical investigation is easier. Computations are also cheaper: there is no need for computing the exponential function in activations, and sparsity can be exploited.''
\end{quote}

A problem with ReLU was that some neurons were never activated, and called ``dying'' or ``dead'', as described in \cite{Maas.2013}:
\begin{quote}
	``However, ReLU units are at a potential disadvantage during optimization because the gradient is 0 whenever the unit is not active. This could lead to cases where a unit never activates as a gradient-based optimization algorithm will not adjust the weights of a unit that never activates initially. Further, like the vanishing gradients problem, we might expect learning to be slow when training ReL networks with constant 0 gradients.''
\end{quote}
To remedy this ``dying'' or ``dead'' neuron problem,
%
% CMES style, rewriting 
the Leaky ReLU, proposed in
\cite{Maas.2013},\footnote{
%	According to Google Scholar, as of 2019.10.13, \cite{Glorot.2011:rd0001} (2011) received 3,656 citations, while \cite{Maas.2013} (2013) received 2,154 citations.
	According to Google Scholar, \cite{Glorot.2011:rd0001} (2011) received 3,656 citations on 2019.10.13 and 8,815 citations on 2022.06.23, whereas \cite{Maas.2013} (2013) received 2,154 and 6,380 citations on these two respective dates.
} had the expression already given previously in Eq.~(\ref{eq:leaky-relu}), and can be viewed as an approximation to the leaky diode in Figure~\ref{fig:I-V-halfwave}.  Both ReLU and Leaky ReLU have been known and used in neuroscience for years before being imported into artificial neural network; see Section~\ref{sc:history} for a historical review.

%{\color{red}
	% HERE, 2020.02.01, from HERE onward, Change notation for network predicted output from $\by$ (without hat, using macro \verb|\by|) to $\bout$ (with wide hat, using macro \verb|\bout|), and use $\expand{\by}$ (with overbar, using macro \verb|\expand{\by}|) for expanded output matrix.  For scalar output, use \verb|\out| to get $\out$.
%}

\begin{figure}[h]
	\centering
	%
	% 2022.12.17
	% add "-eps-converted-to.pdf" for arXiv
	% \includegraphics[width=0.7\linewidth]{figures/parametric-ReLU}
	\includegraphics[width=0.7\linewidth]{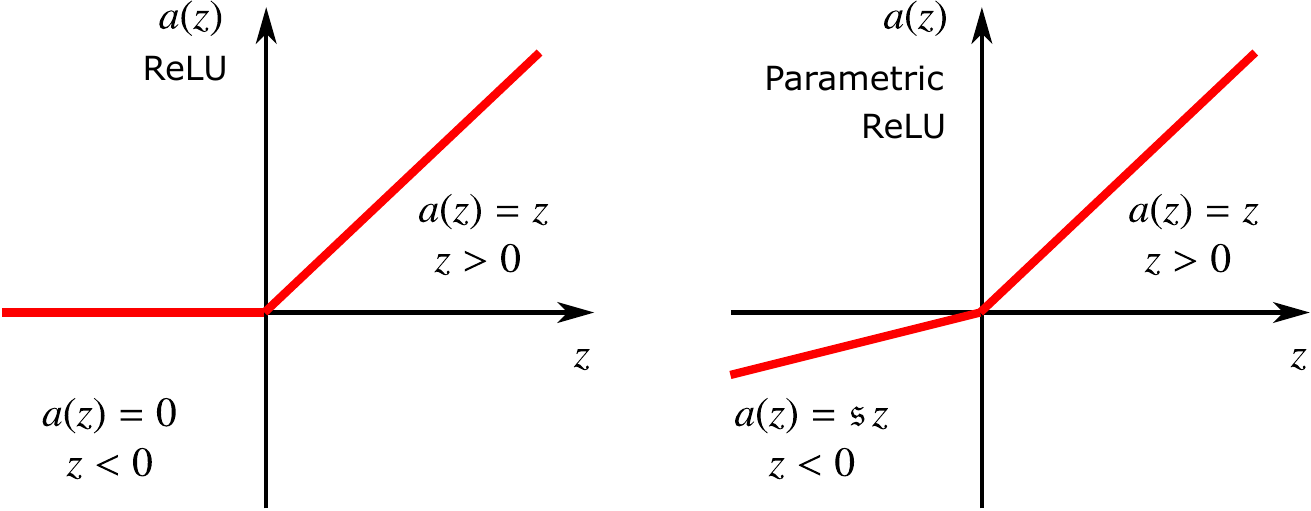}
	\caption{
		\emph{Rectified Linear Unit} (ReLU, left) and \emph{Parametric ReLU} (right)
		(Section~\ref{sc:relu}), in which the slope $\slope$ is a parameter to optimize; see Section~\ref{sc:parametric-ReLU}.  See also Figure~\ref{fig:ReLU} on ReLU.
		% \\
		%{\small (Figure reproduced with permission of the authors.)} 
		%{\footnotesize (Figure reproduced with permission of the authors.)}
	}
	\label{fig:parametric-relu}
\end{figure}

\subsubsection{Parametric rectified linear unit (PReLU)}
\label{sc:parametric-ReLU}

Instead of arbitrarily fixing the slope $\slope$ of the Leaky ReLU at $0.01$ for negative $z$ as in Eq.~(\ref{eq:leaky-relu}), 
%
% CMES style, rewriting
%\vphantom{\cite{He.2015b}}\cite{He.2015b} 
it is proposed to leave this slope $\slope$ as a free parameter to optimize along with the weights and biases \vphantom{\cite{He.2015b}}\cite{He.2015b}; see Figure~\ref{fig:parametric-relu}:
\begin{align}
	a(z) = 
	\max( \slope \, z , z)
	=
	\begin{cases}
	\slope \, z & \text{ for } z \le 0
	\\
	z         & \text{ for } 0 < z
	\end{cases}
	\label{eq:parametric-relu}
\end{align}
and thus the network adaptively learned the parameters to control the leaky part of the activation function.  
Using 
%
% CMES style, rewriting
%their 
the Parametric ReLU in Eq.\eqref{eq:parametric-relu}, 
%\cite{He.2015b} were 
a deep convolutional neural network (CNN) in \cite{He.2015b} was 
able to surpass the level of human performance in image recognition for the first time in 2015; see Figure~\ref{fig:ImageNet-error} on ImageNet competition results over the years.

%\input{06-empty}
% stochastic gradient descent

\begin{figure}[h]
	\centering
	\includegraphics[width=0.5\linewidth]{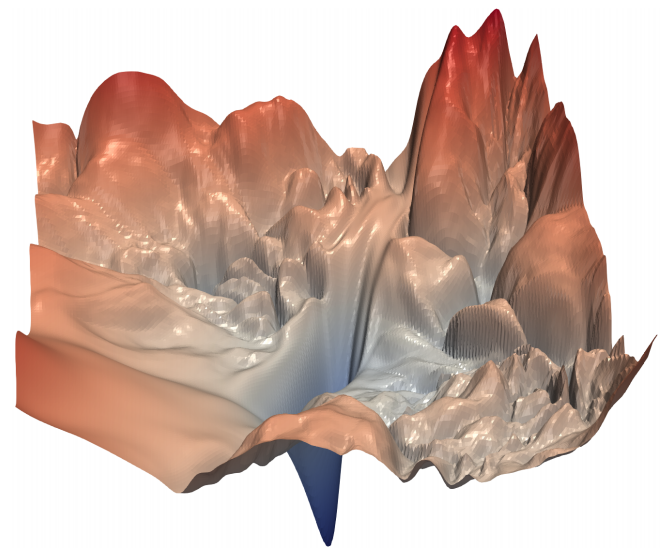}
	\caption{
		\emph{Cost-function landscape} (Section~\ref{sc:training}).  Residual network with 56 layers (ResNet-56) on the CIFAR-10 training set.  Highly non-convex, with many local minima, and deep, narrow valleys \cite{Li.2018}. 
		%
		% CMES style, rewriting 
%		From \cite{Li.2018}.  
		The training error and test error for fully-connected network increased when the number of layers was increased from 20 to 56, Figure~\ref{fig:deep-network-errors}, motivating the introduction of residual network, Figure~\ref{fig:resNet-basic} and Figure~\ref{fig:resNet-full}, Section~\ref{sc:architecture}.
		{\footnotesize (Figure reproduced with permission of the authors.)}
		%{\color{red} ASK PERMISSION 2019.11.25}
	}
	\label{fig:cost-function-landscape}
\end{figure}

% \section{Training, stochastic optimization methods}
\section{Network training, optimization methods}
\label{sc:training}

For network training, i.e., to find the optimal network parameters $\bparam$ that minimize the cost function $J(\bparam)$, we describe here both deterministic optimization methods used in full-batch mode,\footnote{
	A ``full batch'' is a complete training set of examples; see Footnote~\ref{fn:full-batch-minibatch}.
} and stochastic optimization methods used in minibatch\footnote{
	A minibatch is a random subset of the training set, which is called here the ``full batch''; see Footnote~\ref{fn:full-batch-minibatch}.
} mode.  

Figure~\ref{fig:cost-function-landscape} shows the highly non-convex landscape of the cost function of a residual network with 56 layers trained using the CIFAR-10 dataset (Canadian Institute For Advanced Research), a collection of images commonly used to train machine learning and computer vision algorithms, containing 60,000 32x32 color images in 10 different classes, representing airplanes, cars, birds, cats, deer, dogs, frogs, horses, ships, and trucks. Each class has 6,000 images.\footnote{
	See ``CIFAR-10'', Wikipedia,  \href{https://en.wikipedia.org/w/index.php?title=CIFAR-10&oldid=921224113}{version 16:44, 14 October 2019}.
}

Deterministic optimization methods (Section~\ref{sc:deterministic-optimization}) include first-order gradient method (Algorithm~\ref{algo:descent-armijo-deterministic}) and second-order quasi-Newton method (Algorithm~\ref{algo:gradient-quasi-newton-armijo-deterministic}), with line searches based on different rules, introduced by Goldstein, Armijo, and Wolfe.

Stochastic optimization methods (Section~\ref{sc:stochastic-gradient-descent}) include 
\begin{itemize}
	
	\item
	First-order \hyperref[sc:generic-SGD]{stochastic gradient descent} (\hyperref[sc:generic-SGD]{SGD}) methods (Algorithm~\ref{algo:generic-SGD}), with \hyperref[sc:add-on-tricks]{add-on tricks} such as momentum and accelerated gradient
	
	\item
	\hyperref[sc:adaptive-learning-rate-algos]{Adaptive learning-rate algorithms}
	 (Algorithm~\ref{algo:unified-adaptive-learning-rate-2}): \hyperref[para:adam1]{Adam} and variants such as \hyperref[para:amsgrad]{AMSGrad}, \hyperref[para:adamw]{AdamW}, etc. that are popular in the machine-learning community
	
	\item 
	\hyperref[para:adam-criticism]{Criticism of adaptive methods} and SGD resurgence with \hyperref[sc:add-on-tricks]{add-on tricks} such as effective tuning and step-length decay (or annealing) 
	
	\item 
	Classical line search with stochasticity:
	\hyperref[sc:SGD-armijo]{SGD with Armijo line search}
	 (Algorithm~\ref{algo:descent-armijo-stochastic-1}), second-order \hyperref[sc:stochastic-Newton]{Newton method with Armijo-like line search} (Algorigthm~\ref{algo:stochastic-newton})
		
\end{itemize}

\begin{figure}[h]
	\centering
	%
	% 2022.12.17
	% add "-eps-converted-to.pdf" for arXiv
	% \includegraphics[width=0.7\linewidth]{figures/validation-set}
	\includegraphics[width=0.7\linewidth]{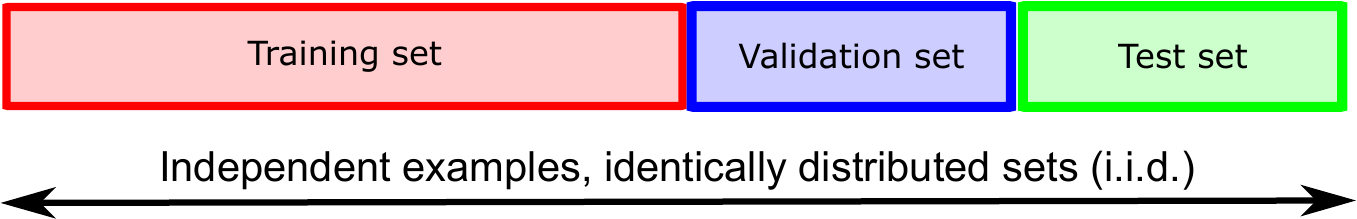}
	\caption{
		\emph{Training set, validation set, test set} (Section~\ref{sc:training-valication-test}). Partition of whole dataset.  The examples are independent.  The three subsets are identically distributed.
		%{\footnotesize (Figure reproduced with permission of the authors.)}
		%{\color{red} ASK PERMISSION 2019.11.25}
	}
	\label{fig:validation-set}
\end{figure}

\subsection{Training set, validation set, test set, stopping criteria}
\label{sc:training-valication-test}
%\begin{rem}
	%\label{rm:training-valication-test}
	%Training set, validation set, test set.
	%{\rm 		
	%}
%\end{rem}

\begin{figure}[h]
	\centering
	\includegraphics[width=0.8\linewidth]{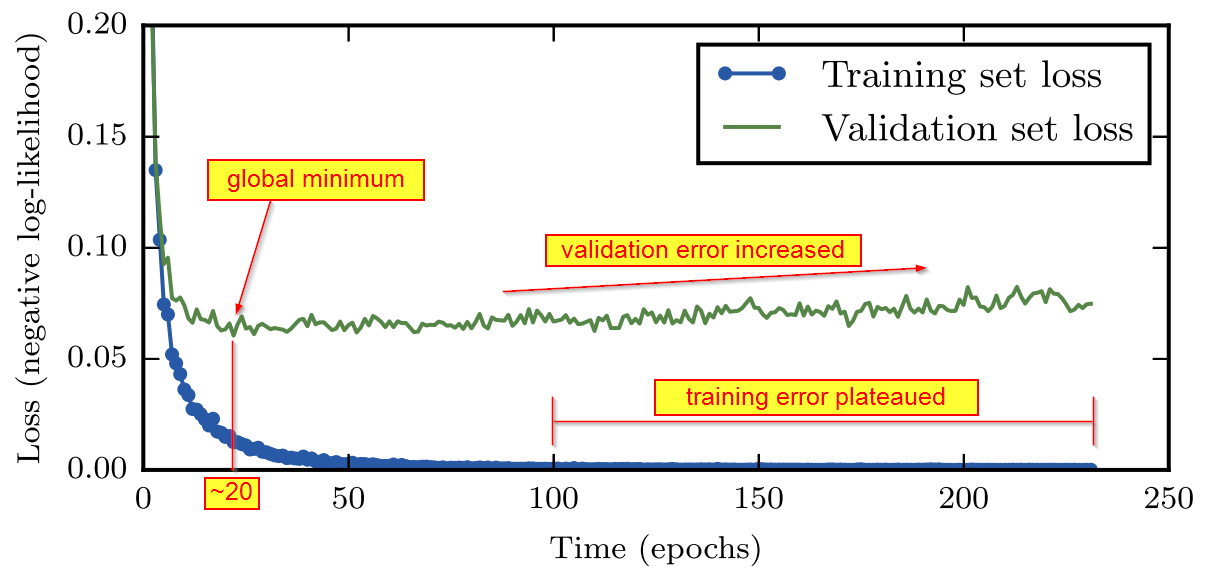}
	\caption{
		\emph{Training and validation learning curves---Classical viewpoint} (Section~\ref{sc:training-valication-test}), i.e., plots of training error and validation errors versus epoch number (time). While the training cost decreased continuously, the validation cost reaches a minimum around epoch 20, then started to gradually increase, forming an ``asymmetric U-shaped curve.''  Between epoch 100 and epoch 240, the training error was essentially flat, indicating convergence. Adapted from \cite{Goodfellow.2016}, p.~239.
		See Figure~\ref{fig:double-descent-risk} (a, left), where the classical risk curve is the classical viewpoint, whereas the modern interpolation viewpoint is on the right subfigure (b).
		{\footnotesize (Figure reproduced with permission of the authors.)}
		%{\color{red} ASK PERMISSION 2019.11.25}
	}
	\label{fig:validation-error}
\end{figure}

\begin{figure}[h]
	\centering
	%
	% 2022.12.17
	% add "-eps-converted-to.pdf" for arXiv
	% \includegraphics[width=0.5\linewidth]{figures/Prechelt-1998-validation-error-2020-01-08}
	\includegraphics[width=0.5\linewidth]{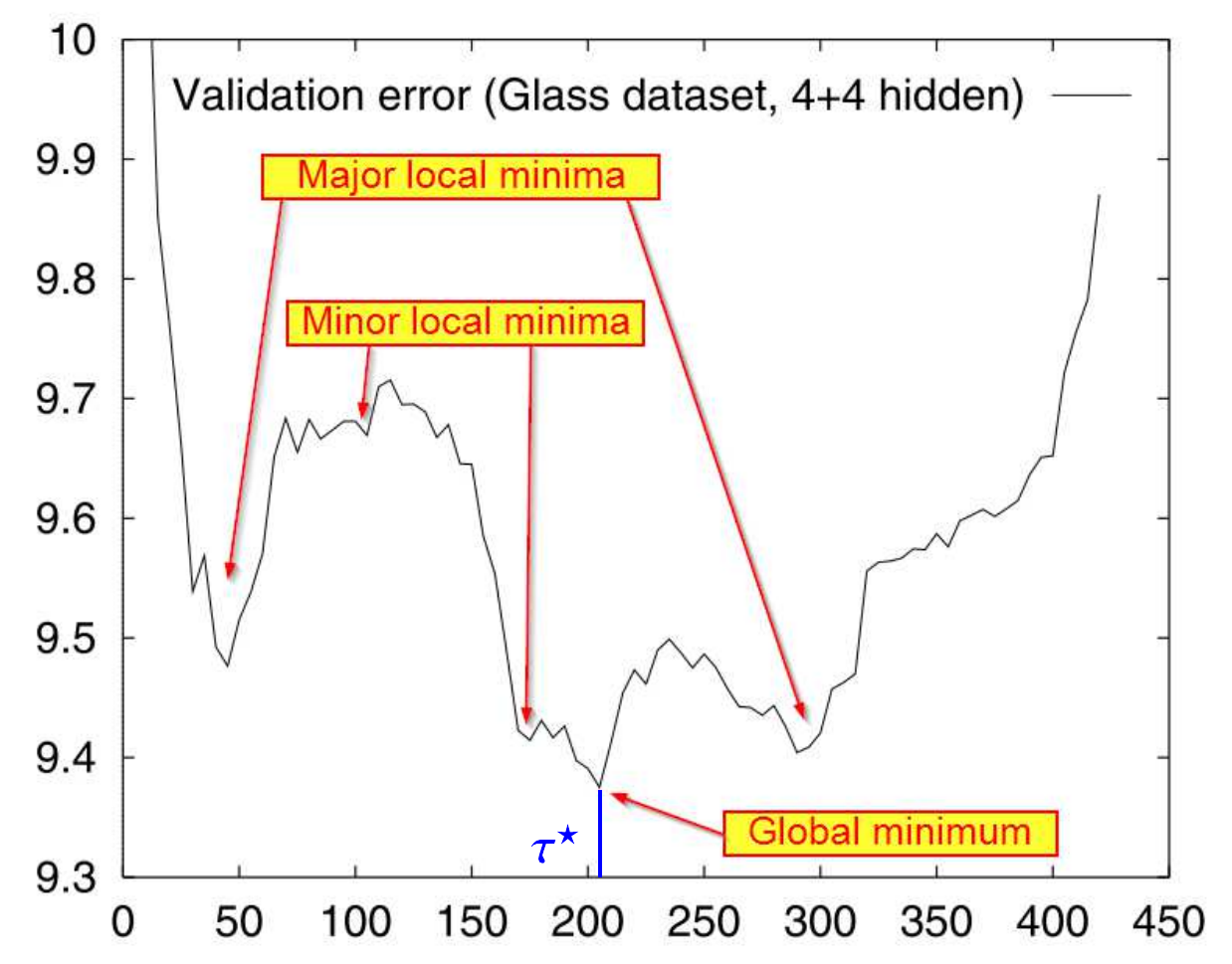}
	\caption{
		\emph{Validation learning curve} (Section~\ref{sc:training-valication-test}, Algorithm~\ref{algo:generic-SGD}). Validation error vs epoch number. Some validation error could oscillate wildly around the mean, resulting in an ``ugly reality''.  The global minimum validation error corresponded to epoch number $\tau^{\star}$.  Since the stopping criteria may miss this global minimum, it was suggested to monitor the validation learning curve to find the epoch $\tau^{\star}$ at which the network parameters $\bparamt_{\tau^\star}$ would be declared optimal.
		Adapted from \cite{Prechelt.1998}, p.~55.
		{\footnotesize (Figure reproduced with permission of the authors.)}
		%{\color{red} ASK PERMISSION 2019.11.25}
	}
	\label{fig:validation-error-ugly}
\end{figure}

\begin{figure}[h]
	\centering
	\includegraphics[width=0.8\linewidth]{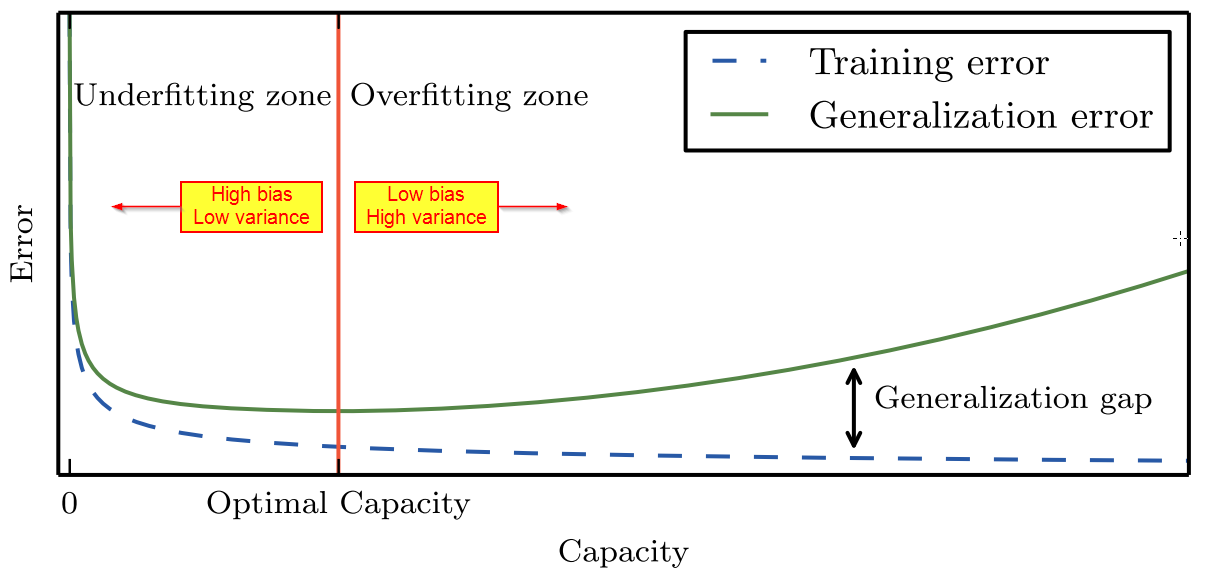}
	\caption{
		\emph{Bias-variance trade-off} (Section~\ref{sc:training-valication-test}).
%		\emph{Training error (cost) and test error versus model capacity}. 
		Training error (cost) and test error versus model capacity. 
		Two ways to change the model capacity: (1) change the number of network parameters, (2) change the values of these parameters (weight decay).  The generalization gap is the difference between the test (generalization) error and the training error.  As the model capacity increases from underfit to overfit, the training error decreases, but the generalization gap increases, past the optimal capacity.  Figure~\ref{fig:overfit} gives examples of underfit, appropriately fit, overfit.  See \cite{Goodfellow.2016}, p.~112.
		The above is the classical viewpoint, which is still prevalent \cite{Belkin_2019}; see Figure~\ref{fig:double-descent-risk} for the modern viewpoint, in which overfitting with high capacity model generalizes well (small test error) in practice.
		{\footnotesize (Figure reproduced with permission of the authors.)}
		%{\color{red} ASK PERMISSION 2019.11.25}
	}
	\label{fig:test-error}
\end{figure}

\begin{figure}[h]
	\centering
	\includegraphics[width=1.0\linewidth]{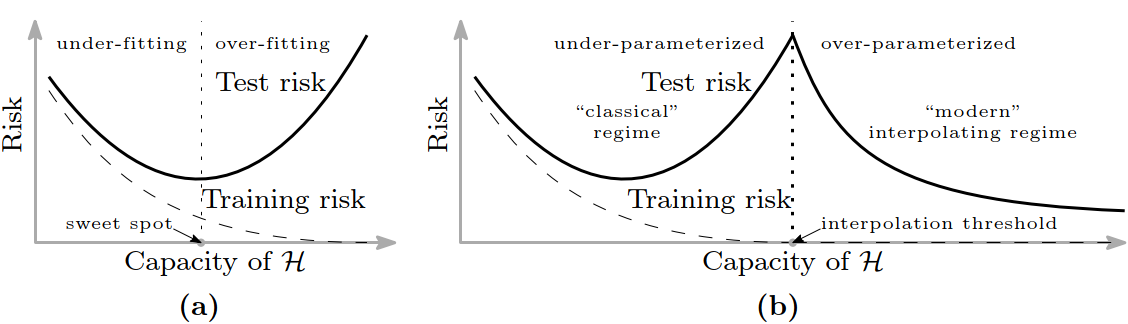}
	\caption{
		\emph{Modern interpolation regime} (Sections~\ref{sc:training-valication-test}, \ref{sc:lack-understanding}).  
		Beyond the interpolation threshold, the test error goes down as the model capacity (e.g., number of parameters) increases, describing the observation that networks with high capacity beyond the interpolation threshold generalize well, even though overfit in training.  Risk = error or cost. Capacity = number of parameters (but could also be increased by weight decay).
		Figures~\ref{fig:validation-error}, \ref{fig:test-error} corresponds to the classical regime, i.e., old method (thinking) \cite{Belkin_2019}.  
		See Figure~\ref{fig:test-error-large-params} for experimental evidence of the modern interpolation regime, and Figure~\ref{fig:network-infinite-width} for a shallow network with infinite width.
		\href{https://www.pnas.org/page/about/rights-permissions}{Permission of NAS}.
%		{\footnotesize (Figure reproduced with permission of the authors.)}
%		{\color{red} ASK PERMISSION 2019.11.25}
	}
	\label{fig:double-descent-risk}
\end{figure}

The classical (old) thinking---starting in 1992 with \cite{geman1992neural} and exemplified by  Figures~\ref{fig:validation-error}, \ref{fig:validation-error-ugly}, \ref{fig:test-error}, \ref{fig:double-descent-risk} (a, left)---would surprise first-time learners that \emph{minimizing the training error is not optimal} in machine learning.  A reason is that
training a neural network is \emph{different} from using ``pure optimization'' since 
%we not only want to decrease 
it is desired to decrease not only
the error during training (called training error, and that's pure optimization), but also the error committed by a trained network on inputs never seen before.\footnote{
	See also \cite{Goodfellow.2016}, p.~268, Section 8.1, ``How learning differs from pure optimization''; that's the classical thinking.
} Such error is called generalization error or test error. This classical thinking, known as the \emph{bias-variance trade-off}, has been included in books since 2001 \cite{hastie2001elements} (p.~194) and even repeated in 2016 \cite{Goodfellow.2016} (p.~268).  Models with lower number of parameters have higher bias and lower variance, whereas models with higher number of parameters have lower bias and higher variance; Figure~\ref{fig:test-error}.\footnote{
	See \cite{hastie2001elements}, p.~11, for a classification example using two methods: (1) linear models and least squares and (2) k-nearest neighbors. ``The linear model makes huge assumptions about structure [high bias] and yields stable [low variance] but possibly inaccurate predictions [high training error]. The method of k-nearest neighbors makes very mild structural assumptions [low bias]: its predictions are often accurate [low training error] but can be unstable [high variance].''
	See also \cite{bishop2006pattern}, p.~151, Figure~3.6.
} 

\begin{figure}[h]
	\centering
	\includegraphics[width=0.6\linewidth]{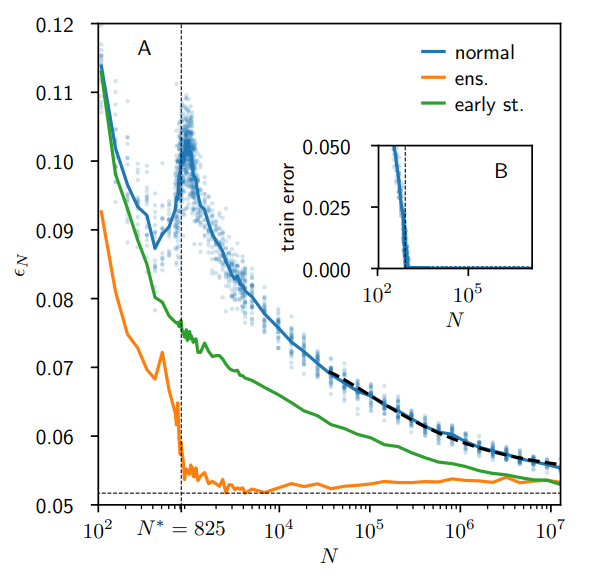}
	\caption{
		\emph{Empirical test error vs Number of paramesters} (Sections~\ref{sc:training-valication-test}, \ref{sc:lack-understanding}).  
		Experiments using the \href{http://yann.lecun.com/exdb/mnist/}{MNIST handwritten digit database} in \cite{Geiger_2020} confirmed the modern interpolation regime in Figure~\ref{fig:double-descent-risk} \cite{Belkin_2019}. 
%		The test error continued to decrease significantly with increasing number of parameters $N$ beyond the interpolation threshold $N^\star = 825$, whereas the classical regime ($N < N^\star$) with the bias-variance trade-off (blue) in Figure~\ref{fig:test-error} was restrictive, and did not generalize as well (larger test error).
		\emph{Blue:} Average over 20 runs.
		\emph{Green:} Early stopping.
		\emph{Orange:} Ensemble average on $n = 20$ samples, trained independently.
		See Figure~\ref{fig:network-infinite-width} for a shallow network with infinite width.
		{\footnotesize (Figure reproduced with permission of the authors.)}
%		{\color{red} ASK PERMISSION 2019.11.25}
	}
	\label{fig:test-error-large-params}
\end{figure}

The modern thinking is exemplified by Figure~\ref{fig:double-descent-risk} (b, right) and Figure~\ref{fig:test-error-large-params}, and does not contradict the intuitive notion that \emph{decreasing the training error to zero is indeed desirable}, as overparameterizing networks beyond the interpolation threshold (zero training error) in modern practice generalizes well (small test error).  In Figure~\ref{fig:test-error-large-params}, the test error continued to decrease significantly with increasing number of parameters $N$ beyond the interpolation threshold $N^\star = 825$, whereas the classical regime ($N < N^\star$) with the bias-variance trade-off (blue) in Figure~\ref{fig:test-error} was restrictive, and did not generalize as well (larger test error).
Beyond the interpolation threshold $N^\star$, variance can be decreased by using ensemble average, as shown by the orange line in Figure~\ref{fig:test-error-large-params}.

Such modern practice was the motivation for research into shallow networks with \emph{infinite} width as a first step to understand how overparameterized networks worked so well; see Figure~\ref{fig:network-infinite-width} and Section~\ref{sc:lack-understanding} ``Lack of understanding on why deep learning worked.''

To develop a neural-network model, a dataset governed by the same probability distribution, such as the CIFAR-10 dataset mentioned above, can be typically divided into three non-overlapping subsets called \emph{training set, validation set}, and \emph{test set}.  The validation set is also called the \emph{development set}, a terminology used in \vphantom{\cite{Wilson.2018}}\cite{Wilson.2018}, in which an effective method of step-length decay was proposed; see Section~\ref{sc:step-length-decay}.  

%
% CMES style, rewriting
%\cite{Prechelt.1998}, p.~61, suggested
It was suggested in \cite{Prechelt.1998}, p.~61, to use 50\% of the dataset as training set, 25\% as validation set, and 25\% as test set.  
%
% CMES style, rewriting
%\cite{Goodfellow.2016}, p.~118, suggested 
On the other hand, while a validation set with size about $1/4$ of the training set was suggested in \cite{Goodfellow.2016}, p.~118, there was no suggestion for the relative size of the test set.\footnote{
	Andrew Ng suggested the following partitions.  For small datasets having less than $10^{4}$ examples, the training/validation/test ratio of $60\% / 20\% / 20\%$ could be used.  For large datasets with order of $10^{6}$ examples, use ratio $98\% / 1\% / 1\%$.  For datasets with much more than $10^{6}$ examples, use ratio $99.5\% / 0.25\% / 0.25\%$.  See Coursera course ``Improving deep neural network: Hyperparameter tuning, regularization and optimization'', at time 4:00, \href{https://www.coursera.org/lecture/deep-neural-network/train-dev-test-sets-cxG1s}{video website}.
}  
See Figure~\ref{fig:validation-set} for a conceptual partition of the dataset.

Examples in the training set are fed into an optimizer to find the network parameter estimate $\bparamt$ that minimizes the cost function estimate $\losst (\bparamt)$.\footnote{
	The word ``estimate'' is used here for the more general case of stochastic optimization with minibatches; see Section~\ref{sc:generic-SGD} on stochastic gradient descent and subsequent sections on stochastic algorithms.  When deterministic optimization is used with the full batch of dataset, then the cost estimate is the same as the cost, i.e., $\losst \equiv \loss$, and the network parameter estimates are the same as the network parameters, i.e., $\bparamt \equiv \bparam$.
}  As the optimization on the training set progresses from epoch to epoch,\footnote{
	An epoch is when all examples in the dataset had been used in a training session of the optimization process.  For a formal definition of ``epoch'', see Section~\ref{sc:generic-SGD} on stochastic gradient descent (SGD) and Footnote~\ref{fn:epoch}.
} examples in the validation set are fed as inputs into the network to obtain the outputs for computing the cost function $\losst_{val} (\bparamt_\tau)$, also called validation error, at predetermined epochs $\{\tau_k\}$ using the network parameters $\{\bparamt_{\tau_k}\}$ obtained from the optimization on the training set at those epochs. 

Figure~\ref{fig:validation-error} shows the different behaviour of the training error versus that of the validation error.  The validation error would decrease quickly initially, reaching a global minimum, then gradually increased, whereas the training error continued to decrease and plateaued out, indicating that the gradients got smaller and smaller, and there was not much decrease in the cost.  From epoch 100 to epoch 240, the traning error was at about the same level, with litte noise.  The validation error, on the other hand, had a lot of noise. 

Because of the ``asymmetric U-shaped curve'' of the validation error, the thinking was that if the optimization process could stop early at the global mininum of the validation error, then the generalization (test) error, i.e., the value of cost function on the test set, would also be small, thus the name ``\emph{early stopping}''.  The test set contains examples that have not been used to train the network, thus simulating inputs never seen before.  The validation error could have oscillations with large amplitude around a mean curve, with many local minima; see Figure~\ref{fig:validation-error-ugly}.

The difference between the test (generalization) error and the validation error is called the generalization gap, as shown in the \emph{bias-variance trade-off} \cite{geman1992neural} 
Figure~\ref{fig:test-error}, which qualitatively delineates these errors versus model capacity, and conceptually explains the optimal model capacity as where the generalization gap equals the training error, or the generalization error is twice the training error.

%{\color{red} [NOTE: 2022.09.08 - I am HERE in updating Section~\ref{sc:training-valication-test} to include the modern mode of thinking in terms of double descent risk curve in Figure~\ref{fig:double-descent-risk}, which is more general than the classical (old) risk curve of Figure~\ref{fig:validation-error}.] ENDNOTE}

\begin{rem}
	\label{rm:generalization-capability}
	{\rm 
		Even the best machine learning generalization capability nowadays still cannot compete with the generalization ability of human babies; see Section~\ref{sc:whats-new} on ``What's new? Teaching machines to think like babies''.
	}
	\hfill$\blacksquare$
\end{rem}

%{\color{red} [NOTE: 2022.09.09 - define bias and variance here.]}

{\bf Early-stopping criteria.}  One criterion is to first define the lowest validation error from epoch 1 up to the current epoch $\tau$ as: 
\begin{align}
	\losst_{val}^\star (\bparamt_{\tau})
	=
	\min_{\tau^\prime \le \tau}
	\{ \losst_{val} (\bparamt_{\tau^\prime}) \}
	\ ,
	\label{eq:validation-error-lowest}
\end{align}
then define the \emph{generalization loss} (in percentage) at epoch $\tau$ as the increase in validation error relative to the minimum validation error from epoch 1 to the present epoch $\tau$:
\begin{align}
	G (\tau) = 100 
	\cdot 
		\left( \frac{\losst_{val} (\bparamt_{\tau})}{\losst_{val}^\star (\bparamt_{\tau})} - 1 
	\right)
	\ .
	\label{eq:generalization-error-index}
\end{align}
\cite{Prechelt.1998} then defined the ``first class of stopping criteria'' as follows: Stop the optimization on the training set when the generalization loss exceeds a certain threshold $\glub$ (generalization loss lower bound):
\begin{align}
	G_\glub : \text{Stop after epoch } \tau \text{ if } G(\tau) > \glub
	\ .
	\label{eq:stopping-criterion-1}
\end{align}
The issue is how to determine the generalization loss lower bound $\glub$ so not to fall into a local minimum, and to catch the global minimum; see Figure~\ref{fig:validation-error-ugly}.
There were many more early-stopping criterion classes in \cite{Prechelt.1998}.  But it is not clear whether all these increasingly sophisticated stopping criteria would work to catch the validation-error global minimum in Figure~\ref{fig:validation-error-ugly}.  

Moreover, the above discussion is for the \emph{classical regime} in Figure~\ref{fig:double-descent-risk} (a).
In the context of the \emph{modern interpolation regime} in Figure~\ref{fig:double-descent-risk} (b), early stopping means that the computation would cease as soon as the training error reaches ``its lowest possible value (typically zero [beyond the interpolation threshold], unless two identical data points have two different labels)'' \cite{Geiger_2020}.  See the green line in Figure~\ref{fig:test-error-large-params}.

%{\color{red} [NOTE: 2022.09.09 - add remark on early stopping in the modern interpolation regime.]}

{\bf Computational budget, learning curves.} A simple method would be to set an epoch budget, i.e., the largest number of epochs for computation sufficiently large for the training error to go down significantly, then monitor graphically both the training error (cost) and the validation error versus epoch number.  These plots are called the \emph{learning curves}; see Figure~\ref{fig:validation-error}, for which an epoch budget of 240 was used.  Select the global minimum of the validation learning curve, with epoch number $\tau^\star$ (Figure~\ref{fig:validation-error}), and use the corresponding network parameters $\bparamt_{\tau^\star}$, which were saved periodically, as optimal paramters for the network.\footnote{
	See also ``Method for early stopping in a neural network'', StackExchange, 2018.03.05, 
	\href{https://stats.stackexchange.com/questions/331821/method-for-early-stopping-in-a-neural-network}{Original website},
	\href{https://web.archive.org/web/20200109015935/https://stats.stackexchange.com/questions/331821/method-for-early-stopping-in-a-neural-network}{Internet archive}.
	\cite{Goodfellow.2016}, p.~287, also suggested to monitor the training learning curve to adjust the step length (learning rate). 
}

\begin{rem}
	\label{sc:pure-optimization-NO}
	{\rm
		Since it is important to monitor the validation error during training,
		%
		% CMES style, rewriting 
%		\cite{Goodfellow.2016} devoted 
		a whole section is devoted in \cite{Goodfellow.2016}
		(Section 8.1, p.~268) to expound on ``How Learning Differs from Pure Optimization''. And also for this reason, it is not clear yet what \emph{global optimization} algorithms such as in \cite{Sampaio.2020} could bring to network training, whereas the stochastic gradient descent (SGD) in Section~\ref{sc:generic-SGD} is quite efficient; see also Section~\ref{sc:adam-criticism} on criticism of adaptive methods.
	}
	$\hfill\blacksquare$
\end{rem}

\begin{rem}
	\label{rm:budget}
	Epoch budget, global iteration budget.
	{\rm 
		For stochastic optimization algorithms---Sections~\ref{sc:stochastic-gradient-descent}, \ref{sc:adaptive-learning-rate-algos}, \ref{sc:SGD-armijo}, \ref{sc:stochastic-Newton}---the epoch counter is $\tau$ and the epoch budget $\tau_{max}$.  Numerical experiments in Figure~\ref{fig:wilson-examples} had an epoch budget of $\tau_{max} = 250$, whereas numerical experiments in Figure~\ref{fig:adamw-examples} had an epoch budget of $\tau_{max} = 1800$.
		The computational budget could be specified in terms of global iteration counter $j$ as $j_{max}$.  Figure~\ref{fig:amsgrad-examples} had a global iteration budget of $j_{max} = 5000$.
	}
	$\hfill\blacksquare$ 
\end{rem}

Before presenting the stochastic gradient-descent (SGD) methods in Section~\ref{sc:stochastic-gradient-descent}, it is important to note that classical deterministic methods of optimization in Section~\ref{sc:deterministic-optimization} continue to be useful in the age of deep learning and SGD.

\begin{quote}
	``One should not lose sight of the fact that [full] batch approaches possess some intrinsic advantages. First, the use full gradient information at each iterate opens the door for many deterministic gradient-based optimization methods that have been developed over the past decades, including not only the full gradient method, but also accelerated gradient, conjugate gradient, quasi-Newton, inexact Newton methods, and can benefit from parallelization.''
	\cite{Bottou.2018:rd0001}, p.~237.
\end{quote}

%{\color{red} HERE 2020.01.08}

%\subsection{Parameter update, learning rate, line search}
\subsection{Deterministic optimization, full batch}
\label{sc:learning-rate}
\label{sc:deterministic-optimization}

Once the gradient $\partial J / \partial \bparamp{\ell} \in
\real^{\widths{\ell} \times [\widths{\ell-1}  + 1]}$ of the cost function $J$ has been computed using backpropagation described in Section~\ref{sc:gradient}, the layer parameters $\bparamp{\ell}$ are updated to decrease cost function $J$ using gradient descent as follows:
\begin{align}
	\bparamp{\ell} \leftarrow \bparamp{\ell} - \epsilon \, \partial J / \partial \bparamp{\ell}
	=
	\bparamp{\ell} - \epsilon \, \bgradp{\ell}
	\ ,
	\text{ with }
	\bgradp{\ell} := \partial J / \partial \bparamp{\ell}
	\in
	\real^{\widths{\ell} \times [\widths{\ell-1}  + 1]}
	\ .
	\label{eq:update-params-layer}
\end{align} 
being the gradient direction, and $\epsilon$ called the learning rate.\footnote{
	See Figure~\ref{fig:Haibe-Kains-irreproducibility} in Section~\ref{sc:irreproducibility} on ``Lack of transparency and irreproducibility of results'' in recent deep-learning papers.
}  
The layer-by-layer update in Eq.~(\ref{eq:update-params-layer}) as soon as the gradient $\bgradp{\ell} = \partial J / \partial \bparamp{\ell}$ had been computed is valid when the learning rate $\learn$ does not depend on the gradient $\bgrad = \partial J / \partial \bparam$ with respect to the whole set of network parameters $\bparam$.  

Otherwise, the update of the whole network parameter $\bparam$ would be carried out after the complete gradient $\bgrad  = \partial J / \partial \bparam$ had been obtained, and the learning rate $\learn$ had been computed based on the gradient $\bgrad$: 
\begin{align}
	\bparam \leftarrow \bparam - \epsilon \, \partial J / \partial \bparam
	=
	\bparam - \epsilon \, \bgrad
	\ ,
	\text{ with }
	\bgrad := \partial J / \partial \bparam
	\in
	\real^{\Tparam \times 1}
	\ ,
	\label{eq:update-params}
\end{align}
where $\Tparam$ is the total number of network paramters defined in Eq.~(\ref{eq:totalParams}).
An example of a learning-rate computation that depends on the complete gradient $\bgrad  = \partial J / \partial \bparam$ is gradient descent with Armijo line search; see Section~\ref{sc:armijo} and line~\ref{lst:line:update-params-backprop} in Algorithm~\ref{algo:backprop}.

\begin{quote}
	``Neural network researchers have long realized that the learning rate is reliably one of the most difficult to set hyperparameters because it significantly affects model performance.'' \cite{Goodfellow.2016}, p.~298.
\end{quote}
In fact, it is well known in the field of optimization, where the learning rate is often mnemonically denoted by $\lambda$, being Greek for ``l'' and standing for ``step length''; see, e.g., Polak (1971) \cite{Polak.1971}.

\begin{quote}
	``We can choose $\epsilon$ in several different ways. A popular approach is to set $\epsilon$ to a small constant. Sometimes, we can solve for the step size that makes the directional derivative vanish. Another approach is to evaluate $f (\bx - \epsilon \nabla_{\bx} f(\bx))$ for several values of $\epsilon$ and choose the one that results in the smallest objective function value. This last strategy is called a {\bf line search}.''  \cite{Goodfellow.2016}, p.~82.
\end{quote}
Choosing an arbitrarily small $\epsilon$, without guidance on how small is small, is not a good approach, since exceedingly slow convergence could result for too small $\epsilon$.  In general, it would not be possible to solve for the step size to make the directional derivative vanish.  There are several variants of line search for computing the step size to decrease the cost function, based on the ``decrease'' conditions,\footnote{
	See, e.g., \cite{Lewis.2000}, \cite{Kolda.2003}. 
} among which some are mentioned below.\footnote{
	See also \cite{Polak.1971}, p.~243.
} 

\begin{rem}
	\label{rm:classic-never-dies}
	Line search in deep-learning training.
	{\rm 
		Line search methods are not only important for use in deterministic optimization with full batch of examples,\footnote{
			\label{fn:full-batch-minibatch}
			A full batch contains all examples in the training set.  There is a confusion in the use of the word ``batch'' in terminologies such as ``batch optimization'' or ``batch gradient descent'', which are used to mean the full training set, and not a subset of the training set; see, e.g., \cite{Goodfellow.2016}, p.~271.  Hence we explicitly use ``full batch'' for full training set, and mini-batch for a small subset of the training set. 
		} but also in stochastic optimization (see Section~\ref{sc:stochastic-gradient-descent}) with random mini-batches of examples \cite{Bottou.2018:rd0001}.  The difficulty of using stochastic gradient coming from random mini-batches is the presence of noise or ``discontinuities''\footnote{
			See, e.g., \cite{Bottou.2018:rd0001}.  Noise is sometimes referred to as ``discontinuities'' as in \cite{Kafka.2019}.
			See also the lecture video ``Understanding mini-batch gradient descent,'' at time 1:20, by Andrew Ng on Coursera \href{https://www.coursera.org/lecture/deep-neural-network/understanding-mini-batch-gradient-descent-lBXu8}{website}.
		} in the cost function and in the gradient.  Recent stochastic optimization methods---such as the sub-sampled Hessian-free Newton method reviewed in \cite{Bottou.2018:rd0001}, the probabilisitic line search in \cite{Mahsereci.2017}, the first-order stochastic Armijo line search in \cite{Paquette.2018}, the second-order sub-sampling line search method in \vphantom{\cite{Bergou.2018}}\cite{Bergou.2018}, quasi-Newton method with probabilitic line search in \cite{Wills.2018}, etc.---where line search forms a key subprocedure, are designed to address or circumvent the noisy gradient problem.  For this reason, claims that line search methods have ``fallen out of favor''\footnote{
			\label{fn:classic-skepticism}
			%
			% CMES style, rewriting
%			\cite{Goodfellow.2016} completely bypassed a 
			In \cite{Goodfellow.2016}, 
			a discussion on line search methods, however brief, was completely bypassed to focus on stochastic gradient-descent methods with learning-rate tuning and scheduling, such as AdaGrad, Adam, etc.  Ironically, it is disconcerting to see these authors, who made important contributions to deep learning, thus helping thawing the last ``AI winter'', regard with skepticism ``most guidance'' on learning-rate selection; see \cite{Goodfellow.2016}, p.~287, and Section~\ref{sc:stochastic-gradient-descent}.  Since then, fully automatic stochastic line-search methods, without tuning runs, have been developed, apparently starting with \cite{Mahsereci.2015}.
			%
			% CMES style, rewriting
%			\cite{Kafka.2019}, who presented 
			In the abstract of \cite{Kafka.2019}, where
			an interesting method using only gradients, without function evaluations, was presented, one reads
%			wrote in their abstract: 
			``Due to discontinuities induced by mini-batch sampling, [line searches] have largely fallen out of favor''.
		} would be misleading, as they may encourage students not to learn the classics.  A classic never dies; it just re-emerges in a different form with additional developments to tackle new problems.  
	}
	$\hfill\blacksquare$
\end{rem}

In view of Remark~\ref{rm:classic-never-dies}, a goal of this section is to develop a feel for some classical deterministic line search methods for readers not familiar with these concepts to prepare for reading extensions of these methods to stochastic line search methods.

%\vspace{5mm}
%\noindent
%{\bf Variant 1:} \emph{Exact line search}.
\subsubsection{Exact line search}
\label{sc:line-search-exact}

Find a positive step length $\epsilon$ that minimizes the cost function $J$ along the descent direction $\bd$ such that the scalar (dot) product between $\bgrad$ and $\bd  \in
\real^{\Tparam \times 1}$:
\begin{align}
	< \bgrad , \bd > 
	= 
	\bgrad \dotprod \, \bd
	=
	\sum_i \sum_j \grad_{ij} d_{ij} < 0
	\ , 
	\label{eq:descent-dir}
\end{align}
is negative, i.e., the descent direction $\bd$ and the gradient $\bgrad$ form an obtuse angle bounded away from $90^\circ$,\footnote{
	Or equivalently, the descent direction $\bd$ forms an acute angle with the gradient (or steepest) descent direction $[- \bgrad = - \partial J / \partial \bparam]$, i.e., the negative of the gradient direction.
} and
\begin{align}
	J( \bparam + \epsilon \, \bd)
	=
	\min_\lambda \{ J(\bparam + \lambda \, \bd ) \, | \, \lambda \ge 0 \} 
	\ .
	\label{eq:linesearch-1}
\end{align}
The minimization problem in Eq.~(\ref{eq:linesearch-1}) can be implemented using the Golden section search (or infinite Fibonacci search) for unimodal functions.\footnote{
	See, e.g., \cite{Polak.1971}, p.~31, for the implementable algorithm, with the assumption that the cost function was convex.  Convexity is, however, not needed for this algorithm to work; only unimodality is needed.  A unimodal function has a unique minimum, and is decreasing on the left of the minimum, and increasing on the right of the minimum. Convex functions are necessarily unimodal, but not vice versa.  Convexity is a particular case of unimodality.
	See also \cite{Luenberger.2016}, p.~216, on Golden section search as infinite Fibonacci search and curve fitting line-search methods.
}
For more general non-convex cost functions, a minimizing step length may be non-existent, or difficult to compute exactly.\footnote{
	See \cite{Polak.1997}, p.~29, for this assertion, without examples.  An example of non-existent minimizing step length $\epsilon = \arg\min_\lambda \{ f(\theta + \lambda d) | \lambda \ge 0 \}$ could be $f$ being a concave function. If we relax the continuity requirement, then it is easy to construct a function with no mininum and no maximum, e.g., $f(x) = |x|$ on open intervals $(-1, 0) \cup (0, +1)$, and $f(-1) = f(0) = f(1) = 0.5$; this function is ``essentially'' convex, except on the set $\{-1, 0, +1\}$ of measure zero where it is discontinuous. An example of a function whose minimum is difficult to compute exactly could be one with a minimum at the bottom of an extremely narrow crack.
}
In addition, a line search for a minimizing step length is only an auxilliary step in an overall optimization algorithm.  It is therefore sufficient to find an approximate step length satisfying some decrease conditions to ensure convergence to a local minimum, while keeping the step length from being too small that would hinder a reasonable advance toward such local minimum.  For these reasons, inexact line search methods (rules) were introduced, first 
%
% CMES style, rewriting
%by \cite{Goldstein.1965}, 
in \cite{Goldstein.1965},
followed by \cite{Armijo.1966}, then \cite{Wolfe.1969} and \cite{Wolfe.1971}.  In view of Remark~\ref{rm:classic-never-dies} and Footnote~\ref{fn:classic-skepticism}, as we present these deterministic line-search rules, we will also immediately recall, where applicable, the recent references that generalize these rules by adding stochasticity for use as a subprocedure (inner loop) for the stochastic gradient-descent (SGD) algorithm.

%{\color{red} HERE 2020.01.11}

%\vspace{5mm}
%\noindent
%{\bf Variant 2:} \emph{Inexact line-search, Goldstein's rule.}
\subsubsection{Inexact line-search, Goldstein's rule}
\label{sc:inexact-line-search-goldstein}
The method is inexact since the search for an acceptable step length would stop before a minimum is reached, once the rule is satisfied.\footnote{
	\label{fn:polak-goldstein}
	%
	% CMES style, rewritiing
%	\cite{Polak.1971}, p.~33, cited 
	The book \cite{Goldstein.1967} was cited in \cite{Polak.1971}, p.~33, but not the papers \cite{Goldstein.1965} and \cite{Goldstein.1967b}, where Goldstein's rule was explicitly presented in the form: Step length $\gamma \rightarrow g(x^k, \gamma) = [ f(x^k) - f(x^k - \gamma \varphi(x^k) )] / \{ \gamma [\nabla f (x^k) , \varphi(x^k) ]\}$, with $\varphi$ being the descent direction, $[a,b]$ the scalar (dot) product between vector $a$ and vector $b$, and $\delta \le g(x^k , \gamma) \le 1-\delta$, if $g(x^k , 1) \le \delta$.  The same relation was given in \cite{Goldstein.1965}, with different notation.
} 
For a fixed constant $\alpha \in (0,\frac12)$, select a learning rate (step length) $\epsilon$ such that\footnote{
	See also \cite{Polak.1971}, p.~33.
}
\begin{align}
	1 - \alpha 
	\le 
	\frac{ J(\bparam  + \epsilon \, \bd ) - J(\bparam) }{\epsilon \, \bgrad \dotprod \, \bd }
	\le
	\alpha
	\ ,
	\label{eq:goldstein-1}
\end{align} 
where both the numerator and the denominator are negative, i.e., $J(\bparam  + \epsilon \, \bd ) - J(\bparam) < 0$ and $\bgrad \dotprod \, \bd < 0$ by Eq.~(\ref{eq:descent-dir}).  Eq.~(\ref{eq:goldstein-1}) can be recast into a slightly more general form:
For $0 < \alpha < \beta < 1$, choose a learning rate $\epsilon$ such that\footnote{
	\label{fn:goldstein-principle}
	See \cite{Polak.1997}, p.~55, and \cite{Ortega.1970}, p.~256, where the equality $\alpha = \beta$ was even allowed, provided the step length $\epsilon$ satisfied the equality $f(\bparam  + \epsilon \, \bd ) - f(\bparam) = \alpha 
	\,
	\epsilon \, \bgrad \dotprod \, \bd$.  But that would make the computation unnecessarily stringent and costly, since the step length $\epsilon$ as the root of this equation has to be solved for accurately.  Again, the idea should be to make the sector bounded by the two lines, $\beta
	\,
	\epsilon \, \bgrad \dotprod \, \bd$ from below and the line $\alpha 
	\,
	\epsilon \, \bgrad \dotprod \, \bd$ from above, as large as possible in inexact line search.  See discussion below Eq.~(\ref{eq:goldstein-2}). 
}

\begin{align}
	\beta
	\,
	\epsilon \, \bgrad \dotprod \, \bd
	\le
	\Delta J (\epsilon)
	\le 
	\alpha 
	\,
	\epsilon \, \bgrad \dotprod \, \bd
	\ , 
	\text{ with }
	\Delta J  (\epsilon)
	:=
	J(\bparam  + \epsilon \, \bd ) - J (\bparam)
	\ .
	\label{eq:goldstein-2}
\end{align}
A reason could be that the sector bounded by the two lines $(1-\alpha) \epsilon \, \bgrad \dotprod \, \bd$ and $\alpha \epsilon \, \bgrad \dotprod \, \bd$ may be too narrow when $\alpha$ is close to 0.5 from below, making $(1-\alpha)$ also close to 0.5 from above. For example, 
%
% CMES style, rewriting
%\cite{Polak.1971}, p.~33 and p.~37, recommended using 
it was recommended in \cite{Polak.1971}, p.~33 and p.~37, to use
$\alpha=0.4$, and hence $1 - \alpha = 0.6$, making a tight sector, but we could enlarge such sector by choosing $0.6 < \beta < 1$.

\begin{figure}[h]
	\centering
	%
	% 2022.12.17
	% add "-eps-converted-to.pdf" for arXiv
	% \includegraphics[width=0.8\linewidth]{figures/line-search-goldstein-c}
	\includegraphics[width=0.8\linewidth]{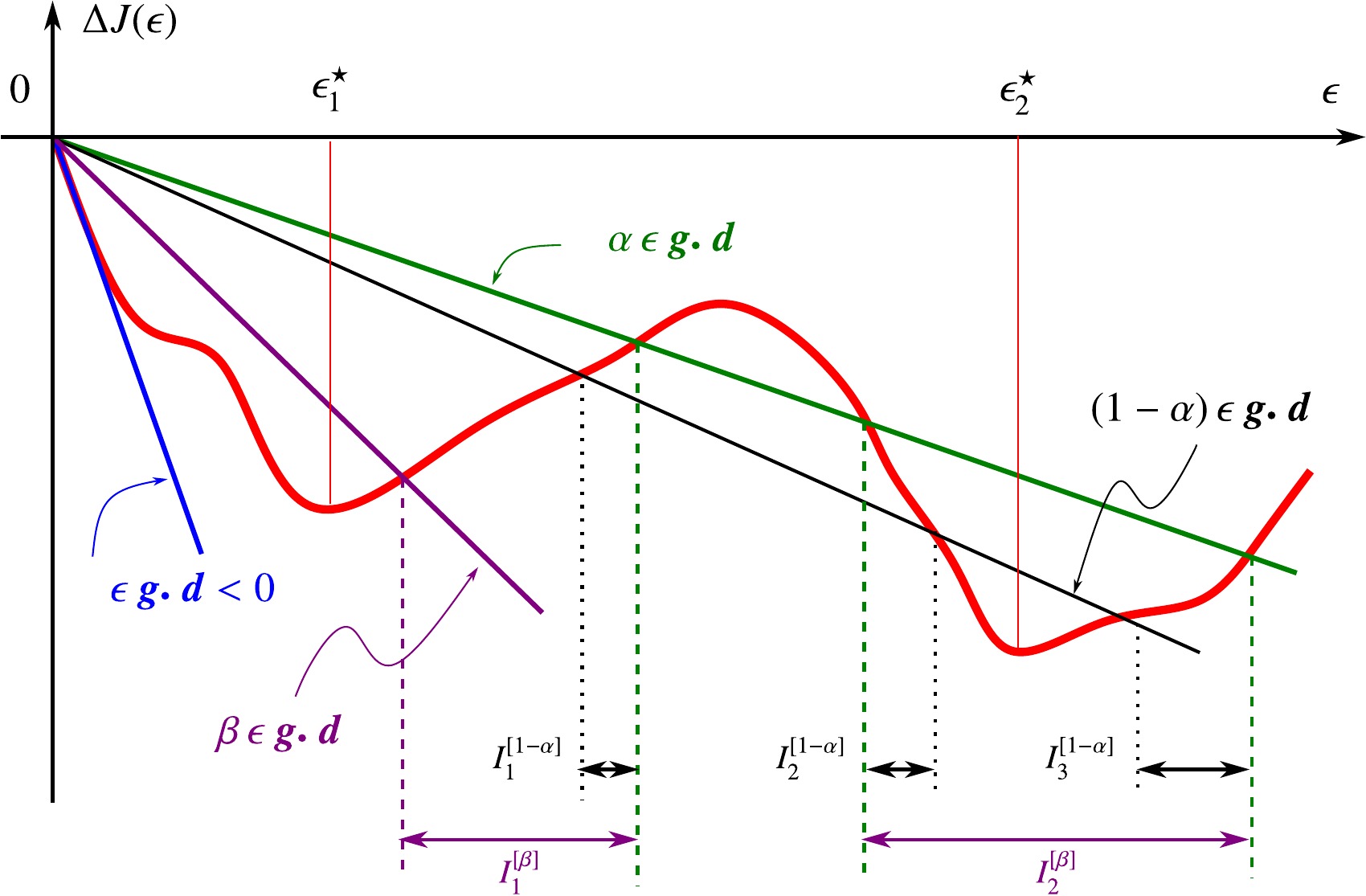}
	\caption{
		\emph{Inexact line search, Goldstein's rule} (Section~\ref{sc:inexact-line-search-wolfe}).  acceptable step lengths would be such that a decrease in the cost function $J$, denoted by $\Delta J$ in Eq.~(\ref{eq:goldstein-2}), falls into an acceptable sector formed by an upper-bound line and a lower-bound line.  the upper bound is given by the straight line $\alpha \,\epsilon \, \bgrad \dotprod \, \bd$ (green), with fixed constant $\alpha \in (0, \frac12)$ and $\epsilon \, \bgrad \dotprod \, \bd < 0$ being the slope to the curve $\Delta J (\epsilon)$ at $\epsilon = 0$. 
		The lower-bound line $(1-\alpha) \epsilon \, \bgrad \dotprod \, \bd$ (black), adopted in \cite{Goldstein.1965} and \cite{Goldstein.1967b}, would be too narrow when $\alpha$ is close to $\frac12$, leaving all local minimizers such as $\epsilon^\star_1$ and $\epsilon^\star_2$ outside of the  acceptable intevals $I^{[1-\alpha]}_1$, $I^{[1-\alpha]}_2$, and $I^{[1-\alpha]}_3$ (black), which are themselves narrow.
		The lower-bound line $\beta \epsilon \, \bgrad \dotprod \, \bd$ (purple) proposed in \cite{Ortega.1970}, p.~256, and \cite{Polak.1997}, p.~55, with $(1-\alpha) < \beta < 1$, would enlarge the acceptable sector, which then may contain the minimizers inside the corresponding acceptable intervals $I^{[\beta]}_1$ and $I^{[\beta]}_2$ (purple).
	}
	\label{fig:goldstein-1}
\end{figure}

The search for an appropriate step length that satisfies Eq.~(\ref{eq:goldstein-1}) or Eq.~(\ref{eq:goldstein-2}) could be carried out by a subprocedure based on, e.g., the bisection method, as suggested in \cite{Polak.1971}, p.~33.   Goldstein'rule---also designated as \emph{Goldstein principle} in the classic book 
%
% CMES style, rewriting
%the classic book by \cite{Ortega.1970}, p.~256,
\cite{Ortega.1970}, p.~256, since it ensured a decrease in the cost function---has been ``used only occasionally'' per Polak (1997) \cite{Polak.1997}, p.~55, largely superceded by Armijo's rule, and has not been generalized to add stochasticity.  On the other hand, the idea behind Armijo's rule is similar to Goldstein's rule, but with a convenient subprocedure\footnote{
	See \cite{Polak.1971}, p.~36, Algorithm 36. 
} to find the appropriate step length.

%{\color{red} HERE 2020.01.11}

%\vspace{5mm}
%\noindent
%{\bf Variant 3:} \emph{Inexact line-search, Armijo's rule.}
\subsubsection{Inexact line-search, Armijo's rule}
\label{sc:inexact-line-search-armijo}
\label{sc:armijo}

Apparently without the knowledge of \cite{Goldstein.1965}, 
%
% CMES style, rewriting
%\cite{Armijo.1966} proposed 
it was proposed in \cite{Armijo.1966}
the following highly popular Armijo step-length search,\footnote{
%	As of 2019.11.01, \cite{Armijo.1966} was cited 1827 times
	As of 2022.07.09, \cite{Armijo.1966} was cited 2301 times
	in various publications (books, papers) according to Google Scholar, and 
%	792 times 
	1028 times
	in archival journal papers according to Web of Science.  There are references that mention the name Armijo, but without referring to the original paper \cite{Armijo.1966}, such as \cite{Nocedal.2006}, clearly indicating that Armijo's rule is a classic, just like there is no need to refer to Newton's original work for Newton's method.
} which recently forms the basis for stochastic line search for use in stochastic gradient-descent algorithm described in Section~\ref{sc:stochastic-gradient-descent}: 
%
% CMES style, rewriting 
%\cite{Paquette.2018} added 
Stochasticity was added to Armijo's rule in \cite{Paquette.2018},
%
% CMES style, rewriting  
%\cite{Bergou.2018} extended 
and the concept was extended to second-order line search \cite{Bergou.2018}.  Line search based on Armijo's rule is also applied to quasi-Newton method for noisy functions in  \vphantom{\cite{Bollapragada.2019}}\cite{Bollapragada.2019}, and to exact and inexact subsampled Newton methods in \vphantom{\cite{Berakas.2019}}\cite{Berakas.2019}.\footnote{
	All of these stochastic optimization methods are considered as part of a broader class known as derivative-free optimization methods \vphantom{\cite{Larson.2019}}\cite{Larson.2019}.  
} 

Armijo's rule is stated as follows: For $\alpha \in (0,1)$, $\beta \in (0,1)$, and $\rho > 0$, use the step length $\epsilon$ such that:\footnote{
	\cite{Ortega.1970}, p.~491, called the constructive technique in Eq.~(\ref{eq:armijo-1}) to obtain the step length $\epsilon$ the Goldstein-Armijo algorithm, since \cite{Goldstein.1965} and \cite{Goldstein.1967b} did not propose a method to solve for the step length, while \cite{Armijo.1966} did.  See also below Eq.~(\ref{eq:goldstein-2}) where it was mentioned that a bisection method can be used with Goldstein's rule.
}
\begin{align}
	\epsilon (\bparam) = \min_a \{\beta^a  \, \rho \, | \, J( \bparam + \beta^j \, \bd ) - J(\bparam) \le \alpha {\beta^a} \rho \, \bgrad \, \dotprod \, \bd \}  
	% \text{ such that the Armijo condition is satisfied}
	\label{eq:armijo-1}
\end{align}
where the decrease in the cost function along the descent direction $\bd$, denoted by $\Delta J$, was defined in Eq.~(\ref{eq:goldstein-2}), and the descent direction $\bd$ is related to the gradient $\bgrad$ via Eq.~(\ref{eq:descent-dir}).  The Armijo condition in Eq.~(\ref{eq:armijo-1})  can be rewritten as
\begin{align}
	J( \bparam + \epsilon \, \bd )
	\le
	J(\bparam)
	+
	\alpha \epsilon
	\, \bgrad \, \dotprod \, \bd
	\ ,
	\label{eq:armijo-3}
\end{align}
which is also known as the Armijo sufficient decrease condition, the first of the two Wolfe conditions presented below; see \cite{Wolfe.1969}, \cite{Polak.1997}, p.~55.\footnote{
	See also \cite{Nocedal.2006}, p.~34, \cite{Luenberger.2016}, p.~230.
}

Regarding the paramters $\alpha$, $\beta$, and $\rho$ in the Armijo's rule Eq.~(\ref{eq:armijo-1}), \cite{Armijo.1966} selected to fix 
\begin{align}
\alpha = \beta = \frac{1}{2} \ , \text{ and } \rho \in (0, +\infty) \ ,
\label{eq:armijo-parameters-1}
\end{align}
and proved a convergence theorem.  In practice, $\rho$ cannot be arbitrarily large.
Polak (1971) \cite{Polak.1971}, p.~36, also fixed $\alpha = \frac12$, but recommended to select $\beta \in (0.5, 0.8)$, based on numerical experiments.\footnote{
	See \cite{Polak.1971}, p.~301.
}, and to select\footnote{
	To satisfy the condition in Eq.~(\ref{eq:descent-dir}), the descent direction $\bd$ is required to satisfy $(- \bgrad) \dotprod \, \bd \ge \rho \parallel \bd \parallel \parallel \bgrad \parallel$, with $\rho > 0$, so to form an obtuse angle, bounded away from $90^\circ$, with the gradient $\bgrad$.
	But $\rho > 1$ would violate the Schwarz inequality, which requires $| \bgrad \dotprod \, \bd | \le \parallel \bd \parallel \parallel \bgrad \parallel$. 
} $\rho = 1$ to minimize the rate $r$ of geometric progression (from the iterate $\bparam_i$, for $i=0,1,2, \ldots$, toward the local minimizer $\bparam^\star$) for linear convergence:\footnote{
	%To alleviate the notation, the layer superscript $(\ell)$ is omitted in the notation for the iterate $\bparam_i$.
	The inequality in Eq.~(\ref{eq:rate-convergence}) leads to linear convergence in the sense that $| J (\bparam_{i+1}) - J(\bparam^\star) | \le r | J (\bparam_{i}) - J(\bparam^\star) |$, for $i=0,1,2, \ldots$, with $| J (\bparam_{0}) - J(\bparam^\star) |$ being a constant.
	See \cite{Polak.1971}, p.~245.
}
\begin{align}
	| J (\bparam_{i+1}) - J(\bparam^\star) |
	\le
	r^i
	| J (\bparam_{0}) - J(\bparam^\star) |
	\ ,
	\text{ with }
	r = 1 - \left(\frac{\rho m}{M}\right)^2
	\ ,
	\label{eq:rate-convergence}
\end{align} 
where $m$ and $M$ are the lower and upper bounds of the eigenvalues\footnote{
	A narrow valley with the minimizer $\bparam^\star$ at the bottom would have a very small ratio $m / M$.  See also the use of ``small heavy sphere'' (also known as ``heavy ball'') method to accelerate convergence in the case of narrow valley in Section~\ref{sc:SGD-momentum} on stochastic gradient descent with momentum.
} of the Hessian $\nabla^2 J$, thus $\frac{m}{M} < 1$.
In summary, \cite{Polak.1971} recommended:
\begin{align}
\alpha = \frac{1}{2} \ , \beta \in (0.5, 0.8) \ , \text{ and } \rho =1 \ .
\label{eq:armijo-parameters-2}
\end{align}

\begin{algorithm}
	{\bf Gradient descent with Armijo line-search, deterministic}
	\\
	\KwData{
		% Layer outputs $\byp{\ell}$, for $\ell=L, \cdots , 1$ 
		\\
		$\bullet$ Select an initial guess $\bparam_0$ (which is part of the network initialization)
		\\
		$\bullet$ Select 3 Armijo parameters $\alpha \in (0,1)$, $\beta \in (0,1)$, and $\rho >0 $, as in Eq.~(\ref{eq:armijo-parameters-1})
		\\
		\hspace{5mm} $\star$ Recommend using $\alpha = \frac12$, $\beta \in (0.5, 0.8)$, and $\rho = 1$, as in Eq.~(\ref{eq:armijo-parameters-2})
	}
	\KwResult{
		Local minimizer $\bparam^\star$ 
		%where gradient is zero, $\parallel {\bgrad (\hat{\bparam})} \parallel = \parallel \partial J (\hat{\bparam})/ \partial \bparam \parallel = 0$ (stopping criterion).
	}
	\vphantom{Blank line}
	Define gradient function $\bgrad (\cdot) := \partial J(\cdot) / \partial \bparam$, obtained through backpropagation; see Figure~\ref{fig:backprop-2}
	\;
	\ding{172} \For{$k = 0,1,2, \ldots$}{
		Compute steepest descent direction $\bd_k = - \bgrad_k = - \bgrad(\bparam_k)$ as in Eq.~(\ref{eq:descent-direction-quasiNewton})
		\;
		
		%\vphantom{Blank line}
		\vspace{2mm}
		
%		% gradient descent
 		\eIf{Descent direction is not zero, $\parallel \bd_k \parallel \ne 0$}{
%			
%			% THEN, gradient descent 
 			\vspace{2mm}
 			$\blacktriangleright$ Compute step length (learning rate) $\epsilon_k$ using Armijo's rule. \label{lst:line:armijo-begin}
 			\\
 			% Set $\mu = \rho$
 			Initialize step length $\epsilon_k = \rho$
 			\;
 			\ding{173} \For{a=1,2,\ldots}{
 				\eIf{Armijo decrease condition Eq.~(\ref{eq:armijo-3}) not satisfied}{
%					
%					% THEN, Armijo
 					$\blacktriangleright$ Decrease step length
 					\\
 					Set $\epsilon_k \leftarrow \beta \epsilon_k$
 					\;
 				}{
%					
%					% ELSE, Armijo
 					$\blacktriangleright$ Armijo decrease condition Eq.~(\ref{eq:armijo-3}) satisfied.
 					\\
 					Stop \ding{173} {\bf for} loop, use current step length $\epsilon_k$.
%					\;
%					
 				} % ENDIF, Armijo
 			} % END For loop, Armijo
 			\label{lst:line:armijo-end}
 			\vspace{2mm}
%			%{\color{red} HERE}
 			$\blacktriangleright$ Update network parameter $\bparam_k$ to $\bparam_{k+1}$ as in Eq.~(\ref{eq:update-params})
 			\\
 			Set $\bparam_{k+1} = \bparam_{k} + \epsilon_k \bd_k$
 			\;
 			\vspace{2mm}
 		}{
%			
%			% ELSE, gradient descent
 			\vspace{2mm}
 			$\blacktriangleright$ Descent direction is zero, $\parallel \bd_k \parallel = 0$
 			\;
 			Set the minimizer $\bparam^\star = \bparam_k$
 			\;
 			Stop \ding{172} {\bf for} loop.
 		} % ENDIF, gradient descent
 		\vspace{2mm}
 		$\blacktriangleright$ Continue to next iteration $(k+1)$
 		\\
 		Set $k \leftarrow k+1$
 		\;
	}

	\vphantom{Blank line} 
	\caption{
		\emph{Steepest gradient descent with Armijo line search, deterministic} (Section~\ref{sc:armijo}, Algorithms~\ref{algo:descent-armijo-stochastic-1}, \ref{algo:stochastic-newton}).
		%The parameter array $\bparam_k$, for $k=0,1,2, \ldots$, in this pseudocode, can represent either all network parameters, or the paramaters of layer $(\ell)$, in which case the pseudocode is applied to update the layer parameters $\bparamp{\ell}$ in Figure~\ref{fig:backprop-2}.  
		Even though Polak (1971) \cite{Polak.1971}, p.~36, recommended to select $\rho = 1$, this parameter is kept here to compare with the stochastic optimization methods in \cite{Paquette.2018} (Algorithm~\ref{algo:descent-armijo-stochastic-1}) and \cite{Bergou.2018} (Algorithm~\ref{algo:stochastic-newton}), where there was no such recommendation. 
		See Table~\ref{tb:armijo-params-notations} for a comparison of the notations used by several authors in relation to Armijo line search.		
	}
	\label{algo:descent-armijo-deterministic}
\end{algorithm}

\begin{algorithm}
	{\bf Quasi-Newton / Newton methods with Armijo line-search, deterministic}
	\\
	\KwData{
		% Layer outputs $\byp{\ell}$, for $\ell=L, \cdots , 1$ 
		\\
		$\bullet$ Select an initial guess $\bparam_0$ (which is part of the network initialization)
		\\
		$\bullet$ Select 3 Armijo parameters $\alpha \in (0,1)$, $\beta \in (0,1)$, and $\rho >0 $, as in Eq.~(\ref{eq:armijo-parameters-1})
		\\
		\hspace{5mm} $\star$ Recommend using $\alpha = \frac12$, $\beta \in (0.5, 0.8)$, and $\rho = 1$, as in Eq.~(\ref{eq:armijo-parameters-2})
		\\
		$\bullet$ Select stabilization parameter $\delta > 0$ if regularized Newton used, Eq.~(\ref{eq:descent-direction-regularized-newton})
	}
	\KwResult{
		Local minimizer $\bparam^\star$
	}
	
	\vphantom{Blank line}
	%Define gradient function $\bgrad (\cdot) := \partial J(\cdot) / \partial \bparam$, obtained through backpropagation; see Figure~\ref{fig:backprop-2}
	%\;
	%Define Hessian function $\bHs (\cdot) := \partial^2 J(\cdot) / (\partial \bparam)^2 $
	%\;
	
	\ding{172} \For{$k = 0,1,2, \ldots$}{
		
		Compute gradient $\bgrad_k = \bgrad(\bparam_k)$ at current network parameter $\bparam_k$
		\;
		
		%\vphantom{Blank line}
		%\vspace{2mm}
		
		% find descent direction
		%\vspace{2mm}
		$\blacktriangleright$ Find descent direction $\bd_k$
		\\
 		\eIf{Gradient not zero, $\parallel \bgrad_k \parallel \ne 0$}{
%			
%			% THEN, gradient descent 
%			%\vspace{2mm}
 			Compute Hessian $\bHs_k = \bHs (\bparam_k)$ at current network parameter $\bparam_k$
 			\;
%			%\vspace{2mm}
 			\eIf{Hessian inverse $\bHs_k^{-1}$ exists}{
%				
%				% THEN hessian inverse existing
%				%\vspace{2mm}
 				$\blacktriangleright$ Compute Newton descent direction  $\bd_k = - \bHs_k^{-1} \bgrad_k$, Eq.~(\ref{eq:descent-direction-newton})
 			}{
%				
%				% ELSE hessian inverse not existing, use gradient descent
%				%\vspace{2mm}
 				$\blacktriangleright$ Hessian inverse not existing	
 				\\
 				\If{Quasi-Newton used}{
 					Set descent direction $\bd_k = - \bgrad_k$, Eq.~(\ref{eq:descent-direction-quasiNewton})
					\;
 				}
%				\\
 				\If{Regularized-Newton used}{
 					Compute descent direction 
					% CMES style - 2022.06.01
%                                       $\bd_k = - \left[ \bHs_k  + \delta \bId \right]^{-1} \bgrad_k$, 
					% the problem of tex capacity exceeded was due to the nested definition of the macros such as \bId in the file
					% macros.tex.  So I explicitly coded up this equation; it worked
                                        $\bd_k = - \left[ \boldsymbol{H}_k  + \delta \boldsymbol{I} \right]^{-1} \boldsymbol{g}_k $
                                        Eq.~(\ref{eq:descent-direction-regularized-newton})
 				\;
 					\label{lst:line:newton-regularized}
 				}
 			} % ENDIF
%			
%			%\vspace{2mm}
 			$\blacktriangleright$ Compute step length $\epsilon_k$ using Armijo's rule Eq.~(\ref{eq:armijo-1}), lines~\ref{lst:line:armijo-begin}-\ref{lst:line:armijo-end} in Algorithm~\ref{algo:descent-armijo-deterministic}.
 			\\
%			
%			
%			%\vspace{2mm}
%			%{\color{red} HERE}
 			$\blacktriangleright$ Update network parameter $\bparam_k$ to $\bparam_{k+1}$:
 			%with step length $\epsilon_k$ and descent direction $\bd_k$
%			%\\
 			Set $\bparam_{k+1} = \bparam_{k} + \epsilon_k \bd_k$, Eq.~(\ref{eq:update-params})
 			\;
%			%\vspace{2mm}
%			
%			% END THEN
%			
 		}{
%			
%			% ELSE, gradient is zero
%			% \vspace{2mm}
 			$\blacktriangleright$ Gradient is zero, $\parallel \bgrad_k \parallel = 0$
 			\;
 			Set the minimizer $\bparam^\star = \bparam_k$;
%			%\;
 			Stop \ding{172} {\bf for} loop.
%			%\vspace{2mm}
%			
 		} % ENDIF, quasi-Newton
		
		%\vspace{2mm}
		$\blacktriangleright$ Continue to next iteration $(k+1)$;
		%\\
		Set $k \leftarrow k+1$
		\;
		
	} 	
	\vphantom{Blank line} 
	\caption{
		\emph{Quasi-Newton / Newton methods with Armijo line search, deterministic.}
		Even though \cite{Polak.1971}, p.~36, recommended to select $\rho = 1$, this parameter is kept here to compare with the pseudocodes by \cite{Paquette.2018} (Algorithm~\ref{algo:descent-armijo-stochastic-1}) and by \cite{Bergou.2018} (Algorithm~\ref{algo:stochastic-newton}), in which there is no such recommendation. 
		See Table~\ref{tb:armijo-params-notations} for a comparison of the notations used by several authors in relation to Armijo line search.		
	}
	\label{algo:gradient-quasi-newton-armijo-deterministic}
\end{algorithm}

The pseudocode for deterministic gradient descent with Armijo line search is Algorithm~\ref{algo:descent-armijo-deterministic}, and the pseudocode for deterministic quasi-Newton / Newton with Armijo line search is Algorithm~\ref{algo:gradient-quasi-newton-armijo-deterministic}. 
When the Hessian $\bHs (\bparam) = \partial^2 J (\bparam) / (\partial \bparam)^2$ is positive definite, the Newton descent direction is:
\begin{align}
	\bd = - \bHs^{-1} (\bparam) \, \bgrad
	\ ,
	\label{eq:descent-direction-newton}
\end{align}
When the Hessian $\bHs (\bparam)$ is not positive definite, e.g., near a saddle point, then quasi-Newton method uses the gradient descent direction $(- \bgrad = - \partial J / \partial \bparam)$ as the descent direction $\bd$ as in Eq.~(\ref{eq:update-params}), 
\begin{align}
	\bd 
	= - \bgrad 
	= - \partial J (\bparam) / \partial \bparam
	\ ,
	\label{eq:descent-direction-quasiNewton}
\end{align}
and regularized Newton method uses a descent direction based on a regularized Hessian of the form: 
\begin{align}
	% CMES style - 2022.06.01
	% \bd = - \left[ \bHs (\bparam) + \delta \bId \right]^{-1} \bgrad
	% commented out the above equation since the nested definition of the macro \bId created the bug of TeX capacity exceeded
	%
	% \bd = - \left[ \boldsymbol{H} (\bparam) + \delta \boldsymbol{I} \right]^{-1} \bgrad
	% the problem is in \bId, not in \bHs
	\bd = - \left[ \bHs (\bparam) + \delta \boldsymbol{I} \right]^{-1} \bgrad
	\ ,
	\label{eq:descent-direction-regularized-newton}
\end{align}
where $\delta$ is a small perturbation parameter (line~\ref{lst:line:newton-regularized} in Algorithm~\ref{algo:gradient-quasi-newton-armijo-deterministic} for deterministic Newton and line~\ref{lst:line:newton-descent-regularized} in Algorithm~\ref{algo:stochastic-newton} for stochastic Newton).\footnote{
	See, e.g., \cite{Polak.1997}, p.~35, and \cite{Goodfellow.2016}, p.~302, where both cited the Levenberg-Marquardt algorithm as the first to use regularized Hessian. 
}

%{\color{red}	HERE 2019.11.12.}
%{\color{red} HERE 2020.01.11}

%\vspace{5mm}
%\noindent
%{\bf Variant 4:} \emph{Inexact line-search, Wolfe's rule.}
\subsubsection{Inexact line-search, Wolfe's rule}
\label{sc:inexact-line-search-wolfe}
%
% CMES style, rewriting and correcting
The rule introduced in \cite{Wolfe.1969} and \cite{Wolfe.1971},\footnote{
%	As of 2019.11.02, \cite{Wolfe.1969} was cited 1012 times 
	As of 2022.07.09, \cite{Wolfe.1969} was cited 1336 times
	in various publications (books, papers) according to Google Scholar, and 
%	442 times
    559 times 
	in archival journal papers according to Web of Science.   
} sometimes called the Armijo-Goldstein-Wolfe's rule (or conditions), particularly in \cite{Lewis.2000} and \cite{Kolda.2003},\footnote{
	%
	% CMES style, rewriting
	The authors of \cite{Lewis.2000} and \cite{Kolda.2003} may not be aware that Goldstein's rule appeared before Armijo's rule, as they cited Goldstein's 1967 book \cite{Goldstein.1967}, instead of Goldstein's 1965 paper \cite{Goldstein.1965}, and referred often to Polak (1971) \cite{Polak.1971}, even though it was written in \cite{Polak.1971}, p.~32, that a ``step size rule [Eq.~(\ref{eq:goldstein-2})] probably first introduced by Goldstein (1967) \cite{Goldstein.1967}'' was used in an algorithm.   See also Footnote~\ref{fn:polak-goldstein}.
} has been extended to add stochasticity \cite{Mahsereci.2017},\footnote{
	An earlier version of the 2017 paper \cite{Mahsereci.2017} is the 2015 preprint \cite{Mahsereci.2015}.
} is stated as follows: For $0 < \alpha < \beta < 1$, select the step length (learning rate) $\learn$ such that (see, e.g., \cite{Polak.1997}, p.~55):
\begin{align}
	&
	J( \bparam + \learn \, \bd )
	\le
	J(\bparam)
	+
	\alpha \epsilon
	\, \bgrad \, \dotprod \, \bd
	\ ,
	\label{eq:wolfe-1}
	\\
	&
	\frac{\partial J ( \bparam + \learn \, \bd )}{\partial \bparam} 
	\, \dotprod \, \bd
	\ge
	\beta
	\, \bgrad \, \dotprod \, \bd
	\ .
	\label{eq:wolfe-2}
\end{align}
The first Wolfe's rule in Eq.~(\ref{eq:wolfe-1}) is the same as the Armijo's rule in Eq.~(\ref{eq:armijo-3}), which ensures that at the updated point $( \bparam + \learn \, \bd )$ the cost function value $J ( \bparam + \learn \, \bd )$ is below the green line $\alpha \epsilon \, \bgrad \, \dotprod \, \bd$ in Figure~\ref{fig:goldstein-1}. 

The second Wolfe's rule in Eq.~(\ref{eq:wolfe-2}) is to ensure that  at the updated point $( \bparam + \learn \, \bd )$ the slope of the cost function cannot fall below the (negative) slope of the purple line $\beta \, \learn \, \bgrad \, \dotprod \, \bd$ in Figure~\ref{fig:goldstein-1}.

For other variants of line search, we refer to \cite{Shi.2005}.

\subsection{Stochastic gradient-descent (1st-order) methods}
\label{sc:stochastic-gradient-descent}

To avoid confusion,\footnote{
	See \cite{Goodfellow.2016}, p.~271, about this terminology confusion.   
	%
	% CMES style, rewriting
	The authors of
	\cite{Bottou.2018:rd0001} used ``stochastic'' optimization to mean optimization using random ``minibatches'' of examples, and ``batch'' optimization to mean optimization using ``full batch'' or full training set of examples.
} we will use the terminology ``full batch'' (instead of just ``batch'') when the entire training set is used for training.   a minibatch is a small subset of the training set.

In fact, as we shall see, and as mentioned in Remark~\ref{rm:classic-never-dies}, classical optimization methods mentioned in Section~\ref{sc:deterministic-optimization} have been developed further to tackle new problems, such as noisy gradients, encountered in deep-learning training with random mini-batches.  There is indeed much room for new research on learning rate since:
\begin{quote}
	``The learning rate may be chosen by trial and error.   This is more of an art than a science, and most guidance on this subject should be regarded with some skepticism.'' \cite{Goodfellow.2016}, p.~287.
\end{quote}
%Indeed, we refer the readers to Remark~\ref{rm:classic-never-dies}.

At the time of this writing, we are aware of two review papers on optimization algorithms for machine learning, and in particular deep learning, aiming particularly at experts in the field: \cite{Bottou.2018:rd0001}, as mentioned above, and 
\vphantom{\cite{Sun.2019}}\cite{Sun.2019}.  Our review complements these two review papers.  We are aiming here at bringing first-time learners up to speed to benefit from, and even to hopefully enjoy, reading these and others related papers.
%\cite{Bottou.2018:rd0001} can be overwhelming due to a relatively heavy mathematical programming language for those outside the field.
To this end, we deliberately avoid the dense mathematical-programming language, not familiar to readers outside the field, as used in \cite{Bottou.2018:rd0001}, while providing more details on algorithms that have proved important in deep learning than \cite{Sun.2019}. 

Listed below are the points that distinguish the present paper from other reviews.  Similar to \cite{Goodfellow.2016}, both \cite{Bottou.2018:rd0001} and \cite{Sun.2019}: 
\begin{itemize}
	
	\item 
	Only mentioned briefly in words the connection between SGD with momentum to mechanics without detailed explanation using the equation of motion of the ``heavy ball'', a name not as accurate as the original name ``small heavy sphere'' by Polyak (1964) \cite{Polyak.1964}.  These references also did not explain how such motion help to accelerate convergence; see Section~\ref{sc:SGD-momentum}.
	
	\item 
	Did not discuss recent practical add-on improvements to SGD such as step-length tuning (Section~\ref{sc:step-length-tuning}) and step-length decay (Section~\ref{sc:step-length-decay}), as proposed in \vphantom{\cite{Wilson.2018}}\cite{Wilson.2018}.  This information would be useful for first-time learners.
	
	\item
	Did not connect step-length decay to simulated annealing, and did not explain the reason for using the name ``annealing''\footnote{
		\label{fn:Sun.2019-annealing}
		%
		% CMES style, rewriting
		The authors of
		\cite{Sun.2019} only cited 
		\cite{Kirkpatrick.1983} for a brief mention of ``simulated annealing'' as an example of ``heuristic optimizers'', with no discussion, and no connection to step length decay.  See also Remark~\ref{rm:metaheuristics} on ``Metaheuristics''.
	} in deep learning by connecting to stochastic differential equation and physics; see Remark~\ref{rm:annealing} in Section~\ref{sc:minibatch-size-increase}.  
	
	\item
	Did not review an alternative to step-length decay by increasing minibatch size, which could be more efficient, as proposed in \vphantom{\cite{SmithSL.2018b}}\cite{SmithSL.2018b}; see Section~\ref{sc:minibatch-size-increase}.
	
	\item 
	Did not point out that the exponential smoothing method (or running average) used in adaptive learning-rate algorithms dated since the 1950s in the field of forecasting.  None of these references acknowledged the contributions 
	%
	% CMES style, rewriting
%	of \cite{Schraudolph.1998} and \cite{Neuneier.1998} who were
	made in \cite{Schraudolph.1998} and \cite{Neuneier.1998}, in which
	%
	% CMES style, rewriting
%	probably the first to bring 
	exponential smoothing from time series in forecasting was probably first brought to machine learning.  See Section~\ref{sc:exponential-smoothing}.
	
	\item 
	Did not discuss recent adaptive learning-rate algorithms such as \hyperref[para:adamw]{AdamW} 
	% 
	% CMES style, rewriting
%	by 
	\cite{Loshchilov.2019}.\footnote{
		%
		% CMES style, rewriting
		The authors of
		\cite{Sun.2019} only cited \cite{Loshchilov.2019} in passing, without reviewing \hyperref[para:adamw]{AdamW}, which was not even mentioned.
	}  These authors also did not discuss the criticism of adaptive methods in \cite{Wilson.2018}; see Section~\ref{sc:adamw}.
	
	\item 
	Did not discuss classical line-search rules---such as \cite{Goldstein.1965}, \cite{Armijo.1966},\footnote{
		The authors of
		\cite{Bottou.2018:rd0001} only cited Armijo (1966) \cite{Armijo.1966} once for a pseudocode using line search.
	} \cite{Wolfe.1969} (Sections~\ref{sc:inexact-line-search-goldstein}, \ref{sc:inexact-line-search-armijo}, \ref{sc:inexact-line-search-wolfe})---that have been recently generalized to add stochasticity, e.g., \cite{Mahsereci.2017}, \cite{Paquette.2018}, \cite{Bergou.2018}; see Sections~\ref{sc:SGD-armijo}, \ref{sc:stochastic-Newton}. 
	
\end{itemize}

\subsubsection{Standard SGD, minibatch, fixed learning-rate schedule}
\label{sc:generic-SGD}

The stochastic gradient descent algorithm, originally introduced 
%
% CMES style rewriting
by 
Robbins \& Monro (1951a)
\cite{Robbins1951a} (another classic) according to many sources,\footnote{
	See, e.g., \cite{Bottou.2018:rd0001}---in which there was a short bio of
	%
	% CMES style rewriting
	Robbins, 
	the first author of \cite{Robbins1951a}---and \cite{Sun.2019} \cite{Paquette.2018}. 
} has been playing an important role in training deep-learning networks:
\begin{quote}
	``Nearly all of deep learning is powered by one very important algorithm: stochastic gradient descent (SGD). Stochastic gradient descent is an extension of the gradient descent algorithm.'' \cite{Goodfellow.2016}, p.~147.
\end{quote} 

{\bf Minibatch.}
\label{para:minibatch}
The number $\Bsize$ of examples in a training set $\Xbb$ could be very large, rendering prohibitively expensive to evaluate the cost function and to compute the gradient of the cost function with respect to the number of parameters, which by itself could also be very large.  At iteration $k$ within a training session $\tau$, let $\Ibbps{k}{\bsize}$ be a randomly selected set of $\bsize$ indices, which are elements of the training-set indices $[\Bsize] = \{1, \ldots, \Bsize\}$. Typically, $\bsize$ is much smaller than $\Bsize$:\footnote{
	\label{fn:imagenet-size-notation}
	As of 2010.04.30, the ImageNet database contained more than 14 million images; see \href{http://www.image-net.org/about-stats}{Original website}, \href{https://web.archive.org/web/20191002051313/http://www.image-net.org/about-stats}{Internet archive}, Figure~\ref{fig:ImageNet-error} and Footnote~\ref{fn:imagenet}.   There is a slight inconsistency in notation in \cite{Goodfellow.2016}, where on p.~148, $m$ and $m^\prime$ denote the number of examples in the training set and in the minibatch, respectively, whereas on p.~274, $m$ denote the number of examples in a minibatch.  In our notation, $m$ is the dimension of the output array $\by$, whereas $\bsize$ (in a different font) is the minibatch size; see Footnote~\ref{fn:output-examples-size-notation}.  In theory, we write $\bsize \le \Bsize$ in Eq.~(\ref{eq:minibatch-1}); in practice, $\bsize \ll \Bsize$.
}  
\begin{quote}
	``The minibatch size $\bsize$ is typically chosen to be a relatively small number of examples, ranging from one to a few hundred. Crucially, $\bsize$ is usually held fixed as the training set size $\Bsize$ grows. We may fit a training set with billions of examples using updates computed on only a hundred examples.'' \cite{Goodfellow.2016}, p.~148.
\end{quote}
Generated as in Eq.~(\ref{eq:minibatch-1}), the random-index sets $\Ibbps{k}{\bsize}$, for $k=1,2,\ldots$, are non-overlapping such that after $k_{max} = \Bsize / \bsize$ iterations, all examples in the training set are covered, and a training session, or training epoch,\footnote{
	\label{fn:epoch}
	An epoch, or training session, $\tau$ is explicitly defined here as when the minibatches as generated in Eqs.~(\ref{eq:minibatch-0})-(\ref{eq:minibatch-2}) covered the whole dataset.  In \cite{Goodfellow.2016}, the first time the word ``epoch'' appeared was in Figure~7.3 caption, p.~239, where it was defined as a ``training iteration'', but there was no explicit definition of ``epoch'' (when it started and when it ended), except indirectly as a ``training pass through the dataset'', p.~274.
	See Figure~\ref{fig:Haibe-Kains-irreproducibility} in Section~\ref{sc:irreproducibility} on ``Lack of transparency and irreproducibility of results'' in recent deep-learning papers.
} is completed (line~\ref{lst:list:SGD-kmax} in Algorithm~\ref{algo:generic-SGD}).
At iteration $k$ of a training epoch $\tau$, the random minibatch $\Bbbps{k}{\bsize}$ is a set of $\bsize$ examples pulled out from the much larger training set $\Xbb$ using the random indices in $\Ibbps{k}{\bsize}$, with the corresponding targets in the set $\Tbbps{k}{\bsize}$: 
\begin{align}
	&
	\Ms{1} = \{ 1 , \ldots , \Bsize \} =: [\Bsize]
	\ , \ k_{max} = \Bsize / \bsize
	\label{eq:minibatch-0}
	\\
	&
	\Ibbps{k}{\bsize} = \{i_{1, k} , \ldots , i_{\bsize, k} \} \subseteq \Ms{k}
	\ , \ 
	\Ms{k+1} = \Ms{k} - \Ibbps{k}{\bsize}
	\text{ for } k = 1,\ldots , k_{max}
	\label{eq:minibatch-1}
	\\
	&
	\Bbbps{k}{\bsize}
	= 
	\{ \bexin{i_{1,k}} , \cdots , \bexin{i_{\bsize,k}}  \}
	\subseteq
	\Xbb 
	=
	\{ \bexin{1} , \cdots , \bexin{\Bsize}  \}
	\label{eq:minibatch-2}
	\\
	&
	\Tbbps{k}{\bsize}
	=
	\{ \bexout{i_{1,k}} , \cdots , \bexout{i_{\bsize,k}}  \}
	\subseteq
	\Ybb 
	=
	\{ \bexout{1} , \cdots , \bexout{\Bsize}  \}
	\label{eq:minibatch-3}
\end{align}
Note that once the random index set $\Ibbps{k}{\bsize}$ had been selected, it was deleted from its superset $\Ms{k}$ to form $\Ms{k+1}$ so the next random set $\Ibbps{j+1}{\bsize}$ would not contain indices already selected in $\Ibbps{k}{\bsize}$.  

Unlike the iteration counter $k$ within a training epoch $\tau$,
the global iteration counter $j$ is not reset to 1 at the beginning of a new training epoch $\tau+1$, but continues to increment for each new minibatch. 
Plots versus epoch counter $\tau$ and plots versus global iteration counter $j$ could be confusing; see Remark~\ref{rm:epoch-vs-iteration} and Figure~\ref{fig:newton-examples}.

{\bf Cost and gradient estimates.}
The cost-function estimate is the average of the cost functions, each of which is the cost function of an example $\bexin{i_k}$ in the minibatch for iteration $k$ in training epoch $\tau$:
\begin{align}
	\losst (\bparam) = \frac{1}{\bsize} \sum_{a=1}^{a=\bsize \le \Bsize} J_{i_{a}} (\bparam)
	\ , 
	\text{ with }
	J_{i_a} (\bparam) = J(f(\bexin{i_a} , \bparam) , \bexout{i_a})
	\ , 
	\text{ and }
	\bexin{i_a} \in \Bbbps{k}{\bsize}
	\ ,
	\bexout{i_a} \in \Tbbps{k}{\bsize}
	\ ,
	\label{eq:cost-estimate}
\end{align}
where we wrote the random index as $i_{a}$ instead of $i_{a,k}$ as in Eq.~(\ref{eq:minibatch-1}) to alleviate the notation. 
The corresponding gradient estimate is:
\begin{align}
	\bgradt (\bparam) 
	= 
	\frac{\partial \losst (\bparam)}{\partial \bparam} 
	=
	\frac{1}{\bsize} \sum_{a=1}^{a=\bsize \le \Bsize}
	\frac{\partial \loss_{i_a} (\bparam)}{\partial \bparam}
	=
	\frac{1}{\bsize} \sum_{a=1}^{a=\bsize \le \Bsize}
	\bgrad_{i_a}
	\ .
	\label{eq:gradient-estimate}
\end{align}

The pseudocode for the standard SGD\footnote{
	See also \cite{Goodfellow.2016}, p.~286, Algorigthm 8.1; \cite{Bottou.2018:rd0001}, p.~243, Algorithm 4.1. 
} is given in Algorithm~\ref{algo:generic-SGD}.  The epoch stopping criterion (line~\ref{lst:list:SGD-budget-not-met} in Algorithm~\ref{algo:generic-SGD}) is usually determined by a computation ``budget'', i.e., the maximum number of epochs allowed.  For example, \cite{Bergou.2018} set a budget of 1,600 epochs maximum in their numerical examples.

{\bf Problems and resurgence of SGD.}
There are several known problems with SGD:
\begin{quote}
	``Despite the prevalent use of SGD, it has known challenges and inefficiencies. First, the direction may not represent a descent direction, and second, the method is sensitive to the step-size (learning rate) which is often poorly overestimated.'' \cite{Paquette.2018}
\end{quote}
For the above reasons, it may not be appropriate to use the norm of the gradient estimate being small as stationarity condition, i.e., where the local minimizer or saddle point is located; see the discussion in \cite{Bergou.2018} and stochastic Newton Algorithm~\ref{algo:stochastic-newton} in Section~\ref{sc:stochastic-Newton}.

Despite the above problems, SGD has been brought back to the forefront state-of-the-art algorithm to beat, surpassing the performance of adaptive methods, as confirmed by three recent papers: \cite{Wilson.2018}, \cite{Aitchison.2019}, \cite{Loshchilov.2019}; see Section~\ref{sc:adam-criticism} on criticism of adaptive methods.  

\vspace{2ex}
%\subsection{Add-on tricks: Momentum, tuning, decaying}
{\bf Add-on tricks to improve SGD.}
\label{sc:add-on-tricks}
The following tricks can be added onto the vanilla (standard) SGD to improve its performance; see also the pseudocode in Algorithm~\ref{algo:generic-SGD}: 
%(1) Momentum and accelerated gradient, (2) step-size tuning, (3) step-size decay or annealing, (4) cyclic annealing.
\begin{itemize}
	
	\item 
	Momentum and accelerated gradient: Improve (accelerate) convergence in narrow valleys, Section~\ref{sc:SGD-momentum}
	
	\item 
	Initial-step-length tuning: Find effective initial step length $\learn_0$, Section~\ref{sc:step-length-tuning}
	
	\item 
	Step-length decaying or annealing: Find an effective learning-rate schedule\footnote{
		See Figure~\ref{fig:Haibe-Kains-irreproducibility} in Section~\ref{sc:irreproducibility} on ``Lack of transparency and irreproducibility of results'' in recent deep-learning papers.
	} to decrease the step length $\learn$ as a function of epoch counter $\tau$ or global iteration counter $j$, cyclic annealing, Section~\ref{sc:step-length-decay}
	
	\item
	Minibatch-size increase, keeping step length fixed, equivalent annealing, Section~\ref{sc:minibatch-size-increase}
				
	\item 
	Weight decay, Section~\ref{sc:weight-decay}	
	%{\color{red} HERE 2020.01.11}
	
\end{itemize}

\begin{figure}[h]
	\centering
	\includegraphics[width=0.5\linewidth]{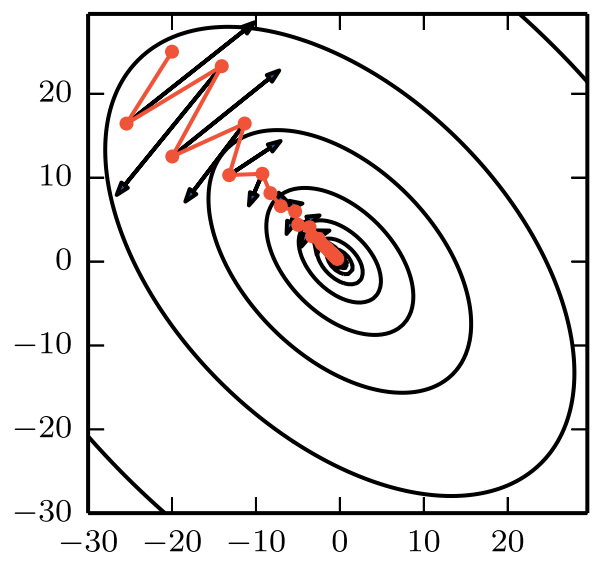}
	\caption{
		\emph{SGD with momentum, small heavy sphere} Section~\ref{sc:SGD-momentum}.  The descent direction (negative gradient, black arrows) bounces back and forth between the steep slopes of a deep and narrow valley.  The small-heavy-sphere method, or SGD with momentum, follows a faster descent (red path) toward the bottom of the valley. See the cost-function landscape with deep valleys in Figure~\ref{fig:cost-function-landscape}. 
		Figure from \cite{Goodfellow.2016}, p.~289.
		{\footnotesize (Figure reproduced with permission of the authors.)}
		%{\color{red} ASK PERMISSION 2019.11.25}
	}
	\label{fig:SGD-momentum}
\end{figure}

\begin{algorithm}
	%{\bf Standard SGD, fixed learning-rate schedule and fixed minibatch size}
	{\bf SGD, standard, momentum, accelerated gradient}
	\\
	\KwData{
		% Layer outputs $\byp{\ell}$, for $\ell=L, \cdots , 1$ 
		\\
		$\bullet$ Select an initial guess $\bparamt_0$ (which is part of the network initialization)
		\\
		$\bullet$ Select variable $\tei$ as epoch $\tau$ or as global iteration $j$, Eq.~(\ref{eq:learning-rate-schedule-b}), and budget $\tau_{max}$ or $j_{max}$ \label{lst:list:SGD-budget-not-met}
		\\
		$\bullet$ Select learning-rate schedule $\learn(\tei) \in \{$ \text{Eq.~(\ref{eq:learning-rate-schedule})-Eq.~(\ref{eq:learning-rate-schedule-4})} $\}$, and parameters $\learn_0, k_c, \decay$ 
		\\
		$\bullet$ Select parameter $\mompar$ for momentum, and $\nesterov$ for Nesterov Accelerated Gradient, Eq.~(\ref{eq:SGD-update-momentum}), if use
		\\
		$\bullet$ Select cosine annealing multiplier $\anneal (\tei)$, Eq.~(\ref{eq:cosine-annealing}), if use
		\\
		$\bullet$ Select a value for $\bsize$ the number of examples in the minibatches
	}
	\KwResult{
		Local minimizer estimate $\bparamt^\star$, satisfying a stopping criterion, Section~\ref{sc:training-valication-test}.
	}
	\vphantom{Blank line}
	%Define gradient function estimate $\bgradt (\cdot) := \partial \losst (\cdot) / \partial \bparam$ as in Eq.~(\ref{eq:gradient-estimate})%, obtained through backpropagation; see Figure~\ref{fig:backprop-2}
	%\;
	Initialize network parameters $\bparamt_1^\star = \bparamt_0$ and global iteration counter $j=1$
	\;
	\ding{172} 
	\For{$\tau = 1,2, \ldots, \tau_{max}$}{
		
		$\blacktriangleright$ Begin training epoch $\tau$ 
		%$\blacktriangleright$ Begin stochastic gradient descent
		\\
		Initialize $\bparamt_1 = \bparamt_\tau^\star$ (from previous training epoch) \label{lst:list:SGD-params-previous-epoch}
		\;
		% gradient descent
		\ding{173}
		\label{lst:list:SGD-kmax}
		\For{$k=1, \ldots , k_{max}$, Eq.~(\ref{eq:minibatch-0})}{
			
			Obtain random minibatch $\Bbbps{k}{\bsize}$ containing $\bsize$ examples, Eq.~(\ref{eq:minibatch-2}) \label{lst:list:SGD-minibatch}
			\;
			Compute steepest-descent direction estimate $\bdt_k = - \bgradt (\bparamt_k)$, Eq.~(\ref{eq:gradient-estimate})
			\;
			Compute learning rate $\learn_{k} = \learn(\tei)$
			\;
			
			$\blacktriangleright$ Update network parameter $\bparamt_k$ to $\bparamt_{k+1}$,   Eq.~(\ref{eq:update-params-combined-tricks}) \label{lst:line:SGD-update}
			\\
			\eIf{$j < j_{max}$ (if use global-iteration budget)}{
				Increment global iteration counter $j \leftarrow j + 1$
			}{
				Stop computation; exit
				\;
			}
				
		} % END For loop
	
		\eIf{Stopping criterion not met (if used, Section~\ref{sc:training-valication-test})}{
			$\blacktriangleright$ Continue to next training epoch $(\tau+1)$
			\\
			Update minimizer estimate for epoch $(\tau+1)$: $\bparamt_{\tau+1}^\star = \bparamt_{k_{max}}$
			\\
			Set $\tau \leftarrow \tau+1$
			\;
		}{
			$\blacktriangleright$ Stopping criterion met
			\\
			Set optimal paramter $\bparamt^\star = \bparamt_\tau$;
			Stop computation, exit \;
		}
				
	} % END For loop over epochs
	
	%\vspace{2mm}
	\If{Stopping criterion NOT used (Section~\ref{sc:training-valication-test})}{
		Find validation-error global minimum at epoch $\tau^\star$, and set optimal parameter $\bparamt^{\star} = \bparamt_{\tau^\star}$ at $\tau^\star$; see Figure~\ref{fig:validation-error-ugly} in Section~\ref{sc:training-valication-test}.
	}
	
	%Accept minimizer estimate from last epoch:
	%\\
	%Set $\bparamt^\star \leftarrow \bparamt^\star_{\tau_{max}}$.

	\vphantom{Blank line} 
	\caption{
		\emph{SGD, standard, momentum, accelerated gradient, cyclic annealing} (Sections~\ref{sc:training-valication-test}, \ref{sc:generic-SGD}, \ref{sc:SGD-momentum}, \ref{sc:SGD-combined-tricks}, \ref{sc:unified-adaptive}, \ref{sc:SGD-armijo}, \ref{sc:stochastic-Newton},  Algorithms~\ref{algo:unified-adaptive-learning-rate-2}, \ref{algo:descent-armijo-stochastic-1}, \ref{algo:stochastic-newton}), with fixed learning-rate schedule $\learn (\tei)$ and fixed minibatch size $\bsize$.  
		The case with constant learning rate $\learn_{k} = \learn_0$, and minibatch-size increasing in Section~\ref{sc:minibatch-size-increase} is not implemented here to keep the pseudocode simple.
		For adaptive learning-rate methods, see Section~\ref{sc:adaptive-learning-rate-algos}. 
		For SGD with stochastic Armijo length search and adaptive minibatch, see Algorithm~\ref{algo:descent-armijo-stochastic-1}, where the outer \ding{172} {\bf for} loop over training epochs is omitted for conciseness, and similarly for Algorithm~\ref{algo:unified-adaptive-learning-rate-2} and Algorithm~\ref{algo:stochastic-newton}.
	}
	\label{algo:generic-SGD}
\end{algorithm}

%{\color{red} HERE 2019.12.29}

%{\bf Momentum.}  
\subsubsection{Momentum and fast (accelerated) gradient}
\label{sc:SGD-momentum}
The standard update for gradient descent is Eq.~(\ref{eq:update-params}) would be slow when encountering deep and narrow valley, as shown in Figure~\ref{fig:SGD-momentum}, and can be replaced by the general update with momentum as follows:
\begin{align}
	\bparamt_{k+1} = \bparamt_{k} 
	- \learn_k \bgradt (\bparamt_{k} + \nesterov_k (\bparamt_k - \bparamt_{k-1}) )
	+ \mompar_k (\bparamt_k - \bparamt_{k-1})
	\ ,
	\label{eq:SGD-update-momentum}
\end{align} 
from which the following methods are obtained (line~\ref{lst:line:SGD-update} in Algorithm~\ref{algo:generic-SGD}): 
\begin{itemize}
	\item
	Standard SGD update Eq.~(\ref{eq:update-params}) with $\nesterov_k = \mompar_k = 0$ \cite{Robbins1951b}
	
	\item 
	SGD with classical momentum: $\nesterov_k = 0$ and $\mompar_k \in (0,1)$ (``small heavy sphere'' or heavy point mass)\footnote{
		Often called by the more colloquial ``heavy ball'' method; see Remark~\ref{rm:heavy-ball}.
	} \cite{Polyak.1964}
	
	\item
	SGD with fast (accelerated) gradient:\footnote{
		Sometimes referred to as Nesterov's Accelerated Gradient (NAG) in the deep-learning literature.
	} $\nesterov_k = \mompar_k \in (0,1)$, Nesterov 
%(\citeyear{Nesterov.1983}, \citeyear{Nesterov.2018})
	(1983 \cite{Nesterov.1983}, 2018 \cite{Nesterov.2018})
\end{itemize}

The continuous counterpart of the parameter update Eq.~(\ref{eq:SGD-update-momentum}) with classical momentum, i.e., when $\nesterov_k \in (0,1)$ and $\mompar_k = 0$, is the equation of motion of a heavy point mass (thus no rotatory inertia) under viscous friction at slow motion (proportional to velocity) and applied force $-\bgradt$ given below with its discretization by finite difference in time, where $h_k$ and $h_{k-1}$ are the time-step sizes \cite{Goudou.2009}:
\begin{align}
	&
	\frac{d^2 \bparamt}{(dt)^2}
	+
	\nu
	\frac{d \bparamt}{dt}
	=
	-
	\bgradt
	\Rightarrow
	\displaystyle
	\frac{\displaystyle \frac{\bparamt_{k+1} - \bparamt_{k}}{h_k} - \frac{\bparamt_{k} - \bparamt_{k-1}}{h_{k-1}}}{h_k}
	+
	\nu
	\frac{\bparamt_{k+1} - \bparamt_{k}}{h_k}
	=
	-
	\bgradt_{k}
	\ ,
	\label{eq:SGD-momentum-2}
	\\
	&
	\bparamt_{k+1} - \bparamt_{k} - \mompar_k (\bparamt_k - \bparamt_{k-1}) = - \learn_k \bgradt_k
	\ ,
	\text{ with }
	\mompar_k = \frac{h_{k-1}}{h_k} \frac{1}{1 + \nu h_k}
	\text{ and }
	\learn_k = (h_k)^2 \frac{1}{1 + \nu h_k}
	\ ,
	\label{eq:SGD-momentum-3}
\end{align}
which is the same as the update Eq.~(\ref{eq:SGD-update-momentum}) with $\nesterov_k = 0$.  The term $\mompar_k (\bparam_k - \bparam_{k-1})$ is often called the ``momentum'' term since it is proportional to (discretized) velocity.  \cite{Polyak.1964} on the other hand explained the term $\mompar_k (\bparam_k - \bparam_{k-1})$ as ``giving inertia to the motion, 
[leading] to motion along the  ``essential''  direction,  i.e.  along  `the bottom of the trough' '', and recommended to select $\mompar_k \in (0.8, 0.99)$, i.e., close to 1, without explanation.  The reason is to have low friction, i.e., $\nu$ small, but not zero friction ($\nu = 0$), since friction is important to slow down the motion of the sphere up and down the valley sides (like skateboarding from side to side in a half-pipe), thus accelerate convergence toward the trough of the valley; from Eq.~(\ref{eq:SGD-momentum-3}), we have
\begin{align}
	h_k = h_{k-1} = h
	\text{ and }
	\nu \in [0, +\infty)
	\Rightarrow
	\mompar_k \in (0,1]
	\ ,
	\text{ with }
	\nu = 0 \Rightarrow \mompar_k = 1
	\label{eq:SGD-momentum-params}
\end{align}

\begin{rem}
	\label{rm:choice-momentum-param}
	{\rm
		The choice of the momentum parameter $\mompar$ in Eq.~(\ref{eq:SGD-update-momentum}) is not trivial.   If $\mompar$ is too small, the signal will be too noisy; if $\mompar$ is too large, ``the average will lag too far behind the (drifting) signal'' \cite{Schraudolph.1998}, p.~212.
		Even though Polyak (1964) \cite{Polyak.1964} recommended to select $\mompar \in (0.8, 0.99)$, as explained above,
		%
		% CMES style rewriting
%		\cite{Goodfellow.2016}, p.~290, reported:
		it was reported in \cite{Goodfellow.2016}, p.~290:
		``Common values of $\mompar$ used in practice include 0.5, 0.9, and 0.99. Like the learning rate, $\mompar$ may also be adapted over time. Typically it begins with a small value and is later raised. Adapting $\mompar$ over time is less important than shrinking $\learn$ over time''.  
		The value of $\mompar = 0.5$ would correspond to relatively high friction $\mu$, slowing down the motion of the sphere, compared to $\mompar = 0.99$.
		
		Figure~\ref{fig:Adam-converge} from \cite{Kingma.2014} shows the convergence of some adaptive learning-rate algorithms: \hyperref[para:adagrad]{AdaGrad}, \hyperref[para:rmsprop]{RMSProp}, 
		\hyperref[sc:SGD-momentum]{SGDNesterov} (accelerated gradient), \hyperref[para:adadelta]{AdaDelta}, \hyperref[para:adam1]{Adam}.  
		 
		In their remarkable paper, the authors of \vphantom{\cite{Wilson.2018}}\cite{Wilson.2018} used a constant momentum parameter $\mompar = 0.9$; see \hyperref[para:adam-criticism]{criticism of adaptive methods} in Section~\ref{sc:adaptive-learning-rate-algos} and Figure~\ref{fig:wilson-examples} comparing \hyperref[sc:SGD-momentum]{SGD}, \hyperref[sc:SGD-momentum]{SGD with momentum}, \hyperref[para:adagrad]{AdaGrad}, \hyperref[para:rmsprop]{RMSProp}, \hyperref[para:adam1]{Adam}.\footnote{
			A nice animation of various optimizers (\hyperref[sc:SGD-momentum]{SGD}, \hyperref[sc:SGD-momentum]{SGD with momentum}, \hyperref[para:adagrad]{AdaGrad}, \hyperref[para:adadelta]{AdaDelta}, \hyperref[para:rmsprop]{RMSProp}) can be found in S. Ruder, `An overview of gradient descent optimization algorithms', updated on 2018.09.02 (\href{https://ruder.io/optimizing-gradient-descent/}{Original website}).
		} %{\color{red} HERE 2020.01.02}.
		
		See Figure~\ref{fig:Haibe-Kains-irreproducibility} in Section~\ref{sc:irreproducibility} on ``Lack of transparency and irreproducibility of results'' in recent deep-learning papers.
	} 
	$\hfill\blacksquare$
\end{rem}

For more insight into the update Eq.~(\ref{eq:SGD-momentum-3}), consider the case of constant coefficients $\mompar_k = \mompar$ and $\learn_k = \learn$, and rewrite this recursive relation in the form:
\begin{align}
	\bparamt_{k+1} - \bparamt_{k}
	= 
	- \learn
	\sum_{i=0}^{k}
	\mompar^i \bgradt_{k-i}
	\ , \text{ using }
	\bparamt_{1} - \bparamt_{0} = - \learn \bgradt_0
	\ ,
	\label{eq:SGD-momentum-explicit}
\end{align}
i.e., without momentum for the first term.  So the effective gradient is the sum of all gradients from the beginning $i = 0$ until the present $i = k$ weighted by the exponential function $\mompar^i$ so there is a fading memory effect, i.e., gradients that are farther back in time have less influence than those closer to the present time.\footnote{
	See also Section~\ref{sc:exponential-smoothing} on time series and exponential smoothing.
}   
The summation term in Eq.~(\ref{eq:SGD-momentum-explicit}) also provides an explanation of how the ``inertia'' (or momentum) term work: (1) Two successive opposite gradients would cancel each other, whereas (2) Two successive gradients in the same direction (toward the trough of the valley) would reinforce each other.  See also \cite{Bertsekas.1995}, pp.~104-105, and \cite{Hinton.2012} who provided a similar explanation:
\begin{quote}
	``Momentum is a simple method for increasing the speed of learning when the objective function contains long, narrow and fairly straight ravines with a gentle but consistent gradient along the floor of the ravine and much steeper gradients up the sides of the ravine. The momentum method simulates a heavy ball rolling down a surface. The ball builds up velocity along the floor of the ravine, but not across the ravine because the opposing gradients on opposite sides of the ravine cancel each other out over time.''
\end{quote}

%{\color{red} HERE 2019.12.30}

In recent years, Polyak (1964) \cite{Polyak.1964} (English version)\footnote{
	Polyak (1964) \cite{Polyak.1964}'s English version appeared before 1979, as cited
	%
	% CMES style rewriting 
%	by 
	\cite{Incerti.1979}, 
%	who used 
	where 
	a similar classical dynamics of a ``small heavy sphere'' or heavy point mass was used to develop an iterative method to solve nonlinear systems.  There, the name Polyak was spelled as ``Poljak'' as in the Russian version.  The earliest citing of the Russian version, with the spelling ``Poljak'' was in \cite{Ortega.1970} and in \cite{Voigt.1971}, but the terminology ``small heavy sphere'' was not used.  See also \cite{Bertsekas.1995}, p.~104 and p.~481, 
%	who cited 
	where
	the Russian version of \cite{Polyak.1964} was cited.
} has often been cited for the classical momentum (``small heavy sphere'') method to accelerate the convergence in gradient descent, but not so before, e.g., the authors of \cite{Rumelhart.1986} \vphantom{\cite{Plaut.1986}}\cite{Plaut.1986} \cite{Jacobs.1988} \cite{Hagiwara.1992} \cite{Hinton.2012} used the same method without citing \cite{Polyak.1964}.  Several books on optimization not related to neural networks, many of them well-known, also did not mention this method: \cite{Polak.1971} \vphantom{\cite{Gill.1981}}\cite{Gill.1981} \cite{Polak.1997} \cite{Nocedal.2006} \cite{Luenberger.2016} \cite{Snyman.2018}.  
%
% CMES style rewriting
%The book by \cite{Priddy.2005} on neural networks did cite 
Both the original Russian version and the English translated version \cite{Polyak.1964} (whose author's name was spelled as ``Poljak'' before 1990) were cited in the book on neural networks \cite{Priddy.2005},
%
% CMES style rewriting
%and referred to 
in which 
another neural-network book \cite{Bertsekas.1995}
%, in which the formulation was discussed.
was referred to for a discussion of the formulation.\footnote{
	See \cite{Priddy.2005}, p.~159, p.~115, and \cite{Bertsekas.1995}, p.~104, respectively.  The name ``Polyak'' was spelled as ``Poljak'' before 1990, \cite{Bertsekas.1995}, p.~481, and sometimes as ``Polyack'', \cite{Goudou.2009}.  See also \vphantom{\cite{Sutskever.2013}}\cite{Sutskever.2013}.
}  

\begin{rem}
	\label{rm:heavy-ball}
	Small heavy sphere, or heavy point mass, is better name.
	{\rm
		Because the rotatory motion is not considered in Eq.~(\ref{eq:SGD-momentum-2}), the name ``small heavy sphere'' given in \cite{Polyak.1964} is more precise than the more colloquial name ``heavy ball'' often given to the SGD with classical momentum,\footnote{
			See, e.g., \cite{Bertsekas.1995}, p.~104, \cite{Priddy.2005}, p.~115, \cite{Goudou.2009}, \cite{Sutskever.2013}.
		} since ``small'' implies that rotatory motion was neglected, and a ``heavy ball'' could be as big as a bowling ball\footnote{
			Or the ``Times Square Ball'', Wikipedia,  \href{https://en.wikipedia.org/w/index.php?title=Times_Square_Ball&oldid=932959767}{version 05:17, 29 December 2019}.
		} for which rotatory motion cannot be neglected.  For this reason, ``heavy point mass'' would be a precise alternative name.
	}
	$\hfill\blacksquare$
\end{rem}

\begin{rem}
	%Strongly convex functions.
	{\rm
		For Nesterov's fast (accelerated) gradient method, many references referred to \cite{Nesterov.1983}.\footnote{
			Reference \cite{Nesterov.1983} cannot be found from the Web of Science as of 2020.03.18, perhaps because it was in Russian, as indicated in Ref.~[35] in \cite{Nesterov.2018}, p.~582, where Nesterov's 2004 monograph was Ref.~[39].
		}  The authors of \cite{Goodfellow.2016}, p.~291, also referred to Nesterov's 2004 monograph, which was mentioned in the Preface of, and the material of which was included in, \cite{Nesterov.2018}. 
		For a special class of strongly convex functions,\footnote{
			A function $f(\cdot)$ is strongly convex if there is a constant $\mu > 0$ such that for any two points $x$ and $y$, we have $f(y) \ge f(x) + \langle \nabla f (x), y-x \rangle + \frac12 \mu \parallel y - x \parallel^2$, where $\langle \cdot , \cdot \rangle$ is the inner (or dot) product, \cite{Nesterov.2018}, p.~74. 
		} the step length can be kept constant, while the coefficients in Nesterov's fast gradient method varied, to achieve optimal performance, \cite{Nesterov.2018}, p.~92.
		``Unfortunately, in the stochastic gradient case, Nesterov momentum does not improve the rate of convergence'' \cite{Goodfellow.2016}, p.~292.  
	}
	$\hfill\blacksquare$
\end{rem}

\subsubsection{Initial-step-length tuning}
\label{sc:step-length-tuning}

The initial step length $\learn_0$, or learning-rate initial value, is one of the two most influential hyperparameters to tune, i.e., to find the best performing values.  During tuning, the step length $\learn$ is kept constant at $\learn_0$ in the parameter update Eq.~(\ref{eq:update-params}) throughout the optimization process, i.e., a fixed step length is used, without decay as in Eqs.~(\ref{eq:learning-rate-schedule}-\ref{eq:learning-rate-schedule-3}) in Section~\ref{sc:step-length-decay}.

%
% CMES style rewriting
%\vphantom{\cite{Wilson.2018}}\cite{Wilson.2018} proposed 
The following simple tuning method was proposed in \cite{Wilson.2018}: 
\begin{quote}
	``To tune the step sizes, we evaluated a logarithmically-spaced grid of five step sizes. If the best performance was ever at one of the extremes of the grid, we would try new grid points so that the best performance was contained in the middle of the parameters. For example, if we initially tried step sizes 2, 1, 0.5, 0.25, and 0.125 and found that 2 was the best performing, we would have tried the step size 4 to see if performance was improved. If performance improved, we would have tried 8
	and so on.''
\end{quote}
The above logarithmically-spaced grid was given by $2^k$, with $k=1, 0, -1, -2, -3$.  This tuning method appears effective as shown in Figure~\ref{fig:wilson-examples} on the CIFAR-10 dataset mentioned above, for which the following values for $\learn_0$ had been tried for different optimizers, even though the values did not always belong to the sequence $\{a^k\}$, but could include close, rounded values:
\begin{itemize}
	
	\item 
	\hyperref[sc:generic-SGD]{SGD} (Section~\ref{sc:generic-SGD}): {2, 1, 0.5 (best), 0.25, 0.05, 0.01}\footnote{
		The last two values $\{0.05, 0.01\}$ did not belong to the sequence $2^k$, with $k$ being integers, since $2^{-3} = 0.125$, $2^{-4} = 0.0625$ and $2^{-5} = 0.03125$.
	}
	
	\item 
	\hyperref[sc:SGD-momentum]{SGD with momentum} (Section~\ref{sc:SGD-momentum}): {2, 1, 0.5 (best), 0.25, 0.05, 0.01}
	
	\item 
	\hyperref[para:adagrad]{AdaGrad} (Section~\ref{sc:adaptive-learning-rate-algos}): {0.1, 0.05, 0.01 (best, default), 0.0075, 0.005}
	
	\item 
	\hyperref[para:rmsprop]{RMSProp} (Section~\ref{sc:adaptive-learning-rate-algos}): {0.005, 0.001, 0.0005, 0.0003 (best), 0.0001}
	
	\item 
	\hyperref[para:adam1]{Adam} (Section~\ref{sc:adaptive-learning-rate-algos}): {0.005, 0.001 (default), 0.0005, 0.0003 (best), 0.0001, 0.00005}
	
\end{itemize}

%step length decay
\subsubsection{Step-length decay, annealing and cyclic annealing}
\label{sc:step-length-decay}

In the update of the parameter $\bparam$  as in Eq.~(\ref{eq:update-params}), the learning rate (step length) $\epsilon$ has to be reduced gradually as a function of either the epoch counter $\tau$ or of the global iteration counter $j$.  Let $\tei$ represents either $\tau$ or $j$, depending on user's choice.\footnote{
	The Avant Garde font $\tei$ is used to avoid confusion with $t$, the time variable used in relation to recurrent neural networks; see Section~\ref{sc:recurrent} on ``Dynamics, sequential data, sequence modeling'', and Section~\ref{sc:dynamic-volterra-series} on ``Dynamics, time dependence, Volterra series''.  Many papers on deep-learning optimizers used $t$ as global iteration counter, which is denoted by $j$ here; see, e.g., \vphantom{\cite{Reddi.2019}}\cite{Reddi.2019}, \cite{Loshchilov.2019}.
}  If the learning rate $\learn$ is a function of epoch $\tau$, then $\learn$ is held constant in all iterations $k = 1,\ldots, k_{max}$ within epoch $\tau$, and we have the relation:
\begin{align}
j = (\tau - 1) * k_{max} + k
\ .
\label{eq:relation-tau-k}
\end{align} 

The following learning-rate scheduling, linear with respect to $\tei$, is one option:\footnote{
	\label{fn:learning-rate-schedule-1}
	See \cite{Goodfellow.2016}, p.~287, where it was suggested that $\tei_c$ in Eq.~(\ref{eq:learning-rate-schedule}) would be ``set to the number of iterations required to make a few hundred passes through the training set,'' and $\epsilon_{\tei_c}$ ``should be set to roughly 1 percent the value of $\epsilon_0$''.  A ``few hundred passes through the training set'' means a few hundred epochs; see Footnote~\ref{fn:learning-rate-schedule-2}.  In \cite{Goodfellow.2016}, p.~286, Algorithm 1 SGD for ``training iteration $k$'' should mean for ``training epoch $k$'', and the learning rate ``$\learn_k$'' would be held constant within ``epoch $k$''.
}
%{\color{blue}
%NOTE: 2019.12.03: the notion of an ``epoch'' was not quite clear to me in my first attempts on machine learning. similarly, your explanation as ``training session'' is hard to grasp. My understanding today is the following (though, I can't offer a reference right now): An ``epoch'' is completed once the entire ``full batch'' has been used to (iteratively) update the parameters during training. 
%}
%{\color{red} [NOTE: 2019.12.06.  actually, the notion of ``epoch'' or training session is clear from the pseudocodes, say Algorithm~\ref{algo:generic-SGD} for SGD for example.  whenever the \ding{173} For loop is done, by satisfying some stopping criterion, that epoch had ended.  this definition does not require that the minibatches had to cover the whole training set.   we probably need to put this definition in the paper.	ENDNOTE]
%}
\begin{align}
&
\epsilon (\tei) 
= 
\begin{cases}
\displaystyle
\left( 1 - \frac{\tei}{\tei_c} \right) \epsilon_0 
+ 
\frac{\tei}{\tei_c} \epsilon_{\tei_c}
&
\text{ for }
0 \le \tei \le \tei_c
\\
\epsilon_{\tei_c}
&
\text{ for }
\tei_c \le \tei
\end{cases}
\label{eq:learning-rate-schedule}
\\
&
\tei = \text{epoch } \tau \text{ or global iteration } j 
\label{eq:learning-rate-schedule-b}
\end{align}
where $\epsilon_0$ is the learning-rate initial value, and $\epsilon_{\tei_c}$ the constant learning-rate value when $\tei \ge \tei_c$.  Other possible learning-rate schedules are:\footnote{
	\label{fn:learning-rate-schedule-2}
	See \cite{Reddi.2019}, p.~3, below Algorithm 1 and just below the equation labeled ``(Sgd)''.  After, say, 400 global iterations, i.e., $\tei = j = 400$, then $\epsilon_{400} = 5\% \, \epsilon_0$ according to Eq.~(\ref{eq:learning-rate-schedule-2}), and $\epsilon_{400} = 0.25\% \, \epsilon_0$ according to Eq.~(\ref{eq:learning-rate-schedule-3}), whereas $\epsilon_{400} = 1\% \, \epsilon_0$ according to Eq.~(\ref{eq:learning-rate-schedule}).  See Footnote~\ref{fn:learning-rate-schedule-1}, and also
	Figure~\ref{fig:Haibe-Kains-irreproducibility} in Section~\ref{sc:irreproducibility} on ``Lack of transparency and irreproducibility of results'' in recent deep-learning papers.
}
\begin{align}
&
\epsilon(\tei) = \frac{\epsilon_0}{\sqrt{\tei}}
\rightarrow 0 \text{ as } \tei \rightarrow \infty
\ ,
\label{eq:learning-rate-schedule-2}
\\
&
\epsilon(\tei) = \frac{\epsilon_0}{\tei}
\rightarrow 0 \text{ as } \tei \rightarrow \infty
\ ,
\label{eq:learning-rate-schedule-3}
\end{align}
with $\tei$ defined as in Eq.~(\ref{eq:learning-rate-schedule}), even though authors such as \cite{Reddi.2019} and \cite{Phuong.2019} used Eq.~(\ref{eq:learning-rate-schedule-2}) and Eq.~(\ref{eq:learning-rate-schedule-3}) with $\tei = j$ as global iteration counter.

%{\color{red} HERE 2020.01.04} 
Another step-length decay method proposed in \cite{Wilson.2018} is to reduce the step length $\learn (\tau)$ for the current epoch $\tau$ by a factor $\decay \in (0,1)$ when the cost estimate $\losst_{\tau - 1}$ at the end of the last epoch $(\tau - 1)$ is greater than the lowest cost in all previous global iterations, with $\losst_j$ denoting the cost estimate at global iteration $j$, and $k_{max} (\tau - 1)$ the global iteration number at the end of epoch $(\tau -1)$:
\begin{align}
\learn (\tau)
=
\begin{cases}
\decay \, \learn (\tau - 1) & \text{ if } \losst_{\tau - 1} > \min\limits_j \{ \losst_j , j = 1 , \ldots , k_{max} (\tau - 1) \}
\\
\learn (\tau - 1) & \text{ Otherwise}
\end{cases}
\label{eq:learning-rate-schedule-4}
\end{align}
Recall, $k_{max}$ is the number of non-overlapping minibatches that cover the training set, as defined in Eq.~(\ref{eq:minibatch-0}). 
\cite{Wilson.2018} set the step-length decay parameter $\decay = 0.9$ in their numerical examples, in particular Figure~\ref{fig:wilson-examples}.

{\bf Cyclic annealing.} In additional to decaying the step length $\learn$, which is already annealing, cyclic annealing is introduced to further reduce the step length down to zero (``cooling''), quicker than decaying, then bring the step length back up rapidly (heating), and doing so for several cycles.  The cosine function is typically used, such as shown in Figure~\ref{fig:adamw-annealing}, as a multiplicative factor $\anneal_k \in [0, 1]$ to the step length $\learn_k$ in the parameter update, and thus the name ``cosine annealing'':
\begin{align}
	\bparamt_{k+1} = \bparamt_{k} - \anneal_k \learn_{k} \bgradt_{k}
	\ ,
	\label{eq:annealing-1}
\end{align}
as an add-on to the parameter update for vanilla SGD Eq.~(\ref{eq:update-params}), or
\begin{align}
	\bparamt_{k+1} = \bparamt_{k} 
	- \anneal_k \learn_k \bgradt (\bparamt_{k} + \nesterov_k (\bparamt_k - \bparamt_{k-1}) )
	+ \mompar_k (\bparamt_k - \bparamt_{k-1})
	\label{eq:annealing-2}
\end{align} 
as an add-on to the parameter update for SGD with momentum and accelerated gradient Eq.~(\ref{eq:SGD-update-momentum}).  The cosine  annealing factor can take the form \cite{Loshchilov.2019}:
\begin{align}
	\anneal_k = 0.5 + 0.5 \cos (\pi T_{cur} / T_p)
	\in [0,1]
	\ ,
	\text{ with }
	T_{cur} := j - \sum_{q=1}^{q=p-1} T_q
	\label{eq:cosine-annealing}
\end{align}
where $T_{cur}$ is the number of epochs from the start of the last warm restart at the end of epoch $\sum_{q=1}^{q=p-1} T_q$, where $\anneal_k = 1$ (``maximum heating''), $j$ the current global iteration counter,  $T_p$ the maximum number of epochs allowed for the current $p$th annealing cycle, during which $T_{cur}$ would go from $0$ to $T_p$, when $\anneal_k = 0$ (``complete cooling'').  Figure~\ref{fig:adamw-annealing} shows 4 annealing cycles, which helped reduce dramatically the number of epochs needed to achieve the same lower cost as obtained without annealing.

Figure~\ref{fig:adamw-examples} shows the effectiveness of cosine annealing in bringing down the cost rapidly in the early stage, but there is a diminishing return, as the cost reduction decreases with the number of annealing cycle.  Up to a point, it is no longer as effective as SGD with weight decay in Section~\ref{sc:weight-decay}.

% {\color{red} HERE 2020.01.25}

{\bf Convergence conditions.}
The sufficient conditions for convergence, for convex functions, are\footnote{
	Eq.~(\ref{eq:SGD-convergence-conditions}) are called the ``stepsize requirements'' in \cite{Bottou.2018:rd0001}, and ``sufficient condition for convergence'' in \cite{Goodfellow.2016}, p.~287, and in \cite{Li.2019}.
	%
	% CMES style rewriting
	Robbins \& Monro (1951b)
	\cite{Robbins1951b} were concerned with solving $M(x) = \alpha$ when the function $M(\cdot)$ is not known, but the distribution of the output, as a random variable, $\yup = \yup(x)$ is assumed known.  For the network training problem at hand, one can think of $M(x) = \parallel \nabla J(x) \parallel$, i.e., the magnitude of the gradient of the cost function $J$ at $x = \parallel \bparamt - \bparam \parallel$, the distance from a local minimizer, and $\alpha = 0$, i.e., the stationarity point of $J(\cdot)$.  In \cite{Robbins1951b}---in which there was no notion of ``epoch $\tau$'' but only global iteration counter $j$---Eq.~(6) on p.~401 corresponds to Eq.~(\ref{eq:SGD-convergence-conditions})$_1$ (first part), and Eq.~(27) on p.~404 corresponds to Eq.~(\ref{eq:SGD-convergence-conditions})$_2$ (second part).  Any sequence $\{ \epsilon_{k} \}$ that satisfied Eq.~(\ref{eq:SGD-convergence-conditions}) was called a sequence of type $1/k$; the convergence Theorem 1 on p.~404 and Theorem 2 on p.~405 indicated that the sequence of step length $\{ \epsilon_{k} \}$ being of type $1/k$ was only one among other sufficient conditions for convergence.  In Theorem 2 of \cite{Robbins1951b}, the additional sufficient conditions were Eq.~(33), $M(x) = \parallel \nabla J(x) \parallel$ non decreasing, Eq.~(34), $M(0) = \parallel \nabla J(0) \parallel = 0$, and Eq.~(35), $M^\prime(0) = \parallel \nabla^2 J(0) \parallel > 0$, i.e., the iterates $\{x_k \, | \, k=1,2,\ldots\}$, fell into a local convex bowl. 
}
\begin{align}
	\sum_{j=1}^{\infty} \epsilon_{j}^2 < \infty
	\ ,
	\text{ and }
	\sum_{j=1}^{\infty} \epsilon_{j} = \infty
	\ .
	\label{eq:SGD-convergence-conditions}
\end{align}
The inequality on the left of Eq.~(\ref{eq:SGD-convergence-conditions}), i.e., the sum of the squared of the step lengths being finite, ensures that the step length would decay quickly to reach the minimum, but is valid only when the minibatch size is fixed.  The equation on the right of Eq.~(\ref{eq:SGD-convergence-conditions}) ensures convergence, no matter how far the initial guess was from the minimum \vphantom{\cite{SmithSL.2018b}}\cite{SmithSL.2018b}.
 
In Section~\ref{sc:minibatch-size-increase}, the step-length decay is shown to be equivalent to minibatch-size increase and simulated annealing in the sense that there would be less fluctuation, and thus lower ``temperature'' (cooling) by analogy to the physics governed by the Langevin stochastic differential equation and its discrete version, which is analogous to the network parameter update.

\subsubsection{Minibatch-size increase, fixed step length, equivalent annealing}
\label{sc:minibatch-size-increase}

The minibatch parameter update from Eq.~(\ref{eq:SGD-update-momentum}), without momentum and accelerated gradient, which becomes Eq.~(\ref{eq:update-params}), can be rewritten to introduce the error due to the use of the minibatch gradient estimate $\bgradt$ instead of the full-batch gradient $\bgrad$ as follows:
\begin{align}
	\bparamt_{k+1} 
	&
	=
	\bparamt_{k} - \learn_{k} \bgradt_k
	\tag{\ref{eq:update-params}}
	\\
	&
	= 
	\bparamt_{k} - \learn_{k} 
	\left[ 
		\bgrad_k 
		+ 
		\left( \bgradt_k - \bgrad_k \right)
	\right]
	\Rightarrow
	\frac{\Delta \bparamt_{k}}{\learn_{k}}	
	=
	\frac{\bparamt_{k+1} - \bparamt_{k}}{\learn_{k}}
	=
	- \bgrad_{k}
	+ \left( \bgrad_k - \bgradt_k \right)
	=
	- \bgrad_{k} + \graderr_k
	\ ,
	\label{eq:update-params-2}
\end{align}
where $\bgrad_{k} = \bgrad (\bparam_{k})$ and $\bgradt_{k} = \bgradt_{b} (\bparam_{k})$, with $b = k$, and $\bgradt_b (\cdot)$ is the gradient estimate function using minibatch $b = 1, \ldots, k_{max}$.

To show that the gradient error has zero mean (average), based on the linearity of the expectation function $\xpc (\cdot) = \langle \cdot \rangle$ defined in Eq.~(\ref{eq:expectation}) (Footnote~\ref{fn:expectation-notation}), i.e., 
\begin{align}
	\langle \alpha \boldsymbol{u} + \beta \boldsymbol{v} \rangle
	=
	\alpha
	\langle \boldsymbol{u} \rangle
	+
	\beta
	\langle \boldsymbol{u} \rangle
	\ ,
	\label{eq:expectation-linearity}
 \end{align}
%we have
\begin{align}
	\langle \graderr_k \rangle
	=
	\langle \bgrad_k - \bgradt_k \rangle
	=
	\langle \bgrad_k \rangle
	-
	\langle \bgradt_k \rangle
	=
	\bgrad_k - \langle \bgradt_k \rangle
	%= 0
	\ ,
	\label{eq:gradient-error-zero-mean-1}
\end{align}
from Eqs.~(\ref{eq:minibatch-0})-(\ref{eq:minibatch-2}) on the definition of minibatches and Eqs.~(\ref{eq:cost-estimate})-(\ref{eq:gradient-estimate}) on the definition of the cost and gradient estimates (without omitting the iteration counter $k$), we have
\begin{align}
	\bgrad_{k} = \bgrad (\bparam_{k})
	&
	=
	\frac{1}{k_{max}} 
	\sum_{b=1}^{b=k_{max}}
	\frac{1}{\bsize}
	\sum_{a=1}^{a=\bsize}
	\bgrad_{i_{a,b}}
	\Rightarrow
	\langle \bgrad_k \rangle
	=
	\bgrad_{k}
	=
	\frac{1}{k_{max}} 
	\sum_{b=1}^{b=k_{max}}
	\frac{1}{\bsize}
	\sum_{a=1}^{a=\bsize}
	\langle \bgrad_{i_{a,b}} \rangle
	=
	\langle \bgrad_{i_{a,b}} \rangle 
	\\
	\bgradt_k 
	&
	=
	\frac{1}{\bsize} \sum_{a=1}^{a=\bsize \le \Bsize}
	\bgrad_{i_{a,k}}
	\Rightarrow
	\langle \bgradt_k \rangle
	=
	\frac{1}{k_{max}} \sum_{k=1}^{k=k_{max}}
	\langle \bgrad_{i_{a,k}} \rangle
 	=
	\frac{1}{k_{max}} \sum_{k=1}^{k=k_{max}}
	\langle \bgrad_k \rangle
	=
	\bgrad_{k}
	\Rightarrow
	\langle \graderr_k \rangle = 0
	\ .
	\label{eq:gradient-error-zero-mean-2}
\end{align}
Or alternatively, the same result can be obtained with:
\begin{align}
	\bgrad_k
	&
	=
	\frac{1}{k_{max}}
	\sum_{b=1}^{b=k_{max}}
	\bgradt_{b} (\bparam_{k})
	\Rightarrow
	%\langle \bgrad_{k} \rangle
	%=
	\bgrad_{k}
	%=
	%\frac{1}{k_{max}}
	%\sum_{b=1}^{b=k_{max}}
	%\langle \bgradt_{b} \rangle
	=
	\langle \bgradt_{b} (\bparam_{k}) \rangle
	=
	\langle \bgradt_{k} (\bparam_{k}) \rangle
	=
	%\ ,
	%\\
	%\Rightarrow
	\langle \bgradt_{k} \rangle
	%&
	%=
	%\frac{1}{\bsize}
	%\sum_{a=1}^{a=\bsize}
	%\langle \bgrad_{i_{a,k}} \rangle
	%=
	%\bgrad_{k}
	%\Rightarrow
	%\langle \graderr_k \rangle = 0
	\Rightarrow
	\langle \graderr_k \rangle = 0
	\ .
	\label{eq:gradient-error-zero-mean-3}
\end{align}

Next, the mean value of the ``square'' of the gradient error, i.e., $\langle \graderr^T \graderr \rangle$, in which we omitted the iteration counter subscript $k$ to alleviate the notation, relies on some identities related to the covariance matrix $\langle \graderr , \graderr \rangle$.  
The mean of the square matrix $\bxup_i^T \bxup_j$, where $\{\bxup_i , \bxup_j\}$ are two random \emph{row} matrices, is the sum of the product of the mean values and the covariance matrix of these matrices\footnote{
	See, e.g., \cite{Gardiner.2004}, p.~36, Eq.~(2.8.3).
}
\begin{align}
	\langle \bxup_i^T \bxup_j \rangle
	=
	\langle \bxup_i \rangle^T
	\langle \bxup_j \rangle
	+
	\langle \bxup_i , \bxup_j \rangle
	\ ,
	\text{ or }
	\langle \bxup_i , \bxup_j \rangle
	=
	\langle \bxup_i^T \bxup_j \rangle 
	-
	\langle \bxup_i \rangle^T
	\langle \bxup_j \rangle
	\ ,
	\label{eq:mean-x-x^T}
\end{align}
where $\langle \bxup_i , \bxup_j \rangle$ is the covariance matrix of $\bxup_i$ and $\bxup_j$, and thus the covariance operator $\langle \cdot , \cdot \rangle$ is bilinear due to the linearity of the mean (expectation) operator $\langle \cdot \rangle$ in Eq.~(\ref{eq:expectation-linearity}):
\begin{align}
	\left\langle 
		\sum_i \alpha_i \boldsymbol{u}_i , \sum_j \beta_j \boldsymbol{v}_j
	\right\rangle
	=
	\sum_i \alpha_i \beta_j 
	\langle \boldsymbol{u}_i , \boldsymbol{v}_j \rangle
	\ ,
	\forall 
	\alpha_i , \beta_j \in \real
	\text{ and } 
	\forall
	\boldsymbol{u}_i , \boldsymbol{v}_j \in \real^n 
	\text{ random}
	\ .
	\label{eq:covariance-matrix-bilinear}
\end{align}
Eq.~(\ref{eq:covariance-matrix-bilinear}) is the key relation to derive an expression for the square of the gradient error $\langle \graderr^T \graderr \rangle$, which can be rewritten as the sum of four covariance matrices upon using Eq.~(\ref{eq:mean-x-x^T})$_1$ and either Eq.~(\ref{eq:gradient-error-zero-mean-2}) or Eq.~(\ref{eq:gradient-error-zero-mean-3}), i.e., $\langle \bgradt_{k} \rangle = \langle \bgrad_{k} \rangle = \bgrad_{k}$, as the four terms $\bgrad_k^T \bgrad_k$ cancel each other out:
\begin{align}
	\langle \graderr^T \graderr \rangle
	=
	\langle (\bgradt - \bgrad)^T (\bgradt - \bgrad) \rangle
	=
	\langle \bgradt , \bgradt \rangle
	-
	\langle \bgradt , \bgrad \rangle
	-
	\langle \bgrad , \bgradt \rangle
	+
	\langle \bgrad , \bgrad \rangle
	\ ,
	\label{eq:gradient-error-squared-1}
\end{align} 
where the iteration counter $k$ had been omitted to alleviate the notation.  Moreover, to simplify the notation further, the gradient related to an example is simply denoted by $\bgrad_{a}$ or $\bgrad_{b}$, with $a, b = 1,\ldots, \bsize$ for a minibatch, and $a, b = 1,\ldots, \Bsize$ for the full batch:
\begin{align}
	\bgradt = \bgradt_{k}
	& 
	=
	\frac{1}{\bsize}
	\sum_{a=1}^{\bsize}
	\bgrad_{i_{a},k}
	=
	\frac{1}{\bsize}
	\sum_{a=1}^{\bsize}
	\bgrad_{i_{a}}
	=
	\frac{1}{\bsize}
	\sum_{a=1}^{\bsize}
	\bgrad_{a}
	\ ,
	\label{eq:gradient-decomp-minibatch}
	\\
	\bgrad
	&
	=
	\frac{1}{k_{max}}
	\sum_{k=1}^{k_{max}}
	\frac{1}{\bsize}
	\sum_{a=1}^{\bsize}
	\bgrad_{i_{a},k}
	=
	\frac{1}{\Bsize}
	\sum_{b=1}^{\Bsize}
	\bgrad_{b}
	\ .
	\label{eq:gradient-decomp-full}
\end{align}
Now assume the covariance matrix of any pair of single-example gradients $\bgrad_{a}$ and $\bgrad_{b}$ depends only on the parameters $\bparam$, and is of the form:
\begin{align}
	\langle \bgrad_{a} , \bgrad_{b}\rangle 
	= 
	\covmat (\bparam) \delta_{ab}
	\ , \forall a, b \in \{1 , \ldots , \Bsize\}
	\ ,
	\label{eq:covariance-matrix}
\end{align}
where $\delta_{ab}$ is the Kronecker delta.
Using Eqs.~(\ref{eq:gradient-decomp-minibatch})-(\ref{eq:gradient-decomp-full}) and Eq.~(\ref{eq:covariance-matrix}) in Eq.~(\ref{eq:gradient-error-squared-1}), we obtain a simple expression for $\langle \graderr^T \graderr \rangle$:\footnote{
	Eq.~(\ref{eq:gradient-error-squared-2}) possesses a simplicity elegance compared to the expression $\langle \alpha^2 \rangle = N (N/B - 1) F (\omega)$ in \cite{SmithSL.2018b}, based on a different definition of gradient, with $N \equiv \Bsize$, $B \equiv \bsize$, $\omega \equiv \bparam$, but $F(\omega) \ne \covmat (\bparam)$.
}
\begin{align}
	\langle \graderr^T \graderr \rangle
	=
	\left( \frac{1}{\bsize} - \frac{1}{\Bsize} \right) \covmat
	\ .
	\label{eq:gradient-error-squared-2}
\end{align}

%{\color{red} HERE 2020.01.22}

%
% CMES style rewriting
The authors of \cite{SmithSL.2018b} introduced the following stochastic differential equation as a continuous counterpart of the discrete parameter update Eq.~(\ref{eq:update-params-2}), as $\learn_{k} \rightarrow 0$:
\begin{align}
	\frac{d \bparam}{dt}
	=
	- \bgrad + \noise (t)
	=
	-
	\frac{dJ}{d \bparam}
	+
	\noise (t)
	\ ,
	\label{eq:SGD-stochastic-diff-eq}
\end{align}
where $\noise (t)$ is the noise function, the continuous counterpart of the gradient error $\graderr_k := \left( \bgrad_k - \bgradt_k \right)$.  The noise $\noise (t)$ is assumed to be Gaussian, i.e., with zero expectation (mean) and with covariance function of the form (see Remark~\ref{rm:annealing} on Langevin stochastic differential equation): 
\begin{align}
	\langle \noise (t) \rangle = 0
	\ ,
	\text{ and }
	\langle \noise (t) \noise (t^\prime) \rangle 
	= 
	\fluct \covmat (\bparam) \delta (t - t^\prime])
	\ ,
	\label{eq:SGD-stochastic-diff-eq-2}
\end{align}
where $\xpc [\cdot] = \langle \cdot \rangle$ is the expectation of a function, $\fluct$ the ``noise scale'' or fluctuation factor, $\covmat (\bparam)$ the same gradient-error covariance matrix in Eq.~(\ref{eq:covariance-matrix}), and $\delta (t - t^\prime)$ the Dirac delta.   Integrating Eq.~(\ref{eq:SGD-stochastic-diff-eq}), we obtain:
\begin{align}
	\int\limits_{t=0}^{t=\learn_k} \frac{d \bparam_k}{dt} dt 
	=
	\bparam_{k+1} - \bparam_{k}
	=
	- \learn_{k} \bgrad_{k} 
	+
	\int\limits_{t=0}^{t=\learn_k} \noise(t) dt 
	\ ,
	\text{ and }
	\left\langle 
	\int\limits_{t=0}^{t=\learn_k} \noise(t) dt 
	\right\rangle 
	=
	\int\limits_{t=0}^{t=\learn_k} \langle \noise(t) \rangle dt = 0 
	\ .
	\label{eq:update-params-5}
\end{align} 
The fluctuation factor $\fluct$ can be identified by equating the square of the error in Eq.~(\ref{eq:update-params-2}) to that in Eq.~(\ref{eq:update-params-5}), i.e.,
\begin{align}
	\learn^2 \langle \graderr^T \graderr \rangle
	=
	\int\limits_{t=0}^{t=\learn} 
	\int\limits_{t^\prime=0}^{t^\prime=\learn}
	\langle \noise(t) \noise(t^\prime) \rangle
	dt^\prime
	dt
	\Rightarrow
	\learn^2 \left( \frac{1}{\bsize} - \frac{1}{\Bsize} \right) \covmat
	=
	\learn \fluct \covmat
	\Rightarrow
	\fluct = \learn  \left( \frac{1}{\bsize} - \frac{1}{\Bsize} \right)
	\ ,
	\label{eq:fluctuation-2}
\end{align}

\begin{figure}[h]
	\centering
	\includegraphics[width=0.9\linewidth]{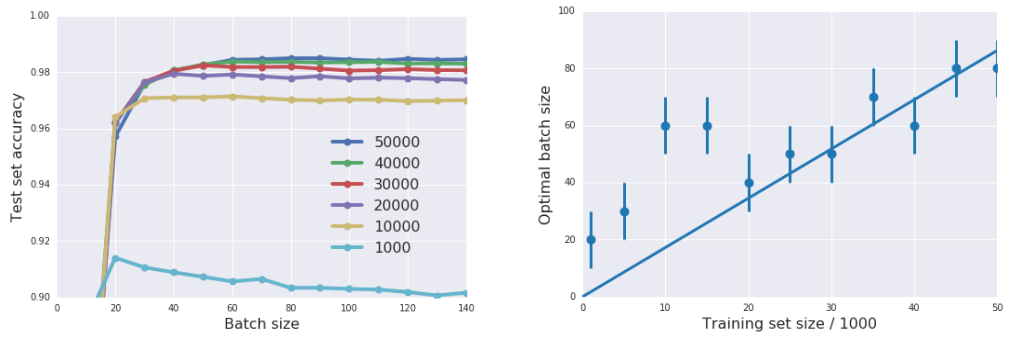}
	\caption{ 
		\emph{Optimal minibatch size vs. training-set size} (Section~\ref{sc:minibatch-size-increase}).  For a given training-set size, the smallest minibatch size that achieves the highest accuracy is optimal. 
		Left figure: The optimal mimibatch size was moving to the right with increasing training-set size $\Bsize$.
		Right figure:
		%
		% CMES style rewriting 
%		For \cite{SmithSL.2018a}, 
		The optimal minibatch size in \cite{SmithSL.2018a} is linearly proportional to the training-set size $\Bsize$ for large training sets (i.e., $\Bsize \rightarrow \infty$), Eq.~(\ref{eq:fluctuation-Smith-2018}), but our fluctuation factor $\fluct$ is independent of $\Bsize$ when $\Bsize \rightarrow \infty$; see Remark~\ref{rm:minibatch-size-training-size}.  
		{\footnotesize (Figure reproduced with permission of the authors.)}
%		{\color{red} ASK PERMISSION 2020.03.09}
	}
	\label{fig:minibatch-size-training-size}
\end{figure}

\begin{rem}
	\label{rm:minibatch-size-training-size}
	{Fluctuation factor for large training set.}
	{\rm
		For large $\Bsize$, our fluctuation factor $\fluct$ is roughly proportional to the ratio of the step length over the minibatch size, i.e., $\fluct \approx \learn / \bsize$.  Thus step-length $\learn$ decay, or equivalenly minibatch size $\bsize$ increase, corresponds to a decrease in the fluctuation factor $\fluct$.  On the other hand, \cite{SmithSL.2018b} obtained their fluctuation factor $\fluctS$ as\footnote{
			In \cite{SmithSL.2018a} and \cite{SmithSL.2018b}, the fluctuation factor was expressed, in original notation, as $g = \learn (N/B - 1)$, where the equivalence with our notation is $g \equiv \fluctS$ (fluctuation factor), $N \equiv \Bsize$ (training-set size), $B \equiv \bsize$ (minibatch size).
		} 
		\begin{align}
			\fluctS 
			=
			\learn 
			\left( \frac{\Bsize}{\bsize} - 1 \right) 
			= 
			\learn
			\Bsize 
			\left( \frac{1}{\bsize} - \frac{1}{\Bsize} \right)
			= 
			\Bsize \fluct
			\label{eq:fluctuation-Smith-2018}
		\end{align}
		since their cost function was not an average, i.e., not divided by the minibatch size $\bsize$, unlike our cost function in Eq.~(\ref{eq:cost-estimate}). When $\Bsize \rightarrow \infty$, our fluctuation factor $\fluct \rightarrow \learn / \bsize$ in Eq.~(\ref{eq:fluctuation-2}), but their fluctuation factor $\fluctS \approx \learn \Bsize / \bsize \rightarrow \infty$, i.e., for increasingly large $\Bsize$, our fluctuation factor $\fluct$ is bounded, but not their fluctuation factor $\fluctS$. 
		\cite{SmithSL.2018a} then went on to show empirically that their fluctation factor $\fluctS$ was proportional to the training-set size $\Bsize$ for large $\Bsize$, as shown in Figure~\ref{fig:minibatch-size-training-size}.
		On the other hand, our fluctuation factor $\fluct$ does not depend on the training set size $\Bsize$.  As a result, unlike \cite{SmithSL.2018a} in Figure~\ref{fig:minibatch-size-training-size}, our optimal minibatch size would not depend of the training-set size $\Bsize$.
	}
	$\hfill\blacksquare$
\end{rem}

\begin{figure}[h]
	\centering
	\includegraphics[width=0.9\linewidth]{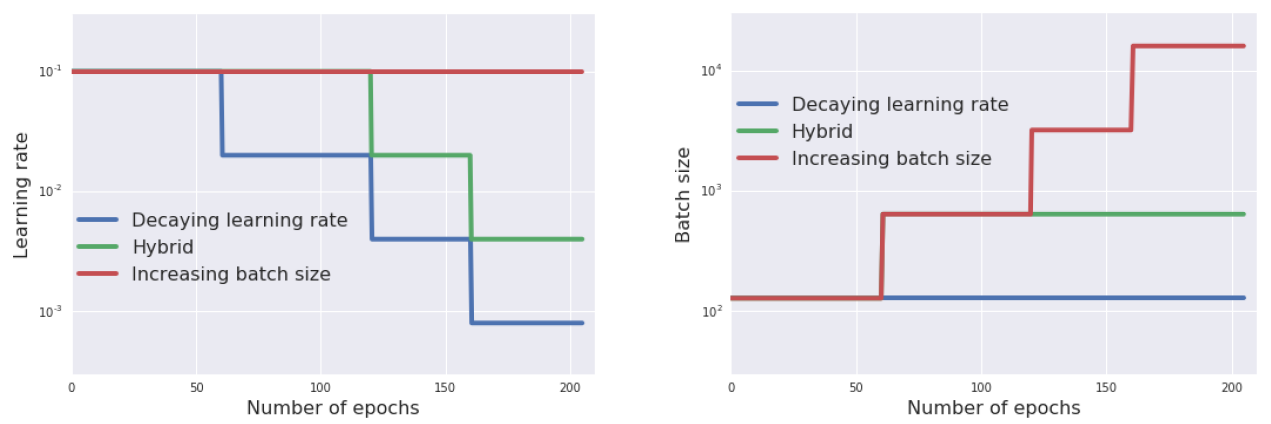}
	\caption{ 
		\emph{Minibatch-size increase vs. step-length decay, training schedules} (Section~\ref{sc:minibatch-size-increase}).
		Left figure: Step length (learning rate) vs. number of epochs.
		Right figure: Minibatch size vs. number of epochs.
%		Three training schedules were used:
		Three learning-rate schedules\protect\footnotemark $\,$were used for training: 
		(1) The step length was decayed by a factor of 5, from an initial value of $10^{-1}$, at specific epochs (60, 120, 160), while the minibatch size was kept constant (blue line);
		(2) Hybrid, i.e., the step length was initially kept constant until epoch 120, then decreased by a factor of 5 at epoch 120, and by another factor of 5 at epoch 160 (green line);
		(3) The step length was kept constant, while the minibatch size was increased by a factor of 5, from an initial value of 128, at the same specific epochs, 60, 120, 160 (red line).
		See Figure~\ref{fig:minibatch-increase-results} for the results using the CIFAR-10 dataset
		\cite{SmithSL.2018b}
		{\footnotesize (Figure reproduced with permission of the authors.)}
%		{\color{red} ASK PERMISSION 2020.03.07}
	}
	\label{fig:minibatch-increase-training-schedules}
\end{figure}

%
% CMES style rewriting
%\cite{SmithSL.2018b} suggested 
It was suggested in \cite{SmithSL.2018b}
to follow the same step-length decay schedules\footnote{
	See Figure~\ref{fig:Haibe-Kains-irreproducibility} in Section~\ref{sc:irreproducibility} on ``Lack of transparency and irreproducibility of results'' in recent deep-learning papers.
} $\learn (\tei)$ in Section~\ref{sc:step-length-decay} to adjust the size of the minibatches, while keeping the step length constant at its initial value $\learn_0$.
To demonstrate the equivalence between decreasing the step length and increasing minibatch size, 
%
% CMES style rewriting
%\cite{SmithSL.2018b} used 
the CIFAR-10 dataset with three different training schedules as shown in Figure~\ref{fig:minibatch-increase-training-schedules} was used in \cite{SmithSL.2018b}.

\footnotetext{
	See Figure~\ref{fig:Haibe-Kains-irreproducibility} in Section~\ref{sc:irreproducibility} on ``Lack of transparency and irreproducibility of results'' in recent deep-learning papers.
}

\begin{figure}[h]
	\centering
	\includegraphics[width=0.9\linewidth]{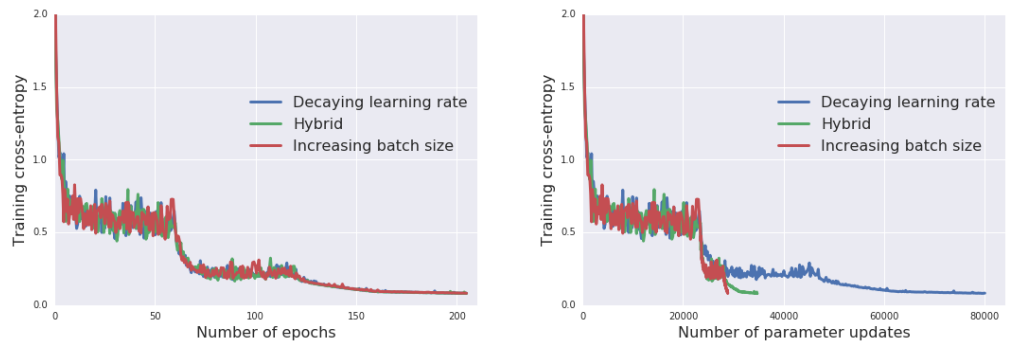}
	\caption{ 
		\emph{Minibatch-size increase, fewer parameter updates, faster comutation}  (Section~\ref{sc:minibatch-size-increase}).  
		For each of the three training schedules in Figure~\ref{fig:minibatch-increase-training-schedules}, the same learning curve is plotted in terms of the number of epochs  (left figure), and again in terms of the number of parameter updates (right figure), which shows the significant decrease in the number of parameter updates, and thus computational cost, for the training schedule with minibatch-size decrease.   The blue curve ends at about 80,000 parameter updates for step-length decrease, whereas the red curve ends at about 29,000 parameter updates for minibatch-size decrease
		\cite{SmithSL.2018b}
		{\footnotesize (Figure reproduced with permission of the authors.)}
%		{\color{red} ASK PERMISSION 2020.03.07}
	}
	\label{fig:minibatch-increase-results}
\end{figure}

\noindent
The results are shown in Figure~\ref{fig:minibatch-increase-results}, where it was shown that the number of updates decreased drastically with minibatch-size increase, allowing for significantly shortening the training wall-clock time.

%{\color{red} HERE 2020.01.23}

\begin{rem}
	\label{rm:annealing}
	Langevin stochastic differential equation, annealing.
	{\rm
		Because the fluctuation factor $\fluct$ was proportional to the step length, and in physics, fluctuation decreases with temperature (cooling), 
		``decaying the learning rate (step length) is simulated annealing''\footnote{
			\label{fn:annealing}
			``In metallurgy and materials science, annealing is a heat treatment that alters the physical and sometimes chemical properties of a material to increase its ductility and reduce its hardness, making it more workable. It involves heating a material above its recrystallization temperature, maintaining a suitable temperature for a suitable amount of time, and then allow slow cooling.''  Wikepedia, `Annealing (metallurgy)', Version \href{https://en.wikipedia.org/w/index.php?title=Annealing_(metallurgy)&oldid=928035952}{11:06, 26 November 2019}.  
			The name ``simulated annealing'' came from the highly cited paper \vphantom{\cite{Kirkpatrick.1983}}\cite{Kirkpatrick.1983}, which received more than 20,000 citations on Web of Science and more than 40,000 citations on Google Scholar as of 2020.01.17.  See also Remark~\ref{rm:metaheuristics} on ``Metaheuristics''.
		}
	    %
	    % CMES style rewriting 
%	    wrote 
	    \cite{SmithSL.2018b}.
		Here, we will connect the step length to ``temperature'' based on the analogy of Eq.~(\ref{eq:SGD-stochastic-diff-eq}), the continuous counterpart of the parameter update Eq.~(\ref{eq:update-params-2}).  In particular, we point exact references that justify the assumptions in Eq.~(\ref{eq:SGD-stochastic-diff-eq-2}). 
		
		Even though 
		%
		% CMES style rewriting
		the authors of
		\cite{SmithSL.2018b} referred to \vphantom{\cite{Li.2017}}\cite{Li.2017} for Eq.~(\ref{eq:SGD-stochastic-diff-eq}), the decomposition of the parameter update in \cite{Li.2017}:
		\begin{align}
			\bparamt_{k+1} 
			&
			= 
			\bparamt_{k} 
			- \learn_{k} \bgrad_{k} 
			+ \sqrt{\learn_{k}} \graderrt_k
			\ , 
			\text{ with }
			\graderrt_k 
			:= 
			\sqrt{\learn_{k}} \left[ \bgrad_{k} - \bgradt_{k} \right] 
			= 
			\sqrt{\learn_{k}} \graderr_k
			\ ,
			\label{eq:update-params-3}
		\end{align}
		with the intriguing factor $\sqrt{\learn_{k}}$ was consistent with the equivalent expression in \cite{Gardiner.2004}, p.~53, Eq.~(3.5.10),\footnote{
			In original notation used in \cite{Gardiner.2004}, p.~53, Eq.~(3.5.10) reads as $\by (t + \Delta t) = \by (t) + \boldsymbol{A} (\by(t), t) \Delta t + \bnSDE (t) \sqrt{\Delta t}$, in which the noise $\bnSDE (t)$ has zero mean, i.e., $\langle \bnSDE \rangle = 0$, and covariance matrix $\langle \bnSDE (t) , \bnSDE (t) \rangle = \boldsymbol{B} (t)$.
		} which was obtained from the Fokker-Planck equation:
		\begin{align}
			% CMES style - 2022.06.01
			% \bparam (t + \Delta t) = \bparam (t) + \boldsymbol{A (\bparam(t), t)} \Delta (t) + \sqrt{\Delta t} \bnSDE (t)
			% cannot put \bparam as argument inside \boldsymbol, so correct as shown below
			\bparam (t + \Delta t) = \bparam (t) + \boldsymbol{A} (\bparam(t), t) \Delta (t) + \sqrt{\Delta t} \bnSDE (t)
			\ , 
			\label{eq:SDE-discrete} 
		\end{align}
		where $\boldsymbol{A} (\bparam (t), t)$ is a nonlinear operator.  The noise term $\sqrt{\Delta t} \bnSDE (t)$ is \emph{not} related to the gradient error as in Eq.~(\ref{eq:update-params-3}), and is Gaussian with zero mean and covariance matrix of the form:
		\begin{align}
			\sqrt{\Delta t} \langle \bnSDE (t)  \rangle 
			= 0
			\ ,
			\text{ and }
			\Delta t
			\langle \bnSDE (t) , \bnSDE (t) \rangle 
			= \Delta t \cmSDE (t)
			\ .
			\label{eq:SDE-discrete-2}
		\end{align}
		The column matrix (or vector) $\boldsymbol{A} (\bparam (t), t)$ in Eq.~(\ref{eq:SDE-discrete}) is called the \emph{drift vector}, and the square matrix $\cmSDE$ in Eq.~(\ref{eq:SDE-discrete-2}) the \emph{diffusion matrix}, \cite{Gardiner.2004}, p.~52.
		Eq.~(\ref{eq:SDE-discrete}) implies that $\bparam ( \cdot )$ is a continuous function, called the ``sample path''. %\cite{Gardiner.2004}, p.~53.
		
		To obtain a differential equation, Eq.~(\ref{eq:SDE-discrete}) can be rewritten as
		\begin{align}
			\frac{\bparam (t + \Delta t) - \bparam (t)}{\Delta t}
			=
			\boldsymbol{A} (\bparam(t), t)
			+
			\frac{\bnSDE (t)}{\sqrt{\Delta t}}
			\ ,
			\label{eq:SDE-discrete-3}
		\end{align}
		which shows that the derivative of $\bparam(t)$ does not exist when taking the limit as $\Delta t \rightarrow 0$, not only due to the factor $1 / \sqrt{\Delta t} \rightarrow \infty$, but also due to the noise $\bnSDE (t)$, \cite{Gardiner.2004}, p.~53.  
		
		The last term $\bnSDE / \sqrt{\Delta t}$ in Eq.~(\ref{eq:SDE-discrete-3}) corresponds to the random force $X$ exerted on a pollen particle by the viscous fluid molecules in the 1-D equation of motion of the pollen particle, as derived by Langevin and in his original notation, \cite{Lemons.1997}:\footnote{
			See also ``Langevin equation'', Wikipedia,  \href{https://en.wikipedia.org/w/index.php?title=Langevin_equation&oldid=929100467}{version 17:40, 3 December 2019}.
		}
		\begin{align}
			m \frac{d^2 x}{dt^2} = - 6 \pi \mu a \frac{dx}{dt} + X (t)
			\Rightarrow
			m \frac{dv}{dt} = - \friction v + X (t)
			\ ,
			\text{ with }
			v = \frac{dx}{dt}
			\text{ and }
			\friction = 6 \pi \mu a
			\ ,
			\label{eq:SDE-langevin-1}
		\end{align}
		where $m$ is the mass of the pollen particle, $x(t)$ its displacement, $\mu$ the fluid viscosity, $a$ the particle radius, $v(t)$ the particle velocity, and $\friction$ the friction coefficient between the particle and the fluid.  
		The random (noise) force $X(t)$ by the fluid molecules impacting the pollen particle is assumed to (1) be independent of the position $x$, (2) vary extremely rapidly compared to the change of $x$, (3) have zero mean as in Eq.~(\ref{eq:SDE-discrete-2}).  The covariance of this noise force $X$ is proportional to the absolute temperature $T$, and takes the form, \vphantom{\cite{Coffey.2004}}\cite{Coffey.2004}, p.~12,\footnote{
			For first-time learners, here a guide for further reading on a derivation of Eq.~(\ref{eq:covariance-noise-X}).  It is better to follow the book \cite{Coffey.2004}, rather than Coffey's 1985 long review paper, cited for equation $\overline{X(t) X(t^\prime)} = 2 \friction k T \delta (t - t^\prime)$ on p.~12, not exactly the same left-hand side as in Eq.~(\ref{eq:covariance-noise-X}), and for which the derivation appeared some 50 pages later in the book.  The factor $2 \friction k T$ was called the \emph{spectral density}.
			The time average $\overline{X(t) X(t^\prime)}$ was defined, and the name \emph{autocorrelation function} introduced on p.~13.  But a particular case of the more general autocorrelation function is $\langle X(t) X(t^\prime) \rangle$ was defined on p.~59, and by the ergodic theorem, $\overline{X(t) X(t^\prime)} = \langle X(t) X(t^\prime) \rangle$ for stationary processes, p.~60, where the spectral density of $X(t)$, denoted by $\Phi_X (\omega)$ was defined as the Fourier transform of the autocorrelation function $\langle X(t) X(t^\prime) \rangle$.  The derivation of Eq.~(\ref{eq:covariance-noise-X}) started from the beginning of Section 1.7, on p.~60, with the result obtained on p.~62, where the confusion of using the same notation $D$ in $2D = 2 \friction kT$, the spectral density, and in $D = kT / \friction$, the diffusion coefficient [\cite{Coffey.2004}, p.~20, \cite{Gardiner.2004}, p.~7] was noted.
		}
		\begin{align}
			\langle X(t), X(t^\prime) \rangle 
			= 
			2 \friction k T \delta (t - t^\prime)
			\ , 
			\label{eq:covariance-noise-X}
		\end{align}
		where $k$ denotes the Boltzmann constant.  
		
		The covariance of the noise $\noise (t)$ in Eq.~(\ref{eq:SGD-stochastic-diff-eq-2}) is similar to the covariance of the noise $X(t)$ in Eq.~(\ref{eq:covariance-noise-X}), and thus the fluctuation factor $\fluct$, and hence the step length $\learn$ in Eq.~(\ref{eq:fluctuation-2}), can be interpreted as being proportional to temperature $T$.  Therefore, decaying the step length $\learn$, or increasing the minibatch size $\bsize$, is equivalent to cooling down the temperature $T$, and simulating the physical annealing, and hence the name \emph{simulated annealing} (see Remark~\ref{rm:metaheuristics}).
		
		Eq.~(\ref{eq:SDE-langevin-1}) cannot be directly integrated to obtain the velocity $v$ in terms of the noise force $X$ since the derivative does not exist, as interpreted in Eq.~(\ref{eq:SDE-discrete-3}).
		Langevin went around this problem by multiplying Eq.~(\ref{eq:SDE-langevin-1}) by the displacement $x(t)$ and take the average to obtain, \cite{Lemons.1997}:
		\begin{align}
			\frac{m}{2} \frac{d z}{dt} 
			=
			- 3 \pi \mu a z + \frac{RT}{N}
			\ ,
			\label{eq:SDE-langevin-2}
		\end{align}
		where $z = d \overline{(x^2)} / dt$ is the time derivative of the mean square displacement, $R$ the ideal gas constant, and $N$ the Avogadro number.  Eq.~(\ref{eq:SDE-langevin-2}) can be integrated to yield an expression for $z$, which led to Einstein's result for Brownian motion.
	} $\hfill\blacksquare$
\end{rem}

\begin{rem}
	\label{rm:metaheuristics}
	Metaheuristics and nature-inspired optimization algorithms.
	{\rm 
		There is a large class of nature-inspired optimization algorithms that implemented the general conceptual \emph{metaheuristics}---such as neighborhood search, multi-start, hill climbing, accepting negative moves, etc.---and that include many well-known methods such as Evolutionary Algorithms (EAs), Artificial Bee Colony (ABC), Firefly Algorithm, etc. 
		\cite{Lones.2014metaheuristics}.
		
		The most famous of these nature-inspired algorithms would be perhaps simulated annealing in \cite{Kirkpatrick.1983}, which is described in \cite{Yang.2014nature}, p.~18, as being ``inspired by the annealing process of metals. It is a trajectory-based search algorithm starting with an initial guess solution at a high temperature and gradually cooling down the system. A move or new solution is accepted if it is better; otherwise, it is accepted with a probability, which makes it possible for the system to escape any local optima'', i.e., the metaheuristic ``accepting negative moves'' mentioned in \cite{Lones.2014metaheuristics}. ``It is then expected that if the system is cooled down slowly enough, the global optimal solution can be reached'', \cite{Yang.2014nature}, p.~18; that's step-length decay or minibatch-size increase, as mentioned above.    See also Footnotes~\ref{fn:Sun.2019-annealing} and \ref{fn:annealing}.
		
		For applications of these nature-inspired algorithms, we cite the following works, without detailed review:
		\cite{Yang.2014nature}
		\vphantom{\cite{Rere.2015simulated}}\cite{Rere.2015simulated} \vphantom{\cite{Rere.2016metaheuristic}}\cite{Rere.2016metaheuristic} \vphantom{\cite{Fong.2018meta}}\cite{Fong.2018meta} \cite{Bozorg-Haddad.2018advanced} \vphantom{\cite{Al-Obeidat.2019combining}}\cite{Al-Obeidat.2019combining} \cite{Bui.2019metaheuristic} \vphantom{\cite{Devikanniga.2019review}}\cite{Devikanniga.2019review} \vphantom{\cite{Mirjalili.2020nature}}\cite{Mirjalili.2020nature}.
	} $\hfill\blacksquare$
\end{rem}

\subsubsection{Weight decay, avoiding overfit}
\label{sc:weight-decay}

%{\color{red} [NOTE: 2022.09.09 - add remark on the role of weight decay in the modern interpolation regime.]}

\begin{figure}[h]
	\centering
	%
	% 2022.12.17
	% add "-eps-converted-to.pdf" for arXiv
	% \includegraphics[width=0.6\linewidth]{figures/Goodfellow-2016-weight-decay-2020-01-14}
	\includegraphics[width=0.6\linewidth]{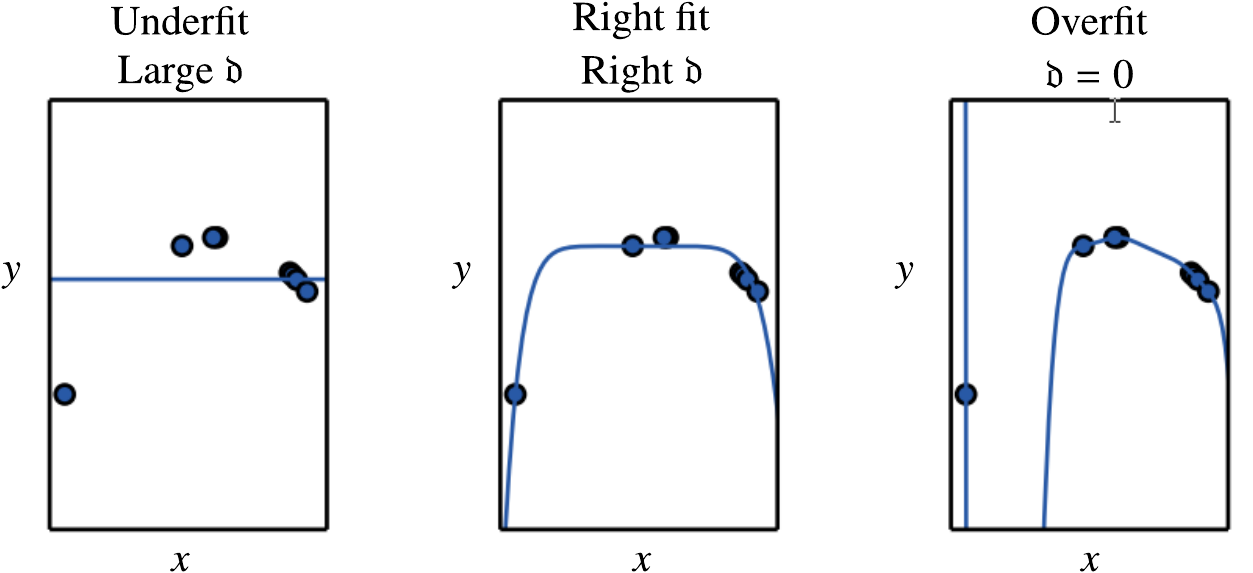}
	\caption{
		\emph{Weight decay} (Section~\ref{sc:weight-decay}).  Effects of magnitude of weight-decay parameter $\decaywt$.  Adapted from \cite{Goodfellow.2016}, p.~116.
		{\footnotesize (Figure reproduced with permission of the authors.)}
		%{\color{red} ASK PERMISSION 2019.11.25}
	}
	\label{fig:weight-decay}
\end{figure}

Reducing, or decaying, the network parameters $\bparam$ (which include the weights and the biases) is one method to avoid overfitting by adding a parameter-decay term to the update equation:
\begin{align}
	\bparamt_{k+1}
	&
	=
	\bparamt_{k} + \blearn_k \bdt_k - \decaywt \bparamt_k 
	%\text{ (element-wise operations)}
	\ ,
	\label{eq:weight-decay}
\end{align}
where $\decaywt \in (0,1)$ is the decay parameter, and there the name ``weight decay'', which is equivalent to SGD with $L_2$ regularization, by adding an extra penalty term in the cost function; see Eq.~(\ref{eq:cost-regularized}) in Section~\ref{sc:adamw} on the adaptive learning-rate method \hyperref[para:adamw]{AdamW}, where such equivalence is explaned following \cite{Loshchilov.2019}. 
``Regularization is any modification we make to a learning algorithm that is intended to reduce its generalization error but not its training error'' \cite{Goodfellow.2016}, p.~117.  Weight decay is only one among other forms of regularization, such as large learning rates, small batch sizes, and dropout, \cite{SmithLN.2018}.  
The effects of the weight-decay parameter $\decaywt$ in avoiding network model overfit is shown in Figure~\ref{fig:weight-decay}.

%
% CMES style rewriting
%\cite{Rognvaldsson.1998} wrote: 
It was written in \cite{Rognvaldsson.1998} that:
``In the neural network community the two most common methods to avoid overfitting are early stopping and weight decay \vphantom{\cite{Plaut.1986}}\cite{Plaut.1986}. Early stopping has the advantage of being quick, since it shortens the training time, but the disadvantage of being poorly defined and not making full use of the available data. Weight decay, on the other hand, has the advantage of being well defined, but the disadvantage of being quite time consuming'' (because of tuning).   For examples of tuning the weight decay parameter $\decaywt$, which is of the order of $10^{-3}$, see \cite{Rognvaldsson.1998} \cite{Loshchilov.2019}.

In the case of weight decay with cyclic annealing, both the step length $\learn_k$ and the weight decay parameter $\decaywt$ are scaled by the  annealing multiplier $\anneal_k$ in the parameter update \cite{Loshchilov.2019}: 
\begin{align}
	\bparamt_{k+1}
	=
	\bparamt_{k} + \anneal_k ( \blearn_k \bdt_k - \decaywt \bparamt_k )
	=
	\bparamt_{k} - \anneal_k ( \blearn_k \bgradt_k + \decaywt \bparamt_k ) 
	%\text{ (element-wise operations)}
	\ .
	\label{eq:weight-decay-cyclic-annealing}
\end{align}
The effectiveness of SGD with weight decay, with and without cyclic annealing, is presented in Figure~\ref{fig:adamw-examples}.

\subsubsection{Combining all add-on tricks}
\label{sc:SGD-combined-tricks}
To have a general parameter-update equation that combines all of the above add-on improvement tricks, start with the parameter update with momentum and accelerated gradient Eq.~(\ref{eq:SGD-update-momentum})
\begin{align}
	\bparamt_{k+1}
	= \bparamt_{k} 
	- \learn_k \bgradt (\bparamt_{k} + \nesterov_k (\bparamt_k - \bparamt_{k-1}) )
	+ \mompar_k (\bparamt_k - \bparamt_{k-1})
	\ ,
	\tag{\ref{eq:SGD-update-momentum}}
\end{align}
and add the weight-decay term $\decaywt \bparamt_k$ from Eq.~(\ref{eq:weight-decay}), then scale both the weight-decay term and the gradient-descent term $(- \learn_{k} \bgradt_{k})$ by the cyclic annealing multiplier $\anneal_k$ in Eq.~(\ref{eq:weight-decay-cyclic-annealing}), leaving the momentum term $\mompar_k (\bparamt_{k} - \bparamt_{k-1})$ alone, to obtain:
\begin{align}
	\bparamt_{k+1}
	= \bparamt_{k} 
	- \anneal_k [ \learn_k \bgradt (\bparamt_{k} + \nesterov_k (\bparamt_k - \bparamt_{k-1}) ) + \decaywt_k \bparamt_{k} ]
	+ \mompar_k (\bparamt_k - \bparamt_{k-1})
	\ ,
	\label{eq:update-params-combined-tricks}
\end{align}
which is included in Algorithm~\ref{algo:generic-SGD}.

%{\color{red} HERE 2020.01.28}

\subsection{Kaiming He initialization}
%Kaiming He's original paper \cite{He.2015:rd0001}

All optimization algorithms discussed above share a crucial step, which crucially affects the convergence of training, especially as neural networks become `deep': 
The \emph{initialization} of the network's parameters $\bparamt_0$ is not only related to the speed of convergence, it ``can determine whether the algorithm converges at all.''\footnote{
See~\cite{Goodfellow.2016}, p.~292.} 
Convergence problems are typically related to the scaling of initial parameters. Large parameters result in large activations, which leads to exploding values during forward and backward propagation, i.e., evaluation of the loss function and computation of its gradient with respect to the parameters.
Small parameters, on the other hand, may result in \emph{vanishing gradients}, i.e., the loss function becomes insensitive to parameters, which causes the training process to stall. 
To be precise, considerations regarding initialization are related to weight-matrices $\boldsymbol W^{(\ell)}$, see Section~\ref{sc:network-layer-details} on ``Network layer, detailed construct''; bias vectors $\boldsymbol b^{(\ell)}$ are usually initialized to zero, which is also assumed in the subsequent considerations.\footnote{  
	Situations favoring a nonzero initialization of weights are explained in~\cite{Goodfellow.2016}, p.~297.
}

The \emph{Kaiming He initialization} provides equally effective as simple means to overcome scaling issues observed when weights are randomly initialized using a normal distribution with fixed standard deviation.
The key idea of the authors~\cite{He.2015:rd0001} is to have the same variance of weights for each of the network's layers. 
As opposed to the \emph{Xavier initialization}\footnote{
	See~\cite{Glorot.2010}.
},
the nonlinearity of activation functions is accounted for.
Consider the $\ell$-th layer of a feedforward neural network, where the vector $\boldsymbol z^{(\ell)}$ follows as affine function of inputs $\boldsymbol y^{(\ell-1)}$
\begin{equation}
	\boldsymbol z^{(\ell)}
	=
	\boldsymbol W^{(\ell)} \boldsymbol y^{(\ell-1)}
	+
	\boldsymbol b^{(\ell)} , 
	\qquad
	\boldsymbol y^{(\ell)}
	=
	\g (\boldsymbol z^{(\ell)}) ,
\end{equation}
where $\boldsymbol W^{(\ell)}$ and $\boldsymbol b^{(\ell)}$ denote the layer's weight matrix and bias vector, respectively.
The output of the layer $\boldsymbol y^{(\ell)}$ is given by element-wise application of the activation function $a$, see Sections~\ref{sc:weigths-biases} and \ref{sc:activation-functions} for a  detailed presentation.
All components of the weight matrix $\boldsymbol W^{(\ell)}$ are assumed to be independent of each other and to share the same probability distribution.
The same holds for the components of the input vector $\boldsymbol y^{(\ell-1)}$ and the output vector $\boldsymbol y^{(\ell)}$.
Additionally, elements of $\boldsymbol W^{(\ell)}$ and $\boldsymbol y^{(\ell)}$ shall be mutually independent.
Further, it is assumed that $\boldsymbol W^{(\ell)}$ and $\boldsymbol y^{(\ell)}$ have zero mean (i.e., expectation, cf. Eq.~\eqref{eq:expectation}) and are symmetric around zero.
In this case, the variance of $\boldsymbol z^{(\ell)} \in \mathbb{R}^{m_{(\ell)} \times 1}$ is given by
\begin{align}
	\Var z^{(\ell)}_i  
%	&= \Var \left( 
%	\boldsymbol W^{(\ell)} \boldsymbol y^{(\ell-1)}
%	+
%	\boldsymbol b^{(\ell)} \right) 
	&= m_{(\ell)} \Var \left( W^{(\ell)}_{ij} y^{(\ell-1)}_j \right) 
	\\
	&= m_{(\ell)} \left( \Var W^{(\ell)}_{ij} \Var y^{(\ell-1)}_j + \mathbb E(W^{(\ell)}_{ij})^2  \Var y^{(\ell-1)}_j  + \Var W^{(\ell)}_{ij} \mathbb E (y^{(\ell-1)}_j )^2 \right) 
	\\
	&= m_{(\ell)} \Var W^{(\ell)}_{ij} \left(  \Var y^{(\ell-1)}_j + \mathbb E( y^{(\ell-1)}_j )^2 \right) ,
\end{align}

%{\color{red} [NOTE: 2022.09.18 - Eq.(177) is not clear.  Need to recall what $m_{(\ell)}$ was (e.g., width of layer $\ell$), and refer to a precise location in the paper where it was defined.  Don't let the readers have to look around, since it is a long paper, a book.  ENDNOTE]}
% AH: 2022.10.04 - done

\noindent
where $m_{(\ell)}$ denotes the width of the $\ell$-th layer and the fundamental relation $\Var (XY) = \Var X \Var Y + \mathbb E(X)^2 \Var Y + \mathbb E(Y)^2 \Var X$ has been used along with the assumption of weights having a zero mean, i.e., $ \mathbb E(W^{(\ell)}_{ij}) = 0$.
The variance of some random variable $X$, which is the expectation of the squared deviation of $X$ from its mean, i.e.,
\begin{equation}
	\Var X = \mathbb E((X - \mathbb E(X))^2) ,
	\label{eq:variance}
\end{equation}
is a measure of the ``dispersion'' of $X$ around its mean value.
The variance is the square of the standard deviation $\stdev$ of the random variable $X$, or, conversely, 
\begin{equation}
	\stdev (X) = \sqrt{\Var X}  = \sqrt{ \mathbb E((X - \mathbb E(X))^2 } .
\end{equation}
As opposed to the variance, the standard deviation of some random variable $X$ has the same physical dimension as $X$ itself.

The elementary relation $\Var X^2  = \mathbb E(X^2) - \mathbb E(X)^2$ gives
\begin{equation}
	\Var z^{(\ell)}_i = m_{(\ell)} \Var W^{(\ell)}_{ij} \mathbb E((y^{(\ell-1)}_j)^2 ) ,
	\label{eq:kaiming_1}
\end{equation}
%

%{\color{red} [NOTE: 2022.09.05 - Relate the variance to the standard deviation using our notation, which most people would understand.] }

Note that the mean of inputs does not vanish for activation functions that are not symmetric about zero as, e.g., the ReLU functions (see Section~\ref{sc:relu}).
For the ReLU activation function, $y = a(x) = max(0,x)$ the mean value of the squared output and the variance of the input are related by
\begin{align}
	\mathbb E(y^2)
	&= \int_{-\infty}^\infty y^2 P(y) \mathrm dy \\
	&= \int_{-\infty}^\infty \max (0, x)^2 P(x) \mathrm dx 
	= \int_{0}^\infty x^2 P(x) \mathrm dx 
	= \frac 1 2 \int_{-\infty}^\infty x^2 P(x) \mathrm dx \\
	&= \frac 1 2 \Var x .
\end{align}
Substituting the above result in Eq.~\eqref{eq:kaiming_1} provides the following relationship among the variances of the inputs to the activation function of two consecutive layers, i.e., $\Var \boldsymbol z^{(\ell)}$ and $\Var \boldsymbol z^{(\ell-1)}$, respectively:
\begin{equation}
	\Var z^{(\ell)}_i = \frac{m_{(\ell)}}{2} \Var W^{(\ell)}_{ij}  \Var z^{(\ell-1)}_j .
\end{equation}
For a network with $L$ layers, the following relation between the variance of inputs $\Var \boldsymbol z^{(1)}$ and outputs $\Var \boldsymbol z^{(L)}$  is obtained:
\begin{equation}
	z^{(L)}_i =  \Var z^{(1)}_j \prod_{\ell = 2}^L \frac{m_{(\ell)}}{2}  \Var W^{(\ell)}_{ij} .
\end{equation}
To preserve the variance through all layers of the network, the following condition must be fulfilled regarding the variance of weight matrices:
\begin{equation}
	 \frac{m_{(\ell)}}{2}  \Var W^{(\ell)}_{ij} = 1 \quad \forall \ell 
	 \qquad \leftrightarrow \qquad
	 \boldsymbol W^{(\ell)} \sim \mathcal N \left( 0 , \frac{2}{m_{(\ell)}} \right) ,
\end{equation}
where $\mathcal N(0, \stdev^2)$ denotes the normal (or Gaussian) distribution with zero mean and
$\stdev^2 = 2 / m_{(\ell)}$ variance.
The above result, which is known as \emph{Kaiming He initialization}, implies that the width of a layer $m_{(\ell)}$ needs to be regarded in the initialization of weight matrices.
Preserving the variance of inputs mitigates exploding or vanishing gradients and improves convergence in particular for deep networks.
The authors of~\cite{He.2015:rd0001} provided analogous results for the parametric rectified linear unit (PReLU, see Section~\ref{sc:parametric-ReLU}).

\subsection{Adaptive methods: Adam, variants, criticism}
\label{sc:adaptive-learning-rate-algos}
\label{sc:adam-amsgrad}

The Adam algorithm was introduced in \cite{Kingma.2014} (version 1), and updated in 2017 (version 9), and has been ``immensely successful in development of several state-of-the-art solutions for a wide range of problems,'' as stated in \cite{Reddi.2019}. 
%
% CMES style rewriting 
%According to \vphantom{\cite{Bock.2018}}\cite{Bock.2018}, 
``In the area of neural networks, the ADAM-Optimizer is one of the most popular adaptive step size methods. It was invented in \cite{Kingma.2014}. The 5865 citations in only three years shows additionally the importance of the given paper''\footnote{
	Reference \cite{Kingma.2014} introduced the Adam algorithm in 2014, received 34,535 citations on 2019.12.11, after 5 years, and a whopping 112,797 citations on 2022.07.11, after an additional period of more than 2.5 years later, according to Google Scholar.
} \vphantom{\cite{Bock.2018}}\cite{Bock.2018}.
%
% CMES style rewriting
The authors of  
\vphantom{\cite{Huang.2019}}\cite{Huang.2019} concurred: ``Adam is widely used in both academia and industry. However, it is also one of the least well-understood algorithms. In recent years, some remarkable works provided us with better understanding of the algorithm, and proposed different variants of it.''  

\begin{algorithm}
	{\bf Unified adaptive learning-rate, momentum, weight decay, cyclic annealing} (for epoch $\tau$)
	\\
	\KwData{
		% Layer outputs $\byp{\ell}$, for $\ell=L, \cdots , 1$ 
		\\
		$\bullet$ Network parameters $\bparamt_0$ obtained from previous epoch $(\tau-1)$
		\\
		{\color{purple} $\bullet$ Select a small stabilization parameter $\delta > 0$, e.g., $10^{-8}$} 
		\\
		$\bullet$ Select $\tei$ as epoch $\tau$ or global iteration $j$, Eq.~(\ref{eq:learning-rate-schedule-b}), and budget $\tau_{max}$ or $j_{max}$ %\label{lst:list:SGD-budget-not-met}
		\\
		$\bullet$ Select learning-rate schedule $\epsilon(\tei) \in \{$ \text{Eqs.~(\ref{eq:learning-rate-schedule})-(\ref{eq:learning-rate-schedule-4}} $\}$, and parameters $\epsilon_0, k_c, \decay$ 
		\\
		$\bullet$ Select $\mompar$ if use standard momentum, and $\nesterov$ if use Nesterov Accelerated Gradient, Eq.~(\ref{eq:SGD-update-momentum})
		\\
		$\bullet$ Select a value for $\bsize$ the number of examples in the minibatches
	}
	\KwResult{
		Updated network parameters $\bparamt_0$ for the next epoch $(\tau+1)$.
	}
	\vphantom{Blank line}
	
	%{\color{purple} Define functions $\bmom_k = \phi_t (\bgradt_1, \ldots , \bgradt_k)$ and $\bmomc_k = \chi_{\phi_t} (\bgradt_1, \ldots , \bgradt_k)$ as in Eq.~(\ref{eq:momentum-0})}
	%\\
	%{\color{purple} Define functions $\bvar_k = \psi_t (\bgradt_1, \ldots , \bgradt_k)$ and $\bvarc_k = \chi_{\psi_t} (\bgradt_1, \ldots , \bgradt_k)$ as in Eq.~(\ref{eq:variance-0})}
	% \\
	% Initialize 1st moment: $\bmom_0 = 0$ 
	% \\
	% Initialize 1st moment: $\bvar_0 = 0$ 
	%\\
	$\blacktriangleright$ Begin training epoch $\tau$
	\\
	$\blacktriangleright$ Initialize $\bparamt_1 = \bparamt_\tau^\star$ (from previous training epoch) (line~\ref{lst:list:SGD-params-previous-epoch} in Algorithm~\ref{algo:generic-SGD})
	%\\
	% Initialize accumulation variables
	
	% gradient descent
	\ding{173}
	\For{$k=1,2, \ldots, k_{max}$}{
		
		Obtain random minibatch $\Bbbps{k}{\bsize}$ containing $\bsize$ examples, Eq.~(\ref{eq:minibatch-2})
		\;
		Compute gradient estimate $\bgradt_k = - \bgradt (\bparamt_k)$ per Eq.~(\ref{eq:gradient-estimate})
		\;
		{\color{purple}
			Compute $\bmom_k = \phi_t (\bgradt_1, \ldots , \bgradt_k)$ and $\bmomc_k = \chi_{\phi_t} (\bgradt_1, \ldots , \bgradt_k)$ as in Eq.~(\ref{eq:momentum-0})
			\;
			Compute $\bvar_k = \psi_t (\bgradt_1, \ldots , \bgradt_k)$ and $\bvarc_k = \chi_{\psi_t} (\bgradt_1, \ldots , \bgradt_k)$ as in Eq.~(\ref{eq:variance-0})
			\;
			Compute learning rate $\blearn_k$ as in Eq.~(\ref{eq:adaptive-learning-rate})
			\;
			Set descent direction $\bdt_k = (- \bmomc_k)$
		}
		\;
		
		\eIf{Stopping-criterion not met (if used, Section~\ref{sc:training-valication-test})}{
			
			% THEN, gradient descent 
			%\vspace{2mm}
			
			%\vspace{2mm}
			%{\color{red} HERE}
			$\blacktriangleright$ Update network parameter $\bparamt_k$ to $\bparamt_{k+1}$: Use Eq.~(\ref{eq:update-params-combined-tricks}) \label{lst:line:adaptive-learning-update}
			\\
			{\color{purple} with Eq.~(\ref{eq:adaptive-learning-rate}), which includes Eq.~(\ref{eq:update-param-adaptive}) and  Eq.~(\ref{eq:adamw-param-update}) for \hyperref[para:adamw]{AdamW}} 
			\label{lst:line:adaptive-learning-update-2}
			\;
			%\vspace{2mm} 
			
		}{
			
			% ELSE, gradient descent
			%\vspace{2mm}
			$\blacktriangleright$ Stopping criterion met
			\\
			Reset initial minimizer-estimate for next training epoch: $\bparamt_0 \leftarrow \bparamt_k$ 
			\;
			Stop \ding{173} {\bf for} loop
			\;
			
		} % ENDIF, gradient descent
		
	} % END For loop	
	
	\vphantom{Blank line} 
	\caption{
		\emph{Unified adaptive learning-rate algorithm} (Section~\ref{sc:unified-adaptive}, \ref{para:adagrad}, \ref{para:rmsprop}, \ref{para:adadelta}, \ref{para:adam1}, \ref{para:amsgrad}, \ref{para:adamw}, Algorithm~\ref{algo:generic-SGD}) which includes momentum, accelerated gradient, weight decay, cyclic annealing. The outer \ding{172} {\bf for} $\tau=1, \ldots , \tau_{max}$ loop over the training epochs, as shown in Algorithm~\ref{algo:generic-SGD} for SGD, is not presented to focus on the \ding{173} $\text{\bf for}$ $k=1,\ldots, k_{max}$ inner loop of one pass over the training set, for a typical training epoch $\tau$.  The differences with Algorithm~\ref{algo:generic-SGD} are highlighted in purple color.  For the parameter update in lines~\ref{lst:line:adaptive-learning-update}-\ref{lst:line:adaptive-learning-update-2}, see also Footnote~\ref{fn:adaptive-algos-param-update}.
	}
	\label{algo:unified-adaptive-learning-rate-2}
\end{algorithm}

\subsubsection{Unified adaptive learning-rate pseudocode}
\label{sc:unified-adaptive}

%
% CMES style rewriting
%\cite{Reddi.2019} suggested a 
It was suggested in \cite{Reddi.2019} a unified pseudocode, adapted in Algorithm~\ref{algo:unified-adaptive-learning-rate-2}, that included not only the standard SGD in Algorithm~\ref{algo:generic-SGD}, but also a number of successful adaptive learning-rate methods: \hyperref[para:adagrad]{AdaGrad}, \hyperref[para:rmsprop]{RMSProp}, \hyperref[para:adadelta]{AdaDelta}, \hyperref[para:adam1]{Adam}, the recent \hyperref[para:amsgrad]{AMSGrad}, \hyperref[para:adamw]{AdamW}.  Our adaptation in Algorithm~\ref{algo:unified-adaptive-learning-rate-2} also includes Nostalgic Adam and AdamX.\footnote{
	\label{fn:adaptive-algos-param-update}
	In lines~\ref{lst:line:adaptive-learning-update}-\ref{lst:line:adaptive-learning-update-2} of Algorithm~\ref{algo:unified-adaptive-learning-rate-2}, use Eq.~(\ref{eq:update-params-combined-tricks}), but replacing scalar learning rate $\learn_{k}$ with matrix learning rate $\blearn_k$ in Eq.~(\ref{eq:adaptive-learning-rate}) to update parameters; the result includes Eq.~(\ref{eq:update-param-adaptive}) for vanilla adaptive methods and Eq.~(\ref{eq:adamw-param-update}) for \hyperref[para:adamw]{AdamW}.
}

%Figure~\ref{fig:Adam-converge} shows the convergence of some adaptive learning-rate algorithms: \hyperref[para:adagrad]{AdaGrad}, \hyperref[para:rmsprop]{RMSProp}, \hyperref[sc:SGD-momentum]{SGDNesterov}, \hyperref[para:adadelta]{AdaDelta}, \hyperref[para:adam1]{Adam}.

Four new quantities are introduced for iteration $k$ in SGD: 
(1) $\bmom_k$ at SGD iteration $k$, as the first moment, and 
(2) its correction $\bmomc_k$
\begin{align}
	\bmom_k = \phi_k (\bgradt_1 , \ldots, \bgradt_k)
	\ , \quad
	\bmomc_k = \chi_{\phi_k} (\bmom_k)
	\ ,
	\label{eq:momentum-0}
\end{align}
and (3) the second moment (variance)\footnote{
	The uppercase letter $\bvar$ is used instead of the lowercase letter $\boldsymbol{v}$, which is usually reserved for ``velocity'' used in a term called ``momentum'', which is added to the gradient term to correct the descent direction. Such algorithm is called gradient descent with momentum in deep-learning optimization literature; see, e.g., \cite{Goodfellow.2016}, Section 8.3.2 Momentum, p.~288.
}  $\bvar_k$ and (4) its correction $\bvarc_k$
\begin{align}
	\bvar_k = \psi_k (\bgradt_1 , \ldots, \bgradt_k)
	\ , \quad
	\bvarc_k = \chi_{\psi_k} (\bvar_k)
	\ .
	\label{eq:variance-0}
\end{align}
The descent direction estimate $\bdt_k$ at SGD iteration $k$ for each training epoch is
\begin{align}
	\bdt_k = - \bmomc_k
	\ .
	\label{eq:adaptive-descent-direction}
\end{align}
The adaptive learning rate $\epsilon_k$ is obtained from rescaling the fixed learning-rate schedule $\epsilon(k)$, also called the ``global'' learning rate particularly when it is a constant, using the 2nd moment $\bvarc_k$ as follows:
\begin{align}
	\blearn_k = \frac{\learn (k)}{\sqrt{\bvarc_k  + \delta}}
	\ ,
	\text{ or }
	\blearn_k = \frac{\learn (k)}{\sqrt{\bvarc_k} + \delta} 
	\text{ (element-wise operations)}
	\label{eq:adaptive-learning-rate}
\end{align}
where $\epsilon (k)$ can be either Eq.~(\ref{eq:learning-rate-schedule})\footnote{
	Suggested in \cite{Goodfellow.2016}, p.~287.
} (which includes $\epsilon(k) = \epsilon_0 =$ constant) or Eq.~(\ref{eq:learning-rate-schedule-2});\footnote{
	Suggested in \cite{Reddi.2019}, p.~3.
} $\delta$ is a small number to avoid division by zero;\footnote{
	AdaDelta and RMSProp used the first form of Eq.~(\ref{eq:adaptive-learning-rate}), with $\delta$ outside the square root, whereas AdaGrad and Adam used the second part, with $\delta$ inside the square root.  AMSGrad, Nostalgic Adam, AdamX did not use $\delta$, i.e., set $\delta = 0$.
} the operations (square root, addition, division) are element-wise, with both $\epsilon_0$ and $\delta = O(10^{-6}) \text{ to } O(10^{-8})$ (depending on the algorithm) being constants.

\begin{rem}
	\label{rm:case-adadelta}
	{\rm 
		A particular case is the AdaDelta algorithm, in which $\bmom_k$ in Eq.~(\ref{eq:momentum-0}) is the second moment for the network parameter increments $\{\Delta \bparam_{i}, i = 1, \ldots , k\}$, and $\bmomc_k$ also in Eq.~(\ref{eq:momentum-0}) the corrected gradient.
	}
	\phantom{blah} $\hfill\blacksquare$
\end{rem}

All of the above arrays---such as $\bmom_k$ and $\bmomc_k$ in Eq.~(\ref{eq:momentum-0}), $\bvar_k$ and $\bvarc_k$ in Eq.~(\ref{eq:variance-0}), and $\bdt_k$ in Eq.~(\ref{eq:adaptive-descent-direction})---together the resulting array $\blearn_k$ in Eq.~(\ref{eq:adaptive-learning-rate}) has the same structure as the network parameter array $\bparam$ in Eq.~(\ref{eq:theta}), with $\Tparam$ in Eq.~(\ref{eq:totalParams}) being the total number of parameters.  The update of the network parameter estimate in $\bparam$ is written as follows:
\begin{align}
	\bparamt_{k+1} 
	= 
	\bparamt_{k} + \blearn_k \odot \bdt_k
	=
	\bparamt_{k} + \blearn_k \bdt_k
	=
	\bparamt_{k} + \Delta \bparamt_{k}
	\text{ (element-wise operations)}
	\ ,
	\label{eq:update-param-adaptive}
\end{align}
where the Hadamard operator $\odot$ (element-wise multiplication) is omitted to alleviate the notation.\footnote{
	There are no symbols similar to the Hadamard operator symbol $\odot$  for other operations such as square root, addition, and division, as implied in Eq.~(\ref{eq:adaptive-learning-rate}), so there is no need to use the symbol $\odot$ just for multiplication.
} 

\begin{rem}
	\label{rm:learning-rate-each-param}
	{\rm The element-wise operations in Eq.~(\ref{eq:adaptive-learning-rate}) and Eq.~(\ref{eq:update-param-adaptive}) would allow each parameter in array $\bparam$ to have its own learning rate, unlike in traditional deterministic optimization algorithms, such as in Algorithm~\ref{algo:descent-armijo-deterministic} or even in the Standard SGD Algorithm~\ref{algo:generic-SGD}, where the same learning rate is applied to all parameters in $\bparam$.}
	$\hfill\blacksquare$
\end{rem}
 
It remains to define the functions $\phi_t$ and $\chi_{\phi_t}$ in Eq.~(\ref{eq:momentum-0}), and $\psi_t$ and $\chi_{\psi_t}$ in Eq.~(\ref{eq:variance-0}) for each of the particular algorithms covered by the unified Algorithm~\ref{algo:unified-adaptive-learning-rate-2}.

{\bf SGD.}  
To obtain Algorithm~\ref{algo:generic-SGD} as a particular case, select the following functions for Algorithm~\ref{algo:unified-adaptive-learning-rate-2}:
\begin{align}
	&
	\phi_k (\bgradt_1 , \ldots, \bgradt_k) = \bgradt_k
	\ , 
	\text{ with }
  	% CMES style - 2022.06.01
	% \chi_{\phi_k} = \bId  \text{ (Identity)}
	% same problem with \bId, which could be that \I had been defined as something else in the TSP style
	% so just code up boldsymbol I explicitly
	\chi_{\phi_k} = \boldsymbol{I}  \text{ (Identity)}
	\Rightarrow
	\bmom_k = \bgradt_k = \bmomc_k
	\ ,
	\\
	&
	% \psi_k (\bgradt_1 , \ldots, \bgradt_k) = \chi_{\psi_k} = \bId
	\psi_k (\bgradt_1 , \ldots, \bgradt_k) = \chi_{\psi_k} = \boldsymbol{I} 
	\text{ and } \delta = 0
	\Rightarrow
	% \bvarc_k = \bId
	\bvarc_k = \boldsymbol{I}
	\text{ and }
	% \blearn_k = \learn(k) \bId
	\blearn_k = \learn(k) \boldsymbol{I}
	\ ,
	\label{eq:unified-2-SGD}
\end{align}
together with learning-rate schedule $\epsilon (k)$ presented in Section~\ref{sc:step-length-decay} on step-length decay and annealing.  In other words, from Eqs.~(\ref{eq:adaptive-descent-direction})-(\ref{eq:update-param-adaptive}), the parameter update reduces to that of the vanilla SGD with the fixed learning-rate schedule $\learn(k)$, without scaling:
\begin{align}
	\bparamt_{k+1} 
	=
	\bparamt_{k} + \blearn_k \bdt_k
	=
	\bparamt_{k} - \learn_k \bgradt_k
	\ .
	\label{eq:unified-SGD-update}
\end{align}
Similarly for SGD with momentum and accelerated gradient (Section~\ref{sc:SGD-momentum}), step-length decay and cyclic annealing (Section~\ref{sc:step-length-decay}), weight decay (Section~\ref{sc:weight-decay}).

\begin{figure}[h]
	\centering
	\includegraphics[width=0.7\linewidth]{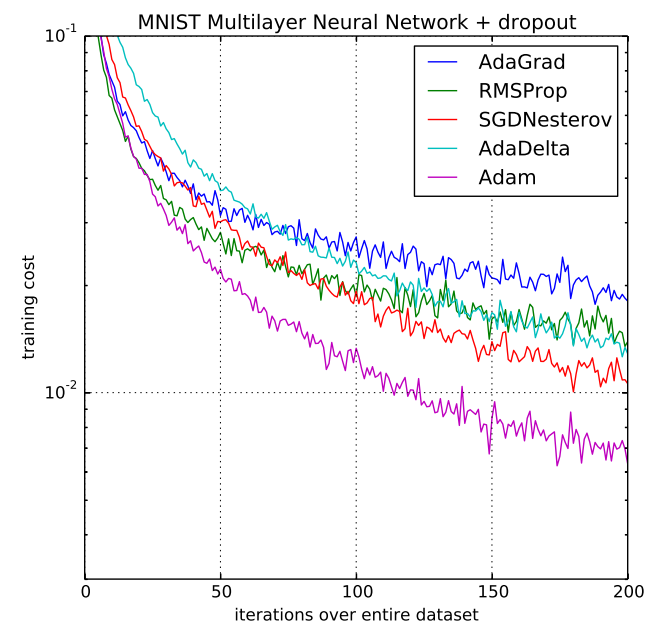}
	\caption{
		\emph{Convergence of adaptive learning-rate algorithms} (Section~\ref{sc:SGD-momentum}):  \hyperref[para:adagrad]{AdaGrad}, \hyperref[para:rmsprop]{RMSProp}, 
		\hyperref[sc:SGD-momentum]{SGDNesterov}, \hyperref[para:adadelta]{AdaDelta}, \hyperref[para:adam1]{Adam} \cite{Kingma.2014}.  
%		From \cite{Kingma.2014}.
		{\footnotesize (Figure reproduced with permission of the authors.)}
		%{\color{red} ASK PERMISSION 2019.11.25}
	}
	\label{fig:Adam-converge}
\end{figure}

%{\color{red} BEGIN capitalize 2020.01.30.}

%{\bf AdaGrad.}
%(Adaptive Gradient.)
\subsubsection{AdaGrad: Adaptive Gradient}
\label{para:adagrad} 
\label{sc:adagrad}

Starting the line of research on adaptive learning-rate algorithms,
%
% CMES style rewriting
the authors of
\vphantom{\cite{Duchi.2011}}\cite{Duchi.2011}\footnote{
	As of 2019.11.28, \cite{Duchi.2011} was cited 5,385 times on Google Scholars, and 1,615 times on Web of Science.
	By 2022.07.11, \cite{Duchi.2011} was cited 10,431 times on Google Scholars, and 3,871 times on Web of Science.
} selected the following functions for Algorithm~\ref{algo:unified-adaptive-learning-rate-2}:
\begin{align}
	&
	\phi_k (\bgradt_1 , \ldots, \bgradt_k) = \bgradt_k
	\ , 
	\text{ with }
  	% CMES style - 2022.06.01
	% \chi_{\phi_k} = \bId \text{ (Identity)}
	% same problem with \bId, which could be that \I had been defined as something else in the TSP style
	% so just code up boldsymbol I explicitly
	\chi_{\phi_k} = \boldsymbol{I} \text{ (Identity)}
	\Rightarrow
	\bmom_k = \bgradt_k = \bmomc_k
	\ ,
	\\
	&
	\psi_k (\bgradt_1 , \ldots, \bgradt_k) = \sum_{i=1}^{k} \bgradt_i \odot \bgradt_i = \sum_{i=1}^{k} (\bgradt_i)^2
	\text{ (element-wise square)}
	\ ,
	\\
	&
  	% CMES style - 2022.06.01
	% same problem with \bId, which could be that \I had been defined as something else in the TSP style
	% so just code up boldsymbol I explicitly
	%
	% \chi_{\psi_k} = \bId
	\chi_{\psi_k} = \boldsymbol{I} 
	\text{ and } \delta = 10^{-7}
	\Rightarrow
	\bvarc_k = \sum_{i=1}^{k} (\bgradt_i)^2
	\text{ and }
	\blearn_k = \frac{\learn (k)}{\sqrt{\bvarc_k} + \delta}
	\text{ (element-wise operations)}
	\ ,
	\label{eq:unified-2-AdaGrad}
\end{align}
leading to an update with adaptive scaling of the learning rate
\begin{align}
	\bparamt_{k+1} 
	=
	\bparamt_{k} - \frac{\learn (k)}{\sqrt{\bvarc_k} + \delta} \bgradt_k
	\text{ (element-wise operations)}
	\ ,
	\label{eq:unified-adagrad-update}
\end{align}
in which each parameter in $\bparamt_k$ is updated with its own learning rate.  For a given network parameter, say, $\param_{pq}$, its learning rate $\learn_{k,pq}$ is essentially $\learn (k)$ scaled with the inverse of the square root of the sum of all historical values of the corresponding gradient component $(pq)$, i.e., $(\delta + \sum_{i=1}^{k} \gradt_{i,pq}^2)^{-1/2}$, with $\delta = 10^{-7}$ being very small.  A consequence of such scaling is that a larger gradient component would have a smaller learning rate and a smaller per-iteration decrease in the learning rate, whereas a smaller gradient component would have a larger learning rate and a higher per-iteration decrease in the learning rate, even though the relative decrease is about the same.\footnote{
	See \cite{Zeiler.2012}. For example, compare the sequence $\{ \frac15 , \frac{1}{5+5} \}$ to the sequence $\{ \frac12 , \frac{1}{2+2} \}$.  
	%
	% CMES style rewriting
	The authors of
	\cite{Goodfellow.2016}, p.~299, mistakenly stated ``The parameters with the largest partial derivative of the loss have a correspondingly rapid decrease in their learning rate, while parameters with small partial derivatives have a relatively small decrease in their learning rate.''
}  Thus progress along different directions with large difference in gradient amplitudes is evened out as the number of iterations increases.\footnote{
	%
	% CMES style rewriting
%	\cite{Zeiler.2012} stated
	It was stated in
	\cite{Zeiler.2012}:  ``progress along each dimension evens
	out over time. This is very beneficial for training deep neural networks since the scale of the gradients in each layer is
	often different by several orders of magnitude, so the optimal
	Learning rate should take that into account.''  Such observation made more sense than saying ``The net effect is greater progress in the more gently sloped directions of parameter space'' as did
	%
	% CMES style rewriting
	the authors of 
	\cite{Goodfellow.2016}, p.~299, who referred to AdaDelta in Section 8.5.4, p.~302, through the work of other authors, but might not read \cite{Zeiler.2012}.
}

Figure~\ref{fig:Adam-converge} shows the convergence of some adaptive learning-rate algorithms: \hyperref[para:adagrad]{AdaGrad}, \hyperref[para:rmsprop]{RMSProp}, 
\hyperref[sc:SGD-momentum]{SGDNesterov}, \hyperref[para:adadelta]{AdaDelta}, \hyperref[para:adam1]{Adam}.

%{\color{red} END capitalize 2020.01.30.}

\subsubsection{Forecasting time series, exponential smoothing}
\label{sc:exponential-smoothing}
%\begin{rem}
	%\label{rm:exponential-smoothing}
	%{\rm 
	%$\hfill\blacksquare$ 
	%}
%\end{rem}
%Forecasting time series, exponential smoothing.
At this point, all of the subsequent adaptive learning-rate algorithms made use of an important technique in forecasting known as exponential smoothing of time series, without using this terminology, but instead referred to such technique as 
	``exponential decaying average'' \cite{Zeiler.2012}, \cite{Goodfellow.2016}, p.~300,
	``exponentially decaying average'' \cite{Bottou.2018:rd0001},
	``exponential moving average'' \cite{Kingma.2014}, \cite{Reddi.2019}, \vphantom{\cite{Chen.2019}}\cite{Chen.2019}, \cite{Sun.2019},
	``exponential weight decay'' \cite{Loshchilov.2019}.
	
	\begin{figure}[h]
		\centering
		\includegraphics[width=0.7\linewidth]{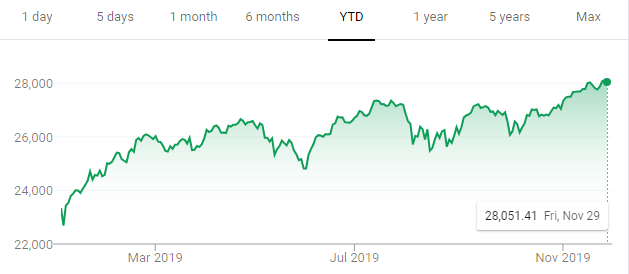}
		\caption{
			\emph{Dow Jones Industrial Average} (DJIA, Section~\ref{sc:exponential-smoothing}) stock index year-to-date (YTD) chart as from 2019.01.01 to 2019.11.30, Google Finance.
		}
		\label{fig:dow-jones}
	\end{figure}
	
	``Exponential smoothing methods have been around since the 1950s, and are still the most popular forecasting methods used in business and industry'' such as ``minute-by-minute stock prices, hourly temperatures at a weather station, daily numbers of arrivals at a medical clinic, weekly sales of a product, monthly unemployment figures for a region, quarterly imports of a country, and annual turnover of a company'' \vphantom{\cite{Hyndman.2008}}\cite{Hyndman.2008}.  See Figure~\ref{fig:dow-jones} for the chart of a stock index showing noise.
	
\begin{figure}[h]
		\centering
		\includegraphics[width=0.7\linewidth]{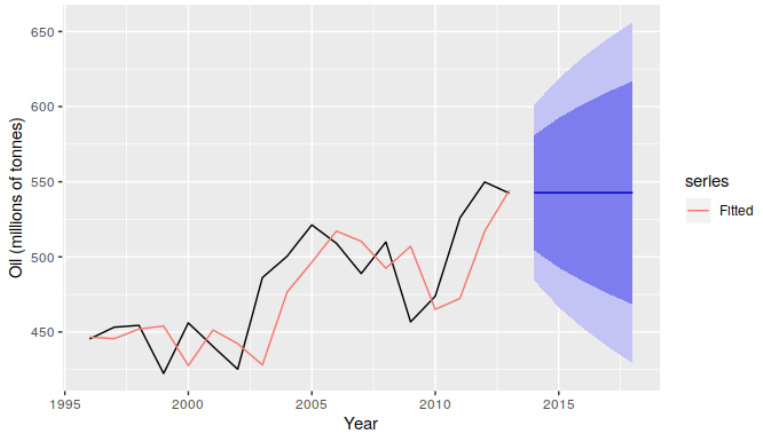}
		\caption{
			\emph{Saudi Arabia oil production during 1996-2013} (Section~\ref{sc:exponential-smoothing}).  Piecewise linear data (black) and fitted curve (red), despite the name ``smoothing''. From \cite{Hyndman.2018}, Chap.~7.
			{\footnotesize (Figure reproduced with permission of the authors.)}
			%{\color{red} ASK PERMISSION}
		}
		\label{fig:arabian-oil}
	\end{figure}

	``Exponential smoothing was proposed in the late 1950s (Brown, 1959; Holt, 1957; Winters, 1960), and has motivated some of the most successful forecasting methods. Forecasts produced using exponential smoothing methods are weighted averages of past observations, with the weights decaying exponentially as the observations get older. In other words, the more recent the observation the higher the associated weight. This framework generates reliable forecasts quickly and for a wide range of time series, which is a great advantage and of major importance to applications in industry'' \cite{Hyndman.2018}, Chap.~7, ``Exponential smoothing''.  See Figure~\ref{fig:arabian-oil} for an example of ``exponential-smoothing'' curve that is not ``smooth''.
	
	For neural networks, early use of exponential smoothing dates back at least to 1998 in \cite{Schraudolph.1998} and \cite{Neuneier.1998}.\footnote{
		We thank Lawrence Aitchison for informing us about these references; see also \cite{Aitchison.2019}.
	}
	
	For adaptive learning-rate algorithms further below (\hyperref[para:rmsprop]{RMSProp}, \hyperref[para:adadelta]{AdaDelta}, \hyperref[para:adam1]{Adam}, etc.), let $\{x(t), \, t=1,2,\ldots\}$ be a noisy raw-data time series as in Figure~\ref{fig:dow-jones}, and $\{s(t), \, t=1,2,\ldots\}$ its smoothed-out counterpart.  The following recurrence relation is an exponential smoothing used to predict $s(t+1)$ based on the known value $s(t)$ and the data $x(t+1)$:
	\begin{align}
		s(t+1) = \beta s(t) + (1 - \beta) x(t+1)
		\ ,
		\text{ with }
		\beta \in [0,1)
		\label{eq:exponential-smoothing-1}
	\end{align} 
	Eq.~(\ref{eq:exponential-smoothing-1}) is a convex combination between $s(t)$ and $x(t+1)$.  a value of $\beta$ closer to 1, e.g., $\beta = 0.9$ and $1-\beta = 0.1$, would weigh the smoothed-out past data $s(t)$ more than the future raw data point $x(t+1)$.   From Eq.~(\ref{eq:exponential-smoothing-1}), we have
	\begin{align}
	&
	s(1) = \beta s(0) + (1 - \beta) x(1)
	\\
	&
	s(2) = \beta^2 s(0) + (1-\beta) [\beta x(1) + x(2)]
	\\[-4mm]
	%\vspace{-3mm}
	\vdots
	\nonumber
	\\[-6mm]
	&
	s(t) = \beta^t s(0) + (1-\beta) \sum_{i=1}^{t} \beta^{t-i} x(i)
	\ ,
	\label{eq:exponential-smoothing-2}
	\end{align}
	where the first term in Eq.~(\ref{eq:exponential-smoothing-2}) is called the bias, which is set by the initial condition:
	\begin{align}
	s(0) = x(0)
	\ , \text{ and as }
	t \rightarrow \infty
	\Rightarrow
	\beta^t x(0) \rightarrow 0
	\ .
	\label{eq:exponential-smoothing-3}
	\end{align}
	For finite time $t$, if the series started with $s(0) = 0 \ne x(0)$, then there is a need to correct the estimate for the non-zero bias $x(0) \ne 0$ (see Eq.~(\ref{eq:unified-2-Adam-2}) and Eq.~(\ref{eq:unified-2-Adam-4}) in the Adam algorithm below).
	The coefficients in the series in Eq.~(\ref{eq:exponential-smoothing-2}) are exponential functions, and thus the name ``exponential smoothing''.
	
	Eq.~(\ref{eq:exponential-smoothing-2}) is the discrete counterpart of the linear part of Volterra series in Eq.~(\ref{eq:linear-volterra-series}), used widely in neuroscientific modeling; see Remark~\ref{rm:volterra-exponential-smoothing}.  See also the ``small heavy sphere'' method or SGD with momentum Eq.~(\ref{eq:SGD-momentum-explicit}).  
	
	It should be noted, however, that for \emph{forecasting} (e.g., \cite{Hyndman.2018}), the following recursive equation, slightly different from Eq.~(\ref{eq:exponential-smoothing-1}), is used instead:
	\begin{align}
	s(t+1) = \beta s(t) + (1 - \beta) {\color{purple} x(t)}
	\ ,
	\text{ with }
	\beta \in [0,1)
	\ ,
	\label{eq:exponential-smoothing-forecasting}
	\end{align}
	where $x(t)$ (shown in purple) is used instead of $x(t+1)$, since if the data at $(t+1)$ were already known, there would be no need to forecast.

%{\color{red} HERE 2020.01.30}

%{\bf RMSProp.} 
\subsubsection{RMSProp: Root Mean Square Propagation}
\label{sc:rmsprop}
\label{para:rmsprop}
%(Root Mean Square Propagation.) 
Since
``\hyperref[para:adagrad]{AdaGrad} shrinks the learning rate according to the entire history of the squared gradient and may have made the learning rate too small before arriving at such a convex structure'',\footnote{
	See \cite{Goodfellow.2016}, p.~300.
} 
%
% CMES style rewriting
The authors of
\cite{Tieleman.2012}\footnote{
	Almost all authors, e.g., \cite{Bottou.2018:rd0001} \cite{Wilson.2018} \cite{Sun.2019}, attributed RMSProp to
	%
	% CMES style rewriting 
	Ref. \cite{Tieleman.2012}, except for Ref. \cite{Goodfellow.2016}, 
	% 
	% CMES style rewriting
%	who only referred to 
	where only
	Hinton's 2012 Coursera lecture was referred to.  Tieleman was Hinton's student; see the video and the lecture notes in \cite{Tieleman.2012}, where Tieleman's contribution was noted as unpublished.
	%
	% CMES style rewriting
	The authors of 
	\cite{Sun.2019} indicated that both RMSProp and AdaDelta (next section) were developed independently at about the same time to fix problems in AdaGrad.
} fixed the problem of continuing decay of the learning rate by introducing RMSProp\footnote{
	Neither \cite{Sun.2019}, nor \cite{Bottou.2018:rd0001}, nor \cite{Goodfellow.2016}, p.~299, provided the meaning of the acronym RMSProp, which stands for ``Root Mean Square Propagation''.  
} with the following functions for Algorithm~\ref{algo:unified-adaptive-learning-rate-2}:
\begin{align}
	&
	\phi_k (\bgradt_1 , \ldots, \bgradt_k) = \bgradt_k
	\ , 
	\text{ with }
  	% CMES style - 2022.06.01
	% same problem with \bId, which could be that \I had been defined as something else in the TSP style
	% so just code up boldsymbol I explicitly
	%
	% \chi_{\phi_k} = \bId \text{ (Identity)}
	\chi_{\phi_k} = \boldsymbol{I} \text{ (Identity)}
	\Rightarrow
	\bmom_k = \bgradt_k = \bmomc_k
	\ ,
	\label{eq:unified-2-RMSProp-1}
	\\
	&
	\bvar_k = \beta \bvar_{k-1} + (1 - \beta) (\bgradt_k)^2
	\text{ with } \beta \in [0,1)
	\text{ (element-wise square)}
	\ ,
	\label{eq:unified-2-RMSProp-2}
\end{align}
\begin{align}
	&
	\bvar_k = (1 - \beta) \sum_{i = 1}^{k} \beta^{k-i} \, (\bgradt_i)^2 =
	\psi_k (\bgradt_1 , \ldots, \bgradt_k)
	\ ,
	\label{eq:unified-2-RMSProp-3}
	\\
	&
  	% CMES style - 2022.06.01
	% same problem with \bId, which could be that \I had been defined as something else in the TSP style
	% so just code up boldsymbol I explicitly
	%
	% \chi_{\psi_k} = \bId
	\chi_{\psi_k} = \boldsymbol{I} 
	\text{ and } \delta = 10^{-6}
	\Rightarrow
	\bvarc_k = \bvar_k
	\text{ and }
	\blearn_k = \frac{\learn (k)}{\sqrt{\bvarc_k + \delta}}
	\text{ (element-wise operations)}
	\ ,
	\label{eq:unified-2-RMSProp-4}
\end{align}
where the running average of the squared gradients is given in Eq.~(\ref{eq:unified-2-RMSProp-2}) for efficient coding, and in Eq.~(\ref{eq:unified-2-RMSProp-3}) in fully expanded form as a series with exponential coefficients $\beta^{k-i}$, for $i=1,\ldots, k$.
Eq.~(\ref{eq:unified-2-RMSProp-2}) is the exact counterpart of exponential smoothing recurrence relation in Eq.~(\ref{eq:exponential-smoothing-1}), and Eq.~(\ref{eq:unified-2-RMSProp-3}) has its counterpart in Eq.~(\ref{eq:exponential-smoothing-2}) if $\bgradt_0 = \bgradt_0^2 = 0$; see Section~\ref{sc:exponential-smoothing} on forecasting time series and exponential smoothing.

Figure~\ref{fig:Adam-converge} shows the convergence of some adaptive learning-rate algorithms: \hyperref[para:adagrad]{AdaGrad}, \hyperref[para:rmsprop]{RMSProp}, 
\hyperref[sc:SGD-momentum]{SGDNesterov}, \hyperref[para:adadelta]{AdaDelta}, \hyperref[para:adam1]{Adam}.

\hyperref[para:rmsprop]{RMSProp} still depends on a global learning rate $\epsilon(k) = \epsilon_0$ constant, a tuning hyperparameter.  Even though \hyperref[para:rmsprop]{RMSProp} was one of the go-to algorithms for machine learning, the pitfalls of \hyperref[para:rmsprop]{RMSProp}, along with other adaptive learning-rate algorithms, were revealed in \cite{Wilson.2018}.

%{\color{red} HERE 2020.01.30}

%{\bf AdaDelta.} 
\subsubsection{AdaDelta: Adaptive Delta (parameter increment)}
\label{sc:adadelta}
\label{para:adadelta}
The name ``AdaDelta'' comes from the adaptive parameter increment $\Delta \bparam$ in Eq.~(\ref{eq:adadelta-param-increment-update}).
In parallel and independently, 
%
% CMES style rewriting
%\cite{Zeiler.2012} proposed 
AdaDelta 
proposed in \cite{Zeiler.2012}
%that 
not only fixed the problem of continuing decaying learning rate of \hyperref[para:adagrad]{AdaGrad}, but also removed the need for a global learning rate $\epsilon_0$, which \hyperref[para:rmsprop]{RMSProp} still used.
By accumulating the squares of the parameter increments, i.e., $(\Delta \bparam_{k})^2$, AdaDelta would fit in the unified framework of Algorithm~\ref{algo:unified-adaptive-learning-rate-2} if the symbol $\bmom_k$ in Eq.~(\ref{eq:momentum-0}) were interpreted as the accumulated 2nd moment $(\Delta \bparam_{k})^2$, per Remark~\ref{rm:case-adadelta}.  
	
%
% CMES style rewriting	
%\cite{Zeiler.2012} made 
%The following observation on the 
The weaknesses of \hyperref[para:adagrad]{AdaGrad} was observed in \cite{Zeiler.2012}: ``Since the magnitudes of gradients are factored out in \hyperref[para:adagrad]{AdaGrad}, this method can be sensitive to initial conditions of the parameters and the corresponding gradients. If the initial gradients are large, the learning rates will be low for the remainder of training. This can be combatted by increasing the global learning rate, making the \hyperref[para:adagrad]{AdaGrad} method sensitive to the choice of learning rate. Also, due to the continual accumulation of squared gradients in the denominator, the learning rate will continue to decrease throughout training, eventually decreasing to zero and stopping training completely.''

%
% CMES style rewriting
%\cite{Zeiler.2012} then introduced 
AdaDelta was then introduced in \cite{Zeiler.2012} as an improvement over AdaGrad with two goals in mind: (1)  to avoid the continuing decay of the learning rate, and (2) to avoid having to specify $\learn (k)$, called the ``global learning rate'', as a constant.  Instead of summing past squared gradients over a finite-size window, which is not efficient in coding, 
%
% CMES style rewriting
%\cite{Zeiler.2012} suggested to use 
exponential smoothing was employed in \cite{Zeiler.2012} for both the squared gradients $(\bgradt_k)^2$ and for the squared increments $(\Delta \bparam_k)^2$, with the increment used in the update $\bparam_{k+1} = \bparam_{k} + \Delta \bparam_k$, by choosing  the following functions for Algorithm~\ref{algo:unified-adaptive-learning-rate-2}:
\begin{align}
	&
	\bvar_k = \beta \bvar_{k-1} + (1 - \beta) (\bgradt_k)^2
	\text{ with } \beta \in [0,1)
	\text{ (element-wise square)}
	\\
	&
	\bvar_k = (1 - \beta) \sum_{i = 1}^{k} \beta^{k-i} \, (\bgradt_i)^2 =
	\psi_k (\bgradt_1 , \ldots, \bgradt_k)
	\ ,
	\\
	&
	\bmom_k = \beta \bmom_{k-1} + (1 - \beta) (\Delta \bparam_k)^2
	\text{ (element-wise square)}
	\\
	&
	\bmom_k = (1 - \beta) \sum_{i = 1}^{k} \beta^{k-i} \, (\Delta \bparam_i)^2
	\ .
	\label{eq:unified-2-AdaDelta}
\end{align}
Thus, exponential smoothing (Section~\ref{sc:exponential-smoothing}) is used for two second-moment series: $\{(\bgradt_i)^2 , \, i = 1,2,\ldots\}$ and $\{(\Delta \bparam_i)^2 , \, i = 1,2,\ldots\}$.
The update of the network parameters from $\bparam_{k}$ to $\bparam_{k+1}$ is carried out as follows:
\begin{align}
	\Delta \bparam_{k}
	=
	-
	\frac{1}{\sqrt{\bvarc_k} + \delta}
	\,
	\bmomc_k 
	\ ,
	\text{ with }
	\bmomc_k = [\sqrt{\bmom_{k-1}} + \delta] \, \bgradt_k 
	\text{ and }
	\bparam_{k+1} = \bparam_{k} + \Delta \bparam_{k}
	\ ,
	\label{eq:adadelta-param-increment-update}
\end{align}
where $\epsilon(k)$ in Eq.~(\ref{eq:unified-2-RMSProp-4}) is fixed to 1 in Eq.~(\ref{eq:adadelta-param-increment-update}), eliminating the hyperparameter $\epsilon(k)$.  Another nice feature of AdaDelta is the consistency of units (physical dimensions), in the sense that the fraction factor of the gradient $\bgradt_k$ in Eq.~(\ref{eq:adadelta-param-increment-update}) has the unit of step length (learning rate):\footnote{
	In spite of the nice features in \hyperref[para:adadelta]{AdaDelta}, neither \cite{Goodfellow.2016}, nor \cite{Bottou.2018:rd0001}, nor \cite{Sun.2019},
	%
	% CMES style rewriting 
%	reviewed 
	had a review of
	\hyperref[para:adadelta]{AdaDelta}, except for citing \cite{Zeiler.2012}, even though 
	%
	% CMES style rewriting
	the authors of
	\cite{Goodfellow.2016}, p.~302, wrote: ``While the results suggest that the family of algorithms with adaptive learning rates (represented by \hyperref[para:rmsprop]{RMSProp} and \hyperref[para:adadelta]{AdaDelta}) performed fairly robustly, no single best algorithm has emerged,'' and ``Currently, the most popular optimization algorithms actively in use include \hyperref[sc:generic-SGD]{SGD}, \hyperref[sc:SGD-momentum]{SGD with momentum}, \hyperref[para:rmsprop]{RMSProp}, \hyperref[para:rmsprop]{RMSProp} with momentum, \hyperref[para:adadelta]{AdaDelta}, and Adam.''
	%
	% CMES style rewriting
	The authors of
	\cite{Bottou.2018:rd0001}, p.~286, did not follow the historical development, briefly reviewed \hyperref[para:rmsprop]{RMSProp}, then cited in passing references for \hyperref[para:adadelta]{AdaDelta} and \hyperref[para:adam1]{Adam}, and then mentioned the ``popular \hyperref[para:adagrad]{AdaGrad} algorithm'' as a ``member of this family''; readers would lose sight of the gradual progress made starting from \hyperref[para:adagrad]{AdaGrad}, to \hyperref[para:rmsprop]{RMSProp}, \hyperref[para:adadelta]{AdaDelta}, then \hyperref[para:adam1]{Adam}, and to the recent \hyperref[para:adamw]{AdamW} in \cite{Loshchilov.2019}, among others, alternating with pitfalls revealed and subsequent fixes, and then more pitfalls revealed and more fixes. 
}
\begin{align}
	\left[ \frac{\sqrt{\bmom_{k-1}} + \delta}{\sqrt{\bvarc_k} + \delta} \right]
	=
	\left[ \frac{\Delta \bparam}{\bgradt} \right]
	=
	[\epsilon]
	\ ,
	\label{eq:adadelta-units}
\end{align}
where the enclosing square brackets denote units (physical dimensions), but that was not the case in Eq.~(\ref{eq:unified-2-RMSProp-4}) of RMSProp:
\begin{align}
	[\blearn_k]
	=
	\left[ \frac{\learn (k)}{\sqrt{\bvarc_k  + \delta}} \right]
	\ne
	[\epsilon]
	\ .
	\label{eq:rmsprop-units}
\end{align}

Figure~\ref{fig:Adam-converge} shows the convergence of some adaptive learning-rate algorithms: \hyperref[para:adagrad]{AdaGrad}, \hyperref[para:rmsprop]{RMSProp}, 
\hyperref[sc:SGD-momentum]{SGDNesterov}, \hyperref[para:adadelta]{AdaDelta}, \hyperref[para:adam1]{Adam}.

Despite this progress, \hyperref[para:adadelta]{AdaDelta} and \hyperref[para:rmsprop]{RMSProp}, along with other adaptive learning-rate algorithms, shared the same pitfalls as revealed in \cite{Wilson.2018}.

%{\bf Adam.} 
\subsubsection{Adam: Adaptive moments}
\label{sc:adam1}
\label{para:adam1}
%[Adaptive moments, i.e., both 1st moment Eq.~(\ref{eq:unified-2-Adam-1}) and 2nd moment Eq.~(\ref{eq:unified-2-Adam-3}).]
Both both 1st moment Eq.~(\ref{eq:unified-2-Adam-1}) and 2nd moment Eq.~(\ref{eq:unified-2-Adam-3}) are adaptive.
To avoid possible large step sizes and non-convergence of \hyperref[para:rmsprop]{RMSProp}, 
%
% CMES style rewriting
%\cite{Kingma.2014} selected 
the following functions were selected for Algorithm~\ref{algo:unified-adaptive-learning-rate-2} \cite{Kingma.2014}:
\begin{align}
	&
	\bmom_k = \beta_1 \bmom_{k-1} + (1 - \beta_1) \bgradt_k
	\ ,
	\text{ with } \beta_1 \in [0,1)
	\text{ and } \bmom_0 = 0
	\ ,
	\label{eq:unified-2-Adam-1}
	\\
	&
	\bmomc_k = \frac{1}{1 - (\beta_1)^k} \, \bmom_k 
	\text{ (bias correction)}
	\ ,
	\label{eq:unified-2-Adam-2}
	\\
	&
	\bvar_k = \beta_2 \bvar_{k-1} + (1 - \beta_2) (\bgradt_k)^2
	\text{ (element-wise square)}
	\ ,
	\text{ with } \beta_2 \in [0,1)
	\text{ and } \bvar_0 = 0
	\ ,
	\label{eq:unified-2-Adam-3}
	\\
	&
	\bvarc_k = \frac{1}{1 - (\beta_2)^k} \, \bvar_k
	\text{ (bias correction)}
	\ ,
	\label{eq:unified-2-Adam-4}
	\\
	&
	\blearn_k = \frac{\learn (k)}{\sqrt{\bvarc_k}  + \delta}
	\text{ (element-wise operations)}
	\ .
	\label{eq:unified-2-Adam-5}
\end{align}
with the following recommended values of the parameters:
\begin{align}
	\beta_1 = 0.9
	\ , 
	\beta_2 = 0.999
	\ ,
	\epsilon_0 = 0.001
	\ ,
	\delta = 10^{-8}
	\ .
	\label{eq:adam-recommended-params}
\end{align}

\begin{rem}
	\label{rm:rmsprop-under-adam}
	{\rm 
		\hyperref[para:rmsprop]{RMSProp} is a particular case of Adam, when $\beta_1 = 0$,  together with the absence of the bias-corrected 1st moment Eq.~(\ref{eq:unified-2-Adam-2}) and bias-corrected 2nd moment Eq.~(\ref{eq:unified-2-Adam-4}).  Moreover, the get \hyperref[para:rmsprop]{RMSProp} from Adam, choose the constant $\delta$ and the learning rate $\blearn_k$ as in Eq.~(\ref{eq:unified-2-RMSProp-4}), instead of Eq.~(\ref{eq:unified-2-Adam-5}) above, but this choice is a minor point, since either choice should be fine.  On the other hand, 
		%
		% CMES style rewriting
%		\cite{Reddi.2019} indicated 
		for deep-learning applications, having the 1st moment (or momentum), and thus requiring $\beta > 0$, would be useful to ``significantly boost the performance'' \cite{Reddi.2019}, and hence an advantage of Adam over \hyperref[para:rmsprop]{RMSProp}.
	}
	$\hfill\blacksquare$
\end{rem}

It follows from Eq.~(\ref{eq:exponential-smoothing-2}) in Section~\ref{sc:exponential-smoothing} on exponential smoothing of time series that the recurrence relation for gradients (1st moment) in Eq.~(\ref{eq:unified-2-Adam-1}) leads to the following series:
\begin{align}
	\bmom_k = (1-\beta_1) \sum_{i=1}^{k} \beta^{k-i} \, \bgradt_i
	\ ,
	\label{eq:adam-exponential-series}
\end{align}
since $\bmom_0 = 0$.  Taking the expectation, as defined in Eq.~(\ref{eq:expectation}), on both sides of Eq.~(\ref{eq:adam-exponential-series}) yields
\begin{align}
	\xpc[\bmom_k] 
	&
	= 
	(1-\beta_1) \sum_{i=1}^{k} \beta^{k-i} \, \xpc[\bgradt_i]
	=
	\xpc[\bgradt_i] \cdot (1-\beta_1) \sum_{i=1}^{k} \beta^{k-i} 
	+ \drift	=
	\xpc[\bgradt_i] \cdot [1-(\beta_1)^2] + \drift
	\ ,
	\label{eq:bias-correction-1}
\end{align}
where $\drift$ is the drift from the expected value, with $\drift = 0$ for stationary random processes.\footnote{
	A random process is stationary when its mean and standard deviation stay constant over time.
}
For non-stationary processes, 
%
% CMES style rewriting
%\cite{Kingma.2014} suggested to keep 
it was suggested in \cite{Kingma.2014} to keep
$\drift$ small by choosing small $\beta_1$ so only past gradients close to the present iteration $k$ would contribute, so to keep any change in the mean and standard deviation in subsequent iterations small. 
By dividing both sides by $[1 - (\beta_1)^2]$, the bias-corrected 1st moment $\bmomc_k$ shown in Eq.~(\ref{eq:unified-2-Adam-2}) is obtained, showing that the expected value of $\bmomc_k$ is the same as the expected value of the gradient $\bgradt_i$ plus a small number, which could be zero for stationary processes:
\begin{align}
	\xpc[\bmomc_k]
	=
	\xpc \left[ \frac{\bmom_k}{1-(\beta_1)^2} \right] 
	=
	\xpc[\bgradt_i] + \frac{\drift}{1-(\beta_1)^2}
	\ .
	\label{eq:bias-correction-2}
\end{align}
The argument to obtain the bias-corrected 2nd moment $\bvarc_k$ in Eq.~(\ref{eq:unified-2-Adam-4}) is of course the same.

The authors of \cite{Kingma.2014} pointed out the lack of bias correction in \hyperref[para:rmsprop]{RMSProp} (Remark~\ref{rm:rmsprop-under-adam}), leading to ``very large step sizes and often divergence'', and provided numerical experiment results to support their point.

Figure~\ref{fig:Adam-converge} shows the convergence of some adaptive learning-rate algorithms: \hyperref[para:adagrad]{AdaGrad}, \hyperref[para:rmsprop]{RMSProp}, 
\hyperref[sc:SGD-momentum]{SGDNesterov}, \hyperref[para:adadelta]{AdaDelta}, \hyperref[para:adam1]{Adam}.  Their results show the superior performance of \hyperref[para:adam1]{Adam} compared to other adaptive learning-rate algorithms.
See Figure~\ref{fig:Haibe-Kains-irreproducibility} in Section~\ref{sc:irreproducibility} on ``Lack of transparency and irreproducibility of results'' in recent deep-learning papers.

%{\bf AMSGrad.}
\subsubsection{AMSGrad: Adaptive Moment Smoothed Gradient}
\label{sc:amsgrad}
\label{para:amsgrad}
%(Adaptive Moment Smoothed Gradient)
%
% CMES style rewriting
%\cite{Reddi.2019}
The authors of \cite{Reddi.2019} stated that \hyperref[para:adam1]{Adam} (and other variants such as \hyperref[para:rmsprop]{RMSProp}, \hyperref[para:adadelta]{AdaDelta}, Nadam) ``failed to converge to an optimal solution (or a critical point in non-convex settings)'' in many applications with large output spaces, and constructed a simple convex optimization for which \hyperref[para:adam1]{Adam} did not converge to the optimal solution.  

An earlier version of \cite{Reddi.2019} received one of the three Best Papers at the ICLR 2018\footnote{
	Sixth International Conference on Learning Representations (\href{https://iclr.cc/Conferences/2018}{Website}).
} conference, in which it was suggested to fix the problem by endowing the mentioned algorithms with ``long-term memory'' of past gradients, and by selecting the following functions for Algorithm~\ref{algo:unified-adaptive-learning-rate-2}:
\begin{align}
	&
	\bmom_k = \beta_{1k} \bmom_{k-1} + (1 - \beta_{1k}) \bgradt_k 
	= \phi_k
	\ ,
	\text{ with }
	\bmom_0 = 0
	\ ,
	\label{eq:unified-2-AMSGrad-1}
	\\
	&
	\beta_{1k} = \beta_{11} \in [0,1) 
	\ ,
	\text{ or }
	\beta_{1k} = \beta_{11} \lambda^{k-1}
	\text{ with }
	\lambda \in (0,1)
	\ ,
	\text{ or }
	\beta_{1k} = \frac{\beta_{11}}{k}
	\ ,
	\label{eq:unified-2-AMSGrad-1b}
	\\
	&
	\bmomc_k = \bmom_k 
	\text{ (no bias correction)}
	\ ,
	\label{eq:unified-2-AMSGrad-2}
	\\
	&
	\bvar_k = \beta_2 \bvar_{k-1} + (1 - \beta_2) (\bgradt_k)^2
	= \psi_k
	\text{ (element-wise square)}
	\ ,
	\text{ with }
	\bvar_0 = 0
	\ ,
	\label{eq:unified-2-AMSGrad-3}
	\\
	&
	\beta_{ 2 } \in [0,1)
	\text{ and }
	\frac{\beta_{11}}{\sqrt{\beta_{ 2 }}} < 1
	\Leftrightarrow
	\beta_2 > (\beta_1)^2
	\ ,
	\label{eq:unified-2-AMSGrad-3b}
	\\
	&
	\bvarc_k = \max ( \bvarc_{k-1} , \bvar_k )
	\text{ (element-wise max, ``long-term memory'')}
	\ ,
	\text{ and } \delta \text{ not used}
	\ ,
	\label{eq:unified-2-AMSGrad-4}
\end{align}
%
% CMES style rewriting
%\cite{Reddi.2019} did not define what
The parameter $\lambda$ was not defined in 
Corollary 1 of \cite{Reddi.2019}; such omission could create some difficulty for first-time leaners.  It has to be deduced from reading the Corollary that $\lambda \in (0,1)$.
For the step-length (or size) schedule $\epsilon (k)$, even though
%
% CMES style rewriting 
%\cite{Reddi.2019} considered 
only Eq.~(\ref{eq:learning-rate-schedule-2}) was considered in \cite{Reddi.2019} for the convergence proofs,  Eq.~(\ref{eq:learning-rate-schedule}) (which includes $\epsilon(k) = \epsilon_0 =$ constant) and Eq.~(\ref{eq:learning-rate-schedule-3}) could also be used.\footnote{
	%
	% CMES style rewriting
	The authors of 
	\cite{Reddi.2019} distinguished the step size $\epsilon(k)$ from the scaled step size $\blearn_k$ in Eq.~(\ref{eq:learning-rate-schedule}) or Eq.~(\ref{eq:learning-rate-schedule-2}), which were called learning rate. 
} 

First-time learners in this field could be overwhelmed by complex-looking equations in this kind of paper, so it would be helpful to elucidate some key results that led to the above expressions, particularly for $\beta_{1k}$, which can be a constant or a function of the iteration number $k$, in Eq.~(\ref{eq:unified-2-AMSGrad-1b}).  

%
% CMES style rewriting
%\cite{Reddi.2019} 
It was 
stated in \cite{Reddi.2019} that ``one typically uses a constant $\beta_{1k}$ in practice (although, the proof requires a decreasing schedule for proving convergence of the algorithm),'' and hence the first choice $\beta_{1k} = \beta_{11} \in (0,1)$.  

The second choice $\beta_{1k} = \beta_{11} \lambda^{k-1}$, with $\lambda \in (0,1)$ and $\epsilon(k) = \epsilon_0 / \sqrt{k}$ as in Eq.~(\ref{eq:learning-rate-schedule-2}), was the result stated in Corollary 1 in \cite{Reddi.2019}, but without proof. We fill this gap here to explain this unusual expression for $\beta_{1k}$.  Only the second term on the right-hand side of the inequality in Theorem 4 needs to be bounded by this choice of $\beta_{1k}$, and is written in our notation as:\footnote{
	To write this term in the notation used in\cite{Reddi.2019}, Theorem 4 and Corollary 1, simply make the following changes in notation: $k \rightarrow t$, $k_{max} \rightarrow T$, $\Tparam \rightarrow d$, $V \rightarrow v$, $\epsilon(k) \rightarrow \alpha_t$.
}
\begin{align}
	\frac{D_\infty^2}{(1-\beta_1)^2}
	\sum_{k=1}^{k_{max}} \sum_{i=1}^{\Tparam}
	\frac{\beta_{1k} \varc_{k,i}^{1/2}}{\epsilon(k)}
	\ ,
	\label{eq:reddi-2nd-term-theorem-4}
\end{align}
where $k_{max}$ is the maximum number of iterations in the \ding{173} {\bf for} loop in Algorithm~\ref{algo:unified-adaptive-learning-rate-2}, and $\Tparam$ is the total number of network parameters defined in Eq.~(\ref{eq:totalParams}).
The factor $D_\infty^2 / (1 - \beta_1)^2$ is a constant, and the following bound on the component $\varc_{k,i}^{1/2}$ is a consequence of an assumption in Theorem 4 in \cite{Reddi.2019}:
\begin{align}
	\parallel \nabla J_k \parallel_\infty \le G_\infty 
	\text{ for all } k \in \{1, \ldots , k_{max}\}
	\Rightarrow
	\varc_{k,i}^{1/2} \le G_\infty 
	\text{ for any } i \in \{1, \ldots , \Tparam\}
	\ ,
	\label{eq:bound-V-compo}
\end{align}
where $\parallel \cdot \parallel_\infty$ is the infinity (max) norm, which is clearly consistent with the use of element-wise maximum components for ``long-term memory'' in Eq.~(\ref{eq:unified-2-AMSGrad-4}). Intuitively, $\varc_{k,i}$ has the unit of gradient squared, and thus $\varc_{k,i}^{1/2}$ has the unit of gradient.  Since $G_\infty$ is the upperbound of the maximum component of the gradient\footnote{
	 The notation ``$G$'' is clearly mnemonic for ``gradient'', and the uppercase is used to designate upperbound.
} $\nabla J_k$ of the cost function $J_k$ at any iteration $k$, it follows that $\varc_{k,i}^{1/2} \le G_\infty$.
We refer to \cite{Phuong.2019}, p.~10, Lemma 4.2, for a formal proof of the inequality in Eq.~(\ref{eq:bound-V-compo}).  Once $\varc_{k,i}^{1/2}$ is replaced by $G_\infty^2$, which is then pulled out as a common factor in expression ($\ref{eq:reddi-2nd-term-theorem-4}$), where upon substituting $\beta_{1k} = \beta_{11} \lambda^{k-1}$, with $\beta_{11} = \beta_1$, and $\epsilon(k) = \epsilon_0 / \sqrt{k}$, we obtain:   
\begin{align}
	(\ref{eq:reddi-2nd-term-theorem-4}) 
	&
	\le 
	\frac{D_\infty^2 G_\infty}{(1-\beta_1)^2}
	\sum_{k=1}^{k_{max}} \sum_{i=1}^{\Tparam}
	\frac{\beta_{1} \sqrt{k} \lambda^{k-1}}{\epsilon_0}
	= 
	\frac{D_\infty^2 G_\infty \beta_{1} \Tparam}{(1-\beta_1)^2 \epsilon_0}
	\sum_{k=1}^{k_{max}}
	\sqrt{k} \lambda^{k-1}
	\\
	&
	\le 
	\frac{D_\infty^2 G_\infty \beta_{1} \Tparam}{(1-\beta_1)^2 \epsilon_0}
	\sum_{k=1}^{k_{max}}
	k \lambda^{k-1}
	=
	\frac{D_\infty^2 G_\infty {\color{blue} \beta_{1}} {\color{purple} \Tparam}}{(1-\beta_1)^2 {\color{purple}\epsilon_0}}
	\frac{1}{(1 - \lambda)^2}
	\ ,
	\label{eq:reddi-2nd-term-theorem-4-bound}
\end{align}
where the following series expansion had been used:\footnote{
	See also \cite{Phuong.2019}, p.~4, Lemma 2.4.
}
\begin{align}
	\frac{1}{(1 - \lambda)^2}
	=
	\sum_{k=1}^{k_{max}}
	k \lambda^{k-1}
	\ .
	\label{eq:series-1-(1-x)^2}
\end{align}
Comparing the bound on the right-hand side of (\ref{eq:reddi-2nd-term-theorem-4-bound}) to the corresponding bound shown in \cite{Reddi.2019}, Corollary 1, second term, it can be seen that two factors, ${\color{black} \Tparam}$ and ${\color{black}\epsilon_0}$ (in purple), were missing in the numerator and in the denominator, respectively.  In addition, there should be no factor ${\color{black} \beta_{1}}$ (in blue) as pointed out 
%
% CMES style rewriting
%by
in 
\cite{Phuong.2019} in their correction of the proof 
%by 
in
\cite{Reddi.2019}. 

On the other hand, there were some slight errors in the theorem statements and in the proofs in \cite{Reddi.2019} that were corrected 
%
% CMES style rewriting
%by
in 
\cite{Phuong.2019}, 
%who 
whose authors
did a good job of not skipping any mathematical details that rendered the understanding and the verification of the proofs obscure and time consuming.  It is then recommended to read \cite{Reddi.2019} to get a general idea on the main convergence results of AMSGrad, then read \cite{Phuong.2019} for the details, together with their variant of AMSGrad called \hyperref[para:adamx]{AdamX}. 

%
% CMES style rewriting
The authors of 
\cite{Bock.2018}, like those of \cite{Reddi.2019}, pointed out errors in the convergence proof 
%of 
in
\cite{Kingma.2014}, and proposed a fix to this proof, but did not suggest any new variant of \hyperref[para:adam1]{Adam}.

\begin{figure}[h]
	\centering
	\includegraphics[width=1.0\linewidth]{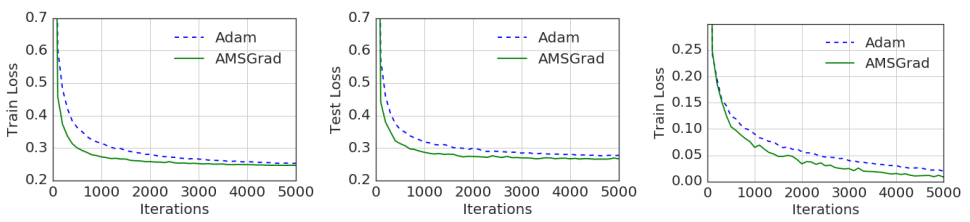}
	\caption{
		\emph{AMSGrad vs Adam, numerical examples} (Sections~\ref{sc:training-valication-test}, \ref{sc:amsgrad}).  The MNIST dataset is used.  The first two figures on the left were the results of using logistic regression (network with one layer with logistic sigmoid activation function), whereas the figure on the right is by using a neural network with three layers (input layer, hidden layer, output layer).  The cost function decreased faster for AMSGrad compared to that of Adam.  For logistic regression, the difference between the two cost values also decreased with the iteration number, and became very small at iteration 5000.  For the three-layer neural network, the cost difference between AMSGrad and Adam stayed more or less constant, as the cost went down to more than one tenth of the initial cost at about 0.3, and after 5000 iterations, the AMSGrad cost ($\approx 0.01$) was about 50\% of the Adam cost ($\approx 0.02$).  See \cite{Reddi.2019}.
		{\footnotesize (Figure reproduced with permission of the authors.)}
		%{\color{red} ASK PERMISSION.}
	}
	\label{fig:amsgrad-examples}
\end{figure}

In the two large numerical experiments on the MNIST dataset in Figure~\ref{fig:amsgrad-examples},\footnote{
	See a description of the NMIST dataset in Section~\ref{sc:vanish-grad} on ``Vanishing and exploding gradients''.  For the difference between logistic regression and neural network, see, e.g., \cite{Dreiseitl.2002}, Raschka, ``Machine Learning FAQ: What is the relation between Logistic Regression and Neural Networks and when to use which?'' \href{https://sebastianraschka.com/faq/docs/logisticregr-neuralnet.html}{Original website}, \href{http://web.archive.org/web/20161204204235/https://sebastianraschka.com/faq/docs/logisticregr-neuralnet.html}{Internet archive}.  See also \cite{Goodfellow.2016}, p.~200, Figure6.8b, for the computational graph of Logistic Regression (one-layer network).
} 
%
% CMES style rewriting
the authors of 
\cite{Reddi.2019} used contant $\beta_{1k} = \beta_{1} = 0.9$, with $\beta_{ 2 } \in \{0.99, 0.999\}$; they chose the step size schedule $\learn(k) = \learn_0 / \sqrt{k}$ in the logistic regression experiment, and constant step size $\learn(k) = \learn_0$ in a network with three layers (input, hidden, output).  There was no single set of optimal parameters, which appeared to be problem dependent. 

%
% CMES style rewriting
The authors of 
\cite{Reddi.2019} also did not provide any numerical example with $\beta_{1k} = \beta_1 \lambda^{k-1}$; such numerical examples can be found, however, in \cite{Phuong.2019} in connection with \hyperref[para:adamx]{AdamX} below.

Unfortunately, when comparing \hyperref[para:amsgrad]{AMSGrad} to \hyperref[para:adam1]{Adam} and \hyperref[para:adamw]{AdamW} (further below), 
%
% CMES style rewriting
%\cite{Gugger.2018} remarked 
it was remarked in \cite{Gugger.2018}
that \hyperref[para:amsgrad]{AMSGrad} generated ``a lot of noise for nothing'', meaning AMSGrad did not live up to its potential and best-paper award when tested on ``real-life problems''.

%{\bf AdamX.}
\subsubsection{AdamX and Nostalgic Adam}
\label{sc:adamx}
\label{para:adamx}
%
% CMES style rewriting
%\cite{Phuong.2019},
{\bf AdamX.} 
The authors of 
\cite{Phuong.2019}, already mentioned above in connection to errors in the proofs 
%by
in 
\cite{Reddi.2019} for \hyperref[para:amsgrad]{AMSGrad}, also pointed out errors in the proofs by \cite{Kingma.2014} (Theorem 10.5), \cite{Bock.2018} (Theorem 4.4), and by others, and suggested a fix for these proofs, and a new variant of \hyperref[para:amsgrad]{AMSGrad} called AdamX.

%
% CMES style rewriting
%The work by 
Reference
\cite{Phuong.2019} is more convenient to read, compared to \cite{Reddi.2019}, as 
%
% CMES style rewriting
%they 
the authors
provided all mathematical details for the proofs, without skipping important details. 

%
% CMES style rewriting
%\cite{Phuong.2019} proposed a
A slight change to Eq.~(\ref{eq:unified-2-AMSGrad-4}) was proposed in \cite{Phuong.2019} as follows:
\begin{align}
	\bvarc_1 = \bvar_1
	\ ,
	\text{ and }
	\bvarc_k 
	= 
	\max \left( \frac{ (1 - \beta_{1,k})^2 }{ (1 - \beta_{1, k-1})^2 } \bvarc_{k-1} , \bvar_k \right)
	\text{ for }
	K \ge 2
	\ ,
	\label{eq:unified-2-AdamX}
\end{align}

In addition, 
%
% CMES style rewriting
%\cite{Phuong.2019} provided 
numerical examples were provided in \cite{Phuong.2019} with $\beta_{1k} = \beta_1 \lambda^{k-1}$, $\beta_1 = 0.9$, $\lambda = 0.001$, $\beta_2 = 0.999$, and $\delta = 10^{-8}$ in Eq.~(\ref{eq:adaptive-learning-rate}), even though 
%their 
the
pseudocode did not use $\delta$ (or set $\delta = 0$).  
%They 
The authors of \cite{Phuong.2019}
showed that both \hyperref[para:amsgrad]{AMSGrad} and AdamX converged with similar results, thus supporting their theoretical investigation, in particular, correcting the errors in the proofs of \cite{Reddi.2019}.

{\bf Nostalgic Adam.}
\label{para:nosadam}
%
% CMES rewriting
The authors of 
\cite{Huang.2019} also fixed the non-convergence of Adam by introducing ``long-term memory'' to the second-moment of the gradient estimates, similar to the work 
%of 
in
\cite{Reddi.2019} on \hyperref[para:amsgrad]{AMSGrad} and 
%of 
in
\cite{Phuong.2019} on \hyperref[para:adamx]{AdamX}.

There are many more variants of \hyperref[para:adam1]{Adam}.  But how are \hyperref[para:adam1]{Adam} and its variants compared to good old SGD with new \hyperref[sc:add-on-tricks]{add-on tricks} ? (See the end of Section~\ref{sc:generic-SGD})

\begin{figure}[h]
	\centering
	\includegraphics[width=0.6\linewidth]{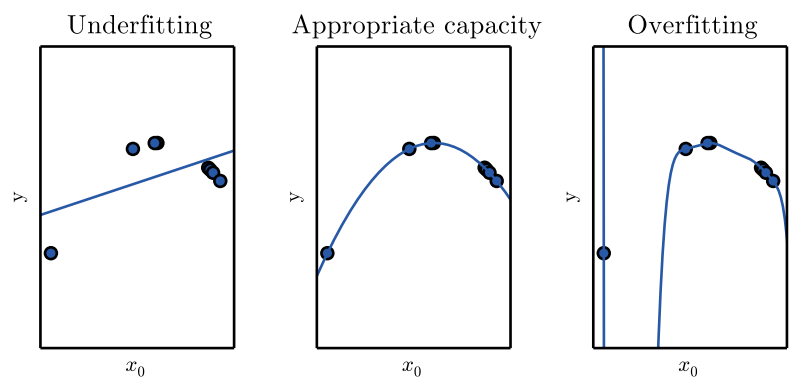}
	\caption{
		\emph{Overfitting} (Section~\ref{sc:adam-criticism}, \ref{para:adamw}).  \emph{Left:} Underfitting with 1st-order polynomial.  Middle: Appropriate fitting with 2nd-order polynomial.  \emph{Right:} Overfitting with 9th-order polynomial.    See \cite{Goodfellow.2016}, p.~110, Figure5.2.
		{\footnotesize (Figure reproduced with permission of the authors.)}
		%{\color{red} ASK PERMISSION.}
	}
	\label{fig:overfit}
\end{figure}

%{\bf Criticism of adaptive methods.}  
\subsubsection{Criticism of adaptive methods, resurgence of SGD}
\label{sc:adam-criticism}
\label{para:adam-criticism}
Yet, despite the claim that \hyperref[para:rmsprop]{RMSProp} is ``currently one of the go-to optimization methods being employed routinely by deep learning practitioners,'' and that ``currently, the most popular optimization algorithms actively in use include \hyperref[sc:SGD-momentum]{SGD}, \hyperref[sc:SGD-momentum]{SGD with momentum}, \hyperref[para:rmsprop]{RMSProp}, RMSProp with momentum, \hyperref[para:adadelta]{AdaDelta}, and \hyperref[para:adam1]{Adam}'',\footnote{
	See \cite{Goodfellow.2016}, pp.~301-302.
}  
%
% CMES style rewriting
%\vphantom{\cite{Wilson.2018}}\cite{Wilson.2018} showed, 
the authors of \cite{Wilson.2018}
through their numerical experiments, that adaptivity can overfit (Figure~\ref{fig:overfit}), and that standard SGD with step-size tuning performed better than adaptive learning-rate algorithms such as \hyperref[para:adagrad]{AdaGrad}, \hyperref[para:rmsprop]{RMSProp}, and \hyperref[para:adam1]{Adam}.  The total number of parameters, $\Tparam$ in Eq.~(\ref{eq:totalParams}), in deep networks could easily exceed 25 times the number of output targets $m$ (Figure~\ref{fig:network3b}), i.e.,\footnote{
	See \cite{Wilson.2018}, p.~4, Section 3.3 ``Adaptivity can overfit''.
} 
\begin{align}
	\Tparam \ge 25 m
	\ ,
	\label{eq:total-params}
\end{align}
making it prone to overfit without employing special techniques such as regularization or weight decay (see \hyperref[para:adamw]{AdamW} below).

\begin{figure}[h]
	\centering
	\includegraphics[width=0.9\linewidth]{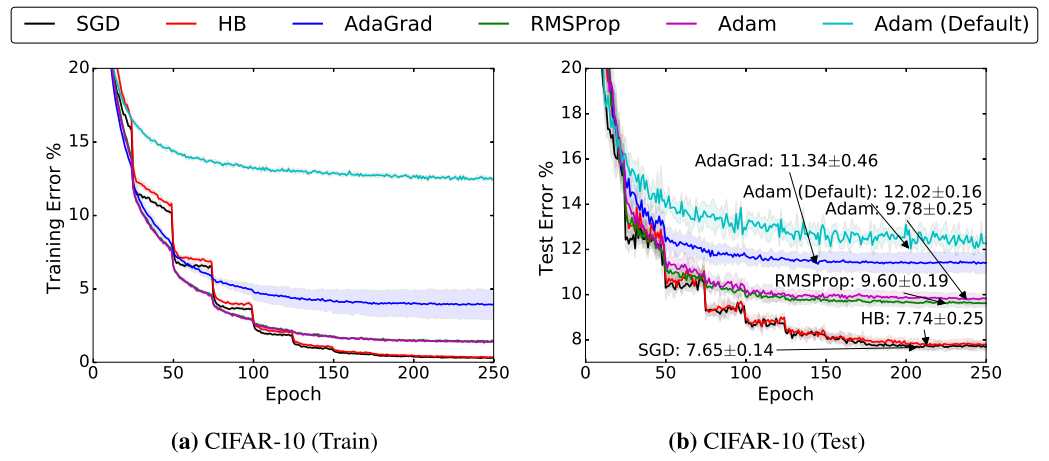}	
	\caption{
		\emph{\hyperref[sc:generic-SGD]{Standard SGD} and \hyperref[sc:SGD-momentum]{SGD with momentum} vs \hyperref[para:adagrad]{AdaGrad}, \hyperref[para:rmsprop]{RMSProp}, \hyperref[para:adam1]{Adam} on CIFAR-10 dataset} (Sections~\ref{sc:training-valication-test}, \ref{sc:SGD-momentum}, \ref{sc:adam-criticism}).   
		From \cite{Wilson.2018}, 
		%
		% CMES style rewriting
%		who proposed 
		where 
		a method for step-size tuning and step-size decaying
		was proposed 
		to achieve lowest training error and generalization (test) error for both \hyperref[sc:generic-SGD]{Standard SGD} and \hyperref[sc:SGD-momentum]{SGD with momentum} (``Heavy Ball'' or better yet ``\hyperref[sc:SGD-momentum]{Small Heavy Sphere}'' method) compared to adaptive methods such as \hyperref[para:adagrad]{AdaGrad}, \hyperref[para:rmsprop]{RMSProp}, \hyperref[para:adam1]{Adam}. 
		{\footnotesize (Figure reproduced with permission of the authors.)}
		%{\color{red} ASK PERMISSION.}
	}
	\label{fig:wilson-examples}
\end{figure}

%
% CMES style rewriting
%\cite{Wilson.2018} observed 
It was observed in \cite{Wilson.2018}
that adaptive methods tended to have larger generalization (test) errors\footnote{
	See \cite{Goodfellow.2016}, p.~107, regarding training error and test (generalization) error.  ``The ability to perform well on previously unobserved inputs is called generalization.''
} compared to SGD: ``We observe that the solutions found by adaptive methods generalize worse (often significantly worse) than SGD, even when these solutions have better training performance,'' (see Figure~\ref{fig:wilson-examples}), and concluded that:
\begin{quote}
	``Despite the fact that our experimental evidence demonstrates that adaptive methods are not advantageous for machine learning, the \hyperref[para:adam1]{Adam} algorithm remains incredibly popular. We are not sure exactly as to why, but hope that our step-size tuning suggestions make it easier for practitioners to use standard stochastic gradient methods in their research.''
\end{quote}
The work of \cite{Wilson.2018} has encouraged researchers who were enthusiastic with adaptive methods to take a fresh look at SGD again to tease something more out of this classic method.\footnote{
	See, e.g., \vphantom{\cite{Xing.2018}}\cite{Xing.2018}, where 
	%
	% CMES style rewriting
%	who, while not referring directly to \cite{Wilson.2018}, referred to \cite{SmithSL.2018b}, who referred to \cite{Wilson.2018}. 
    \cite{Wilson.2018} was not referred to directly, but through a reference to \cite{SmithSL.2018b}, in which there was a reference to \cite{Wilson.2018}.  
}

\begin{figure}[h]
	\centering
	%
	% 2022.12.17
	% add "-eps-converted-to.pdf" for arXiv
	% \includegraphics[width=0.48\linewidth]{figures/Loshchilov-2019-AdamW-2019-12-12}
	\includegraphics[width=0.48\linewidth]{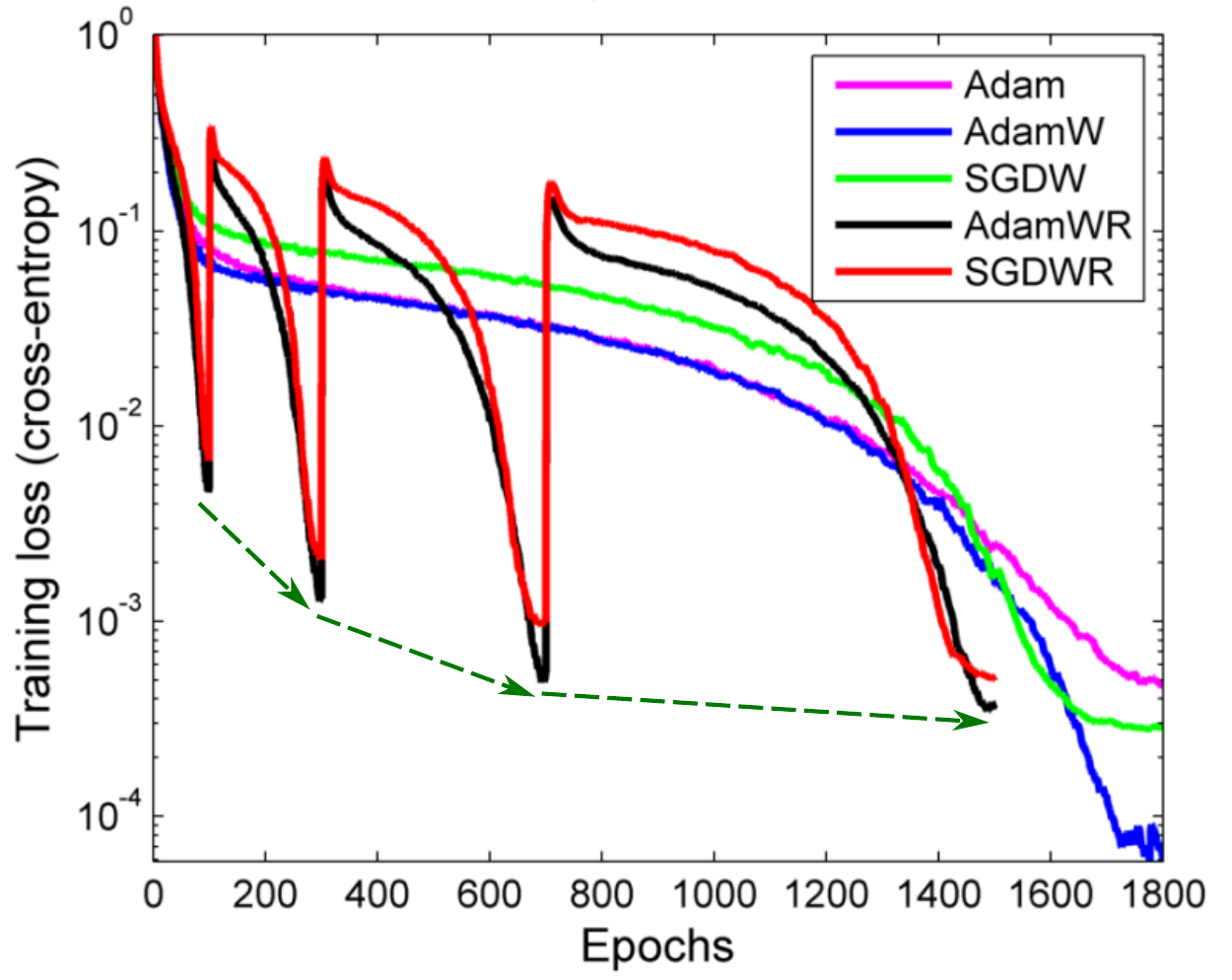}
	\includegraphics[width=0.48\linewidth]{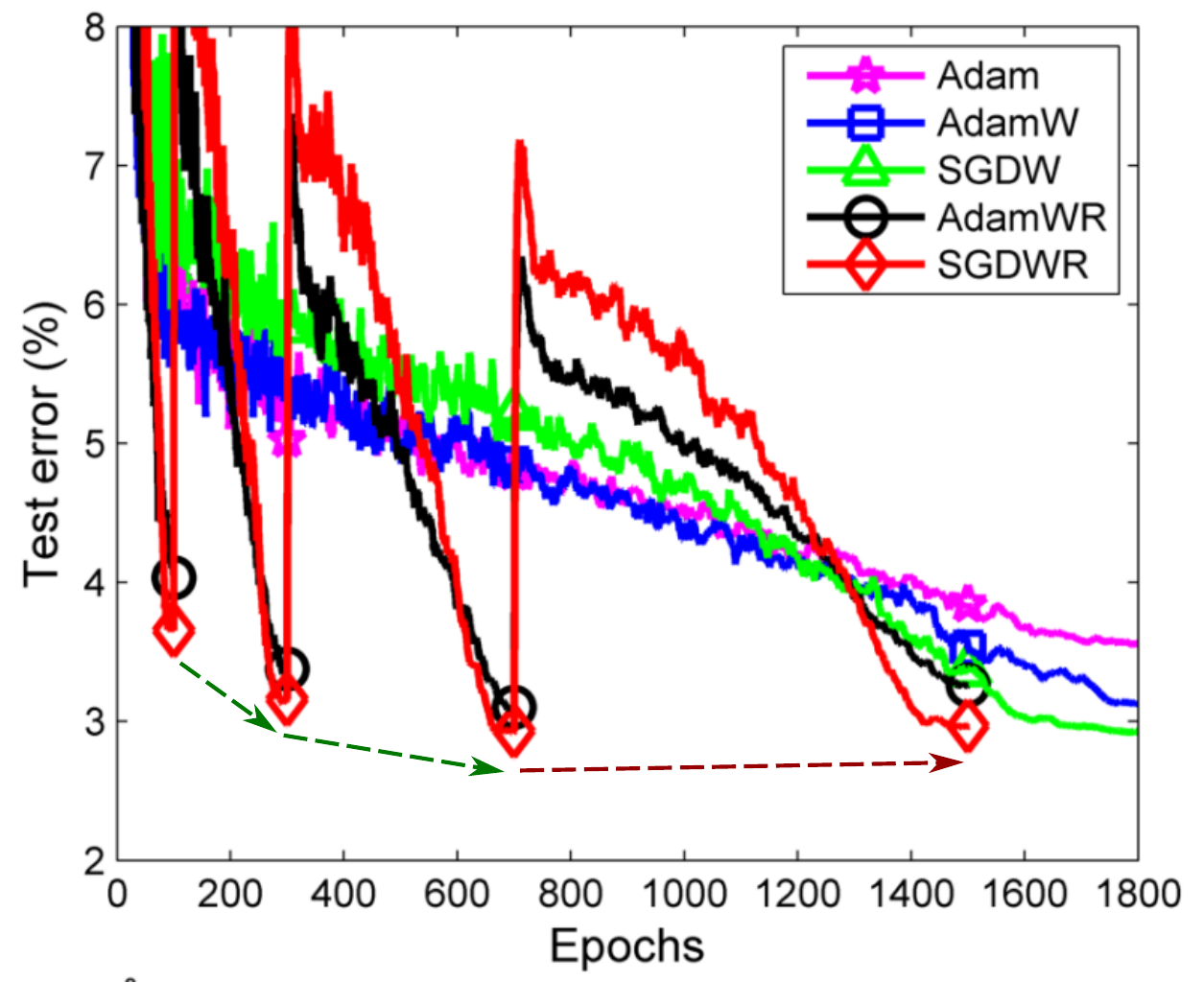}	
	\caption{
		\emph{AdamW vs Adam, SGD, and variants on CIFAR-10 dataset} (Sections~\ref{sc:training-valication-test}, \ref{sc:adamw}). While AdamW achieved lowest training loss (error) after 1800 epochs, the results showed that SGD with weight decay (SGDW) and with warm restart (SGDWR) achieved lower test (generalization) errors than \hyperref[para:adam1]{Adam}, \hyperref[para:adamw]{AdamW}, AdamWR. See Figure~\ref{fig:adamw-annealing} for the scheduling of the annealing multiplier $\anneal_k$, for which the epoch numbers $(100, 300, 700, 1500)$ for complete cooling ($\anneal_k = 0$) coincided with the same epoch numbers for the sharp minima.  There was, however, a diminishing return beyond the 4th cycle as indicated by the dotted arrows, for both training error and test error, which actually increased at the end of the 4th cycle (right subfigure, red arrow), see Section~\ref{sc:training-valication-test} on early-stopping criteria and Remark~\ref{rm:annealing-cyclic-limitation}.    Adapted from \cite{Loshchilov.2019}. 
		{\footnotesize (Figure reproduced with permission of the authors.)}
		%{\color{red} ASK PERMISSION.}
	}
	\label{fig:adamw-examples}
\end{figure}

%{\bf AdamW.}
\subsubsection{AdamW: Adaptive moment with weight decay}
\label{sc:adamw}
\label{para:adamw}
%(Adaptive moment with weight decay.)
%
% CMES style rewriting
The authors of
\cite{Loshchilov.2019}, aware of the work 
%of
in 
\cite{Wilson.2018}, wrote: 
It was suggested in
\cite{Wilson.2018} 
%``suggested 
``that adaptive gradient methods do not generalize as well as SGD with momentum when tested on a
diverse set of deep learning tasks, such as image classification, character-level language modeling and constituency parsing.''  In particular, 
%
% CMES style rewriting
%\cite{Loshchilov.2019} showed 
it was shown in \cite{Loshchilov.2019}
that `` a major factor of the poor generalization of the most popular adaptive gradient method, \hyperref[para:adam1]{Adam}, is due to the fact that $L_2$ regularization is not nearly as effective for it as for SGD,'' and proposed to move the weight decay from the gradient ($L_2$ regularization) to the parameter update (original weight decay regularization).  So what is ``$L_2$ regularization'' and what is ``weight decay'' ?  (See also Section~\ref{sc:weight-decay} on weight decay.)

\begin{figure}
	\centering
	%
	% 2022.12.17
	% add "-eps-converted-to.pdf" for arXiv
	% \includegraphics[width=0.48\linewidth]{figures/Loshchilov-2019-annealing-2019-12-21}	
	\includegraphics[width=0.48\linewidth]{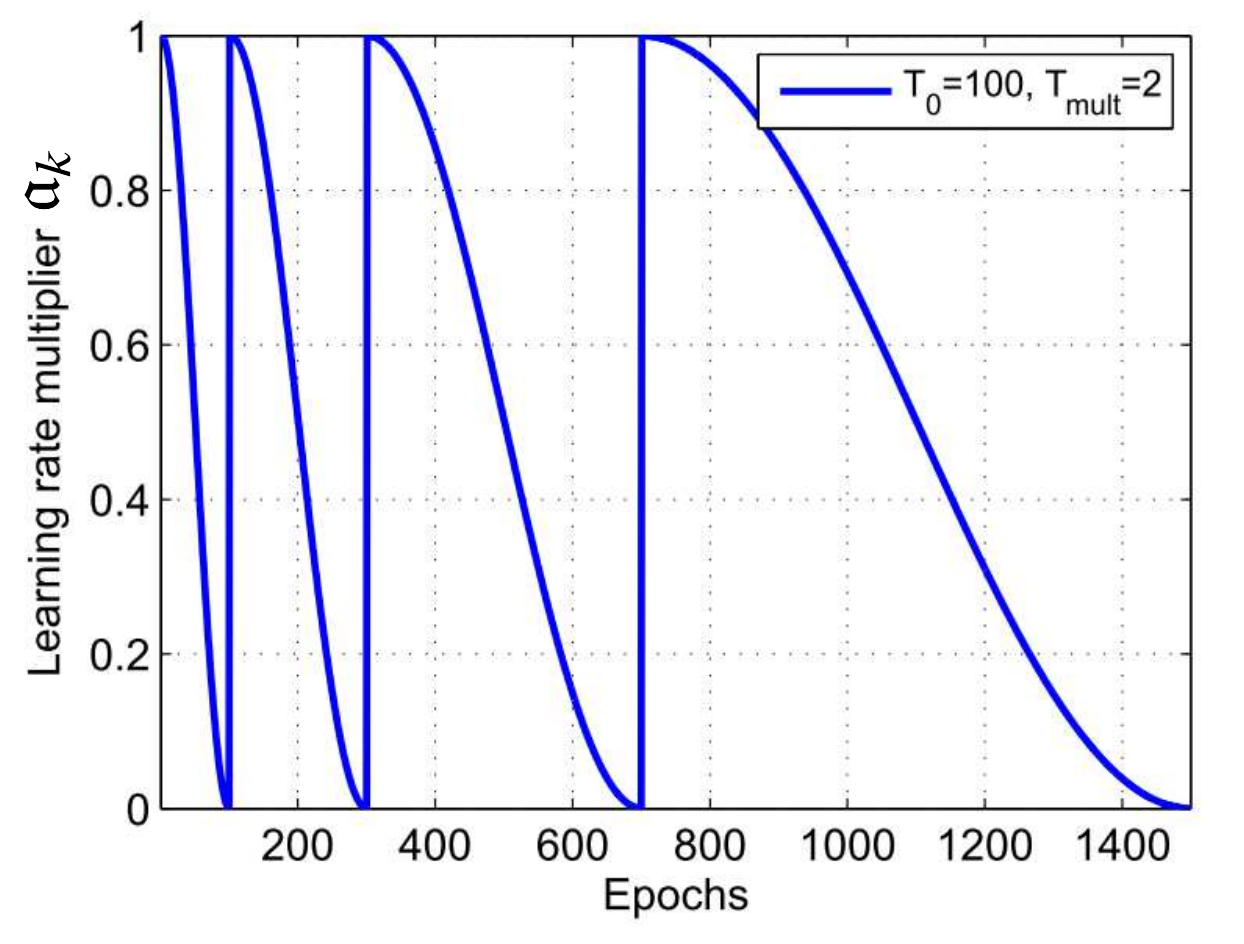}	
	\caption{
		\emph{Cosine annealing} (Sections~\ref{sc:step-length-decay}, \ref{para:adamw}). Annealing factor $\anneal_k$ as a function of epoch number. Four annealing cycles $p=1,\ldots,4$, with the following schedule for $T_p$ in Eq.~(\ref{eq:cosine-annealing}): 
		(1) Cycle 1, $T_1 = 100$ epochs, epoch 0 to epoch 100, 
		(2) Cycle 2, $T_2 = 200$ epochs, epoch 101 to epoch 300,
		(3) Cycle 3, $T_3 = 400$ epochs, epoch 301 to epoch 700,
		(4) Cycle 4, $T_4 = 800$ epochs, epoch 701 to epoch 1500.
		From \cite{Loshchilov.2019}.   See Figure~\ref{fig:adamw-examples} in which the curves for AdamWR and SGDWR ended at epoch 1500.    
		{\footnotesize (Figure reproduced with permission of the authors.)}
		%{\color{red} ASK PERMISSION.}
	}
	\label{fig:adamw-annealing}
\end{figure}

Briefly, ``$L_2$ regularization'' is aiming at decreasing the overall weights to avoid overfitting, which simply means that the network model tries to fit through as many data points as possible, including noise.\footnote{
	See \cite{Goodfellow.2016}, p.~107, Section 5.2 on ``Capacity, overfitting and underfitting'', and p.~115 provides a good explanation and motivation for regularization, as in Gupta 2017, `Deep Learning: Overfitting', 2017.02.12,  \href{https://towardsdatascience.com/deep-learning-overfitting-846bf5b35e24}{Original website}, \href{https://web.archive.org/web/20190201174735/https://towardsdatascience.com/deep-learning-overfitting-846bf5b35e24?gi=f70587e2080}{Internet archive}. 
}

\begin{align}
	\losstr (\bparamt) = \losst (\bparamt) + \frac{\decaywt}{2} (\parallel \bparamt \parallel_2)^2
	\label{eq:cost-regularized}
\end{align}
The magnitude of the coefficient $\decaywt$ regulates (or regularizes) the behavior of the network: $\decaywt = 0$ would lead to overfitting (Figure~\ref{fig:overfit} right), a moderate $\decaywt$ may yield appropriate fitting (Figure~\ref{fig:overfit} middle), a large $\decaywt$ may lead to underfitting (Figure~\ref{fig:overfit} left).  The gradient of the regularized cost $\losstr$ is then
\begin{align}
	\bgradtr
	:=
	\frac{\partial \losstr}{\partial \bparamt}
	=
	\frac{\partial \losst}{\partial \bparamt}
	+
	\decaywt \bparamt
	=
	\bgradt + \decaywt\bparamt
	\label{eq:gradient-regularized}
\end{align}
and the update becomes:
\begin{align}
	\bparamt_{k+1} = \bparamt_k - \learn_k ( \bgradt_k + \decaywt \bparamt_k ) = (1 - \learn_k \decaywt) \bparamt_k - \learn_k \bgradt_k
	\ ,
	\label{eq:update-regularized}
\end{align}
which is equivalent to decaying the parameters (including the weights in) $\bparamt_k$ when $(1 - \learn_k  \decaywt) \in (0,1)$, but with varying decay parameter $\decaywt_k = \learn_k \decaywt \in (0,1)$ depending on the step length $\learn_k$, which itself would decrease toward zero.

The same equivalence between $L_2$ regularization Eq.~(\ref{eq:cost-regularized}) and weight decay Eq.~(\ref{eq:update-regularized}), which is linear with respect to the gradient, cannot be said for adaptive methods due to the nonlinearity with respect to the gradient in the update procedure using Eq.~(\ref{eq:adaptive-learning-rate}) and Eq.~(\ref{eq:update-param-adaptive}).  See also the parameter update in lines~\ref{lst:line:adaptive-learning-update}-\ref{lst:line:adaptive-learning-update-2} of the unified pseudocode for adaptive methods in Algorithm~\ref{algo:unified-adaptive-learning-rate-2} and Footnote~\ref{fn:adaptive-algos-param-update}.  Thus,
%
% CMES style rewriting 
%\cite{Loshchilov.2019} proposed 
it was proposed in \cite{Loshchilov.2019}
to explicitly add weight decay to the parameter update Eq.~(\ref{eq:update-param-adaptive}) for AdamW (lines~\ref{lst:line:adaptive-learning-update}-\ref{lst:line:adaptive-learning-update-2} in Algorithm~\ref{algo:unified-adaptive-learning-rate-2}) as follows:
\begin{align}
	\bparamt_{k+1}
	=
	\bparamt_{k} + \anneal_k ( \blearn_k \bdt_k - \decaywt \bparamt_k ) 
	\text{ (element-wise operations)}
	\ ,
	\label{eq:adamw-param-update}
\end{align}
where the parameter $\anneal_k \in [0,1]$ is an annealing multiplier defined in Section~\ref{sc:step-length-decay} on weight decay:
\begin{align}
	\anneal_k = 0.5 + 0.5 \cos (\pi T_{cur} / T_p)
	\in [0,1]
	\ ,
	\text{ with }
	T_{cur} := j - \sum_{q=1}^{q=p-1} T_q
	\ ,
	\tag{\ref{eq:cosine-annealing}}
\end{align}

%{\color{red} HERE 2019.12.20}

%
% CMES style rewriting
%\cite{Loshchilov.2019} reported 
The
results of 
%their 
the
numerical experiments on the CIFAR-10 dataset 
using \hyperref[para:adam1]{Adam}, AdamW, SGDW (Weight decay), AdamWR (Warm Restart), and SGDWR
were reported in Figure~\ref{fig:adamw-examples} \cite{Loshchilov.2019}.

\begin{rem}
	\label{rm:annealing-cyclic-limitation}
	Limitation of cyclic annealing.
	{\rm 
		The 5th cycle of annealing not shown in Figure~\ref{fig:adamw-examples} would end at epoch $1500 + 1600 = 3100$, which is well beyond epoch budget of $1800$.  In view of the diminishing return in the decrease of the training error at the end of each cycle, in addition to an increase in the test error by the end of the 4th cycle, as shown in Figure~\ref{fig:adamw-examples}, it is unlikely that it is worthwhile to start the 5th cycle, since not only the computation would be more expensive, due to the warm restart, the increase in the test error indicated that the end of the 3rd cycle was optimal, and thus the reason for \cite{Loshchilov.2019} to stop at the end of the 4th cycle. 
	} $\hfill\blacksquare$
\end{rem}

\begin{figure}
	\centering
	%
	% 2022.12.17
	% add "-eps-converted-to.pdf" for arXiv
	% \includegraphics[width=1.0\linewidth]{figures/Aitchison-2019-Fig-2020-01-29}	
	\includegraphics[width=1.0\linewidth]{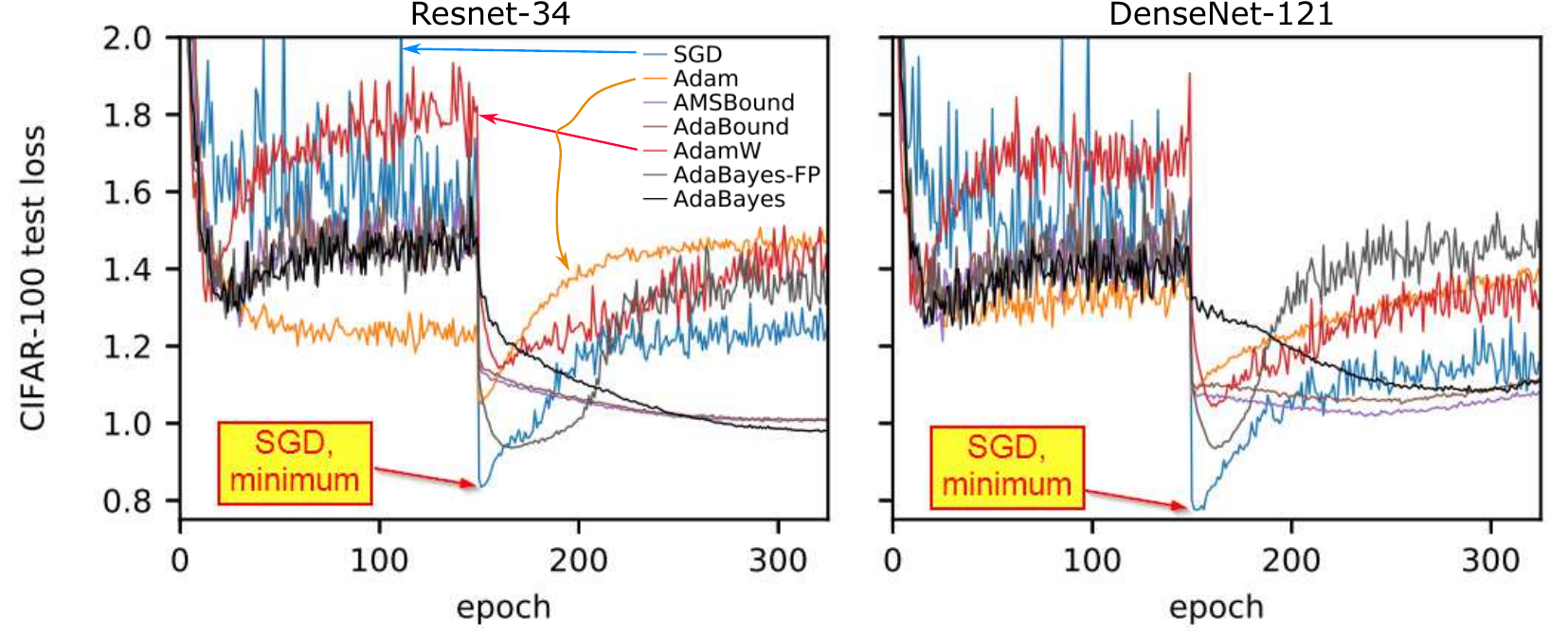}	
	\caption{
		\emph{CIFAR-100 test loss using Resnet-34 and DenseNet-121} (Section~\ref{para:adamw}). Comparison between various optimizers, including \hyperref[sc:adam1]{Adam} and \hyperref[sc:adamw]{AdamW}, showing that \hyperref[sc:generic-SGD]{SGD} achieved the lowest global minimum loss (blue line) compared to all adaptive methods tested as shown \cite{Aitchison.2019}.   See also Figure~\ref{fig:Aitchison-table} and Section~\ref{sc:training-valication-test} on early-stopping criteria. 
		{\footnotesize (Figure reproduced with permission of the authors.)}
		%{\color{red} ASK PERMISSION.}
	}
	\label{fig:Aitchison-figure}
\end{figure}

\begin{figure}
	\centering
	\includegraphics[width=0.7\linewidth]{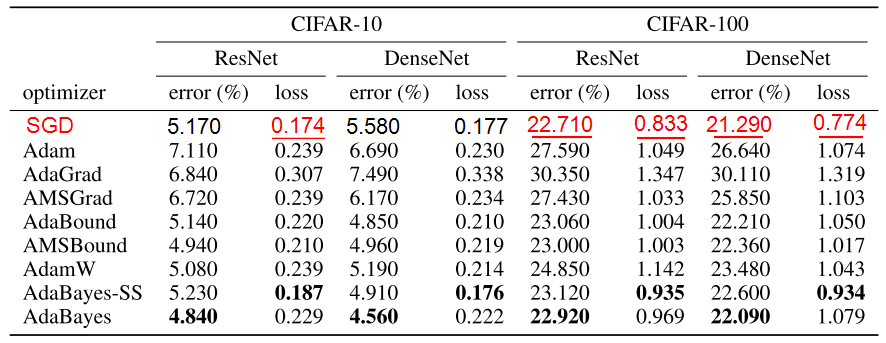}	
	\caption{
		\emph{SGD frequently outperformed all adaptive methods} (Section~\ref{para:adamw}). The table contains the global minimum for each optimizer, for each of the two datasets CIFAR-10 and CIFAR-100, using two different networks.  For each network, an error percentage and the loss (cost) were given.  Shown in red are the lowest global minima obtained by \hyperref[sc:generic-SGD]{SGD} in the corresponding columns.  Even in the three columns in which \hyperref[sc:generic-SGD]{SGD} results were not the lowest, two \hyperref[sc:generic-SGD]{SGD} results were just slightly above those of \hyperref[sc:adamw]{AdamW} (1st and 3rd columns), and one even smaller (4th column).   \hyperref[sc:generic-SGD]{SGD} clearly beat \hyperref[sc:adam1]{Adam}, \hyperref[sc:adagrad]{AdaGrad}, \hyperref[sc:amsgrad]{AMSGrad} \cite{Aitchison.2019}.    See also Figure~\ref{fig:Aitchison-figure}.
		{\footnotesize (Figure reproduced with permission of the authors.)}
		%{\color{red} ASK PERMISSION.}
	}
	\label{fig:Aitchison-table}
\end{figure}

The results for test errors in Figure~\ref{fig:adamw-examples} appeared to confirm the \hyperref[para:adam-criticism]{criticism} 
%
% CMES style rewriting
%by 
in
\cite{Wilson.2018} that adaptive methods brought about ``marginal value'' compared to the classic \hyperref[sc:generic-SGD]{SGD}.  Such observation was also in agreement with \cite{Aitchison.2019}, where it was stated:
\begin{quote}
	``In our experiments, either AdaBayes or AdaBayes-SS outperformed other adaptive methods, including AdamW (Loshchilov \& Hutter, 2017), and Ada/AMSBound (Luo et al., 2019), though SGD frequently outperformed all adaptive methods.'' (See Figure~\ref{fig:Aitchison-figure} and Figure~\ref{fig:Aitchison-table})
\end{quote}
If \hyperref[para:amsgrad]{AMSGrad} generated ``a lot of noise for nothing'' compared to \hyperref[sc:adam1]{Adam} and \hyperref[sc:adamw]{AdamW}, according to \cite{Gugger.2018}, then does ``marginal value'' mean that adaptive methods in general generated a lot of noise for not much, compared to \hyperref[sc:generic-SGD]{SGD} ?  

The work 
%
% CMES style rewriting
%of
in 
\cite{Wilson.2018}, \cite{Aitchison.2019}, and \cite{Loshchilov.2019} proved once more that a classic like SGD introduced 
by Robbins \& Monro (1951b) 
\cite{Robbins1951b} never dies, and would be motivation to generalize classical deterministic first and second-order optimization methods together with line search methods to add stochasticity.  We will review in detail two papers along this line: \cite{Paquette.2018} and \cite{Bergou.2018}.

\begin{algorithm}
	{\bf SGD with Armijo line-search and adaptive minibatch} (for training epoch $\tau$)
	\\
	\KwData{
		% Layer outputs $\byp{\ell}$, for $\ell=L, \cdots , 1$ 
		\\
		$\bullet$ Parameter estimate $\bparamt^\star_{\tau}$ from previous epoch $(\tau-1)$ (line~\ref{lst:list:SGD-params-previous-epoch} in Algorithm~\ref{algo:generic-SGD})
		\\
		$\bullet$ Select 3 Armijo parameters $\alpha \in (0,1)$, $\beta \in (0,1)$, $\rho >0 $, and {\color{purple} reliability bound $\delta_1 = \delta > 0$}
		\\
		{\color{purple} $\bullet$ Select probability $p_f \in (0,1]$ for cost estimate $\losst$ to be close to true cost $\loss$}
		\\
		{\color{purple} $\bullet$ Select probability $p_g \in (0,1]$ for gradient estimate $\bgradt$ to be close to true gradient $\bgrad$}  
	}
	\KwResult{
		{\color{purple} Parameter estimate for next epoch $(\tau+1)$: $\bparamt_{\tau+1}^\star$.} \label{lst:list:SGD-Armijo-result}
	}
	\vphantom{Blank line}
	Define cost{\color{purple} -estimate function $\losst (\bparamt , \epsilon , p_J)$ using adaptive minibatch, Eq.~(\ref{eq:cost-gradient-adaptive-minibatch})}
	\;
	Define gradient{\color{purple} -estimate function $\bgradt (\bparamt , \epsilon , p_g )$ using adaptive minibatch, Eq.~(\ref{eq:cost-gradient-adaptive-minibatch})}
	\;
	$\blacktriangleright$ Begin training epoch $\tau$ (see Algorithm~\ref{algo:generic-SGD} for standard SGD)
	\\
	Initialize parameter estimate $\bparamt_1 = \bparamt^\star_{\tau}$ and step length $\epsilon_1 = \rho$ 
	%\vspace{2mm}
	\;
	{\color{purple} \ding{173} \ding{174}} \For{$k = 1,2, \ldots$}{
		
		{\color{purple} Compute gradient estimate $\bgradt_k = \bgradt (\bparamt_k , \epsilon_k , p_g )$ with adaptive minibatch} 
		\;
		
		Set descent-direction {\color{purple} estimate} $\bdt_k = - \bgradt_k$ as in Eq.~(\ref{eq:stochastic-steepest-descent-direction})
		\;
		
		{\color{purple} Compute cost estimates $\losst (\bparamt_k, \epsilon_{k} , p_J )$ and $\losst (\bparamt_k + \epsilon_k \bdt_k , \epsilon_{k} , p_J )$  with adaptive minibatch}
		\;

		%\vphantom{Blank line}
		%\vspace{2mm}
		
		% gradient descent
		\eIf{Stopping criterion not satisfied}{
			
			% THEN, gradient descent 
			%\vspace{2mm}
			$\blacktriangleright$ Compute step length (learning rate) $\epsilon_k$ using stochastic Armijo's rule.
			\\
			%\ding{174} \For{j=1,2,\ldots}{
			
			%empty, no for loop for stochastic Armijo
			
			%} % END For loop, Armijo
			
			\eIf{{\color{purple} Stochastic} Armijo decrease condition Eq.~(\ref{eq:stochastic-armijo-3}) not satisfied}{
				
				% THEN, Armijo
				$\blacktriangleright$ Decrease step length: 
				Set $\epsilon_k \leftarrow \beta \epsilon_k$
				\;
				
			}{
				
				% ELSE, Armijo
				$\blacktriangleright$ {\color{purple} Stochastic} Armijo decrease condition Eq.~(\ref{eq:stochastic-armijo-3}) satisfied.
				\\
				$\blacktriangleright$ Update network parameter $\bparamt_k$ to $\bparamt_{k+1}$:
				Set $\bparamt_{k+1} \leftarrow \bparamt_{k} + \epsilon_k \bd_k$
				\;
				{ % BEGIN Purple
					\color{purple}
					$\blacktriangleright$ Increase the next step length $\epsilon_{k+1}$:
					%\\
					Set $\epsilon_{k+1} = \max\{ \rho , (\epsilon_{k} / \beta) \}$
					\;
					%\vspace{2mm}
					$\blacktriangleright$ Check for reliability of step length
					\\
					\eIf{Step length reliable, $\epsilon_{k} \parallel \bgradt \parallel^2 \ge \delta_k^2$}{
						
						% THEN, reliability
						$\blacktriangleright$ Increase reliability bound:
						%\\
						Set $\delta_{k+1}^2 \leftarrow \delta_k^2 / \beta$
						\;
						
					}{
						
						% ELSE, reliability
						$\blacktriangleright$ Decrease reliability bound:
						% {\color{red} HERE }
						%\\
						Set $\delta_{k+1}^2 \leftarrow \beta \delta_k^2$
						\;
						
					} % END, reliability
				} % END Purple
				
				% no for loop for stochastic Armijo
				%Stop \ding{174} {\bf for} loop (end Armijo line search).
				
			} % ENDIF, Armijo
			
		}{
			
			% ELSE, gradient descent
			%\vspace{2mm}
			$\blacktriangleright$ Stopping criterion satisfied; 
			\\
			$\blacktriangleright$ {\color{red} Update parameter estimate for next epoch $(\tau+1)$}:
			$\bparamt_{\tau+1}^\star \leftarrow \bparamt_k$
			\\
			$\blacktriangleright$ Stop {\color{purple} \ding{173} \ding{174}} {\bf for} loop (end minimization for training epoch $\tau$).
			
		} % ENDIF, gradient descent
		
		%\vspace{2mm}
		$\blacktriangleright$ Continue to next iteration $(k+1)$:
		Set $k \leftarrow k+1$
		\;
	}

	\vphantom{Blank line} 
	\caption{
		\emph{SGD with Armijo line search and adaptive minibatch} (Section~\ref{sc:SGD-momentum}, \ref{sc:SGD-armijo}, \ref{sc:stochastic-Newton}, Algorithms~\ref{algo:descent-armijo-deterministic},
		\ref{algo:generic-SGD}, \ref{algo:stochastic-newton}).
		From \cite{Paquette.2018}. The differences compared to the deterministic gradient descent with Armijo line search in Algorithm~\ref{algo:descent-armijo-deterministic}, including the \ding{173} \ding{174} {\bf for} loop, are highlighted in purple; See Remark~\ref{rm:paquette-2018}.  For the related stochastic methods with fixed minibatch, see the standard SGD Algorithm~\ref{algo:generic-SGD} and the Newton descent with Armijo-like line search Algorithm~\ref{algo:stochastic-newton}. Table~\ref{tb:armijo-params-notations} shows a comparison of the notations used by several authors in relation to Armijo line search.		
	}
	\label{algo:descent-armijo-stochastic-1}
\end{algorithm}

%\subsubsection{SGD with Armijo line search and adaptive minibatch}
\subsection{SGD with Armijo line search and adaptive minibatch}
\label{sc:SGD-armijo}
In parallel to the deterministic choice of step length based on Armijo's rule in Eq.~(\ref{eq:armijo-1}) and Eq.~(\ref{eq:armijo-3}), we have the following respective stochastic version proposed by \cite{Paquette.2018}:\footnote{
	It is not until Section 4.7 in \cite{Paquette.2018} that this version is presented for the general descent of nonconvex case, whereas the pseudocode in their Algorithm 1 at the beginning of their paper, and referred to in Section 4.5, was restricted to steepest descent for convex case.
}
\begin{align}
	&
	\epsilon (\bparam) = \min_j \{\beta^j  \, \rho \, | \, \losst ( \bparam + \beta^j \, \bdt ) - \losst (\bparam) \le \alpha {\beta^j} \rho \, \bgradt \, \dotprod \, \bdt \}
	\ ,
	\label{eq:stochastic-armijo-1}
	\\
	&
	\losst (\bparam + \epsilon \bdt)
	\le
	\losst (\bparam)
	+
	\alpha \epsilon
	\, \bgradt \, \dotprod \, \bdt
	\ ,
	\label{eq:stochastic-armijo-3}
\end{align}
where the overhead tilde of a quantity designates an estimate of that quantity based on a randomly selected minibatch, i.e., $\losst$ is the cost estimate, $\bdt$ the descent-direction estimate, and $\bgradt$ the gradient estimate, similar to those in Algorithm~\ref{algo:generic-SGD}.  

There is a difference though: The standard SGD in Algorithm~\ref{algo:generic-SGD} uses a fixed minibatch for the computation of the cost estimate and the gradient estimate, whereas Algorithm~\ref{algo:descent-armijo-stochastic-1} 
%
% CMES style rewriting
%by  
in
\cite{Paquette.2018} uses adaptive subprocedure to adjust the size of the minibatches to achieve a desired (fixed) probability $p_J$ that the cost estimate is close to the true cost, and a desired probability $p_g$ that the gradient estimate is close to the true gradient.  These adaptive-minibatch subprocedures are also functions of the learning rate (step length) $\learn$, conceptually written as:
\begin{align}
	\losst = \losst (\bparam , \learn , p_J)
	\text{ and }
	\bgradt = \bgradt (\bparam , \learn , p_g)
	\ ,
	\label{eq:cost-gradient-adaptive-minibatch}
\end{align} 
which are the counterparts to the fixed-minibatch procedures in Eq.~(\ref{eq:cost-estimate}) and Eq.~(\ref{eq:gradient-estimate}), respectively.  

\begin{rem}
	\label{rm:paquette-2018}
	{\rm
		Since the appropriate size of the minibatch depends on the gradient estimate, which is not known and which is computed based on the minibatch itself, the adaptive-minibatch subprocedures for cost estimate $\losst$ and for gradient estimate $\bgradt$ in Eq.~(\ref{eq:cost-gradient-adaptive-minibatch}) contain a loop, started by guessing the gradient estimate, to gradually increase the size of the minibatch by adding more samples until certain criteria are met.\footnote{
			See \cite{Paquette.2018}, p.~7, below Eq.~(2.4).
		}  
		
		In addition, since both the cost estimate $\losst$ and the gradient estimate $\bgradt$ depend on the step size $\epsilon$,  the Armijo line-search loop to determined the step length $\epsilon$---denoted as the \ding{173} for loop in the deterministic Algorithm~\ref{algo:descent-armijo-deterministic}---is combined with the iteration loop $k$ in Algorithm~\ref{algo:descent-armijo-stochastic-1}, where these two combined loops are denoted as the \ding{173} \ding{174} {\bf for} loop.
	}
	$\hfill\blacksquare$
\end{rem}

The same relationship between $\bdt$ and $\bgradt$ as in Eq.~(\ref{eq:descent-dir}) holds:
\begin{align} 
	\bgradt \, \dotprod \, \bdt
	=
	\sum_i \sum_j \gradt_{ij} \downt_{ij} < 0
	\ . 
	\label{eq:descent-dir-2}
\end{align}
For SGD, the descent-direction estimate $\bdt$ is identified with the steepest-descent direction estimate $(-\bgradt)$: 
\begin{align}
	\bdt = - \bgradt
	\ ,
	\label{eq:stochastic-steepest-descent-direction}
\end{align}
For Newton-type algorithms, such as in \cite{Bergou.2018} \cite{Wills.2018}, the descent direction estimate $\bdt$ is set to equal to the Hessian estimate $\bHst$ multiplied by the steepest-descent direction estimate $(-\bgradt)$:
\begin{align}
	\bdt = \bHst \cdot (- \bgradt)
	\ ,
	\label{eq:stochastic-Newton-direction}
\end{align}

\begin{rem}
	\label{rm:paquette-2018-reliability}
	{\rm 
		In the SGD with Armijo line search and adaptive minibatch Algorithm~\ref{algo:descent-armijo-stochastic-1}, the reliability parameter $\delta_k$ and its use is another difference between Algorithm~\ref{algo:descent-armijo-stochastic-1} and  Algorithm~\ref{algo:descent-armijo-deterministic}, the deterministic gradient descent with Armijo line search, and similarly for Algorithm~\ref{algo:generic-SGD}, the standard SGD.  
		%
		% CMES style rewritiing
%		\cite{Paquette.2018} provided the 
		The
		reason was provided in \cite{Paquette.2018}: Even when the probability of gradient estimate and cost estimate is near 1, it is not guaranteed that the expected value of the cost at the next iterate $\bparamt_{k+1}$ would be below the cost at the current iterate $\bparamt_k$, due to arbitrary increase of the cost.  
		``Since random gradient may not be representative of the true gradient the function estimate accuracy and thus the expected improvement needs to be controlled by a different quantity,''  $\delta_k^2$.
	} $\hfill\blacksquare$
\end{rem}

%
% CNES style rewriting
The authors of 
\cite{Paquette.2018} provided a rigorous convergence analysis of their proposed Algorithm~\ref{algo:descent-armijo-stochastic-1}, but had not implemented their method, and thus had no numerical results at the time of this writing.\footnote{
	Based on our private communications with the authors of \cite{Paquette.2018} on 2019.11.16.
}
Without empirical evidence that the algorithm works and is competitive compared to SGD (see adaptive methods and their criticism in Section~\ref{sc:adam-criticism}), there would be no adoption.

%{\color{red}  HERE 2019.12.10 }

\begin{minipage}{\linewidth}
	\begin{table}[H]
		\centering
		\caption{
			\emph{Armijo parameters} (Section~\ref{sc:armijo}, Algorithms~\ref{algo:descent-armijo-deterministic}, \ref{algo:descent-armijo-stochastic-1}, \ref{algo:stochastic-newton}). Comparing our notations here as used in Eq.~(\ref{eq:armijo-1}) and Eq.~(\ref{eq:armijo-3}) to those of other authors: \cite{Polak.1971} \cite{Paquette.2018} \cite{Bergou.2018} \cite{Armijo.1966}.  
			In \cite{Armijo.1966} \cite{Polak.1971} \cite{Paquette.2018}, the parameters are for the first-order term $(-\bgrad) \dotprod \, \bd = \parallel \bd \parallel^2$ (for steepest descent, $\bd = - \bgrad = - \partial J / \partial \bparam$) in the Taylor series expansion of $\Delta J = J(\bparam + \epsilon \bd) - J(\bparam)$.  
			In \cite{Bergou.2018}, the parameters are for the second-order term in the Taylor series expansion, leading to the cube of the norm of the descent direction, $\parallel \bd \parallel^3$.  For the stochastic optimization algorithms presented in this paper, the 4th parameter $\delta$ is introduced to represent the reliability parameter $\delta_0$ of \cite{Paquette.2018}  (Algorithm~\ref{algo:descent-armijo-stochastic-1}) and the stability parameter $\epsilon$ of \cite{Bergou.2018}  (Algorithm~\ref{algo:stochastic-newton}). 
			%While the subscript $0$ in $\mu_0$ represents the initial element in a sequence in \cite{Paquette.2018}, $\mu_0$ is only a fixed constant, not part of any sequence, in \cite{Bergou.2018}. 
			Deterministic algorithms in \cite{Armijo.1966} and \cite{Polak.1971} do not have this 4th parameter.
		}
		\begin{tabularx}{0.95\textwidth}{c *{6}{Y}}
			\toprule[3pt]
			Parameter type		
			& \multicolumn{1}{c}{This paper}  
			& \multicolumn{1}{c}{Polak \cite{Polak.1971}}
			& \multicolumn{1}{c}{Paquette \cite{Paquette.2018}} 
			& \multicolumn{1}{c}{Bergou \cite{Bergou.2018}}
			& \multicolumn{1}{c}{Armijo \cite{Armijo.1966}} 
			\\
			\cmidrule(lr){1-1} \cmidrule(lr){2-2} \cmidrule(lr){3-3} \cmidrule(lr){4-4} \cmidrule(lr){5-5} \cmidrule(lr){6-6}
			Fixed factor 		
			& \multicolumn{1}{c}{$\alpha$}  
			& \multicolumn{1}{c}{$\alpha$} 
			& \multicolumn{1}{c}{$\theta$} 
			& \multicolumn{1}{c}{$\frac16$} 
			& \multicolumn{1}{c}{$\frac12$}
			\\
			Varying-power factor	
			& \multicolumn{1}{c}{$\beta$}  
			& \multicolumn{1}{c}{$\beta$} 
			& \multicolumn{1}{c}{$\gamma^{-1}$} 
			& \multicolumn{1}{c}{$\theta$} 
			& \multicolumn{1}{c}{$\frac12$}
			\\
			Fixed factor	
			& \multicolumn{1}{c}{$\rho$}  
			& \multicolumn{1}{c}{$\rho$} 
			& \multicolumn{1}{c}{$\alpha_{max}$} 
			& \multicolumn{1}{c}{$\eta$} 
			& \multicolumn{1}{c}{$\alpha$}
			\\
			\cmidrule(lr){1-1} \cmidrule(lr){2-2} \cmidrule(lr){3-3} \cmidrule(lr){4-4} \cmidrule(lr){5-5} \cmidrule(lr){6-6}
			Reliability	/ Stability
			& \multicolumn{1}{c}{$\delta$}  
			& \multicolumn{1}{c}{---} 
			& \multicolumn{1}{c}{$\delta_0$} 
			& \multicolumn{1}{c}{$\epsilon$} 
			& \multicolumn{1}{c}{---}
			\\
			\midrule[2pt]
			\\
		\end{tabularx}
		\label{tb:armijo-params-notations}
	\end{table}
\end{minipage}

\begin{figure}[h]
	\centering
	\includegraphics[width=0.48\linewidth]{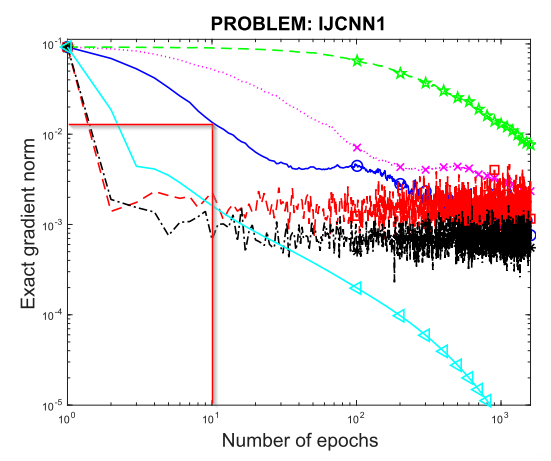}
	\includegraphics[width=0.48\linewidth]{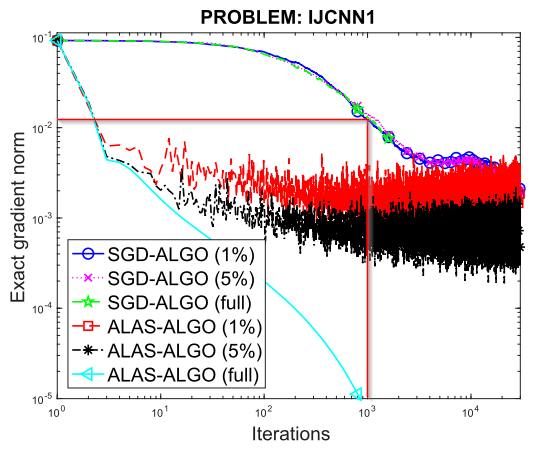}
	\caption{
		\emph{Stochastic Newton with Armijo-like 2nd order line search} (Section~\ref{sc:stochastic-Newton}).  IJCNN1 dataset from the LIBSVM library. Three batch sizes were used (1\%, 5\%, 100\%) for both SGD and ALAS (stochastic Newton Algorithm~\ref{algo:stochastic-newton}). The exact gradient norm for each of these six cases was plotted against the training epochs on the left, and against the iteration numbers on the right.  An epoch is the number of non-overlapping minibatches (and thus iterations) to cover the whole training set. One epoch for a minibatch size of s\% (respectively 1\%, 5\%, 100\%) of the training set is equivalent to $100 / s$ (respectively 100, 20, 1) iterations.  Thus, for SGD-ALGO (1\%), as shown, 10 epochs is equivalent to 1,000 iterations, with the same gradient norm.  The markers on the curves were placed every 100 epochs (left) and 800 iterations (right).  For the same number of epochs, say 10, SGD with smaller minibatches yielded lower gradient norm.  The same was true for Algorithm~\ref{algo:stochastic-newton} for number of epochs less than 10, but the gradient norm plateaued out after that with a lot of noise.  Second-order Algorithm~\ref{algo:stochastic-newton} converged faster than 1st-order SGD.
		See \cite{Bergou.2018}.  
		{\footnotesize (Figure reproduced with permission of the authors.)}
		%{\color{red} ASK PERMISSION.}
	}
	\label{fig:newton-examples}
\end{figure}

\subsection{Stochastic Newton method with 2nd-order line search}
\label{sc:stochastic-Newton}
The stochastic Newton method in Algorithm~\ref{algo:stochastic-newton}, described in \cite{Bergou.2018}, is a generalization of the deterministic Newton method in Algorithm~\ref{algo:gradient-quasi-newton-armijo-deterministic} to add stochasticity via random selection of minibatches and Armijo-like 2nd-order line search.\footnote{
	The Armijo line search itself is 1st order; see Section~\ref{sc:deterministic-optimization} on full-batch deterministic optimization.
}  

Upon a random selection of a minibatch as in Eq.~(\ref{eq:minibatch-2}), the computation of the estimates for the cost function $\losst$ in Eq.~(\ref{eq:cost-estimate}), the gradient $\bgradt$ in Eq.~(\ref{eq:gradient-estimate}), and the Hessian $\bHst$ (new quantity) can proceed similar to that used for the standard SGD in Algorithm~\ref{algo:generic-SGD}, recalled below for convenience:
\begin{align}
	\Bbbp{\bsize}
	= 
	\{ \bexin{i_1} , \cdots , \bexin{i_\bsize}  \}
	\subseteq
	\Xbb 
	=
	\{ \bexin{1} , \cdots , \bexin{\Bsize}  \}
	\ ,
	\tag{\ref{eq:minibatch-2}}
\end{align}
\begin{align}
	\losst (\bparam) = \frac{1}{\bsize} \sum_{k=1}^{k=\bsize \le \Bsize} J_{i_k} (\bparam)
	\ , 
	\text{ with }
	J_{i_k} (\bparam) = J(f(\bexin{i_k} , \bparam) , \bexout{i_k})
	\ , 
	\text{ and }
	\bexin{i_k} \in \Bbbp{\bsize}
	\ ,
	\bexout{i_k} \in \Tbbp{\bsize}
	\ ,
	\tag{\ref{eq:cost-estimate}}
\end{align}
\begin{align}
	\bgradt (\bparam) 
	= 
	\frac{\partial \losst (\bparam)}{\partial \bparam} 
	=
	\frac{1}{\bsize} \sum_{k=1}^{k=\bsize \le \Bsize}
	\frac{\partial \losst_{i_k} (\bparam)}{\partial \bparam}
	\ ,
	\tag{\ref{eq:gradient-estimate}}
\end{align}
\begin{align}
	\bHst (\bparam)
	= 
	\frac{\partial^2 \losst (\bparam)}{(\partial \bparam)^2} 
	=
	\frac{1}{\bsize} \sum_{k=1}^{k=\bsize \le \Bsize}
	\frac{\partial^2 \losst_{i_k} (\bparam)}{(\partial \bparam)^2}
	\ .
	\label{eq:hessian-estimate}
\end{align}
In the above computation, the minibatch in Algorithm~\ref{algo:stochastic-newton} is fixed, not adaptive such as in Algorithm~\ref{algo:descent-armijo-stochastic-1}. 

If the current iterate $\bparam_{k}$ is in a flat region (or plateau) or at the bottom of a local convex bowl, then the smallest eigenvalue $\Heigent$ of the Hessian estimate $\bHst$ would be close to zero or positive, respectively, and the gradient estimate would be zero (line~\ref{lst:line:newton-plateau} in Algorithm~\ref{algo:stochastic-newton}):
\begin{align}
	\Heigent \ge - \delta^{1/2}
	\text{ and }
	\parallel \bgradt \parallel = 0
	\ ,
	\label{eq:stochastic-newton-plateau}
\end{align}
where $\delta$ is a small positive number.  In this case, no step (or update) would be taken, which is equivalent to setting step length to zero or descent direction to zero\footnote{
	See Step 2 of Algorithm 1 in \cite{Bergou.2018}.
}, then go to the next iteration $(k+1)$.  Otherwise, i.e., the conditions in Eq.~(\ref{eq:stochastic-newton-plateau}) are not met, do the remaining steps.

If the current iterate $\bparam_{k}$ is on the downward side of a saddle point, characterized by the condition that the smallest eigenvalue $\Heigent_k$ is clearly negative  (line~\ref{lst:line:newton-negative-eigenvalue} in Algorithm~\ref{algo:stochastic-newton}):
\begin{align}
	\Heigent_k < - \delta^{1/2} < 0
	\ ,
	\label{eq:stochastic-newton-negative-eigenvalue}
\end{align}
then find the eigenvector $\bHevect_k$ corresponding to $\Heigent_k$, and scale it such that its norm is equal to the absolute value of this negative smallest eigenvalue, and such that this eigenvector forms an obtuse angle with the gradient $\bgradt_k$, then select such eigenvector as the descent direction $\bdt_k$:
\begin{align}
	\bHst_k \bHevect_k = \Heigent_k \bHevect_k 
	\text{ such that }
	\parallel \bHevect_k \parallel = - \Heigent_k
	\text{ and }
	\bHevect_k \, \dotprod \, \bgradt_k < 0
	\Rightarrow
	\bdt_k = \bHevect_k
	\label{eq:hessian-evect}
\end{align}

When the iterate $\bparam_{k}$ is in a local convex bowl, the Hessian $\bHst_k$ is positive definite, i.e., the smallest eigenvalue $\Heigent_k$ is strictly positive, then use the Newton direction as descent direction $\bdt_k$ (line~\ref{lst:line:newton-descent} in Algorithm~\ref{algo:stochastic-newton}): 
\begin{align}
	\Heigent_k > \parallel \bgradt_k \parallel^{1/2} > 0
	\Rightarrow
	\bHst_k \bdt_k = - \bgradt_k
	\label{eq:newton-descent}
\end{align}

The remaining case is when the iterate $\bparam_{k}$ is close to a saddle point such that the smallest eigenvalue $\Heigent_k$ is bounded below by $-\delta^{1/2}$ and above by $\parallel \bgradt_k \parallel^{1/2}$, then $\Heigent_k$ is nearly zero, and thus the Hessian estimate $\bHst_k$ is nearly singular.  Regularize (or stabilize) the Hessian estimate, i.e, move its smallest eigenvalue away from zero, by adding a small perturbation diagonal matrix using the sum of the bounds $\delta^{1/2}$ and $\parallel \bgradt_k \parallel^{1/2}$.  The regularized (or stabilized) Hessian estimate is no longer nearly singular, thus invertible, and can be used to find the Newton descent direction (lines~\ref{lst:line:newton-descent-not-selected-yet}-\ref{lst:line:newton-descent-regularized} in Algorithm~\ref{algo:stochastic-newton} for stochastic Newton and line~\ref{lst:line:newton-regularized} in Algorithm~\ref{algo:gradient-quasi-newton-armijo-deterministic} for deterministic Newton):
\begin{align}
	-\delta^{1/2} \le \Heigent_k \le \parallel \bgradt_k \parallel^{1/2}
	\Rightarrow
	0 < \Heigent_k + \parallel \bgradt_k \parallel^{1/2} + \delta^{1/2}
	\Rightarrow
	\left[ 
  	% CMES style - 2022.06.01
	% same problem with \bId, which could be that \I had been defined as something else in the TSP style
	% so just code up boldsymbol I explicitly
	%
		% \bHst_k + \left( \parallel \bgradt_k \parallel^{1/2} + \delta^{1/2} \right) \bId 
		\bHst_k + \left( \parallel \bgradt_k \parallel^{1/2} + \delta^{1/2} \right) \boldsymbol{I} 
	\right]
	\bdt_k
	=
	- \bgradt_k
	\ .
	\label{eq:newton-descent-regularized}
\end{align}

If the stopping criterion is not met, use Armijo's rule to find the step length $\epsilon_{k}$ to update the parameter $\bparam_{k}$ to $\bparam_{k+1}$, then go to the next iteration $(k+1)$. (line~\ref{lst:list:newton-stop-not-met} in Algorithm~\ref{algo:stochastic-newton}).
%
% CMES style rewriting
The authors of  
\cite{Bergou.2018} provided a detailed discussion of their stopping criterion.   Otherwise, the stopping criterion is met, accept the current iterate $\bparamt_{k}$ as local minimizer estimate $\bparamt^\star$, stop the Newton-descent \ding{173} {\bf for} loop to end the current training epoch $\tau$.

From Eq.~(\ref{eq:armijo-1}), the deterministic 1st-order Armijo's rule for steepest descent can be written as:
\begin{align}
	J( \bparam_k + \learn_k \, \bd_k )
	\le
	J(\bparam_k)
	-
	\alpha \beta^a \rho
	\parallel \bd \parallel^2
	\ ,
	\text{ with } 
	\bd = - \bgrad
	\ ,
	\label{eq:armijo-4}
\end{align}
with $a$ being the minimum power for Eq.~(\ref{eq:armijo-4}) to be satisfied.  In Algorithm~\ref{algo:stochastic-newton}, the Armijo-like 2nd-order line search reads as follows:
\begin{align}
	\losst ( \bparamt_k + \learn_k \, \bdt_k )
	\le
	\losst (\bparamt_k)
	-
	\frac16
	(\beta^a)^3 \rho
	\parallel \bdt_k \parallel^3
	\ ,
	\label{eq:armijo-like-2nd-order}
\end{align}
with $a$ being the minimum power for Eq.~(\ref{eq:armijo-like-2nd-order}) to be satisfied.  The parallelism between Eq.~(\ref{eq:armijo-like-2nd-order}) and Eq.~(\ref{eq:armijo-4}) is clear; see also Table~\ref{tb:armijo-params-notations}.  

Figure~\ref{fig:newton-examples} shows the numerical results of Algorithm~\ref{algo:stochastic-newton} on the IJCNN1 dataset (\cite{Prokhorov.2001}) from the Library for Vector Support Machine (LIBSVM) library by \cite{Chang.2011}.  It is not often to see plots versus epochs side by side with plots versus iterations.  Some papers may have only plots versus iterations (e.g., \cite{Reddi.2019}); other papers may rely only on plots versus epochs to draw conclusions (e.g., \cite{Loshchilov.2019}).  Thus Figure~\ref{fig:newton-examples} provides a good example to see the differences, as noted in Remark~\ref{rm:epoch-vs-iteration}.  

\begin{rem}
	\label{rm:epoch-vs-iteration}
	Epoch counter vs global iteration counter in plots.
	{\rm
		When plotted gradient norm versus epochs (left of Figure~\ref{fig:newton-examples}), the three curves for \hyperref[sc:generic-SGD]{SGD} were separated, with faster convergence for smaller minibatch sizes Eq.~(\ref{eq:minibatch-0}), but the corresponding three curves fell on top of each other when plotted versus iterations (right of Figure~\ref{fig:newton-examples}).  The reason was the scale on the horizontal axis was different for each curve, e.g., 1 iteration for full batch was equivalent to 100 iterations for minibatch size at 1\% of full batch. While the plot versus iterations was the zoom-in view, but for each curve separately. To compare the rates of convergence among different algorithms and different minibatch sizes, look at the plots versus epochs, since each epoch covers the whole training set.  It is just an optical illusion to think that \hyperref[sc:generic-SGD]{SGD} with different minibatch sizes had the same rate of convergence.
	}
	$\hfill\blacksquare$
\end{rem}

The authors of \cite{Bergou.2018} planned to test their Algorithm~\ref{algo:stochastic-newton} on large datasets such as the CIFAR-10, and report the results in 2020.\footnote{
	Per our private correspondence as of 2019.12.18.
}

Another algorithm along the same line as Algorithm~\ref{algo:descent-armijo-stochastic-1} and Algorithm~\ref{algo:stochastic-newton} is the stochastic quasi-Newton method proposed in \cite{Wills.2018}, where the stochastic Wolfe line search of \cite{Mahsereci.2017} was employed, but with no numerical experiments on large datasets such as CIFAR-10, etc.

% line~\ref{lst:list:newton-armijo-rule} in Algorithm~\ref{algo:stochastic-newton}

% {\color{red} HERE 2019.12.14}

%\vspace{3mm}
\begin{algorithm}
	{\bf Stochastic Newton with 2nd-order line-search and fixed minibatch} (for training epoch $\tau$)
	\\
	\KwData{
		% Layer outputs $\byp{\ell}$, for $\ell=L, \cdots , 1$ 
		\\
		$\bullet$ Parameter estimate $\bparamt^\star_{\tau}$ from previous epoch $(\tau-1)$ (line~\ref{lst:list:SGD-params-previous-epoch} in Algorithm~\ref{algo:generic-SGD})
		\\
		$\bullet$ Select 3 Armijo paramters $\alpha \in (0,1)$, $\beta \in (0,1)$, $\rho >0 $, and {\color{purple} stability parameter $\delta > 0$}
	}
	\KwResult{
		Parameter estimate for next epoch $(\tau+1)$: $\bparamt_{\tau+1}^\star$. (line~\ref{lst:list:SGD-Armijo-result} in Algorithm~\ref{algo:descent-armijo-stochastic-1})
	}
	\vphantom{Blank line}
	%Define cost{\color{purple} -estimate function $\losst (\bparamt)$, as in Eq.~(\ref{eq:cost-estimate}) } 
	%\;
	%Define gradient{\color{purple} -estimate function $\bgradt (\bparamt)$, as in Eq.~(\ref{eq:gradient-estimate})}
	%\;
	%Define Hessian{\color{purple} -estimate function $\bHst (\bparamt)$, as in Eq.~(\ref{eq:hessian-estimate})}
	%\;
	$\blacktriangleright$ Begin training epoch $\tau$. 
	Set $\bparamt_1 = \bparamt_\tau^\star$
	\;
	\ding{173} \For{$k = 1,2, \ldots$}{
		
		Obtain minibatch $\Bbbps{k}{\bsize}$, Eq.~(\ref{eq:minibatch-2}) (line~\ref{lst:list:SGD-minibatch} in Algorithm~\ref{algo:generic-SGD})
		\;
		
		Compute gradient estimate $\bgradt_k = \bgradt (\bparamt_k)$, Eq.~(\ref{eq:gradient-estimate})
		\;
		
		Compute Hessian estimate $\bHst_k = \bHst(\bparamt_k)$, Eq.~(\ref{eq:hessian-estimate}), and its smallest eigenvalue $\Heigent_k$
		%{\color{red} HERE 2019.12.11}
		\;
				
		%\vphantom{Blank line}
		%\vspace{2mm}
			
		\eIf{Plateau or local minimizer, $\Heigent_k \ge - \delta^{1/2}$ and $\parallel \bgradt_k \parallel = 0$, Eq.~(\ref{eq:stochastic-newton-plateau}) \label{lst:line:newton-plateau}}{
			
			% THEN plateau
			$\blacktriangleright$ Take zero step length. Go to next iteration $(k+1)$, line~\ref{lst:line:newton-next-iteration}.
			
		}{ % ELSE plateau
		
			\If{Downward side of saddle point, $\Heigent_k < - \delta^{1/2}$, Eq.~(\ref{eq:stochastic-newton-negative-eigenvalue} \label{lst:line:newton-negative-eigenvalue})}{
				$\blacktriangleright$ Compute descent direction as eigenvector $\bHevect_k$ for $\Heigent_k$, Eq.~(\ref{eq:hessian-evect}): 
				%\\
				Set $\bdt_k = \bHevect_k$
				\;
			}
			
			\If{Convex bowl $\Heigent_k > \parallel \bgradt_k \parallel^{1/2}$ \label{lst:line:newton-descent}}{
				$\blacktriangleright$ Compute Newton descent direction, Eq.~(\ref{eq:newton-descent}): $\bHst_k \bdt_k = - \bgradt_k$, 
			}
			
			\If{$\bdt_k$ not yet selected, $-\delta^{1/2} \le \Heigent_k \le \parallel \bgradt_k \parallel^{1/2}$ \label{lst:line:newton-descent-not-selected-yet}}{
				$\blacktriangleright$ Compute regularized Newton descent direction, Eq.~(\ref{eq:newton-descent-regularized}) \label{lst:line:newton-descent-regularized}
			}
			
			% gradient descent
			\eIf{Stopping criterion not met \label{lst:list:newton-stop-not-met}}{
				
				% THEN, gradient descent 
				%\vspace{2mm}
				$\blacktriangleright$ Compute step length $\epsilon_k$ using Armijo's rule. \label{lst:list:newton-armijo-rule}
				\\
				% Set $\mu = \rho$
				Initialize step length $\epsilon_k \leftarrow \rho$
				\;
				% Begin For loop Armijo
				\ding{174} \For{a=1,2,\ldots}{
					
					% DO Armijo
					\eIf{Armijo decrease condition Eq.~(\ref{eq:armijo-like-2nd-order}) not satisfied}{
						
						% THEN, Armijo
						$\blacktriangleright$ Decrease step length:
						%\\
						Set $\epsilon_k \leftarrow \beta \epsilon_k$
						\;
						
					}{
						
						% ELSE, Armijo
						$\blacktriangleright$ Armijo decrease condition Eq.~(\ref{eq:armijo-like-2nd-order}) met.
						\\
						$\blacktriangleright$ Update network parameter $\bparamt_k$ to $\bparamt_{k+1}$:
						%\\
						Set $\bparamt_{k+1} = \bparamt_{k} + \epsilon_k \bdt_k$
						\;
						
					} % ENDIF, Armijo
					
					%\vspace{2mm}
					%{\color{red} HERE}
				
				} % END For loop Armijo

				%\vspace{2mm}
				
			}{
				
				% ELSE, gradient descent
				%\vspace{2mm}
				$\blacktriangleright$ Stopping criterion met.
				%\;
				Update minimizer estimate for epoch $(\tau+1)$: $\bparamt_{\tau+1}^\star \leftarrow \bparamt_k$
				\;
				Stop \ding{173} {\bf for} loop. (End training epoch $\tau$.)
				
			} % ENDIF, gradient descent
		
		} % ENDIF plateau

		\vspace{2mm}
		$\blacktriangleright$ Continue to next iteration $(k+1)$ \label{lst:line:newton-next-iteration}:
		%\\
		Set $k \leftarrow k+1$
		\;
	}

	\vphantom{Blank line} 
	\caption{
		\emph{Stochastic Newton method with 2nd-order Armijo-like line-search and fixed minibatch} (Sections~\ref{sc:armijo}, \ref{sc:stochastic-Newton}, Algorithms~\ref{algo:generic-SGD}, \ref{algo:descent-armijo-stochastic-1}).
		Adapted from \cite{Bergou.2018}.
		See Algorithm~\ref{algo:gradient-quasi-newton-armijo-deterministic} for \emph{deterministic} Newton and quasi-Newton algorithm with Armijo line search and fixed minibatch, Algorithm~\ref{algo:generic-SGD} for Standard SGD, Algorithm~\ref{algo:descent-armijo-stochastic-1} for SGD with Armijo line search and adaptive minibatch, and
		Table~\ref{tb:armijo-params-notations} for a comparison of the notations used by several authors in relation to the Armijo line search.		
	}
	\label{algo:stochastic-newton}
\end{algorithm}

At the time of this writing, due to lack of numerical results with large datasets commonly used in the deep-learning community such as CIFAR-10, CIFAR-100 and the likes, for testing, and thus lack of comparison of performance in terns of cost and accuracy against Adam and its variants, our assessment is that \hyperref[sc:generic-SGD]{SGD} and its variants, or \hyperref[sc:adam1]{Adam} and its better variants, particularly \hyperref[sc:adamw]{AdamW}, continue to be the prevalent methods for training.

Time constraint did not allow us to review other stochastic optimization methods such as that with the gradient-only line search in \cite{Kafka.2019} and \cite{Snyman.2018} could not be reviewed here. 

%{\color{red} HERE 2020.02.01}

% 2019.12.07, commented out this subsection on regularization
% \subsection{Regularization}

% \subsubsection{Penalty regularization}

% \subsubsection{Batch normalization}

% \subsubsection{Dropout}

%\input{08-empty}
% LSTM, recurrent neural network

% \section{Dynamics, recurrent neural networks}
% time series
\section{Dynamics, sequential data, sequence modeling}
\label{sc:recurrent}
% 2019.12.04
% Mark: 08-recurrent.tex, note on higher order derivatives

\subsection{Recurrent Neural Networks (RNNs)}
\label{sc:RNN}
In many fields of physics, the respective governing equations that describe the response of a system to (external) stimuli follow a common pattern. 
The temporal and/or spatial change of some quantity of a system is balanced by sources that cause the change, which is why we refer to equations of this kind as balance relations.
The balance of linear momentum in mechanics, for instance, establishes a relationship between the temporal change of a body's linear momentum and the forces acting on the body. 
Along with kinematic relations and constitutive laws, the balance equations provide the foundation to derive the equations of motion of some mechanical system.
For linear problems, the equations of motion constitute a system of second-order ODEs in appropriate (generalized) coordinates~$\dv$, which, in case of continua, is possibly obtained by some spatial discretization,
\begin{equation}
	% \Mm \ddot \dv + \Dm \dot \dv + \Cm \dv = \fv .
	\Mm \ddt \dv + \Dm \dt \dv + \Cm \dv = \fv .
	\label{eq:struct-dyn-1}
\end{equation}
In the above equation, $\Mm$ denotes the mass matrix, $\Dm$ is a damping matrix and $\Cm$ is the stiffness matrix; $\fv$ is the vector of (generalized) forces. 
The equations of motion can be rewritten as a system of first-order ODEs by introducing the vector of (generalized) velocities~$\vv$, 
%%
%\begin{equation}
%\dot \dv = \vv , \qquad
%\dot \vv = \Mm^{-1} \left( \fv - \Dm \vv - \Cm \dv  \right) ,
%\end{equation}
%>>>>>>> Stashed changes
%
\begin{equation}
	\begin{bmatrix}
	\dt \dv	\\ 	\dt \vv
	\end{bmatrix}
	=
	\underbrace{
	\begin{bmatrix}
	\boldsymbol 0	&	\I \\
	-\Mm^{-1} \Cm	&	-\Mm^{-1} \Dm
	\end{bmatrix}
	}_{\Am}
	\begin{bmatrix}
	\dv	\\ 	\vv
	\end{bmatrix}
	+ 
	\underbrace{
	\begin{bmatrix}
	\boldsymbol 0	\\	\Mm^{-1} 
	\end{bmatrix}
	}_{\Bm} 
	\fv
	\ .
	\label{eq:struct-dyn-2}
\end{equation}
In control theory, $\Am$ and $\bv = \Bm \fv$ are referred to as state matrix and input vector, respectively.\footnote{
	\label{fn:linear-time-invariant-system}
	The state-space representation of time-continuous LTI-systems in control theory, see, e.g., Chapter~3 of~\cite{Brogan.1990}, ``State Variables and the State Space Description of Dynamic Systems'' is typically written as 
%	\begin{equation*}
	$\dt \xv = \Am \xv + \Bm \uv$, with the output equation 
%	\qquad
	$\yv = \boldsymbol{C} \bx + \boldsymbol{D} \uv$.
%	\end{equation*}
	The state vector is denoted by $\xv$, $\uv$ and $\yv$ are the vectors of inputs and outputs, respectively.
	The ODE describes the (temporal) evolution of the system's state.
	The output of the system $\yv$ is a linear combination of states $\xv$ and the inputs $\uv$.
}
If we gather the (generalized) coordinates and velocities in the state vector $\qv$,  
\begin{equation}
	\qv = \begin{bmatrix}
	\dv^T 	&	\vv^T 
	\end{bmatrix}^T ,
\end{equation}
we obtain a compact representation of the equations of motion, which relates the temporal change of the system's state $\dt \qv$ to its current state $\qv$ and the input $\bv$ linearly by means of
\begin{equation} 
	\label{eq:eom-1st_order}
	\dt \qv %= f(\hv, \bv)
	= \Am \qv + \Bm \fv
	= \Am \qv + \bv .
\end{equation}
%
%{\color{red}
%	[NOTE: 2019.08.03.
%	the above equation
%	can be written as $\dt \hv = \Am \hv + \Bm \boldsymbol{u} = \Am \hv + \bv$, based on the re-written Eq.~(\ref{eq:struct-dyn-2}) with the matrix $\Bm$; see Footnote \ref{fn:linear-time-invariant-system}.
%	ENDNOTE]
%}
%
%
%{\color{blue}
%Note: Goodfellow doesn't use a bold symbol $f$, in mechanics, however, we would. On the other hand, $\fv$ is already used for the force vector
%}

%<<<<<<< Updated upstream
%{\color{red}
%NOTE: 2019.02.03.  you can specialize the Newmark method to the trapezoidal rule by selecting particular values for the parameters before proceeding.
%}
%
%=======
We 
%discover 
found
similar relations in neuroscience,\footnote{
See Section~\ref{sc:dynamic-volterra-series} on ``Dynamic, time dependence, Volterra series''.
} where the dynamics of neurons 
%is 
was
accounted for, e.g., in the pioneering work 
%
% CMES style rewriting
%of~
\cite{Hopfield.1984}, 
%who 
whose author
modeled a neuron as electrical circuit with capacitances.
%
% CMES style rewriting
%In his work on back-propagation, \cite{Pineda.1987} considered time-continuous RNNs. %, whose structure is close to what is considered canonical today.
Time-continuous RNNs were considered in a paper on back-propagation \cite{Pineda.1987}.
The temporal change of an RNN's state is related to the current state $\yv$ and the input $\xv$ by
%{\color{blue} 
%[NOTE: 2019.08.05.
%link to section on neuroscience
%ENDNOTE]
%}
%
\begin{equation}
\dt \yv = - \yv + a( \Wm \yv ) + \xv ,
\end{equation}
where $a$ denotes a non-linear activation function as, e.g., the sigmoid function, and $\Wm$ is the weight matrix that describes the connection among neurons.\footnote{
See the general time-continuous neural network with a continuous delay described by Eq.~\eqref{eq:continuous-RNN-general}.
}
%\footnote{
%{\color{blue}
%TODO: \cite{Dimirovski.2017}
%}
%}

%We discover similar equations in computational cybernetics, e.g., in studies on the response of recurrent artificial neural networks that have a time-varying delay, see~\cite{Dimirovski.2017},
%%
%>>>>>>> Stashed changes
%\begin{equation}
%\dot \zv = - \Am \zv + f (\Wm \zv (t - h(t)) + \jv) ,
%\end{equation}
%where $\zv$ is the state vector, $f$ denotes the activation function, matrices $\Am$, $\Wm$ represent the state matrix and connection weights, respectively, and $\jv$ is the bias.  
%The function $h(t)$ represents a time-depending delay.
%%

Returning to mechanics, we are confronted with the problem that the equations of motion do not admit closed-form solutions in general.
To construct approximate solution, time-integration schemes need to be resorted to, where we 
%exemplarily 
mention a few examples such as Newmark's method 
%(see~\cite{Newmark.1959}), 
\cite{Newmark.1959},
the Hilber-Hughes-Taylor (HHT) method 
%by~
\cite{Hilber.1977}, and 
the generalized-$\alpha$ method 
%by~
\cite{Chung.1993}.
For simplicity, we have a closer look at the classical trapezoidal rule, in which the time integral  of some function $f$ over a time step $\deltat = t_{n+1} - t_n$ is approximated by the average of the function values $f(t_{n+1})$ and $f(t_{n})$.
If we apply the trapezoidal rule to the system of ODEs in Eq.~\eqref{eq:eom-1st_order}, we obtain a system of algebraic relations that reads
\begin{equation}
\qvt{n+1} %= \hv_n + \frac{\deltat}{2} \left( f(\hv_n, \bv_n) + f(\hv_{n+1}, \bv_{n+1}) \right) 
= \qvt{n} + \frac{\deltat}{2} \left( \Am \qvt{n} + \bvt{n} + \Am \qvt{n+1} + \bvt{n+1} \right) .
\end{equation}
Rearranging the above relation for the new state gives the update equation for the state vector,
\begin{equation} \label{eq:eom-update_1}
\qvt{n+1} = \left( \I - \frac{\deltat}{2} \Am \right)^{-1} \left( \I + \frac{\deltat}{2} \Am \right) \qvt{n} + \frac{\deltat}{2} \left( \I - \frac{\deltat}{2} \Am \right)^{-1} \left( \bvt{n} + \bvt{n+1} \right) ,
\end{equation}
which determines the next state $\qvt{n+1}$ in terms of the state $\qvt{n}$ as well as the inputs $\bvt{n}$ and $\bvt{n+1}$.\footnote{
	We can regard the trapezoidal rule as a combination of Euler's explicit and implicit methods.
	The \emph{explicit Euler} method approximates time-integrals by means of rates (and inputs) at the beginning of a time step.
	The next state (at the end of a time step) is obtained from previous state and the previous input as
	%{\color{red}
	%[NOTE: 2019.07.25.  it is awkward to put equation numbers in the footnotes since the order of the equation numbers is out of whack.  it is best to use the command ``$\backslash$nonumber'' to remove the equation numbers in footnotes (which i did), or simply write the equations within the dollar signs.]
	%}
	%\begin{equation}
	$\qv_{n+1} = \qv_n + \deltat \, f(\qv_n, \bv_n) $
	%= \hv_n + \deltat \left( \Am \, \hv_n + \bv_n \right)
	$= \left( \I + \deltat \, \Am \right) \qv_n + \deltat \, \bv_n .$
	%\nonumber
	%\end{equation}
	On the contrary, the \emph{implicit Euler} method uses rates (and inputs) at the end of a time step, which leads to the update relation
%	\begin{equation}
	$\qv_{n+1} = \qv_n + \deltat \, f(\qv_{n+1}, \bv_{n+1}) $
	%= \hv_n + \deltat \left( \Am \, \hv_{n+1} + \bv_{n+1} \right)
	$= \left( \I - \deltat \, \Am \right)^{-1} \qv_n + \deltat \, \bv_{n+1} . $
%	\nonumber
%	\end{equation}
}
To keep it short, we introduce the matrices $\Wm$ and $\Um$ as
\begin{equation}
	\Wm = \left( \I - \frac{\deltat}{2} \Am \right)^{-1} \left( \I + \frac{\deltat}{2} \Am \right) , \qquad
	\Um = \frac{\deltat}{2} \left( \I - \frac{\deltat}{2} \Am \right)^{-1} ,
\end{equation}
which allows us to rewrite 
%relation
Eq.~\eqref{eq:eom-update_1} as
\begin{equation}
	\label{eq:eom-update_2} 
	\qvt{n+1} = \Wm \qvt{n} + \Um \left( \bvt{n} + \bvt{n+1} \right) .
\end{equation}

The update equation of time-discrete RNNs is similar to the discretized equations of motion
Eq.~\eqref{eq:eom-update_2}.
Unlike feed-forward neural networks, the state $\hv$ of an RNN at the $n$-th time step,\footnote{
Though the elements $\xvt{n}$ of a sequence are commonly referred to as ``time steps'', the nature of a sequence is not necessarily temporal.
The time step index $t$ then merely refers to a position within some given sequence.
}
which is denoted by $\hvt{n}$, does not only depend on the current input $\xvt{n}$, but also on the  state $\hvt{n-1}$ of the previous time step $n-1$. 
Following the notation 
%
% CMES style rewriting
%of~
in
\cite{Goodfellow.2016}, we introduce a \emph{transition function} $f$ that produces the new state,
\begin{equation}
	\hvt{n} = f ( \hvt{n-1}, \xvt{n}; \thetav ) .
	\label{eq:RNN-1}
\end{equation}

\begin{rem}
	\label{rm:hidden-cell}
	{\rm
		%
		% CMES style, rewriting
%		\cite{Goodfellow.2016} distinguish 
		In \cite{Goodfellow.2016}, there is a distinction
		between the ``hidden state'' of an RNN cell at the $n$-th step denoted by $\bh^{[n]}$, where ``$h$'' is mnemonic for ``hidden'', and the cell's output $\yvt{n}$.
		The output of a multi-layer RNN is a linear combination of the last layer's hidden state.
		Depending on the application, the output is not necessarily computed at every time step, but the network can ``summarize'' sequences of inputs to produce an output after a certain number of steps.\footnote{
		cf.~\cite{Goodfellow.2016}, p.~371, Figure~10.5.} 
%		the output of the RNN cell at state $[n]$, called the ``hidden state'' and denoted by $\bh^{[n]}$, where ``$h$'' is mnemonic for ``hidden'', can also be written as $\by^{[n]}$, consistent with the systematic use of $\by$ for outputs.
%		both notations $\bh^{[n]}$ and $\by^{[n]}$ are equivalent
		If the output is identical to the hidden state,
		\begin{align}
		\by^{[n]}
		\equiv
		\bh^{[n]} ,
		\label{eq:equiv-y-h-cell}
		\end{align}
		$\bh^{[n]}$ and $\by^{[n]}$ can be used interchangeably.
		In the current Section~\ref{sc:recurrent} on ``Dynamics, sequential data sequence modeling'',
		the notation $\bh^{[n]}$ is used, whereas in Section~\ref{sc:feedforward} on ``Static, feedforward networks'', the notation $\byp{\ell}$ is used to designate the output of the ``hidden layer'' $(\ell)$, keeping in mind the equivalence in Eq.~(\ref{eq:equiv-y-h}) in Remark~\ref{rm:hidden}.  
		whenever necessary, readers are reminded of the equivalence in Eq.~(\ref{eq:equiv-y-h-cell}) to avoid possible confusion when reading deep-learning literature.
	}
	\hfill$\blacksquare$ 
\end{rem}

\begin{figure}[H]
	\centering
	%
	% 2022.12.17
	% add "-eps-converted-to.pdf" for arXiv
	% \includegraphics[width=\linewidth]{./figures/discrete-RNN-1}
	\includegraphics[width=\linewidth]{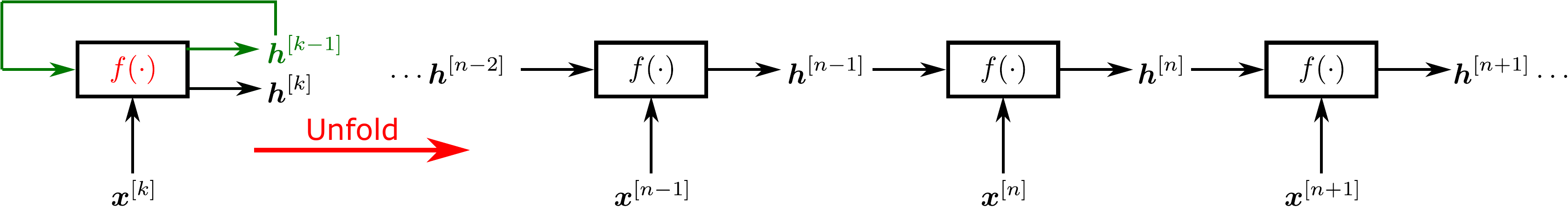}
	\caption{
		\emph{Folded and unfolded discrete RNN}
		(Section~\ref{sc:RNN}, \ref{sc:dynamic-volterra-series}).
		\emph{Left:}
		\emph{Folded discrete RNN}
		at configuration (or state) number $[k]$, where $k$ is an integer, with input $\bx^{[k]}$ to a multilayer neural network $f(\cdot) = f^{(1)} \circ f^{(2)} \circ \cdots \circ f^{(L)} (\cdot)$ as in Eq.~(\ref{eq:compositions}), having a feedback loop $\bh^{[k-1]}$ with delay by one step, to produce output $\bh^{[k]}$.
		\emph{Right:}
		\emph{Unfolded discrete RNN}, where the feedback loop is unfolded, centered at $k = n$, as represented by Eq.~(\ref{eq:RNN-1}).
		This graphical representation, with $f(\cdot)$ being a multilayer neural network, is more general than Figure~10.2 in
		\cite{Goodfellow.2016}, p.~366, and is a particular case of Figure~10.13b in \cite{Goodfellow.2016}, p.~388.
		See also the continuous RNN explained in Section~\ref{sc:dynamic-volterra-series} on ``Dynamic, time dependence, Volterra series'', Eq.~(\ref{eq:continuous-RNN-general}), Figure~\ref{fig:continuous-RNN-delay}, for which we refer readers to Remark~\ref{rm:hidden-cell} and the notation equivalence $\by^{[k]} \equiv \bh^{[k]}$ in Eq.~(\ref{eq:equiv-y-h-cell}).	
%		{\color{blue}
%		[NOTE: 2019.08.20: why do we distinguish $n$ and $k$?]
%		}
	}
	\label{fig:our-RNN}
\end{figure}

The above relation is illustrated as a circular graph in Figure~\ref{fig:our-RNN} (left), where the delay is explicit in the superscript of $\hvt{n-1}$.  
The hidden state $\hvt{n-1}$ at $n - 1$, in turn, is a function of the hidden state $\hvt{n-2}$ and the input $\xvt{n-1}$,
\begin{equation} 
\hvt{n} = f ( f (\hvt{n-2}, \xvt{n-1}; \thetav), \xvt{n}; \thetav ) .
%= f ( f ( \ldots ( f  (\hvt{0}, \xvt{1})
\label{eq:rnn}
\end{equation}
Continuing this \emph{unfolding} process repeatedly until we reach the beginning of a sequence, the recurrence can be expressed as a function $\gt{n}$,
\begin{align} 
	\hvt{n}  = 
	\gt{n} (\xvt{n} , \xvt{n-1}, \xvt{n-2}, \ldots , \xvt{2}, \xvt{1}, \hvt{0}; \thetav) =
	\gt{n} (\{ \xvt{k} , k = 1, \ldots, n \}, \hvt{0}; \thetav) ,	
	\label{eq:rnn_unfolded}
\end{align}
which takes the entire sequence up to the current step $n$, i.e., $\{ \xvt{k} , k = 1, \ldots, n \}$ (along with an initial state $\hvt{0}$ and parameters $\thetav$), as input to compute the current state $\hvt{n}$.
The unfolded graph representation is illustrated in Figure~\ref{fig:our-RNN} (right).

%\cite{Goodfellow.2016}
%\sout{introduce a black square to indicate a delay of a single step}.

%{\color{red}
%NOTE: 2019.07.21.  this delay can be made continuous in the continuous RNN as explained in Section~\ref{sc:dynamic-volterra-series} and Eq.~(\ref{eq:continuous-RNN-general}).
%ENDNOTE}

%
%\begin{figure}[H]
%	\centering
%	\includegraphics{./figures/Goodfellow-RNN}
%	\caption{
%		Discrete RNN with input $\xv$, hidden state $\hv$ and no outputs (left), represented by Eq.~(\ref{eq:RNN-1}).
%		The black square indicates a delay of one \sout{time} step, i.e., the hidden state is computed from the current input and the hidden state from the previous \sout{time} step. 
%		The same network can be represented as an unfolded graph (right), where each node corresponds to a particular \sout{time} step.
%		% {\color{blue}\cite[p.~366]{Goodfellow.2016}}
%		\cite{Goodfellow.2016}, p.~366.
%		see also the continuous RNN explained in Section~\ref{sc:dynamic-volterra-series}, Eq.~(\ref{eq:continuous-RNN-general}), Figure~\ref{fig:continuous-RNN-delay}.
%	}
%	\label{fig:Goodfellow-RNN}
%\end{figure} 

%2019.12.08
 %Mark: 08-recurrent.tex, Comment on RNN figure
 
%{\color{red}
%	[NOTES: 2019.07.31. for other choices of fonts, see DATES \hyperlink{2019.07.30. LaTeX fonts}{2019.07.30. LaTeX fonts}. ENDNOTE]
%}
%
As an example, consider the default \emph{(``vanilla'')} single-layer RNN provided by PyTorch\footnote{
See PyTorch documentation: Recurrent layers, \href{https://pytorch.org/docs/stable/nn.html\#recurrent-layers}{Original website}
(\href{https://web.archive.org/web/20221113090517/https://pytorch.org/docs/stable/nn.html\#recurrent-layers}{Internet archive})
}  
and TensorFlow\footnote{
See TensorFlow API: TensorFlow Core r1.14: tf.keras.layers.SimpleRNN, \href{https://www.tensorflow.org/api_docs/python/tf/keras/layers/SimpleRNN}{Original website}
(\href{https://web.archive.org/web/20220405022833/https://www.tensorflow.org/api_docs/python/tf/keras/layers/SimpleRNN}{Internet archive})
}, 
which is also described in~\cite{Goodfellow.2016}, p.~370:
\begin{align} 
	\label{eq:our-RNN-1}
%	\begin{aligned}
	\zvt{n} &= \bv + \Wm \hvt{n-1} + \Um \xvt{n} , \nonumber \\
	\hvt{n} &= a ( \zvt{n} ) = \tanh ( \zvt{n} ) .
%	\end{aligned}
\end{align}
First, $\zvt{n}$ is formed from an affine transformation of the current input $\xvt{n}$, the previous hidden state $\hvt{n-1}$ and the bias $\bv$ using weight matrices $\Um$ and $\Wm$, respectively. 
Subsequently, the hyperbolic tangent is applied to the elements of $\zvt{n}$ as activation function that produces the new hidden state $\hvt{n}$.

\begin{figure}[h]
	\centering
	%
	% 2022.12.17
	% add "-eps-converted-to.pdf" for arXiv
	% \includegraphics[width=\textwidth]{./figures/discrete-RNN-2b}
	\includegraphics[width=\textwidth]{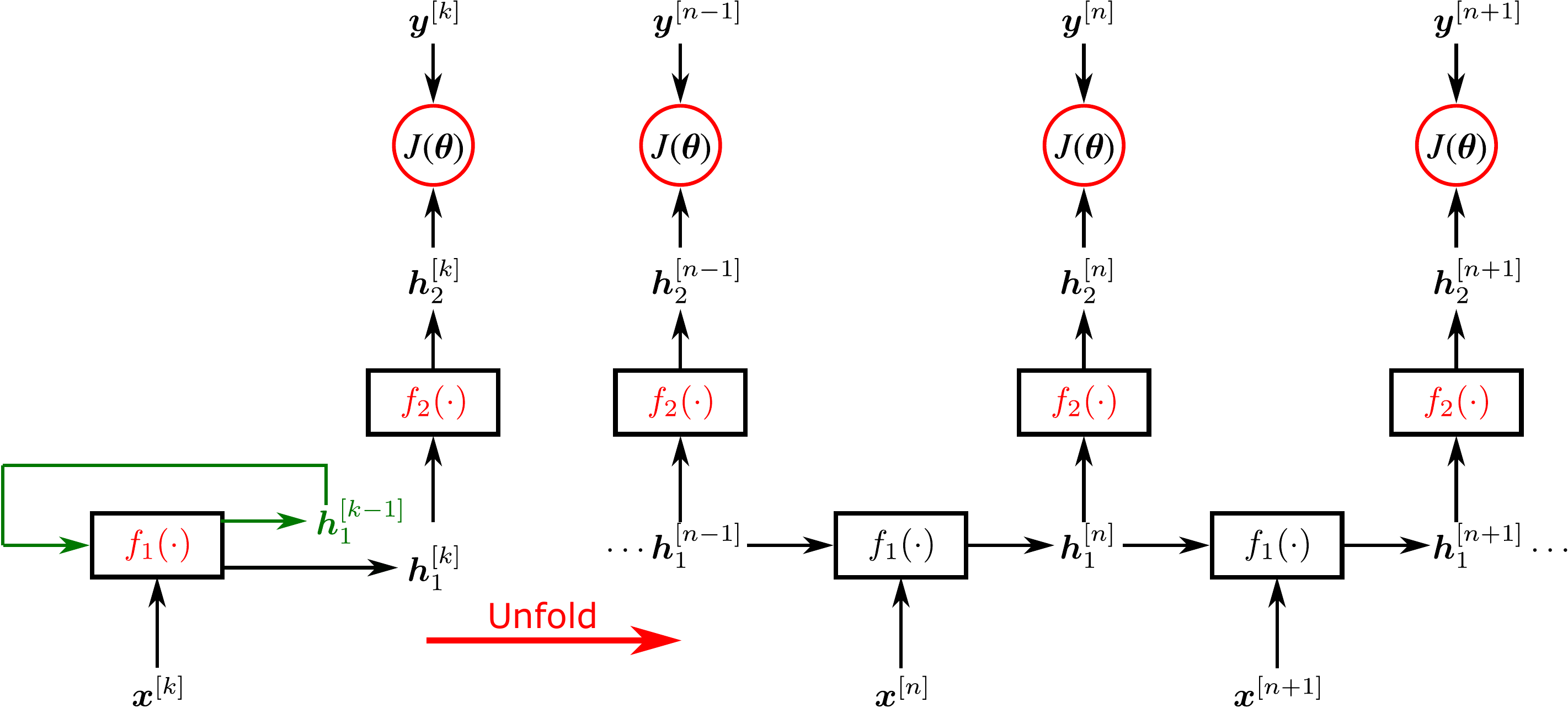}
	\caption{
		\emph{RNN with two multilayer neural networks (MLNs),} (Section~\ref{sc:RNN}) denoted by $f_1(\cdot)$ and $f_2(\cdot)$, whose outputs are fed into the loss function for optimization.  This RNN is a generalization of the RNN in Figure~\ref{fig:our-RNN}, and includes the RNN in Figure~10.3 in \cite{Goodfellow.2016}, p.~369, as a particular case, where Eq.~(\ref{eq:RNN_hidden_state}) of the first layer is simply $\bh_1^{[n]} = f_1 (\bx^{[n]}) = a_1 (\bz_1^{[n]})$, with $\bz_1^{[n]} = \bv + \Wm \hvt{n-1}_1 + \Um \xvt{n}$ and $a_1(\cdot) = \tanh(\cdot)$, whereas Eq.~(\ref{eq:RNN_hidden_state}) is $\bh_2^{[n]} = f_2 (\bx^{[n]}) = a_2 (\bz_2^{[n]})$, with $\bz_2^{[n]} = \cv + \Vm \hvt{n}_1$ and $a_2(\cdot) = \text{softmax}(\cdot)$ for the second layer. 
		The general relation for both MLNs is $\bh_j^{[n]} = f_j (\hvt{n-1}_{j}, \hvt{n}_{j-1}) = a_j (\bz_j^{[n]})$, for $j = 1,2$.
		%{\color{blue} NOTE: 2012.12.08: 
		%There are multiple mistakes in the graph: subscript missing for $\bh_j^{[n]}$; $\hvt{n-1}_{2}$ needs to be replaced by $\hvt{n+1}_{2}$
		%}
		%{\color{red} 2020.02.10. corrected.  DONE.}
	}
	\label{fig:our-RNN-2}
\end{figure}

A common design pattern of RNNs adds a linear output layer to the simplistic example in Figure~\ref{fig:our-RNN}, i.e., the RNN has a recurrent connection between its hidden units, which represent the state $\hv$,\footnote{
For this reason, $\hv$ is typically referred to as the \emph{hidden state} of an RNN.}
and produces an output at each time step.\footnote{
Such neural network is \emph{universal}, i.e., any function computable by a \emph{Turing machine} can be computed by an RNN of finite size, see~\cite{Goodfellow.2016}, p.~368.
}
Figure~\ref{fig:our-RNN-2} shows a two-layer RNN, which extends our above example by a second layer, i.e., the first layer is identical to Eq.~\eqref{eq:our-RNN-1},
\begin{align}
%	\begin{aligned}
	\zvt{n}_1 &= \bv + \Wm \hvt{n-1}_1 + \Um \xvt{n} , \nonumber \\
	\hvt{n}_1 &= a_1 ( \zvt{n}_1 ) = \tanh ( \zvt{n}_1 ) .
%	\end{aligned}
\end{align}
and the second layer forms a linear combination of the first layer's output $\hvt{n}_1$ and the bias $\cv$ using weights $\Vm$. Assuming the output $\hvt{n}_2$ is meant to represent probabilities, it is input to a $\softmax(\cdot)$ activation (a derivation of which is provided in Remark~\ref{rm:softmax} in Section~\ref{sc:classification}):
\begin{align}
%	\begin{aligned}
	\zvt{n}_2 &= \cv + \Vm \hvt{n}_1 , \nonumber \\
	\hvt{n}_2 &= a_2 ( \zvt{n}_2 ) 
	= \softmax ( \zvt{n}_2 ) 
	\ ,
%	\end{aligned}
\end{align}
which was given in Eq.~\eqref{eq:softmax}; see \cite{Goodfellow.2016}, p.~369.
The parameters of the network are the weight matrices $\Wm$, $\Um$ and $\Vm$, as well as the biases $\bv$ and $\cv$ of the recurrence layer and the output layer, respectively. 

%{\color{red} 2022/10/25 - we do we have a $\softmax$ function here? I think it creates more confusion here than necessary.}
%
%{\color{blue} [NOTE: 2022.11.06 - why would softmax be confusing?  See \cite{Goodfellow.2016}, Fig.~10.3, p.~369 (pdf p.~392), where softmax was used in the second layer.  ENDNOTE]}
% 2019.12.08
% Mark: 08-recurrent.tex: RNN classification, loss function

Irrespective of the number of layers, the hidden state $\hvt{n}_j$ of the $j$-th layer is gennerally computed from the previous hidden state $\hvt{n-1}_j$ and the input to the layer $\hvt{n}_{j-1}$,
%
%\begin{align}
%%\avt{n} &= \bv + \Wm \hvt{n-1} + \Um \xvt{n} , \\
%%\hvt{n} &= \tanh \avt{n} , \\
%\hvt{n} &= \tanh \left( \bv + \Wm \hvt{n-1} + \Um \xvt{n} \right) , \label{eq:RNN_h} \\[-1ex]
%\ovt{n} &= \cv + \Vm \hvt{n} , \label{eq:RNN_o}
%\end{align}

\begin{equation} \label{eq:RNN_hidden_state}
\hvt{n}_j = f_j(\hvt{n-1}_{j}, \hvt{n}_{j-1}) = a_j (\zvt{n}_j ) .%, \qquad \zvt{n}_j = \left( \bv + \Wm \hvt{n-1} + \Um \xvt{n} \right) , \label{eq:RNN_h} % \\[-1ex]
%\ovt{n} &= \cv + \Vm \hvt{n} , \label{eq:RNN_o}
\end{equation}
%
%where the hyperbolic tangent is used as activation function and, $\ovt{n}$ denotes the output of the linear (output) layer. % that has weights $\Vm$ and biases $\cv$ as parameters.
Other design patterns for RNNs show, e.g., recurrent connections between the hidden units but produce a single output only.
RNNs may also have recurrent connections from the output at one time step to the hidden unit of next time step.

Comparing the recurrence relation in Eq.~\eqref{eq:rnn} and its unfolded representation in Eq.~\eqref{eq:rnn_unfolded}, we can make the following observations:
\begin{itemize}
	\item The unfolded representation after $n$ steps $\gt{n}$ can be regarded as a factorization into repeated applications of $f$. 
	Unlike $\gt{n}$, the transition function $f$ does not depend on the length of the sequence and always has the same input size. 
	\item The same transition function $f$ with the same parameters $\thetav$ is used in every time step. 
	\item A state $\hvt{n}$ contains information about the whole past sequence.
\end{itemize}

%RNNs typically have a state $\hv$ that is represented by the hidden units of the network.

\begin{rem}
	\label{rm:depth-of-RNN}
	Depth of RNNs.
	{\rm
		For the above reasons and Figures~\ref{fig:our-RNN}-\ref{fig:our-RNN-2}, ``RNNs, once unfolded in time, can be seen as very deep feedforward networks in which all the layers share the same weights'' \cite{LeCun.2015:rd0001}.
		See Section~\ref{sc:depth} on network depth and Remark~\ref{rm:depth-definitions}.
	}
$\hfill\blacksquare$
\end{rem}

By nature, RNNs are typically employed for the processing of sequential data $\xvt{1}, \ldots, \xvt{\tau}$, where the sequence length $\tau$ typically need not be constant.
To process data of variable length, parameter sharing is a fundamental concept that characterizes RNNs. 
Instead of using separate parameters for each time step in a sequence, the same parameters are shared across several time-steps.
The idea of parameter sharing does not only allows us to process sequences of variable length (and possibly not seen during training), the \emph{``statistical strength''}\footnote{
see \cite{Goodfellow.2016}, p.~363.
} 
is also shared across different positions in time, which is important if relevant information occurs at different positions within a sequence.
A fully-connected feedforward neural network that takes each element of a sequence as input instead needs to learn all its rules separately for each position in the sequence. 

Comparing the update equations Eq.~\eqref{eq:eom-update_2} and Eq.~\eqref{eq:our-RNN-1}, we note the close resemblance of dynamic systems and RNNs.
%
% 2019.12.09:
% Mark: 08-recurrent.tex, graph dynamical systems
%
Let aside the non-linearity of the activation function and the presence of the bias vector, both have state vectors with recurrent connections to the previous states. 
Employing the trapezoidal rule for the time-discretization, we find a recurrence in the input, which is not present in the type of RNNs described above.
The concept of parameter sharing in RNNs translates into the notion of time-invariant systems in dynamics, i.e., the state matrix $\Am$ does not depend on time. 
In the computational mechanics context, typical outputs of a simulation could be, e.g., the displacement of a structure at some specified point or the von-Mises stress field in its interior.
The computations required for determining the output from the state (e.g., nodal displacements of a finite-element model) depend on the respective nature of the output quantity and need not be linear. 
 
%{\color{blue}Applications, examples, etc. }

The crucial challenge in RNNs is to learn long-term dependencies, i.e., relations among distant elements in input sequences.
For long sequences, we face the problem of vanishing or exploding gradients when training the network by means of back-propagation.
To understand vanishing (or exploding) gradients, we can draw analogies between RNNs and dynamic systems once again.
For this purpose, we consider an RNN without inputs whose activation function is the identity function: 
\begin{equation}
\hvt{n} = \Wm \hvt{n-1} .
\end{equation}
From the dynamics point of view, the above update equation corresponds to a linear autonomous system, whose time-discrete representation is given by
\begin{equation}
\hv_{n+1} = \Wm \hv_n .
\end{equation}
Clearly, the equilibrium state of the above system is the trivial state $\hv = \mathbf 0$.
Let $\hv_0$ denote a perturbation of the equilibrium state. 
An equilibrium state is called \emph{Lyapunov stable} if trajectories of the system, i.e., the states at times $t \geq 0$, remain bounded.
%\footnote{
%	{\color{red} NOTE: 2022.07.15 - Alex needs to complete this footnote.}
%TODO: notion of stability in RNNs~\cite{Goodfellow.2016}.
%BIBO stability with inputs
%}
If the trajectory eventually arrives at the equilibrium state for $t \rightarrow \infty$ (i.e., the trajectory is attractive), the equilibrium of the system is called \emph{asymptotically stable}.
In other words, an initial perturbation $\hv_0$ from the equilibrium state (i.e., the initial state) vanishes over time in the case of asymptotic stability.
Linear time-discrete systems are asymptotically stable if all eigenvalues $\lambda_i$ of $\Wm$ have an absolute value smaller than one.
From the unfolded representation in Eq.~\eqref{eq:rnn_unfolded} (see also Figure~\ref{fig:our-RNN} (right)), it is understood that we observe a similar behavior in the RNN described above.
At step $n$, the initial state $\hvt{0}$ has been multiplied $n$ times with the weight matrix $\Wm$, i.e.,
\begin{equation}
\hvt{n} = \Wm^n \hvt{0} .
\end{equation}
If eigenvalues of $\Wm$ have absolute values smaller than one, $\hvt{0}$ exponentially decays to zero in long sequences. 
On the other hand, we encounter exponentially increasing values for eigenvalues of $\Wm$ with magnitudes greater than one, which is equivalent to an unstable system in dynamics. 
When performing back-propagation to train RNNs, gradients of the loss function need to be passed backwards through the unfolded network, where  gradients are repeatedly multiplied with $\Wm$ as is the initial state $\hvt{0}$ in the forward pass.
The exponential decay (or increase) therefore causes gradients to vanish (or explode) in long sequences, which makes it difficult to learn long-term dependencies among distant elements of a sequence.

\begin{figure}[h]
	\centering
	\includegraphics[width=0.9\textwidth]{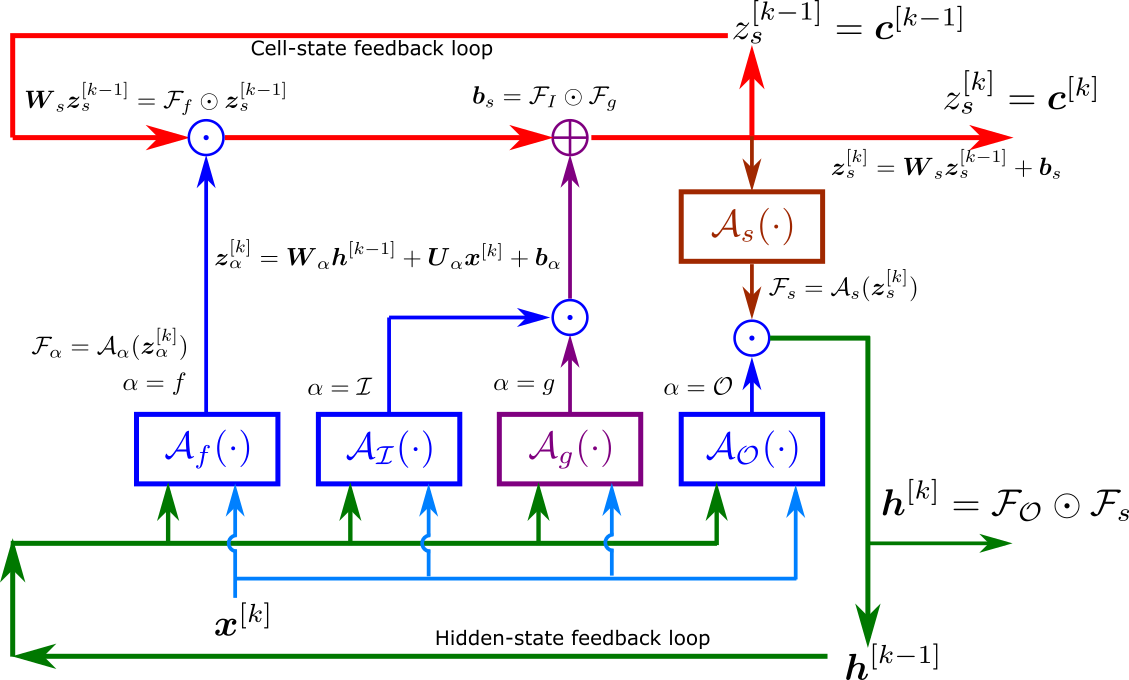}
	\caption{
		\emph{Folded Recurrent Neural Network (RNN) with Long Short-Term Memory (LSTM) cell} (Section~\ref{sc:LSTM}, \ref{sc:Wang-optimal-RNN}).  
		The cell state at $[k]$ is denoted by $\boldsymbol{z}_s^{[k]} \equiv \boldsymbol{c}^{[k]}$. Two feedback loops, one for cell state $\boldsymbol{z}_s$ and one for hidden state $\boldsymbol{h}$, with one-step delay $[k-1]$. 
		The key unified recurring relation is $\mathcal F_\alpha = \mathcal A_\alpha ( \boldsymbol{z_\alpha^{[k]}})$, with $\alpha \in \{ s \text{ (state)}, f \text{ (forget)}, \mathcal I \text{ (Input)}, g \text{ (external input)}, \mathcal O \text{ (Output)} \} $, where $\mathcal A_\alpha$ is a sigmoidal activation (squashing) function, and $\boldsymbol{z}_\alpha^{[k]}$ is a linear combination of some inputs with weights plus biases at cell state $[k]$.
		See Figure~\ref{fig:Olah-lstm_chain} for unfolded RNN with LSTM cells, and
		also Figure~\ref{fig:Wang-LSTM_cell} in Section~\ref{sc:Wang-Sun-2018}. 
%		and Figure~\ref{fig:Mohan-RNN-LSTM} in Section~\ref{sc:Mohan-2018}.
		%The LSTM cell state is also denoted by $\boldsymbol{z}_s^{[k]} \equiv \boldsymbol{c}^{[k]}$ for convenience.
	}
	\label{fig:our-lstm_cell}
\end{figure}

\subsection{Long Short-Term Memory (LSTM) unit}
\label{sc:LSTM}
%
%\begin{figure}[H]
%	\centering
%	\includegraphics[width=0.7\textwidth]{./figures/LSTM}
%	\caption{
%	{\color{blue}TODO}
%	}
%	\label{fig:eom_graph}
%\end{figure}

The vanishing (exploding) gradient problem prevents us from effectively learning long-term dependencies in long input sequences by means of conventional RNNs.
Gated RNNs as the long short-term memory (LSTM) and networks based on the gated recurrent unit (GRU) have proven to successfully overcome the vanishing gradient problem in diverse applications.
The common idea of gated RNNs is to create paths through time along which gradients neither vanish nor explode.
Gated RNNs can accumulate information in their state over many time steps, but, once the information has been used, they are capable to forget their state by, figuratively speaking, ``closing gates'' to stop the information flow.
This concept bears a resemblance to residual networks, which introduce skip connections to circumvent vanishing gradients in deep feed-forward networks; see Section~\ref{sc:architecture} on network ``Architecture''.
%{\color{blue}TODO: reference to the section on residual networks.}

\begin{rem}
	\label{rm:explain-short-term}
	\emph{\it What is ``short-term'' memory?}
	{\rm
	The vanishing gradient at the earlier states of an RNN (or layers in the case of a multilayer neural network) makes it that information in these earlier states (or layers) did not propagate forward to contribute to adjust the predicted outputs to track the labeled outputs, so to decrease the loss.  A state $[k]$ with two feedback loops is depicted in Figure~\ref{fig:our-lstm_cell}.  The reason for information in the earlier states not propagating forward was because the weights in these layers did not change much (i.e., did not learn), due to very small gradients, due to repeated multiplications of numbers with magnitude less than 1.  As a result, information in earlier states (or ``events'') played little role in decreasing the loss function, and thus had only ``short-term'' effects, rather than the needed long-term effect to be carried forward to the output layer.  Hence, we had a short-term memory problem.
	See also Remark~\ref{rm:vanish-gradient-MLP} in Section~\ref{sc:backprop} on back-propagation for vanishing or exploding gradient in multilayer neural networks. 
	} 
	\hfill$\blacksquare$
\end{rem}

%
% CMES style rewriting
In their pioneering work on LSTM,
the authors of 
%\cite{Hochreiter.1997:rd0001} presented a mechanism to forget states by introducing additional states, paths and self-loops to allow information (inputs, gradients) to flow over a long duration.
\cite{Hochreiter.1997:rd0001} presented a mechanism that allows information (inputs, gradients) to flow over a long duration by introducing additional states, paths and self-loops.
The additional components are encapsulated in so-called \emph{LSTM cells}. 
LSTM cells are the building blocks for LSTM networks, where they are connected recurrently to each other analogously to hidden neurons in conventional RNNs.
The introduction of a \emph{cell state}\footnote{
	The cell state is denoted with the variable $s$ in \cite{Goodfellow.2016}, p.~399, Eq.~(10.41).
} $\cv$ is one of the key ingredients to LSTM. 
The schematic cell representation (Figure~\ref{fig:our-lstm_cell}) shows that the cell state can propagate through an LSTM cell without much interference, which is why this path is described as ``conveyor belt'' for information 
%
% CMES style rewriting
%by~
in
\cite{Olah.2015}.

Another way to explain that could contribute to elucidate the concept is: Since information in RNN cannot be stored for a long time, over many subsequent steps, LSTM cell corrected this short-term memory problem by remembering the inputs for a long time:
\begin{quote}
	 ``A special unit called the memory cell acts like an accumulator or a gated leaky neuron: it has a connection to itself at the next time step that has a weight of one, so it copies its own real-valued state and accumulates 
	 the external signal, but this self-connection is multiplicatively gated by another unit that learns to decide when to clear the content of the memory.''
	 \cite{LeCun.2015:rd0001}.
\end{quote}

In a unified manner, the various relations in the original LSTM unit depicted in Figure~\ref{fig:our-lstm_cell} can be expressed in a single key generic recurring-relation that is more easily remembered:
\begin{align}
	\mathcal F_\alpha (\bx , \bh )= \mathcal A_\alpha ( \boldsymbol{z}_\alpha)
	\text{ with } \alpha \in \{ s \text{ (state)}, f \text{ (forget)}, \mathcal I \text{ (Input)}, g \text{ (external input)}, \mathcal O \text{ (Output)} \} 
	\ , 
	\label{eq:lstm-generic-relation}
\end{align}
where 
$\bx = \bx^{[k]}$ is the input at cell state $[k]$,
$\bh = \bh^{[k-1]}$ the hidden variable at cell state $[k-1]$,
$\mathcal A_\alpha$ (with ``$\mathcal A$'' being mnemonic for ``activation'') is a sigmoidal activation (squashing) function---which can be either the logistic sigmoid function  or the hyperbolic tangent function (see Section~\ref{sc:logistic-sigmoid})---and $\bz_\alpha = \boldsymbol{z}_\alpha^{[k]}$ is a linear combination of some inputs with weights plus biases at cell state $[k]$.
The choice of the notation in Eq.~\eqref{eq:lstm-generic-relation} is to be consistent with the notation in the relation $\widetilde{\boldsymbol y} = f ({\boldsymbol x}) = \g ({\boldsymbol W} {\boldsymbol x} + {\boldsymbol b}) = \g (\boldsymbol{z})$ in the caption of Figure~\ref{fig:neuron2}.

In Figure~\ref{fig:our-lstm_cell},
two types of squashing functions are used: One type (three blue boxes with $\alpha \in \{ f, \mathcal I, \mathcal O \}$) squashes inputs into the range $(0,1)$ (e.g., the logistic sigmoid, Eq.~\eqref{eq:logistic-sigmoid}), and the other type (purple box with $\alpha = g$, brown box with $\alpha = s$) squashes inputs into the range $(-1, +1)$ (e.g., the hyperbolic tangent, Eq.~\eqref{eq:hyperbolic-tangent}). 
The \emph{gates} are the activation functions $\mathcal A_\alpha$ with $\alpha \in \{ f, \mathcal I, g , \mathcal O \}$ (3 blue and 1 purple boxes), with argument containing the input $\bx^{[k]}$ (through $\bkarsp{z}{\alpha}{[k]}$).

\begin{rem}
	{\rm
		The activation function $\mathcal A_s$ (brown box in Figure~\ref{fig:our-lstm_cell}) is a hyperbolic tangent, but is not called a gate, since it has the cell state $\bkarp{c}{[k]}$, but not the input $\bx^{[k]}$, as argument.  In other words, a gate has to take in the input $\bx^{[k]}$ in its argument.
	} 
	$\hfill\blacksquare$
\end{rem}

There are two feedback loops, each with a delay by one step.
The cell-state feedback loop (red) at the top involves the LSTM cell state $\boldsymbol{z}_s^{[k-1]} \equiv \boldsymbol{c}^{[k-1]}$, with a delay by one step.  Since the cell state $\boldsymbol{z}_s^{[k-1]} \equiv \boldsymbol{c}^{[k-1]}$ is \emph{not} squashed by a sigmoidal activation function, vanishing or exploding gradient is avoided; the ``short-term'' memory would last longer, thus the name ``Long Short-Term Memory''. 
See Remark~\ref{rm:vanish-gradient-MLP} on vanishing or exploding gradient in back-propagation, and Remark~\ref{rm:explain-short-term} on short-term memory.

\begin{figure}[h]
	\centering
	\includegraphics[width=1\linewidth]{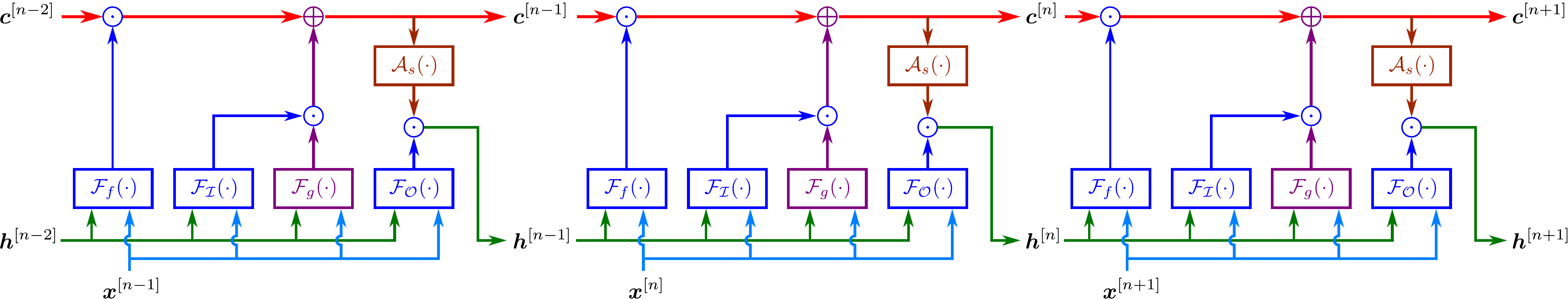}
	\caption{
		\emph{Unfolded RNN with LSTM cells} (Sections~\ref{sc:Wang-Sun-2018}, \ref{sc:LSTM}, \ref{sc:Mohan-POD}): 
		In this unfolded RNN, the cell states are centered at the LSTM cell $[k=n]$, preceded by the LSTM cell $[k=n-1]$, and followed by the LSTM cell $[k=n+1]$.
		See Eq.~(\ref{eq:lstm-cell-relation}) for the recurring relation among the successive cell states, and Figure~\ref{fig:our-lstm_cell} for a {\it folded} RNN with details of an LSTM cell.
		Unlike conventional RNNs, in which the hidden state is repeatedly multiplied with its shared weights, the additional cell state of an LSTM network can propagate over several time steps without much interference. For this reason, LSTM networks typically perform significantly better on long sequences as compared to conventional RNNs, which suffer from the problem of vanishing gradients when being trained. 
		See also 
		%Figure~\ref{fig:Mohan-RNN-LSTM} 
		Figure~\ref{fig:Mohan-LSTM-BiLSTM}
		in Section~\ref{sc:Mohan-reduced-POD}.
%		Section~\ref{sc:Mohan-2018}.
		%{\color{red} [NOTE: 2019.08.20, LSTM cell $[k=n-1]$, LSTM cell $[k=n]$, LSTM cell $[k=n+1]$.   2019.07.20. we need to give equation number in figure caption, such as the one added above, so to quickly identify which equation the figure corresponds to.]}
	}
	\label{fig:Olah-lstm_chain}
\end{figure}

In the hidden-state feedback loop (green) at the bottom, the combination $\boldsymbol{z}_{\mathcal O}^{[k]}$ of the input $\bx^{[k]}$ and the previous hidden state $\bh^{[k-1]}$ is squashed into the range $(0,1)$ by the processor $\mathcal F_{\mathcal O}$ to form a factor that filters out less important information from the cell state $\boldsymbol{z}_s^{[k]} \equiv \boldsymbol{c}^{[k]}$, which had been squashed into the range $(-1, +1)$.
See Figure~\ref{fig:Olah-lstm_chain} for an unfolded RNN with LSTM cells.
See also Appendix~\ref{app:another-lstm-block-diagram} and Figure~\ref{fig:lstm-cell-goodfellow} for an alternative block diagram.

As the term suggests, the presence of gates that control the information flow are a further key concept in gated RNNs and LSTM, in particular.
Gates are constructed from a linear layer with a sigmoidal function (logistic sigmoid or hyperbolic tangent) as activation function that squashes the components of a vector into the range $(0,1)$ (logistic sigmoid) or $(-1,+1)$ (hyperbolic tangent). 
A component-wise multiplication of the sigmoid's output with the cell state controls the evolution of the cell state (forward pass) and the flow of its gradients (backward pass), respectively. 
Multiplication with 0 suppresses the propagation of a component of the cell state, whereas a gate value of 1 allows a component to pass.

To understand the function of LSTM, we follow the paths information is routed through an LSTM cell. 
At time $n$, we assume that the hidden state $\hvt{n-1}$ and the cell state $\cvt{n-1}$ from the previous time step along with the current input $\xvt{n}$ are given. 
The hidden state $\hvt{n-1}$ and the input $\xvt{n}$ are inputs to the \emph{forget gate}, i.e., a fully connected layer with a sigmoid non-linearity, and the general expression Eq.~\eqref{eq:lstm-generic-relation}, with $k=n$ (see Figure~\ref{fig:Olah-lstm_chain}), becomes (Figure~\ref{fig:our-lstm_cell}, where the superscript $[k]$ on $\mathcal{F}_\alpha$ was omitted to alleviate the notation)
\begin{equation}
	\alpha = f 
	\Rightarrow
	\mathcal F_f^{[n]} = \mathcal A_f ( \boldsymbol{z}_f^{[n]})
	\rightarrow
	\fvt{n} = \sigmoid ( \zvt{n}_f ), \text{ with }
	\zvt{n}_f = \Wmf \hvt{n-1} + \Umf \xvt{n} + \bvf .
	%\fvt{n} = \sigmoid \left( \Wmf \hvt{n-1} + \Umf \xvt{n} + \bvf \right) .
	\label{eq:forget-gate}
\end{equation}

The weights associated with the hidden state and the cell state are $\Wmf$ and $\Umf$, respectively; the bias vector of the forget gate is denoted by $\bvf$.
The forget gate determines which and to what extent components of the previous cell state $\cvt{n-1}$ are to be kept subsequently.
Knowing which information of the cell state to keep, the next step is to determine how to update the cell state.
For this purpose, the hidden state $\hvt{n-1}$ and the input $\xvt{n}$ are input to a linear layer with a hyperbolic tangent as activation function, called the \emph{external input gate}, and the general expression Eq.~\eqref{eq:lstm-generic-relation}, with $k=n$, becomes (Figure~\ref{fig:our-lstm_cell})
\begin{equation}
	\alpha = g 
	\Rightarrow
	\mathcal F_g^{[n]} = \mathcal A_g ( \boldsymbol{z}_g^{[n]})
	\rightarrow
	\gvt{n} = \tanh ( \zvt{n}_g ) , \text{ with }
	\zvt{n}_g = \Wmg \hvt{n-1} + \Umg \xvt{n} + \bvg .
	%\gvt{n} = \tanh \left( \Wmg \hvt{n-1} + \Umg \xvt{n} + \bvg \right) .
\end{equation}
Again, $\Wmg$ and $\Umg$ are linear weights and $\bvg$ represents the bias.
The output vector of the $\tanh$ layer, which is also referred to as \emph{cell gate}, can be regarded as a candidate for updates to the cell state.

The actual updates are determined by a component-wise multiplication of the candidate values $\gvt{n}$ with the \emph{input gate}, which has the same structure as the forget gate but has its own parameters (i.e., $\Wmi$, $\Umi$, $\bvi$), and the general expression Eq.~\eqref{eq:lstm-generic-relation} becomes (Figure~\ref{fig:our-lstm_cell})
\begin{equation}
	\alpha = i 
	\Rightarrow
	\mathcal F_i^{[n]} = \mathcal A_i ( \boldsymbol{z}_i^{[n]})
	\rightarrow
	\ivt{n} = \sigmoid ( \zvt{n}_i ) , \text{ with }
	\zvt{n}_i = \Wmi \hvt{n-1} + \Umi \xvt{n} + \bvi \ .
	%\ivt{n} = \sigmoid \left( \Wmi \hvt{n-1} + \Umi \xvt{n} + \bvi \right) .
\end{equation}
The new cell state $\cvt{n}$ is formed by summing the scaled (by the forget gate) values of the previous cell state and scaled (by the input gate) values of the candidate values,
\begin{equation}
	\cvt{n} = \fvt{n} \odot \cvt{n-1} + \ivt{n} \odot \gvt{n} \ ,
	\label{eq:lstm-cell-relation}
\end{equation}
where the component-wise multiplication of matrices, which is also known by the name Hadamard product, is indicated by a ``$\odot$''.  

\begin{rem}
	\label{rm:lstm-cell-state}
	{\rm
		The path of the cell state $\cvt{k}$ is reminiscent of the identity map that jumps over some layers to create a residual map inside the jump in the building block of a residual network; see Figure~\ref{fig:resNet-basic} and also Remark~\ref{rm:residual-network} on the identity map in residual networks.
		$\hfill\blacksquare$
	}
\end{rem}

%{\color{red} NOTE: 2022.11.07 - I am HERE to connect all these gate expressions to the symbols used in the general expression  Eq.~\eqref{eq:lstm-generic-relation}, i.e., $\mathcal F_\alpha = \mathcal A_\alpha ( \boldsymbol{z}_\alpha^{[k]})$. ENDNOTE}

Finally, the hidden state of the LSTM cell is computed from the cell state $\cvt{n}$. 
The cell state is first squashed into the range $(-1, +1)$ by a hyperbolic tangent $\tanh$, i.e., the general expression Eq.~\eqref{eq:lstm-generic-relation} becomes (Figure~\ref{fig:our-lstm_cell})
\begin{align}
	\alpha = s 
	\Rightarrow
	\mathcal F_s^{[n]} 
	= \mathcal A_s ( \boldsymbol{z}_s^{[n]})
	= \mathcal A_s ( \boldsymbol{c}^{[n]})
	= \tanh \boldsymbol{c}^{[n]} \ , \text{ with }
	\boldsymbol{z}_s^{[n]} \equiv \boldsymbol{c}^{[n]} \ , 	
\end{align}
before the result $\mathcal F_s^{[n]}$ is multiplied with the output of a third sigmoid gate, i.e., the \emph{output gate}, for which the general expression Eq.~\eqref{eq:lstm-generic-relation} becomes (Figure~\ref{fig:our-lstm_cell})
\begin{equation}
	\alpha = \mathcal{O} 
	\Rightarrow
	\mathcal F_\mathcal{O}^{[n]} = \mathcal A_\mathcal{O} ( \boldsymbol{z}_\mathcal{O}^{[n]})
	\rightarrow
	\ovt{n} = \sigmoid ( \zvt{n}_o ) , \text{ with }
	\zvt{n}_o = \Wmo\hvt{n-1} + \Umo\xvt{n} + \bvo 
	\ .
	%\ovt{n} = \sigmoid \left( \Wmo\hvt{n-1} + \Umo\xvt{n} + \bvo \right) ,
\end{equation}

Hence, the output, i.e., the new hidden state $\hvt{n}$, is given by
\begin{equation}
\hvt{n}
= \mathcal F_\mathcal{O}^{[n]} \odot \mathcal F_s^{[n]}
=  \ovt{n} \odot \tanh \cvt{n} .
\end{equation}
For the LSTM cell, we get an intuition for respective choice of the activation function. 
The hyperbolic tangent is used to normalize and center information that is to be incorporated into the cell state or the hidden state.
The forget gate, input gate and output gate make use of the sigmoid function, which takes values between 0 and 1, to either discard information or allow information to pass by.
%{\color{blue}
%LSTM cell: additional internal recurrence + external recurrence of RNN
%}
%

%
%Comparing the unrolled graph of a conventional RNN (Figure~\ref{fig:Goodfellow-RNN_with_output}) with that of 

\begin{figure}[h]
	\centering
	\includegraphics[width=0.9\textwidth]{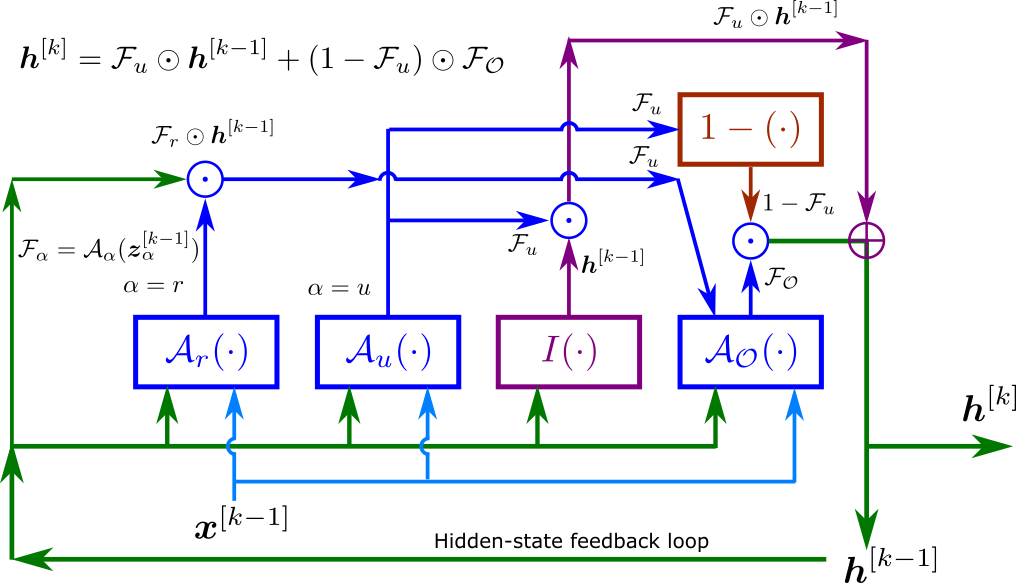}
	\caption{
		\emph{Folded RNN with Gated Recurrent Unit (GRU)} (Section~\ref{sc:GRU}).  
		The cell state at $[k-1]$, i.e., $(\bx^{[k-1]} , \bh^{[k-1]})$ are inputs to produce the hidden state  $\bh^{[k]}$. One feedback loop for the hidden state $\bh$, with one-step delay $[k-1]$. 
		The key unified recurring relation is $\mathcal F_\alpha = \mathcal A_\alpha ( \boldsymbol{z_\alpha^{[k-1]}})$, with $\alpha \in \{ r \text{ (reset)}, u \text{ (update)}, \mathcal O \text{ (Output)} \} $, where $\mathcal A_\alpha$ is a logistic sigmoi activation function, and $\bz_\alpha^{[k-1]}$ is a linear combination of some inputs with weights plus biases at cell state $[k-1]$.
		Compare to the LSTM cell in Figure~\ref{fig:our-lstm_cell}. 
		%		and Figure~\ref{fig:Mohan-RNN-LSTM} in Section~\ref{sc:Mohan-2018}.
		%The LSTM cell state is also denoted by $\boldsymbol{z}_s^{[k]} \equiv \boldsymbol{c}^{[k]}$ for convenience.
	}
	\label{fig:GRU-cell}
\end{figure}

\subsection{Gated Recurrent Unit (GRU)}
\label{sc:GRU}

%TO COMPLETE

In a unified manner, the various relations in the GRU depicted in Figure~\ref{fig:GRU-cell} can be expressed in a single key generic recurring-relation similar to the LSTM Eq.~\eqref{eq:lstm-generic-relation}:
\begin{align}
	\mathcal F_\alpha (\bx , \bh )= \mathcal A_\alpha ( \boldsymbol{z}_\alpha)
	\text{ with } \alpha \in \{ r \text{ (reset)}, u \text{ (update)}, \mathcal O \text{ (Output)} \} 
	\ , 
	\label{eq:gru-generic-relation}
\end{align}
where 
$\bx = \bx^{[k-1]}$ is the input at cell state $[k-1]$,
$\bh = \bh^{[k-1]}$ the hidden variable at cell state $[k-1]$,
$\mathcal A_\alpha$ the logistic sigmoid activation function, and $\bz_\alpha = \boldsymbol{z}_\alpha^{[k-1]}$ a linear combination of some inputs with weights plus biases at cell state $[k-1]$.

To facilitate a direct comparison between the GRU cell and the LSTM cell,
the locations of the boxes (gates) in the GRU cell in Figure~\ref{fig:GRU-cell} are identical to those in the LSTM in Figure~\ref{fig:our-lstm_cell}.  It can be observed that in the GRU cell (1) There is no feedback loop for the cell state, (2) The input is $\bx^{[k-1]}$ (instead of $\bx^{[k]}$), (3) The \emph{reset gate} replaces the LSTM forget gate, (4) The \emph{update gate} replaces the LSTM input gate, (5)  The \emph{identity map} (no effect on $\bh^{[k-1]}$) replaces the LSTM external input gate, (6) The complement of the update gate, i.e., $(1 - \mathcal F_u)$ replaces the LSTM state activation $\mathcal A_s = \tanh$.  There are fewer activations in the GRU cell compared to the LSTM cell.

The GRU was introduced in \cite{cho2014properties}, and tested against LSTM and $\tanh$-RNN in \cite{chung2014empirical}, with concise GRU schematics, and thus not easy to follow, unlike Figure~\ref{fig:GRU-cell}.  The GRU relations below follow \cite{Goodfellow.2016}, p.~400.

The hidden variable $\bh^{[n]}$, with $k=n$, is obtained from the GRU cell state at $[n-1]$, including the input $\bx^{[n-1]}$, and is a convex combination of $\bh^{[n-1]}$ and the GRU \emph{output-gate} effect $\mathcal F_\mathcal{O}^{[n-1]}$, using the GRU \emph{update-gate} effect $\mathcal{F}_u^{[n-1]}$ as coefficient
\begin{align}
	\bh^{[n]} = \mathcal{F}_u^{[n-1]} \odot \bh^{[n-1]} + (1 - \mathcal{F}_u^{[n-1]}) \odot \mathcal{F}_\mathcal{O}^{[n-1]}
	\ .
\end{align}

For the GRU \emph{update-gate} effect $\mathcal F_u^{[n-1]}$, the generic relation Eq.~\eqref{eq:gru-generic-relation} becomes (Figure~\ref{fig:GRU-cell}, where the superscript $[k-1]$ on $\mathcal{F}_\alpha$ was omitted to alleviate the notation), with $\alpha = u$ 
\begin{equation}
	\mathcal F_u^{[n-1]} = \mathcal A_u ( \boldsymbol{z}_u^{[n-1]})
	\rightarrow
	\bkars{u}{[n-1]} = \sigmoid (\bkarsp{z}{u}{[n-1]}) , \text{ with }
	\zvt{n-1}_u = \bkars{W}{u} \hvt{n-1} + \bkars{U}{u} \xvt{n-1} + \bkars{b}{u} 
	\ .
	%\ovt{n} = \sigmoid \left( \Wmo\hvt{n-1} + \Umo\xvt{n} + \bvo \right) ,
\end{equation}

For the GRU \emph{output-gate} effect $\mathcal F_\mathcal{O}^{[n-1]}$, the generic Eq.~\eqref{eq:gru-generic-relation} becomes (Figure~\ref{fig:GRU-cell}), with $\alpha = \mathcal{O}$, 
\begin{equation}
	\mathcal F_\mathcal{O}^{[n-1]} = \mathcal A_\mathcal{O} ( \boldsymbol{z}_\mathcal{O}^{[k-1]})
	\rightarrow
	\ovt{n} = \sigmoid ( \zvt{n-1}_o ) , \text{ with }
	\zvt{n-1}_o = \Wmo (\mathcal F_r^{[n-1]} \odot \hvt{n-1}) + \Umo\xvt{n-1} + \bvo 
	.
	%\ovt{n} = \sigmoid \left( \Wmo\hvt{n-1} + \Umo\xvt{n} + \bvo \right) ,
\end{equation}

For the GRU \emph{reset-gate} effect $\mathcal F_r^{[n-1]}$, the generic Eq.~\eqref{eq:gru-generic-relation} becomes (Figure~\ref{fig:GRU-cell}), with $\alpha = r$ 
\begin{equation}
	\mathcal F_r^{[n-1]} = \mathcal A_r ( \boldsymbol{z}_r^{[n-1]})
	\rightarrow
	\bkarp{r}{[n-1]} = \sigmoid ( \zvt{n-1}_r ) , \text{ with }
	\zvt{n-1}_r = \bkars{W}{r} \hvt{n-1} + \bkars{U}{r} \xvt{n-1} + \bkars{b}{r} 
	\ .
	%\ovt{n} = \sigmoid \left( \Wmo\hvt{n-1} + \Umo\xvt{n} + \bvo \right) ,
\end{equation}

%Refer to Remark~\ref{rm:attention-kernel-PINN}

\begin{rem}
	{\rm
		GRU has fewer activation functions compared to LSTM, and is thus likely to be more efficient, even though it was stated in \cite{chung2014empirical} that no concrete conclusion could be made as to ``which of the two gating units was better.''  See Remark~\ref{rm:attention-kernel-PINN} on the use of GRU to solve hyperbolic problems with shock waves.
	}
	$\hfill\blacksquare$
\end{rem}

%{\color{red} NOTE: 2022.11.10 - I am HERE.  ENDNOTE}

%\subsection{Sequence-to-sequence models, Attention, Transformer unit}
\subsection{Sequence modeling, attention mechanisms, Transformer}
\label{sc:sequence-modeling_attention_transfomer}

%\noindent
%{\color{red} [NOTE: 2022.11.04 - I changed the section title to follow closely the language used in \cite{Vaswani.2017:rd0001}.
%	
%It would help readers to have a short paragraph, right at the beginning of this section, describing the relationship between sequence modeling, attention mechanism, and Transformer architecture, before diving into the details of each topic.  The two main references for attention mechanism are cited in \cite{Vaswani.2017:rd0001}:
%
%%\cite{bahdanau2014neural} 
%\cite{Bahdanau.2015} = [2] Dzmitry Bahdanau, Kyunghyun Cho, and Yoshua Bengio. Neural machine translation by jointly learning to align and translate. CoRR, abs/1409.0473, 2014.	
%
%\cite{kim2017structured} = [19] Yoon Kim, Carl Denton, Luong Hoang, and Alexander M. Rush. Structured attention networks, in International Conference on Learning Representations, 2017.  
%
%I added below a short opening paragraph for this section, based on \cite{Vaswani.2017:rd0001}. 
%
%I also suggest we have two subsubsections, as shown below and in the Table of Contents.
%
%\noindent
%ENDNOTE]}

RNN and LSTM have been well-established for use in sequence modeling and ``transduction'' problems such as language modeling and machine translation.  Attention mechanism, introduced in 
%\cite{bahdanau2014neural} 
\cite{Bahdanau.2015}
\cite{kim2017structured}, allowed for ``modeling of dependencies without regard to their distance in the input and output sequences,'' and has been used together with RNNs \cite{Vaswani.2017:rd0001}.   Transformer is a much more efficient architecture that only uses an attention mechanism, but without the RNN architecture, to ``draw global dependencies between input and output.''   
%We explain in details each concept in what follows.
Each of these concepts is discussed in detail below.

\subsubsection{Sequence modeling, encoder-decoder}
\label{sc:sequence-encoder-decoder-attention}
%{\color{red} NOTE: 2022.11.04 - This subsubsection was more about encoder-decoder, but little about the key player ``attention mechanism,'' which should be defined and mentioned early on in this subsubsection, not until the end.  I added the sentence in red.  Ref. \cite{kim2017structured}, which is more recent and well regarded (see the OpenReview link in \cite{kim2017structured}), should also be used. ENDNOTE}

%RNNs and LSTM, in particular, have become well-established methods in sequence-to-sequence problems as, e.g., machine translation. 
The term \emph{neural machine translation} describes the approach of using a single neural network to translate a sentence.\footnote{
	See, e.g., \cite{Bahdanau.2015}.
}
Machine translation is a special kind of sequence-to-sequence modeling problem, in which a \emph{source sequence} is ``translated'' into a \emph{target sequence}. 
Neural machine translation typically relies on \emph{encoder--decoder} architectures.\footnote{
	\emph{Autoencoders} are a special kind of encoder-decoder networks, which are trained to reproduce the input sequence, see Section~\ref{sc:autoencoder}.
}  

The encoder network converts (encodes) the essential information of the input sequence $\sourceseq{} = \{ \xvt{k}, k = 1, \ldots, n \}$ into an intermediate, typically fixed-length vector representation, which is also referred to as \emph{context}.
The decoder network subsequently generates (decodes) the output sequence $\targetseq{} = \{ \yvt{k}, k = 1, \ldots, n \}$ from the encoded intermediate vector.
The intermediate context vector in the encoder--decoder structure allows for input and output sequences of different length.
Consider, for instance, an RNN composed from LSTM cells (Section \ref{sc:LSTM}) as encoder network that generates the intermediate vector by sequentially processing the elements (words, characters) of the input sequence (sentence).
The decoder is a second RNN that accepts the fixed-length intermediate vector generated by the encoder, e.g., as the initial hidden state, and subsequently generates the output sequence (sentences, words) element by element (words, characters).\footnote{
	For more details on different types of RNNs, see, e.g., \cite{Goodfellow.2016}, p.~385, Section 10.4 ``Encoder-decoder sequence-to-sequence architectures.''
}

To make things clearer, we briefly sketch the structure of a typical RNN encoder-decoder model following \cite{Bahdanau.2015}. 
The encoder $\encoder$ is composed from a recurrent neural network $\encoderrnn$ and  some non-linear feedforward function $\encoderff$. 
In terms of our notation, let $\xvt{k}$ denote the $k$-th element in the sequence of input vectors, $\sourceseq{}$, and let $\encoderhidden{k-1}$ denote the corresponding hidden state at time $k-1$, the hidden state at time $k$ follows from the recurrence relation (Figure~\ref{fig:our-RNN})
\begin{equation} \label{eq:encoder_rnn}
%	h_{t} = f\left(x_{t}, h_{t-1}\right)
	\encoderhidden{k} = \encoderrnn ( \encoderhidden{k-1}, \xvt{k} ) .
\end{equation} 
%The non-linear function $f$ referred to as transition function in Section~\ref{sc:RNN}. 
In Section~\ref{sc:RNN}, we referred to $\encoderrnn$ as the transition function (denoted by $f$ in Figure~\ref{fig:our-RNN}).
%{\color{red} NOTE: 2022.11.05 - I expanded the last sentence for clarity; please check.  ENDNOTE}
%AH 2022.11.16 - no, that is not correct. the recurrent network represents the transition function, not the feed-forward one
The context vector $\cv$ is generated by another generally non-linear function $\encoderff$, which takes the sequence of hidden states $\xvt{k}$, $k=1,\ldots,n$ as input:
% 2021.01.22
%\newcommand{\enc}{\mathfrak E}
%\newcommand{\be}{\bkar \enc}
%\newcommand{\besp}[2]{\bkarsp \enc {#1} {#2}}
%
%$\be$
%$\besp {blah} {bob}$
\begin{equation}
	\cv = \encoderff (\{ \hvt{k} , k=1,\ldots,n \} ) .
\end{equation}
Note that $\encoderff$ could as well be a function of just the final hidden state in the encoder-RNN.\footnote{
	See, e.g., \cite{Goodfellow.2016}, p.~385.}

From a probabilistic point of view, the joint probability of the entire output sequence (i.e., the ``translation'') $\outseq$ can be decomposed into conditional probabilities of the $k$-th output item $\yvt{k}$ given its predecessors $\yvt{1}, \ldots, \yvt{k-1}$ and the input sequence $\inpseq$, which in term can be approximated using the context vector $\cv$:
\begin{equation}
	P( \yvt{1}, \ldots, \yvt{n} ) 
	= \prod_{k=1}^{n} P ( \yvt{k} \mid \yvt{1}, \ldots, \yvt{k-1} , \mathbf x )
	\approx \prod_{k=1}^{n} P ( \yvt{k} \mid \yvt{1}, \ldots, \yvt{k-1} , \cv ) .
\end{equation}
%
%\emph{``defines a probability over the translation by decomposing the joint probability into the ordered conditionals"}, see~\cite{}.
Accordingly, the decoder is trained to predict the next item (word, character) in the output sequence $\yvt{k}$ given the previous items $\yvt{1}, \ldots, \yvt{k-1}$ and the context vector $\cv$.
In analogy to the encoder $\encoder$, the decoder $\decoder$ comprises a recurrence function $\decoderrnn$ and a non-linear feedforward function $\decoderff$. 
%
%\begin{equation}
%	p(\mathbf{y})=\prod_{t=1}^{T} p\left(y_{t} \mid\left\{y_{1}, \cdots, y_{t-1}\right\}, c\right)
%\end{equation}
%
Practically, RNNs provide an intuitive means to realize functions of variable-length sequences, since the current hidden state in RNNs contains information of all previous inputs. 
Accordingly, the decoder's hidden state $\decoderhidden{k}$ at step $k$ follows from the recurrence relation $\decoderrnn$ as
\begin{equation}
	\decoderhidden{k} = \decoderrnn (\decoderhidden{k-1}, \yvt{k-1}, \cv).
	\label{eq:decoder_hidden_state}
\end{equation}
To predict the conditional probability of the next item $\yvt{k}$ by means of the function $\decoderff$, the decoder can therefore use only the previous item $\yvt{k-1}$ and the current hidden state $\mathbf{s}^{[k]}$ as inputs (along with the context vector $\cv$) rather than all previously predicted items $\yvt{1}, \ldots, \yvt{k-1}$:
%The hidden state of an RNN contains information of all previous 
%Instead of all previous items in the 
%Suppose the decoder is a RNN, it predicts the conditional probability of the next item $\yvt{k}$ from the previous item $\yvt{k-1}$, its  hidden state $\mathbf{s}^{[k]}$ and the context vector $\cv$:
%
\begin{equation} 
	P ( \yvt{k} \mid \yvt{1}, \ldots, \yvt{k-1} , \cv ) = \decoderff (\yvt{k-1}, \decoderhidden{k}, \cv) .
	\label{eq:decoder_output_probability}
\end{equation}
We have various choices of how the context vector $\cv$ and inputs $\yvt{k}$ are fed into to the decoder-RNN.\footnote{
See, e.g., \cite{Goodfellow.2016}, p.~386.
}
The context vector $\cv$, for instance, can either be used as the decoder's initial hidden state $\decoderhidden{1}$ or, alternatively, as the first input.

%
% CMES style rewriting
\subsubsection{Attention}
\label{sc:attention}
%The \emph{attention mechanism} is implemented in the decoder \cite{Bahdanau.2015}.

As the authors of \cite{Bahdanau.2015} emphasized, the encoder \emph{``needs to be able to compress all the necessary information of a source sentence into fixed-length vector''}.
For this reason, long sentences pose a challenge in neural machine translation, in particular, as sentences to be translated are longer than the sentences  networks have seen during training, which was confirmed by the observations 
%
% CMES style rewriting
%of~
in
\cite{Cho.2014}.
To cope with long sentences,
%
% CMES style rewriting 
%\cite{Bahdanau.2015} proposed 
an encoder--decoder architecture, \textit{``which learns to align and translate jointly,''}  was proposed in \cite{Bahdanau.2015}.
Their approach is motivated by the observation that individual items of the target sequence correspond to different parts of the source sequence. 
To account for the 
%``distance'' among inputs and outputs items in the respective sequences, 
fact that only a subset of the source sequence is relevant when generating a new item of the target sequence,
%
% CMES style rewriting
%\cite{Bahdanau.2015} introduce 
two key ingredients (alignment and translation) to the conventional encoder--decoder architecture described above were introduced in \cite{Bahdanau.2015}, and will be presented below.
%
%{\color{blue} TODO: bi-directional RNN replaces RNN of the encoder; we obtain annotations rather than hidden states}
%

%{\color{red} [NOTE: 2022.11.04 - I recalled I mentioned this point before: If there is the first ingredient, there should be the second ingredient to avoid confusion.  Where is the second ingredient?  OK, I saw ``a second key ingredient'' above Eq.~\eqref{eq:alignment_model}, so I added the word ``key'' to the first ingredient, and changed ``a'' to ``the'' for ``the second key ingredient,'' unless there was another ``second ingredient''; please check.  I also italicized the first and second key ingredients to help with the search. ENDNOTE]}

% AH: 2022.11.07: OK, so I removed the note

\ding{42}
The \emph{first key ingredient} to their concept of ``alignment'' 
%that only a subset of the input sequence is relevant when generating a new item of the output sequence is realized by 
is the idea of 
using a distinct context vector $\cv_k$ for each output item $\yvt{k}$ instead of a single context $\cv$.
%Formally, this idea is reflected in the context  
Accordingly, the recurrence relation of the decoder,  Eq.~\eqref{eq:decoder_hidden_state}, is modified and takes the context $\cv_k$ as argument
%
%{\color{red} [NOTE: 2022.07.19 - After the phrase ``The first ingredient'' was used, readers would expect to see ``The second ingredient'', and the ``The third ingredient'' etc. It is confusing not to see ``The second ingredient'', which should be indentified explicitly in subsequent text. ENDNOTE]}
%
%\begin{equation}
%	\decoderhidden{k} = g (\decoderhidden{k-1}, \yvt{k-1}, \cv_k ) ,
%\end{equation}
%
\begin{align}
	\decoderhidden{k} = g (\decoderhidden{k-1}, \yvt{k-1}, \cv_k ) ,
\end{align}
as is the conditional probability of the output items, Eq.~\eqref{eq:decoder_output_probability},
\begin{equation}
	P ( \yvt{k} \mid \yvt{1}, \ldots, \yvt{k-1} , \inpseq) 
	= P ( \yvt{k} \mid \yvt{1}, \ldots, \yvt{k-1} , \cv_k )
	\approx \decoderff (\yvt{k-1}, \decoderhidden{k}, \cv_k) ,
\end{equation}
i.e., it is conditioned on distinct context vectors $\cv_k$ for each output $\yvt{k}$, $k=1, \ldots, n$.

The $k$-th context vector $\cv_k$ is supposed to capture the information of that part of the source sequence $\inpseq$ which is relevant to the $k$-th target item $\yvt{k}$.
For this purpose, $\cv_k$ is computed as weighted sum of all hidden states $\encoderhidden{l}$,  $l=1, \ldots, n$ of the encoder: 
%
%
%
%The $k$-th context $\cv_k$, in turn, is computed as weighted sum of all hidden states $\encoderhidden{l}$,  $l=1, \ldots, n$ of the encoder:\footnote{
	%		\cite{Bahdanau.2015} also referred to the hidden states $\encoderhidden{k}$ as ``annotations''.
	%}
\begin{equation} 
	\cv_k = \sum_{l=1}^n \alignmentweights \encoderhidden{l} .
	\label{eq:attention_weighting}
\end{equation}

The $k$-th hidden state of a conventional RNN obeying the recurrence given by Eq.~\eqref{eq:encoder_rnn} only includes information about the preceding items ($1, \ldots, k$) in the source sequence, since the remaining items ($k+1, \ldots, n$) still remain to be processed. 
When generating the $k$-th output item, however, we want information about all source items, before \emph{and} after, to be contained in the $k$-th hidden state $\encoderhidden{k}$.

For this reason, 
%
% CMES style rewriting
%\cite{Bahdanau.2015} proposed 
using a \emph{bidirectional RNN}\footnote{
	See \cite{Schuster.1997}.
} as encoder was proposed in \cite{Bahdanau.2015}. 
A bidirectional RNN combines two RNNs, i.e., a \emph{forward RNN} and \emph{backward RNN}, which independently process the source sequence in the original and in reverse order, respectively.
The two RNNs generate corresponding sequences of forward and backward hidden states,
\begin{equation}
	\encoderhiddenfwd{k} = \encoderrnnfwd ( \encoderhiddenfwd{k-1}, \xvt{k} ) , \qquad
	\encoderhiddenrev{k} = \encoderrnnrev ( \encoderhiddenrev{k-1}, \xvt{n-k} ) .
\end{equation}
In each step, these vectors are concatenated to a single hidden state vector $\encoderhidden{k}$, which the authors of \cite{Bahdanau.2015} refer to as ``\emph{annotation}'' of the $k$-th source item:
\begin{equation}
	\encoderhidden{k} = \left[ \left( \encoderhiddenfwd{k} \right)^T , \left( \encoderhiddenrev{k} \right)^T \right]^T .
\end{equation}
They mentioned \emph{``the tendency of RNNs to better represent recent inputs''} as reason why the annotation $\encoderhidden{k}$ focuses around the $k$-th encoder input $\xvt{k}$.

\ding{42}
As the \emph{second key ingredient},  
%
% CMES style rewriting
the authors of 
\cite{Bahdanau.2015} proposed a so-called \emph{``alignment model''}, i.e., a function $\alignmentfun$ to compute weights $\alignmentweights$ needed for the context $\cv_k$, Eq.~\eqref{eq:attention_weighting},
\begin{equation}
	\alignment = \alignmentfun ( \decoderhidden{k-1}, \encoderhidden{l}) ,
	\label{eq:alignment_model}
\end{equation}
which is meant to quantify (``score'') the relation, i.e., the \emph{alignment}, between the $k$-th decoder output (target) and inputs \emph{``around''} the $l$-th position of the source sequence. 
The score is computed from the decoder's hidden state of the previous output $\decoderhidden{k-1}$ and the annotation $\encoderhidden{l}$ of the $l$-th input item, so it \emph{``reflects the importance of the annotation $\encoderhidden{l}$ with respect to the previous hidden state $\decoderhidden{k-1}$ in deciding the next state $\decoderhidden{k}$ and generating $\yvt{k}$.''}

The alignment model is represented by a feedforward neural network, which is jointly trained along with all other components of the encoder--decoder architecture.
The weights of the annotations, in turn, follow from the alignment scores upon exponentiation and normalization (through $\softmax(\cdot)$ (Section~\ref{sc:classification}) along the second dimension):
%For this purpose, a feedforward neural network is jointly trained with all components of the encoder--decoder architecture. 
\begin{equation}
	\alignmentweights = \frac{\exp \left( \alignment \right)}{\sum_{j=1}^{n} \exp \left( \aligments{kj} \right)} .
\end{equation}
%
% CMES style rewriting
%\cite{Bahdanau.2015} interpreted 
The weighting in Eq.~\eqref{eq:attention_weighting} is interpreted as a way \emph{``to compute an expected annotation, where the expectation is over possible alignments"} \cite{Bahdanau.2015}. 
From this perspective, $\alignmentweights$ represents the probability of an output (target) item $\yvt{k}$ being aligned to an input (source) item $\xvt{l}$. 

In neural machine translation %(English-to-French translation), 
%
% CMES style rewriting 
%\cite{Bahdanau.2015} could show 
it was possible to show
that 
%their 
the
attention model in \cite{Bahdanau.2015} significantly outperformed conventional encoder--decoder architectures, which encoded the entire source sequence into a single fixed-length vector.
In particular, this proposed approach turned out to perform better in translating long sentences, where it could achieve performance on par with phrase-based statistical machine translation approaches of that time. 

\subsubsection{Transformer architecture}
\label{sc:Transformer}
Despite improvements, the attention model
%
% CMES style rewriting 
%of 
in
\cite{Bahdanau.2015} shared the fundamental drawback intrinsic to all RNN-based models: The sequential nature of RNNs is adverse to parallel computing making training less efficient as with, e.g., feed-forward or convolutional neural networks, which lend themselves to massive parallelization.
To overcome this drawback, 
%
% CMES style rewriting
%\cite{Vaswani.2017:rd0001} proposed 
a novel model architecture, which entirely dispenses with recurrence, was proposed in \cite{Vaswani.2017:rd0001}. 
As the title \emph{``Attention Is All You Need''} already reveals, their approach to neural machine translation is exclusively based on the concept of attention (and some feedforward-layers), which is repeatedly used in the proposed architecture referred to as \emph{``Transformer''}.

\begin{figure}[h]
\centering
\includegraphics[]{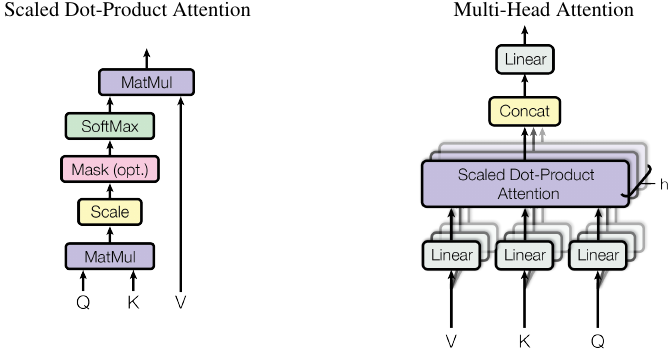}
\caption{
	\emph{Scaled dot-product attention and multi-head attention} (Section~\ref{sc:Transformer}). 
	Scaled-dot product attention (left) is the elementary building block of the Transformer model. 
	It compares query vectors (Q) against a set of key vectors (K) to produce a context vector by weighting to value vectors (V) that correspond to the keys. 
	For this purpose, $\softmax(\cdot)$ function is applied to the inner product (MatMul) of the query and key vectors (scaled by a constant).
	The output of the $\softmax(\cdot)$ represent the weighting by which value vectors are scaled taking the inner product (MatMul), see Eq.\eqref{eq:scaled-dot-product-attention}.
	To prevent attention functions of the decoder, which generates the output sequence item by item, from using future items of the output sequence, a masking layer is introduced. By the masking, all scores beyond the current time/position are set to $- \infty$. 
%	{\color{red} [NOTE: 2022.07.20 - Exactly 3 years later; see NOTE date below.  Please check what I just wrote in the caption.  What is the difference between the left figure (attention?) and the right figure (multi-head attention?). ENDNOTE]}
%	{\color{red} [NOTE: 2019.07.20. we need to give equation number in figure caption, such as the one added above, so to quickly identify which equation the figure corresponds to.]}
\newline
	Multi-head attention (right) combines several ($h$) scaled-dot product attention functions in parallel, each of which is referred to as ``head''.
	For this purpose, queries, keys and values are projected by means of a head-specific linear layers (Linear), whose outputs are input to the individual scaled dot-product attention functions, see Eq.\eqref{eq:projection_multi_head_attention}.
	The context vectors produced by each head are concatenated before being fed into one more linear layer, see Eq.\eqref{eq:multi_head_attention}.
	{
		\footnotesize (Figure reproduced with permission of the authors.)
	}
}
\label{fig:Vaswani-Attention}
\end{figure}

\begin{figure}[h]
	\centering
	\includegraphics[scale=0.875]{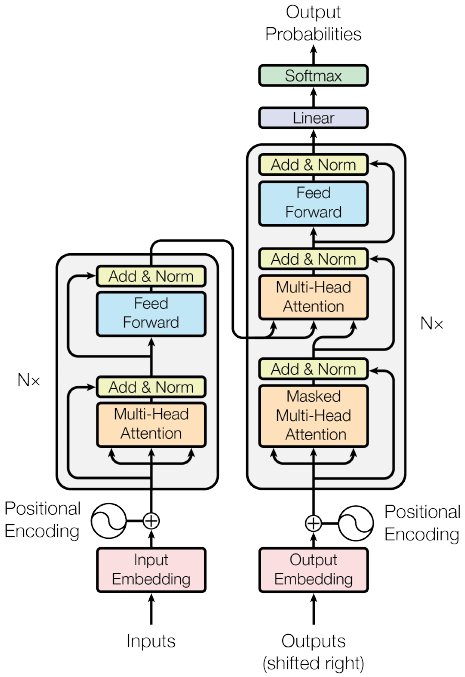}
	\caption{
		\emph{Transformer architecture} (Section~\ref{sc:Transformer}).
		The Transformer is a sequence-to-sequence model without recurrent connections. Encoder and decoder are entirely built upon scaled dot-product attention.
		Items of source and target sequences are numerically represented as vectors, i.e., embeddings.
		Positional encodings furnish embeddings with information on their positions within the respective sequences.
		The encoder stack comprises $N$ layers, each of which is composed from two sub-layers: The first sub-layer is a multi-head attention function used as a \emph{self-attention}, by which relations among the items of one and the same sequence are learned.
%		As with all sub-layers, a residual connection and normalization (Eq.\eqref{eq:layer_normalization}) complete the sub-layer.
		The second sub-layer is a \emph{position-wise} fully-connected network.
		The decoder stack is also built from $N$ layers consisting of three sub-layers. 
		The first and third sub-layers are identical to the encoder except for masking in the self-attention function.
		The second sub-layer is a multi-head attention function using the encoder's output as keys and values to relate items of the source and target sequences.
%		Once again, the sub-layer is complemented by a residual connection and a normalization layer.
%		The decoder's output is projected onto a vector of the dimension of the ``vocabulary'', over which the $\softmax(\cdot)$ function gives probabilities. 
		{
			\footnotesize (Figure reproduced with permission of the authors.)
		}
	}
	\label{fig:Vaswani-Transformer}
\end{figure}

In what follows, we describe the individual components of the Transformer architecture, see Figure~\ref{fig:Vaswani-Transformer}. 
Among those, the \emph{scaled dot-product attention}, see Figure~\ref{fig:Vaswani-Attention}, is a fundamental building block.
Scaled-dot product attention, which is represented by a function $\attention$, compares a query vector $\query \in \real^{d_k}$ and a set of $m$ key vectors $\key_i \in \real^{d_k}$ to determine the weighting of value vectors $\val_i \in \real^{d_v}$ corresponding to the keys.
As opposed to the additive alignment model used 
%by~
in
\cite{Bahdanau.2015}, see Eq.~\eqref{eq:alignment_model}, scaled-dot product attention combines query and key vectors in a multiplicative way.

Let $\keys = [\key_1 , \ldots, \key_m]^T \in \real^{m \times d_k}$ and $\values = [\val_1 , \ldots, \val_m]^T \in \real^{m \times d_v}$ denote key and value matrices formed from the individual vectors, where query and key vectors share the dimension $d_k$ and value vectors are $d_v$-dimensional. 
The attention model produces a context vector $\context \in \real^{d_v}$ by weighting the values $\val_i$, $i=1,\ldots,m$, according to the multiplicative alignment of the query $\query$ with keys $\key_i$, $i=1,\ldots,m$:
%where $d_k$ is the dimension of the query and key vectors the attention mechanism is 
%
%Scaled dot-product attention:
%\begin{equation}
%	\context
%	= \attention(\query, \keys, \values) 
%	= \frac{1}{\sqrt{d_k}} \sum_{i=1}^m \frac{\exp (\query \cdot \key_i )}{\sum_{j=1}^m \exp \left( \query \cdot \key_j \right)} \, \val_i
%	= \frac{1}{\sqrt{d_k}} \softmax \left( \query^T \keys^T  \right) \values .
%%	\attention(\queries, \keys, \values) = \softmax \left( \frac{\queries \cdot \keys}{\sqrt{d_k}} \right) \values .
%%	\head = \attention(\queries \wq, \keys \wk, \values \wv)
%	\label{eq:scaled-dot-product-attention}
%\end{equation}
\begin{align}
	\context
	= \attention(\query, \keys, \values) 
	= \frac{1}{\sqrt{d_k}} \sum_{i=1}^m \frac{\exp (\query \cdot \key_i )}{\sum_{j=1}^m \exp \left( \query \cdot \key_j \right)} \, \val_i
	= \frac{1}{\sqrt{d_k}} \softmax \left( \query^T \keys^T  \right) \values .
	\label{eq:scaled-dot-product-attention}
\end{align} 
Scaling with the square root of the query/key dimension is supposed to prevent pushing the $\softmax(\cdot)$ function to regions of small gradients for large $d_k$ as scalar products grow with the dimension of queries and keys. 
The above attention model can be simultaneously applied to multiple queries. 
For this purpose, let $\queries = [\query_1 , \ldots, \query_k]^T \in \real^{k \times d_k}$ and $\contexts = [\context_1 , \ldots, \context_k]^T \in \real^{k \times d_v}$ denote matrices of query and context vectors, respectively. 
We can rewrite the attention model using matrix multiplication as follows:
\begin{equation}
	\contexts
	= \attention(\queries, \keys, \values) 
%	= \frac{1}{\sqrt{d_k}} \sum_{i=1}^m \frac{\exp (\query \cdot \key_i )}{\sum_{j=1}^m \exp \left( \query \cdot \key_j \right)} \, \val_i
	= \frac{1}{\sqrt{d_k}} \softmax \left( \queries \keys^T  \right) \values  \in \real^{k \times d_k} .
	\label{eq:scaled_dot_product_attention}
\end{equation} 
Note that $\queries \keys^T$ gives a $k \times m$ matrix, for which the $\softmax(\cdot)$ is computed along the second dimension.

Based on the concept of scaled dot-product attention,
%
% CMES style rewriting 
%\cite{Vaswani.2017:rd0001} proposed 
the idea of using multiple attention functions in parallel rather than just a single one was proposed in \cite{Vaswani.2017:rd0001}, see Figure~\ref{fig:Vaswani-Attention}.
In their concept of \emph{``Multi-Head Attention''}, each ``head'' represents a separate context $\contexts_j$ computed from scaled dot-product attention, Eq.~\eqref{eq:scaled_dot_product_attention}, on queries $\queries_j$, keys $\keys_j$ and values $\values_j$, respectively. 
The inputs to the individual scaled dot-product attention functions $\queries_j$, $\keys_j$ and $\values_j$, $j=1,\ldots,h$, in turn, are head-specific (learned) projections of the queries $\queriesmh \in \real^{k \times d}$, keys $\keysmh \in \real^{m \times d}$ and values $\valuesmh \in \real^{m \times d}$.
Assuming that queries $\query_i$, $\key_i$ and $\val_i$ share the dimension $d$, the projections are represented by matrices $\wq \in \real^{d \times d_k}$, $\wk \in \real^{d \times d_k}$ and $\wv \in \real^{d \times d_v}$:\footnote{
	%
	% CMES style rewriting and correction
	Unlike the use of notation in 
%	As opposed to 
	\cite{Vaswani.2017:rd0001}, 
	we use different symbols for the arguments of the scaled dot-product attention function, Eq.~\eqref{eq:scaled_dot_product_attention}, and those of the multi-head attention, Eq.~\eqref{eq:multi_head_attention}, to emphasize their distinct dimensions.
}
\begin{equation}
	\queries_j = \queriesmh \wq \in \real^{k \times d_k} , \qquad 
	\keys_j = \keysmh \wk \in \real^{m \times d_k} , \qquad
	\values_j =  \valuesmh \wv \in \real^{m \times d_k} .
	\label{eq:projection_multi_head_attention}
\end{equation}
Multi-head attention combines the individual {``heads''} $\contexts_i$ through concatenation (along the second dimension) and subsequent projection by means of $\wo \in \real^{d_v h \times d}$,
%The respective contexts $\contexts_i$, i.e., heads, are concatenated and 
%
%
\begin{equation}
	\heads = \multiheadattention( \queriesmh, \keysmh, \valuesmh ) = [\contexts_1 , \ldots, \contexts_h] \wo \in \real^{k \times d}, \qquad
	\contexts_i = \attention(\queries_i, \keys_i, \values_i ) , 
	\label{eq:multi_head_attention}
\end{equation}
where $h$ denotes the number of heads, i.e., individual scaled dot-product attention functions used in parallel.
Note that the output of multi-head attention $\heads$ has the same dimensions of as the input queries $\queriesmh$, i.e., $\heads, \queriesmh \in \real^{k \times d}$. 

To understand why the projection is essential in the Transformer architecture, we shift our attention (no pun intended) to the encoder-structure illustrated in Figure~\ref{fig:Vaswani-Transformer}. 
The encoder combines a stack of $N$ identical layers, which, in turn, are composed from two sub-layers each. 
To stack the layers without further projection, all inputs and outputs of the encoder layers share the same dimension.
The same holds true for each of the sub-layers, which are also designed to preserve the input dimensions.
The first sub-layer is a multi-head \emph{self-attention} function, in which the input sequence attends to itself. 
The concept of self-attention is based on the idea to relate different items of a single sequence to generate a representation of the input, i.e., \emph{``each position in the encoder can attend to all positions in the previous layer of the encoder.''}

In the context of multi-head attention, self-attention implies that one and the same sequence multiply serves as queries, keys and values, respectively.
Let $\sourceseq{\ell} = [ \encoderinput{\ell}{1}, \ldots, \encoderinput{\ell}{n} ] \in \real^{n \times d}$ denote the input to the self-attention (sub-)layer and $\heads \in \real^{n \times d}$ the corresponding output, where $n$ is the length of the sequence and  $d$ is the dimension of single items (\emph{``positions''}), self-attention can be expressed in terms of the multi-head attention function as
\begin{equation}
	\heads = \selfattention (\sourceseq{\ell} ) = \multiheadattention ( \sourceseq{\ell}, \sourceseq{\ell} , \sourceseq{\ell} ) \in \real^{n \times d} .
	\label{eq:transformer_self_attention}
\end{equation}
%
% CMES style rewriting
The authors of \cite{Vaswani.2017:rd0001} introduced a residual connection around the self-attention (sub-)layer, which, in view of the same dimensions of heads and queries, reduces to a simple addition. 

To prevent values from growing upon summation, the residual connection is followed by \emph{layer normalization} as proposed 
%
% CMES style rewriting
%by~
in
\cite{Ba.2016}, which scales the input to zero mean and unit variance:
\begin{equation}
	\layernorm ( \norminp ) = \frac{1}{\stdev} \left( \norminp - \mu \ones \right),
	\label{eq:layer_normalization}
\end{equation}
where $\mu$ and $\stdev$ denote the mean value and the standard deviation of the input components; 
$\ones$ is a matrix of ones with the same dimension as the input $\norminp$.
The normalized output of the encoder's first sub-layer therefore follows from the sum of inputs to self-attention function $\sourceseq{\ell} $ and its outputs $\heads$ as
\begin{equation}
	\outputmh{\ell} = \layernorm(\sourceseq{\ell} + \heads ) \in \real^{n \times d} . 
\end{equation}

The output of the first sub-layer is the input to the second sub-layer within the encoder stack, i.e.,
%
%The second sub-layer is 
a \emph{``position-wise feed-forward network.''}
{Position-wise} means that a fully connected feedforward network (see Section \ref{sc:feedforward} on ``Static, feedforward networks''), which is subsequently represented by the function $\transformerffn$, is applied to each item (i.e., \emph{``position''}) of the input sequence $\ffninput{\ell}{i} \in \outputmh{\ell} = [ \ffninput{\ell}{1} , \ldots , \ffninput{\ell}{n} ]^T \in \real^{n \times d}$ \emph{``separately and identically.''}
In particular, 
%
% CMES style rewriting
%\cite{Vaswani.2017:rd0001} used 
a network with a single hidden layer and a linear rectifier unit (see Section \ref{sc:relu} on ``Rectified linear function (ReLU)'') as activation function was used in \cite{Vaswani.2017:rd0001} to compute the vectors $\ffnoutput{\ell}{i}$:\footnote{
	Eq.~\eqref{eq:transformer-ffn} is meant to reflect the idea off \emph{position-wise} computations of the second sub-layer, since single vectors $\ffninput{\ell}{i}$ are input to the feedforward network. 
	From the computational point of view, however, all ``positions'' can be processed simultaneously using the same weights and biases as in Eq.~\eqref{eq:transformer-ffn}:
%	\begin{equation}
		$\ffnlayer{\ell}  = \max(\mathbf 0, \outputmh{\ell} \weightffnstransformerat + \biasffntransformera) \weightffnstransformerbt + \biasffntransformerb \in \real^{n \times d} .$
%	\end{equation}
	Note that the addition of bias vectors $\biasffntransformera$, $\biasffntransformerb$ needs to be computed for all positions, i.e., they are added row-wisely to the matrix-valued projections of the layer inputs by means $\weightffnstransformerat$ and $\weightffnstransformerbt$, respectively. 
}
\begin{equation} \label{eq:transformer-ffn}
	\ffnoutput{\ell}{i} = \transformerffn (\ffninput{\ell}{i})
	= \weightffnstransformerb \max(\mathbf 0, \weightffnstransformera \ffninput{\ell}{i} + \biasffntransformera) + \biasffntransformerb \in \real^{d} ,
\end{equation}
%\begin{equation}
%	\ffnlayer{\ell} 
%	= \transformerffn (\outputmh{\ell})
%	= \max(\mathbf 0, \outputmh{\ell} \weightffnstransformera + \biasffntransformera) \weightffnstransformerb + \biasffntransformerb \in \real^{n \times d},
%\end{equation}
%
%{\color{blue}
%	TODO: not correct, see dimension of bias! the addition is broadcasted over the first dimension! do we want to introduce a notation for that?
%}
%
where $\weightffnstransformera \in \real^{h \times d_{ff}}$, $\weightffnstransformerb \in \real^{d_{ff} \times h}$ denote the connection weights and $\biasffntransformera \in \real^{h}$ and $\biasffntransformerb \in \real^{d}$ are bias vectors.
The individual outputs $\ffnoutput{\ell}{i}$, which correspond to the respective items $\encoderinput{\ell}{i}$ of the input sequence $\sourceseq{\ell}$, form the matrix $\ffnlayer{\ell} = [ \ffnoutput{\ell}{1}, \ldots, \ffnoutput{\ell}{n} ]^T \in \real^{n \times d}$.

As for the first sub-layer, 
%
% CMES style rewriting
the authors of 
\cite{Vaswani.2017:rd0001} introduced a residual connection followed by layer normalization around the feedforward network.
The ouput of the encoder's second sublayer, which, at the same time, is the output of the encoder layer, is given by
\begin{equation}
	\encoderout{\ell} =  \layernorm ( \outputmh{\ell} + \ffnlayer{\ell} ) \in \real^{n \times d} .
\end{equation}

Within the transformer architecture (Figure~\ref{fig:Vaswani-Transformer}), the encoder is composed from $N$ encoder layers, which are ``stacked'', i.e.,
the output of the $\ell$-th layer is input to the subsequent layer $\sourceseqlayer{\ell+1} = \encodertransformerlayer{\ell}$.
Let $\encodertransformer{\ell}$ denote the $\ell$-th encoder layer composed from the layer-specific self-attention function $\selfattention^{(\ell)}$, which takes $\sourceseqlayer{\ell} \in \real^{n \times d}$ as input, and the layer-specific feedforward network $\transformerffn^{(\ell)}$, the layer's output $\encodertransformerlayer{\ell}$ is computed as follows:
\begin{equation}
	\encodertransformerlayer{\ell} = \encodertransformer{\ell}(\sourceseqlayer{\ell}) = \layernorm (\transformerffnlayer{\ell} ( \layernorm (\sourceseqlayer{\ell} + \selfattentionl (\sourceseqlayer{\ell}))) + \layernorm (\sourceseqlayer{\ell} + \selfattentionl (\sourceseqlayer{\ell})) ) ,
\end{equation} 
%
%\noindent
%{\color{red} [NOTE: 2022.09.29 - The symbol $\encodertransformer{\ell}$ was defined, but not the symbol $\selfattentionl$, which makes the presentation obscure to read, and which should be defined.  ENDNOTE]}
%
%
%\noindent
or, alternatively, using a step-wise representation,
\begin{align}
	\headslayer{\ell} &= \selfattentionl( \sourceseqlayer{\ell} ) \in \real^{n \times d} , \label{eq:transformer-encoder-selfattention}
	\\
	\outputmhlayer{\ell} &= \layernorm (\sourceseqlayer{\ell} + \headslayer{\ell}) \in \real^{n \times d} ,\\
	\outputffnlayer{\ell} &= \transformerffnlayer{\ell} (\outputmhlayer{\ell}) \in \real^{n \times d} ,\\
	\encodertransformerlayer{\ell} &= \layernorm (\outputmhlayer{\ell} + \outputffnlayer{\ell}) \in \real^{n \times d} .
\end{align}
Note that inputs and outputs of all components of an encoder layer share the same dimensions, which facilitates several layers to be stacked without additional projections in between.

The decoder's structure within the transformer architecture is similar to that of the encoder, see the right part in Figure~\ref{fig:Vaswani-Transformer}.
As the encoder, it is composed from $N$ identical layers, each of which combines three sub-layers (as opposed to two sub-layers in the encoder).
In addition to the self-attention sub-layer (first sub-layer) and the fully-connected position-wise feed-forward network (third sub-layer) 
%{\color{red} [NOTE: Add the phrase ``(a sub-layer)'' here for clarity.]}
, the decoder inserts a 
%{\color{red} [NOTE: I added the article ``a'' here.]} 
second sub-layer that \emph{``performs multi-head attention over the output of the encoder stack''} in between.
%{\color{red} [NOTE: I rewrote for clarity; please check and remove the stricken out text.]}. 
Attending to outputs of the encoder enables the decoder to relate items of the source sequence to items of the target sequence.
Just as in the encoder, residual connections are introduced around each of the decoder's sub-layers.

Let $\decoderinputfirstlayer{\ell} \in \real^{n \times d}$ denote the input to the $\ell$-th decoder layer, which, for $\ell > 1$, is the output $\decodertransformerlayer{\ell-1} \in \real^{n \times d}$ of the previous layer.
The output $\decodertransformerlayer{\ell}$ of the $\ell$-th decoder layer is then obtained through the following computations. 
The first sub-layer performs (multi-headed) self-attention over items of the output sequence, i.e., it establishes a relation among them:
\begin{align}
	\decoderheadsfirstlayer{\ell} &= \selfattentionl( \decoderinputfirstlayer{\ell} ) \in \real^{n \times d} ,  \label{eq:transformer-decoder-selfattention}
	\\
	\outputmhfirstlayer{\ell} &= \layernorm (\targetseqlayer{\ell} + \decoderheadsfirstlayer{\ell}) \in \real^{n \times d} .
\end{align}
The second sub-layer relates items of the source and the target sequences by means of multi-head attention:
\begin{align}
	\decoderinputsecondlayer{\ell} &= \multiheadattention^{(\ell)} (\encodertransformerlayer{\ell}, \encodertransformerlayer{\ell} , \outputmhfirstlayer{\ell})  \in \real^{n \times d} , 
	\label{eq:transformer-decoder-mhattention}
	\\
	\outputmhsecondlayer{\ell} &= \layernorm (\outputmhfirstlayer{\ell} + \decoderinputsecondlayer{\ell}) \in \real^{n \times d} .
\end{align}
The third sub-layer is a fully-connected feed-forward network (with a single hidden layer) that is applied to each element of the sequence (position-wisely)
\begin{align}
	\decoderinputthirdlayer{\ell} &= \transformerffnlayer{\ell} (\outputmhsecondlayer{\ell}) \in \real^{n \times d} ,\\
	\decodertransformerlayer{\ell} &= \layernorm (\outputmhsecondlayer{\ell} + \decoderinputthirdlayer{\ell}) \in \real^{n \times d} .
\end{align}
The output of the last ($\ell = N$) decoder-layer within the decoder-stack is projected onto a vector that has the dimension of the ``vocabulary'', i.e., the set of all feasible output items.
Taking the $\softmax(\cdot)$ over the components of the vector produces probabilities over elements of the vocabulary to be item of the output sequence (see Section~\ref{sc:classification}).
%
%\alex{TODO: explain the different inputs during training (whole squences) and use (token by token)}

A Transformer-model is trained on the complete source and target sequences, which are input to the encoder and the decoder, respectively. 
The target sequence is shifted right by one position, such that a special token indicating the start of a new sequence can be placed at the beginning, see Fig.~\ref{fig:Vaswani-Transformer}.
To prevent the decoder from attending to future items of the output sequence, the (multi-headed) self-attention sub-layer needs to be \emph{``masked''}.
%For this purpose, the (multi-headed) self-attention sub-layer needs to be \emph{``masked''} in order \emph{``to preserve the auto-regressive property''}, i.e., 
% here
%The self-attention sub-layers of the encoder are modified \emph{``to prevent leftward information flow in the decoder to preserve the auto-regressive property.''}
The masking is realized by setting those inputs to the $\softmax(\cdot)$ function of the scaled dot-product attention, see Eq.~\eqref{eq:scaled-dot-product-attention}, for which query vectors are aligned with keys that correspond to items in the output sequence beyond the respective query's position, set to $- \infty$. 
As loss function, the negative log-likelihood is typically used; see Section~\ref{sc:maximum-likelihood} on maximum likelihood (probability cost) and Section~\ref{sc:classification} on classification loss functions.

As the transformer architecture does not have recurrent connections,
%
% CMES style rewriting 
%\cite{Vaswani.2017:rd0001} added 
\emph{positional encodings} were added to the inputs of both encoder and decoder \cite{Vaswani.2017:rd0001}, see Figure~\ref{fig:Vaswani-Transformer}.
The positional encodings supply the individual items of the inputs to the encoder and the decoder with information about their positions within the respective sequences.
%
% CMES style rewriting
For this purpose, the authors of  
\cite{Vaswani.2017:rd0001} proposed to add vector-valued positional encodings $\posenc_i \in \reals^{d}$, which have the same dimension $d$ as the input embedding of an item, i.e., its numerical representation as a vector.
They used sine and cosine functions of an items position ($i$) within respective sequence, whose frequency varies (decreases) with the component index ($j$):
\begin{equation}
	p_{i, 2j} = \sin \theta_{ij} , \qquad
	p_{i, 2j-1} = \cos \theta_{ij}  , \qquad
	\theta_{ij} = \frac{j}{10000^{2i/d}} , \qquad 
	j = 1, \ldots, d .
	\label{eq:transformer_positional_encoding}
\end{equation}

The trained Transformer-model produces one item of the output sequence at a time.
Given the input sequence and all outputs already generated in previous steps, the Transformer predicts probabilities for the next item in the output sequence.
Models of this kind are referred to as ``auto-regressive''.

The authors of~\cite{Vaswani.2017:rd0001} varied parameters of the Transformer model to study the importance of individual components. 
Their ``base model'' used $N=6$ encoder and decoder layers.
Inputs and outputs of all sub-layers are sequences of vectors, which have a dimension of $d=512$.
All multi-head attention functions of the Transformer, see Eq.\eqref{eq:transformer-encoder-selfattention}, Eq.\eqref{eq:transformer-decoder-selfattention} and Eq.\eqref{eq:transformer-decoder-mhattention}, have $h = \num{8}$ heads each.
Before the alignment scores are computed by each head, the dimensions of queries, keys and values were reduced to $d_k = d_v = \num{64}$.
The hidden layer of the fully-connected feedforward network, see Eq.\eqref{eq:transformer-ffn}, was chosen as $d_{ff} = \num{2048}$ neurons wide. 
With this set of (hyper-)parameters, the Transformer model features a total of \num{65e6} parameters.
The training cost (in terms of FLOPS) of the attention-based Transformer was shown to be (at least) two orders of magnitudes smaller than comparable models at that time while the performance was maintained.
A refined (``big'') model was able to outperform all previous approaches (based on RNNs or convolutional neural networks) in English-to-French and English-to-German translation tasks.

As of 2022, the \emph{Generative Pre-trained Transformer 3} (GPT-3) model~\cite{GPT-3}, which is based on the Transformer architecture, belongs to the most powerful language models. 
GPT-3 is an autoregressive model that produces text from a given (initial) text prompt, whereby it can deal with different tasks as translation, question-answering, cloze-tests\footnote{
	A cloze is ``a form of written examination in which candidates are required to provide words that have been omitted from sentences, thereby demonstrating their knowledge and comprehension of the text'', see \href{https://en.wiktionary.org/w/index.php?title=cloze&oldid=65140547}{Wiktionary version 11:23, 2 January 2022}.
} and word-unscrambling, for instance.
The impressive capabilities of GPT-3 are enabled by its huge capacity of \num{175} \emph{billion} parameters, which is 10 times more than preceding language models.

\begin{rem}
	\label{rm:attention-kernel-PINN}
	Attention mechanism, kernel machines, physics-informed neural networks (PINNs).
	{\rm
		In \cite{tsai2019transformer}, a new attention architecture (mechanism) was proposed by using kernel machines discussed in Section~\ref{sc:kernel-machines}, whereas in \cite{rodriguez2022physics}, the gated recurrent units (GRU, Section~\ref{sc:GRU}) and the attention mechanism (Section~\ref{sc:sequence-encoder-decoder-attention}) were used in conjunction with Physics-Informed Neural Networks (PINNs, Section~\ref{sc:PINN-frameworks}) to solve hyperbolic problems with shock waves; Remark~\ref{rm:PINN-attention} and Remark~\ref{rm:PINN-solid-mechanics}.
	}
	$\hfill\blacksquare$
\end{rem}

\section{Kernel machines (methods, learning)}
\label{sc:kernel-machines}

%{\color{red} [NOTE: 2022.09.14 - I made this part into a section (instead of a subsubsection) to prepare to move it to before the section on deep-learning libraries.  In addition, I can then have the latitude to have subsections in this section.  ENDNOTE]}

Researchers have observed that as the number of parameters increased beyond the interpolation threshold, or as the number of hidden units in a layer (i.e., layer width) increased, the test error decreased, i.e., such network generalized well; see Figures~\ref{fig:double-descent-risk} and \ref{fig:test-error-large-params}.   So, as a first step to try to understand why deep-learning networks work (Section~\ref{sc:lack-understanding} on ``Lack of understanding''), it is natural to study the limniting case of infinite layer width first, since it would be relatively easier than the case of finite layer width
\cite{bahri2019towards}; see Figure~\ref{fig:network-infinite-width}. 

In doing so, a connection between networks with infinite width and  the kernel machines or kernel methods, was revealed \cite{anan2021anew} \cite{lee2018deep} \cite{jacot2018neural}.  See also the connection between kernel methods and Support Vector Machines (SVM) in Footnote~\ref{fn:support-vector-machine}.

``A neural network is a little bit like a Rube Goldberg machine. You don't know which part of it is really important.  ... reducing [them] to kernel methods---because kernel methods don't have all this complexity---somehow allows us to isolate the engine of what's going on'' \cite{anan2021anew}.

Quanta Magazine described the discovery of such connection as the
2021 breakthrough in computer science \cite{quanta2021breakthroughs}.

%At least on a conceptual level, the connection between networks with infinite width and the kernel methods makes sense based on the relation between the linear combination of weights and states and the kernel formulation in the continuous temporal summation in neuroscience as explained in Section~\ref{sc:dynamic-volterra-series}.\footnote{
	%	We wrote Section~\ref{sc:dynamic-volterra-series} roughly two years before we learned about this connection in Aug 2022.
	%}

Covariance functions, or covariance matrices, are kernels in Gaussian processes (Section~\ref{sc:Gaussian-process}), an important class of methods in machine learning \cite{rasmussen2006gaussian} \cite{bishop2006pattern}.
A kernel method in terms of the time variable was discussed in Section~\ref{sc:dynamic-volterra-series} in connection with the continuous temperal summation in neuroscience.  Our aim here is only to provide first-time learners background material on kernel methods in terms of space variables (specifically the ``Setup'' in \cite{belkin2018understand}) in preparation to read more advanced references mentioned in this section, such as \cite{belkin2018understand} \cite{jacot2018neural} \cite{lee2020finite} etc. 

%{\color{red} [NOTE: 2022.09.10 - additional refs for kernel methods \cite{hastie2017elements}, \cite{berlinet2004reproducing}; to expand the discussion.  ENDNOTE]}

\subsection{Reproducing kernel: General theory}
\label{sc:reproducing-kernel-defn}
A kernel $\K(\cdot,y)$ is called \emph{reproducing} if its scalar ($L_2$) product with a function $f(y)$ reproduces the same function $f(\cdot)$ itself \cite{aronszajn1950theory}:
\begin{align}
	f(x) 
	= \langle f (y) , \K (x , y) \ , \rangle_y
	= \int f(y) \K (x,y) dy 	
	\label{eq:reproducing-kernel}
\end{align}
where the subscript $y$ in the scalar product $\langle \cdot , \cdot  \rangle_y$ indicates the integrand.

%\subsection{Kernels constructed from basis functions}

%basis functions
%$\bfun , \bfuns{i} \bbfuns{j}$
%
%Gram matrix
%$\Gram_{ij} , \bGram$
%
%Inverse of Gram matrix
%$\iGram_{lm} , \biGram$

Let $\{ \bfuns{1} (x) , \ldots , \bfuns{n} (x) \}$ be a set of linearly independent basis functions.  A function $f(x)$ can be expressed in this basis as follows:
\begin{align}
	f(x) = \sum_{k=1}^{n} \zeta_k \bfuns k (x) \ .
	\label{eq:function-in-basis}
\end{align}
The scalar product of two functions $f$ and $g$ is given by:
\begin{align}
	g(x) 
	= \sum_{k=1}^{n} \eta_k \bfuns k (x) 
	\Rightarrow
	\langle f, g \rangle 
	= \sum_{i,j=1}^{n} \zeta_i \eta_i 
	\langle \bfuns i , \bfuns j \rangle 
	= \sum_{i,j=1}^{n} \zeta_i \eta_i \Grams i j \ ,
	\label{eq:scalar-product-of-functions}
\end{align}
where
\begin{align}
	\Grams i j =
	\langle \bfuns i , \bfuns j \rangle \ , \quad
	\bGram = [ \Grams i j ] > 0 \in \real^{n \times n}
	\label{eq:Gram-matrix}
\end{align}
is the Gram matrix,\footnote{
	The stiffness matrix in the displacement finite element method is a Gram matrix.
} which is strictly positive definite, with its inverse (also strictly positive definite) denoted by 
\begin{align}
	\biGram = \bGram^{-1} = [\iGrams i j] > 0 \in \real^{n \times n} \ .
	\label{eq:inverse-Gram}
\end{align} 

Then a reproducing kernel can be written as \cite{aronszajn1950theory}\footnote{
	See Eq.~(6), p.~346, in \cite{aronszajn1950theory}.
}
\begin{align}
	\K(x,y) 
	= \sum_{i,j=1}^{n} \iGrams{i}{j} \bfuns i (x) \bfuns j (y)
	= \bbfunV^T (x) \biGram \bbfunV (y) \ , \text{ with }
	\bbfunV^T (x) = [ \bfuns 1 (x) , \ldots , \bfuns n (x)] \in \real^{1 \times n} \ .
	\label{eq:kernel-basis-functions}
\end{align}

\begin{rem}
	\label{rm:reproducing-kernel}
	{\rm It is easy to verify that the function $\K (x, y)$ in Eq.~\eqref{eq:kernel-basis-functions} is a reproducing kernel:
	\begin{align}
		\langle f(y) , \K(y,x) \rangle_y 
		& 
		= \langle \sum_{i} \zeta_i \bfuns{i} (y) , \sum_{j,k} \iGrams{j}{k}  \bfuns{j} (y) \bfuns{k} (x) \rangle_y
		= \sum_{i} \zeta_i \sum_{j,k} \langle \bfuns{i} (y) , \bfuns{j} (y) \rangle_y \iGrams{j}{k} \bfuns{k} (x)
		\\
		&
		= \sum_{i} \zeta_i \sum_{j,k} \Grams{i}{j} \iGrams{j}{k} \bfuns{k} (x)
		= \sum_{i} \zeta_i \sum_{k} \delta_{ik} \bfuns{k} (x)
		= \sum_{i} \zeta_i \bfuns{i} (x) = f(x) \ ,
		\label{eq:reproducing-kernel-1}
	\end{align}
	using Eqs.~\eqref{eq:Gram-matrix} and \eqref{eq:inverse-Gram}.}
	$\hfill\blacksquare$
\end{rem}

From Eq.~\eqref{eq:scalar-product-of-functions}, the norm of a function $f$ can be defined as
\begin{align}
	\parallel f \parallel_\K^2 = \langle f , f \rangle 
	= \sum_{i,j=1}^{n} \zeta_i \Gram_{ij} \zeta_j 
	= \boldsymbol{\zeta}^T \bGram \boldsymbol{\zeta}
	= \boldsymbol{\zeta}^T \biGram^{-1} \boldsymbol{\zeta}
	\ , \text{ with }
	\boldsymbol{\zeta}^T = [\zeta_1 , \ldots , \zeta_n ] \ ,
	\label{eq:kernel-norm-finite} 
\end{align}
where $\parallel \cdot \parallel_\K$ denotes the norm induced by the kernel $\K$, and where the matrix notation in Eq.~\eqref{eq:matrices_x_y} was used.

When the basis functions in $\bbfunV$ are mutually orthogonal, then the Gram matrix $\bGram$ in Eq.~\eqref{eq:Gram-matrix} is diagonal, and the reproducing kernel $\K$ in Eq.~\eqref{eq:kernel-basis-functions} takes the simple form:
\begin{align}
	\K (x,y) = \sum_{k=1}^{n} \iGram_k \bfuns{k} (x) \bfuns{k} (y) 
	\ , \text{ with }
	\iGram_k = \iGrams{i}{k} \delta_{ik} = \iGram_{(kk)} > 0 \ ,
	\text{ for } k = 1, \ldots, \infty \ ,
	\label{eq:kernel-orthogonal-basis}
\end{align}
where $\delta_{ij}$ is the Kronecker delta, and
where the summation convention on repeated indices, except when enclosed in parentheses, was applied.  In the case of infinite-dimensional space of functions, Eq.~\eqref{eq:kernel-orthogonal-basis}, Eq.~\eqref{eq:function-in-basis} and Eq.~\eqref{eq:scalar-product-of-functions} would be written with $n \rightarrow \infty$:\footnote{
	Eq.~\eqref{eq:infinite-dim-function-space-1}$_{1}$, and
	Eqs.~\eqref{eq:infinite-dim-function-space-2}$_{1,2}$ were given as Eq.~(5.46), Eq.~(5.47), and Eq.~(5.45), respectively, in \cite{hastie2017elements}, pp.~168-169, where the basis functions $\bfuns{i} (x)$ were the ``eigen-functions,'' and where the general case non-orthogonal basis presented in Eq.~\eqref{eq:kernel-basis-functions} was not discussed.  Technically, Eq.~\eqref{eq:infinite-dim-function-space-2}$_2$ is accompanied by the conditions $\iGram_i \ge 0$, for $i = 1 , \ldots , \infty$, and $\sum_{k=1}^{\infty} (\iGram_k)^2 < \infty$, i.e., the sum of the squared coefficients is finite \cite{hastie2017elements}, p.~188.
}   
\begin{align}
	&
	f(x) = \sum_{k=1}^{\infty} \zeta_k \bfuns k (x) 
	\ , \quad
	\langle f, g \rangle 
	= \sum_{k=1}^{\infty} \zeta_k \eta_k \Gram_k
	= \sum_{k=1}^{\infty} \zeta_k \eta_k / \iGram_k
	= {\bkar \zeta}^T \bGram {\bkar \eta}
%	\ , \quad
	\label{eq:infinite-dim-function-space-1}
	\\
	&
	\parallel f \parallel_\K^2 
	= \sum_{k=1}^{\infty} (\zeta_k)^2  \Gram_k
	= \sum_{k=1}^{\infty} (\zeta_k)^2 / \iGram_k
	= {\bkar \zeta}^T \bGram {\bkar \zeta}
	\ , \quad
	\K (x,y) = \sum_{k=1}^{\infty} \iGram_k \bfuns{k} (x) \bfuns{k} (y)
	\ .
	\label{eq:infinite-dim-function-space-2}
\end{align}

Let $L(y, f(x))$ be the loss (cost) function, with $x$ being the data, $f(x)$ the predicted output, and $y$ the label.  Consider the regularized minimization problem:
\begin{align}
%	&
	\min_{f} \left[ \sum_{i=1}^{n} L(y_i , f(x_i) ) + \lambda \parallel f \parallel_\K^2 \right]
	\label{eq:regularized-minimization-1}
%	\\
%	&
	=
%	\min_{\zeta_i, i = 1, \ldots , \infty} 
	\min_{ \{\zeta_i \}_1^\infty }
	\left[ \sum_{i=1}^{n} L(y_i , \sum_{j=1}^{\infty} \zeta_j \bfuns{j} (x_i) ) 
	+ \lambda \sum_{j=1}^\infty (\zeta_j)^2 / \iGram_j 
	\right] \ ,
%	\label{eq:regularized-minimization-2} 
\end{align}  
which is a ``ridge'' penalty method\footnote{
	See \cite{hastie2017elements}, Eq.~(3.41), p.~61, which is in Section 3.4.1 on ``Ridge regression,'' which ``shrinks the regression coefficients by imposing a penalty on them.''  Here the penalty is imposed on the kernel-induced norm of $f$, i.e., $\parallel f \parallel_\K^2$.
} with $\lambda$ being the penalty coefficient (or regularization parameter), with the aim of forcing the norm of the minimizer, i.e, $\parallel f^\star \parallel^2$, to be as small as possible, by penalizing the objective function (cost $L$ plus penalty $\lambda \parallel f \parallel_\K^2$) if it were not.\footnote{
	In classical regularization, the loss function (1st term) in Eq.~\eqref{eq:regularized-minimization-1} is called the ``empirical risk'' and the penalty term (2nd term) the ``stabilizer'' \cite{evgeniou2000regularization}.
}
What is remarkable is that even though the minimization problem is \emph{infinite} dimensional, the minimizer (solution) $f^\star$ of Eq.~\eqref{eq:regularized-minimization-1} is \emph{finite} dimensional \cite{hastie2017elements}:
\begin{align}
	f^\star (x) =  \sum_{i=1}^{n} \alpha_i^\star \K (x, x_i)  \ ,
	\label{eq:minimizer-function}
\end{align}
where $\K( \cdot , x_i)$ is a basis function, and $\alpha_i^\star$ the corresponding coefficient, for $i = 1, \ldots , n$.

\begin{rem}
	Finite-dimensional solution to infinite-dimentional problem.
	{\rm
		Since 
		$f(x)$ is expressed as in Eq.~\eqref{eq:infinite-dim-function-space-1}$_1$, and
		$\K (x,y)$ as in Eq.~\eqref{eq:infinite-dim-function-space-2}$_2$, it follows that
		\begin{align}
			\langle \K(y, x_i) , f (y) \rangle_y = f(x_i)
			\ ,
		\end{align}
		following the same argument as in Remark~\ref{rm:reproducing-kernel}.  As a result, 
		\begin{align}
			\langle \K (y , x_i) , \K(y, x_j) \rangle_y = \K(x_i , x_j) \ , \text{ with } i,j = 1, \ldots , n
			\ .
		\end{align}
		Thus $\bkar \K = \left[ \K (x_i , x_j) \right] \in \real^{n \times n}$ is the Gram matrix of the set of functions $\{ \K (\cdot , x_1) , \ldots , \K (\cdot , x_n)\}$.  To show that $\bkar \K > 0$, i.e., positive definite, for any set $\{ a_1 , \ldots , a_n \}$, consider
		\begin{align}
			&
			\sum_{i,j=1}^{n} a_i \K(x_i, x_j) a_j
			= \sum_{i,j=1}^{n} a_i \sum_{p=1}^\infty \iGram_p \bfuns{p} (x_i) \bfuns{p} (x_j) a_j 
			=  \sum_{p=1}^\infty \iGram_p (b_p)^2 \ge 0 \ ,
			\\
			&
			b_p := \sum_{i=1}^n a_i \bfuns{p} (x_i) \ , \text{ for } p = 1 , \ldots , \infty \ ,
		\end{align}
		which is equivalent to the matrix $\bkar \K$ being positive definite, i.e., $\bkar \K > 0$,\footnote{
			See, e.g., \cite{berlinet2004reproducing}, p.~11.
		} and thus the functions $\{ \K (\cdot , x_1) , $ $ \ldots ,$ $ \K (\cdot , x_n)\}$ are linearly independent, and form a basis, making expression such as Eq.~\eqref{eq:minimizer-function} possible.
		
%		Consider the following orthogonal decomposition of a function $f(x)$:
%		\begin{align}
%			f (x) = g(x) + h(x) 
%			\ , \text{ with }
%			g (x) = \sum_{i=1}^{n} \alpha_i \K(x, x_i) 
%			\text{ and }
%			\langle g(x) , h(x)  \rangle_x = 0 \ , \text{ i.e., }
%			g \perp h \ .
%			\label{eq:perturned-function}
%		\end{align}
		A goal now is to show that the solution to the \emph{infinite}-dimensional regularized minimization problem Eq.~\eqref{eq:regularized-minimization-1} is 
%		of the form $g(x)$ in Eq.~\eqref{eq:perturned-function}$_2$, i.e., 
		\emph{finite} dimensional, for which the coefficients $\alpha_i^\star$, $i = 1 , \ldots , n$, in Eq.~\eqref{eq:minimizer-function} are to be determined.  It is also remarkable that the solution of the form Eq.~\eqref{eq:minimizer-function} holds in general for \emph{any} type of differentiable loss function $L$ in Eq.~\eqref{eq:regularized-minimization-1}, and not necessarily restricted to the squared-error loss \cite{girosi1998equivalence} \cite{evgeniou2000regularization}.
		
		For notation compactness, let the objective function (loss plus penalty) in Eq.~\eqref{eq:regularized-minimization-1} be written as
		\begin{align}
			\krb{L} [f] := \sum_{k=1}^{n} L( y_k , f(x_k) ) + \lambda \parallel f \parallel_\K^2 
			= \sum_{k=1}^{n} L(y_k , \sum_{j=1}^{\infty} \zeta_j \bfuns{j} (x_k) ) 
			+ \lambda \sum_{k=1}^\infty (\zeta_k)^2 / \iGram_j
			\ ,
			\label{eq:regularized-objective-function}
		\end{align}
		and set the derivative of $\krb{L}$ with respect to the coefficients $\zeta_p$, for $p = 1, \ldots, \infty$, to zero to solve for these coefficients:
		\begin{align}
			&
			\frac{\partial \krb{L}}{\partial \zeta_p}
			= - \sum_{k=1}^{n} \frac{\partial {L} ( y_k , f(x_k) )}{\partial f} \frac{\partial f (x_k)}{\partial \zeta_p} + 2 \lambda \frac{\partial \parallel f \parallel_\K}{\partial \zeta_p}
			=  - \sum_{k=1}^{n} \frac{\partial {L} ( y_k , f(x_k) )}{\partial f} \bfuns{p} (x_k) + 2 \lambda \frac{\zeta_p}{\iGram_p} = 0 
			\\
			&
			\Rightarrow
			\zeta_p^\star = \frac{\iGram_p}{2 \lambda} \sum_{k=1}^{n} \frac{\partial {L} ( y_k , f(x_k) )}{\partial f} \bfuns{p} (x_k) 
			=  \iGram_p \sum_{k=1}^{n} \alpha^\star_k \bfuns{p} (x_k)
			\ , \text{ with }
			\alpha^\star_k :=  \frac{1}{2 \lambda} \frac{\partial {L} ( y_k , f(x_k) )}{\partial f} \ ,
			\label{eq:minimizer-coeff-alpha-star}
			\\
			&
			\Rightarrow
			f^\star (x) = \sum_{k=1}^{\infty} \zeta_k^\star \bfuns{k} (x) 
			= \sum_{k=1}^{\infty} \iGram_k \sum_{i=1}^{n} \alpha^\star_i \bfuns{k} (x_i) \bfuns{k} (x)
			= \sum_{i=1}^{n} \alpha^\star_i \sum_{k=1}^{\infty} \iGram_k  \bfuns{k} (x_i) \bfuns{k} (x) 
			= \sum_{i=1}^{n} \alpha^\star_i \K(x_i, x) \ ,
			\label{eq:minimizer-function-2}
		\end{align}
		where the last expression in Eq.~\eqref{eq:minimizer-function-2} came from using the kernel expression in Eq.~\eqref{eq:infinite-dim-function-space-2}$_2$, and the end result is Eq.~\eqref{eq:minimizer-function}, i.e., the solution (minimizer) $f^\star$ is of finite dimension.\footnote{
			See also \cite{hastie2017elements}, p.~169, Eq.~5.50, and p.~185.
		} 	
	}
	$\hfill\blacksquare$
\end{rem}

For the squared-error loss, 
\begin{align}
	&
	L(\bkar y , f(\bkar x)) = \parallel \bkar y - f(\bkar x) \parallel_\K^2 = \sum_{k=1}^{n} \left[ y_k - f(x_k) \right]^2 = \left[ \bkar y - \bkar \K \bkar \alpha^\star \right]^T \left[ \bkar y - \bkar \K \bkar \alpha^\star \right] \ , \text{ with }
	\\
	&
	\bkar y^T = \left[ y_1 , \ldots , y_n \right] \in \real^{1 \times n} 
	\ , \quad
	\bkar \K = \left[ \Ks{ij} \right] = \left[ \K (x_i , x_j )\right] \in \real^{n \times n}
	\ , \quad  
	\bkar \alpha^{\star T} = \left[ \alpha^\star_1 , \ldots , \alpha^\star_n \right] \in \real^{n \times n} \ ,
\end{align}
the coefficients $\alpha^\star_k$ in Eq.~\eqref{eq:minimizer-function-2} (or Eq.~\eqref{eq:minimizer-function}) can be computed using Eq.~\eqref{eq:minimizer-coeff-alpha-star}$_2$:
\begin{align}
	&
	\alpha^\star_k = \frac{1}{\lambda} \left[ y_k - f^\star (x_k) \right]
	\Rightarrow
	\lambda \alpha^\star_k = y_k - \sum_{j=1}^{n} \Ks{kj} \alpha^\star_j
	\Rightarrow
	\left[ \bkar \K + \lambda \bkar I \right] \bkar \alpha^\star = \bkar y
%	\\
%	& 
	\Rightarrow
	\bkar \alpha^\star = \left[ \bkar \K + \lambda \bkar I \right]^{-1} \bkar y \ .
	\label{eq:minimizer-coeff-alpha-star-2}
\end{align}
It is then clear from the above that the ``Setup'' section in \cite{belkin2018understand} simply corresponded to the particular case where the penalty parameter was zero:
\begin{align}
	\lambda = 0 \Rightarrow y_k = f^\star (x_k) = \sum_{j=1}^{n} \Ks{kj} \alpha_j^\star
	\ ,
\end{align}
i.e., $f^\star$ is interpolating.

For technical jargon such as Reproducing Kernel Hilbert Space (RKHS), Riesz Representation Theorem, $\K (\cdot, x_i)$ as a representer of evaluation at $x_i$, etc. to describe several concepts presented above, see \cite{aronszajn1950theory} \cite{wahba1990spline} \cite{berlinet2004reproducing} \cite{hastie2017elements}.\footnote{
	A succinct introduction to Hilbert space and the Riesz Representation theorem, with detailed proofs, starting from the basic definitions, can be found in \cite{adler2021hilbert}.
}

%		\begin{align}
%			\parallel f^\star \parallel_\K^2
%			&
%			= \langle f^\star , f^\star \rangle
%			= \langle \sum_{i=1}^{\infty} \alpha_i^\star \K( y , x_i)  ,  \sum_{j=1}^{\infty} \alpha_j^\star \K(y , x_j) \rangle_y
%			= \sum_{i,j=1}^{\infty} \alpha_i^\star \alpha_j^\star \K(x_i , x_j) 
%			\\
%			&
%			= \bkar {\alpha^\star}^T \bkar \K \bkar {\alpha^\star}
%			\ , \text{ with }
%			\bkar \K = \left[ \K (x_i , x_j) \right]
%			\ .
%		\end{align}

%\cite{bertsekas2000enhanced}

%\cite{wahba1990spline}

\begin{table}[h]
	\centering
	\caption{
		\emph{Some reproducing kernels} (Section~\ref{sc:kernel-machines}). See  \cite{evgeniou2000regularization}, \cite{bishop2006pattern}, p.~296, p.~305, \cite{schaback2006kernel}. 
	}
	\begin{tabularx}{0.7\textwidth}{l *{2}{Y}}
		\toprule[3pt]
		Regularization network		
		& \multicolumn{1}{l}{Kernel function}  
		\\
		\cmidrule(lr){1-1} \cmidrule(lr){2-2} 
		Gaussian Radial Basis Function 		
		& \multicolumn{1}{l}{$\exp \left[ - \parallel \bkar x - \bkar y \parallel^2 \right]$}   
		\\
		Exponential (Laplacian) 		
		& \multicolumn{1}{l}{$\exp \left[ - \parallel \bkar x - \bkar y \parallel \right]$}   
		\\
		Inverse multiquadric	
		& \multicolumn{1}{l}{$\left[ \parallel \bkar x - \bkar y \parallel^2 + c^2 \right]^{-1/2}$}  
		\\
		Multiquadric	
		& \multicolumn{1}{l}{$\left[ \parallel \bkar x - \bkar y \parallel^2 + c^2 \right]^{1/2}$}  
		\\
		Thin plate spline (a)	
		& \multicolumn{1}{l}{$ \parallel \bkar x - \bkar y \parallel^{2n + 1} $}  
		\\
		Thin plate spline (b)	
		& \multicolumn{1}{l}{$ \parallel \bkar x - \bkar y \parallel^{2n} \log \parallel \bkar x - \bkar y \parallel $}  
		\\
		Multilayer perceptron (for some values of $\theta$)	
		& \multicolumn{1}{l}{$\tanh \left( \bkar x \cdot \bkar y - \theta \right)$}  
		\\
		Polynomial of degree $d$	
		& \multicolumn{1}{l}{$\left[ 1 + \bkar x \cdot \bkar y \right]^d $}  
		\\
		\midrule[2pt]
		%		\\
	\end{tabularx}
	\label{tb:reproducing-kernels}
\end{table}

\subsection{Exponential functions as reproducing kernels}
\label{sc:kernel-exponential-functions}
%{\color{red} [NOTE: 2022.09.17 - Notation for standard deviation using the macro \verb*|\stdev| yielding the symbol $\stdev$.  We need to replace the symbol \verb*|\sigma| for the logistic sigmoid by the macro \verb*|\sigmoid|, yielding the symbol $\sigmoid$, already defined in both files \verb*|macros_alex.tex| and \verb*|macrosAlex.tex| by a search and replace.   After that, we can discuss to decide which symbol for which quantity to use by changing the definition of the macros.   Also the macro \verb*|\stress| yields the symbol $\stress$.  ENDNOTE]}

\begin{figure}[h]
	\centering
	\includegraphics[width=0.48\linewidth]{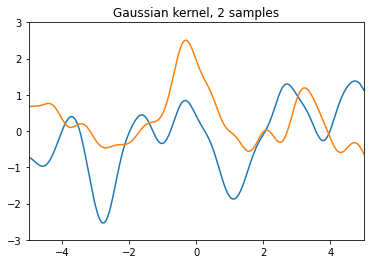}
	\includegraphics[width=0.48\linewidth]{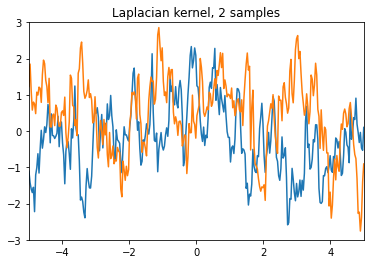}
	\caption{
		\emph{Gaussian process priors} (Section~\ref{sc:Gaussian-process}).
		\emph{Left:} Two samples with Gaussian kernel. 
		\emph{Right:} Two samples with Laplacian kernel.
		Parameters for both kernels: 
		Kernel precision (inverse of variance) $\gamma = \sigma^{-2} = 0.2$ in Eq.~\eqref{eq:exponential-kernel-1},
		isotropic noise variance $\nu^2 \bkar{I} = 10^{-6} \bkar{I}$ added to covariance matrix $\bkars{C}{\by \by}$ of output $\by$ and isotropic weight covariance matrix $\bkars{C}{\bkar{w} \bkar{w}} = \bkar{\Sigma} = \bkar{I}$ in Eq.~\eqref{eq:Gram-matrix-perturned}.
	}
	\label{fig:Gaussian-process-prior}
\end{figure}

A list of reproducing kernels is given in, e.g., \cite{girosi1998equivalence} \cite{evgeniou2000regularization}, such as those in Table~\ref{tb:reproducing-kernels}.
Two reproducing kernels with exponential function were used to understand how deep learning works \cite{belkin2018understand}, and are listed in Eq.~\eqref{eq:exponential-kernel-1}: (1) the popular smooth Gaussian kernel $\K_G$ in in Eq.~\eqref{eq:exponential-kernel-1}$_1$, and (2) the non-smooth Laplacian (exponential) kernel\footnote{
	\label{fn:Laplacian-kernel-Brownian-motion}
	In \cite{bishop2006pattern}, p.~305, the kernel $\K_L$ in Eq.~\eqref{eq:exponential-kernel-1}$_2$ was called ``exponential kernel,'' without a reference to Laplace, but referred to ``the Ornstein-Uhlenbeck process originally introduced by Uhlenbeck and Ornstein (1930) to describe Brownian motion.''  Similarly, in \cite{rasmussen2006gaussian}, p.~85, the term ``exponential covariance function'' (or kernel) was used for the kernel $k(r) = \exp (-r / \ell)$, with $r = | x - y |$, in connection with the Ornstein-Uhlenbeck process.  Even though the name ``kernel'' came from the theorie of integral operator \cite{rasmussen2006gaussian}, p.~80, the attribution of the exponential kernel to Laplace came from the \href{https://en.wikipedia.org/w/index.php?title=Laplace_distribution&oldid=1106362547}{Laplace probability distribution} (Wikepedia, version 06:55, 24 August 2022), also called the ``double exponential'' distribution, but not from the different kernel used in the Laplace transform. See also Remark~\ref{rm:Gaussian-Laplacian-kernels} and Figure~\ref{fig:Gaussian-process-prior}.
} $\K_L$ in Eq.~\eqref{eq:exponential-kernel-1}$_2$: 
%
%\noindent
%Popular smooth Gaussian kernel:
\begin{align}
	\K_G ( \bx , \by) 
	= \exp 
	\left( - \frac{\parallel  \bx - \by  \parallel^2}{2 \stdev^2} \right) \ , \quad
	\K_L (\bx , \by) 
	= \exp 
	\left( - \frac{\parallel \bx - \by \parallel}{\stdev} \right) \ ,
	\label{eq:exponential-kernel-1}
\end{align} 
where $\stdev$ is the standard deviation.  The method in Remark~\ref{rm:reproducing-kernel} is not suitable to show that these exponential functions are reproducing kernels.  We provide a verificatrion of the reproducing property of the Laplacian kernel in Eq.~\eqref{eq:exponential-kernel-1}$_2$.

\begin{rem}
	Laplacian kernel is reproducing.
	{\rm
		Consider the Laplacian kernel in Eq.~\eqref{eq:exponential-kernel-1}$_2$ for the scalar case $(x,y)$ with $\stdev = 1$ for simplicity, without loss of generality.  The goal is to show that the reproducing property in Eq.~\eqref{eq:reproducing-kernel}$_1$ holds for such kernel expressed in Eq.~\eqref{eq:kernel-Laplacian-scalar-1}$_1$ below.  
		\begin{align}
			&
			\K(x,y) = \exp \left[ - | x - y | \right]
			\Rightarrow 
			\K^\prime (x,y) := 
			\frac{\partial \K(x,y)}{\partial y} = 
			\begin{cases}
				- \K (x,y) \text{ for } y > x
				\\
				\phantom{-} \K (x,y) \text{ for } y < x
			\end{cases}
		\label{eq:kernel-Laplacian-scalar-1}
			\\
			&
			\Rightarrow
			\K^{\prime\prime} (x,y) :=
			\frac{\partial^2 \K(x,y)}{(\partial y)^2}
			= \K (x,y) \ , \text{ for } y \ne x \ .
			\label{eq:kernel-Laplacian-scalar-2}
		\end{align}
		The method is by using integration by parts and by using a function norm different from that in Eq.~\eqref{eq:reproducing-kernel}$_2$; see \cite{berlinet2004reproducing}, p.~8.
		Now start with the integral in Eq.~\eqref{eq:reproducing-kernel}$_2$, and do integration by parts:
		\begin{align}
			&
%			\displaystyle
			\int f(y) \K (x,y) dy 
			= \int f(y) \K^{\prime\prime} (x,y) dy
			= \int \left[ f \K^\prime \right]^\prime dy - \int f^\prime \K^\prime dy
			= \left[ f \K^\prime \right]_{y = -\infty}^{y = + \infty} - \int f^\prime \K^\prime dy
			\\
			&
%			\displaystyle
			\left[ f \K^\prime \right]_{y = -\infty}^{y = + \infty}
			= \left[ f (+\K) \right]_{y = -\infty}^{y = x^-} + \left[ f (-\K) \right]_{y = x^+}^{y = +\infty} 
%			- \int f^\prime \K^\prime dy 
			= f(x^-) \K (x^- , x) + f(x^+) \K (x^+ , x) 
%			- \int f^\prime \K^\prime dy
			= 2 f(x)
			\\
			&
%			= 2 f(x) - \int f^\prime \K^\prime dy
			\Rightarrow
			\int f (y) \K (x,y) dy + \int f^\prime (y) \K^\prime (x,y) dy = 2 f(x) \ ,
			\label{eq:kernel-Laplacian-int-by-parts}
		\end{align}
		where $x^- = x - \epsilon$ and $x^+ = x + \epsilon$ with $\epsilon > 0$ being very small.
		The scalar product on the space of functions that are differentiable almost everywhere, i.e.,
		\begin{align}
			\langle f , g \rangle = \frac12 \left[\int f (y) g (y) dy + \int f^\prime (y) g^\prime (y) dy \right] \ ,
			\label{eq:differentiable-functions-product}
		\end{align} 
		together with Eq.~\eqref{eq:kernel-Laplacian-int-by-parts}, and $g (y) = \K (x,y)$, show that the Laplacian kernel is reproducing.\footnote{
			It is possible to define the Laplacian kernel in Eq.~\eqref{eq:kernel-Laplacian-scalar-1}$_1$ with the factor $\frac12$, which then will not appear in the definition of the scalar product in Eq.~\eqref{eq:differentiable-functions-product}.  See also \cite{berlinet2004reproducing}, p.~8.
		}
	}
	{$\hfill\blacksquare$}
\end{rem}

%\noindent
%Non-smooth Laplacian (exponential) kernel:
%\begin{align}
%	\K_L (\bx , \by) 
%	= \exp 
%	\left( - \frac{\parallel \bx - \by \parallel}{\stdev} \right)
%\end{align}

%\subsection{Positive-definite functions are reproducing kernels}

%\begin{rem}
%	\label{rm:kernel-methods}
%	{\rm \emph{Kernel methods or kernel machines}.  
	%		It is appropriate here to briefly describe the kernel methods.
	%	}
%\end{rem}

\subsection{Gaussian processes}
\label{sc:Gaussian-process}
The Kalman filter, well-known in engineering, is an example of a Gaussian-process model.
%		A well-known Gaussian process in engineering is the Kalman filter.
See also Remark~\ref{rm:PINN-patent} in Section~\ref{sc:PINN-frameworks} on the 2021 US patent on Physics-Informed Learning Machine that was based on Gaussian processes (GPs),\footnote{
	Non-Gaussian processes, such as in \cite{yaida2020non} \cite{sendera2021non}, are also important, but more advanced, and thus beyond the scope of the present review. 
	See also the Gaussian-Process Summer-School videos (Youtube) \href{https://www.youtube.com/playlist?list=PLZ_xn3EIbxZHoq8A3-2F4_rLyy61vkEpU}{2019} and \href{https://www.youtube.com/playlist?list=PLZ_xn3EIbxZGcqHGFj-P_SI6OCXy8TfoL}{2021}. 
	We thank David Duvenaud for noting that we did not review non-Gaussian processes.
} which possess the ``most pleasant resolution imaginable'' to the question of how to computationally deal with infinite-dimensional objects like functions \cite{rasmussen2006gaussian}, p.~2: 
\begin{quote}
	``If you ask only for the properties of the function at a finite number of points, then the inference from the Gaussian process will give you the same answer if you ignore the infinitely many other points, as if you would have taken them into account!  And these answers are consistent with any other finite queries you may have.  One of the main attractions of the Gaussian process framework is precisely that it unites a sophisticated and consistent view with computational tractability.''
\end{quote}

A simple example of a Gaussian process is the linear model $f(\cdot)$ below \cite{bishop2006pattern}:
\begin{align}
	&
	y = f(x) 
	= \sum_{k=0}^{k=n} w_k x^k 
	=
	\bkar{w}^T \bkar{\phi} (x)
	\ ,  \text{ with }
	\bkar{w}^T 
	= \lfloor w_0 , \ldots , w_n \rfloor \in \real^{1 \times (n+1)}
	\ , \ 
	w_k \sim {\mathcal N} (0, 1) \ ,
	\label{eq:gaussian-process-1}
	\\
	& 
	\bkar{\phi} (x) = [ \phi_0 (x) , \ldots , \phi_n(x) ]^T
	\in \real^{(n+1) \times 1}
	\ ,  
	 \text{ and }
	\phi_k (x) = x^k \ , 
	\label{eq:gaussian-process-2}
\end{align}
with random coefficients $w_k$,  $k=0, \ldots , n$, being normally distributed with zero mean $\mu = 0$ and unit variance $\sigma^2 = 1$, and with basis functions $\{ \phi_k (x) = x^k , k=0, \ldots , n  \}$.  

More generally, $\bkar{\phi} (x)$ could be any basis of nonlinear functions in $x$, and the weights in $\bkar{w} \in \real^{n \times 1}$ could have a joint Gaussian distribition with zero mean and a given covariance marix $\bkar{C}_{\bkar{w} \bkar{w}} = \bkar{\Sigma} \in \real^{n \times n}$, i.e, $\bkar{w} \sim \mathcal{N} (\bkar{0}, \bkar{\Sigma})$.   If $w_k$, $k = 1, \ldots , n$, are idependently and identically distributed (i.i.d.), with variance $\sigma^2$, then the covariance matrix is diagonal (and is called ``isotropic'' \cite{bishop2006pattern}, p.~84), i.e., $\bkar{\Sigma} = \sigma^2 \bkar{I}$ and $\bkar{w} \sim \mathcal{N} (\bkar{0}, \sigma^2 \bkar{I})$, with $\bkar{I}$ being the identity matrix.

Formally, a \emph{Gaussian process} is a probability distribution over the functions $f(x)$ such that, for any given arbitrary set of input training points $\bx = [ x_1 , \ldots, x_m ]^T \in \real^{m \times 1}$, the set of output values of $f(\cdot)$ at input training points, i.e., $\by = [y_1 , \ldots , y_m ]^T = [ f(x_1) , \ldots , f(x_m) ]^T \real^{m \times 1}$, such that $y_i = f(x_i)$, which from Eq.~\eqref{eq:gaussian-process-1} can be written as
\begin{align}
	\by = (\bkar{w}^T \bkar{\Phi})^T = \bkar{\Phi}^T \bkar{w}
	\ , \quad
	y_j = f(x_j) = \sum_{i=1}^{n} w_i \phi_i (x_j)
	\ , \ j = 1, \ldots , m
	\ , \text{ such that } 
%	w_i \sim \mathcal{N} (0, \stdev^2)
	\bkar{w} \sim \mathcal{N} (\bkar{0} , \bkar{\Sigma})
	\ ,
	\label{eq:gaussian-process-3}
	%			\ , \quad
	%			\bkar{w} = [w_1 , \ldots , w_N]^T 
	%			\ , \quad
	%			\bkar{\Phi} = \left[ \phi_i (x_j) \right] \in \real^{N \times M}
\end{align}
%with weights $\bkar{w} = [w_1 , \ldots , w_N]^T \in \real^{N \times 1}$, such that $w_i \sim \mathcal{N} (0, \stdev^2)$, 
and the \emph{design} matrix $\bkar{\Phi} = \left[ \phi_i (x_j) \right] \in \real^{n \times m}$, 
has a joint probability distribution; see \cite{bishop2006pattern}, p.~305.  

Another way to put it succinctly, a Gaussian process describes a distribution over functions, and is defined as a collection of random variables (representing the values of the function $f(\bx)$ at location $\bx$), such that any finite subset of which has a joint probability distribution \cite{rasmussen2006gaussian}, p.~13. 

%		a Gaussian process is collection of random variables, such that any finite subset of these variables has a joint probability distribution; see, e.g., \cite{wilson2014covariance}, p.~34, for a formal proof for the case $N=1$.

The multivariate (joint probability) Gaussian distribution for an $m \times 1$ matrix $\by$, with mean $\bkar{\mu}$ (element-wise expectation of $\by$) and covariance matrix $\bkars{C}{\by \by}$  (element-wise expectation of $\by \by^T$) is written as
\begin{align}
	&
	\mathcal{N} (\by | \bkar{\mu} , \bkars{C}{\by \by})
	=
	\frac{1}{(2 \pi)^{m/2} \det \bkars{C}{\by \by}}
	\exp \left[ -\frac{1}{2} \left(\by - \bkar{\mu} \right)^T \bkarsp{C}{\by \by}{-1}  \left(\by - \bkar{\mu} \right) \right]
	\ , \text{ with }
	\label{eq:multivariate-gaussian-distribution-1}
	\\
	&
	\bkar{\mu} = \mathbb{E} [\by] = \mathbb{E} \left[\bkar{w}^T\right] \bkar{\Phi} = \bkar{0}
	\ , 
	\label{eq:multivariate-gaussian-distribution-2}
	\\
	&
	\bkars{C}{\by \by} = \mathbb{E} \left[ \by \by^T \right]
	= \bkar{\Phi}^T \mathbb{E} \left[ \bkar{w} \bkar{w}^T \right] \bkar{\Phi}
	= \bkar{\Phi}^T \bkar{C}_{\bkar{w} \bkar{w}} \bkar{\Phi}
	= \bkar{\Phi}^T \bkar{\Sigma} \bkar{\Phi}
	= \left[ \K (x_i , x_j )\right] \in \real^{m \times m}
%	= \stdev^2 \bkar{\Phi}^T \bkar{\Phi} \in \real^{m \times m}
	\ ,
	\label{eq:multivariate-gaussian-distribution-3}
\end{align}
where $\bkar{C}_{\bkar{w} \bkar{w}} = \bkar{\Sigma} \in \real^{n \times n}$ is a given covariance matrix of the weight matrix $\bkar{w} \in \real^{n \times 1}$. The covariance matrix $\bkars{C}{\by \by}$ in Eq.~\eqref{eq:multivariate-gaussian-distribution-3} has the same mathematical structure as Eq.~\eqref{eq:kernel-basis-functions}, and is therefore a reproducing kernel, with kernel function $k (\cdot , \cdot)$
\begin{align}
	\text{cov} ( y_i , y_j ) 
	= \text{cov} ( f(x_i) , f(x_j) )
	= \mathbb{E} \left[ f(x_i) , f(x_j) \right]
	= \K (x_i , x_j) 
	= \sum_{p=1}^{n} \sum_{q=1}^{n} \phi_p(x_i) \Sigma_{pq} \phi_q(x_j) 
	\ .
	\label{eq:kernel-gaussian-process-0}
\end{align}
In the case of an isotropic covariance matrix $\bkars{C}{\bkar{w} \bkar{w}}$, the kernel function $k(\cdot , \cdot)$ takes a simple form:\footnote{
	The ``precision'' is the inverse of the variance, i.e., $\sigma^{-2}$ \cite{bishop2006pattern}, p.~304.
}
\begin{align}
	&
	\bkars{C}{\bkar{w} \bkar{w}} = \sigma^2 \bkar{I}
	\Rightarrow
	\K ( y_i , y_j ) = \K ( f(x_i) , f(x_j) ) 
	= \sigma^2 \sum_{p=1}^{n} \phi_p(x_i) \phi_p(x_j) 
	= \sigma^2 \bkar{\phi}^T (x_i) \bkar{\phi} (x_j)
	\ ,
	\label{eq:kernel-gaussian-process-1}
	\\
	&
	\text{with }
	\bkar{\phi} (x) = \left[ \phi_1 (x) , \ldots , \phi_n (x) \right]^T \in \real^{n \times 1}
	\ .
	\label{eq:kernel-gaussian-process-2}
\end{align}
The Gaussian (normal) probability distribution $\mathcal{N} (\by | \bkar{\mu} , \bkars{C}{\by \by})$ in Eq.~\eqref{eq:multivariate-gaussian-distribution-1} is the \emph{prior} probability distribution for $\by$, before any conditioning with observed data (i.e., before specifying the actual observed values of the outputs in $\by = f(\bx)$).

\begin{figure}[h]
	\centering
	\includegraphics[width=0.95\linewidth]{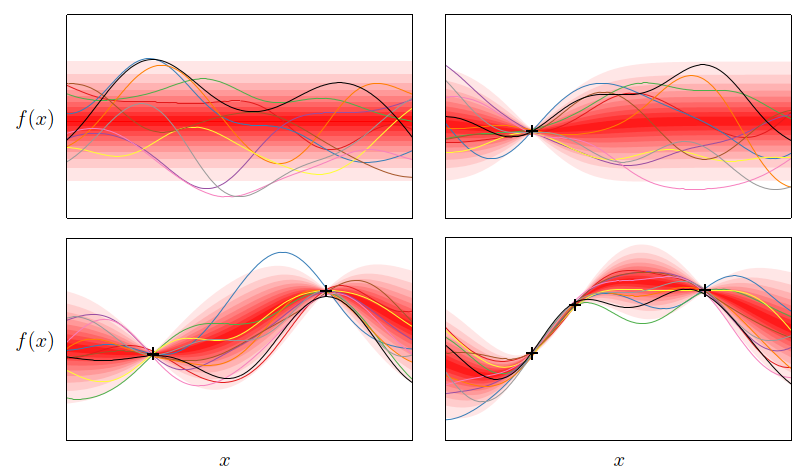}
	\caption{
		\emph{Gaussian process prior and posterior samplings, Gaussian kernel} (Section~\ref{sc:Gaussian-process}).
		\emph{Top left:} Gaussian-prior samples
%		 (Remark~\ref{rm:Gaussian-prior-sampling}). 
		(Section~\ref{rm:Gaussian-prior-sampling}).
		The shaded red zones represent the predictive density of at each input location.
		\emph{Top right:} Gaussian-posterior samples with 1 data point.
		\emph{Bottom left:} Gaussian-posterior samples with 2 data points.
		\emph{Bottom right:} Gaussian-posterior samples with 3 data points \cite{duvenaud2014automatic}.
		See Figure~\ref{fig:GP-Gaussian-kernel-3-data-points-noise} for the noise effects and 
		Figure~\ref{fig:GP-animation} for animation of GP priors and posteriors.
		{
			\footnotesize (Figure reproduced with permission of the authors.)
		} 
	}
	\label{fig:GP-Gaussian-kernel-3-data-points}
\end{figure}

\begin{figure}[h]
	\centering
	\includegraphics[width=0.70\linewidth]{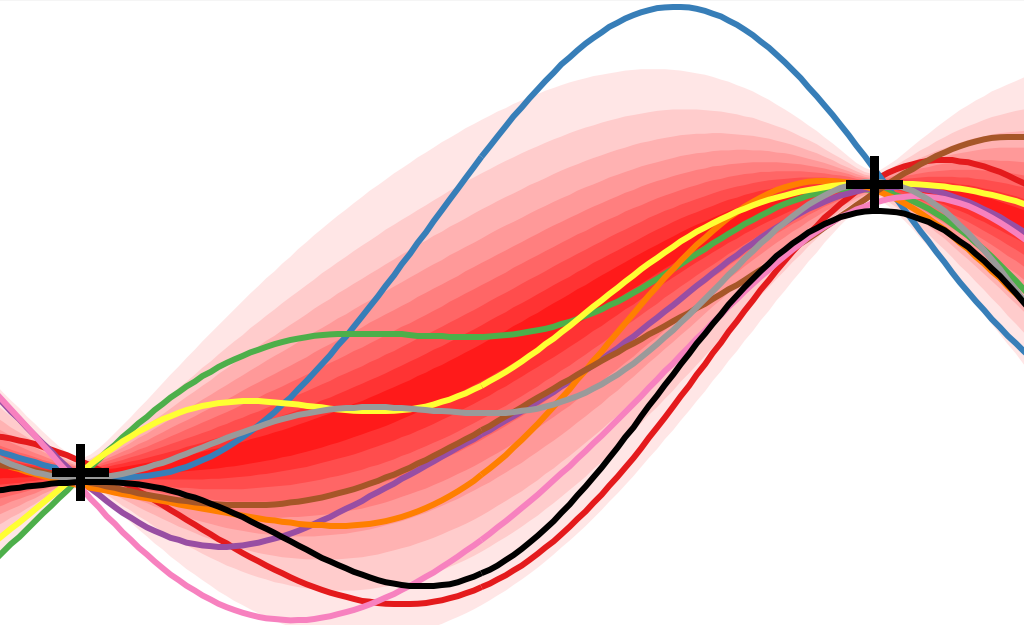}
	\caption{
		\emph{Gaussian process posterior samplings, noise effects} (Section~\ref{sc:Gaussian-process}).  Not all sampled curves in Figure~\ref{fig:GP-Gaussian-kernel-3-data-points} went through the data points, such as the black line in the present zoomed-in view of the bottom-left subfigure \cite{duvenaud2014automatic}.
		It is easy to make the sampled curves passing closer to the data points simply by reducing the noise variance $\nu^2$ in Eqs.~\eqref{eq:Gaussian-posterior-0}, \eqref{eq:Gaussian-posterior-2}, \eqref{eq:Gaussian-posterior-3}.
		See Figure~\ref{fig:GP-animation} for animation of GP priors and posteriors.
		{
			\footnotesize (Figure reproduced with permission of the authors.)
		} 
	}
	\label{fig:GP-Gaussian-kernel-3-data-points-noise}
\end{figure}

\begin{figure}[h]
	\centering
	\includegraphics[width=0.80\linewidth]{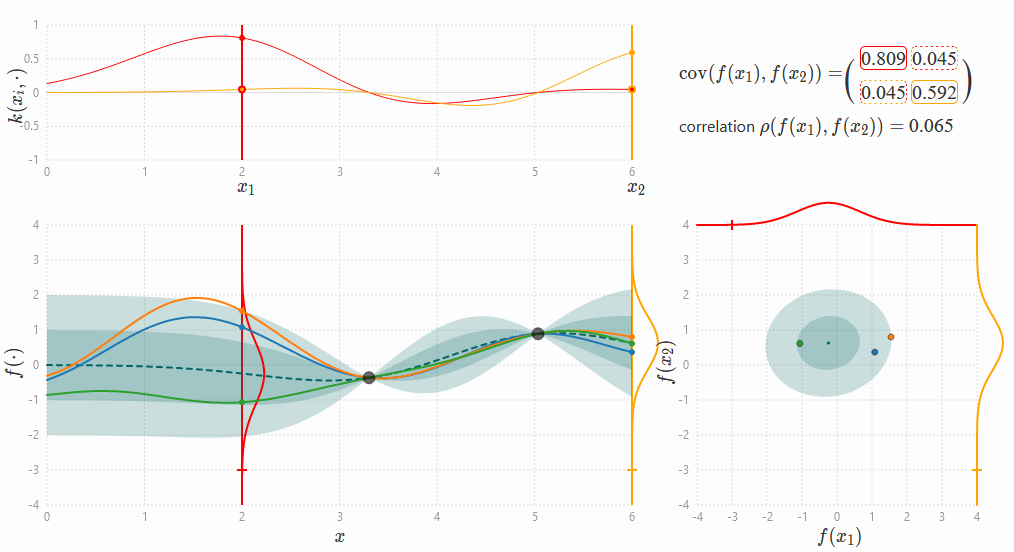}
	\caption{
		\emph{Gaussian process posterior samplings, animation} (Section~\ref{sc:Gaussian-process}). 
		\href{http://www.infinitecuriosity.org/vizgp/}{Interactive Gaussian Process Visualization}, \href{http://www.infinitecuriosity.org/}{Infinite curiosity}. 
		Click on the plot area to specify data points. 
		See Figures~\ref{fig:GP-Gaussian-kernel-3-data-points} and \ref{fig:GP-Gaussian-kernel-3-data-points-noise}.
	}
	\label{fig:GP-animation}
\end{figure}

\begin{rem}
	\label{rm:Gaussian-processes-key-points}
	Zero mean.
	{\rm
		In a Gaussian process, the joint distribution Eq.~\eqref{eq:multivariate-gaussian-distribution-1} over the outputs, i.e., the $n$ random variables in $\by = f (\bx) \in \real^{n \times 1}$, is defined completely by ``second-order statistics,'' i.e., the mean $\bkar{\mu} = \bkar{0}$ and the covariance $\bkars{C}{\by \by}$.
		In practice, the mean of $\by = f(\bx)$ is not available a priori, and ``so by symmetry, we take it to be zero''  \cite{bishop2006pattern}, p.~305, or equivalently, specify that the weights in $\bkar{w} \in \real^{n \times 1}$ have zero mean, as in Eqs.~\eqref{eq:gaussian-process-1}, \eqref{eq:gaussian-process-3}, \eqref{eq:multivariate-gaussian-distribution-2}.
	}
	$\hfill\blacksquare$
\end{rem}

\subsubsection{Gaussian-process priors and sampling}
%\begin{rem}
	\label{rm:Gaussian-Laplacian-kernels}
	\label{rm:Gaussian-prior-sampling}
	\label{sc:Gaussian-prior-sampling}
%	Gaussian-prior sampling.
	{\rm   
		Instead of defining the kernel function $\K (\cdot , \cdot)$ by selecting a basis of functions as in Eq.~\eqref{eq:kernel-gaussian-process-0}, the kernel function can be directly defined using the analytical expressions in Table~\ref{tb:reproducing-kernels} and Section~\ref{sc:kernel-exponential-functions}.
		Figure~\ref{fig:Gaussian-process-prior} provides a comparison of samples from two Gaussian-process priors, one with Gaussian kernel $\K_G$ and one with Laplacian kernel $\K_L$, as given in Eq.~\eqref{eq:exponential-kernel-1} with precision $\gamma = \stdev^{-2}$.
		Since the Gram matrix $\bkar{K} = \left[ \K (x_i , x_j) \right]$ for a kernel $\K$ is positive semidefinite\footnote{
			That the Gram matrix is positive semidefinite should be familiar with practitioners of the finite element method, in which the Gram matrix is the stiffness matrix (for elliptic differential operators), and is positive semidefinite before applying the essential boundary conditions.
		} (\cite{bishop2006pattern}, p.~295), to make the Gram matrix $\bkar{K} = \bkars{C}{\by \by}$ in Eq.~\eqref{eq:gaussian-process-3} positive definite before performing a Choleski decomposition, a noise with isotropic variance $\nu^2 \bkar{I}$ is added to $\bkar{K}$ (\cite{bishop2006pattern}, p.~314), meaning that the noise is the same and independent in each direction.
		Then the sample function values can be obtained as follows (\cite{bishop2006pattern}, p.~528):
		\begin{align}
			\bkars{C}{\by \by , \nu}
			= \bkarsp{\Phi}{\nu}{T} \bkar{\Sigma} \bkar{\Phi}_\nu
			=
			\bkar{K} + \nu^2 \bkar{I} = \bkar{L} \bkar{L}^T 
			\ , \text{ with }
			\bkar{w} \sim \mathcal{N} (\bkar{0} , \bkar{\Sigma} = \bkar{I})
			\ , \text{ then }
			\by = \bkarsp{\Phi}{\nu}{T} \bkar{w} = \bkar{L} \bkar{w}
			\ .
			\label{eq:Gram-matrix-perturned}
		\end{align}
		In other words, the GP prior samples in Figure~\ref{fig:Gaussian-process-prior} were drawn from the Gaussian distribution with zero mean and covariance matrix $\bkars{C}{\by \by , \nu}$ in Eq.~\eqref{eq:Gram-matrix-perturned}:
		\begin{align}
%			\bkars{f}{\star} 
%			\by_\star
			\byt
%			:= f(\bx_\star)
			:= f(\bxt)
			\sim \mathcal{N} ( \bkar{y} | \bkar{\mu}  , \bkars{C}{\by \by , \nu} )
			= \mathcal{N} ( \bkar{0} ,  \bkar{K} + \nu^2 \bkar{I})
			\ , 
			\label{eq:Gaussian-prior}
		\end{align}
		where 
%		$\bx_\star$ 
		$\bxt$
		contains the test inputs, and 
%		$\bkars{f}{\star} = f(\bx_\star)$ 
%		$\by_\star = f(\bx_\star)$
		$\byt = f(\bxt)$
		contains the predictive values of $f(\cdot)$ at $\bxt$.
	
		It can be observed from Figure~\ref{fig:Gaussian-process-prior} that samples obtained with the Gaussian kernel were smooth, with slow variations, whereas samples obtained with the Laplacian kernel had high jiggling with rapid variations, appropriate to model Brownian motion; see Footnote~\ref{fn:Laplacian-kernel-Brownian-motion}.
	}
%	$\hfill\blacksquare$
%\end{rem}

%\begin{rem}
%	Gaussian-posterior sampling.
%	{\rm 
%		\cite{rasmussen2006gaussian}, p.~200.
%	}
%\end{rem}

\subsubsection{Gaussian-process posteriors and sampling}
\label{sc:Gaussian-posterior-sampling}
Let $\by = \left[ y_1 , \ldots , y_m \right]^T \in \real^{m \times 1}$ be the observed data (or target values) at the training points $\bx = \left[ x_1 , \ldots , x_m \right]^T \in \real^{m \times 1}$, and let 
%$\bxt \in \real^{m_\star \times 1}$ 
$\bxt \in \real^{\mt \times 1}$
contain the $\mt$ test input points, with the predictive values in 
%$\bkars{f}{\star} = f(\bx_\star) \in \real^{m_\star \times 1}$.  
$\byt = f(\bxt) \in \real^{\mt \times 1}$.
The combined function values in the matrix 
%$\left[ \by , \bkars{f}{\star} \right]^T \in \real^{(m + m_\star) \times 1}$, 
$\left[ \by , \byt \right]^T \in \real^{(m + \mt) \times 1}$,
as random variables, are distributed ``normally'' (Gaussian), i.e.,
\begin{align}
	\begin{Bmatrix}
		\f (\bx)
		\\
		\f (\bxt)
	\end{Bmatrix}
	=
	\begin{Bmatrix}
		\by
		\\
%		\bfstar
		\byt
	\end{Bmatrix}
	=
	\mathcal{N}
	\left( 
		\begin{Bmatrix}
			\bkar{\mu} (\bx)
			\\
			\bkar{\mu} (\bxt)
		\end{Bmatrix}
		\ ,
		\begin{bmatrix}
			\bkar{K} (\bx , \bx) + \nu^2 \bkar{I} & \bkar{K} (\bx , \bxt)
			\\
			\bkar{K}^T (\bx , \bxt) & \bkar{K} (\bxt , \bxt)
		\end{bmatrix}
	\right)
	\ .
	\label{eq:Gaussian-posterior-0}
\end{align}
The Gaussian-process posterior distribution, i.e., the conditional Gaussian distribution for the test output $\byt$ given the training data $(\bx , \by = f(\bx))$ is then (See Appendix~\ref{app:Gaussian-distribution-conditional} for the detailed derivation, which is simpler than in \cite{bishop2006pattern} and in \cite{mises1964mathematical})\footnote{
	\label{fn:Gaussian-distribution-conditional}
	See, e.g., \cite{rasmussen2006gaussian}, p.~16, \cite{bishop2006pattern}, p.~87, \cite{duvenaud2014automatic}, p.~4.  The authors of \cite{rasmussen2006gaussian}, in their Appendix A.2, p.~200, referred to \cite{mises1964mathematical} ``sec.~9.3'' for the derivation of Eqs.~\eqref{eq:Gaussian-posterior-2}-\eqref{eq:Gaussian-posterior-3}, but there were several sections numbered ``9.3'' in \cite{mises1964mathematical}; the correct referencing should be \cite{mises1964mathematical}, Chapter XIII ``More on distributions,'' Sec.~9.3 ``Marginal distributions and conditional distributions,'' p.~427.
}  
\begin{align}
	&
	p ( \byt \, | \, \bx , \by , \bxt )
	=
	\mathcal{N} (\byt \, | \, \bmut , \bkars{C}{\byt \byt} )
	\ ,
	\label{eq:Gaussian-posterior-1}
	\\
	&
	\bmut = \bmu (\bxt) + \bkar{K} (\bxt , \bx) \left[ \bkar{K} (\bx , \bx) + \nu^2 \bkar{I} \right]^{-1} \left[ \by - \bmu (\bx) \right]
	= \bkar{K} (\bxt , \bx) \bkar{K}^{-1} (\bx , \bx) \ \by
	\ ,
	\label{eq:Gaussian-posterior-2}
	\\
	&
	\bkars{C}{\byt \byt} = \bkar{K} (\bxt , \bxt) - \bkar{K} (\bxt , \bx) \left[ \bkar{K} (\bx , \bx) + \nu^2 \bkar{I} \right]^{-1} \bkar{K} (\bx , \bxt)
	\ ,
	\label{eq:Gaussian-posterior-3}
\end{align}
where the mean was set to zero by Remark~\ref{rm:Gaussian-processes-key-points}.
In Figure~\ref{fig:GP-Gaussian-kernel-3-data-points}, the number $m$ of training points varied from 1 to 3.
The Gaussian posterior sampling follows the same method as in Eq.~\eqref{eq:Gram-matrix-perturned}, but with the covariance matrix in Eq.~\eqref{eq:Gaussian-posterior-3}:\footnote{
	The Matlab \href{https://github.com/duvenaud/phd-thesis/blob/62ff4d61f27e7f83bc968324cc1f864a1ea7344c/code/plot_oned_gp.m}{code} for generating Figures~\ref{fig:GP-Gaussian-kernel-3-data-points} and \ref{fig:GP-Gaussian-kernel-3-data-points-noise} was provided courtesy of David Duvenaud.  On line 25 of the code, the noise variance $\nu^2$ was set as {\tt sigma = 0.02}, which was much larger than $\nu^2 = 10^{-6}$ used in Figure~\ref{fig:Gaussian-process-prior}.
}
\begin{align}
	\bkars{C}{\byt \byt}
	= \bkart{L} \bkart{L}^T 
	\ , \text{ with }
	\bkar{w} \sim \mathcal{N} (\bkar{0} , \bkar{\Sigma} = \bkar{I})
	\ , \text{ then }
	\byt = \bkart{L} \bkar{w}
	\ .
	\label{eq:Gram-matrix-perturned-2}
\end{align}

%\noindent
%{\color{red} [NOTE: 2022.11.22 - I am HERE, finishing up Gaussian processes.  ENDNOTE]}

\section{Deep-learning libraries, frameworks, platforms}
\label{sc:libraries}

Among factors that drove the resurgence of AI, see Section~\ref{sc:resurgence}, the availability of effective computing hardware and libraries that facilitate leveraging that hardware for DL-purposes have played and continue to play an important role in terms of both expansion of research efforts and dissemination in applications.
Both commercial software, but primarily open-source libraries, which are backed by major players in software industry and academia, have emerged over the last decade, see e.g., Wikipedia's ``Comparison of deep learning software'' \href{https://en.wikipedia.org/w/index.php?title=Comparison_of_deep_learning_software&oldid=1105085605}{version 12:51, 18 August 2022}.

\begin{figure}[h]
	\centering
	\includegraphics[width=0.9\linewidth]{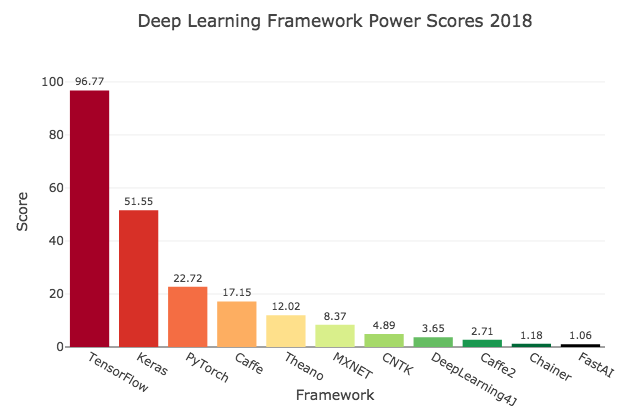}
	\caption{
		Top deep-learning libraries in 2018 by the ``Power Score'' in
		\cite{Hale.2018:rd0001}.  By 2022, using Google Trends, the popularity of different frameworks is significantly different; see Figure~\ref{fig:Google_trends-ML_frameworks}.
%		{\color{red}
%		NOTE: once this review paper is done, write to Hale on the webpage of this article with the following message: ``My colleague and I are writing a review paper, in which we used a figure of yours.  Based on a previous request by Tony Holdroyd (around 2018), to whom you gave permission to use a bar chart, we assume that you would give permission to use the figure.  Once the article is published in the Computer Modeling in Engineering and Science (http://www.techscience.com/cmes/), we will provide the link to the online article in this webpage.''
%		}
	}
	\label{fig:Hale-top-DL-libraries}
\end{figure}

Figure~\ref{fig:Hale-top-DL-libraries} compares the popularity (as of 2018) of various software frameworks of the DL-realm by means of their  ``Power Scores''.\footnote{
	See~\cite{Hale.2018:rd0001}.
}
The ``Power Score'' metric was computed from the occurrences of the respective libraries on 11 different websites, which range from scientific ones, e.g., as world's largest storage for research articles and preprints \href{https://arxiv.org}{arXiv} to social media outlets as, e.g., \href{https://linkedin.com}{Linkedin}.

The impressive pace at which DL research and applications progress is also reflected in changes the software landscape has been subjected to. 
As of 2018, TensorFlow was clearly dominant, see Figure~\ref{fig:Hale-top-DL-libraries}, whereas Theano was the only library around among those only five years earlier according to the author of the 2018 study.

%\cite{Clark.2018:rd0001}, \cite{Hale.2018:rd0001}, 
%``Comparison of deep learning software'', Wikipedia,  \href{https://en.wikipedia.org/w/index.php?title=Comparison_of_deep_learning_software&oldid=878809707}{version 02:01, 17 January 2019}

\begin{figure}[h]
	\centering
	\includegraphics[width=0.8\textwidth]{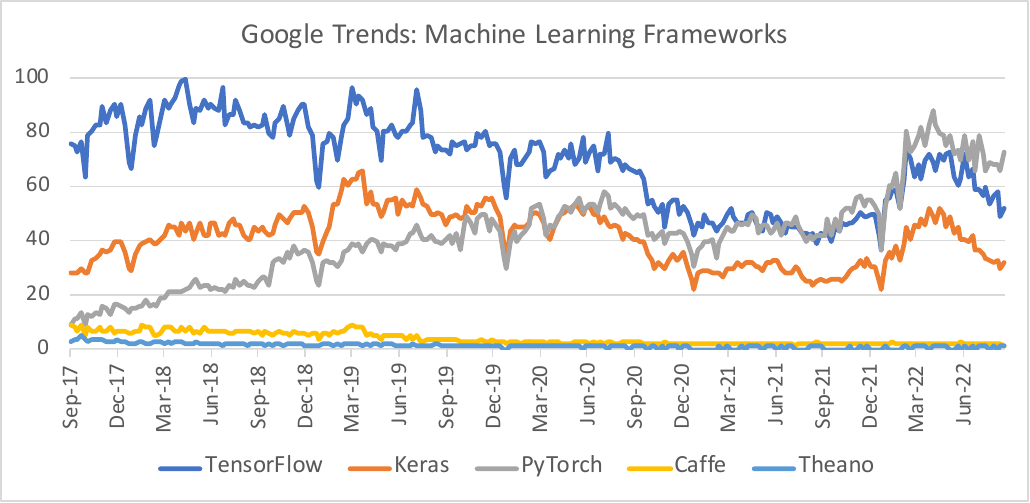}
	\caption{
		{\it Google Trends of deep-learning software libraries} (Section~\ref{sc:libraries}).
		The chart shows the popularity of five DL-related software libraries most ``powerful'' in 2018 over the last 5 years (as of July 2022).   See also Figure~\ref{fig:Hale-top-DL-libraries} for the rankings of DL frameworks in 2018.
	}
	\label{fig:Google_trends-ML_frameworks}
\end{figure}

As of August 2022, the picture has once again changed quite a bit.
Using Google Trends as metric,\footnote{
	See \href{https://bit.ly/3KPgWFx}{this link} for the latest Google Trends results corresponding to Figure~\ref{fig:Google_trends-ML_frameworks}. 
} PyTorch, which was in third place in 2018, has taken the leading position from TensorFlow as most popular DL-related software framework, see Figure~\ref{fig:Google_trends-ML_frameworks}.

Although the individual software frameworks do differ in terms of functionality, scope and internals, their overall purpose are clearly the same, i.e., to facilitate creation and  training of neural networks and to harness the computational power of parallel computing hardware as, e.g., GPUs.
For this reason, libraries share the following ingredients, which are essentially similar but come in different styles:

\begin{itemize}
	\item 
		\emph{Linear algebra:} In essence, DL boils down to algebraic operations on large sets of data arranged in multi-dimensional arrays, which are supported by all software frameworks, see Section~\ref{sc:network-layer-details}.\footnote{
		In the context of DL, multi-dimensional arrays are also referred to a tensors, although the data often lacks the defining properties of tensors as algebraic objects.
		The software framework \emph{TensorFlow} even reflects that by its name.
	}
	\item 
		\emph{Back-propagation:} Gradient-based optimization relies on efficient evaluation of derivatives of loss functions with respect to network parameters. 
		The representation of algebraic operations as computational graphs allows for automatic differentiation, which is typically performed in \emph{reverse-mode}, hence, back-propagation, see Section~\ref{sc:backprop}.
	\item 
		\emph{Optimization:} DL-libraries provide a variety of optimization algorithms that have proven effective in training of neural networks, see Section~\ref{sc:training}.
	\item 
		\emph{Hardware-acceleration:} Training deep neural networks is computationally intensive and requires adequate hardware, which allows algebraic computations to be performed in parallel.
		DL-software frameworks support various kinds of parallel/distributed hardware ranging from multi-threaded CPUs to GPUs to DL-specific hardware as TPUs.
		Parallelism is not restricted to algebraic computations only, but data also has to be efficiently loaded from storage and transferred to computing units.
	\item
		\emph{Frontend and API:} Popular DL-frameworks provide an intuitive API, which supports accessibility for first-time learners and dissemination of novel methods to scientific fields beyond computer science. 
		\emph{Python} has become the prevailing programming language in DL, since it is more approachable for less-proficient developers as compared to languages traditionally popular in computational science as, e.g., C++ or Fortran.
		High-level APIs provide all essential building blocks (layers, activations, loss functions, optimizers, etc.) for both construction and training of complex network topologies and fully abstract the underlying algebraic operations from users.
\end{itemize}

In what follows, a brief description of some of the most popular software frameworks is given.

%\begin{figure}[h]
%	\centering
%	\includegraphics[width=0.9\linewidth]{figures/Clark-top-DL-libraries}
%	\caption{
%		Top 16 deep-learning libraries. 
%		\cite{Clark.2018:rd0001}
%		{\color{red} ASK PERMISSION}
%	}
%	\label{fig:Clark-top-DL-libraries}
%\end{figure}

\subsection{TensorFlow}
\emph{TensorFlow}~\cite{tensorflow2015-whitepaper} is a free and open-source software library which is being developed by the \emph{Google Brain} research team, which, in turn, is part of Google's AI division.
TensorFlow emerged from Google's proprietary predecessor ``DistBelief'' and was released to public in November 2015.
Ever since its release, TensorFlow has rapidly become the most popular software framework in the field of deep-learning and maintains a leading position as of 2022, although it has been outgrown in popularity by its main competitor PyTorch particularly in research. 

In 2016, Google presented its own AI accelerator hardware for TensorFlow called ``Tensor Processing Unit'' (TPU), which is built around an application-specific integrated circuited (ASIC) tailored to computations needed in training and evaluation of neural networks.
DeepMind's grandmaster-beating software AlphaGo, see Section~\ref{sc:motivation} and Figure~\ref{fig:protein-folding-Go-game}, was trained using TPUs.\footnote{
	See Google's announcement of TPUs~\cite{Jouppi.2016}.
}
TPUs were made available to the public as part of ``Google Cloud'' in 2018.
A single fourth-generation TPU device has a peak computing power of \num{275} teraflops for \num{16} bit floating point numbers (\texttt{bfloat16}) and \num{8} bit integers (\texttt{int8}).
A fourth-generation cloud TPU ``pod'', which comprises \num{4096} TPUs offers a peak computing power of \num{1.1} exaflops.\footnote{
	Recall that the `exa'-prefix translates into a factor of \num{1e18}.
	For comparison, Nvidia's latest H100 GPU-based accelerator has a half-precision floating point (\texttt{bfloat16}) performance of \num{1} petaflops. 
}

\subsubsection{Keras}
\emph{Keras}~\cite{keras.2015} plays a special role among the software frameworks discussed here.
As a matter of fact, it is not a full-featured DL-library, much rather Keras can be considered as an interface to other libraries providing a high-level API, which was originally built for various backends including TensorFlow, Theano and the (now deprecated) Microsoft Cognitive Toolkit (CNTK).
As of version 2.4, TensorFlow is the only supported framework.
Keras, which is free and also open-source, is meant to further simplify experimentation with neural networks as compared to TensorFlow's lower level API.

\subsection{PyTorch}
PyTorch~\cite{pytorch.2019}, which is a free and open-source library, which was originally released to public in January 2017.\footnote{
	See this blog post on the history of PyTorch~\cite{Soumith.2022} and the YouTube talk of Yann LeCun, PyTorch co-creator Sousmith Chintala, Meta's PyTorch lead Lin Qiao and Meta's CTO Mike Schroepfer~\cite{PyTorch.YT.2022}.
}
As of 2022, PyTorch has evolved from a research-oriented DL-framework to a fully fledged environment for both scientific work and industrial applications, which, as of 2022 has caught up with, if not surpassed, TensorFlow in popularity.
Primarily addressing researchers in its early days, PyTorch saw a rapid growth not least for the---at that time---unique feature of \emph{dynamic} computational graphs, which allows for great flexibility and simplifies creation of complex network architectures. 
As opposed to its competitors as, e.g., TensorFlow, computational graphs, which represent compositions of mathematical operations and allow for automatic differentiation of complex expressions, are created on the fly, i.e., at the very same time as operations are performed.
\emph{Static} graphs, on the other hand, need to be created in a first step, before they can be evaluated and automatically differentiated.  Some examples of applying PyTorch to computational mechanics are provided in the next two remarks.

\begin{figure}[h]
	\centering
	\includegraphics[width=0.5\textwidth]{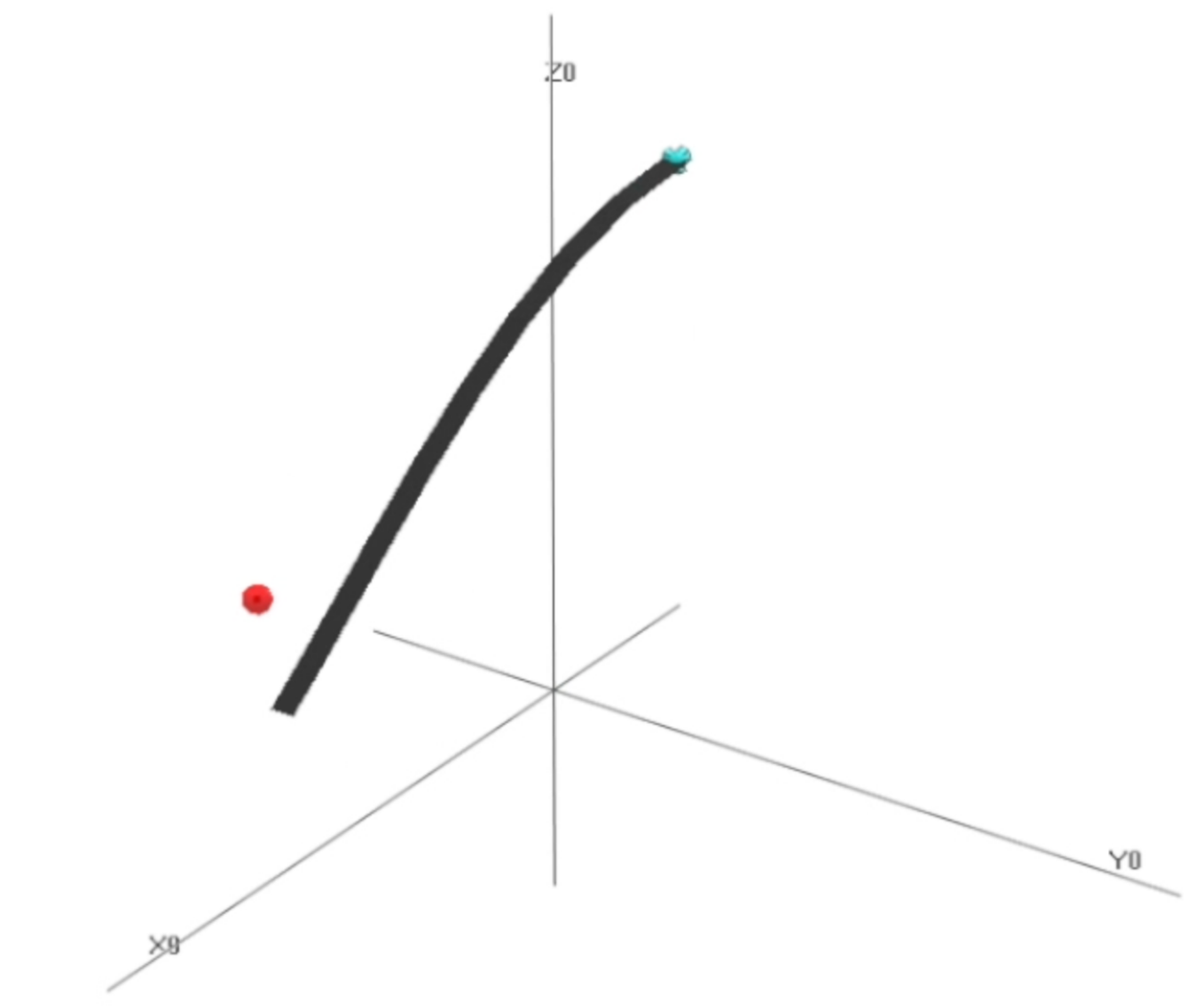}
	\caption{
		%			{\it Reinforcement learning to control of a large-deformable beam} 
		{\it Positioning and pointing control of large deformable beam}
		(Section~\ref{sc:libraries}, Remark~\ref{rm:reinforcement-learning}). 
		Reinforcement learning.
		The agent is trained to align the tip of the flexible beam with the target position (red ball).
		For this purpose, the agent can move the base of the cantilever; the environment returns the negative Euclidean distance of the beam's tip to the target position as ``reward'' in each time-step of the simulation. 
		\href{https://www.techscience.com/uploads/video/cmes/2022/reinforcement\%20learning.mp4}{Simulation video}.
		See also \href{https://gitlab.com/alexander.humer/cmes-dl-review}{GitLab} repository for code and video.
	}
	\label{fig:humer-rl-flexible-beam}
\end{figure}

\begin{rem}
	\label{rm:reinforcement-learning}
	{Reinforcement Learning} (RL) 
	{\rm is a branch of machine-learning, in which computational methods and DL-methods naturally come together. 
	Owing to the progress in DL, reinforcement learning, which has its roots in the early days of cybernetics and machine learning, see, e.g., the survey~\cite{Kaelbling1996}, has gained attention in the fields of automatic control and robotics again. 
	In their opening to a more recent review, the authors of~\cite{Arulkumaran2017} expect no less than \emph{``deep reinforcement-learning is poised to revolutionize artificial intelligence and represents a step toward building autonomous systems with a higher level understanding of the visual world.''}
	RL is based on the concept that an autonomous agent learns complex tasks by trial-and-error.
	Interacting with its environment, the agent receives a reward if it succeeds in solving a given tasks.
	Not least to speed up training by means of the parallelization, simulation has become are key ingredient to modern RL, where agents are typically trained in virtual environments, i.e., simulation models of the physical world.
	Though (computer) games are classical benchmarks, in which DeepMind's AlphaGo and  AlphaZero models excelled humans (see Section~\ref{sc:open}, Figure~\ref{fig:protein-folding-Go-game}), deep RL has proven capable of dealing with real-world applications in the field of control and robotics, see, e.g.,~\cite{Suenderhauf2018}.
	Based on the PyTorch's introductory tutorial (see \href{https://pytorch.org/tutorials/intermediate/reinforcement_q_learning.html}{Original website}) of a the classic cart-pole problem, i.e., an inverted pendulum (pole) mounted to a moving base (cart), we developed a RL-model for the control of large-deformable beams, see Figure \ref{fig:humer-rl-flexible-beam} and the \href{https://gitlab.com/alexander.humer/cmes-dl-review/-/blob/main/rl-flexible-beam/rl-flexible-beam.mp4}{video illustrating the training progress}.
	For some large-deformable beam formulations, 
	see, e.g., \cite{simo1988dynamics}, \cite{humer2013dynamic},
	 \cite{steinbrecher2017numerical}, \cite{humer2020b}.
%	{\rm 
%		{\color{red} NOTE: 2022.10.28 - Blah blah blah.... Link the GitLab repository here.  Need some figures. ENDNOTE}
%		See, e.g., \cite{simo1988dynamics} and \cite{steinbrecher2017numerical} for the formulation of geometrically-exact large-deformable beam.
%	}
	}
	$\hfill\blacksquare$
\end{rem}

\subsection{JAX}
JAX~\cite{jax.2018} is a free and open-source research project driven by Google. Released to the public in 2018, JAX is one of the more recent software frameworks that have emerged during the current wave of AI.
It is being described as being ``Autograd and XLA'' and as ``a language for expressing and composing transformations of numerical programs'', i.e., JAX focuses on accelerating evaluations of algebraic expression and, in particular, gradient computations. 
As a matter of fact, its core API, which provides a mostly NumPy-compatible interface to many mathematical operations, is rather trimmed-down in terms of DL-specific functions as compared to the broad scope of functionality offered by TensorFlow and PyTorch, for instance, which, for this reason, are often referred to as \emph{end-to-end} frameworks.
JAX, on the other hand, considers itself as system that facilitates ``transformations'' like gradient computation, just-in-time compilation and automatic vectorization of compositions of functions on parallel hardware as GPUs and TPUs.
A higher-level interface to JAX' functionality, which is specifically made for ML-purposes, is available through the \emph{FLAX} framework~\cite{flax.2020}. FLAX rovides many fundamental building blocks essential for creation and training of neural networks.

\subsection{Leveraging DL-frameworks for scientific computing}
\label{sc:dl-frameworks-fem}
Software frameworks for deep-learning as, e.g., PyTorch, TensorFlow and JAX share several features which are also essential in scientific computing, in general, and finite-element analysis, in particular.
These DL-frameworks are highly optimized in terms of vectorization and parallelization of algebraic operations. 
Within finite-element methods, parallel evaluations can be exploited in several respects: 
First and foremost, residual vectors and (tangent) stiffness matrices need to be repeatedly evaluated for all  elements of finite-element mesh, into which the domain of interest is discretized.
Secondly, the computation of each of these vectors and matrices is based upon numerical quadrature (see Section~\ref{sc:quadrature-weight-correction-modeling} for a DL-based approach to improve quadrature), which, from an algorithmic point of view, is computed as a weighted sum of integrands evaluated at a finite set of points.
A further key component that proves advantages in complex FE-problems is automatic differentiation, which, in the form of backpropagation (i.e., reverse-mode automatic differentiation), is the backbone of gradient-based training of neural networks, see Section~\ref{sc:backprop}.
In the context of FE-problems in solid mechanics, automatic differentiation saves us from deriving and implementing derivatives of potentials, whose variation and linearization with respect to (generalized) coordinates give force vectors and tangent-stiffness matrices.\footnote{
	GPU-computing and automatic differentiation are by no means new to scientific computing not least in the field of computational mechanics.
	\emph{Project Chrono} (see \href{https://www.projectchrono.org}{Original website}), for instance, is well known for its support of GPU-computing in problems of flexible multi-body and particle systems. 
	The general purpose finite-element code \emph{Netgen/NGSolve}~\cite{ngsolve.2014} (see \href{https://www.ngsolve.org}{Original website}) offers a great degree of flexibility owing to its automatic differentiation capabilities.
	Well-established commercial codes, on the other hand, are often built on a comparatively old codebase, which dates back to times before the advent of GPU-computing.
}

\begin{figure}[h]
	\centering
	\includegraphics[width=0.8\textwidth]{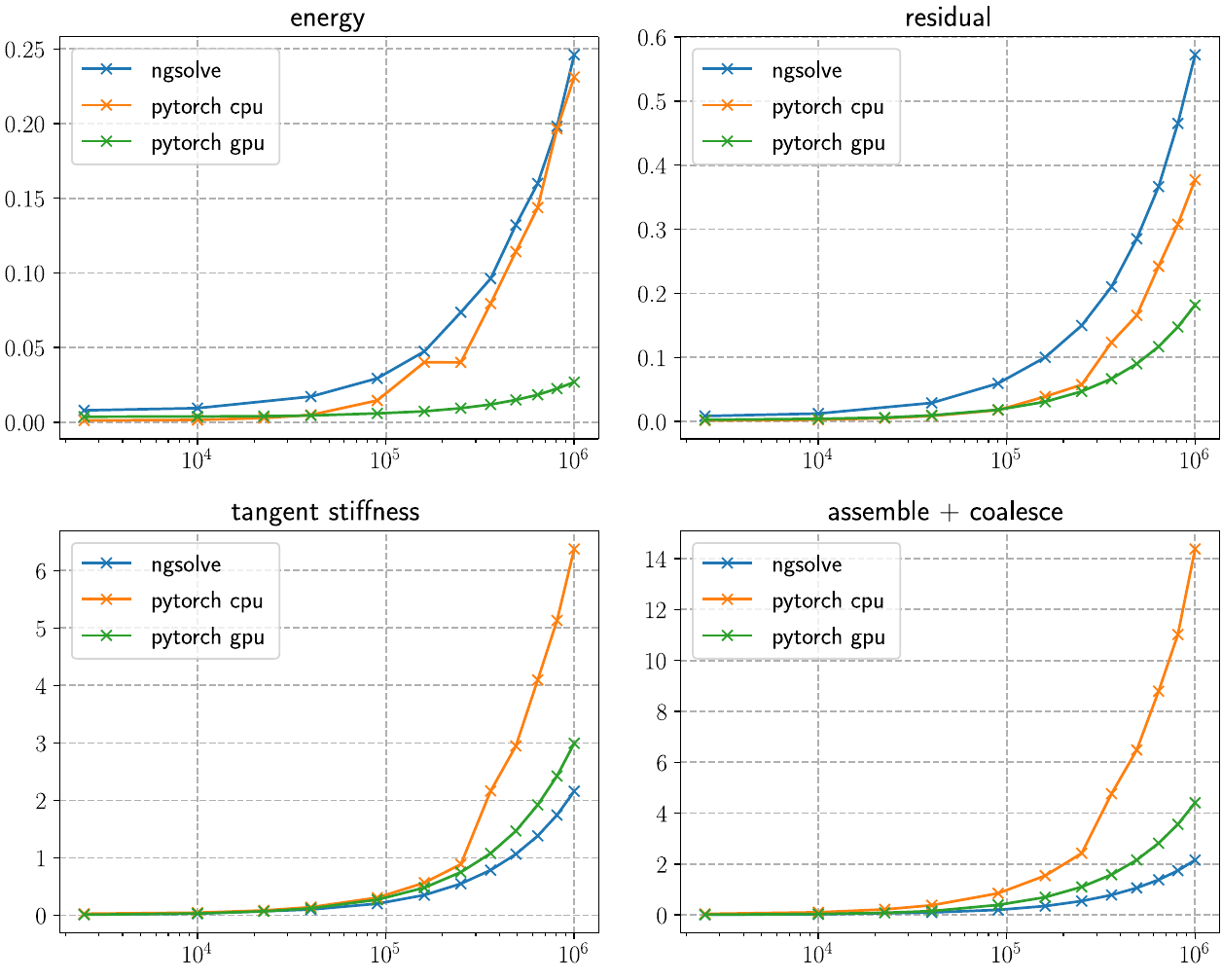}
	\caption{
		\emph{DL-frameworks in nonlinear finite-element problems} (Section~\ref{sc:dl-frameworks-fem}).
		The computational efficiency of a PyTorch-based (Version 1.8) finite-element code implemented was compared against the state-of-the-art general purpose Netgen/NGSolve~\cite{ngsolve.2014} for a problem of nonlinear elasticity, see the \href{https://gitlab.com/alexander.humer/cmes-dl-review/-/blob/main/asme_idetc_msndc_2021/presentation_idetc.pdf}{slides of the presentation} and the corresponding \href{https://gitlab.com/alexander.humer/cmes-dl-review/-/blob/main/asme_idetc_msndc_2021/humer_idetc.mp4}{video}.
		The figures show timings (in seconds) for evaluations of the strain energy (top left), the internal forces (residual, top right) and element-stiffness matrices (bottom left) and the global stiffness matrix (bottom right) against the number of elements.
		Owing to PyTorch's parallel computation capacity, the simple Python implementation could compete with the highly-optimized finite-element code, in particular, as computations were moved to a GPU (NVIDIA Tesla V100). 
	}
	\label{fig:timings-pytorch-ngsolve}
\end{figure}

The potential of modern DL-software frameworks in conventional finite-element problems was studied in \cite{humer.2021}, where Netgen/NGSolve, which is a highly-optimized, OpenMP-parallel finite-element code written in C++, was compared against PyTorch and JAX implementations.
In particular, the computational efficiency of the computing and assembling vectors of internal forces and tangent stiffness matrices of a hyperelastic solid was investigated.
On the same (virtual) machine, it turned out that both PyTorch and JAX can compete with Netgen/NGSolve when computations are performed on CPU, see the timings shown in Figure \ref{fig:timings-pytorch-ngsolve}.
Moving computations to a GPU, the Python-based DL-frameworks outperformed Netgen/NGSolve in the evaluation of residual vectors. 
Regarding tangent-stiffness matrices, which are obtained through (automatic) second derivatives of the strain-energy function with respect to nodal coordinates, both PyTorch and JAX showed (different) bottlenecks, which, however, are likely to be sorted out in future releases. 

\subsection{Physics-Informed Neural Network (PINN) frameworks}
\label{sc:DeepXDE-PINN}
\label{sc:PINN-frameworks}

In laying out the roadmap for ``Simulation Intelligence'' (SI) the authors of \cite{lavin2021simulation} considered PINN as a key player in the first of the nine SI ``motifs,'' called ``Multi-physics \& multi-scale modeling.''

%{\color{red} [NOTE: 2022.10.02 - Adding this section on PINN frameworks, which include DeepXDE \cite{lu2021deepxde}.  PINN is an important topic; see \cite{lavin2021simulation}.
%2021.10.28 - add and comment on the reference below in Section~\ref{sc:libraries} ``Deep-learning libraries, frameworks, platforms'': 
%		%	OR at the end in Section~\ref{sc:resurgence-2} ``Resurgence of AI and current state'' : 
%		\begin{itemize}
%				
%				\item Lu et al. 2021,   DeepXDE: A deep learning library for solving differential equations. SIAM Review, 63(1), 208-228 \cite{lu2021deepxde}, and
%				
%				%	\item Higham 2019, Deep learning: An introduction for applied mathematicians. SIAM Review, 61(4), 860-891 \cite{higham2019deep}.   DONE in Footnote~\ref{fn:review-papers}.
%				
%			\end{itemize}
%In particular, point out the differences between our review paper and the above reference.
%ENDNOTE]}

\begin{figure}[h]
	\centering
	\includegraphics[width=0.95\textwidth]{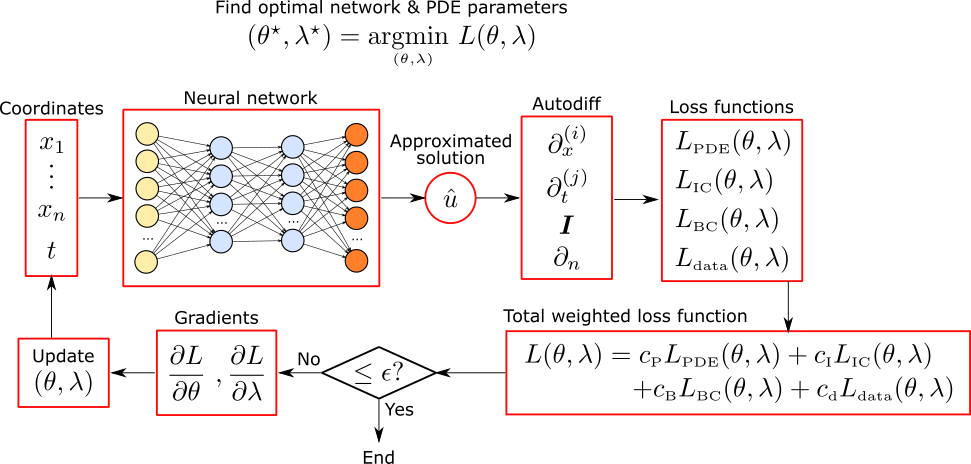}
	\caption{
		{\it Physics-Informed Neural Networks (PINN) concept} (Section~\ref{sc:DeepXDE-PINN}). The goal is to find the optimal network parameters $\theta^\star$ (weights) and PDE parameters $\lambda^\star$ that minimize the total weighted loss function $L(\theta , \lambda)$, which is a linear combination of four loss functions: (1) The residual of the PDE, $L_{\scriptscriptstyle \text{PDE}}$, (2) Loss due to initial conditions, $L_{\scriptscriptstyle \text{IC}}$, (3) Loss due to boundary conditions , $L_{\scriptscriptstyle \text{BC}}$, (4) Loss due to known (labeled) data, $L_{\scriptscriptstyle \text{data}}$, with $c_{\scriptscriptstyle \text{P}}, c_{\scriptscriptstyle \text{I}}, c_{\scriptscriptstyle \text{B}}, c_{\scriptscriptstyle \text{d}}$ being the combination coefficients. 
		With the space-time coordinates $(x_1, \ldots, x_n, t)$ as inputs, the neural network produces an approximated multi-physics solution $\hat u = \{u, v, p, \phi \}$, of which the derivatives, estimated by automatic differentiation, are used to evaluate the loss functions.  If the total loss $L$ is not less than a tolerance, its gradients with respect to the parameters $(\theta, \lambda)$ are used to update these parameters in a descent direction toward a local minimum of $L$ 
		\cite{cai2021fluid}.
	}
	\label{fig:PINN-concept}
\end{figure}

The PINN method to solve differential equations (ODEs, PDEs) aims at training a neural network to minimize the total weighted loss $L$ in
Figure~\ref{fig:PINN-concept}, which describes the PINN concept in more technical details.  As an example, the PDE residual $L_{\scriptscriptstyle \text{PDE}}$ for incompressible flow can be written as follows \cite{cai2021fluid}: 
\begin{align}
	&
	L_{\scriptscriptstyle \text{PDE}} = 
	\frac{1}{N_f} \sum_{i=1}^{N_f} \sum_{j=1}^{4} \left[ r_j ( \bkarsp{x}{f}{(i)}  , \karsp{t}{f}{(i)} ) \right]^2 \ , 
	\text{ with }
	\\
	&
	\bkar{r} = [r_1 , r_2 , r_3 ]^T = \frac{\partial \bkar{u}}{\partial t} + \left( \bkar{u} \dotprod \nabla \right) \bkar{u} + \nabla p - \frac{1}{Re} \nabla^2 \bkar{u}
	\in \real^{3 \times 1} \ ,  \text{ and }
	\label{eq:incompressible-flow-1}
%	\\
%	&
	r_4 = \nabla \dotprod \bkar{u} \ ,
%	\label{eq:incompressible-flow-2}
\end{align}
where $N_f$ is the number of residual (collocation) points, which could be in the millions generated randomly,\footnote{
	For incompressible flow past a cylinder, the computational domain  of dimension $[-7.5, 28.5] \times [-20, 20] \times [0, 12.5]$---with coordinates $(x, y, z)$ non-dimensionalized by the diameter of the cylinder, with axis along the $z$ direction, and going through the point $(x,y) = (0,0)$---contained $3 \times 10^6$ residual (collocation) points \cite{cai2021fluid}.
} the residual $\bkar{r} = [r_1 , r_2 , r_3 ]^T$ in Eq.~\eqref{eq:incompressible-flow-1}$_1$ is the left-hand side of the balance of linear momentum (the right-hand side being zero), and the residual $r_4$ in Eq.~\eqref{eq:incompressible-flow-1}$_2$ is the left-hand side of the incomrepssibility constraint, and $(\bkarsp{x}{f}{(i)}  , \karsp{t}{f}{(i)})$ the space-time coordinates of the collocation point $(i)$.

%\cite{cai2021fluid} : Review of PINN applied to fluid mechanics

Some review papers on PINN are 
\cite{cuomo2022scientific} 
\cite{karniadakis2021physics}, with the latter being more general than \cite{cai2021fluid}, which was restricted to fluid mechanics, and touching on many different fields.
Table~\ref{tb:PINN-frameworks} lists PINN frameworks that are currently actively developed, and a few selected \emph{solvers} among which are summarized below.

%\cite{raissi2017physics-1} : Original Part 1 of PINN paper on arXiv.
%\cite{raissi2017physics-2} : Original Part 2 of PINN paper on arXiv.

\begin{table}[h]
	\centering
	\caption{
		\emph{PINN frameworks} (Section~\ref{sc:PINN-frameworks}) being actively developed \cite{karniadakis2021physics} \cite{cuomo2022scientific}. 
%		In a \emph{solver}, users only need to define the problem, which is solved by the software.
		A \emph{solver} solves the problem defined by users.
		A \emph{wrapper} does not solve, but only wraps low-level functions from other libraries (e.g., PyTorch) into high-level functions that are convenient for users to implement PINN to solve the problem.
	}
	\begin{tabularx}{0.92\textwidth}{l *{4}{Y}}
		\toprule[3pt]
		Framework site \& Ref.		
		& 
		\multicolumn{1}{l}{Usage} 
		& 
		\multicolumn{1}{l}{Language}
		& 
		\multicolumn{1}{l}{Backend}      
		\\
		\cmidrule(lr){1-1} \cmidrule(lr){2-2} \cmidrule(lr){3-3} \cmidrule(lr){4-4}
		\href{https://deepxde.readthedocs.io/en/latest/}{DeepXDE} \cite{lu2021deepxde} 		
		& 
		\multicolumn{1}{l}{Solver} 
		& 
		\multicolumn{1}{l}{Python}
		& 
		\multicolumn{1}{l}{TensorFlow, PyTorch, JAX, Paddle}  
		\\
		\href{https://developer.nvidia.com/modulus}{NVIDIA Modulus (SimNet)} \cite{hennigh2020nvidia} 		
		& 
		\multicolumn{1}{l}{Solver} 
		& 
		\multicolumn{1}{l}{Python}
		& 
		\multicolumn{1}{l}{TensorFlow}
		\\
		\href{https://github.com/analysiscenter/pydens}{PyDEns} \cite{koryagin2019pydens} 		
		& 
		\multicolumn{1}{l}{Solver} 
		& 
		\multicolumn{1}{l}{Python}
		& 
		\multicolumn{1}{l}{TensorFlow}
		\\
		\href{https://github.com/NeuroDiffGym/neurodiffeq}{NeuroDiffEq} \cite{chen2020neurodiffeq} 		
		& 
		\multicolumn{1}{l}{Solver} 
		& 
		\multicolumn{1}{l}{Python}
		& 
		\multicolumn{1}{l}{PyTorch}
		\\
		\href{https://neuralpde.sciml.ai/dev/}{NeuralPDE} \cite{rackauckas2017differentialequations} 		
		& 
		\multicolumn{1}{l}{Solver} 
		& 
		\multicolumn{1}{l}{Julia}
		& 
		\multicolumn{1}{l}{Julia}
		\\
		\href{https://www.sciann.com/}{SciANN} \cite{haghighat2021sciann} 		
		& 
		\multicolumn{1}{l}{Wrapper} 
		& 
		\multicolumn{1}{l}{Python}
		& 
		\multicolumn{1}{l}{TensorFlow}
		\\
		\href{https://kailaix.github.io/ADCME.jl/latest/}{ADCME} \cite{xu2020adcme} 		
		& 
		\multicolumn{1}{l}{Wrapper} 
		& 
		\multicolumn{1}{l}{Julia}
		& 
		\multicolumn{1}{l}{TensorFlow}
		\\
		\href{https://gpytorch.ai/}{GPytorch} \cite{gardner2018gpytorch} 		
		& 
		\multicolumn{1}{l}{Wrapper} 
		& 
		\multicolumn{1}{l}{Python}
		& 
		\multicolumn{1}{l}{PyTorch}
		\\
		\href{https://github.com/google/neural-tangents}{Neural Tangents} \cite{novak2020fast} 		
		& 
		\multicolumn{1}{l}{Wrapper} 
		& 
		\multicolumn{1}{l}{Python}
		& 
		\multicolumn{1}{l}{JAX}
		\\
		\midrule[2pt]
		%		\\
	\end{tabularx}
	\label{tb:PINN-frameworks}
\end{table}

%\cite{lu2021deepxde} : DeepXDE - ``We also present a Python library for PINNs, DeepXDE, which is designed to serve both as an educational tool to be used in the classroom as well as a research tool for solving problems in computational science and engineering. Specifically, DeepXDE can solve forward problems given initial and boundary conditions, as well as inverse problems given some extra measurements. DeepXDE supports complex-geometry domains based on the technique of constructive solid geometry and enables the user code to be compact, resembling closely the mathematical formulation. We introduce the usage of DeepXDE and its customizability, and we also demonstrate the capability of PINNs and the user-friendliness of DeepXDE for five different examples. More broadly, DeepXDE contributes to the more rapid development of the emerging scientific machine learning field.''

\noindent
\ding{42}
\href{https://deepxde.readthedocs.io/en/latest/}{DeepXDE} \cite{lu2021deepxde}, one of the first PINN framework and a \emph{solver} (Table~\ref{tb:PINN-frameworks}) was developed in Python, with a TensorFlow backend, for both teaching and research. This framework can solve both forward problems (``given initial and boundary conditions'') and inverse problems (`` given some extra measures''), with domains having complex geometry.  According to the authors, DeepXDE is user-friendly, with compact user code resembling the problem mathematical formulation, customizable to different types of mechanics problem.  The site contains many published papers with a large number of demo problems: Poisson equation, Burgers equation, diffusion-reaction equation, wave propagation equation, fractional PDEs, etc.  In addition, there are demos on inverse problems and operator learning.  Three more backends beyond TensorFlow, which was reported in \cite{karniadakis2021physics}, have been added to DeepXDE: PyTorch, JAX, Paddle.\footnote{
	See Section \href{https://deepxde.readthedocs.io/en/latest/user/installation.html\#working-with-different-backends}{Working with different backends}.  See also \cite{cuomo2022scientific}.
}

\noindent
\ding{42}
\href{https://github.com/NeuroDiffGym/neurodiffeq}{NeuroDiffEq} \cite{chen2020neurodiffeq}, a \emph{solver}, was developed about the same time as DeepXDE, with the backend being PyTorch.  Even though it was written that the authors were ``actively working on extending NeuroDiffEq to support three spatial dimensions,'' this feature is not ready, and can be worked around by including the 3D boundary conditions in the loss function.\footnote{
	``All you need is to import {\tt GenericSolver} from {\tt neurodiffeq.solvers}, and {\tt Generator3D} from {\tt neurodiffeq.generators}. The catch is that currently there is no reparametrization defined in {\tt neurodiffeq.conditions} to satisfy 3D boundary conditions,'' which can be hacked into the loss function ``by either adding another element in your equation system or overwriting the {\tt additional\_loss} method of {\tt GenericSolve}.''
	Private communication with a developer of NeuroDiffEq on 2022.10.08.
}  Even though in principle, NeuroDiffEq can be used to solve PDEs of interest to engineering (e.g., Navier-Stokes solutions), there were no such examples in the official documentation, except for a 2D Laplace equation and a 1D heat equation.  The backend is limited to PyTorch, and the site did not list any papers, either by the developers or by others, using this framework.

\begin{figure}[h]
	\centering
	\includegraphics[width=0.44\textwidth]{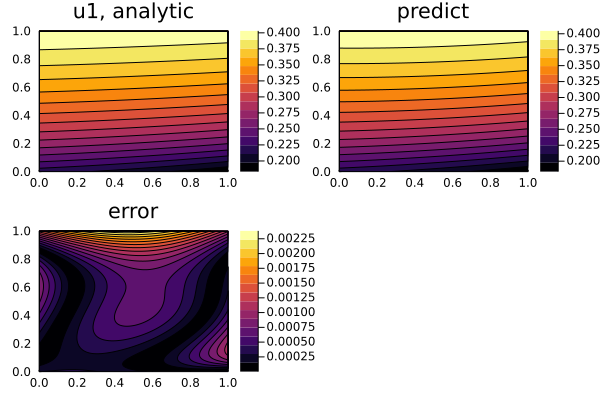}
	\includegraphics[width=0.44\textwidth]{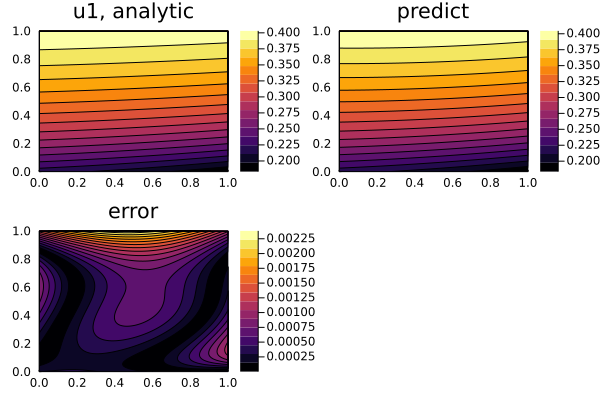}
	\caption{
		{Coupled nonlinear hyperbolic equations}
%		\href{https://neuralpde.sciml.ai/dev/}{\it NeuralPDE}
%		\textit{NeuralPDE} 
		(Section~\ref{sc:PINN-frameworks}). 
		Analytical solution, predicted solution by \href{https://neuralpde.sciml.ai/dev/}{NeuralPDE} \cite{rackauckas2017differentialequations} and error for the coupled nonlinear hyperbolic equations in Eq.~\eqref{eq:NeuralPDE-hyperbolic}.
	}
	\label{fig:NeuralPDE-predict-error}
\end{figure}

\noindent
\ding{42}
\href{https://neuralpde.sciml.ai/dev/}{NeuralPDE} \cite{rackauckas2017differentialequations}, a \emph{solver},
was developed in a relatively new language \href{https://julialang.org/}{Julia}, which is 20 years younger than Python, has a speed edge over Python in machine learning, but does not in data science.\footnote{
	There are many comparisons of Julia versus Python on the web; one is \href{https://blog.boot.dev/python/python-vs-julia/}{Julia vs Python: Which is Best to Learn First?}, by By Zulie Rane on 2022.02.05, updated on 2022.10.01.
}  Demos are given for ODEs, generic PDEs, such as the coupled nonlinear hyperbolic PDEs of the form:
\begin{align}
	&
	\frac{\partial^2 u}{(\partial t)^2} = \frac{a}{n} \frac{\partial}{\partial x} (x^n \frac{\partial u}{\partial x}) + u f (\frac{u}{w}) \ , \quad
%	\\
%	&
	\frac{\partial^2 w}{(\partial t)^2} = \frac{b}{n} \frac{\partial}{\partial x} (x^n \frac{\partial u}{\partial x}) + w f (\frac{u}{w}) \ ,
	\label{eq:NeuralPDE-hyperbolic}
\end{align}
with $f$ and $g$ being arbitrary functions. There are initial and boundary conditions, and exact solution to find the error of the numerical solution, Figure~\ref{fig:NeuralPDE-predict-error}.  The site did not list any papers, either by the developers or by others, using this framework.

Additional PINN software packages other than those in Table~\ref{tb:PINN-frameworks} are listed and summarized in \cite{cuomo2022scientific}.

\begin{rem}
	\label{rm:PINN}
	PINN and activation functions.
	{\rm 
		Deep neural networks (DNN), having at least two hidden layers, with ReLU activation function (Figure~\ref{fig:ReLU}) were shown to correspond to linear finite element interpolation \cite{he2020relu}, since piecewise linear functions can be written as DNN with ReLU activation functions \cite{arora2016understanding}.
		
		But using the \emph{strong} form such as the PDE in Eq.~\eqref{eq:NeuralPDE-hyperbolic}, which has the second partial derivative with respect to $x$, and since the second derivative of ReLU is zero, activation functions such as the logistic sigmoid (Figure~\ref{fig:sigmoid}), hyperbolic tangent (Figure~\ref{fig:tanh}), or the swish function (Figure~\ref{fig:swish}) are recommended.  Because of the presence of the second partial derivatives with respect to the spatial coordinates  in the general PDE Eq.~\eqref{eq:general-PDE}, particularized to the 2D Navier-Stokes Eq.~\eqref{eq:navier-stokes-2D}:
 		\begin{align}
			&
			\bkars{u}{t} + {\mathcal N} [\bkar{u} ; \bkar{\lambda}] = \bkar{0} \, \text{ for } \bkar{x} \in \Omega \, t \in [0, T] \ , \text{ with}
			\label{eq:general-PDE}
			\\
			&
			\bkars{u}{t} = 
			\begin{Bmatrix}
				u_t
				\\
				v_t
			\end{Bmatrix}  
			\ , \quad 
			{\mathcal N} [\bkar{u} ; \bkar{\lambda}] =  
			\begin{Bmatrix}
				\lambda_1 (u u_x + v u_y) + p_x - \lambda_2 (u_{xx} + u_{yy})
				\\
				\lambda_1 (u v_x + v v_y) + p_y - \lambda_2 (v_{xx} + v_{yy})
			\end{Bmatrix}
			\ , \quad
			\bkar{\lambda} = 
			\begin{Bmatrix}
				\lambda_1
				\\
				\lambda_2
			\end{Bmatrix} \ ,
			\label{eq:navier-stokes-2D}
		\end{align}
		for which the hyperbolic tangent ($\tanh$) was used as activation function \cite{raissi2019physics}.
	}
	$\hfill\blacksquare$
\end{rem}

%\cite{hennigh2020nvidia} : NVIDIA Modulus (SimNet)

\begin{rem}
	\label{rm:VPINN}
	Variational PINN.
	{\rm
		Similar to the finite element method, in which the \emph{weak} form, not the strong form as in Remark~\ref{rm:PINN}, of the PDE allows for a reduction in the requirement of differentiability of the trial solution, and is discretized with numerical integration used to evaluate the resulting coefficients for various matrices (e.g, mass, stiffness, etc), PINN can be formulated using the \emph{weak} form, instead of the strong form such as Eq.~\eqref{eq:navier-stokes-2D}, at the expense of having to perform numerical integration (quadrature) \cite{kharazmi2019variational} \cite{kharazmi2021hp} \cite{berrone2022variational}.   
		
		Examples of 1-D PDEs were given in \cite{kharazmi2019variational} in which the activation function was a sine function defined over the interval $(-1,1)$.
		To illustrate the concept, consider the following simple 1-D problem \cite{kharazmi2019variational} (which could model the axial displacement $u(x)$ of an elastic bar under distributed load $f(x)$, with prescribed end displacements):
		\begin{align}
%			&
			- u^{\prime\prime} (x) = - \frac{d^2 u(x)}{(dx)^2} = f(x), \text{ for } x \in (-1, 1) \ , \quad u(-1) = g \ , \quad u(1) = h \ ,
%			\\
%			&
			\label{eq:elastic-bar}
		\end{align}
	
		\noindent
		where $g$ and $h$ are constants.  Three variational forms of the strong form Eq.~\eqref{eq:elastic-bar} are:
		
		\begin{align}
			&
			\textsc{a}_k (u,v )= \textsc{b} (f,v) := \int_{-1}^{+1} f(x) v(x) dx \ , \forall v \text{ with } v(-1) = v(+1) = 0
			\ , \text{ and }
			\label{eq:variational-0}
			\\
			&
			\textsc{a}_1 (u,v) := - \int_{-1}^{+1} u^{\prime\prime} (x) v (x) dx 
			\ ,
			\label{eq:variational-1}
			\\
			&
			\textsc{a}_2 (u,v) := \int_{-1}^{+1} u^{\prime} (x) v^{\prime} (x) dx
			\ ,
			\label{eq:variational-2}
			\\
			&
			\textsc{a}_3 (u,v) := - \int_{-1}^{+1} u (x) v^{\prime\prime} (x) dx + \left. u(x) v^\prime (x) \right|_{x=-1}^{x=+1}
			\ ,
			\label{eq:variational-3}
		\end{align}
		where the familiar symmetric operator $\textsc{a}_2$ in Eq.~\eqref{eq:variational-2} is the weak form, with the non-symmetric operator $\textsc{a}_1$ in Eq.~\eqref{eq:variational-1} retaining the second derivation on the solution $u(x)$, and the non-symmetric operator $\textsc{a}_3$  in Eq.~\eqref{eq:variational-3} retaining the second derivative on the test function $v(x)$ in addition to the boundary terms.
		Upon replacing the solution $u(x)$ by its neural network (NN) approximation $\kars{u}{NN} (x)$ obtained using one single hidden layer ($L=1$ in Figure~\ref{fig:network3b}) with $y = \karp{f}{(1)} (x)$ and layer width $\kars{m}{(1)}$, and using the sine activation function on the interval $(-1, +1)$:
		\begin{align}
			\kars{u}{NN} (x) = y(x) = \karp{f}{(1)} (x) 
			= \sum_{i=1}^{\kars{m}{(1)}}  \kars{u}{NN_i} (x)
			= \sum_{i=1}^{\kars{m}{(1)}}  c_i \sin (w_i x + b_i)
			\ ,
			\label{eq:neural-approximation}
		\end{align}
		which does not satisfy the essential boundary conditions (whereas the solution $u(x)$ does), the loss function for the VPINN method is then the squared residual of the variational form plus the squared residual of the essential boundary conditions:
		\begin{align}
			&
			L(\theta) = \left[ \textsc{a}_k (\kars{u}{NN}, v) - \textsc{b} (f,v) \right]^2 
			+ 
			\left[ \kars{u}{NN} (x) - u (x) \right]_{x \in \{-1, +1\}}^2
			\ , \text{ and }
			\\
			&
			\theta^\star = \left\{ \bkar w^\star , \bkar b^\star , \bkar c^\star \right\} = \underset{\theta}{\text{argmin}} \  L(\theta)
			\ ,
			\label{eq:VPINN}
		\end{align}
		where $\theta^\star = \left\{ \bkar w^\star , \bkar b^\star , \bkar c^\star \right\}$ are the optimal parameters for Eq.~\eqref{eq:neural-approximation},
		with the goal of enforcing the variational form and essential boundary conditions in Eq.~\eqref{eq:variational-0}.
		More details are in \cite{kharazmi2019variational} \cite{kharazmi2021hp}. 
		
		For a symmetric variational form such as $\textsc{a}_2 (u , v)$ in Eq.~\eqref{eq:variational-2}, a potential energy exists, and can be minimized to obtain the neural approximate solution $u_{NN} (x)$:
		\begin{align}
			&
			J(u) = \frac12 \textsc{a}_2 (u , u) - \textsc{b} (f, u) 
			\ , \quad 
			\tilde J(\theta) = \frac12 \textsc{a}_2 (u_{NN} , u_{NN}) - \textsc{b} (f, u_{NN})
			\ ,
			\\
			&
			\theta^\star = \left\{ \bkar w^\star , \bkar b^\star , \bkar c^\star \right\} = \underset{\theta}{\text{argmin}} \  L(\theta)
			\ , \quad
			L(\theta) = \tilde J(\theta) + \left[ \kars{u}{NN} (x) - u (x) \right]_{x \in \{-1, +1\}}^2
			\ ,
		\end{align}
		which is similar to the approach taken in \cite{he2020relu}, where the ReLU activation function (Figure~\ref{fig:ReLU}) was used, and where a constraint on the NN parameters was used to satisfy an essential boundary condition.
	}
	$\hfill\blacksquare$
\end{rem}

\begin{rem}
	\label{rm:PINN-convergence-problem}
	PINN, kernel machines, training, convergence problems.
	{\rm 
		There is a relationship between PINN and kernel machines in Section~\ref{sc:kernel-machines}.  Specifically, the neural tangent kernel \cite{jacot2018neural}, which ``captures the behavior of fully-connected neural networks in the infinite width limit during training via gradient descent'' was used to understand when and why PINN failed to train \cite{wang2020when}, whose authors found a ``remarkable discrepancy in the convergence rate of the different loss components contributing to the total training error,'' and proposed a new gradient descent algorithm to fix the problem.  
		
		It was often reported that PINN optimization converged to ``solutions that lacked physical behaviors,'' and ``reduced-domain methods improved convergence behavior of PINNs''; see \cite{rohrhofer2022understanding}, where a dynamical system of the form below was studied:
		\begin{align}
			\frac{d u_{NN} (x)}{dx} = f(u_{NN} (x))
			\ , \quad
			L(\theta) = \frac1N \sum_{i=1}^{N} \left[ \frac{d u_{NN} (x^{(i)})}{dx} - f(u_{NN} (x^{(i)})) \right]^2
			\ ,
		\end{align}
		with $L(\theta)$ being called the ``physics'' loss function.  An example with $f(u(x)) = u(x) \left[ 1 - u^2 (x) \right]$ was studied to show that the ``physics loss optimization predominantly results in convergence issues, leading to incorrectly learned system dynamics''; see \cite{rohrhofer2022understanding}, where it was found that ``solutions corresponding to nonphysical system dynamics [could]
		be dominant in the physics loss landscape and optimization,'' and that ``reducing the computational
		domain [lowered] the optimization complexity and chance of getting trapped with nonphysical solutions.'' 
		
		See also \cite{erichson2019physics} for incorporating the Lyapunov stability concept into PINN formulation for CFD to ``improve the generalization error and reduce the prediction uncertainty.''
	}
	$\hfill\blacksquare$
\end{rem}

\begin{rem}
	\label{rm:PINN-attention}
	PINN and attention architecture.
	{\rm
		In \cite{rodriguez2022physics}, PIANN, Physics-Informed Attention-based Neural Network, was proposed to connect PINN to attention architecture, discussed in  Section~\ref{sc:Transformer} to solve hyperbolic PDE with shock wave.  See Remark~\ref{rm:attention-kernel-PINN} and Remark~\ref{rm:PINN-solid-mechanics}.
	}
	$\hfill\blacksquare$
\end{rem}

\begin{rem}
	\label{rm:PINN-patent}
	``Physics-Informed Learning Machine'' (PILM) 2021 US Patent 
	{\rm
		\cite{raissi2021physics}.
		First note that the patent title used the phrase ``learning machine,'' instead of ``machine learning,'' indicating that the emphasis of the patent appeared to be on ``machine,'' instead of on ``learning'' \cite{raissi2021physics}.  PINN was not mentioned, as it was first invented 
%		by Lagaris et al. 
		in
		\cite{lagaris1998artificial} \cite{lagaris2000neural}, which were cited by the patent authors in their original PINN paper \cite{raissi2019physics}.\footnote{
			The 2019 paper \cite{raissi2019physics} was a merger of a two-part preprint \cite{raissi2017physics-1} \cite{raissi2017physics-2}. 
		}
		The abstract of this 2021 PILM US Patent \cite{raissi2021physics} reads as follows:
		%For a succinct description of PINN understandable by general readers, the abstract of the 2021 US patent in \cite{raissi2021physics} gives the following:
		\begin{quote}
			``A method for analyzing an object includes modeling the object with a differential equation, such as a linear partial differential equation (PDE), and sampling data associated with the differential equation.  The method uses a probability distribution device to obtain the solution to the differential equation. The method eliminates use of discretization of the differential equation.''
		\end{quote} 
		The first sentence is nothing new to the readers.  In the second sentence, a ``probability distribution device'' could be replaced by a neural network, which would make PILM into PINN. This patent mainly focused on the Gaussian processes (Section~\ref{sc:Gaussian-process}), as an exemple of probability distribution (see Figure~4 in \cite{raissi2021physics}).  The third sentence would be the claim-to-fame of PILM, and also of PINN.
	}
	$\hfill\blacksquare$
\end{rem}

\begin{rem}
	Using PINN frameworks.
	{\rm
		While undergraduates with limited knowledge on the theory of the Finite Element Method could run FE Analysis of complicated structures and complex domain geometries on a laptop using commercial FE codes, solving problems with exceedingly simple domain geometry using a PINN framework such as DeepXDE does require knowledge of governing PDEs, initial and boundary conditions, artificial neural networks and frameworks (such as PyTorch, TensorFlow, etc.), the Python language, and having a more powerful computer.  In addition, because there are many parameters to fiddle with, beyond the sample problems posted on the DeepXDE website, first-time users could encounter disappointment and doubt when trying to solve a new problem. 
		It is not clear when the PINN methods would reach the level of FE commercial codes that undergraduates could use, or would they just fade away after an initial period of excitement like the meshless methods before them.
	}
	$\hfill\blacksquare$
\end{rem}

%\input{10-empty}
% Oishi 2017 paper
%{\color{red}
%[NOTE: avoid using ``data-driven computational mechanics'', which is the title of a paper based on ``distance minimization'' and variational method.]
%}

% \section{Application 1: Quadrature for finite elements}
\section{Application 1: Enhanced numerical quadrature for finite elements}
\label{sc:integration}
\label{sc:Oishi-2}
The results and deep-learning concepts used in \cite{Oishi.2017:rd9648} were presented in Section~\ref{sc:Oishi-summary} above. In this section, we discuss some details of the formulation.

The finite element method (FEM) has become the most important numerical method for the approximation of solutions to partial differential equations, in particular, the governing equations in solid mechanics. 
As for any mesh-based method, the discretization of a continuous (spatial, temporal) domain into a finite-element mesh, i.e., a disjoint set of finite elements, is a vital ingredient that affects the quality of the results and therefore has emerged as a field of research on its own.
%\footnote{
%see, e.g., \cite{Zienkiewicz.2013:rd0001}, Chapter ``8 Automatic mesh generation''.
%}
Being based on the weak formulation of the governing balance equations, numerical integration is a second key ingredient of FEM, in which integrals over the physical domain of interest are approximated by the sum of integrals over the individual finite elements.
In real-world problems, regularly shaped elements, e.g., triangles and rectangles in 2-D, tetrahedra and hexahedra in 3-D, typically no longer suffice to represent the complex shape of bodies or physical domains.
By distorting basic element shapes, finite elements of more arbitrary shapes are obtained, while the interpolation functions of the ``parent'' elements can be retained.\footnote{
	See, e.g., \cite{Zienkiewicz.2013:rd0001}, p.61, Section ``3.5 Isoparametric form'' and
	p.170, Section ``6.5 Mapping: Parametric forms.''
%	See, e.g., \cite{Zienkiewicz.2013:rd0001}, Chapter \sout{``5 Mapped elements and numerical integration – `infinite' and `singularity elements'{}''}.
%	{\color{red} [NOTE: 2020.05.14.  NO, ``infinite elements'' are in \cite{Zienkiewicz.2013:rd0001}, p.221, Section 7.6, and ``singular elements'' are in \cite{Zienkiewicz.2013:rd0001}, p.224, Section 7.7, whereas Chap 5 in \cite{Zienkiewicz.2013:rd0001}, p.115, does not have the above title, but ``Field Problems: A Multidimensional Finite Element Method''.   On the other hand, ``Numerical Integration'' is in \cite{Zienkiewicz.2013:rd0001}, Section 6.8, p.178.  We also need to give the page number to the section, as done here. Please check; are we using the same edition ? ENDNOTE]}
}
The mapping represents a coordinate transformation by which the coordinates of a {``parent'' or} ``reference'' element are mapped onto distorted, possibly curvilinear, physical coordinates of the actual elements in the mesh.
Conventional finite element formulations use polynomials as interpolation functions,
%\footnote{
%	\sout{{\color{blue} TODO: isoparametric concept}}
%  See, e.g., \cite{Zienkiewicz.2013:rd0001}, p.170, Section ``6.5 Mapping: Parametric forms''.
%}
for which Gauss-Legendre quadrature is the most efficient way to integrate numerically.
Efficiency in numerical quadrature is immediately related the (finite) number of integration points that is required to exactly integrate a polynomial of a given degree.\footnote{
Using Gauss-Legendre quadrature, $p$ integration points integrate polynomials up to a degree of $2p-1$ exactly.}
For distorted elements, the Jacobian of the transformation describing the distortion of the parent element renders the integrand of, e.g., the element stiffness matrix non-polynomial.
Therefore, the integrals generally are not integrated exactly using Gauss-Legendre quadrature, but the accuracy depends, roughly speaking, on the degree of distortion. 

\subsection{Two methods of quadrature, 1-D example}
\label{sc:Oishi-1D-example}
To motivate their approaches, 
%
% CMES style rewriting
the authors of 
\cite{Oishi.2017:rd9648} presented 
an illustrative 1-D example of a simple integral, which was analytically integrated as
\begin{equation} 
\int_{-1}^1 x^{10} \dx = \left[ \frac{1}{11} \, x^{11} \right]_{-1}^1 = \frac{2}{11} .
\end{equation}
Using 2 integration points, Gauss-Legendre quadrature yields significant error 
\begin{equation} \label{eq-oishi-example}
\int_{-1}^1 x^{10} \dx \approx 1 \left( - \frac{1}{\sqrt 3} \right)^{10} + 1 \left( \frac{1}{\sqrt 3} \right)^{10} = \frac{2}{243} ,
\end{equation}
which is owed to the insufficient number of integration points.

Method 1 to improve accuracy, which is reflected in Application 1.1 (see Section~\ref{sc:Oishi-1.1}), is to increase the number of integration points.
In the above example, 6 integration points are required to obtain the exact value of the integral.
By increasing the accuracy, however, we sacrifice computational efficiency due the need for 6 evaluations of the integrand instead of the original 2 evaluations.

Method 2 is to
retain 2 integration points, and to adjust the quadrature weights at the integration points instead. 
If the same quadrature weights of $243/11$ are used instead of $1$, the integral evaluates to the exact result:
\begin{equation}
\int_{-1}^1 x^{10} \dx = \frac{243}{11} \left( - \frac{1}{\sqrt 3} \right)^{10} + \frac{243}{11} \left( \frac{1}{\sqrt 3} \right)^{10} = \frac{2}{11} .
\end{equation}
By adjusting the quadrature weights rather than the number of integration points, which is the key concept of Application 1.2 (see Section~\ref{sc:Oishi-1.2}), the computational efficiency of the original approach Eq.~\eqref{eq-oishi-example} is retained.

\begin{figure}[h]
	\centering
	\includegraphics[width=0.24\textwidth]{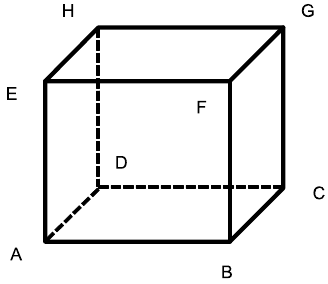}
	\includegraphics[width=0.74\textwidth]{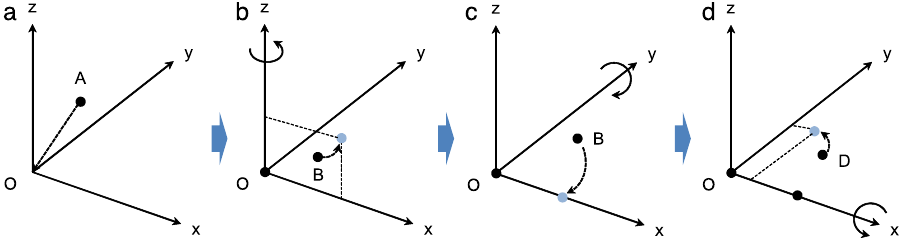}
	\caption{
		{\it Normalization procedure for hexahedra} (Section~\ref{sc:Oishi-1D-example}).
%		{\color{red} [NOTE: 2020.06.09. this figure is not referred to anywhere in the text, and should belong to Subsection~\ref{sc:Oishi-1.1}, not the beginning of Section~\ref{sc:Oishi-2}. ENDNOTE]}
		Numerical integration is performed in local element coordinates. The accuracy of Gauss-Legendre quadrature only depends on the element shape, i.e., it is invariant with respect to rigid-body motion and uniform stretching deformation. 
		For this reason, 
		%
		% CMES style rewriting
%		\cite{Oishi.2017:rd9648} introduced 
		a normalization procedure was introduced involving one translation and three rotations (second from left to right) for linear hexahedral elements, whose nodes are labelled as shown for the regular hexahedron (left) \cite{Oishi.2017:rd9648}.  
		(1) The element is displaced such that node $A$ coincides with the origin $O$ of the global frame. 
		(2) The hexahedron is rotated about the $z$-axis of the global frame to place node $B$ in the $xz$-plane, and then
		(3) rotated about the $y$-axis such that node $B$ lies on the $x$-axis. 
		(4) A third rotation about the $x$-axis relocates node $D$  to the $xy$-plane of the global frame. 
		(5) Finally (not shown), the element is scaled by a factor of $1/l_0$, where $l_0 = (AB + AD)/2$.
		 {\footnotesize (Figure reproduced with permission of the authors.)}
	}
	\label{fig:Oishi-normalization}

\end{figure}

%As a data basis, \cite{Oishi.2017:rd9648} generate \num{20000} arbitrarily distorted hexahedral elements by randomly displacing nodes of the reference element.
%
% CMES rewriting
%In their study,~\cite{Oishi.2017:rd9648} considered
In this study \cite{Oishi.2017:rd9648}, 
hexahedral elements with linear shape functions were considered. 
To exactly integrate the element stiffness matrix of an undistorted element $2 \times 2 \times 2 = 8$ integration points are required.\footnote{
	Conventional linear hexahedra are known to suffer from locking, see, e.g.,~\cite{Zienkiewicz.2013:rd0001} Section ``10.3.2 Locking,'' 
%	{\color{red} [NOTE: 2020.05.19. in \cite{Zienkiewicz.2013:rd0001}, p.291, the Section with the title ``Locking'' is Section 9.3.2, not 10.3.2. Section 10.3, p.317, discusses locking with the title ``Two-field incompressible elasticity (u-p form)''. ENDNOTE] } 
	which can be alleviated by ``reduced integration,'' i.e., using a single integration point ($1 \times 1 \times 1$).
	The concept 
	%
	% CMES style rewriting
%	of~
	in \cite{Oishi.2017:rd9648}, however, immediately translates to higher-order shape functions and non-conventional finite element formulations (e.g., mixed formulation). 
}
Both applications summarized subsequently required the integrand, i.e., the element shape to be identified in an  unique way. 
Gauss-Legendre quadrature is performed in the local coordinates of the reference element with accuracy invariant to both rigid-body motion and uniform stretching of the actual elements.
To account for these invariances,
%
% CMES style rewriting
%~\cite{Oishi.2017:rd9648} proposed
a ``normalization'' of elements was proposed in \cite{Oishi.2017:rd9648} (Figure~\ref{fig:Oishi-normalization}), i.e., hexahedra were re-located to the origin of a frame of reference, re-oriented along with the coordinate planes and scaled by means of the average length of two of its edges.

\begin{figure}[h]
	\centering
	\includegraphics[width=0.4\textwidth]{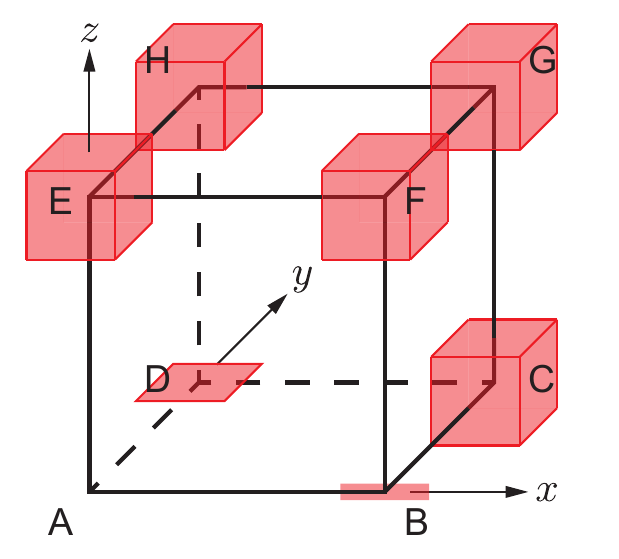}
	\caption{{\it Creation of randomly distorted elements}
		(Section~\ref{sc:integration}). Hexahedra forming the training and validation sets are created by randomly displacing the nodes of a regular hexahedral. To comply with the normalization procedure, node A remains fixed, node B is shifted along the $x$-axis and node C is displaced with the $xy$-plane. For each of the remaining nodes (E, F, G, H), all three nodal coordinates are varied randomly. 
		%
		% CMES style rewriting
%		\cite{Oishi.2017:rd9648} grouped 
		The elements are grouped according to the maximum \emph{possible} nodal displacement $d \in \left\{ 0.1, 0.2, 0.3, 0.4, 0.5 \right\}$, from which each of the 18 nodal displacements was obtained upon multiplication with a random number $r_i \in [0,1]$, $i=1,\ldots,18$ \cite{Oishi.2017:rd9648}, see Eq.~\eqref{eq:Oishi-random}. Each of the 8 nodal zones in which the corresponding node can be placed randomly is shown in red; all nodal zones are cubes, except for Node B (an interval on the $x$ axis) and for Node C (a square in the plane $(x,y)$).  
}
	\label{fig:Oishi-2017-regular-hex-boxes}
\end{figure}

To train the neural networks involved in their approaches, 
%
% CMES style rewriting
%\cite{Oishi.2017:rd9648} created
a large set of distorted elements was created by randomly displacing seven nodes of a regular cube \cite{Oishi.2017:rd9648},
\begin{equation}
	\begin{aligned}
		%&B(1\pm r_1 d, 0, 0) , \quad
		%&&C(1\pm r_2 d, 1\pm r_3 d, \pm r_4 d) , 
		%&&D(\pm r_5 d, 1\pm r_6 d, 0) , 
		%&&E(\pm r_7 d, \pm r_8 d, 1 \pm r_9 d) , 
		B(1\pm r_1 d, 0, 0) , \quad
		C(1\pm r_2 d, 1\pm r_3 d, \pm r_4 d) , \quad 
		D(\pm r_5 d, 1\pm r_6 d, 0) , \quad
		E(\pm r_7 d, \pm r_8 d, 1 \pm r_9 d) ,
		\\
		F(1\pm r_{10}d,\pm r_{11}d,1\pm r_{12}d) , \quad
		G(1\pm r_{13}d,1\pm r_{14}d,1\pm r_{15}d) , \quad
		H(\pm r_{16}d,1\pm r_{17}d,1\pm r_{18}d) ,
	\end{aligned}
\label{eq:Oishi-random}
\end{equation}
where $r_i \in [0,1]$, $i = 1, \ldots, 18$ were 18 random numbers, see Figure~\ref{fig:Oishi-2017-regular-hex-boxes}. 
The elements were collected into five groups according to five different degrees of maximum distortion (maximum \emph{possible} nodal displacement) $d \in \left\{ 0.1, 0.2, 0.3, 0.4, 0.5 \right\}$.
Elements in the set $d = 0.5$ would only be highly distorted with $r_i$ having values closer to 1, but may only be slightly distorted with $r_i$ having values closer to 0.\footnote{
%	{\color{red} [NOTE: 2020.05.20. past tense would be better, at least for negative things. ENDNOTE]}
	In fact, 
%	\cite{Oishi.2017:rd9648} provided 
	a somewhat ambiguous description of the random generation of elements was provided in \cite{Oishi.2017:rd9648}. On the one hand, 
%	they 
	the authors 
	stated that a {\it``\ldots coordinates of nodes are changed using a uniform random number $r$ \ldots,''} and did not distinguish the random numbers in Eq.~\eqref{eq:Oishi-random} by subscripts. 
	On the other hand, 
	they
%	the authors 
	noted that exaggerated distortion may occur if nodal coordinates of an element were changed independently, and introduced the constraints on the distortion mentioned above in that context. 
	If the same random number $r$ were used for all nodal coordinates, all elements generated would exhibit the same mode of distortion.
} 
To avoid large distortion, the displacement of each node was restricted to the range of \num{0.5} to \num{2}, and the angle between adjacent faces must lie within the range of \SI{90}{\degree} to \SI{120}{\degree}.
Applying the normalization procedure, an element was characterized by a total of 18 
%{\color{red} [NOTE 2020.05.20.  should we say ``random'', to be more specific than ``non-trivial'' ? ENDNOTE]} 
nodal coordinates randomly distributed, but in a specific manner according to Eq.~(\ref{eq:Oishi-random}). 
%
%{\color{blue}
%[NOTE: 2020.01.21: it's not clear from the paper, whether ``uniform'' means that the same random number is used for all nodes (and nodal coordinates). 
%}
%

To quantify the quadrature error, 
%
% CMES style rewriting
the authors of 
\cite{Oishi.2017:rd9648} introduced $e(q)$, a relative measure of accuracy for the components of the stiffness matrix $k_{ij}$ as a function of the number integration points $q$ along each local coordinate,\footnote{
	For example, for a $2 \times 2 \times 2$ integration with a total of $8$ integration points, $q = 2$.
} defined as
%
%To quantify the quadrature error of the stiffness matrix with components $k_{ij}$, \cite{Oishi.2017:rd9648} define a relative error as function of the number of quadrature points $q$ as
\begin{equation} \label{eq:oishi-error}
e(q) = \frac{1}{\max_{i,j} \vert k_{ij}^{q_{\rm max}} \vert } \sum_{i,j} \vert k_{ij}^{q}  - k_{ij}^{q_{\rm max}} \vert \ ,
\end{equation} 
where $k_{ij}^{q}$ denotes the component in the $i$-th row and $j$-th column of the stiffness matrix obtained with $q$ integration points.
The error $e(q)$ is measured with respect to reference values $k_{ij}^{q_{\rm max}}$, which are obtained using the Gauss-Legendre quadrature with $q_{\rm max} = 30$ integration points, i.e., a total of $30^3 = \num{27000}$ integration points for 3-D elements.
%
%
%
%{\color{red}
%[NOTE: 2019.01.02, we need to mention that \cite{Oishi.2017:rd9648} actually used the sigmoid function to obtain their results, and not ReLU, which provided similar results.  their few tests indicated that ReLU was more efficient than the sigmoid function, which was already implemented in their code.
%refer also to Footnote~\ref{fn:for-experts}.
%ENDNOTE]
%}

\subsection{Application 1.1: Method 1, Optimal number of integration points}
\label{sc:Oishi-1.1}
The details of this particular deep-learning application, mentioned briefly in Section \ref{sc:motivation} on motivation via applications of deep learning---specifically Section \ref{sc:Oishi-summary}, item \ref{Oishi:method1}---are provided here.
% below is new
The idea is to have a neural network predict, for each element (particularly distorted elements), the number of integration points that provides accurate integration within a given error tolerance $\etol$, \cite{Oishi.2017:rd9648}. 

\begin{figure}[h]
	\centering
	\includegraphics[width=0.44\textwidth]{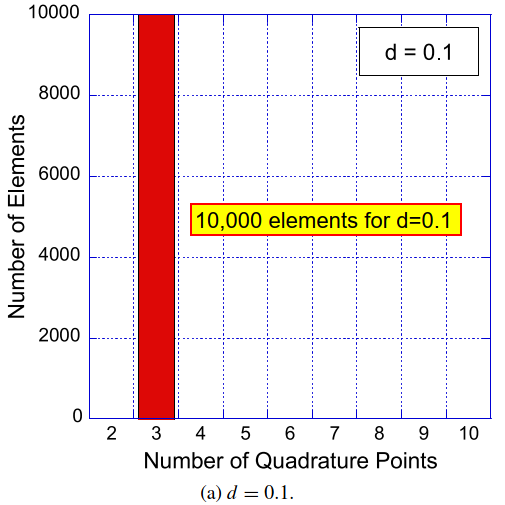}
	\includegraphics[width=0.45\textwidth]{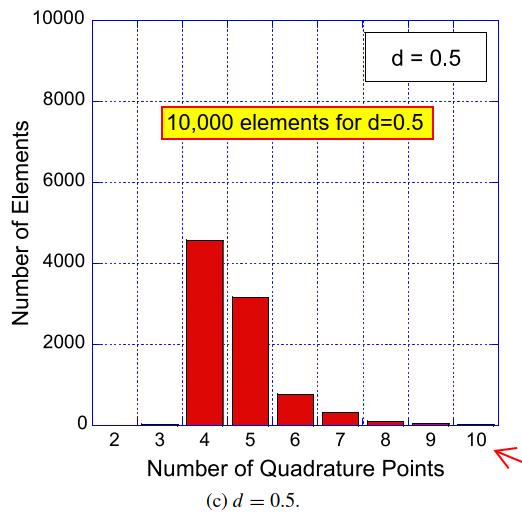}
	\caption{\textit{Method 1, Optimal number of integration points, feasibility} (Section~\ref{sc:Oishi-method-1-feasibility}). Distribution of minimum numbers of integration points on a local coordinate axes for a maximum error of $\etol=10^{-3}$ among 10,000 elements generated randomly using the method in Figure~\ref{fig:Oishi-2017-regular-hex-boxes}. 
	For $d=0.1$, all elements were only slightly distorted, and  required 3 integration points each.
	For $d=0.5$, close to 5,000 elements required 4 integration points each; very few elements required 3 or 10 integration points \cite{Oishi.2017:rd9648}.
		%\\
		{\footnotesize (Figure reproduced with permission of the authors.)}
	}
	\label{fig:Oishi-Fig-8c}
\end{figure}

\subsubsection{Method 1, feasibility study}
\label{sc:Oishi-method-1-feasibility}
%
% CMES style rewriting
%In their example,~\cite{Oishi.2017:rd9648} required 
In the example in \cite{Oishi.2017:rd9648},
the quadrature error is required to be smaller than $\etol = \num{1e-3}$. 
For this purpose, a fully-connected \hyperref[sc:feedforward]{feed-forward neural network} with $N$ hidden layers of $50$ neurons each was used. 
%The 
The non-trivial nodal coordinates were fed as inputs to the network, i.e., $\xv = \yv^{(0)} \in \mathbb R^{18 \times 1}$.
This neural network performed a classification task,\footnote{
%	{\color{red} [NOTE: 2020.05.27.  the long content of this footnote, which remains here, was moved to Section~\ref{sc:classification} in Section~\ref{sc:cost-function} on cost functions.  ENDNOTE]}
	%
	% CMES style rewriting
	The authors of 
	\cite{Oishi.2017:rd9648} used the squared-error loss function (Section~\ref{sc:mean-squared-error}) for the classification task, for which the softmax loss function can also be used, as discussed in Section~\ref{sc:classification}.
	\label{footnote:classification}
}
where each class corresponded to the minimum number of integration points $q$ along a local coordinate axis for a maximum error $\etol = 10^{-3}$.  Figure~\ref{fig:Oishi-Fig-8c} presents the distribution of 10,000 elements generated randomly for two degrees of maximum possible distortion, $d=0.1$ and $d=0.5$, using the method of Figure~\ref{fig:Oishi-2017-regular-hex-boxes}, and classified the minimum number of integration points.
%
% CMES style rewriting  
%\cite{Oishi.2017:rd9648} also presented 
Similar results for $d=0.3$ were presented in \cite{Oishi.2017:rd9648}, in which the conclusion  
%Their conclusion 
that ``as the shape is distorted, more integration points are needed'' is expected.

%{\color{red} [NOTE 2020.06.11.  with the above sub-section, we may not need this paragraph at all.  ENDNOTE]}
%With the minimum number of integration points being 2, the output layer was the matrix $\bout \in \mathbb R^{(q_{\rm max}-1) \times 1}$, where $q_{\rm max}$ was the maximum number of integration points allowed, with $q_{\rm max} = 9$ in the example provided in \cite{Oishi.2017:rd9648}.  {\color{red} [NOTE 2020.05.30.  i rewrote this sentence; please check.  there was something wrong with the phrase ``With the minimum number of integration points being 2'', which does not seem to fit in the sentence.  there is no relationship between ``2'' and $q_{max} = 9$, and why $q_{max} -1$ as the number of rows in $\bout$ ?  it is not clear where \cite{Oishi.2017:rd9648} mentioned $q_{max} = 9$ and $q_{max} -1$ as dimension of the output matrix $\bout$.  Figure8 in \cite{Oishi.2017:rd9648}, with the caption ``Number of quadrature points required to converge within prescribed error ($10^{-3}$) versus distribution of numbers of elements'', the number of integration points could be 10, i.e., $q_{max} = 10$. i looked into \cite{Oishi.2017:rd9648}, Section 3.5.1.2 ``Training phase'', p.339.  ENDNOTE]}

\subsubsection{Method 1, training phase}
\label{sc:Oishi-method-1-training}
To train the network,
%
% CMES style rewriting
%~\cite{Oishi.2017:rd9648} generated 
\num{2000} randomly distorted elements were generated for each of the five degrees of maximum distortion, $d \in \{0.1, 0.2, 0.3, 0.4, 0.5\}$, to obtain a total of \num{10000} elements \cite{Oishi.2017:rd9648}.
Before the network was trained, the minimum number of integration points $\qopt$ to meet the accuracy tolerance $\etol$ was determined for each element.

\begin{figure}[h]
	\centering
	\includegraphics[width=\textwidth]{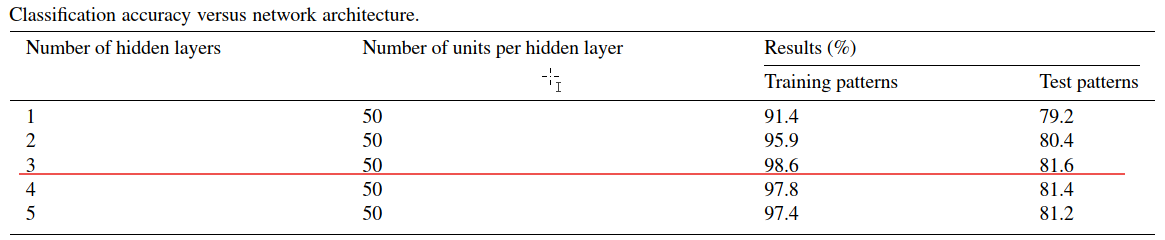}
	\caption{\textit{Method 1, Optimal network architecture for training} (Section~\ref{sc:Oishi-method-1-training}). The number of hidden layers varies from 1 to 5, keeping the number of neurons per hidden layer constant at 50.  The network with 3 hidden layers provided the highest accuracy for both the training set (``patterns'') at 98.6\% and for the validation set (``test patterns'') at 81.6\%.  Increase the network depth does not necessarily increase the accuracy \cite{Oishi.2017:rd9648}.
		%\\
		{\footnotesize (Table reproduced with permission of the authors.)}
	}
	\label{fig:Oishi-tab1}
\end{figure}

The whole dataset was partitioned\footnote{
	%
	% CMES style rewriting
%	\cite{Oishi.2017:rd9648} did not indicate 
	There was no indication in \cite{Oishi.2017:rd9648} on
	how these 5,000 elements were selected from the total of 10,000 elements, perhaps randomly.
} into a training set $\Xbb = \{ \bexin{1} , \cdots , \bexin{\Bsize}  \}$, $\Ybb = \{ \bexout{1} , \cdots , \bexout{\Bsize}  \}$ of $\Bsize = \num{5000}$ elements and an equally large validation set,\footnote{
	\label{fn:oishi-test-patterns}
	In contrast to the terminology of the present paper, in which ``training set'' and ``validation set'' are used  (Section~\ref{sc:training-valication-test}), 
	%
	% CMES style rewriting
%	\cite{Oishi.2017:rd9648} used 
	the terms ``training patterns'' and ``test patterns'', respectively, were used in \cite{Oishi.2017:rd9648}. 
	The ``test patterns'' were used in the training process, since
	%
	% CMES style rewriting
	the authors of  
	\cite{Oishi.2017:rd9648} \emph{``\ldots terminated the training before the error for test patterns started to increase''} (p.331).
	These ``test patterns'', based on their use as stated, correspond to the elements of the \emph{validation set} in the present paper, Figure~\ref{fig:Oishi-tab2}.  Technically,
	%
	% CMES style rewriting 
%	\cite{Oishi.2017:rd9648} did not have a 
	there was no test set in \cite{Oishi.2017:rd9648}; see Section~\ref{sc:training-valication-test}.
} by which 
%
% CMES style rewriting
%\cite{Oishi.2017:rd9648} monitored 
the training progress was monitored \cite{Oishi.2017:rd9648}, as described in Section~\ref{sc:training-valication-test}.
%
% CMES style rewriting
The authors of 
\cite{Oishi.2017:rd9648} explored 
several network architectures (with depth ranging from 1 to 5, keeping the width fixed at 50) to determine the optimal structure of their classifier network, which used the \hyperref[sc:logistic-sigmoid]{{logistic sigmoid}} as activation function.\footnote{
%{
%\color{blue}
%
% CMES style rewriting
The first author of 
\cite{Oishi.2017:rd9648} provided the information on the activation function through a private communication to the authors on 2018 Nov 16.
Their tests with the \hyperref[sc:relu]{ReLU} did not show improved performance (in terms of accuracy) as compared to the logistic sigmoid.
}\textsuperscript{,}\footnote{
	\label{fn:oishi-classification}
%	{\color{red} [NOTE 2020.06.12.  at the beginning of their paper,  \cite{Oishi.2017:rd9648} did mention the squared error loss functions in their Eqs.~(3) and (4), which are simpler than the softmax function, and more familiar to engineers.  they surely used these squared error loss functions. ENDNOTE]}
	Even though 
	%
	% CMES style rewritiing
%	\cite{Oishi.2017:rd9648} used 
	the squared-error loss function (Section~\ref{sc:mean-squared-error}) was used in \cite{Oishi.2017:rd9648}, we also discuss the softmax loss function for classification tasks; see Section~\ref{sc:classification}.  
	%Neither the loss function (maximum likelihood/cross-entropy loss?) nor the optimization method are mentioned in the paper.
	%} 
}
Their optimal feed-forward network, composed of 3 hidden layers of 50 neurons each, correctly predicted the number of integration points needed for 
% \SI{96.6}{\percent}
98.6\% 
%	[NOTE 2020.06.07.  based on Row 3 in Figure~\ref{fig:Oishi-tab1}, which is Table 1 in \cite{Oishi.2017:rd9648}, the correct number should be 98.6\%. --- 2020.06.05.  should check this number since i got 98.6\% below, not 96.6\%; there are 2 digits in agreement, the 1st and the 3rd, but not the 2nd.  could this be a misprint ?  need to check Table (b) in Figure~\ref{fig:Oishi-tab2}.  ENDNOTE]} 
of the elements in the training set, and for \SI{81.6}{\percent} of the elements in the validation set, Figure~\ref{fig:Oishi-tab1}.

\subsubsection{Method 1, application phase}
\label{sc:Oishi-method-1-application}
\begin{figure}[h]
	\centering
	\includegraphics[width=\textwidth]{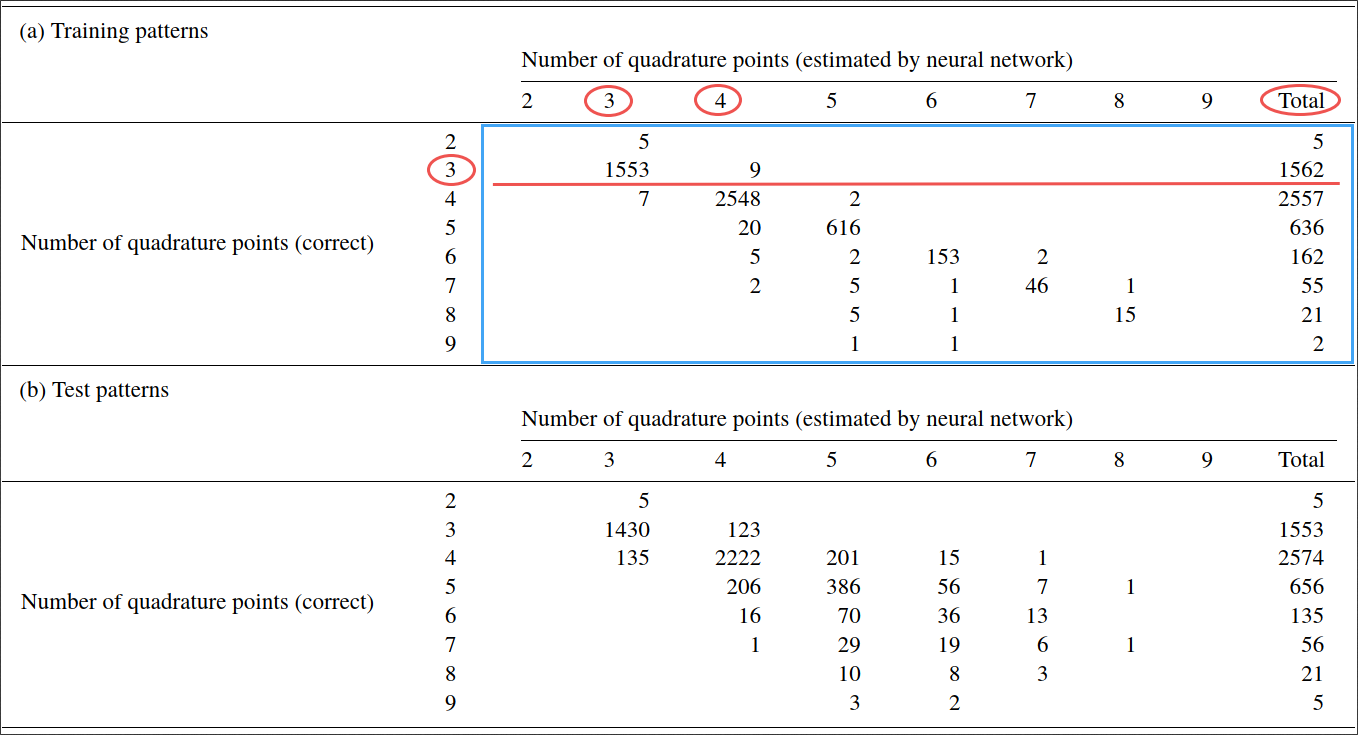}
	\caption{\textit{Method 1, application phase} (Section~\ref{sc:Oishi-method-1-application}). The numbers of quadrature points predicted by the neural network was compared to the minimum numbers of quadrature points for maximum error $\etol=10^{-3}$ \cite{Oishi.2017:rd9648}.  Table (a) shows the results for the training set (``patterns''), and Table (b) for the validation set.
		%\\
		{\footnotesize (Table reproduced with permission of the authors.)}
	}
	\label{fig:Oishi-tab2}
\end{figure}

The correct number of quadrature points and the corresponding number of points predicted by the neural network are illustrated in Figure~\ref{fig:Oishi-tab2} for both the training set (``patterns'') in Table (a) and the validation set (``test patterns'') in Table (b).

%{\color{red} [NOTE: 2020.06.07.  i also expanded the paragraph below and added another one further below.  2020.06.05. The paragraph below was intended for the caption of Figure~\ref{fig:Oishi-tab2}, but made the caption too long for latex to handle (the figure just did not show up).  as a results, i moved it in the main text below. ENDNOTE]}

Both the training set and the validation set, each had \num{5000} distorted element shapes.
As an example to interpret these tables, take Table (a), Row 2 (red underline, labeled ``3'' in red circle) of the matrix (blue box): Out of a Total of \num{1562} element shapes (last column) in the training set that were ideally integrated using 3 quadrature points (in red circle), the neural network correctly estimated a need of 3 quadrature points  (Column 2, labeled ``3'' in red circle) for \num{1553} element shapes, and 4 quadrature points (Column 3, labeled ``4'' in red circle) for \num{9} element shapes.  That's an accuracy of 99.4\% for Row 2.  The accuracy varies, however, by row, 0\% for Row 1 (2 integration points), 99.6\% for Row 3, ..., 71.4\% for Row 7, 0\% for Row 8 (9 integration points).
The numbers in column ``Total'' add up to 
5 + 1562 + 2557 + 636 + 162 + 55 + 21 + 2 = 5000
elements in the training set.
The diagonal coefficients add up to  
1553 + 2548 + 616 + 153 + 46 + 15 = 4931 elements with correctly predicted number of integration points, yielding the overall accuracy of
4931 / 5000 = 98.6\% in training, Figure~\ref{fig:Oishi-tab1}.

For Table (b) in Figure~\ref{fig:Oishi-tab2}, the numbers in column ``Total'' add up to
5 + 1553 + 2574 + 656 + 135 + 56 + 21 + 5 = 5005 elements in the validation set, which should have 5000, as written by \cite{Oishi.2017:rd9648}.  Was there a misprint ?
The diagonal coefficients add up to
1430 + 2222 + 386 + 36 + 6 = 4080 elements with correctly predicted number of integration points, yielding the accuracy of
4080 / 5000 = 81.6\%, which agrees with Row 3 in Figure~\ref{fig:Oishi-tab1}.  As a result of this agreement, the number of elements in the validation set (``test patterns'') should be 5000, and not 5005, i.e., there was a misprint in column ``Total''.

%\noindent
%{\color{red} [NOTE: 2020.06.12. i am done with Section 9.2, Method 1, above.  now continue to work on Section 9.3, Method 2, below. ENDNOTE]}

\begin{figure}[h]
	\centering
	\includegraphics[]{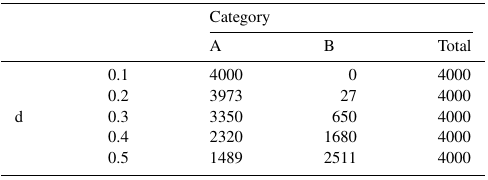}
	\caption{
		\textit{Method 2, Quadrature weight correction, feasibility} (Section~\ref{sc:quadrature-weight-correction-feasibility}). 
		Each element was tested 1 million times with randomly generated sets of quadrature weights.  There were 4000 elements in each of the 5 groups with different degrees of maximum distortion, $d$.
		Quadrature weight correction effectiveness increased with element distortion. Weakly distorted elements ($d = 0.1$) did not have any improvement, and thus belonged to Category A (error ratio $R_{\rm error} \geq 1$).  As $d$ increased, the size of Category A decreased, while the size of Category B increased.   Among the \num{4000} elements in the group with  $d = 0.5$, the stiffness matrices of \num{2511} elements could be integrated more accurately by correcting their quadrature weights  (Category $B$, $R_{\rm error} < 1$) \cite{Oishi.2017:rd9648}.
		{\footnotesize (Table reproduced with permission of the authors.)}
	}
	\label{fig:Oishi-2017-weight-correction-classes}
\end{figure}

\subsection{Application 1.2: Method 2, optimal quadrature weights} 
\label{sc:Oishi-1.2}
The details of this particular deep-learning application, mentioned briefly in Section \ref{sc:motivation} on motivation via applications of deep learning, Section \ref{sc:Oishi-summary}, item \ref{Oishi:method2}, are provided here.
As an alternative to increasing the number of quadrature points,
%
% CMES rewriting 
the authors of 
\cite{Oishi.2017:rd9648} proposed
to compensate for the quadrature error introduced by the element distortion by adjusting the quadrature weights at a fixed number of quadrature points. 
For this purpose, they introduced correction factors $\{ w_{i,j,k}\}$ that were predicted by a neural network, and were multipliers for the corresponding standard quadrature weights $\{ H_{i,j,k} \}$ of an undistorted hexahedron. 
To exactly compute the components of the stiffness matrix, undistorted linear hexahedra require eight quadrature points ($i=j=k=2$) at local positions $\xi, \eta, \zeta = \pm 1 / \sqrt{3}$ with uniform weights $H_{i,j,k} = 1$.

The data preparation here was similar to that in Method 1, Section~\ref{sc:Oishi-1.1}, except that 
%
% CMES rewriting
%\cite{Oishi.2017:rd9648}  generated 
20,000 randomly distorted elements were generated, with \num{4000} elements in each of the five groups, each group having a different degree of maximum distortion $d \in \{ 0.1, 0.2, 0.3, 0.4, 0.5 \}$, as depicted in Figure~\ref{fig:Oishi-2017-regular-hex-boxes} \cite{Oishi.2017:rd9648}.  

\begin{figure}[h]
	\centering
	\includegraphics[width=0.8\linewidth]{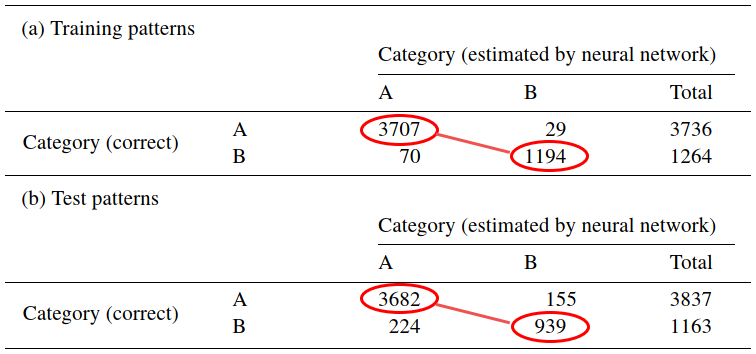}
	\caption{
		\textit{Method 2, training phase, classifier network} (Section~\ref{sc:quadrature-weight-correction-modeling}). 
		%A neural network is trained to predict whether the error in numerical quadrature can be reduced by correcting the quadrature weights (Category B) or not (Category A). 
		%The feedforward neural network is a binary classifier that takes \num{18} non-trivial normalized nodal coordinates as input and has a single output, i.e., the conditional probability of an element admitting improvements by weight correction (Category B). 
		The training and validation sets comprised \num{5000} elements each, of which \num{3707} and \num{3682}, respectively, belonged to Category A (no improvements upon weight correction).
		A \emph{first} neural network with 4 hidden layers of \num{30} neurons correctly classified $(3707 + 1194) / 5000 \approx \SI{98}{\percent}$ elements in the training set (a) and $(3682 + 939) / 5000 \approx \SI{92}{\percent}$ elements in the validation set (b).
		See also Figure~\ref{fig:Oishi-error} in Section~\ref{sc:Oishi-summary} \cite{Oishi.2017:rd9648}.
		{\footnotesize (Table reproduced with permission of the authors.)}
		%	{\color{red} 2020.03.19.  I used a new figure above to highlight the numbers.}
	}
	\label{fig:Oishi-2017-weight-correction-classification-net}
\end{figure}

\subsubsection{Method 2, feasibility study}
\label{sc:Oishi-method-2-feasibility}
\label{sc:quadrature-weight-correction-feasibility}
Using the above 20,000 elements, the feasibility of improving integration accuracy by quadrature weight correction was established in Figure~\ref{fig:Oishi-2017-weight-correction-classes}.
To obtain these results, a brute-force search was used:
For each of the \num{20000} elements, 
%
% CMES style rewriting
%\cite{Oishi.2017:rd9648} tested 
%\num{1000000} 
1 million sets of random correction factors $\{ w_{i,j,k}\} \in [\num{0.95}, \num{1.05}]$ were tested to find the optimal set of correction factors for that single element \cite{Oishi.2017:rd9648}.
The effectiveness of quadrature weight correction was quantified by the error reduction ratio defined by
\begin{equation}
	R_{\rm error} = \frac{\sum_{i,j} | k_{ij}^{q, opt} - k_{ij}^{q_{\rm max}}|}{\sum_{i,j} | k_{ij}^{q} - k_{ij}^{q_{\rm max}}|} ,
	\label{eq:Oishi-error-reduction-ratio}
\end{equation}
i.e., the ratio between the quadrature error defined in Eq.~\eqref{eq:oishi-error} obtained using the \emph{optimal} (``\emph{opt}'') corrected quadrature weights and the quadrature error obtained using the standard quadrature weights of Gauss-Legendre quadrature.
Accordingly, a ratio $R_{\rm error} < 1$ indicates that the modified quadrature weights improved the accuracy.
For each element, the optimal correction factors that yielded the smallest error ratio, i.e.,
\begin{equation}
\{ w_{i,j,k}^{opt} \} = \arg \min_{w_{i,j,k}} R_{\rm error} ,
\end{equation}
were retained as target values for training, and identified with the superscript ``\emph{opt}'', standing for ``optimal''.  The corresponding optimally integrated coefficients in the element stiffness matrix are denoted by $k_{ij}^{q, opt}$, with $i, j = 1, 2, \cdots$, as appeared in Eq.~\eqref{eq:Oishi-error-reduction-ratio}.

It turns out that a reduction of the quadrature error by correcting the quadrature weights is not feasible for all element shapes. 
Undistorted elements, for instance, which are already integrated exactly using standard quadrature weights, naturally do not admit improvements. 
%
% CMES style rewriting
%\cite{Oishi.2017:rd9648} classified 
These 20,000 elements were classified into two categories A and B \cite{Oishi.2017:rd9648}, Figure~\ref{fig:Oishi-2017-weight-correction-classes}.
Quadrature weight correction was not effective for Category A ($R_{\rm error} \geq 1$), but effective for Category B ($R_{\rm error} < 1$).
Figure~\ref{fig:Oishi-2017-weight-correction-classes} shows that elements with a higher degree of maximum distortion were more likely to benefit from the quadrature weight correction as compared to weakly distorted elements.  Recall that among the elements of the group $d=0.5$ were elements that were only slightly distorted (due to the random factors $r_{ij}$ in Eq.~\eqref{eq:Oishi-random} being close to zero), and therefore would not benefit from quadrature weight correction (Category A); there were 1489 such elements, Figure~\ref{fig:Oishi-2017-weight-correction-classes}.
%, i.e., elements for which weight correction is ineffective (category A, $R_{\rm error} \geq 1$) and elements that 

\subsubsection{Method 2, training phase}
\label{sc:Oishi-method-2-training}
\label{sc:quadrature-weight-correction-modeling}
Because the effectiveness of the quadrature weight correction strongly depends on the degree of maximum distortion, $d$, 
%
% CMES style rewriting
the authors of 
\cite{Oishi.2017:rd9648} proposed a two-stage approach for the correction of quadrature weights, which relied on two fully-connected feedforward neural networks.

In the first stage, a \emph{first} neural network, a binary classifier, was trained to predict whether an element shape admits improved accuracy by quadrature weight correction (Category B) or not (Category A). 
The neural network to perform the classification task took the \num{18} non-trivial nodal coordinates obtained upon the proposed normalization procedure for linear hexahedra as inputs, i.e., $\xv = \yv^{(0)} \in \mathbb R^{18 \times 1}$.
The output of the neural network was single scalar $\out \in \mathbb R$, where $\out = 0$ indicated an element of Category A and  $\out = 1$ an element of Category B.

%
% CMES style rewriting
%\cite{Oishi.2017:rd9648} selected \num{10000} 
Out of the \num{20000} elements generated, \num{10000} elements were selected to train the classifier network, for which both the training set and the validation set comprised \num{5000} elements each \cite{Oishi.2017:rd9648}.\footnote{
	Even though
	%
	% CMES style rewriting 
%	\cite{Oishi.2017:rd9648} did not give 
	a reason for not using the entire set of \num{20000} elements was not given, it could be guessed that 
%	they 
	the authors of \cite{Oishi.2017:rd9648}
	would want the size of the training set and of the validation set to be the same as that in Method 1, Section~\ref{sc:Oishi-1.1}.  Moreover, even though details were not given, the selection of these 10,000 elements would likely be a random process.
}
The optimal neural network in terms of classification accuracy for this application had 4 hidden layers with 30 neurons per layer; see  Figure~\ref{fig:Oishi-2017-weight-correction-classification-net}.
The trained neural network succeeded in predicting the correct category for \SIlist{98; 92}{\percent} of the elements in the training set and in the validation set, respectively.

\begin{figure}[h]
	\centering
	\includegraphics[]{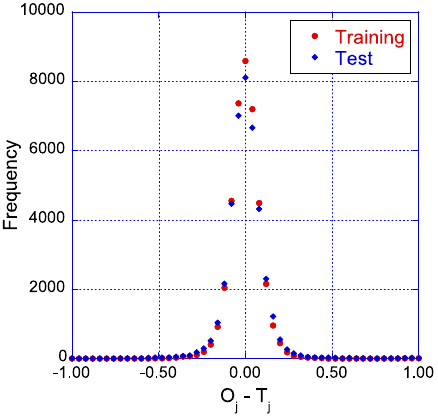}
	\caption{{\it Method 2, training phase, regression network} (Section~\ref{sc:quadrature-weight-correction-modeling}). A \emph{second} neural network estimated 8 correction factors $\{ w_{i,j,k}\}$, with $i,j,k \in \{1, 2\}$, to be multiplied by the standard quadrature weights for each element.  Distribution of normalized errors, i.e., the normalized differences between the predicted weights (outputs) $O_j$ and the true weights $T_j$ for the elements of the training set (red) and the test set (blue).
	For both sets, which consist of $5000 \times 8 = \num{40000}$ correction factors each, the error has a mean of zero and seems to obey a normal distribution  \cite{Oishi.2017:rd9648}.
	{\footnotesize (Figure reproduced with permission of the authors.)}
	}
	\label{fig:Oishi-2017-weight-correction-error-distribution}
\end{figure}

In the second stage, a \emph{second} neural network was trained to predict the corrections to the quadrature weights for all those elements, which allowed a reduction of the quadrature error. 
Again, the \num{18} non-trivial nodal coordinates of a normalized hexahedron were input to the neural network, i.e., $\xv = \yv^{(0)} \in \mathbb R^{18 \times 1}$.
The outputs of the neural network $\bout \in \mathbb R^{8 \times 1}$ represented the eight correction factors to the standard weights $\{ w_{i,j,k}^{opt} \}$, $i,j,k=1,2$, of the Gauss-Legendre quadrature. 
%
% CMES style rewriting
The authors of 
\cite{Oishi.2017:rd9648} stated that \num{10000} elements with an error reduction ration $R_{\rm error} < 1$ formed equally large training set and validation set, comprising \num{5000} elements each.\footnote{
	According to Figure~\ref{fig:Oishi-2017-weight-correction-classes}, only \num{4868} out of in total \num{20000} elements generated belonged to Category B, for which $R_{\rm error} < 1$ held. 
	%
	% CMES style rewriting
%	\cite{Oishi.2017:rd9648} did not provide 
	Further details on the \num{10000} elements that were being used for training and validation were not proviced in \cite{Oishi.2017:rd9648}.
}  
A neural network with 5 hidden layers of 
%
% don't use \num{} for numbers having only two digits
%\num{50}
50
neurons was reported to perform best in predicting the corrections to the quadrature weights. %\footnote{As \cite{Oishi.2017:rd9648} do not provide}
The normalized error\footnote{
	No details on the normalization procedure were provided in the paper.
}  distribution of the correction factors are illustrated in Figure~\ref{fig:Oishi-2017-weight-correction-error-distribution}.
%
%\begin{figure}[h!]

\subsubsection{Method 2, application phase}
\label{sc:Oishi-method-2-application}
The effectiveness of the numerical quadrature with corrected quadrature weights was already presented in Figure~\ref{fig:Oishi-error} in Section~\ref{sc:Oishi-summary}, %Figure~\ref{fig:Oishi-2017-weight-correction-results}, 
%which shows the error-reduction ratio $R_{\rm error}$ achieved for the training set (a) and for the validation set (b).
with the distribution of the error-reduction ratio $R_{\rm error}$ defined in Eq.~\eqref{eq:Oishi-error-reduction-ratio} for the training set (patterns) (a) and the validation set (test patterns) (b).  Further explanation is provided here for Figure~\ref{fig:Oishi-error}.\footnote{
	The caption of Figure~\ref{fig:Oishi-error}, if it were in Section~\ref{sc:Oishi-method-2-application}, would begin with \emph{``Method 2, application phase''} in parallel to the caption of Figure~\ref{fig:Oishi-tab2}.
}

The red bars (``Optimized'') in Figure~\ref{fig:Oishi-error} represent the distribution of the error-reduction ratio $R_{\rm error}$ in Eq.~\eqref{eq:Oishi-error-reduction-ratio}, using the optimal correction factors, which were themselves obtained by a brute force method (Sections~\ref{sc:Oishi-method-2-feasibility}-\ref{sc:Oishi-method-2-training}).  While these optimal correction factors were used as targets for training (Figure~\ref{fig:Oishi-error} (a)),
they were used only to compute the error-reduction ratios $R_{\rm error}$ for elements in the validation set  
(Figure~\ref{fig:Oishi-error} (b)).

The blue bars (``Estimated by Neuro'') correspond to the error reduction ratios achieved with the corrected quadrature weights that were predicted by the trained neural network.
 
The error-reduction ratios $R_{\rm error} < 1$ indicate improved quadrature accuracy, which occurred for \SI{99}{\percent} of the elements in the training set, and for \SI{97}{\percent} of the elements in the validation set.
Very few elements had their accuracy worsened ($R_{\rm error} > 1$) when using the predicted quadrature weights. 

There were no red bars (optimal weights) with $R_{\rm error} > 1$ since only elements that admitted improved quadrature accuracy by optimal quadrature weight correction were included in the training set and validation set. 
The blue bars with error ratios $R_{\rm error} > 1$ corresponded to the distorted hexahedra for which the accuracy worsened as compared to the standard weights of Gauss-Legendre quadrature. 

%According to \cite{Oishi.2017:rd9648}, the corrected weights predicted by the neural network improved ($R_{\rm error} < 1$) accuracy of numerical quadrature for \SI{99}{\percent} of the elements in the training set and \SI{97}{\percent} of the elements in the validation set.
%Improved quadrature accuracy was obtained for \SI{99}{\percent} of the elements in the training set, and for \SI{97}{\percent} of the elements in the validation set.

%Note that only elements that admitted improved quadrature by weight correction were included in the training set and validation set; thus was the reason why no red bars (optimal weights) appeared for error ratios $R_{\rm error} > 1$.
%

%
% CMES style rewriting
The authors of 
\cite{Oishi.2017:rd9648} concluded their paper with a discussion on the computational efforts, and in particular the evaluation of trained neural networks in the application phase.
As opposed to computational mechanics, where we are used to double precision floating-point arithmetics, deep neural networks have proven to perform well with reduced numerical precision.\footnote{
%	\color{blue}See \cite{} ADD reference. {\color{red} [NOTE 2020.06.15.  we need to cite these refs here:
	See, e.g.,
	\vphantom{\cite{gupta2015deep}}\cite{gupta2015deep} \vphantom{\cite{courbariaux2016binarized}}\cite{courbariaux2016binarized} \vphantom{\cite{de2017understanding}}\cite{de2017understanding}.  
%	ENDNOTE]}
}
To speed up the evaluation of the trained networks, 
%
% CMES style rewriting
%\cite{Oishi.2017:rd9648} simply removed 
the least significant bits from all parameters (weights, biases) and the inputs were simply removed \cite{Oishi.2017:rd9648}.
In both the estimation of the number of quadrature points and the prediction of the weight correction factors, half precision floating-point numbers (16 bit) turn out to show sufficient accuracy almost on par with single precision floats (32 bit).

\section{Application 2: Solid mechanics, multi-scale, multi-physics}
\label{sc:solid}
\label{sc:Wang-Sun-2018-2}
The results and deep-learning concepts used in \cite{Wang.2018:rd3109} were presented in Section~\ref{sc:Wang-Sun-2018} further above. In this section, we discuss some details of the formulation.

%
%\begin{figure}[h]
%	\centering
	%  \includegraphics[width=0.7\linewidth]{figures/multiphysics.png}
%	\includegraphics[width=0.8\linewidth]{figures/Wang-multiphysics.pdf}
%	\caption{
%		\emph{Porous-media multiphysics block diagram.}
%		Information flow and dependencies in multiscale problems of poromechanics: black arrows represent definitions or (accepted) first principles; red arrows stand for phenomenological relations that are not based on first principles.
%		This block diagram is a more complex version of the single-physics block diagram in Figure~\ref{fig:Wang-2018-single-physics}.
%		\cite{Wang.2018:rd3109}.
%		{\footnotesize (Figure reproduced with permission of the authors.)}
		%{\color{red} ASK PERMISSION } 
		%${\color{blue} [NOTE: 2018.10.16, I replaced the figure by a vectorgraphics (PDF) version. ENDNOTE]}
	%}
	%\label{fig:multiphysics}
%\end{figure}
%

%\cite{Wang.2018:rd3109}
%\href{https://drive.google.com/open?id=1XIJ2AwRuknEcatb1A2BeRor_qmtTIl6Q}{(pdf)}
%

\subsection{Multiscale problems}
\label{sc:Wang-multiscale-problems}
Multiscale problems are characterized by the fact that couplings of physical processes occurring on different scales of length and/or time needs to be considered.
In the field of computational mechanics, multiscale models are often used to accurately capture the constitutive behavior on a macroscopic length scale, since resolving the entire domain under consideration on the smallest relevant scale if often intractable. 
To reduce the computational costs, multiscale techniques as, e.g., coupled DEM-FEM or coupled FEM-FEM (known as FEM$^2$),\footnote{
	%\color{blue} TODO: find better references than Wang's
	DEM = Discrete Element Method.  FEM = Finite Element Method.
	See references in \cite{Wang.2018:rd3109}.
} have been proposed for bridging length scales by deducing constitutive laws from micro-scale models for Representative Volume Elements (RVEs).\footnote{
	RVEs are also referred to as \emph{representative elementary volumes} (REVs) or, simply, \emph{unit cells}.
}
These approaches no longer require the entire macroscopic domain to be resolved on the micro-scale.
In FEM$^2$, for instance, the microscale model to deduce the constitutive behavior on the macroscopic scale is evaluated at the quadrature points of the microscale model. 

\begin{figure}[h]
	\centering
	\includegraphics[width=0.9\linewidth]{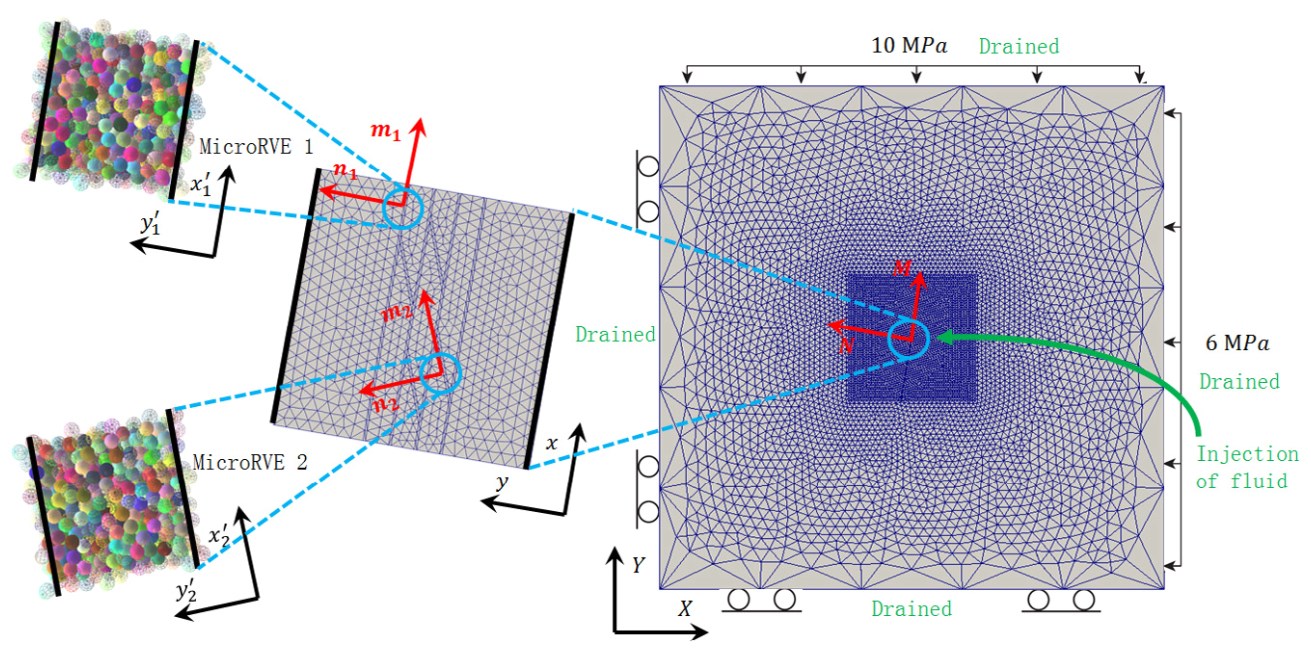}
	\caption{
		\emph{Three scales in data-driven fault-reactivation simulations} (Sections~\ref{sc:Wang-Sun-2018}, \ref{sc:Wang-multiscale-problems}, \ref{sc:Wang-strong-discontinuities}). %\ref{sc:Wang-results}).
		Relative orientation of Representative Volume Elements (RVEs).
		\emph{Left:} Microscale ($\mu$) RVE using Discrete Element Method (DEM), Figure~\ref{fig:Wang-micro-RVE} and Row 1 of Figure~\ref{fig:Wang-RNN-FEM}.
		\emph{Center:} Mesoscale (cm) REV using FEM; Row 2 of Figure~\ref{fig:Wang-RNN-FEM}.
		\emph{Right:} Field-size macroscale (km) FEM model; Row 3 of Figure~\ref{fig:Wang-RNN-FEM}
		\cite{Wang.2018:rd3109}.
%		{\color{red} [NOTE: Figure15, pdf p.26.]}
		{\footnotesize (Figure reproduced with permission of the authors.)}
	}
	\label{fig:Wang-DEM-FEM-three-scales}
\end{figure}

The multiscale problem in the mechanics of porous media tackled in \cite{Wang.2018:rd3109} is represented in Figure~\ref{fig:Wang-DEM-FEM-three-scales}, where the relative orientations among the three models at microscale, mesoscale, and macroscale in Figure~\ref{fig:Wang-RNN-FEM} are indicated.  The method of analysis (DEM or FEM) in each scale is also indicated in the figure.

\begin{figure}[h]
	\centering
	\includegraphics[width=0.9\textwidth]{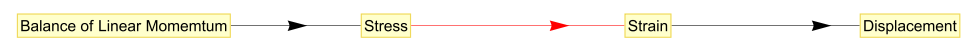}
	\caption{
		\emph{Single-physics block diagram} (Section~\ref{sc:Wang-data-drive-constitutive}).
		%		(Section~\ref{sc:Wang-Sun-2018})
		Single physics is an easiest way to see the role of deep learning in modeling complex nonlinear constitutive behavior (stress-strain relation, red arrow), as first realized
		%
		% CMES style rewriting 
%		by 
		in
		\cite{Ghaboussi.1991:rd0001}, where balance of linear momentum and strain-displacement relation are definitions or accepted ``universal principles'' (black arrows) 
		%The approach is extended to multiphysics in \cite{Wang.2018:rd3109} for porous media, with the block diagram given in Figure~\ref{fig:multiphysics}.
		%\\
		\cite{Wang.2018:rd3109}
		{\footnotesize (Figure reproduced with permission of the authors.)}
	}
	\label{fig:Wang-2018-single-physics}
\end{figure}

% 2020.08.29
% not necessary to present these figures for now
%Wang-2018-micro-RVE-TE1-test-data
%Wang-2018-micro-RVE-TE1-RNN-LSTM-prediction

\subsection{Data-driven constitutive modeling, deep learning}
\label{sc:Wang-data-drive-constitutive}
Despite the diverse approaches proposed, multiscale problems remain a computationally challenging task, which
%
% CMES style rewriting  
%\cite{Wang.2018:rd3109} tackled 
was tackled in \cite{Wang.2018:rd3109}
by means of a hybrid data-driven method by combining deep neural networks and conventional constitutive models. 
To illustrate the hierarchy among the relations of models and to identify, 
%
% CMES style rewriting
the authors of 
\cite{Wang.2018:rd3109} used directed graphs, which also indicated the nature of the individual relations by the colors of the graph edges. Black edges correspond to ``universal principles'' whereas red edges represent phenomenological relations, see, e.g., the classical problem in solid mechanics shown in Figure~\ref{fig:Wang-2018-single-physics}.
Within classical mechanics, the balance of linear momentum is axiomatic in nature, i.e., it represents a well-accepted premise that is taken to be true.  
The relation between the displacement field and the strain tensor represents a definition.
The constitutive law describing the stress response, which, in the elastic case, is an algebraic relation among stresses and strains, is the only phenomenological part in the {``single-physics''} solid mechanics problem and, therefore, highlighted in red.
%{\color{blue}
%2020.04.15. The question what may be considered phenomenological or not may easily become controversial. One could object that universal principles are phenomenological by definition since they cannot be proven within a theory and therefore must rely on observations. But I would rather not like to launch a discussion \ldots
%}
%

In many engineering problems, stress-strain relations of, possibly nonlinear, elasticity, which are parameterized by a set of elastic moduli, can be used in the modeling. 
For heterogeneous materials as, e.g., in composite structures, even the ``single physics'' problem of elastic solid mechanics may necessitate multiscale approaches, in which constitutive laws are replaced by RVE simulations and homogenization. 
%
% CMES style rewriting
%\cite{Wang.2018:rd3109} extended 
This approach was extended to multiphysics models of porous media, in which multiple scales needed to be considered \cite{Wang.2018:rd3109}.
%The directed graph in Figure~\ref{fig:multiphysics} already reveals much of the complexity and the dependencies in the hydro-mechanical coupling of the dual-porosity dual-permeability problems. 
%As in the single physics problem (Figure~\ref{fig:Wang-2018-single-physics}), red edges represent phenomenological relations that are not based on first principles or definitions.
The counterpart of Figure~\ref{fig:Wang-2018-single-physics} for the mechanics of porous media is complex, could be confusing for readers not familiar with the field, does not add much to the understanding of the use of deep learning in this study, and therefore not included here; see \cite{Wang.2018:rd3109}.  

The hybrid approach 
%
% CMES style rewriting
%of \cite{Wang.2018:rd3109},
in \cite{Wang.2018:rd3109}, 
which 
%they 
was 
described as graph-based machine learning model, retained those parts of the model which represented universal principles or definitions (black arrows). 
Phenomenological relations (red arrows), which, in conventional multiscale approaches, followed from microscale models, were replaced by computationally efficient data-driven models.
In view of the path-dependency of the constitutive behavior in the poromechanics problem considered, 
%
% CMES style rewriting
%\cite{Wang.2018:rd3109} 
it was
proposed in \cite{Wang.2018:rd3109}
to use \hyperref[sc:RNN]{recurrent neural networks} (RNNs), Section~\ref{sc:RNN}, constructed with  \hyperref[sc:LSTM]{Long Short-Term Memory} (LSTM) cells, Section~\ref{sc:LSTM}.

\subsection{Multiscale multiphysics problem: Porous media}
\label{sc:multiscale-multiphysics}
The problem of hydro-mechanical coupling in deformable porous media with multiple permeabilities is characterized by the presence of two or more pore systems with different typical sizes or geometrical features of the host matrix \cite{Wang.2018:rd3109}.
The individual pore systems may exchange fluid depending on whether the pores are connected or not. 
If the (macroscopic) deformation of the solid skeleton is large, plastic deformation and cracks may occur, which result in anisotropic evolution of the effective permeability.  
As a consequence, problems of this kind are not characterized by a single effective permeability, and, to identify the material parameters on the macroscopic scale, micro-structural models need to be incorporated. 

%
% CMES style rewriting
The authors of 
\cite{Wang.2018:rd3109} considered a saturated porous medium, which features two dominant pore scales: The regular solid matrix was characterized by micropores, whereas macropores may result, e.g., from cracks and fissures.
Both the volume and the partial densities of each constituent in the mixture of solid, micropores, macropores and voids were characterized by the porosity, i.e., the (local) ratio of pores and the total volume, as well as the fractions of the respective pore systems.  

\subsubsection{Recurrent neural networks for scale bridging}
\label{sc:Wang-RNN-scale-bridging}
%
% CMES style rewriting 
%\cite{Wang.2018:rd3109} used 
Recurrent neural networks (RNNs, Section~\ref{sc:RNN}), which are equivalent to ``very deep feedforward networks'' (Remark~\ref{rm:depth-of-RNN}), were used in \cite{Wang.2018:rd3109} as a scale-bridging method to efficiently simulate multiscale problems of poroplasticity.
In Figure~\ref{fig:Wang-RNN-FEM}, Section~\ref{sc:Wang-Sun-2018}, three scales were considered: Microscale ($\mu$), mesoscale (cm), macroscale (km).

%{
%	\color{purple} [OLD CAPTION of Figure~\ref{fig:Wang-RNN-FEM}: lation multiphysics in \cite{Wang.2018:rd3109}]
%}
%{\color{red}
%	[NOTE: 2020.03.14. The OLD CAPTION of of Figure~\ref{fig:Wang-RNN-FEM} (in purple just above) was mangled for some reason, probably due to cutting and pasting; why not use the same title as in \cite{Wang.2018:rd3109} as shown above.]
%} 

%{\color{red}
%	NOTE: 2020.04.07.  (i initially incorporated the text below into the caption of Figure~\ref{fig:Wang-RNN-FEM}, but Latex cannot process it, so i moved this text out of the caption, to use it somewhere else.)
	
	At the microscale, Discrete Element Method (DEM) is used to simulate a mesoscale
	Representative Volume Element (RVE) that consists of a cubic pack of microscale
	particles, with different loading conditions to generate a training set for a mesoscale
	RNN with LSTM architecture to model mesoscale constitutive response to produce loading
	histories $(\epsilon_{meso} , \stress^\prime_{meso})$ on demand.
	
	At the mesocale, Finite Element Method (FEM), combined with mesoscale loading histories
	$(\epsilon_{meso} $, $ \stress^\prime_{meso})$ produced by the above mesoscale RNN, are
	used to model a macroscale RVE, subjected to different loading conditions to generate a
	training set for a macroscale RNN with LSTM architecture to model macroscale
	constitutive response to produce loading histories $(\epsilon_{macro} , \stress^\prime_{macro})$ on demand.
	
	At the macroscale, Finite Element Method (FEM), combined with macroscale loading
	histories $(\epsilon_{macro} ,$ $\stress^\prime_{macro})$ produced by the above macroscale RNN, are used to model an actual domain at kilometer size (macroscale).

%}

%{\color{red} [NOTE: 2020.07.30.  Fabric tensor of the 1st kind of rank 2, \cite{kanatani1984distribution}; temporarily put here to get the equation label to refer to in the motivation Section~\ref{sc:Wang-Sun-2018} ENDNOTE]}

\begin{figure}[h]
	\centering
	\includegraphics[width=0.9\linewidth]{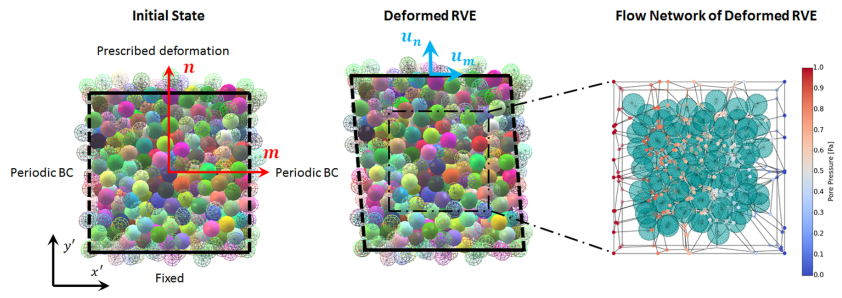}
	\caption{
		\emph{Microscale RVE} (Sections~\ref{sc:Wang-microstructure-data}, \ref{sc:Wang-optimal-RNN}, \ref{sc:Wang-strong-discontinuities}). %\ref{sc:Wang-results}).
		A $10 \text{ cm} \times 10 \text{ cm} \times 5 \text{ cm}$ box of identical spheres of $0.5 \text{ cm}$ diameter (Figure~\ref{fig:Wang-RNN-FEM}, row 1, Remark~\ref{rm:Wang-micro-RVE}), subjected to imposed displacements.
		(a) Initial configuration of the granular assembly; (b) Imposed displacement, deformed configuration; (c) Flow network generated from deformed configuration used to predict anisotropic effective permeability \cite{Wang.2018:rd3109}.
		See also Figure~\ref{fig:Wang-DEM-FEM-three-scales} and Figure~\ref{fig:Wang-micro-RVE-training-data}.
%		{\color{red} [NOTE: Figure16, pdf p.27.]}
		{\footnotesize (Figure reproduced with permission of the authors.)}
	}
	\label{fig:Wang-micro-RVE}
\end{figure}

\subsubsection{Microstructure and principal direction data}
\label{sc:Wang-microstructure-data}
To train the mesoscale RNN with LSTM units (which 
%
% CMES style rewriting
%\cite{Wang.2018:rd3109} 
was called 
the ``Mesoscale data-driven constitutive model'' in \cite{Wang.2018:rd3109}), incorporating microstucture data---such as the fabric tensor $\boldsymbol{F}$ of the 1st kind of rank 2 (motivated in Section~\ref{sc:Wang-Sun-2018}, Figure~\ref{fig:Wang-coordination-number} and Figure~\ref{fig:Wang-prediction-LSTM-microstructure}) defined in Eq.~\eqref{eq:Wang-fabric-tensor} \cite{kanatani1984distribution}---into the training set, whose data were generated by a discrete element assembly as in Figure~\ref{fig:Wang-micro-RVE} depicting a microscale RVE, was important to obtain good network prediction, as mentioned in Section~\ref{sc:Wang-Sun-2018} in relation to Figure~\ref{fig:Wang-prediction-LSTM-microstructure}:
\begin{align}
	\boldsymbol{F}
	:=
	\boldsymbol{A}_F \cdot \boldsymbol{A}_F
	:=
	\frac{1}{N_c}
	\sum_{c=1}^{c=N_c}
	\boldsymbol{n}_c \otimes \boldsymbol{n}_c
	\ ,
	\label{eq:Wang-fabric-tensor}
\end{align}
where $N_c$ is the number of contact points (which is the same as the coordination number $CN$), and $\boldsymbol{n}_c$ the unit normal vector at contact point $c$.

\begin{rem}
	\label{rm:Wang-micro-RVE}
	{\rm
		Even though in Figure~\ref{fig:Wang-RNN-FEM} (row 1) in Section~\ref{sc:Wang-Sun-2018}, the microscale RVE was indicated to be of micron size, but the microscale RVE in Figure~\ref{fig:Wang-micro-RVE} was of size $10 \text{ cm} \times 10 \text{ cm} \times 5 \text{ cm}$ with particles of $0.5 \text{ cm}$ in diameter, many orders of magnitude larger.
		See Remark~\ref{rm:Wang-meso-RVE}.
	}
$\hfill\blacksquare$
\end{rem}

Other microstructure data such as the porosity and the coordination number (or number of contact point), being scalars and did not incorporate directional data like the fabric tensor, did not help to improve the accuracy of the network prediction, as noted in the caption of Figure~\ref{fig:Wang-prediction-LSTM-microstructure}.

To enforce objectivity of constitutive models realized as neural networks, (the history of) principal strains and incremental rotation parameters that describe the orientation of principal directions served as inputs to the network.
Accordingly, principal stresses and incremental rotations for the principal directions were outputs of 
%
% CMES style rewriting
%what~\cite{Wang.2018:rd3109} 
what was
referred to as \emph{Spectral RNNs} \cite{Wang.2018:rd3109}, which preserved objectivity of constitutive models.

%\noindent
%{\color{red} HERE 2020.08.03.}

\begin{figure}[h]
	\centering
	\includegraphics[width=0.95\linewidth]{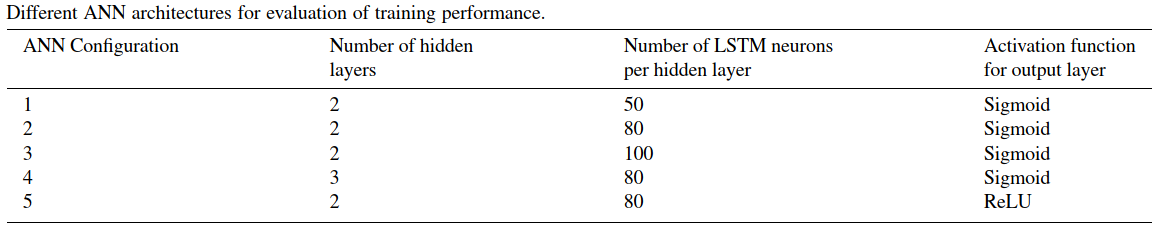}
	\caption{
		\emph{Optimal RNN-LSTM architecture} (Section~\ref{sc:Wang-optimal-RNN}). 
		5 different configurations of RNNs with LSTM units 
		\cite{Wang.2018:rd3109}.
		{\footnotesize (Table reproduced with permission of the authors.)}
	}
	\label{fig:Wang-optimal-RNN-table}
\end{figure}

\begin{figure}[h]
	\centering
	\includegraphics[width=0.45\linewidth]{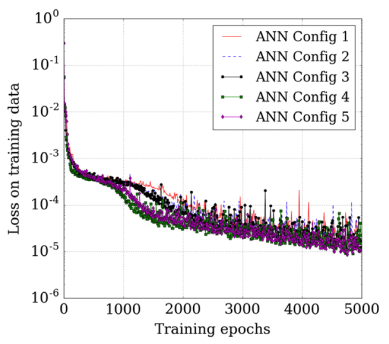}
	\includegraphics[width=0.45\linewidth]{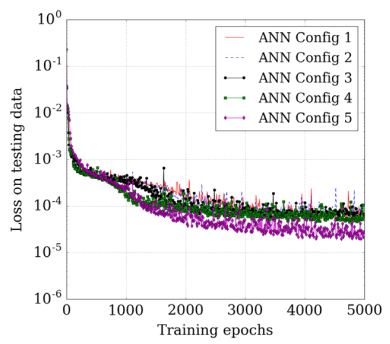}
	\caption{
		\emph{Optimal RNN-LSTM architecture} (Section~\ref{sc:Wang-optimal-RNN}). 
		Training error and test errors for 5 different configurations of RNN with LSTM units, see Figure~\ref{fig:Wang-optimal-RNN-table}
		\cite{Wang.2018:rd3109}
		{\footnotesize (Figure reproduced with permission of the authors.)}
	}
	\label{fig:Wang-optimal-RNN}
\end{figure}

\subsubsection{Optimal RNN-LSTM architecture}
\label{sc:Wang-optimal-RNN}
Using the same discrete element assembly of microscale RVE in Figure~\ref{fig:Wang-micro-RVE} to generate data, 
%
% CMES style rewriting
the authors of 
\cite{Wang.2018:rd3109} tried out 5 different configurations of RNNs with LSTM units, with 2 or 3 hidden layers, 50 to 100 LSTM units (Figure~\ref{fig:Wang-LSTM_cell}, Section~\ref{sc:Wang-Sun-2018}, and Figure~\ref{fig:our-lstm_cell} for the detailed original LSTM cell), and either logistic sigmoid or ReLU as activation function, Figure~\ref{fig:Wang-optimal-RNN-table}.  

Configuration 1 has 2 hidden layers with 50 LSTM units each, and with the logistic sigmoid as activation function.
Config 2 is similar, but with 80 LSTM units per hidden layer.
Config 3 is similar, but with 100 LSTM units per hidden layer.
Config 4 is similar to Config 2, but with 3 hidden layers.
Config 5 is similar to Config 4, but with ReLU activation function.

\begin{figure}[h]
	\centering
	\includegraphics[width=0.45\linewidth]{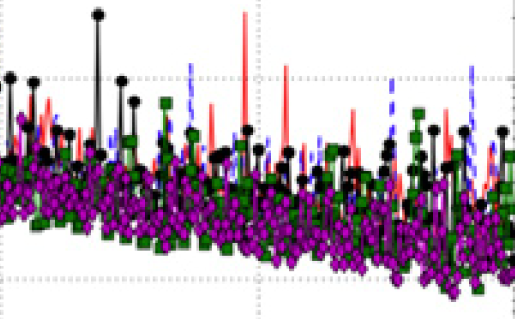}
	\includegraphics[width=0.45\linewidth]{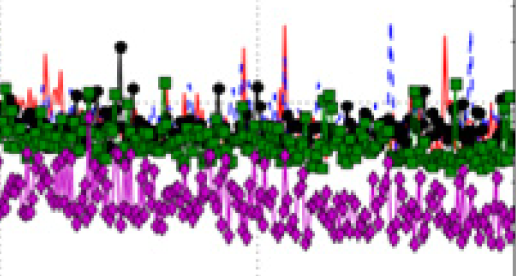}
	\caption{
		\emph{Optimal RNN-LSTM architecture} (Section~\ref{sc:Wang-optimal-RNN}). 
		Training error (a) and testing error (b),
		close-up views of Figure~\ref{fig:Wang-optimal-RNN} from epoch 3000 to epoch 5000:
		Config 5 (purple line with dots) was optimal, with smaller errors than those of Config 2 (blue dashed line).  
		%Moreover, ReLU is computationally more efficient than logistic sigmoid.
		See Figure~\ref{fig:Wang-optimal-RNN-table} for config details
		\cite{Wang.2018:rd3109}  
		{\footnotesize (Figure reproduced with permission of the authors.)}
	}
	\label{fig:Wang-optimal-RNN-zoom}
\end{figure}

The training error and test error obtained from using these 5 configurations are shown in Figure~\ref{fig:Wang-optimal-RNN}.  The zoomed-in views of the training error and test error from epoch 3000 to epoch 5000 in Figure~\ref{fig:Wang-optimal-RNN-zoom} show that Config 5 was the optimal with smaller errors, and since ReLU would be computationally more efficient than the logistic sigmoid. 
But 
%
% CMES style rewriting
%\cite{Wang.2018:rd3109} selected, however, 
Config 2 was, however, selected in \cite{Wang.2018:rd3109}, whose authors cited that the discrepancy was ``not significant'', and that Config 2 gave ``good training and prediction performances''.

\begin{rem}
	{\rm
		The above search for an optimal network architecture is similar to searching for an appropriate degree of a polynomial function for a best fit, avoiding overfit and underfit, over a given set of data points in a least-square curve fittings.  See Figure~\ref{fig:overfit} in Section~\ref{sc:adam-criticism} for an explanation of underfit and overfit, and Figure~\ref{fig:Oishi-tab1} in Section~\ref{sc:Oishi-2} for a similar search of an optimal network for numerical integration by ANN. 
	}
	$\hfill\blacksquare$
\end{rem}

\begin{rem}
	\label{rm:Wang-layer-number} 
	{\rm
		Referring to Remark~\ref{rm:layer-definitions} and the neural network in Figure~\ref{fig:Wang-RNN-FEM} and to our definition of action depth as total number of action layers $L$ in Figure~\ref{fig:network3b} in Section~\ref{sc:graphical-representation} and Remark~\ref{rm:depth-definitions} in Section~\ref{sc:depth-size}, it is clear that a network layer in \cite{Wang.2018:rd3109} is a state layer, i.e., an input matrix $\byp{\ell}$, with $\ell = 0, \ldots, L$.
		Thus all configs in Figure~\ref{fig:Wang-optimal-RNN-table} with 2 hidden layers have an action depth of $L=3$ layers and a state depth of $L+1=4$ layers, whereas Config 4 with 3 hidden layers has an action depth of $L=4$ layers and a state depth of $L+1=5$ layers. 
		On the other hand, since RNNs with LSTM units were used, in view of Remark~\ref{rm:depth-of-RNN}, these networks were equivalent to ``very deep feedforward networks''. 
		%See also Remark~\ref{rm:layer-definitions}.
	}
$\hfill\blacksquare$
\end{rem}

\begin{rem}
	\label{rm:Wang-optimal-RNN-LSTM-micro-meso-RVE}
	{\rm 
		%
		% CMES style rewriting
%		\cite{Wang.2018:rd3109} used 
		The same \emph{selected} architecture of RNN with LSTM units on both the microscale RVE (Figure~\ref{fig:Wang-micro-RVE}, Figure~\ref{fig:Wang-micro-RVE-training-data}) and the mesoscale RVE (Figure~\ref{fig:Wang-mesoscale-RVE-imposed-displacements}, Figure~\ref{fig:Wang-mesoscale-RVE}) was used to produce the mesoscale RNN with LSTM units (``Mesoscale data-driven constitutive model'') and the macroscale RNN with LSTM units (``Macroscale data-driven constitutive model''), respectively \cite{Wang.2018:rd3109}.
	}
$\hfill\blacksquare$
\end{rem}

%\noindent
%{\color{red} HERE 2020.08.04.}

\subsubsection{Dual-porosity dual-permeability governing equations}
\label{sc:Wang-balance-equations}
\begin{figure}[h]
	\centering
	\includegraphics[width=0.45\linewidth]{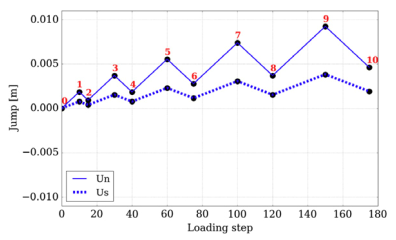}
	\includegraphics[width=0.50\linewidth]{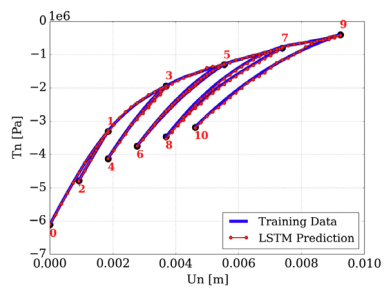}
	\caption{
		\emph{Mesoscale RNN with LSTM units.  Traction-separation law} (Sections~\ref{sc:Wang-optimal-RNN}, \ref{sc:Wang-strong-discontinuities}). 
		\emph{Left:}
		Sequence of imposed displacement jumps on microscale RVE (Figure~\ref{fig:Wang-micro-RVE}), normal (solid line) and tangential (dotted line, with $u_s \equiv u_m$ in Figure~\ref{fig:Wang-micro-RVE}, center). 
		\emph{Right:}
		Normal traction vs. normal displacement.  Cyclic loading and unloading. Microscale RVE training data (blue) vs. Mesoscale RNN with LSTM prediction (red, Section~\ref{sc:Wang-optimal-RNN}), with mean squared error $3.73 \times 10^{-5}$.
		See also Figure~\ref{fig:Wang-results_macroscale_RNN} on the macroscale RNN with LSTM units
		\cite{Wang.2018:rd3109}  
		{\footnotesize (Figure reproduced with permission of the authors.)}
	}
	\label{fig:Wang-micro-RVE-training-data}
\end{figure}

The governing equations for media with dual-porosity dual-permeability in this section are only applied to macroscale (field-size) simulations, i.e., not for simulations with the microscale RVE (Figure~\ref{fig:Wang-micro-RVE}, Figure~\ref{fig:Wang-micro-RVE-training-data}) and with the mesoscale RVE (Figure~\ref{fig:Wang-mesoscale-RVE-imposed-displacements}, Figure~\ref{fig:Wang-mesoscale-RVE}).

For field-size simulations, assuming stationary conditions, small deformations, incompressibility, no mass exchange among solid and fluid constituents, the problem is governed by the balance of linear momentum and the balance of fluid mass in micropores and macropores, respectively. 
The displacement field of the solid $\uv$, the micropore pressure $\pmicro$ and the macropore pressure $\pmacro$ constitute the primary unknowns of the problem.
The (total) Cauchy stress tensor $\cauchy$ is the sum of the effective stress tensor $\cauchyeff$ on the solid skeleton and the pore fluid pressure $p_f$, which was in between $\pmacro$ and $\pmicro$, and assumed to be a convex combination of the latter two pore pressures \cite{Wang.2018:rd3109}, 
\begin{equation}
	\cauchy = \cauchyeff - p_f \I \Rightarrow
	\cauchyeff = \cauchy + \left\{\fracmacro \, \pmacro + \left( 1- \fracmacro \right) \pmicro \right\} \I ,
	\label{eq:effective-stress}
\end{equation}
where $\fracmacro$ denoted the ratio of macropore volume over total pore volume, and $\left( 1- \fracmacro \right)$ the ratio of micropore volume over total pore volume.  

The \emph{balance of linear momemtum} equation is written as
\begin{align}
	\text{div } \cauchy + \rho \bgrav 
	= 
	\czero (\bvtm - \bvtM)
	\ ,
	\label{eq:Wang-balance-linear-momentum}
\end{align}
where 
$\rho$ is the total mass density, 
$\bgrav$ the acceleration of gravity,
\begin{align}
	\bvtM := \bvM - \bvelo
	\ , \quad
	\bvtm := \bvm - \bvelo
	\ ,
	\label{eq:Wang-relative-velocities}
\end{align}
%$\bvtM = \bvM - \bvelo$ the velocity of the macropore fluid \emph{relative} to the velocity $\bvelo$ of the solid, similarly for $\bvtm = \bvm - \bvelo$.
where $\bvelo$ is the velocity of the solid skeleton, 
$\bvM$ and $\bvm$ are the fluid velocities in the macropores and in the micropores, respectively.
%the fluid relative velocities in the macropores and in the micropores, respectively, with respect to the solid skeleton.
In Eq.~\eqref{eq:Wang-balance-linear-momentum}, the coefficient $\czero$ of fluid mass transfer between macropores and micropores was assumed to obey the ``semi-empirical'' relation \cite{Wang.2018:rd3109}:
\begin{equation}
	\czero = \frac{\bar \alpha}{\mu_f} \left( \pmacro - \pmicro \right) ,
	\label{eq:Wang-fluid-transfer}
\end{equation}
with $\bar \alpha$ being a parameter that characterized the interface permeability between macropores and micropores, and $\mu_f$ the dynamic viscosity of the fluid.

In the 1-D case, Darcy's law is written as
\begin{align}
    \frac{q(x,t)}{\rho_f}
    =
	v(x, t) 
	= - \frac{k}{\mu_f} \frac{\partial p(x,t)}{\partial x}
	\ ,
	\label{eq:darcy-law-1}
\end{align}
where $q$ is the fluid mass flux ($\text{kg} / (\text{m}^2 \text{s})$),
$\rho_f$ the fluid mass density ($\text{kg / m}^3$),
$v$ the fluid velocity ($\text{m/s}$),
$k$ the medium permeability ($\text{m}^2$),
$\mu_f$ the fluid dynamic viscosity ($\text{Pa} \cdot s = \text{N s} / \text{m}^2$),
$p$ the pressure ($\text{N} / \text{m}^2$), and
$x$ the distance ($\text{m}$).

\begin{rem}
	\label{rm:Wang-alpha-bar-1}
	Dimension of $\bar \alpha$ in Eq.~\eqref{eq:Wang-fluid-transfer}.
	{\rm 
		Each term in the balance of linear momentum Eq.~\eqref{eq:Wang-balance-linear-momentum} has force per unit volume ($F/L^3$) as dimension, which is therefore the dimension of the right-hand side of Eq.~\eqref{eq:Wang-balance-linear-momentum}, where  $\czero$ appears.  As a result, $\czero$ has the dimension of mass ($M$) per unit volume ($L^3$) per unit time ($T$):
		\begin{align}
			\left[ \czero (\bvtm - \bvtM) \right] 
			= \frac{F}{L^3} = \frac{M}{L^2 T^2}
			\Rightarrow
			\left[ \czero \right] = \frac{M}{L^2 T^2} \frac{T}{L} = \frac{M}{L^3 T}
			\ .
			\label{eq:Wang-c0-dimension-1}
		\end{align}
		Another way to verify is to identify the right-hand side of Eq.~\eqref{eq:Wang-balance-linear-momentum} with the usual inertia force per unit volume:
		\begin{align}
			\left[ \czero (\bvtm - \bvtM) \right]
			=
			\left[
			\rho \frac{\partial \bvelo}{\partial t} 
			\right]
			\Rightarrow
			\left[ \czero \right] \left[ \bvelo \right]
			=
			\left[ \rho \right] \frac{\left[ \bvelo \right]}{T}
			\Rightarrow
			\left[ \czero \right] 
			=
			\frac{\left[ \rho \right]}{T}
			=
			\frac{M}{L^3 T}
			\ .
			\label{eq:Wang-c0-dimension-2}
		\end{align}
		The empirical relation Eq.~\eqref{eq:Wang-fluid-transfer} adopted 
		% 
		% CMES style rewriting
%		by  
		in \cite{Wang.2018:rd3109}
		implies that the dimension of $\bar \alpha$ was
		\begin{align}
			\left[ \bar \alpha \right] 
			=
			\frac{\left[ \czero \right] \left[ \mu_f \right]}{\left[ (p_M - p_m) \right]}
			=
			\frac{\left( M / (L^3 T) \right) \left( F T / L^2 \right)}{F / L^2}
			=
			\frac{M}{L^3}
			\ ,
			\label{eq:Wang-dimension-alpha-bar}
		\end{align}
		i.e., mass density, whereas permeability has the dimension of area ($L^2$), as seen in Darcy's law Eq.~\eqref{eq:darcy-law-1}.
		It is not clear why
		%
		% CMES style rewriting 
%		\cite{Wang.2018:rd3109} wrote 
		it is written in \cite{Wang.2018:rd3109}
		that $\bar \alpha$ characterized the ``interface permeability between the macropores and the micropores''.
		A reason could be that the ``semi-empirical'' relation Eq.~\eqref{eq:Wang-fluid-transfer} was assumed to be analogous to Darcy's law Eq.~\eqref{eq:darcy-law-1}.  
		Moreover, as a result of Eq.~\eqref{eq:Wang-dimension-alpha-bar}, the dimension of $\mu_f / \bar \alpha$ is therefore the same as that of the kinematic viscosity $\nu_f = \mu_f / \rho$.
		If $\bar \alpha$ had the dimension of permeability, then
		the choice of the right-hand side of the balance of linear momentum Eq.~\eqref{eq:Wang-balance-linear-momentum} was inconsistent, dimensionally speaking.
		See Remark~\ref{rm:Wang-alpha-bar-2}.
	}
	$\hfill\blacksquare$
\end{rem}

\begin{figure}[h]
	\centering
	\includegraphics[width=0.8\linewidth]{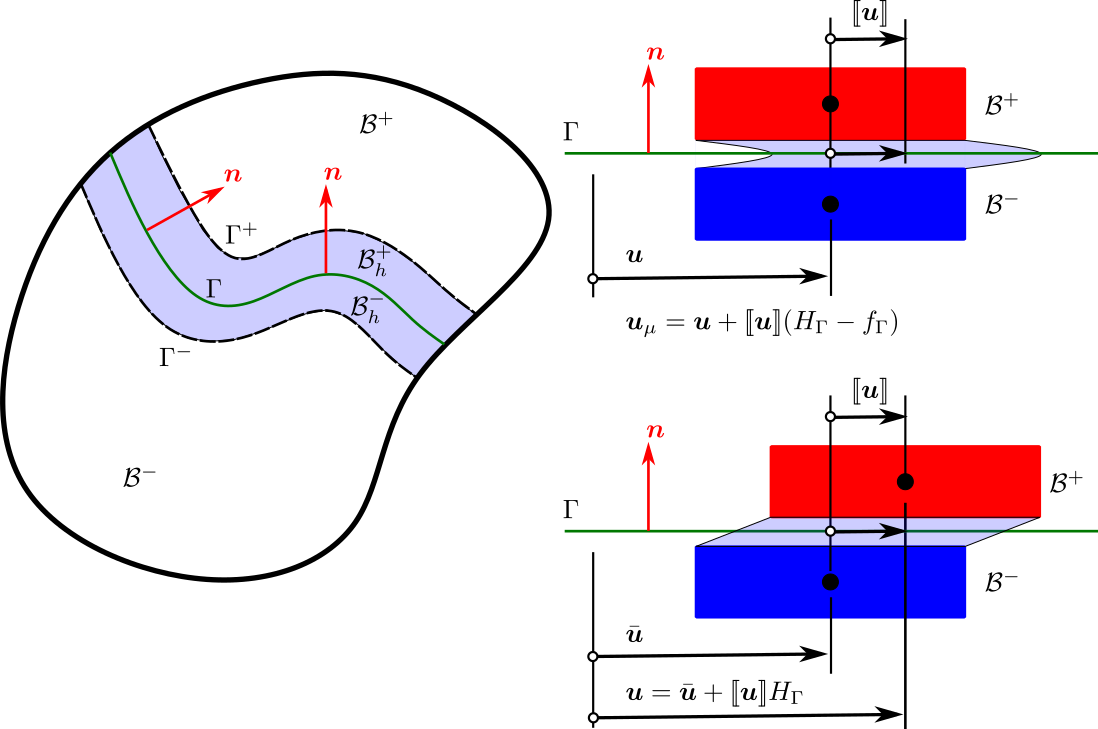}
	\caption{
		\emph{Continuum with embedded strong discontinuity} (Section~\ref{sc:Wang-strong-discontinuities}).
		Domain $\domain = \domp + \cup \domp -$ with embedded discontinuity surface $\surface$, running through the middle of a narrow band (light blue) $\doms h = ( \domps{+}{h} \cup \domps{-}{h} ) \subset \domain$ between the parallel surfaces $\surfp +$ and $\surfp -$. 
		Objects behind $\Gamma$ in the negative direction of the normal $\bnorm$ to $\Gamma$ are designated with the minus sign, and those in front of $\Gamma$ with the plus sign. 
		The narrow band $\doms h$ represents an embedded strong discontinuity, as shown in the mesoscale RVE in Figure~\ref{fig:Wang-mesoscale-RVE}, where the discretized strong discontinuity zones were a network of straight narrow bands.
		\emph{Top right:} No sliding, tnterpretation of $\uvloc = \uv + \uvjump (H_\Gamma - f_\Gamma)$ in \cite{Wang.2018:rd3109}.
		\emph{Bottom right:} Sliding, interpretation of $\uv = \bar \uv + \jump \uv H_\Gamma$ in \cite{borja2000finite}. 
		%See also \cite{borja2000finite}.
		%\cite{Wang.2018:rd3109}  
		%{\footnotesize (Figure reproduced with permission of the authors.)}
	}
	\label{fig:Wang-domain-discontinuity}
\end{figure}

The (local) porosity $\phi$ is the ratio of the void volume $dV_v$ over the total volume $dV$.  Within the void volume, let $\psi$ be the percentage of macropores, and $(1 - \psi)$ the percentage of micropores; we have:
\begin{align}
	\phi = \frac{dV_v}{dV} \ , \quad
	\psi = \frac{dV_M}{dV_v} \ , \quad
	1 - \psi = \frac{dV_m}{dV_v} \ .
\end{align}
The absolute macropore flux $\bqM$ and the absolute micropore flux $\bqm$ are defined as follows:
\begin{align}
	\esM = \rho_f \phi \psi \ , \quad
	\bqM = \esM \bvM \ , \quad
	\esm = \rho_f \phi (1 - \psi) \ , \quad
	\bqm = \esm \bvm
	\ ,
\end{align}
There is a conservation of fluid transfer between the macropores and the micropores across any closed surface $\Gamma$, i.e.,
\begin{align}
	\int_\Gamma 
	(\bqM + \bqm) \, \dotprod \, \bnorm \,
	d \Gamma
	= 0
	\Rightarrow
	\diver (\bqM + \bqm) = 0
	\ .
	\label{eq:Wang-conservation}
\end{align}

Generalizing Eq.~\eqref{eq:darcy-law-1} to 3-D,
Darcy's law in tensor form governs the fluid mass fluxes $\fluxmacro$, $\fluxmicro$ is written as
\begin{equation}
	\bqtM = \esM \bvtM 
	=
	%\fluxmacro = 
	- \rho_f \frac{\permeabmacro}{\mu_f} \cdot \left( \nabla \pmacro - \rho_f \gv \right) , \quad
	\bqtm = \esM \bvtm 
	=
	%\fluxmicro = 
	- \rho_f \frac{\permeabmicro}{\mu_f} \cdot \left( \nabla \pmicro - \rho_f \gv \right) ,
	\label{eq:darcy-law-2}
\end{equation}
where $\permeabmacro$, $\permeabmicro$ denote the permeability tensors on the respective scales and $\gv$ is the gravitational acceleration.

From Eq.~\eqref{eq:Wang-conservation}, assuming that 
\begin{align}
	\diver \bqM = - \diver \bqm = \czero
	\ ,
	\label{eq:Wang-conservation-2}
\end{align}
where $\czero$ is given in Eq.~\eqref{eq:Wang-fluid-transfer}, then two more governing equations are obtained:
\begin{align}
	&
	\diver \bqM = \esM \diver \bvelo + \diver \bqtM = \czero
	\label{eq:Wang-conservation-2a}
	\ ,
	\\
	&
	\diver \bqm = \esm \diver \bvelo + \diver \bqtm = - \czero
	\ ,
	\label{eq:Wang-conservation-2b}
\end{align}
agreeing with \cite{Wang.2018:rd3109}.

\begin{rem}
	\label{rm:Wang-alpha-bar-2}
	{\rm
		It can be verified from Eq.~\eqref{eq:Wang-conservation-2} that the dimension of $\czero$ is
		\begin{align}
			\left[ \czero \right]
			=
			\left[ \diver \bqM \right]
			=
			\frac{\left[ \bqM \right]}{L}
			=
			\frac{\left[ \rho \right] \left[ \bvelo \right]}{L}
			=
			\frac{M}{L^3 T}
			\ ,
		\end{align}
		agreeing with Eq.~\eqref{eq:Wang-c0-dimension-2}.  In view of Remark~\ref{rm:Wang-alpha-bar-1}, for all three governing Eq.~\eqref{eq:Wang-balance-linear-momentum}, Eqs.~\eqref{eq:Wang-conservation-2a}-\eqref{eq:Wang-conservation-2b} to be dimensionally consistent, $\bar \alpha$ in Eq.~\eqref{eq:Wang-fluid-transfer} should have the same dimension as mass density, as indicated in Eq.~\eqref{eq:Wang-dimension-alpha-bar}.
		Our consistent notation for fluxes $(\bqM, \bqm)$ and $(\bqtM, \bqtm)$ differ in meaning compared to \cite{Wang.2018:rd3109}.
	}
	\phantom{space}$\hfill\blacksquare$
\end{rem}
\begin{figure}[h]
	\centering
	\includegraphics[width=0.9\linewidth]{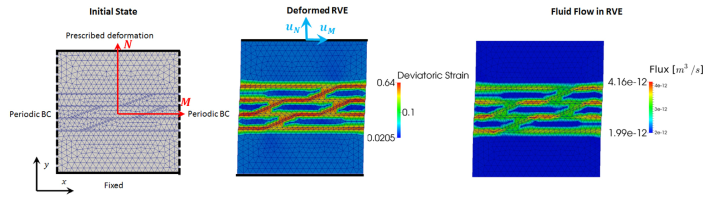}
	\caption{
		\emph{Mesoscale RVE} (Sections~\ref{sc:Wang-optimal-RNN}, \ref{sc:Wang-strong-discontinuities}). A 2-D domain of size $1 \text{ m} \times 1 \text{ m}$ (Remark~\ref{rm:Wang-meso-RVE}).
		See Figure~\ref{fig:Wang-RNN-FEM} (row 2, left), and also Figure~\ref{fig:Wang-domain-discontinuity} for a conceptual representation.
		Embedded strong discontinuity zones where damage occurred formed a network of straight narrow bands surrounded by elastic material.  Both the strong discontinuity narrow bands and the elastic domain were discretized into finite elements.
		Imposed displacements $(\uv_N , \uv_M)$, with $\uv_S \equiv \uv_N$, at the top (center).
		See Figure~\ref{fig:Wang-mesoscale-RVE} for the deformation (strains and displacement jumps)
		\cite{Wang.2018:rd3109}.  
		{\footnotesize (Figure reproduced with permission of the authors.)}
	}
	\label{fig:Wang-mesoscale-RVE-imposed-displacements}
\end{figure}

\begin{rem}
	\label{rm:Wang-field-size-simulation}
	{\rm 
		For field-size simulations, the above equations do not include the changing size of the pores, which were assumed to be of constant size, and thus constant porosity, in \cite{Wang.2018:rd3109}.  As a result, the collapse of the pores that leads to nonlinearity in the stress-strain relation observed in experiments (Figure~\ref{fig:Majella-stress-strain}) is not modelled in \cite{Wang.2018:rd3109}, where the nonlinearity essentially came from the embedded strong discontinuities (displacement jumps) and the associated traction-separation law obtained from DEM simulations using the micro RVE in Figure~\ref{fig:Wang-micro-RVE} to train the meso RNN with LSTM; see Section~\ref{sc:Wang-strong-discontinuities}.
		See also Remark~\ref{rm:Wang-no-nonlinear-stress-strain}.
	}
\phantom{space}$\hfill\blacksquare$
\end{rem}

\subsubsection{Embedded strong discontinuities, traction-separation law}
\label{sc:Wang-strong-discontinuities}
%
% discontinuities
\begin{figure}[h]
	\centering
	\includegraphics[width=0.9\linewidth]{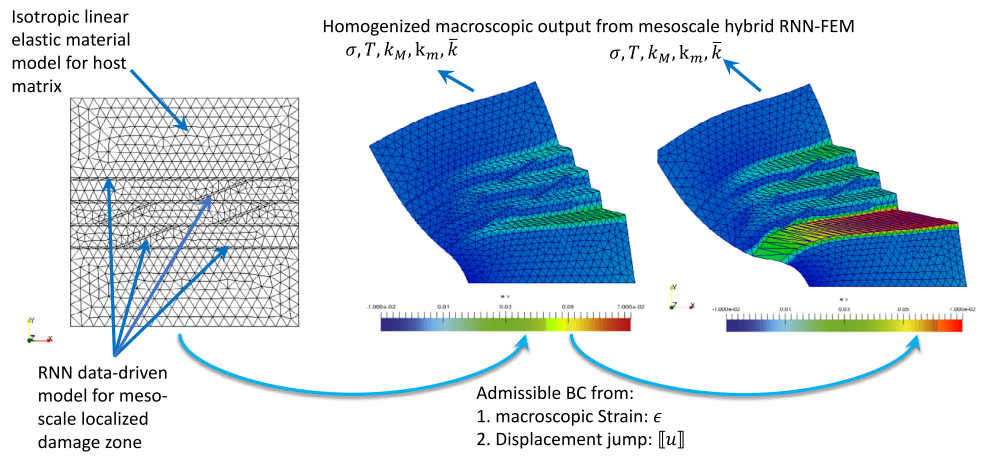}
	\caption{
		\emph{Mesoscale RVE} (Section~\ref{sc:Wang-optimal-RNN}). Strains and displacement jumps
		\cite{Wang.2018:rd3109}  
		{\footnotesize (Figure reproduced with permission of the authors.)}
	}
	\label{fig:Wang-mesoscale-RVE}
\end{figure}

Strong discontinuities are embedded at both the mesocale and the macroscale. 
Once a fault is formed through cracks in rocks, it could become inactive (no further slip) due to surrounding stresses, friction between the two surfaces of a fault, cohesive bond, and low fluid pore pressure.  A fault can be reactivated (onset of renewed fault slip) due to changing stress state, loosened fault cohesion, high fluid pore-pressure.  Conventional models for fault reactivation are based on effective stresses and Coulomb law \cite{sibson1985note}:\begin{align}
	\tau \ge \tau_p 
	= C + \mu \stress^\prime 
	= C + \mu (\stress - p) 
	\ ,
	\label{eq:coulomb-law} 
\end{align}
where
$\tau$ is the shear stress along the fault line, 
$\tau_p$ the critical shear stress for the onset of fault reactivation,
$C$ the cohesion strength, 
$\mu$ the coefficient of friction,
$\stress^\prime$ the effective stress normal to the fault line,
$\stress$ the normal stress, and
$p$ the fluid pore pressure.  
%
% CMES style rewriting
The authors of 
\vphantom{\cite{passelegue2018fault}}\cite{passelegue2018fault} demonstrated that increase in fluid injection rate led to increase in peak fluid pressure and, as a result, fault reactivation, as part of a study on why there was an exponential increase in seismic activities in Oklahoma, due to wastewater injection for use in hydraulic fracturing (i.e., fracking) \cite{kuchment2019even}. 

But criterion Eq.~\eqref{eq:coulomb-law} involves only stresses, with no displacement, and thus cannot be used to quantify the amount of fault slip.  To allow for quantitative modeling of fault slip, or displacement jump, in a displacement-driven FEM environment, the so-called ``cohesive traction-separation laws,'' expressing traction (stress) vector on fault surface as a function of fault slip, similar to those used in modeling cohesive zone in nonlinear fracture mechanics \cite{park2011cohesive}, is needed.  But these classical ``cohesive traction-separation law'' are not appropriate for handling loading-unloading cycles.

To model a continuum with displacement jumps, i.e., embedded strong discontinuities, 
%
% CMES style rewriting
%\cite{Wang.2018:rd3109} represent 
the traction-separation law was represented in \cite{Wang.2018:rd3109} as
\begin{align}
	\bT ( \jump \uv) 
	= 
	\bstress^\prime ( \jump \uv) \dotprod \bnorm
	\ ,
	\label{eq:Wang-traction-separation}
\end{align}
where $\bT$ is the traction vector on the fault surface, 
$\uv$ the displacement field,
$\jump \cdot$ the jump operator, making
$\jump \uv$ the displacement jump (fault slip, separation),
$\bstress^\prime$ the effective stress tensor at the fault surface,
$\bnorm$ the normal to the fault surface.
To obtain the traction-separation law represented by the function $\bT ( \jump \uv)$, a neural network can be used provided that data are available for training and testing.

%
% CMES style rewriting
It was assumed in
\cite{Wang.2018:rd3109} that cracks were pre-existing and did not propagate (see the mesoscale RVE in Figure~\ref{fig:Wang-mesoscale-RVE}), then set out to use the microscale RVE in Figure~\ref{fig:Wang-micro-RVE} to generate training data and test data for the mesocale RNN with LSTM units, called the ``Mesoscale data-driven constitutive model'', to represent the traction-separation law for porous media.  Their results are shown in Figure~\ref{fig:Wang-micro-RVE-training-data}.

\begin{rem}
	\label{rm:Wang-not-realistic}
	{\rm
		The microscale RVE in Figure~\ref{fig:Wang-micro-RVE} did not represent any real-world porous rock sample such the Majella limestone with macroporosity $11.4\% \approx 0.1$ and microporosity $19.6\% \approx 0.2$ shown in Figure~\ref{fig:Majella-pore-structure} in Section~\ref{sc:Wang-Sun-2018}, but was a rather simple assembly of mono-disperse (identical) solid spheres with no information on size and no clear contact force-displacement relation; see \cite{zhang2007accurate}.  Another problem was that realistic porosity of 0.1 or 0.2 could not be achieved with this microscale RVE, yielding a porosity above 0.3, which was the total porosity (= macroporosity + microporosity) of the highly porous Majella limestone. 
		A goal of \cite{Wang.2018:rd3109} was only to demonstrate the methodology, not presenting realistic results.
	}
$\hfill\blacksquare$
\end{rem}

\begin{figure}[h]
	\centering
	\includegraphics[width=0.9\linewidth]{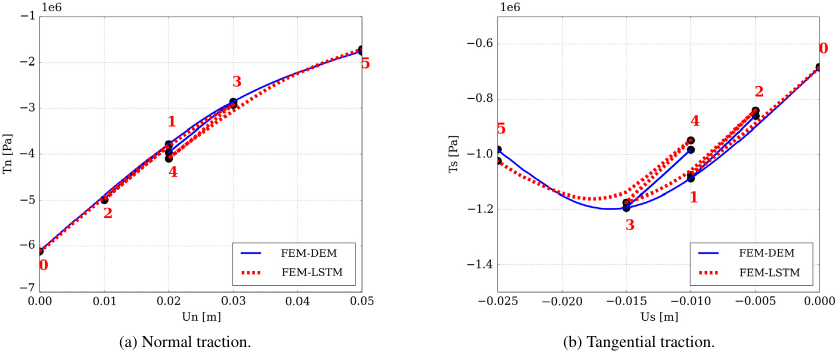}
	\caption{
		\emph{Mesoscale RVE} (Section~\ref{sc:Wang-strong-discontinuities}).
		Validation of coupled FEM and RNN with LSTM units (FEM-LSTM, red dotted line) against coupled FEM and DEM (FEM-DEM, blue line) to analyze the mesoscale RVE in Figure~\ref{fig:Wang-mesoscale-RVE} under a sequence of imposed displacement jumps at the top (represented by numbers).
		(a) Normal traction (Tn) vs normal displacement (Un).
		(b) Tangential traction (Ts) vs tangential displacement (Us) 
		\cite{Wang.2018:rd3109}.
%		{\color{red} [NOTE: Figure25, pdf p.34.]}
		{\footnotesize (Figure reproduced with permission of the authors.)}
	}
	\label{fig:Wang-results_mesoscale_RNN}
\end{figure}

\begin{rem}
	\label{rm:Wang-meso-RVE}
	{\rm
		Even though in Figure~\ref{fig:Wang-RNN-FEM} (row 2) in Section~\ref{sc:Wang-Sun-2018}, the mesoscale RVE was indicated to be of centimeter size, but the mesoscale RVE in Figure~\ref{fig:Wang-mesoscale-RVE} was of size $1 \text{ m} \times 1 \text{ m}$, many orders of magnitude larger.
		See Remark~\ref{rm:Wang-micro-RVE}.
	}
	$\hfill\blacksquare$
\end{rem}

%\noindent
%{\color{red} [NOTE 2020.08.16.  it is better to do the results first (Section 5 on ``Numerical experiments'' in \cite{Wang.2018:rd3109}) to see whether there is a need to have this section on strong discontinuities.  ENDNOTE]}

To analyze the mesoscale RVE in Figure~\ref{fig:Wang-mesoscale-RVE} (Figure~\ref{fig:Wang-DEM-FEM-three-scales}, center) and the macroscale (field-size) model (Figure~\ref{fig:Wang-DEM-FEM-three-scales}, right) by finite elements, both with embedded strong discontinuities (Figure~\ref{fig:Wang-domain-discontinuity}),
%
% CMES style rewriting
the authors of 
\cite{Wang.2018:rd3109} adopted a formulation that looked similar to \cite{borja2000finite} to represent strong discontinuities, which could result from fractures or shear bands, by the \emph{local} displacement field $\uvloc$ as\footnote{
	The subscript $\mu$ in $\uvloc$ probably meant ``micro'', and was used to designate the \emph{local} nature of $\uvloc$.
}
\begin{equation}
	\uvloc = \uv + \uvjump \left( H_\Gamma - f_\Gamma \right)
	\ , \quad
	\bar \uv := \uv - \uvjump f_\Gamma
	\Rightarrow
	\uvloc = \bar \uv + \uvjump H_\Gamma
	\ ,
	\label{eq:Wang-displacement-jump}
\end{equation}
which differs from the global smooth displacement field $\uv$ by the displacement jump vector $\uvjump$ across the singular surface $\Gamma$ that represents a discontinuity, multiplied by the function $(H_\Gamma - f_\Gamma)$, where $H_\Gamma$ is the Heaviside function, such that $H_\Gamma = 0$ in $\domp{-}$  and $H_\Gamma = 1$ in $\domp{+}$, and $f_\Gamma$ a smooth ramp function equal to zero in $({\domp -} - {\domps - h})$ and going smoothly up to 1 in $({\domp +} - {\domps + h})$, as defined in \cite{borja2000finite}.  
%\sout{The jump $\uvjump$ is smoothed using the Heaviside function $H_\Gamma$ and a smooth ramp function $f_\Gamma$}.
So Eq.~\eqref{eq:Wang-displacement-jump} means that the displacement field $\uv$ only had a smooth ``bump'' with support being the band $\doms h$, as a result of a local displacement jump, with \emph{no} fault sliding, as shown in the top right subfigure in Figure~\ref{fig:Wang-domain-discontinuity}.

\begin{figure}[h]
	\centering
	\includegraphics[width=0.9\linewidth]{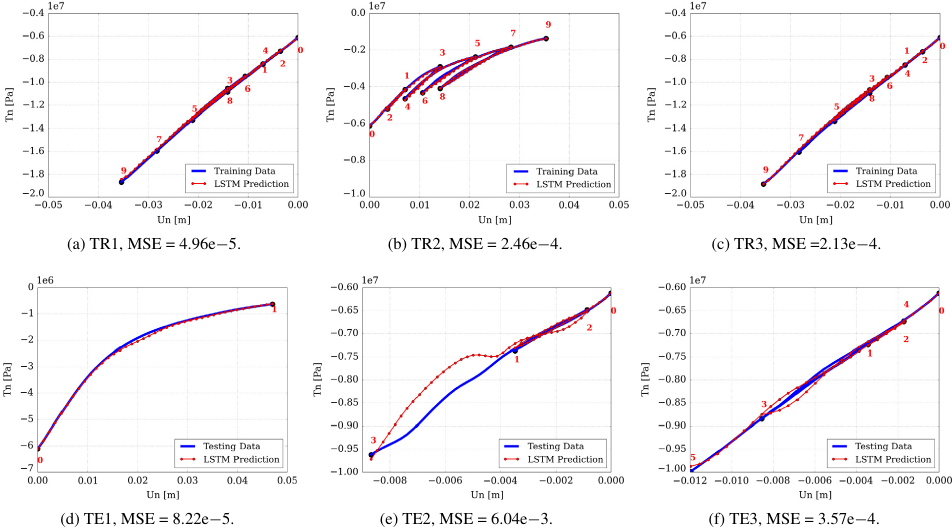}
	\caption{
		\emph{Macroscale RNN with LSTM units} (Section~\ref{sc:Wang-strong-discontinuities}).
		%		Imposed displacement jumps on mesoscale RVE.
		Normal traction (Tn) vs imposed displacement jumps (Un) on mesoscale RVE (Figure~\ref{fig:Wang-mesoscale-RVE-imposed-displacements}, Figure~\ref{fig:Wang-mesoscale-RVE}).
		\emph{Blue:} Training data (TR1-TR3) and test data (TE1-TE3) from the mesoscale FEM-LSTM model, where numbers indicate the sequence of loading-unloading steps similar to those in Figure~\ref{fig:Wang-micro-RVE-training-data} for the mesoscale RNN with LSTM units.
		\emph{Red:} Corresponding predictions of the trained macroscale RNN with LSTM units.
		The mean squared error (MSE) was used as loss function
		\cite{Wang.2018:rd3109}.
		{\footnotesize (Figure reproduced with permission of the authors.)}
	}
	\label{fig:Wang-results_macroscale_RNN}
\end{figure}

%
% CMES style rewriting
%\cite{borja2000finite} started from 
Eq.~\eqref{eq:Wang-displacement-jump}$_3$ was the starting point in \cite{borja2000finite}, but \emph{without} using the definition of $\bar \uv$ in Eq.~\eqref{eq:Wang-displacement-jump}$_2$:\footnote{
	\cite{borja2000finite}, Eq.~(2.1).
}
\begin{align}
	\uv = \bar \uv + \uvjump H_\Gamma
	\Rightarrow
	\dot \uv = \dot{\bar \uv} + \jump{\dot \uv} H_\Gamma
	\ , 
	\label{eq:Borja-displacement-jump}
\end{align}
with
$\uv$ being the total displacement field (including the jump),  
$\bar \uv$ the smooth part of $\uv$, and the overhead dot representing the rate or increment.  As such, Eq.~\eqref{eq:Borja-displacement-jump} can describe the sliding between $\doms -$ and $\doms +$, as shown in the bottom right subfigure in Figure~\ref{fig:Wang-domain-discontinuity}.
Assuming that the jump $\jump \uv$ has zero gradient in $\doms +$, take the gradient of rate form in Eq.~\eqref{eq:Borja-displacement-jump}$_2$ and symmetrize to obtain the small-strain rate\footnote{
	\cite{borja2000finite}, Eq.~(2.2).
}
\begin{align}
	(\nabla \jump {\uv} = 0 
	\text{ and }
	\nabla H_\Gamma = \bnorm \delta_\Gamma)
	\Rightarrow
	\dot {\boldsymbol{\epsilon}} 
	= 
	\text{sym} (\nabla \dot \uv) + \text{sym} (\jump{\dot \uv} \otimes \bnorm \delta_\Gamma)
	\ ,
	\text{ with }
	\text{sym} (\bv) := \frac12 (\bv + \bv^T)
	\ .
	\label{eq:Borja-strain-rate}
\end{align}
Later in \cite{borja2000finite}, an equation that looked similar to Eq.~\eqref{eq:Wang-displacement-jump}$_1$, but in rate form, was introduced:\footnote{
	\cite{borja2000finite}, Eq.~(2.15).
}
\begin{align}
	\dot \uv 
	= \dot{\bar \uv} + \jump{\dot \uv} H_\Gamma
	= (\dot{\bar \uv} + \jump{\dot \uv} f_\Gamma) + \jump{\dot \uv} (H_\Gamma - f_\Gamma)
	= \dot{\tilde \uv} + \jump{\dot \uv} (H_{\Gamma} - f_\Gamma), 
	\text{ with }
	\dot{\tilde \uv} := \dot{\bar \uv} + \jump{\dot \uv} f_\Gamma
	\ ,
	\label{eq:Borja-smooth-velocity}
\end{align}
where $\tilde \uv$, defined with the term $+ \jump \uv$, is not the same as $\bar \uv$ defined with the term $- \jump \uv$ in Eq.~\eqref{eq:Wang-displacement-jump}$_2$ of \cite{Wang.2018:rd3109}, even though
Eq.~\eqref{eq:Wang-displacement-jump}$_3$ looked similar to
Eq.~\eqref{eq:Borja-smooth-velocity}$_1$.
From Eq.~\eqref{eq:Borja-smooth-velocity}$_3$, the small-strain rate $\dot \epsilon$ in Eq.~\eqref{eq:Borja-strain-rate}$_3$ in terms of the smoothed velocity $\dot{\tilde{\uv}}$ is\footnote{
	\cite{borja2000finite}, Eq.~(2.17).
}
\begin{align}
	\dot {\boldsymbol{\epsilon}}
	=
	\text{sym} (\nabla \dot{\tilde{\uv}})
	+
	\text{sym} (\jump {\dot{{\uv}}} \otimes \bnorm \delta_\Gamma)
	-
	\text{sym} (\jump {\dot{{\uv}}} \otimes \nabla f_\Gamma)
	\ ,
	\label{eq:Borja-strain-rate-2}
\end{align}
which, when removing the overhead dot, is similar to, but different from, the small strain expression in \cite{Wang.2018:rd3109}, written as
\begin{align}
	{\boldsymbol{\epsilon}}
	=
	\text{sym} (\nabla {{\uv}})
	+
	\text{sym} (\jump {{{\uv}}} \otimes \bnorm \delta_\Gamma)
	-
	\text{sym} (\jump {{{\uv}}} \otimes \nabla f_\Gamma)
	\ ,
	\label{eq:Wang-strain-rate}
\end{align}  
where the first term was $\uv$ and not $\tilde{\uv} = \uv + \jump \uv f_\Gamma$ in the notation of \cite{Wang.2018:rd3109}.\footnote{
	Recall that $\uv$ in \cite{Wang.2018:rd3109} (Eq.~\eqref{eq:Wang-displacement-jump}$_1$), the ``large-scale (or conformal) displacement field'' without displacement jump, is equivalent to $\bar \uv$ in \cite{borja2000finite} (Eq.~\eqref{eq:Borja-displacement-jump}$_1$), but is of course not the $\bar \uv$ in \cite{Wang.2018:rd3109} (Eq.~\eqref{eq:Wang-displacement-jump}$_2$).
}

Typically, in this type of formulation \cite{borja2000finite}, once the traction-separation law $\bT ( \jump \uv)$ in Eq.~\eqref{eq:Wang-traction-separation} was available (e.g., Figure~\ref{fig:Wang-micro-RVE-training-data}), then given the traction $\bT$, the displacement jump $\jump \uv$ was solved for using Eq.~\eqref{eq:Wang-traction-separation} at each Gauss point within a constant-strain triangular (CST) element \cite{Wang.2018:rd3109}.

At this point, it is no longer necessary to review further this continuum formulation for displacement jumps to return to the training of the macroscale RNN with LSTM units, which 
%
% CMES style rewriting
the authors of 
\cite{Wang.2018:rd3109} called the ``Macroscale data-driven constitutive model'' (Figure~\ref{fig:Wang-RNN-FEM}, row 3, right), using the data generated from the simulations using the mesoscale RVE (Figure~\ref{fig:Wang-mesoscale-RVE}) and the mesoscale RNN with LSTM units, called the ``Mesoscale data-driven constitutive model'' (Figure~\ref{fig:Wang-RNN-FEM}, row 2, right), obtained earlier.

The mesoscale RNN with LSTM units (``mesoscale data-driven constitutive model'') was first validated using the mesoscale RVE with embedded discontinuities (Figure~\ref{fig:Wang-mesoscale-RVE}), discretized into finite elements, and subjected to imposed displacements at the top.  This combined FEM and RNN with LSTM units on the mesoscale RVE is denoted FEM-LSTM, with results compared well with those obtained from the coupled FEM and DEM (denoted as FEM-DEM), as shown in Figure~\ref{fig:Wang-results_mesoscale_RNN}.

Once validated, the FEM-LSTM model for the mesocale RVE was used to generate data to train the macroscale RNN with LSTM units (called ``Macroscale data-driven constitutive model'') by imposing displacement jumps at the top of the mesoscale RVE (Figure~\ref{fig:Wang-mesoscale-RVE-imposed-displacements}), very much like what was done with the microscale RVE (Figure~\ref{fig:Wang-micro-RVE}, Figure~\ref{fig:Wang-micro-RVE-training-data}), just at a larger scale.

%\noindent
%{\color{red} HERE 2020.09.10.}

%Figure~\ref{fig:Wang-results_mesoscale_RNN} compares meso-scale simulations with the microscopic scale being represented by the DEM model and the trained micro-scale RNN model, respectively.
 
The accuracy of the macroscale RNN with LSTM units (``Macroscale data-driven constitutive model'') is illustrated in Figure~\ref{fig:Wang-results_macroscale_RNN}, where the normal tractions under displacement loading were compared to results obtained with the mesoscale RVE (Figure~\ref{fig:Wang-mesoscale-RVE-imposed-displacements}, Figure~\ref{fig:Wang-mesoscale-RVE}), which was used for generating the training data.  Once established, the macroscale RNN with LSTM units is used in field-size macroscale simulations.  Since there is no further interesting insights into the use of deep learning, we stop of review of \cite{Wang.2018:rd3109} here.

\begin{rem}
	\label{rm:Wang-no-nonlinear-stress-strain}
	No non-linear stress-strain relation.
	{\rm 
		In the end, 
		%
		% CMES style rewriting
		the authors of 
		\cite{Wang.2018:rd3109} only used Figure~\ref{fig:Majella-pore-structure} to motivate the double porosity (in Majella limestone) in their macroscale modeling and simulations, which did not include the characteristic non-linear stress-strain relation found experimentally in Majella limestone as shown in Figure~\ref{fig:Majella-stress-strain}.  All nonlinear responses considered in \cite{Wang.2018:rd3109} came from the nonlinear traction-separation law obtained from DEM simulations in which the particles themselves were elastic, even though the Hertz contact force-displacement relation was nonlinear \cite{vu1999accurate} \vphantom{\cite{vu2001normal}}\cite{vu2001normal} \cite{zhang2007accurate}.
		See Remark~\ref{rm:Wang-field-size-simulation}.
	}
$\hfill\blacksquare$
\end{rem}

\begin{rem}
	\label{rm:PINN-solid-mechanics}
	Physics-Informed Neural Networks (PINNs) applied to solid mechanics.
	{\rm 
		The PINN method discussed in Section~\ref{sc:PINN-frameworks} has been applied to problems in solid mechanics \cite{haghighat2021physics}: Linear elasticity (square plate, plane strain, trigonometric body force, with exact solution), nonlinear elasto-plasticity (perforated plate with circular hole, under plane-strain condition and von-Mises elastoplasticity, subjected to uniform extension, showing localized shear band). 
		Less accuracy was encountered for solutions that presented discontinuiies (localized high gradients) in the materials properties or at the boundary conditions; see Remark~\ref{rm:attention-kernel-PINN} and Remark~\ref{rm:PINN-attention}. 
	}
	$\hfill\blacksquare$
\end{rem}

% Mohan (?) 2018 paper
\begin{figure}[h]
	\centering
	\includegraphics[width=0.9\linewidth]{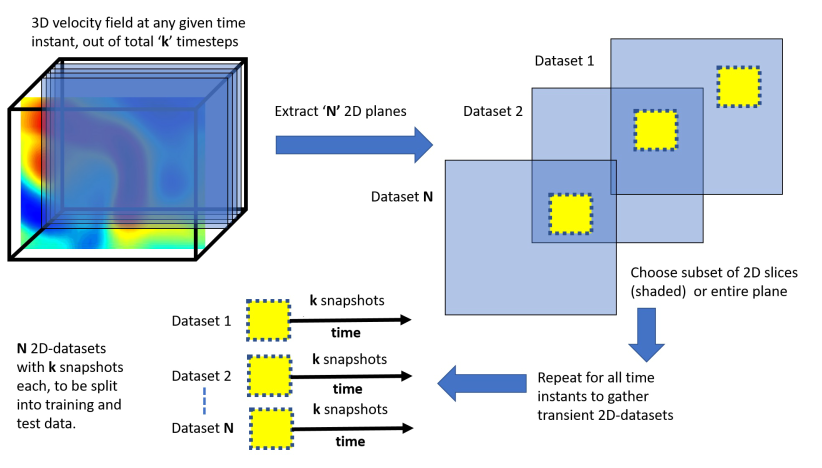}
	\caption{ 
		\emph{2-D datasets for training neural networks} (Sections~\ref{sc:Mohan-2018}, \ref{sc:Mohan-POD}). %\ref{sc:Mohan-data-preparation}, \ref{sc:Mohan-reduced-POD-procedure})
		Extract 2-D datasets from 3-D turbulent flow field evolving in time. From the 3-D flow field, extract $N$ equidistant 2-D planes (slices).  Within each 2-D plane, select a region (yellow square), and $k$ temporal snapshots of this region as it evolves in time to produce a dataset.  
		%There were thus $N$ datasets, each containing $k$ snapshots of the same region within each 2-D plane.
		Among these $N$ datasets, each containing $k$ snapshots of the same region within each 2-D plane, the majority of the datasets is used for training, and the rest for testing; see Remark~\ref{rm:Mohan-reduced-order-POD}.
		For each dataset, the reduced POD basis consists of $m \ll k$ POD modes with highest eigenvalues of a matrix constructed from the $k$ snapshots (Figure~\ref{fig:Mohan-reduced-order-POD-basis})
		%See Figure~\ref{fig:Mohan-reduced-order-POD-basis} for the concept of reduced-order POD basis.
		\cite{Mohan.2018}. 
		\footnotesize (Figure reproduced with permission of the author.)
	}
	\label{fig:Mohan-2D-datasets}
\end{figure}

\section{Application 3: Fluids, turbulence, reduced-order models}
\label{sc:fluid}
\label{sc:Mohan-2018-2}
The general ideas behind the work 
%
% CMES style rewriting
%of 
in
\cite{Mohan.2018} were presented in Section~\ref{sc:Mohan-2018} further above. In this section, we discuss some details of the formulation, starting with a brief primer on Proper Orthogonal Decomposition (POD) for unfamiliar readers.

\begin{figure}[h]
	\centering
	\includegraphics[width=0.9\linewidth]{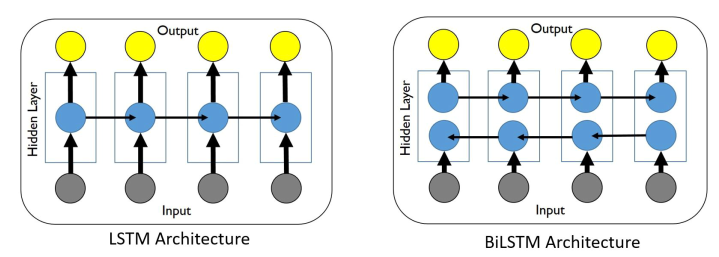}
	\caption{ 
		\emph{LSTM unit and BiLSTM unit} (Sections~\ref{sc:Wang-Sun-2018}, \ref{sc:Mohan-2018}, \ref{sc:LSTM}, \ref{sc:Mohan-reduced-POD}).
		Each blue dot is an original LSTM unit (in \emph{folded} form Figure~\ref{fig:our-lstm_cell} in Section~\ref{sc:LSTM}, without peepholes as shown in Figure~\ref{fig:Wang-LSTM_cell}), thus a single hidden layer.
		The above LSTM architecture (left) in \emph{unfolded} form corresponds to Figure~\ref{fig:Olah-lstm_chain}, with the inputs at state $[k]$ designated by $\bx^{[k]}$ and the corresponding outputs by $\bh^{[k]}$, for $k = \ldots , n-1, n, n+1, \ldots$. 
		In the BiLSTM architecture (right), there are two LSTM units in the hidden layer, with the forward flow of information in the bottom LSTM unit, and the backward flow in the top LSTM unit
		\cite{Mohan.2018}. 
		\footnotesize (Figure reproduced with permission of the author.)
	}
	\label{fig:Mohan-LSTM-BiLSTM}
\end{figure}

\begin{figure}[h]
	\centering
	\includegraphics[width=0.9\linewidth]{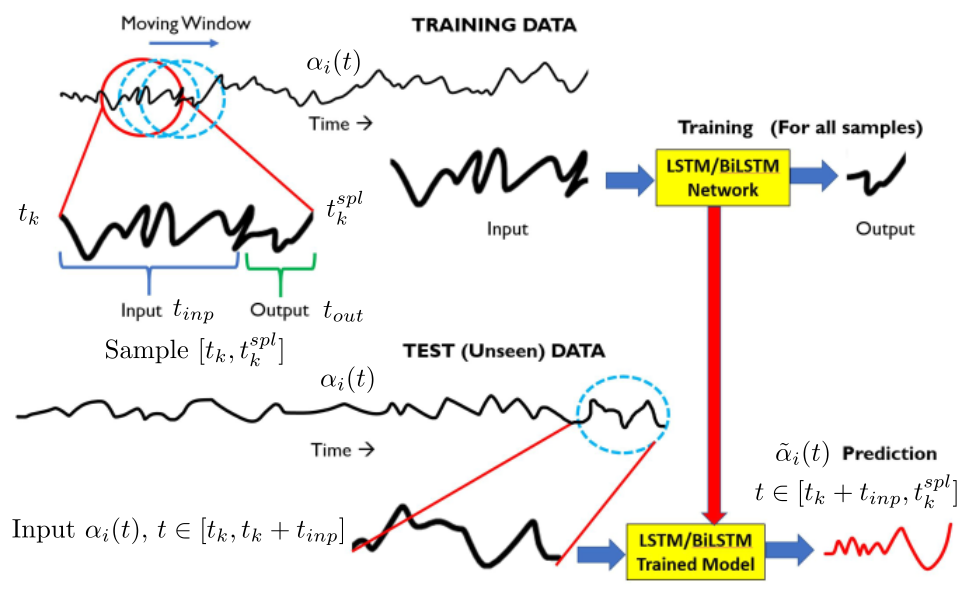}
	\caption{ 
		\emph{LSTM/BiLSTM training strategy} (Sections~\ref{sc:Mohan-reduced-POD-method}, \ref{sc:Mohan-reduced-POD-procedure}).
		From the 1-D time series $\alpha_i(t)$ of each dominant mode $\phi_i$, for $i=1, \ldots , m$, use a moving window to extract thousands of samples $\alpha_i(t)$, $t \in [t_k , t_k^{spl}]$, with $t_k$ being the time of snapshot $k$. Each sample is subdivided into an input signal $\alpha_i(t)$, $t \in [t_k , t_k + t_{inp}]$ and an output signal $\alpha_i(t)$, $t \in [t_k + t_{inp} , t_k^{spl}]$, with $t_k^{spl} - t_k = t_{inp} + t_{out}$ and $0 < t_{out} \le t_{inp}$. The windows can be overlapping. These thousands of input/output pairs were then used to train LSTM-ROM networks, in which LSTM can be replaced by BiLSTM (Figure~\ref{fig:Mohan-LSTM-BiLSTM}).
		The trained LSTM/BiLSTM-ROM networks were then used to predict $\tilde{\alpha}_i(t)$, $t \in [t_k + t_{inp} , t_k^{spl}]$ of the test datasets, given $\alpha_i(t)$, $t \in [t_k , t_k + t_{inp}]$.
		\cite{Mohan.2018}. 
		\footnotesize (Adapted with permission of the author.)
	}
	\label{fig:Mohan-LSTM-training}
\end{figure}

\begin{figure}[h]
	\centering
	% option [b] in subfigure is for "bottom" aligmment
 	\begin{subfigure}[b]{0.48\textwidth}
		\centering
		\includegraphics[clip=true, trim=2mm 2mm 2mm 2mm,width=0.92\linewidth]{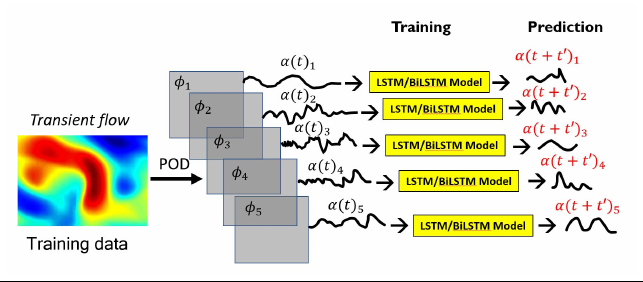}
		\caption{
			Multiple-network method
		}
	\label{fig:Mohan-nn-separate}
	\end{subfigure}
	\ 
	\begin{subfigure}[b]{0.48\textwidth}
		\centering
		\includegraphics[clip=true, trim=1.5mm 1.5mm 2mm 0,width=\linewidth]{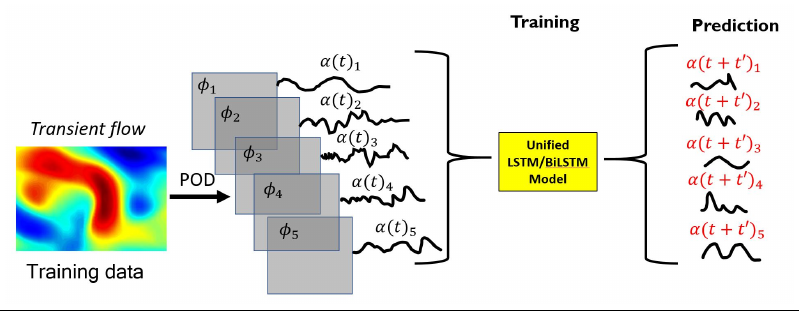}
		\caption{
			Single-network method
		}
	\label{fig:Mohan-nn-unified}
	\end{subfigure}
	\caption{
		\emph{Two methods of developing LSTM-ROM} (Section~\ref{sc:Mohan-2018}, \ref{sc:Mohan-reduced-POD-procedure}).
		For each physical problem (ISO or MHD):
		(a) Each dominant POD mode has a network to predict $\alpha_i (t + t^\prime)$, with $t^\prime$, given $\alpha_i(t)$;
		(b) All $m$ POD dominant modes share the same network to predict $\{\alpha_i (t + t^\prime), \ i=1, \ldots, m\}$,  with $t^\prime$, given $\{\alpha_i (t), \ i=1, \ldots, m\}$
		\cite{Mohan.2018}. 
		\footnotesize (Figure reproduced with permission of the author.)
%		{\color{red} 2020.03.14.  need caption.}
	}
	\label{fig:Mohan-nn}
\end{figure}

\begin{figure}[h]
	\centering
	\includegraphics[width=0.9\linewidth]{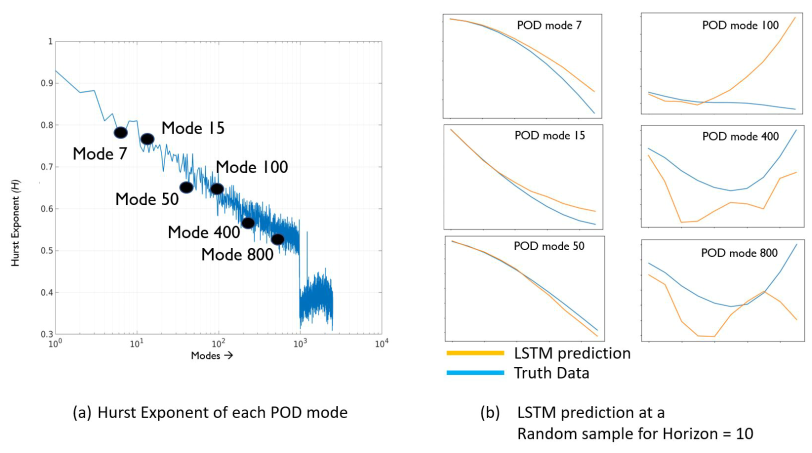}
	\caption{ 
		\emph{Hurst exponent vs POD-mode rank for Isotropic Turbulence (ISO)} (Sections~\ref{sc:Mohan-results}).
		POD modes with larger eigenvalues (Eq.~\eqref{eq:Mohan-eigenvalue-problem-discrete}) are higher ranked, and have lower rank number, e.g., POD mode rank 7 has larger eigenvalue, and thus more dominant, than POD mode rank 50.  
		The Hurst exponent, even though fluctuating, trends downward with the POD mode rank, but not monotonically, i.e., for two POD modes sufficiently far apart (e.g., mode 7 and mode 50), a POD mode with lower rank generally has a lower Hurst exponent.
		The first 800 POD modes for the ISO problem have the Hurst exponents higher than 0.5, and are thus persistent \cite{Mohan.2018}.
		\footnotesize (Adapted with permission of the author.)
	}
	\label{fig:Mohan-Hurst-exponent-ISO}
\end{figure}

\subsection{Proper orthogonal decomposition (POD)}
\label{sc:Mohan-POD}
The presentation of the \emph{continuous} formulation of POD in this section follows \cite{zhai2007analysis}.
Consider the separation of variables of a time-dependent function $u (\bx, t)$, which could represent a component of a velocity field or a magnetic potential in a 3-D domain $\domain$, where $\bx \in \domain$, and $t$ a time parameter:
\begin{align}
	u(\bx, t) = \alpha(t) \phi(\bx)
	\ ,
	\label{eq:Mohan-separation-variables} 
\end{align} 
where $\alpha (t)$ is the time-dependent amplitude, and $\phi(\bx)$ a function of $\bx$ representing the most typical spatial structure of $u(\bx, t)$.  The goal is to find the function $\phi(\bx)$ that maximizes the square of the amplitude $\alpha(t)$ (or component of $u(\bx, t)$ along $\phi(\bx)$):
\begin{align}
	\alpha(t) 
	& = \frac{(u, \phi)}{(\phi, \phi)} \ , \text{ with } (u, \phi) := \int_\domain u(\by , t) \phi(\by) d \domain (\by) 
	\ ,
	\label{eq:Mohan-separation-amplitude-1}
	\\
	(\phi, \phi) 
	& = 1 \Rightarrow \alpha(t) = (u , \phi)
	\ ,
	\label{eq:Mohan-separation-amplitude-2}
\end{align}
i.e., if $\phi(\bx)$ is normalized to 1, then the amplitude $\alpha(t)$ is the spatial scalar product between $u (\bx, t)$ and $\phi (\bx)$, with $(\cdot , \cdot)$ designating the spatial scalar product operator. 
Let $\langle \cdot \rangle$ designate the time average operator, i.e., 
\begin{align}
	\langle f(t) \rangle = \frac{1}{T} \int_{t=0}^{t=T} f(t) \, dt
	\ ,
	\label{eq:Mohan-time-average-operator}
\end{align}
where $T$ designates the maximum time.  The goal now is to find $\phi(\bx)$ such that the time average of the square of the amplitude $\alpha(t)$ is maximized:
\begin{align}
	\phi(\bx) 
	= \underset{\phi}{\text{argmax}} (\langle \alpha^2(t) \rangle) 
	= \underset{\phi}{\text{argmax}} (\langle (u, \phi)^2 \rangle)
	\ ,
	\label{eq:Mohan-phi-argmax}
\end{align}
which is equivalent to maximizing the amplitude $\alpha(t)$, and thus the information content of $u(\bx, t)$ in $\phi(\bx)$, which in turn is also called ``coherent structure''.  The square of the amplitude in Eq.~\eqref{eq:Mohan-phi-argmax} can be written as
\begin{align}
	\lambda := \alpha^2(t) = (u, \phi)^2 = \int\limits_{\domain} \left[ \int\limits_{\domain} u(\bx, t) u(\by, t) \phi(\by) \, d\by \right] \, \phi(\bx) \, d \bx
	\ ,
	\label{eq:Mohan-square-amplitude}
\end{align}  
so that $\lambda$ is the component (or projection) of the term in square brackets in Eq.~\eqref{eq:Mohan-square-amplitude} along the direction $\phi(\bx)$.  The component $\lambda$ is maximal if this term in square brackets is colinear with (or ``parallel to'') $\phi(\bx)$, i.e.,
\begin{align}
	\int\limits_{\domain} u(\bx, t) u(\by, t) \phi(\by) \, d\by = \lambda \phi(\bx)
	\Rightarrow
	\int\limits_{\domain} \langle u(\bx, t) u(\by, t) \rangle \phi(\by) \, d\by = \lambda \phi(\bx)
	\ ,
	\label{eq:Mohan-eigenvalue-problem}
\end{align}
which is a \emph{continuous} eigenvalue problem with the eigenpair being $(\lambda , \phi(\bx))$.  In practice, the dynamic quantity $u(\bx, t)$ is sampled at $k$ discrete times $\{t_1 , \ldots , t_k\}$ to produce $k$ snapshots, which are functions of $\bx$, assumed to be linearly independent, and ordered in matrix form as follows 
%$ \{ u(\bx, t_1) \ldots , u(\bx, t_k) \}$
\begin{align}
	\{ u(\bx, t_1) \ldots , u(\bx, t_k) \} 
	=: \{ u_1 (\bx) , \ldots , u_k (\bx)\} 
	=: \uv (\bx)
	\ .
	\label{eq:Mohan-snapshots-vector}
\end{align}
The coherent structure $\phi(\bx)$ can be expressed on the basis of the snapshots as
\begin{align}
	\phi(\bx) 
	= \sum_{i=1}^{i=k} \beta_i u_i (\bx)
	= \bpc \, \dotprod \, \uv
	\ , 
	\text{ with }
	\bpc = \{ \pcs 1 , \ldots , \pcs k \}
	\ .
	\label{eq:Mohan-phi-snapshots}
\end{align}
As a result of the discrete nature of Eq.~\eqref{eq:Mohan-snapshots-vector} and Eq.~\eqref{eq:Mohan-phi-snapshots}, the eigenvalue problem in Eq.~\eqref{eq:Mohan-eigenvalue-problem} is discretized into
\begin{align}
	\boldsymbol{C} \bpc = \lambda \bpc
	\ , \text{ with }
	\boldsymbol{C} := \left[ \frac{1}{k}  \int\limits_{\domain} \uv (\by) \otimes \uv (\by) d \by \right]
	\ ,
	\label{eq:Mohan-eigenvalue-problem-discrete}
\end{align}
where the matrix $\boldsymbol{C}$ is symmetric, positive definite, leading to positive eigenvalues in $k$ eigenpairs $(\lambda_i , \bpc_i)$, with $i = 1, \ldots, k$.
With $\phi(\bx)$ now decomposed into $k$ linearly independent directions $\phi_i (\bx) := \pcs i u_i (\bx)$ according to Eq.~\eqref{eq:Mohan-phi-snapshots}, the dynamic quantity $u(\bx, t)$ in Eq.~\eqref{eq:Mohan-separation-variables} can now be written as a linear combination of $\phi_i (\bx)$, with $i=1, \ldots, k$, each with a different time-dependent amplitude $\alpha_i (t)$, i.e.,
\begin{align}
	u(\bx, t) = \sum_{i=1}^{i=k} \alpha_i (t) \phi_i (\bx)
	\ ,
	\label{eq:Mohan-POD}
\end{align}
which is called a proper orthogonal decomposition of $u(\bx, t)$, and is recorded in Figure~\ref{fig:Mohan-reduced-order-POD-basis} as ``Full POD reconstruction''.   Technically, Eq.~\eqref{eq:Mohan-phi-snapshots} and Eq.~\eqref{eq:Mohan-POD} are approximations of infinite-dimensional functions by a finite number of linearly-independent functions.  

Usually, a subset of $m \ll k$ POD modes are selected such that the error committed by truncating the basis as done in Eq.~\eqref{eq:Mohan-reduced-POD-basis} would be small compared to  Eq.~\eqref{eq:Mohan-POD}, and recalled here for convenience:
\begin{align}
	u (\bx, t) 
	\approx 
	\sum_{j=1}^{j=m} \phi_{i_j} (\bx) \alpha_{i_j} (t)
	\ , 
	\text{ with } m \ll k
	\text{ and } i_j \in \{ 1 , \ldots , k\}
	\ .
	\tag{\ref{eq:Mohan-reduced-POD-basis}}
\end{align}
One way is to select the POD modes corresponding to the highest eigenvalues (or energies) in Eq.~\eqref{eq:Mohan-eigenvalue-problem}; see Step~\ref{it:Mohan-extract} in Section~\ref{sc:Mohan-reduced-POD-procedure}.

\begin{rem}
	\label{rm:Mohan-reduced-order-POD}
	Reduced-order POD.
	{\rm 
		Data for two physical problems were available from numerical simulations: (1) the Force Isotropic Turbulence (ISO) dataset, and (2) the Magnetohydrodynamic Turbulence (MHD) dataset \vphantom{\cite{graham2016web}}\cite{graham2016web}.
		For each physical problem,
		%
		% CMES style rewriting
		the authors of 
		\cite{Mohan.2018} employed 
		$N=6$ equidistant 2-D planes (slices, Figure~\ref{fig:Mohan-2D-datasets}), with 5 of those 2-D planes used for training, and 1 remaining 2-D plane used for testing (see Section~\ref{sc:training-valication-test}).  
		The same sub-region of the 6 equidistant 2-D plane (yellow squares in Figure~\ref{fig:Mohan-2D-datasets}) was used to generate 6 training / testing datasets.
		For each training / testing dataset, 
		%
		% CMES style rewriting
%		\cite{Mohan.2018} employed 
		$k= 5,023$ snapshots for the ISO dataset and $k=1,024$ snapshots for the MHD dataset were used in \cite{Mohan.2018}.  The reason was because the ISO dataset contained 5,023 time steps, whereas the MHD dataset contained 1,024 time steps. So the number of snapshots was the same as the number of time steps. These snapshots were reduced to $m \in \{5 , \ldots , 10\}$ POD modes with highest eigenvalues (thus energies), which were much fewer than the original $k$ snapshots, since $m \ll k$.
		See Remark~\ref{rm:Mohan-input-output-length}.
	}
$\hfill\blacksquare$
\end{rem}

\begin{rem}
	\label{rm:Mohan-another-POD}
	{\rm 
		\emph{Another method of finding the POD modes} without forming the symmetric matrix $\boldsymbol{C}$ in Eq.~\eqref{eq:Mohan-eigenvalue-problem-discrete} is by using the Singular Value Decomposition (SVD) directly on the rectangular matrix of the sampled snapshots, discrete in both space and time.  
		The POD modes are then obtained from the left singular vectors times the corresponding singular values.  A reduced POD basis is obtained next based on an information-content matrix.
		See \cite{zhai2007analysis},
		%
		% CMES style rewriting 
%		who applied 
		where
		POD was applied to efficiently solve nonlinear electromagnetic problems governed by Maxwell's equations with nonlinear hysteresis at low frequency (10 kHz), called static hysteresis,
		discretized by a finite-element method. 
		See also Remark~\ref{rm:Mohan-POD-different-excitations}.
	}
$\hfill\blacksquare$
\end{rem}

%{\color{red} HERE 2020.10.14}

\subsection{POD with LSTM-Reduced-Order-Model}
\label{sc:Mohan-reduced-POD}
Typically, once the dominant POD modes of a physical problem (ISO or MHD) were identified, a reduced-order model (ROM) can be obtained by projecting the governing partial differential equations (PDEs) onto the basis of the dominant POD modes using, e.g., Galerkin projection (GP). 
Using this method,
%
% CMES style rewriting
the authors of 
\cite{zhai2007analysis} employed full-order simulations of the governing electro-magnetic PDE with certain input excitation to generate POD modes, which were then used to project similar PDE with different parameters and solved for the coefficients $\alpha_i (t)$ under different input excitations.

\subsubsection{Goal for using neural network}
\label{sc:Mohan-reduced-POD-method}
Instead of using GP on the dominant POD modes of a physical problem to solve for the coefficients $\alpha_i (t)$ as described above,
%
% CMES style rewriting 
%\cite{Mohan.2018} used 
deep-learning neural network was used in \cite{Mohan.2018} to predict the next value of $\alpha_i (t + t^\prime)$, with $t^\prime > 0$, given the current value $\alpha_i (t)$, for $i=1, \ldots, m$.

To achive this goal,
%
% CMES style rewriting 
%\cite{Mohan.2018} used 
LSTM/BiLSTM networks (Figure~\ref{fig:Mohan-LSTM-BiLSTM}) were trained using thousands of paired short input / output signals obtained by segmenting the time-dependent signal $\alpha_i (t)$ of the dominant POD mode $\phi_i$ \cite{Mohan.2018}.

\subsubsection{Data generation, training and testing procedure}
\label{sc:Mohan-reduced-POD-procedure}
The following procedure was adopted in \cite{Mohan.2018} to develop their LSTM-ROM for two physical problems, ISO and MHD; see Remark~\ref{rm:Mohan-reduced-order-POD}.
For each of the two physical problems (ISO and MHD), the following steps were used:
\begin{enumerate}
	
	\item
	From the 3-D computational domain of a physical problem (ISO or MHD), select $N$ equidistant 2-D planes that slice through this 3-D domain, and select the same subregion for all of these planes, the majority of which is used for the training datasets, and the rest for the test datasets (Figure~\ref{fig:Mohan-2D-datasets} and Remark~\ref{rm:Mohan-reduced-order-POD} for the actual value of $N$ and the number of training datasets and test datasets employed in \cite{Mohan.2018}).

	%\item
	%From the planes, select a test dataset - this will be the dataset which will be modeled by the DL-ROM after it ``learns'' the dynamics in the training datasets.
	
	\item
	\label{it:Mohan-extract}
	For each of the training datasets and test datasets,
	extract from $k$ snapshot a few $m \ll k$ dominant POD modes $\phi_i$, $i=1, \ldots, m$ 
	(with the highest energies / eigenvalues) 
	and their corresponding coefficients $\alpha_i (t)$, $i = 1, \ldots , m$, by solving the eigenvalue problem Eq.~\eqref{eq:Mohan-eigenvalue-problem-discrete}, then use Eq.~\eqref{eq:Mohan-phi-snapshots} to obtain the POD modes $\phi_i$, and Eq.~\eqref{eq:Mohan-separation-amplitude-2} to obtain $\alpha_i (t)$ for use in Step~\ref{it:Mohan-input-output-pairs}.
	
	\item
	\label{it:Mohan-input-output-pairs}	
	The time series of the coefficient $\alpha_i(t)$ of the dominant POD mode $\phi_i (\bx)$ of a \emph{training} dataset is chunked into thousands of small samples with $t \in [t_k , t_k^{spl}]$, where $t_k$ is the time of the \emph{kth} snapshot, by a moving window.  Each sample is subdivided into two parts: The input part with time length $t_{inp}$ and the output part with time length $t_{out}$, Figure~\ref{fig:Mohan-LSTM-training}.  These thousands of input/output pairs were then used to train LSTM/BiLSTM networks in Step~\ref{it:Mohan-train-LSTM-BiLSTM}.
	See Remark~\ref{rm:Mohan-input-output-length}.
	
%	, with $t \in [t_k , t_k + t_{inp}]$, where $t_k$ is the time of the \emph{kth} snapshot, and $0 < t_{inp}$, as input, use the trained LSTM/BiLSTM networks to predict the values of $\tilde{\alpha}_i(t)$, with $t \in [t_k + t_{inp} , t_k^{spl}]$, where $t_k^{spl}$ is the time at the end of the sample, such that the sample length is $t_k^{spl} - t_k = t_{inp} + t_{out}$ and $0 < t_{out} \le t_{inp}$; see Figure~\ref{fig:Mohan-LSTM-training}.
		
%	{\color{red} HERE 2020.11.08}
	
	\item
	\label{it:Mohan-train-LSTM-BiLSTM}
	Use the input/output pairs generated from the training datasets in Step~\ref{it:Mohan-input-output-pairs} to train LSTM/BiLSTM-ROM networks. Two methods were considered in \cite{Mohan.2018}:
	\begin{enumerate}
		
		\item
%		\emph{Network model 1:}
%		One network \emph{per} dominant POD mode.
		\emph{Multiple-network method:}
		Use a separate RNN for each of the $m$ dominant POD modes to separately predict 
		%trajectories of participation vectors 
		the coefficient $\alpha_i (t)$, for $i = 1,\ldots,m$, given a sample input, see Figure~\ref{fig:Mohan-nn-separate} and Figure~\ref{fig:Mohan-LSTM-training}.
		Hyperparameters (layer width, learning rate, batch size) are tuned for the most dominant POD mode and reused for training the other neural networks.
		
		\item
%		\emph{Network model 2:}
%		One network for \emph{all} dominant POD modes.
		\emph{Single-network method:}
		Use the same RNN to predict the coefficients $\alpha_i (t)$, $i = 1,\ldots, m$ of all $m$ dominant POD modes at once, given a sample input, see Figure~\ref{fig:Mohan-nn-unified} and Figure~\ref{fig:Mohan-LSTM-training}.
		
	\end{enumerate}
	The single-network method better captures the inter-modal interactions that describe the energy transfer from larger to smaller scales.
	Vortices that spread over multiple dominant POD modes also support the single-network method, which does not artificially constrain flow features to separate POD modes.
	
	\item
	Validation: 
	Input/output pairs similar to those for training in Step~\ref{it:Mohan-input-output-pairs} were generated from the \emph{test} dataset for validation.\footnote{
%		The ``test'' dataset in \cite{Mohan.2018} should really be called the \emph{validation} dataset to be consistent with the definition of training / validation / test datasets in Section~\ref{sc:training-valication-test}.
		%
		% CMES style rewriting
		The authors of 
		\cite{Mohan.2018} did not have a validation dataset as defined in Section~\ref{sc:training-valication-test}.
	}
	With a short time series of the coefficient $\alpha_i(t)$ of the dominant POD mode $\phi_i (\bx)$ of the \emph{test} dataset, with $t \in [t_k , t_k + t_{inp}]$, where $t_k$ is the time of the \emph{kth} snapshot, and $0 < t_{inp}$, as input, use the trained LSTM/BiLSTM networks in Step~\ref{it:Mohan-train-LSTM-BiLSTM} to predict the values of $\tilde{\alpha}_i(t)$, with $t \in [t_k + t_{inp} , t_k^{spl}]$, where $t_k^{spl}$ is the time at the end of the sample, such that the sample length is $t_k^{spl} - t_k = t_{inp} + t_{out}$ and $0 < t_{out} \le t_{inp}$; see Figure~\ref{fig:Mohan-LSTM-training}. Compute the error between the predicted value $\tilde{\alpha}_i(t + t_{out})$ and the target value of $\alpha_i(t + t_{out})$ from the test dataset. Repeat the same prediction procedure for all $m$ dominant POD modes chosen in Step~\ref{it:Mohan-extract}.
	See Remark~\ref{rm:Mohan-input-output-length}.

	\item
	At time $t$, use the predicted coefficients $\tilde{\alpha}_i(t)$ together with the dominant POD modes $\phi_i(\bx)$, for $i=1, \ldots, m$, to compute the flow field dynamics $u(\bx, t)$ using Eq.~\eqref{eq:Mohan-reduced-POD-basis}.
		
\end{enumerate}

%	{\color{red} HERE 2020.11.08}

%\subsection{Data preparation}
%\label{sc:Mohan-data-preparation}

%\subsection{Discussion of results}
\subsection{Memory effects of POD coefficients on LSTM models}
\label{sc:Mohan-results}
%{\color{blue}
%	[NOTE: 2012.12.11. what is meant by ``U-velocity field''? END NOTE]
%}

%
The results and deep-learning concepts used in \cite{Mohan.2018} were presented in the motivational Section~\ref{sc:Mohan-2018} above, 
%Figure~\ref{fig:Mohan-result_isotropic}, Figure~\ref{fig:Mohan-result_mhd}, 
Figure~\ref{fig:Mohan-results}. In this section, we discuss some details of the formulation.

\begin{rem}
	\label{rm:Mohan-input-output-length}
	{\rm 
		Even though 
		%
%		\cite{Mohan.2018} gave 
		the value of $t_{inp} = t_{out} = 0.1 \, sec$ was given in \cite{Mohan.2018} as an example, all LSTM/BiLSTM networks in their numerical examples were trained using input/output pairs with $t_{inp} = t_{out} = 10 \times \Delta t$, i.e., 10 time steps for both input and output samples.  With the overall simulation time for both physical problems (ISO and MHD) being $2.056 \, secs$, the time step size is $\Delta t_{\text{ISO}} = 2.056 / 5023 = 4.1 \times 10^{-4} \, sec$ for ISO, and $\Delta t_{\text{MHD}} = 2.056 / 1024 = 2 \times 10^{-3} \, sec = 5 \Delta t_{\text{ISO}}$ for MHD. See also Remark~\ref{rm:Mohan-reduced-order-POD}. 
	}
\phantom{space}$\hfill\blacksquare$
\end{rem}

%{\color{red} HERE 2020.11.08}

%
% CMES style rewriting
%\cite{Mohan.2018} mentioned that they used 
The ``U-velocity field for all results'' was mentioned in \cite{Mohan.2018},  
%but did not provide 
but without
a definition of ``U-velocity'', which was possibly the $x$ component of the 2-D velocity field in the 2-D planes (slices) used for the datasets, with ``V-velocity'' being the corresponding $y$ components. 

The BiLSTM networks in the numerical examples were not as accurate as the LSTM networks for both physical problems (ISO and MHD), despite having more computations; see Figure~\ref{fig:Mohan-LSTM-BiLSTM} above and Figure~\ref{fig:Mohan-results} in Section~\ref{sc:Mohan-2018}.  
%
% CMES style rewriting
The authors of 
\cite{Mohan.2018} conjectured that a reason could be due to the randomness nature of turbulent flows, as opposed to the high long-term correlation found in natural human languages, for which BiLSTM was designed to address.

Since LSTM architecture was designed specifically for sequential data with memory, 
%
% CMES style rewriting
%\cite{Mohan.2018} 
it was
sought in \cite{Mohan.2018} to quantify whether there was ``memory'' (or persistence) in the time series of the coefficients $\alpha_i(t)$ of the POD mode $\phi_i(\bx)$.
%\cite{Mohan.2018} further performed a case study regarding the relationship between the accuracy of their LSTM-ROM models, the length of the predictive horizon and the persistence of histories of individual POD modes.
To this end, the Hurst exponent\footnote{
	``Hurst exponent'', Wikipedia, 
	\href{https://en.wikipedia.org/w/index.php?title=Hurst_exponent&oldid=986283721}{version 22:09, 30 October 2020}.
} was used to quantify the presence or absence of long-term memory in the time series $\mathcal{S}_i (k,n)$:
\begin{align}
	\mathcal{S}_i (k,n) 
	&
	:= 
	\{ \alpha_i(t_{k+1}) , \ldots , \alpha_i(t_{k+n}) \}
	,
	\ \underset{i, n \ \text{fixed}}{\mathbb{E}} \left( \frac{R_i (k,n)}{\stdev_i (k,n)} \right)  
	=
	C_i e^{H_i}
	, 
	\\
	\ R_i (k,n) 
	&
	= \max \mathcal{S}_i (k,n) - \min \mathcal{S}_i (k,n)
	,
	\ \stdev_i (k,n) = \text{stdev} \ [\mathcal{S}_i (k,n)] 
	, 
	\label{eq:Mohan-Hurst-exponent}
\end{align} 
where 
$\mathcal{S}_i (k,n)$ is a sequence of $n$ steps of $\alpha_i(t)$, starting at snapshot time $t_{k+1}$,
$\mathbb{E}$ is the expectation of the ratio of range $R_i$ over standard deviation $\stdev_i$ for many samples $\mathcal{S}_i (k,n)$, with different values of $k$ keeping $(i, n)$ fixed,
$R_i (k,n)$ the range of sequence $\mathcal{S}_i (k,n)$,
$\stdev_i (k,n)$ the standard deviation of sequence $\mathcal{S}_i (k,n)$,
$C_i$ a constant,
$H_i$ the Hurst exponent for POD mode $\phi_i(\bx)$.

A Hurst coefficient of $H = 1$ indicates persistent behavior, i.e., an upward trend in a sequence is followed by an upward trend.
If $H = 0$ the behavior that is represented by time series data is anti-persistent, i.e., an upward trend is followed by a downward trend (and vice versa). 
The case $H = 0.5$ indicates random behavior, which implies a lack of memory in the underlying process.

%
% CMES style rewriting
%\cite{Mohan.2018} studied 
The effects of prediction horizon and persistence on the prediction accuracy of LSTM network were studied in \cite{Mohan.2018}.  
Horizon is the number of steps after the input sample that a LSTM model would predict the values of $\alpha_i(t)$, and is proportional to the output time length $t_{out}$, assuming constant time step size.
Persistence refers to the amount of correlation among subsequent realizations within sequential data, i.e., the presence of the long-term memory.

To this end, they selected one dataset (training or testing), and followed the multiple-network method in Step~\ref{it:Mohan-train-LSTM-BiLSTM} of Section~\ref{sc:Mohan-reduced-POD-procedure} to develop a different LSTM network model for each POD mode with ``non-negligible eigenvalue''.  For both the ISO (Figure~\ref{fig:Mohan-Hurst-exponent-ISO}) and MHD problems, the 800 highest ranked POD modes were used.  

A baseline horizon of 10 steps were used, for which the prediction errors were $\{1.08, 1.37, 1.94, 2.36, $ $ 2.03, 2.00\}$ for POD ranks $\mathcal{R} = \{7, 15, 50, 100, 400, 800\}$, and for Hurst exponents $\{0.78, 0.76, 0.652, $ $ 0.653,  0.56,0.54 \}$, respectively.  So the prediction error increased from 1.08 for POD rank 7 to 2.36 for POD rank 100, then decreased slightly for POD ranks 400 and 800.  The time histories of the corresponding coefficients $\alpha_i(t)$, with $i \in \mathcal{R}$ on the right of Figure~\ref{fig:Mohan-Hurst-exponent-ISO} provided only some qualitative comparison between the predicted values and the true values, but did not provide the scale of the actual magnitude of $\alpha_i(t)$, nor the time interval of these plots.  For example, the magnitude of $\alpha_{800} (t)$ could be very small compared to that of $\alpha_7(t)$, given that the POD modes were normalized in the sense of Eq.~\eqref{eq:Mohan-separation-amplitude-2}.
Qualitatively, the predicted values compared well with the true values for POD ranks 7, 15, 50.
So even though there was a divergence between the predicted value and the true value for POD ranks 100 and 800, but if the magnitude of $\alpha_{100} (t)$ and $\alpha_{800} (t)$ were very small compared to those of the dominant POD modes, there would be less of a concern.

Another expected result was that for the same POD mode rank lower than 50, the error increased dramatically with the prediction horizon.  For example, for POD rank 7, the errors were $\{1.08, 6.86, 15.03,$ $ 19.42, 21.40 \}$ for prediction horizons of $\{10, 25, 50, 75, 100 \}$ steps, respectively.  Thus the error (at 21.40) for POD rank 7 with horizon 100 steps was more than ten times higher than the error (at 2.00) for POD rank 800 with horizon 10 steps.   For POD mode rank 50 and higher, the error did not increase as much with the horizon, but stayed at about the same order of magnitude as the error for horizon 25 steps.

A final note is whether the above trained LSTM-ROM networks could produce accurate prediction for flow dynamics with different parameters such as Reynolds number, mass density, viscosity, geometry, initial conditions, etc., particularly that both the ISO and MHD datasets were created for a single Reynolds number, which was not mentioned in \cite{Mohan.2018}, who mentioned that their method (and POD in general) would work for a ``narrow range of Reynolds numbers'' for which the flow dynamics is qualitatively similar, and for ``simplified flow fields and geometries''. 

\begin{rem}
	\label{rm:Mohan-POD-different-excitations}
	Use of POD-ROM for different systems.
	{\rm
		%
		% CMES style rewriting
		The authors of
		\cite{zhai2007analysis} studied  
		The flexibility of POD reduced-order models to solve nonlinear electromagnetic problems by varying the excitation form (e.g., square wave instead of sine wave) and by using the undamped (without the first-order time derivative term) snapshots in the simulation of the damped case (with the first-order time derivative term).
		They demonstrated via numerical examples involving nonlinear power-magnetic-component simulations that the reduced-order models by POD are quite flexible and robust.  See also Remark~\ref{rm:Mohan-another-POD}.
	}
$\hfill\blacksquare$
\end{rem}

\begin{rem}
	\label{rm:pyMOR}
	pyMOR - Model Order Reduction with Python.
	{\rm 
		Finally, we mention the software pyMOR,\footnote{
			See pyMOR website \href{https://pymor.org/}{https://pymor.org/}.
		} which is ``a software library for building model order reduction applications with the Python programming language. Implemented algorithms include reduced basis methods for parametric linear and non-linear problems, as well as system-theoretic methods such as balanced truncation or IRKA (Iterative Rational Krylov Algorithm). All algorithms in pyMOR are formulated in terms of abstract interfaces for seamless integration with external PDE (Partial Differential Equation) solver packages. Moreover, pure Python implementations of FEM (Finite Element Method) and FVM (Finite Volume Method) discretizations using the NumPy/SciPy scientific computing stack are provided for getting started quickly.''
		It is noted that pyMOR includes POD and ``Model order reduction with artificial neural networks'', among many other methods; see the documentation of pyMOR.
		Clearly, this software tool would be applicable to many physical problems, e.g., solids, structures, fluids, electromagnetics, coupled electro-thermal simulation, etc.
	}
$\hfill\blacksquare$
\end{rem}

%{\color{red} [NOTE: 2022.01.12 - Review Choi 2021 and Mohan's 2nd paper.]}

%
\begin{figure}[h]
	\centering
	\includegraphics[width=0.8\linewidth]{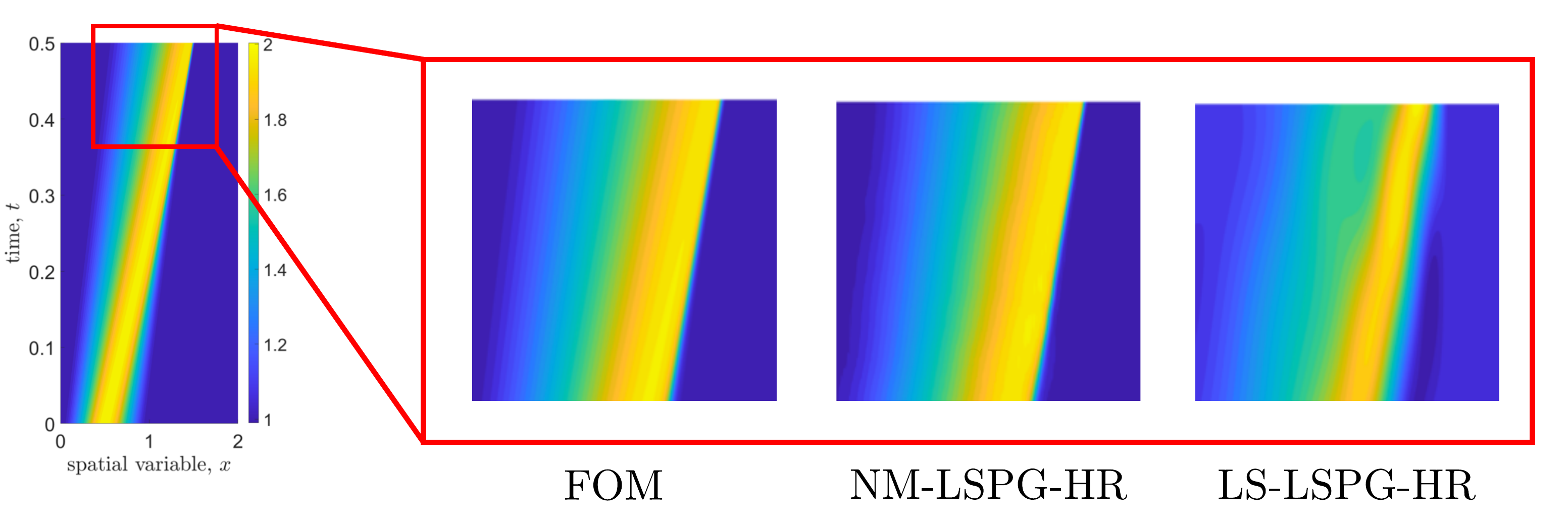}
	\caption{\emph{Space-time solution of inviscid 1D-Burgers' equation} (Section~\ref{sc:Choi-1d-example}).
		The solution shows a characteristic steep spatial gradient, which shifts and further steepens in the course of time.
		The FOM solution (left) and the solution of the proposed hyper-reduced ROM (center), in which the solution subspace is represented by a nonlinear manifold in the form of a feedforward neural network (Section \ref{sc:feedforward}) (NM-LSPG-HR), show an excellent agreement, whereas the spatial gradient is significantly blurred in the solution obtained with a hyper-reduced ROM based on a linear subspace (LS-LSPG-HR) (right).
		The FOM is a obtained upon a finite-difference approximation in the spatial domain with $N_s = 1001$ grid points (i.e., degrees of freedom); the backward Euler scheme is employed for time-integration using a step size $\Delta t = \num{1e-3}$. 
		Both ROMs only use $n_s = 5$ generalized coordinates. The NM-LSPG-HR achieves a maximum relative error of less than \SI{1}{\percent}, while the maximum relative error of the LS-LSPG-HR is approximately \SI{6}{\percent}.
		\footnotesize (Figure reproduced with permission of the authors.)
	}
	\label{fig:Choi-2021-results-1d}
\end{figure}
\subsection{Reduced order models and hyper-reduction}
\label{sc:ROM-hyper}
Reduction of computational expense also is the main aim of \emph{nonlinear manifold reduced-order models} (NM-ROMs), which have recently been proposed in \cite{kim.2020a},\footnote{
	Note that the authors also published a second condensed, but otherwise similar version of their article, see \cite{kim.2020b}.
} 
%
% CMES style rewriting
%Their 
where the
approach belongs to the class of \emph{projection-based methods}. 
Projection-based methods rely on the idea that solutions to physical simulations lie in a subspace of small dimensionality as compared to the dimensionality of high-fidelity models, which we obtain upon discretization (e.g., by finite elements) of the governing equations.
In classical projection-based methods, such \emph{``intrinsic solution subspaces''} are spanned by a set of appropriate basis vectors that capture the essential features of the full-order model (FOM), i.e., the subspace is assumed to be linear.
We refer to \cite{Benner.2015} for a survey on projection-based linear subspace methods for parametric systems.\footnote{
	As opposed to the classical \emph{non-parametric} case, in which all parameters are fixed, \emph{parametric} methods aim at creating ROMs which account for (certain) parameters of the underlying governing equations to vary in some given range. 
	Optimization of large-scale systems, for which repeated evaluations are computationally intractable, is a classical use-case for methods in parametric MOR, see Benner \emph{et al.}~\cite{Benner.2015}.
}
%\paragraph{Prerequisites:} definition of ROM

\subsubsection{Motivating example: 1D Burger's equation}
\label{sc:Choi-1d-example}
The effectiveness of linear subspace methods is directly related to the dimensionality of the basis to represent solutions with sufficient accuracy.
Advection-dominated problems and problems with solutions that exhibit large ({``sharp''}) gradients, however, are characterized by a large Kolmogorov $n$-width,\footnote{
	Mathematically, the dimensionality of linear subspace that `best' approximates a nonlinear manifold is described by the \emph{Kolmogorov $n$-width}, see, e.g., \cite{Greif.2019} for a formal definition.
} which is adverse to linear subspace methods. 
As examples, 
%\citeauthor{kim.2020a}~\cite{kim.2020a} 
%
% CMES style rewriting
%Kim \emph{et al.} 
The authors of
\cite{kim.2020a}
mention hyperbolic equations with large Reynolds number, Boltzmann transport equations and traffic flow simulations.
Many approaches to construct efficient ROMs in adverse problems are based on the idea to enhance the ``solution representability'' of the linear subspace, e.g., by using adaptive schemes tailored to particular problems.
Such problem-specific approaches, however, suffer from limited generality and a the necessity of a-priori knowledge as, e.g., the (spatial) direction of advection.
In view of these drawbacks,
%
% CMES style rewriting 
%\cite{kim.2020a} advocate 
the transition to a solution representation by \emph{nonlinear manifolds} rather than using  linear subspaces in projection-based ROMs was advocated in \cite{kim.2020a}.

Burger's equation serves as a common prototype problem in numerical methods for nonlinear partial differential equations (PDEs) and MOR, in particular.
The inviscid Burgers' equation in one spatial dimension is given by
\begin{equation}
	\frac{\partial u(x, t; \mu)}{\partial t} + u(x,t;\mu) \frac{\partial u}{\partial x} = 0 , \qquad
	x \in \Omega = [0, 2] , \qquad
	t \in [0, T] .
	\label{eq:Choi-Burgers-1d}
\end{equation}
Burgers' equation is a first-order hyperbolic equation which admits the formation of shock waves, i.e., regions with steep gradients in field variables, which propagate in the domain of interest.
Its left-hand side corresponds to material time-derivatives in Eulerian descriptions of continuum mechanics. 
In the balance of linear momentum, for instance, the field $u$ represents the velocity field. 
%
%\citeauthor{kim.2020a}~\cite{kim.2020a}
%
% CMES style rewriting 
%Kim \emph{et al.} \cite{kim.2020a} assume 
Periodic boundary conditions and non-homogeneous initial conditions were assumed \cite{kim.2020a}:
\begin{equation}
	u(2, t; \mu) = u(0, t; \mu) , \qquad
	u(x, 0; \mu) = \begin{cases}
		1 + \frac{\mu}{2} \left( \sin \left( 2 \pi x - \frac{\pi}{2} \right) + 1 \right)	&	\text{if} \; 0 \leq x \leq 1 \\
		1 	&	\text{otherwise}
	\end{cases} ,
\end{equation}
The above initial conditions are governed by a scalar parameter $\mu \in \spaceD = [0.9 , 1.1]$. 
In parametric MOR, ROMs are meant to be valid for not just a single value of the parameter $\mu$, but it is supposed to be valid for a range of 
%
% corrected
%different parameters 
values of $\mu$
in the domain $\spaceD$. 
For this purpose, the reduced-order space, irrespective of whether a linear subspace of FOM or a nonlinear manifold is used, is typically constructed using data obtained for different values of the parameter $\mu$.
%
% CMES style rewritiing
In 
%their
the 
example in \cite{kim.2020a},
%\citeauthor{kim.2020a}~\cite{kim.2020a} 
%Kim \emph{et al.} \cite{kim.2020a} use 
the solutions for the parameter set $\mu \in [{\num{0.9}, \num{1.1}}]$ were used to construct the individual ROMs. 
Note that solution data corresponding to the parameter value $\mu = 1$, for which the ROMs were evaluated, was not used in this example.  

% 2022.06.30
% The code for Figure~\ref{fig:Choi-2021-results-1d} was here, but moved above Section 11.4 so the last figures fall into this section, and not Section 12.
Figure~\ref{fig:Choi-2021-results-1d} shows a zoomed view of solutions to the above problem obtained with the FOM (left), the proposed nonlinear manifold-based ROM (NM-LSPG-HR, center) and a conventional ROM, in which the full-order solution is represented by a linear subspace.
The initial solution is characterized by a ``bump'' in the left half of the domain, which is centered at $x = 1/2$. 
The advective nature of Burgers' problem causes the bump to move right, which also results in slopes of the bump to increase in movement direction but decrease on the averted side.
The zoomed view in Figure~\ref{fig:Choi-2021-results-1d} shows the region at the end ($t = 0.5$) of the considered time-span, in which the (negative) gradient of the solution has already steepened significantly. 
With as little as $n_s = 5$ generalized coordinates, the proposed nonlinear manifold-based approach (NM-LSPG-HR) succeeds in reproducing the FOM solution, which is obtained by a finite-difference approximation of the spatial domain using $N_s = \num{1001}$ grid points; time-integration is performed by means of the backward Euler scheme with a constant step size of $\Delta t = \num{1e-3}$, which translates into a total of $n_t = 500$ time steps.

The ROM based on a linear subspace of the full-order solution (LS-LSPG-HR) fails to accurately reproduce the steep spatial gradient that develops over time, see Figure \ref{fig:Choi-2021-results-1d}, right
Instead, the bump is substantially blurred in the linear subspace-based ROM as compared to the FOM's solution (left). 
%
%To quantify the accuracy of ROMs, \citeauthor{kim.2020a}~\cite{kim.2020a}
%
% CMES style rewriting
%Kim \emph{et al.}~\cite{kim.2020a} consider 
The maximum error over all time steps $t_n$ relative to the full-order solution defined as:
\begin{equation}
	\text{Maximum relative error} = \max_{n \in \left\{ 1, \ldots, n_t \right\}} \frac{\left\Vert \xvtilde(t_n; \mu) - \xv(t_n; \mu) \right\Vert_2}{\left\Vert \xv(t_n; \mu) \right\Vert_2} ,
	\label{eq:Choi-rel-error}
\end{equation}

%{\color{red} [NOTE: 2022.07.29 - the symbol $\mathbb N(N_t)$ is obscure, and must be explained.  What is $\mathbb N$? (Set of natural numbers.) What is $N_t$? ENDNOTE]}

\noindent
where $\xv$ and $\xvtilde$ denote FOM and ROM solution vectors, respectively, was considered in \cite{kim.2020a}. 
In terms of the above metric, the proposed nonlinear-manifold-based ROM achieves a maximum error of approximately \SI{1}{\percent}, whereas the linear-subspace-based ROM shows a maximum relative error of \SI{6}{\percent}. 
For the given problem, the linear-subspace-based ROM is approximately \num{5} to \num{6} times faster than the FOM. 
The nonlinear-manifold-based ROM, however, does not achieve any speed unless \emph{hyper-reduction} is employed. 
Hyper-reduction (HR) methods provide means to efficiently evaluated nonlinear terms in ROMs without evaluations of the FOM (see Hyper-reduction, Section \ref{sc:hyper-reduction}).
Using hyper-reduction, a factor \num{2} speed-up is achieved for both the nonlinear-manifold and linear-subspace-based ROMs, i.e., the effective speed-ups amount to factors of \num{2} and \numrange[range-phrase=--]{9}{10}, respectively. 

%
% CMES style
%\citeauthor{kim.2020a}~\cite{kim.2020a} propose to represent the solution-manifold by 
The solution manifold can be represented by means of a shallow, sparsely connected feed-forward neural network \cite{kim.2020a}
(see Statics, feedforward networks, Sections \ref{sc:feedforward}, \ref{sc:depth-size}).
The network is trained in an unsupervised manner using the concept of {autoencoders} (see Autoencoder, Section \ref{sc:autoencoder}). 

\subsubsection{Nonlinear manifold-based (hyper-)reduction}

The NM-ROM approach proposed in \cite{kim.2020a} addressed nonlinear dynamical systems, whose evolution 
%is
was 
governed by a set of nonlinear ODEs, which 
%have 
had
been obtained by a semi-discretization of the spatial domain, e.g., by means of finite elements: 
\begin{equation}
%	\dot 
	\dt
	\xv 
	= \frac{\mathrm d \xv}{\mathrm d t} = \fv (\xv, t; \muv) , \qquad
	\xv(0; \muv) = \xv_0(\muv) .
	\label{eq:NL_dynamical_system}
\end{equation}
In the above relation, $\xv(t; \muv)$ denotes the parameterized solution of the problem, where $\muv \in \spaceD \subseteq \mathbb R^{n_\mu}$ are an $n_\mu$-dimensional vector of parameters; $\xv_0(\muv)$ is the initial state. 
The function $\fv$ represents the rate of change of the state, which is assumed to be nonlinear in the state $\xv$ and possibly also in its other arguments, i.e., time $t$ and the vector of parameters $\muv$:\footnote{
	In solid mechanics, we typically deal with second-order ODEs, which can be converted into a system of first-order ODEs by including velocities in the state space. 
	For this reason, we prefer to use the term `rate' rather than `velocity' of $\xv$ in what follows. 
}
\begin{equation}
	\xv : [0, T] \times \spaceD \rightarrow \realsN , \qquad 
	\fv : \realsN \times [0, T] \times \spaceD \rightarrow \realsN .
\end{equation}
%A uniform time discretization, which is characterized by a step size $\Delta t$, is assumed; the solution at time $t_n = t_{n-1} + \Delta t = n \Delta t$ is denoted by $\xv_n = \xv(t_n; \muv)$.
%In the development of their method, the authors considered implicit time integration schemes, where the \emph{backward Euler} method,  
%\begin{equation}
%	\xv_n - \xv_{n-1} = \Delta t \fv_n ,
%	\label{eq:backward_euler}
%\end{equation}
%serves as role model.
%The above integration rule implies that span of rates $\fv_n$, which are given by evaluations (`snapshots') of the nonlinear function $\fv$, is included in the span of states (`solution snapshots') $\xv_n$:
%\begin{equation}
%	\spanop \left\{  \fv_n \right\} \subseteq \spanop \left\{  \xv_{n-1}, \xv_n \right\} 
%	\qquad \rightarrow \qquad
%	\spanop \left\{  \fv_1, \ldots, \fv_{N_t} \right\} \subseteq \spanop \left\{  \xv_{0}, \ldots , \xv_{N_t} \right\} .
%\end{equation}
%

%{\color{red} [NOTE: 2022.08.02 - Please check the last sentence above regarding the ``other arguments $(t ; \mu)$'' and the expressions below it.]}

The fundamental idea of any projection-based ROM is to approximate the original solution space of the FOM by a comparatively low-dimensional space $\subspace$.  
In view of the aforementioned shortcomings of linear subspaces, the authors of \cite{kim.2020a} 
%propose
proposed 
a representation in terms of a \emph{nonlinear manifold}, which 
%is 
was
described by the vector-valued function $\gv$, whose dimensionality 
%is
was 
supposed to be much smaller than that of the FOM:
\begin{equation}
	\subspace = \left\{  \gv (\hat \vv) \; | \; \hat \vv \in \realsn \right\} , \qquad
	\gv : \realsn \rightarrow \realsN , \qquad 
	\dim(\subspace) = n_s \ll N_s .
\end{equation}
Using the nonlinear function $\gv$, an approximation $\xvtilde$ to the FOM's solution $\xv$ 
%is 
was
constructed using a set of generalized coordinates $\xvred \in \realsn$:
\begin{equation}
	\xv \approx \xvtilde = \xvref + \gv(\xvred) ,
	\label{eq:state_nm-rom}
\end{equation}
where $\xvref$ 
%denotes
denoted 
a (fixed) reference solution.
The rate of change $\dot \xv$ 
%is
was 
approximated by
\begin{equation}
	\frac{\mathrm d \xv}{\mathrm dt} \approx \frac{\mathrm d \xvtilde}{\mathrm dt} 
	= \Jm_g (\xvred) \frac{\mathrm d \xvred}{\mathrm dt} , \qquad
	\Jm_g (\xvred) = \frac{\partial \gv}{\partial \xvred} \in \reals^{N_s \times n_s} ,
	\label{eq:rate_nm-rom}
\end{equation}
where the Jacobian $\Jm_g$ 
%spans
spanned 
the tangent space to the manifold at $\xvred$.
%Initial 
The initial
conditions for the generalized coordinates 
%are
were 
given by $\xvred_0 = \gv^{-1}(\xv_0 - \xvref)$, where $\gv^{-1}$ 
%denotes
denoted 
the inverse function to $\gv$.

Note that \emph{linear subspace} methods are included in the above relations as the special case where $\gv$ is a linear function, which can be written in terms of a \emph{constant} matrix $\Phim \in \reals^{N_s \times n_s}$, i.e., $\gv (\xvred) = \Phim \xvred$.
In this case, the approximation to the solution in Eq.~\eqref{eq:state_nm-rom} and their respective rates in Eq.~\eqref{eq:rate_nm-rom} are given by 
\begin{equation}
	\gv (\xvred) = \Phim \xvred \Rightarrow
	\xv \approx \xvtilde = \xvref + \Phim \xvred , \qquad
	\frac{\mathrm d \xv}{\mathrm dt} \approx \frac{\mathrm d \xvtilde}{\mathrm dt} = \Phim \frac{\mathrm d \xvred}{\mathrm dt} .
	\label{eq:LS-ROM}
\end{equation}
As opposed to the nonlinear manifold Eq.~\eqref{eq:rate_nm-rom}, the tangent space of the (linear) solution manifold is constant, i.e., $\Jm_g = \Phim$; see \cite{Benner.2015}.
%Linear projection-based methods essentially differ in how solution subspaces are constructed, see, e.g.,  \citeauthor{Benner.2015}~\cite{Benner.2015}.
%
% CMES style rewriting
%Benner \emph{et al.}~\cite{Benner.2015}.
%Exemplary, 
We also mention the example of eigenmodes as a classical choice for basis vectors of reduced subspaces \cite{Craig.1968}. 
%\emph{Substructuring methods} are all about enriching subspaces constructed from mode shapes of individual components by appropriate `interface' modes that describe the interaction among components, see, e.g., the pioneering work of \citeauthor{Craig.1968}~\cite{Craig.1968}.
%
% CMES style rewriting
%Craig \emph{et al.}~\cite{Craig.1968}.  

%TODO: POD (connect to above).

%\citeauthor{kim.2020a}~\cite{kim.2020a}
%
% CMES style rewriting
%Kim \emph{et al.}~
The authors of 
\cite{kim.2020a} 
defined a \emph{residual function} $\rvtilde : \realsn \times \realsn \times \reals \times \spaceD \rightarrow \realsN$ in the reduced set of coordinates by rewriting the governing ODE Eq.~\eqref{eq:NL_dynamical_system} and substituting the approximation of the state Eq.~\eqref{eq:state_nm-rom} and its rate Eq.~\eqref{eq:rate_nm-rom}:\footnote{
	The tilde above the symbol for the residual function $\rvtilde$ in Eq.~\eqref{eq:NL-Galerkin-residual} was used to indicate an approximation, consistent with the use of the tilde in the approximation of the state in Eq.~\eqref{eq:state_nm-rom}, i.e., $\xv \approx \xvtilde$.   
}
%\begin{equation}
%	\rv(\dot \xv, \xv, t; \muv) = \dot \xv - \fv(\xv, t; \muv) .
%\end{equation}
\begin{equation}
%	\rvtilde(\dot \xvred, \xvred, t; \muv)
    \rvtilde(\dt \xvred, \xvred, t; \muv)
	= \Jm_g (\xvred) \dt \xvred - \fv (\xvref + \gv(\xvred), t; \muv) 
	%\rv(\Jm_g (\xvred) \dot \xvred, \xvref + \gv (\xvred), t; \muv)
	\label{eq:NL-Galerkin-residual}
\end{equation}
As $N_s > n_s$, the system of equations $\rvtilde(\dt \xvred, \xvred, t; \muv)  = \mathbf 0$ is over-determined and no solution exists in general. 
For this reason, a least-squares solution that minimized the square of the residual's Euclidean norm, which was denoted by $\Vert ( \cdot ) \Vert_2$, was sought instead \cite{kim.2020a}: 
\begin{equation}
	\dt \xvred = \argmin_{\vvhat \in \realsn} \left\Vert \rvtilde (\vvhat, \xvred, t; \muv) \right\Vert_2^2 .
	\label{eq:NL-Galerkin-min}
\end{equation}
Requiring the derivative of the (squared) residual to vanish, we obtain the following set of equations,\footnote{
	As the authors of~\cite{kim.2020a} omitted a step-by-step derivation, we introduce it here for the sake of completeness.
}
%{\color{red} [NOTE: Did ``we'' obtain Eq.~\eqref{eq:NL_dynamical_system_proj} or did the authors of \cite{kim.2020a} obtain it?  If ``we'' obtained this equation, then it is best to mention that this equation (or the detailed derivation) was not present in \cite{kim.2020a}.]}
\begin{equation}
	\frac{\partial}{\partial \dt \xvred} \left\Vert \rvtilde ( \dt \xvred, \xvred, t; \muv) \right\Vert_2^2
	= 2 \Jm_g^T \left( \Jm_g \dt \xvred - \fv (\xvref + \gv(\xvred), t; \muv)  \right)
	= \mathbf 0 ,
	\label{eq:NL_dynamical_system_proj}
\end{equation}
which can be rearranged for the rate of the reduced vector of generalized coordinates as:
\begin{equation}
	\dt \xvred = \invJm_g (\xvred) \fv (\xvref + \gv(\xvred), t; \muv) , \qquad
	\xvred (0; \muv) = \xvred_0(\muv) , \qquad
	\invJm_g = ( \Jm_g^T \Jm_g )^{-1} \Jm_g^T .
	\label{eq:NM-ROM}
\end{equation}
The above system of ODEs, in which $\invJm_g$ denotes the \emph{Moore-Penrose inverse} of the Jacobian $\Jm_g \in \reals^{N_s \times n_s}$, which is also referred to as \emph{pseudo-inverse}, constitutes the ROM corresponding to the FOM governed by Eq.~\eqref{eq:NL_dynamical_system}. 
Equation \eqref{eq:NL_dynamical_system_proj} reveals that the minimization problem is equivalent to a projection of the full residual onto the low-dimensional subspace $\subspace$ by means of the Jacobian $\Jm_g$.
Note that the rate of the generalized coordinates $\dt \xvred$ lies in the same tangent space, which is spanned by $\Jm_g$. 
Therefore, the authors of \cite{kim.2020a} described the projection Eq.~\eqref{eq:NL_dynamical_system_proj} as \emph{Galerkin projection} and the ROM Eq.~\eqref{eq:NM-ROM} as \emph{nonlinear manifold (NM) Galerkin ROM}. 
To construct the solutions, suitable time-integration schemes needed to be applied to the semi-discrete ROM Eq.~\eqref{eq:NM-ROM}.

%
% CMES style rewriting
%Kim et al.~\cite{kim.2020a}, also present 
An alternative approach was also presented in \cite{kim.2020a} for the construction of ROMs, in which time-discretization was performed prior to the projection onto the low-dimensional solution subspace. 
For this purpose a uniform time-discretization with a step size $\Delta t$ was assumed; the solution at time $t_n = t_{n-1} + \Delta t = n \Delta t$ was denoted by $\xv_n = \xv(t_n; \muv)$.
%
% CMES style rewriting
%In the development of their method, the authors considered implicit time integration schemes, among which they chose the \emph{backward Euler} method, i.e.,
To develop the method for implicit integration schemes, the \emph{backward Euler} method was chosen as example \cite{kim.2020a}:
\begin{equation}
	\xv_n - \xv_{n-1} = \Delta t \fv_n ,
	\label{eq:backward_euler}
\end{equation}
%as role model.
The above integration rule implies that span of rate $\fv_n$, which is given by evaluating (or by taking the `snapshot' of) the nonlinear function $\fv$, is included in the span of the state (`solution snapshot') $\xv_n$:
\begin{equation}
	\spanop \left\{  \fv_n \right\} \subseteq \spanop \left\{  \xv_{n-1}, \xv_n \right\} 
	\qquad \rightarrow \qquad
	\spanop \left\{  \fv_1, \ldots, \fv_{N_t} \right\} \subseteq \spanop \left\{  \xv_{0}, \ldots , \xv_{N_t} \right\} .
	\label{eq:spans_f_x}
\end{equation}

%{\color{red} [NOTE: 2022.08.02 - Please check the rewritten last sentence above Eq.~\eqref{eq:spans_f_x} to use singular, instead of plural, for vector quantities.]}

For the backward Euler scheme Eq.~\eqref{eq:backward_euler}, a residual function was defined in \cite{kim.2020a} as the difference 
%\begin{equation}
%	\rv (\xv_n; \xv_{n-1}, \muv) = \xv - \xv_{n-1} - \Delta t \fv_n .
%\end{equation}
%%
%\begin{equation}
%	\begin{aligned}
%		\rv_{BE}^n (\xvred_n; \xvred_{n-1}, \muv)
%		&= \rvtilde_{BE}^n (\xvref + \gv(\xvred_n); \xvref + \gv(\xvred_{n-1}), \muv) \\
%		&= \gv(\xvred_n) - \gv(\xvred_{n-1}) - \Delta t \fv(\xvref + \gv(\xvred_n) , t_n; \muv)
%	\end{aligned}
%\end{equation}
%
\begin{equation}
	\rvtilde_{BE}^n (\xvred_n; \xvred_{n-1}, \muv) = \gv(\xvred_n) - \gv(\xvred_{n-1}) - \Delta t \fv (\xvref + \gv(\xvred_n), t_n; \muv) .
	\label{eq:NM-LSPG-r}
\end{equation}
Just as in the time-continuous domain, the system of equations $\rvtilde_{BE}^n(\xvred_n; \xvred_{n-1}, \muv) = \mathbf 0$ is over-determined and, hence, was reformulated as a least-squares problem for the generalized coordinates $\xvred_n$:
\begin{equation}
	\xvred_n = \frac 1 2 \argmin_{\vvhat \in \realsn} \left\Vert \rvtilde_{BE}^n(\vvhat; \xvred_{n-1}, \muv) \right\Vert_2^2 .
	\label{eq:NM-LSPG-ROM}
\end{equation}

\begin{figure}[h]
	\centering
	\includegraphics[width=0.8\textwidth,clip=true, trim=0 0cm 0 0]{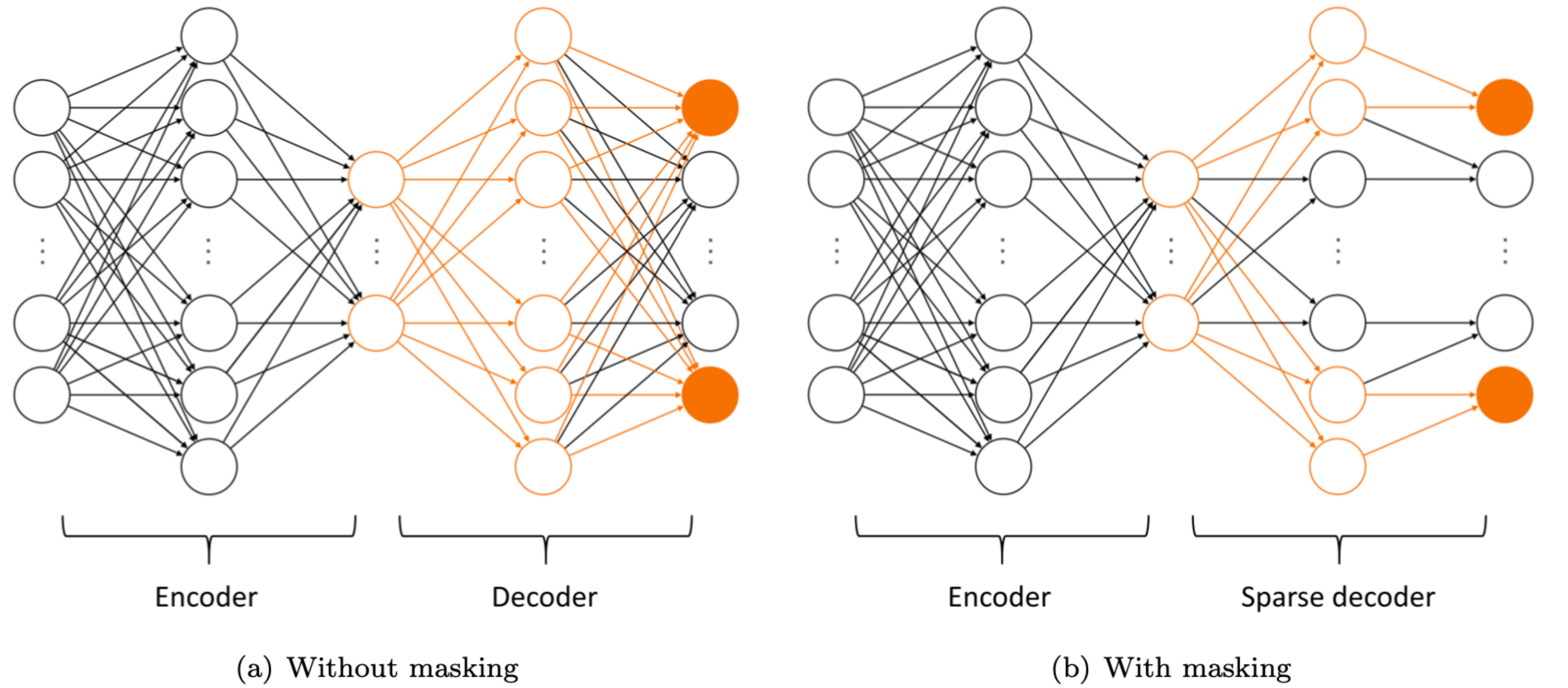}
	\caption{
		\emph{Dense vs. shallow decoder networks} (Section~\ref{sc:autoencoder}). 
		%		Neurons (``nodes'') and connections  (``edges") in orange lie in the ``active" path from selected outputs (orange) to the decoders' input, i.e., contribute to the respective outputs.
		Contributing neurons (orange ``nodes'') and connections  (orange ``edges'') lie in the ``active'' paths arriving at the selected outputs (solid orange ``nodes'') from the decoder's inputs.
		%		In dense networks (a), every neuron of some particular layer has connections to all neurons of both the previous and the succeeding layers (except for neurons of input and output layers, of course). 
		In dense networks as the one in (a), each neuron in a  layer is connected to all other neurons in both the preceeding layer (if it exists) and in the succeeding layer (if it exists).  
		Fully-connected networks are characterized by dense weight matrices, see Section~\ref{sc:network-layer-details}.
		In sparse networks as the decoder in (b), several connections among successive layers are dropped, resulting in sparsely populated weight matrices.
		\footnotesize (Figure reproduced with permission of the authors.)
	}
	\label{fig:Choi-2021-masking-1}
\end{figure}

%To solve the least-squares problem, the Gauss-Newton method with starting point $\xvred_{n-1}$ is applied, which, unlike the time-continuous case, results in a projection involving not only $\Jm_g$, but also the Jacobian of the nonlinear function $\Jm_f = \partial \fv / \partial \xv_n$.\footnote{
%	\label{fn:LSPG}
%	\begin{equation}
%		\begin{aligned}
%			\rvtilde_{BE}^n (\xvred_n; \xvred_{n-1}, \muv) 
%			&\approx \rvtilde_{BE}^n (\xvred_{n-1}; \xvred_{n-1}, \muv) + \left. \frac{\partial \rvtilde_{BE}^n}{\partial \xvred_n} \right|_{\xvred_n = \xvred_{n-1}} \left( \xvred_{n} - \xvred_{n-1} \right) \\
%			&= - \Delta t \fv_{n-1} + \left( \Jm_g (\xvred_{n-1}) - \Delta t \Jm_f ( \xvref + \gv(\xvred_{n-1}) ) \Jm_g (\xvred_{n-1}) \right) \left( \xvred_n - \xvred_{n-1} \right) \\
%			&= - \Delta t \fv_{n-1} + \left( \I - \Delta t \Jm_f ( \xvref + \gv(\xvred_{n-1}) ) \right) \Jm_g (\xvred_{n-1}) \left( \xvred_n - \xvred_{n-1} \right)
%		\end{aligned}
%	\label{eq:NM-LSPG-residual}
%	\end{equation}
%%
%so we obtain:
%	\begin{equation}
%	\Am^T \left( - \Delta t \fv_{n-1} + \Am \left( \xvred_n - \xvred_{n-1} \right)  \right) = \mathbf 0 , \qquad
%		\Am = \left( \I - \Delta t \Jm_f ( \xvref + \gv(\xvred_{n-1}) ) \right) \Jm_g (\xvred_{n-1}) 
%	\end{equation}
%%
%\begin{equation}
%	\xvred_n = \xvred_{n-1} + \Am^\dagger \fv_{n-1} , \qquad
%	\Am^\dagger = ( \Am^T \Am )^{-1} \Am^T 
%\end{equation}
%}
To solve the least-squares problem, the Gauss-Newton method with starting point $\xvred_{n-1}$ was applied, i.e., the residual Eq.~\eqref{eq:NM-LSPG-r} was expanded into a first-order Taylor polynomial in the generalized coordinates \cite{kim.2020a}:
%\begin{equation}
%	\begin{aligned}
%			\rvtilde_{BE}^n (\xvred_n; \xvred_{n-1}, \muv) 
%			&\approx \rvtilde_{BE}^n (\xvred_{n-1}; \xvred_{n-1}, \muv) + \left. \frac{\partial \rvtilde_{BE}^n}{\partial \xvred_n} \right|_{\xvred_n = \xvred_{n-1}} \left( \xvred_{n} - \xvred_{n-1} \right) \\
%			&= - \Delta t \fv_{n-1} + \left( \Jm_g (\xvred_{n-1}) - \Delta t \Jm_f ( \xvref + \gv(\xvred_{n-1}) ) \Jm_g (\xvred_{n-1}) \right) \left( \xvred_n - \xvred_{n-1} \right) \\
%			&= - \Delta t \fv_{n-1} + \left( \I - \Delta t \Jm_f ( \xvref + \gv(\xvred_{n-1}) ) \right) \Jm_g (\xvred_{n-1}) \left( \xvred_n - \xvred_{n-1} \right) .
%		\end{aligned}
%\label{eq:NM-LSPG-residual}
%\end{equation}
\begin{align}
	\rvtilde_{BE}^n (\xvred_n; \xvred_{n-1}, \muv) 
	&\approx \rvtilde_{BE}^n (\xvred_{n-1}; \xvred_{n-1}, \muv) + \left. \frac{\partial \rvtilde_{BE}^n}{\partial \xvred_n} \right|_{\xvred_n = \xvred_{n-1}} \left( \xvred_{n} - \xvred_{n-1} \right) 
	\nonumber
	\\
	&= - \Delta t \fv_{n-1} + \left( \Jm_g (\xvred_{n-1}) - \Delta t \Jm_f ( \xvref + \gv(\xvred_{n-1}) ) \Jm_g (\xvred_{n-1}) \right) \left( \xvred_n - \xvred_{n-1} \right) 
	\nonumber
	\\
	&= - \Delta t \fv_{n-1} + \left( \I - \Delta t \Jm_f ( \xvref + \gv(\xvred_{n-1}) ) \right) \Jm_g (\xvred_{n-1}) \left( \xvred_n - \xvred_{n-1} \right) .
	\label{eq:NM-LSPG-residual}
\end{align}
Unlike the time-continuous case, the Gauss-Newton method results in a projection involving not only the Jacobian $\Jm_g$ defined in Eq.~\eqref{eq:rate_nm-rom}, but also the Jacobian of the nonlinear function $\fv$, i.e., $\Jm_f = \partial \fv / \partial \xv_n$.
Therefore, the resulting reduced set of algebraic equations obtained upon a projection of the fully discrete FOM was referred to as \emph{nonlinear manifold least-squares Petrov-Galerkin (NM-LSPG)} ROM \cite{kim.2020a}.

\subsubsection{Autoencoder}
\label{sc:autoencoder}
Within
%
% CMES style rewriting 
%their
the 
NS-ROM approach, 
%
% CMES style rewriting
%Kim et al.~\cite{kim.2020a} proposed to construct 
it was proposed in \cite{kim.2020a} to construct
$\gv$ by training an \emph{autoencoder}, i.e., a neural network that reproduces its input vector:
\begin{equation}
	\xv \approx \xvtilde = \autoenc (\xv) = \dec(\enc(\xv)).
\end{equation}
As the above relation reveals, autoencoders are typically composed from two parts, i.e., $\autoenc (\xv) = (\dec \circ \enc)(\xv)$: 
The \emph{encoder} `codes' inputs $\xv$ into a so-called \emph{latent state} $\hv = \enc(\xv)$ (not to be confused with hidden states of neural networks).
The \emph{decoder} then reconstructs (`decodes') an approximation of the input from the latent state, i.e., $\xvtilde = \dec(\hv)$.
Both encoder and decoder can be represented by neural networks, e.g., feed-forward networks as done in 
%
% CMES style rewriting
%Kim et al.~\cite{kim.2020a}.
\cite{kim.2020a}.
As 
%
% CMES style rewriting
%Goodfellow et al.~\cite{Goodfellow.2016} 
the authors of \cite{Goodfellow.2016}
noted, an autoencoder is {``not especially useful''} if it exactly learns the identity mapping for all possible inputs $\xv$.
Instead, autoencoders are typically restricted in some way, e.g., by reducing the dimensionality of the latent space as compared to the dimension of inputs, i.e., $\dim(\hv) < \dim(\xv)$.
The restriction forces an autoencoder to focus on those aspects of the input, or input `features', which are essential for the reconstruction of the input.
The encoder network of such \emph{undercomplete autoencoder}\footnote{
	See 
	%
	% CMES style rewriting
%	Goodfellow et al.~\cite{Goodfellow.2016}, 
	\cite{Goodfellow.2016},
	Chapter 14, p.493.
} 
performs a \emph{dimensionality reduction}, which is exactly 
% don't use "we" to avoid misunderstanding that we did this
%what we aim for in
the aim of
projection-based methods for constructing ROMs.
Using nonlinear encoder/decoder networks, undercomplete autoencoders can be trained to represent low-dimensional subspaces of solutions to high-dimensional dynamical systems governed by Eq.~\eqref{eq:NL_dynamical_system}.\footnote{
	In fact, linear decoder networks combined with MSE-loss are equivalent to the \emph{Principal Components Analysis} (PCA) (see 
	%
	% CMES style rewriting
%	Goodfellow et al.~\cite{Goodfellow.2016}, 
    \cite{Goodfellow.2016},
	p.494), which, in turn, is equivalent to the discrete variant of POD by means of singular value decomposition (see Remark \ref{rm:Mohan-another-POD}).
}
In particular, the decoder network represents the nonlinear manifold $\subspace$, which described by the function $\gv$ that maps the generalized coordinates of the ROM $\xvred$ onto the corresponding element $\xvtilde$ of FOM's solution space.
Accordingly, the generalized coordinates are identified with the autoencoder's latent state, i.e., $\xvred = \hv$; the encoder network represent the inverse mapping $\gv^{-1}$, which {``captures the most salient features''} of the FOM.
 %
 
%
% CMES style rewriting
%Kim et al.~\cite{kim.2020a} normalized 
The input data, which was formed by the snapshots of solutions $\Xm = [\xv_1, \ldots, \xv_m]$, where $m$ denoted the number of snapshots, was normalized when training the network \cite{kim.2020a}. 
The shift by the referential state $\xvref$ and scaling by $\xvscale$, i.e.,
\begin{equation}
 	\xvnormal = \xvscale \odot \left( \xv - \xvref \right) \in \mathbb R^{N_s} ,
 	\label{eq:ROM_normalization}
\end{equation}
is such that each component of the normalized input ranges from $[-1, 1]$ or $[0,1]$.\footnote{
	The authors of \cite{kim.2020a} did not provide further details.
}
The encoder maps a FOM's (normalized) solution snapshot onto a corresponding low-dimensional latent state $\xvred$, i.e., vector of generalized coordinates:
\begin{equation}
	\xvred = \gv^{-1}(\xv) = \enc (\xvnormal) = \enc(\xvscale \odot \left( \xv - \xvref \right)) \in \mathbb R^{n_s} .
\end{equation}
The decoder reconstructs the high-dimensional solution $\xvtilde$ from the low-dimensional state $\xvred$ inverting the normalization Eq.~\eqref{eq:ROM_normalization}:
\begin{equation}
	\xvtilde =  \gv (\xvred) = \xvref + \dec(\xvred) \oslash \xvscale \in \mathbb R^{N_s} ,
	\label{eq:ROM_decoder}
\end{equation}
where the operator $\oslash$ denotes the element-wise division.
The composition of encoder and decoder give the autoencoder, i.e.,
\begin{equation}
	\xvtilde = \autoenc (\xv) = \xvref + \dec( \enc(\xvscale \odot (\xv - \xvref )) ) \oslash \xvscale ,
\end{equation} 
which is trained in an \emph{unsupervised} way to (approximately) reproduce states $\nfwidetilde{\Xm} = [\xvtilde_1, \ldots, \xvtilde_m] $ from input snapshots $\Xm$.
To train autoencoders, the \emph{mean squared error} (see Sec. \ref{sc:mean-squared-error}) is the natural choice for the loss function, i.e.,
$J = \Vert \Xm - \nfwidetilde{\Xm} \Vert_F / 2 \rightarrow \min$, where the Frobenius norm of matrices is used. 

Both the encoder and the decoder are feed-forward neural networks with a single hidden layer. 
As activation function, 
%
% CMES style rewriting
%Kim et al.~\cite{kim.2020a} 
it was 
proposed in \cite{kim.2020a} to use the sigmoid function (see Figure~\ref{fig:sigmoid} and Section \ref{sc:sigmoid-history}) or the swish function (see Figure~\ref{fig:swish}). 
It remains unspecified, however, which of the two were used in the numerical examples in \cite{kim.2020a}. 
Though the representational capacity of neural networks increases with depth, which is what `deep learning' is all about, 
%
% CMES style rewriting
%Kim et al.~\cite{kim.2020a} 
the authors of \cite{kim.2020a}
deliberately used shallow networks to minimize the (computational) complexity of computing the decoder's Jacobian $\Jm_g$.
To further improve computational efficiency, 
%the authors
they 
proposed to use a {``masked/sparse decoder''}, in which the weight matrix that mapped the hidden state onto the output $\xvtilde$ was sparse. 
As opposed to a fully populated weight matrix of fully-connected networks (Figure~\ref{fig:Choi-2021-masking-1} (a)), in which each component of the output vector depends on all components of the hidden state, the outputs of a sparse decoder (Figure~\ref{fig:Choi-2021-masking-1} (b)) only depend on the selected components of the hidden state. 
% {\color{red} [NOTE: 2022.08.05 - Please check the above corrected paragraph.]}
% 2022.12.05: OK
%

\begin{figure}[h]
	\centering
	\includegraphics[width=0.8\textwidth]{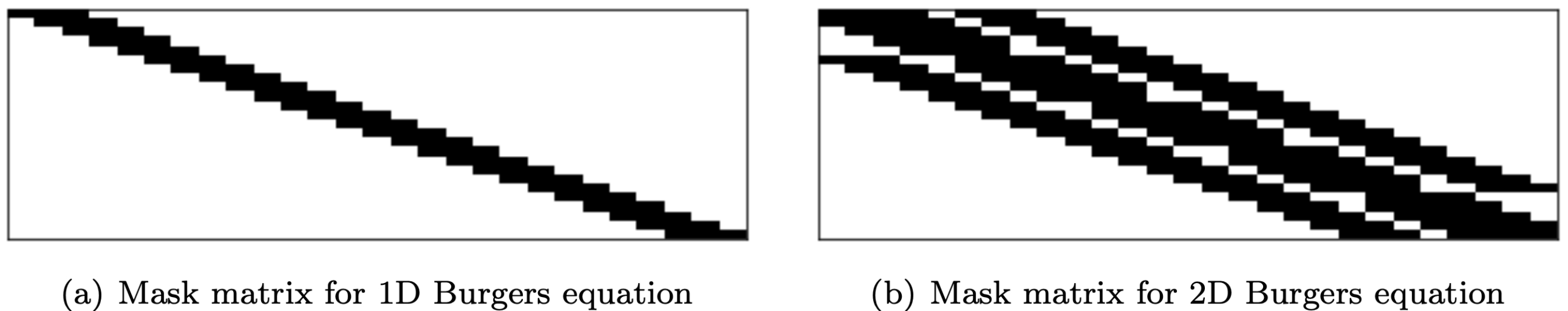}
	\caption{
		\emph{Sparsity masks} (Section~\ref{sc:autoencoder}) used to realize sparse decoders in one- and two-dimensional problems. The structure of the respective binary-valued mask matrices $\Sm$ is inspired by grid-points required in the finite-difference approximation of the Laplace operator in one and two dimensions, respectively.
		\footnotesize (Figure reproduced with permission of the authors.)
	}
	\label{fig:Choi-2021-masks}
\end{figure}

Using our notation for feedforward networks and activation functions introduced in Section~\ref{sc:network-layer-details}, the encoder network has the following structure:
\begin{equation}
	\begin{aligned}
		\xvred = \enc (\xvnormal) := &\Wm_{\rm en}^{(2)} \yv^{(1)} + \bv_{\rm en}^{(2)} , \qquad
		\yv^{(1)} = \sigmoid(\zv^{(1)}) , \\
		\zv^{(1)} = &\Wm_{\rm en}^{(1)} \yv^{(0)} + \bv_{\rm en}^{(1)} , \qquad 
		\yv^{(0)} = \xvnormal ,
	\end{aligned}
\end{equation}
where $\Wm_{\rm de}^{(i)}$, $i=1,2$ are dense matrices.
The masked decoder, on the contrary,  is characterized by sparse connections between the hidden layer and the output layer, which is realized as element-wise multiplication of a dense weight matrix $\Wm_{\rm de}^{(2)}$ with a binary-valued ``mask matrix'' $\Sm$ reflecting the connectivity among the two layers: 
\begin{equation}
	\begin{aligned}
		\xvtilde \odot \xvscale - \xvref = \dec (\xvred) := \Sm \, \odot \, &\Wm_{\rm de}^{(2)} \yv^{(1)} + \bv_{\rm de}^{(2)} , \qquad
		\yv^{(1)} = \sigmoid(\zv^{(1)}) , \\
		\zv^{(1)} = &\Wm_{\rm de}^{(1)} \yv^{(0)} + \bv_{\rm de}^{(1)} , \qquad 
		\yv^{(0)} = \xvred .
	\end{aligned}
\end{equation}
The structure of the sparsity mask $\Sm$, in turn, is inspired by the pattern (``stencil") of grid-points involved in a finite-difference approximation of the Laplace operator, see Figure~\ref{fig:Choi-2021-masks} for the one- and two-dimensional cases.

\subsubsection{Hyper-reduction}
\label{sc:hyper-reduction}
Irrespective of how small the dimensionality of the ROM's solution subspace $\subspace$ is, we cannot expect a reduction in computational efforts in nonlinear problems. 
Both the time-continuous NM-Galerkin ROM in Eq.~\eqref{eq:NM-ROM} and the time-discrete NM-LSPG ROM in Eq.~\eqref{eq:NM-LSPG-ROM} require repeated evaluations of the nonlinear function $\fv$ and, in case implicit time-integration schemes, also its Jacobian $\Jm_f$, whose computational complexity is determined by the FOM's dimensionality.
The term \emph{hyper-reduction} subsumes techniques in model-order reduction to overcome the necessity for evaluations that scale with the FOM's size.
Among hyper-reduction techniques, the \emph{Discrete Empirical Interpolation Method} (DEIM)~\cite{Chaturantabut.2010} and \emph{Gauss-Newton with Approximated Tensors} (GNAT)~\cite{Carlberg.2011} have gained significant attention in recent years.

%
% CMES style rewriting
%In their approach, 
%\citeauthor{kim.2020a} \cite{kim.2020a} 
%Kim \emph{et al.}~\cite{kim.2020a} used 
In \cite{kim.2020a}, 
a variant of GNAT relying on solution snapshots for the approximation of the nonlinear residual term $\rvtilde$ was used, and therefore the reason to extend the GNAT acronym with ``SNS'' for \emph{'solution-based subspace'} (SNS), i.e., \emph{GNAT-SNS}, see \cite{Choi.2020}.
The idea of DEIM, GNAT and their SNS variants takes up the leitmotif of projection-based methods: The approximation of the full-order residual $\rvtilde$ is in turn linearly interpolated by a low-dimensional vector $\rvred$ and an appropriate set of basis vectors $\phiv_{r, i} \in \realsN$, $i = 1, \ldots, n_r$: 
%
%GNAT-SNS:
\begin{equation}
	\rvtilde \approx \Phim_r \rvred , \qquad
	\Phim_r = \left[ \phiv_{r,1}, \ldots, \phiv_{r,n_r} \right] \in \reals^{N_s \times n_r}, \qquad
	n_s \leq n_r \ll N_s .
	\label{eq:residual_interpolation_hr}
\end{equation}

In DEIM methods, the residual $\rvtilde$ of the time-continuous problem in Eq.~\eqref{eq:NL-Galerkin-residual} is approximated, whereas $\rvtilde$ it to be replaced by $\rvtilde_{BE}^n$ in the above relation when applying GNAT, which builds upon Petrov-Galerkin ROMs as in Eq.~\eqref{eq:NM-LSPG-residual}.
Both DEIM and GNAT variants use \emph{gappy POD} to determine the basis $\Phim_r$ for the interpolation of the nonlinear residual.
Gappy POD originates in a method for image reconstruction proposed 
%
% CMES style rewriting
%by 
%\citeauthor{Everson.1995} \cite{Everson.1995}
%Everson \emph{et al.}~\cite{Everson.1995}
\cite{Everson.1995}  
under the name of a 
%\emph{Karhunen–Lo{\`{e}}ve procedure}.
%Karhunen–Lo\`eve procedure.
Karhunen-Lo\`eve procedure, in which images were reconstructed from individual pixels, i.e., from \emph{gappy data}.
In the present context of MOR, gappy POD aims at reconstructing the full-order residual $\rvtilde$ from a small, $n_r$-dimensional subset of its components. 

The matrix $\Phim_r$ was computed by a singular value decomposition (SVD) on the snapshots of data. 
Unlike original DEIM and GNAT methods, their SNS variants did not use snapshots of the nonlinear residual $\rvtilde$ and $\rvtilde_{BE}^n$, respectively, but SVD was performed on solution snapshots $\Xm$ instead. 
The use of solution snapshots was motivated by the fact that the span of snapshots of the nonlinear term $\fv$ was included in the span of of solution snapshots for conventional time-integration schemes, see Eq.~\eqref{eq:spans_f_x}.
%The vector of \emph{``generalized coordinates of the nonlinear residual term''} $\rvred$.
%In other words, $\rvred$ comprises $n_r$ components of the FOM's residual $\rvtilde$, by which the remaining $N_s - n_r$ components are linearly interpolated using the basis $\Phim_r$.
The vector of \emph{``generalized coordinates of the nonlinear residual term''} minimizes the square of the Euclidean distance of selected components of full-order residual $\rvtilde$ and respective components of its reconstruction:
\begin{equation}
	\rvred = \argmin_{\vvhat \in \realsr} \frac 1 2 \left\Vert \Zm^T \left( \rvtilde- \Phim_r \vvhat \right) \right\Vert_2^2 , 
	\quad
	\Zm^T = \left[ \ev_{p_1}, \ldots, \ev_{p_{n_z}}  \right]^T \in \reals^{n_z \times N_s} , 
	\quad
	n_s \leq n_r \leq n_z \ll N_s .
	\label{eq:hyperred}
\end{equation}

The matrix $\Zm^T$, which was referred to as \emph{sampling matrix}, extracts a set of components from the full-order vectors $\vv \in \realsN$. 
For this purpose, the sampling matrix was built from unit vectors $\ev_{p_i} \in \realsN$, having the value one at the $p_i$-th component, which corresponded to the component of $\vv$ to be selected (`sampled'), and the value zero otherwise. 
The components selected by $\Zm^T$ are equivalently described by the (ordered) set of \emph{sampling indices} $\mathcal I = \left\{ p_1, \ldots, p_{n_z} \right\}$, which are represented by the vector $\iv = [
	p_1 , \ldots , p_{n_z} ]^T \in \reals^{n_z}$.
As the number of components of $\rvtilde$ used for its reconstruction may be larger than the dimensionality of the reduced-order vector $\rvred$, i.e., $n_z \geq n_r$, Eq.~\eqref{eq:hyperred} generally constitutes a least-squares problem, the solution of which follows as
\begin{equation}
	\rvred = ( \Zm^T \Phim_r )^\dagger \Zm^T \rvtilde .
	\label{eq:residual-HR_red}
\end{equation} 

Substituting the above result in Eq.~\eqref{eq:residual_interpolation_hr}, 
%
% rewritten to avoid using "we" if the result was in the original paper
%we can interpolate 
the FOM's residual can be interpolated using an \emph{oblique projection matrix} $\Pmcal$ as 
\begin{equation}
	\rvtilde \approx \Pmcal \rvtilde , \qquad
	\Pmcal = \Phim_r ( \Zm^T \Phim_r )^\dagger \Zm^T \in \reals^{N_s \times N_s} .
	\label{eq:residual-HR}
\end{equation}
The above representation in terms of the projection matrix $\Pmcal$ somewhat hides the main point of hyperreduction.
In fact, we do not apply $\Pmcal$ to the full-order residual $\rvtilde$, which would be tautological.
Unrolling the definition of $\Pmcal$, we note that $\Zm^T \rvtilde \in \reals^{n_z}$ is a vector containing only a small subset of components of the full-order residual $\rvtilde$.
In other words, to evaluate the approximation in Eq.~\eqref{eq:residual-HR} only $n_z \ll N_s$ of components of $\rvtilde$ need to be computed, i.e., the computational cost no longer scales with the FOM's dimensionality. 

Several methods have been proposed to efficiently construct a suitable set of sampling indices $\mathcal I$, see, e.g., \cite{Chaturantabut.2010, Carlberg.2011, Carlberg.2013}.
These methods share the property of being \emph{greedy algorithms}, i.e., algorithms that sequentially ({``inductively''}) create optimal sampling indices using some suitable metric. % as, e.g., optimality in terms of reconstruction error. 
%\emph{Greedy algorithms} 
%Given a set of $j-1$ sampling indices $\mathcal I = {p_1 , \dots, p_{j-1}}$, the original DEIM algorithm by \citeauthor{Chaturantabut.2010}~\cite{Chaturantabut.2010},
%
% CMES style rewriting
%Chaturantabut \emph{et al.}~\cite{Chaturantabut.2010},
For instance, the authors of \cite{Chaturantabut.2010} selected the $j$-th index corresponding to the component of the gappy reconstruction of the $j$-th POD-mode $\phivhat_{r,j}$ showing the largest error compared to the original mode $\phiv_{r,j}$.
For the reconstruction, the first $j-1$ POD-modes were used, i.e., 
\begin{equation}
	\phiv_{r,j} \approx \phivtilde_{r,j} = \Phim_r \phivhat_{r,j} , \qquad 
	\Phim_r = \left[ \phiv_{r,1}, \ldots, \phiv_{r, j-1} \right] \in \reals^{N_s \times j-1}.
\end{equation}
The reduced-order vector $\phivhat_{r,j} \in \reals^{j-1}$ minimizes the square of the Euclidean distance between $j-1$ components (selected by $\Zm^T \in \reals^{j \times N_s}$) of the $j$-th mode $\phiv_{r,j}$ and its gappy reconstruction $\phivtilde_{r,j}$:
\begin{equation}
		\phivhat_{r,j} = \argmin_{\vvhat \in \reals^{j-1}} \frac 1 2 \left\Vert \Zm^T \left( \phiv_{r,j} - \Phim_r \vvhat \right) \right\Vert^2_2 .
		\label{eq:deim-mode-reconstruction}
\end{equation}

The key idea of the greedy algorithm is to select additional indices to minimize the error of the gappy reconstruction. 
Therefore, the component of the reconstructed mode that differs most (in terms of magnitude) from the original mode defines the $j$-th sampling index:
\begin{equation}
	p_j = \argmax_{i \in \{1, \ldots, N_s\} \backslash \mathcal I} \left\vert ( \phiv_{r,j} - \Phim_r (\Zm^T \Phim_r)^\dagger \Zm^T \phiv_{r,j} )_i \right\vert .
\end{equation}
A pseudocode representation of the greedy approach for selecting sampling indices is given in Algorithm~\ref{algo:deim}.
To start the algorithm, the first sampling index is chosen according to the largest (in terms of magnitude) component of the first POD-mode, i.e., $p_1 = \arg \max_{i \in \{1, \ldots, N_s\}} \vert \left( \phiv_{r,1} \right)_i \vert$.

\begin{algorithm}
	{\bf DEIM algorithm for construction of sampling indices} \\
	\KwData{POD-basis $\{  \phiv_{r,1}, \ldots, \phiv_{r,n_r} \}$, in descending order according to singular values  }
	\KwResult{Set of sampling indices $\mathcal I = \left\{ p_1, \ldots, p_{n_z} \right\}$} 
%	Initialize set of sampling indices $\mathcal I \leftarrow \emptyset$
%	\;
	$\blacktriangleright$ Initialization \\
	Select first sampling index $p_1 = \arg \max_{i \in \{1, \ldots, N_s\}} \vert \left( \phiv_{r,1} \right)_i \vert$ 
	\;
	Set $\Phim_r \leftarrow [\phiv_{r,1}] $ , $\Zm \leftarrow [ \ev_{p_1} ]$ ,  $\iv = [p_1]$ , $\mathcal I = \left\{ p_1 \right\}$ \;
	\For{$j = 2, \ldots, n_z$}{
		$\blacktriangleright$ Select $j$-th sampling index $p_j$ \\
		Solve $(\Zm^T \Phim_r ) \vvhat = \Zm^T \phiv_{r,j}$ for $\vvhat$ \hspace{1em} (cf. Eq.~\eqref{eq:deim-mode-reconstruction}) \; 
		Set $p_j = \arg \max_{i \in \{1, \ldots, N_s\} \backslash \mathcal I} \vert \left( \phiv_{r,j} - \Phim_r \vvhat \right)_i \vert$ \;
		Update $\Phim_r \leftarrow \begin{bmatrix}
			\Phim_r & \phiv_{r,j}
		\end{bmatrix} $ , 
	$\Zm \leftarrow \begin{bmatrix}
		 \Zm & \ev_{p_j}
	\end{bmatrix}$ , 
	$\iv \rightarrow \begin{bmatrix}
		\iv^T &	p_j
	\end{bmatrix}^T$ ,
	$\mathcal I \leftarrow \mathcal I \cup \left\{ p_j \right\}$  \;
	}
	\vphantom{Blank line} 
	\caption{
		\emph{DEIM algorithm for construction of sampling indices} (Section~\ref{sc:hyper-reduction}).
		Starting from the initial ($j=1$) sampling index $p_1$, a sequence of indices $\mathcal I = \{ p_1 , \ldots, p_{n_z} \}$ is iteratively constructed in a \emph{greedy} way: Given the first index $p_1$, subsequent indices are iteratively chosen according to the error in the gappy reconstruction of POD-modes. In particular, the $j$-th sampling index $p_j$ corresponds to that component of the $j$-th POD mode, for which the gappy reconstruction by means of the $j-1$ preceding modes shows the largest error.
	}
	\label{algo:deim}
\end{algorithm}

%\begin{itemize}
%	\item online/offline computations
%\end{itemize}
%%
%To obtain a hyper-reduced Galerkin ROM, \citeauthor{kim.2020a}~\cite{kim.2020a}
%
% CMES style rewriting
%Kim \emph{et al.}~\cite{kim.2020a} 
The authors of \cite{kim.2020a} 
substituted the residual vector $\rvtilde$ by its gappy reconstruction, i.e., $\Pmcal \rvtilde$ in the minimization problem in Eq.~\eqref{eq:NL-Galerkin-min}.
Unlike the original Galerkin ROM in Eq.~\eqref{eq:NL-Galerkin-min}, the rate of the reduced vector of generalized coordinate does not minimize the (square of the) FOM's residual $\rvtilde$, but the corresponding reduced residual $\rvred$, which are related by $\rvtilde = \Phim_r \rvred$:
\begin{equation}
	\dt \xvred 
	= \argmin_{\vvhat \in \realsn} \left\Vert \rvred(\vvhat, \xvred, t; \muv)  \right\Vert_2^2
	= \argmin_{\vvhat \in \realsn} \left\Vert ( \Zm^T \Phim_r )^\dagger \Zm^T \rvtilde (\vvhat, \xvred, t; \muv) \right\Vert_2^2 .
	\label{eq:NL-Galerkin-HR-min}
\end{equation}
From the above minimization problem, 
%we determine 
the ROM's ODEs were determined in \cite{kim.2020a} by taking the derivative with respect to the reduced vector of generalized velocities $\vvhat$, which was evaluated at $\vvhat = \dt \xvred$, i.e.,\footnote{
	Note again that the step-by-step derivations of the ODEs in Eq.~\eqref{eq:NM-ROM-HR} were not given by the authors of~\cite{kim.2020a}, which is why we provide it in our review paper for the sake of clarity.
}
%\begin{equation}
%	\begin{aligned}
%		{\color{red} \dtb{\xvred}}
%	2 \rvred(\dot \xvred, \xvred, t; \muv) 
%	&
%	\left. 
%	\frac{\partial \rvred (\vvhat, \xvred, t; \muv)}{\partial \vvhat} \right|_{\vvhat = \dot \xvred}
%	= 2 ( \Zm^T \Phim_r )^\dagger \Zm^T \rvtilde \left\{  ( \Zm^T \Phim_r )^\dagger \Zm^T \frac{\partial \rvtilde}{\partial \dot \xvred} \right\} 
%	\\
%	&= 2 ( \Zm^T \Phim_r )^\dagger \Zm^T \left( \Jm_g \dot \xvred - \fv (\xvref + \gv(\xvred), t; \muv) \right) ( \Zm^T \Phim_r )^\dagger \Zm^T \Jm_g 
%	\\
%	&= 2 \left( ( \Zm^T \Phim_r )^\dagger \Zm^T \Jm_g \right)^T ( \Zm^T \Phim_r )^\dagger \Zm^T \left( \Jm_g \dot \xvred - \fv (\xvref + \gv(\xvred), t; \muv) \right) = \mathbf 0 ,
%	\end{aligned}
%	\label{eq:NM-ROM-HR-derivation}
%\end{equation}
%
\begin{align}
%			{\color{red} \dtb{\xvred}}
	2 \rvred(\dtb \xvred, \xvred, t; \muv) 
	&
	\left. 
	\frac{\partial \rvred (\vvhat, \xvred, t; \muv)}{\partial \vvhat} 
	\right|_{\vvhat = {\dtbs \xvred}}
%	\right|_{\vvhat = \dot \xvred}
	= 2 ( \Zm^T \Phim_r )^\dagger \Zm^T \rvtilde \left\{  ( \Zm^T \Phim_r )^\dagger \Zm^T \frac{\partial \rvtilde}{\partial \dtb \xvred} \right\} 
	\nonumber
	\\
	&= 2 ( \Zm^T \Phim_r )^\dagger \Zm^T \left( \Jm_g \dtb \xvred - \fv (\xvref + \gv(\xvred), t; \muv) \right) ( \Zm^T \Phim_r )^\dagger \Zm^T \Jm_g 
	\nonumber
	\\
	&= 2 \left( ( \Zm^T \Phim_r )^\dagger \Zm^T \Jm_g \right)^T ( \Zm^T \Phim_r )^\dagger \Zm^T \left( \Jm_g \dtb \xvred - \fv (\xvref + \gv(\xvred), t; \muv) \right) = \mathbf 0 ,
	\label{eq:NM-ROM-HR-derivation}
\end{align}

%{\color{red} [NOTE: 2022.08.04 - Big dot problem with the ``aligned'' environment solved; see the first symbol $\dtb{\xvred}$ in the top expression of Eq.~\eqref{eq:NM-ROM-HR-derivation}.  Small dot problem at the evaluation point $\vvhat = \dt \xvred$ for the derivative also solved.]}

\noindent
where the definition of the residual vector $\rvtilde$ has been introduced in Eq.~\eqref{eq:NL-Galerkin-residual}.
We therefore obtain the following linear system of equations for $\dt \xvred$,
\begin{equation}
	\left( ( \Zm^T \Phim_r )^\dagger \Zm^T \Jm_g \right)^T ( \Zm^T \Phim_r )^\dagger \Zm^T \Jm_g \dt \xvred 
	= \left( ( \Zm^T \Phim_r )^\dagger \Zm^T \Jm_g \right)^T ( \Zm^T \Phim_r )^\dagger \Zm^T \fv (\xvref + \gv(\xvred), t; \muv) ,
\end{equation}
which, using the notion of the pseudo-inverse, is resolved for $\dt \xvred$ to give, complemented by proper initial conditions $\xvred_0(\muv)$, the governing systems of ODEs of the \emph{hyper-reduced nonlinear manifold, least-squares-Galerkin} (NM-LS-Galerkin-HR) ROM 
%
% CMES style
%proposed by \citeauthor{kim.2020a}~\cite{kim.2020a}:
\cite{kim.2020a}:
\begin{equation}
	 \dt \xvred 
	= \left( ( \Zm^T \Phim_r )^\dagger \Zm^T \Jm_g \right)^\dagger ( \Zm^T \Phim_r )^\dagger \Zm^T \fv (\xvref + \gv(\xvred), t; \muv) , \qquad
	\xvred(0; \muv) = \xvred_0(\muv) .
	\label{eq:NM-ROM-HR}
\end{equation}
%
%Taking the derivative with respect to the reduced vector of generalizied velocities $\vvhat$, we obtain the following vector-valued equations, which needs to be resolved for $\dot \xvred$
\begin{rem}
	{ 
		\rm
		\emph{Equivalent minimization problems.}
		Note the subtle difference between the minimization problems that govern the ROMs with and without hyper-reduction, see Eqs.~\eqref{eq:NL-Galerkin-min} and \eqref{eq:NL-Galerkin-HR-min}, respectively: For the case without hyper-reduction, see Eq.~\eqref{eq:NL-Galerkin-min}, the minimum is sought for the approximate \emph{full-dimensional} residual $\rvtilde$. 
		In the hyper-reduced variant Eq.~\eqref{eq:NL-Galerkin-HR-min},  
%		\citeauthor{kim.2020a}~\cite{kim.2020a} 
		%
		% CMES style rewriting
%		Kim \emph{et al.}~\cite{kim.2020a} 
		the authors of \cite{kim.2020a}
		aimed, however, at minimizing the projected residual $\rvred$, which was related to its full-dimensional counterpart by the residual basis matrix $\Phim_r$, i.e., $\rvtilde = \Phim_r \rvred$.
		Using the full-order residual also in the hyper-reduced ROM translates into the following minimization problem:
		\begin{equation}
			\dt \xvred 
			= \argmin_{\vvhat \in \realsn} \left\Vert \Phim_r \rvred(\vvhat, \xvred, t; \muv)  \right\Vert_2^2
			= \argmin_{\vvhat \in \realsn} \left\Vert \Phim_r ( \Zm^T \Phim_r )^\dagger \Zm^T \rvtilde(\vvhat, \xvred, t; \muv) \right\Vert_2^2 .
	%		= \argmin_{\vvhat \in \realsn} \left\Vert ( \Zm^T \Phim_r )^\dagger \Zm^T \rvtilde (\vvhat, \xvred, t; \muv) \right\Vert_2^2 .
		\end{equation}
		Repeating the steps of the derivation in Eq.~\eqref{eq:NM-ROM-HR-derivation} then gives
%		\begin{equation}
%			2 \Phim_r ( \Zm^T \Phim_r )^\dagger \Zm^T \left( \Jm_g \dot \xvred - \fv (\xvref + \gv(\xvred), t; \muv) \right) \Phim_r ( \Zm^T \Phim_r )^\dagger \Zm^T \Jm_g = \mathbf 0
%		\end{equation}
%	%
%		\begin{equation}
%			\left( \Phim_r ( \Zm^T \Phim_r )^\dagger \Zm^T \Jm_g \right)^T \left( \Phim_r ( \Zm^T \Phim_r )^\dagger \Zm^T \Jm_g \dot \xvred - \Phim_r ( \Zm^T \Phim_r )^\dagger \Zm^T \fv (\xvref + \gv(\xvred), t; \muv) \right) = \mathbf 0
%		\end{equation}
%	%
%		\begin{equation}
%			\dot \xvred =
%			\left( \Phim_r ( \Zm^T \Phim_r )^\dagger \Zm^T \Jm_g \right)^\dagger \Phim_r ( \Zm^T \Phim_r )^\dagger \Zm^T \fv (\xvref + \gv(\xvred), t; \muv) 
%		\end{equation}
	%
		\begin{equation}
			\begin{aligned}
				\dtb {\xvred}
				&= \left( \Phim_r ( \Zm^T \Phim_r )^\dagger \Zm^T \Jm_g \right)^\dagger \Phim_r ( \Zm^T \Phim_r )^\dagger \Zm^T \fv (\xvref + \gv(\xvred), t; \muv) \\
				&= \left( ( \Zm^T \Phim_r )^\dagger \Zm^T \Jm_g \right)^\dagger \Phim_r^\dagger \Phim_r ( \Zm^T \Phim_r )^\dagger \Zm^T \fv (\xvref + \gv(\xvred), t; \muv) \\
				&= \left( ( \Zm^T \Phim_r )^\dagger \Zm^T \Jm_g \right)^\dagger ( \Zm^T \Phim_r )^\dagger \Zm^T \fv (\xvref + \gv(\xvred), t; \muv) ,
			\end{aligned}
		\end{equation}
	i.e., using the identity $\Phim_r^\dagger \Phim_r = \I_{n_r}$, we recover exactly the same result as in the hyper-reduced case in Eq.~\eqref{eq:NM-ROM-HR}.
	The only requirement is that the residual basis vectors $\Phi_{r,j}$, $j=1,\ldots,n_r$ need to be linearly independent. 
	%
%		\begin{equation}
%			\dot \xvred =
%			\left( ( \Zm^T \Phim_r )^\dagger \Zm^T \Jm_g \right)^\dagger \Phim_r^\dagger \Phim_r ( \Zm^T \Phim_r )^\dagger \Zm^T \fv (\xvref + \gv(\xvred), t; \muv)
%		\end{equation}
	%
%		\begin{equation}
%			\dot \xvred =
%			\left( \Zm^T \Jm_g \right)^\dagger \Zm^T \fv (\xvref + \gv(\xvred), t; \muv) 
%		\end{equation}
%	double check that!
	}
		%
	%	\begin{equation}
	%		\dot \xvred = \left\{  \Phim_r  (\Zm^T \Phim_r )^\dagger \Zm^T \Jm_g(\xvred) \right\}^\dagger  \Phim_r ( \Zm^T \Phim_r )^\dagger \Zm^T \fv (\xvref + \gv(\xvred), t; \muv) , \qquad
	%		\xvred(0; \muv) = \xvred_0(\muv) .
	%	\end{equation}
	$\hfill\blacksquare$
\end{rem}

\begin{rem}{Further reduction of the system operator?}
	{ \rm
			At first glance, the operator in the ROM's governing equations in Eq.~\eqref{eq:NM-ROM-HR} appears to be further reducible:
			\begin{equation}
					\left( ( \Zm^T \Phim_r )^\dagger \Zm^T \Jm_g  \right)^\dagger ( \Zm^T \Phim_r )^\dagger
					= (  \Zm^T \Jm_g )^\dagger ( \Zm^T \Phim_r ) ( \Zm^T \Phim_r )^\dagger .
			\end{equation}
			Note that the product $( \Zm^T \Phim_r ) ( \Zm^T \Phim_r )^\dagger$ generally does not, however, evaluate to identity, since our particular definition of the pseudo inverse of $\Am \in \reals^{m \times n}$ is a \emph{left inverse}, for which $\Am^\dagger \Am = \I_n$ holds, but $\Am \Am^\dagger \neq \I_m$.
	}
	\hphantom{x}$\hfill\blacksquare$
\end{rem}
%
%The minimization problem gives the following reduced ($n_s$-dimensional) system of ODEs complemented by boundary conditions for the reduced-order vector of generalized coordinates $\xvred$: 
%%
%\begin{equation}
%	\dot \xvred = \left\{ ( \Zm^T \Phim_r )^\dagger \Zm^T \Jm_g(\xvred) \right\}^\dagger ( \Zm^T \Phim_r )^\dagger \Zm^T \fv (\xvref + \gv(\xvred), t; \muv) , \qquad
%	\xvred(0; \muv) = \xvred_0(\muv) .
%	\label{eq:NM-ROM-HR}
%\end{equation}
%

We first consider linear subspace methods, for which the Jacobian $\Jm_g(\xvred) = \Phim$ that spans the tangent space of the solution manifold is constant, see Eq.~\eqref{eq:LS-ROM}. 
Substituting $\Phim \xvred$ for $\gv(\xvred)$ and $\Phim$ for $\Jm_g$, the above set of ODEs in Eq.~\eqref{eq:NM-ROM-HR} governing the NM-Galerkin-HR ROM reduces to the corresponding \emph{linear subspace} (LS-Galerkin-HR) ROM:
\begin{equation}
	\dt \xvred = \left\{ ( \Zm^T \Phim_r )^\dagger \Zm^T \Phim \right\}^\dagger ( \Zm^T \Phim_r )^\dagger \Zm^T \fv (\xvref + \Phim \xvred, t; \muv)  .
\end{equation}
Note that, in linear subspace methods, the operator $\left\{ ( \Zm^T \Phim_r )^\dagger \Zm^T \Phim \right\}^\dagger ( \Zm^T \Phim_r )^\dagger$ is independent of the solution and, hence, \emph{``can be pre-computed once for all''}. 
The products $\Zm^T \Phim_r$, $\Zm^T \Phim$ and $\Zm^T \fv$ need not be evaluated explicitly.

\begin{figure}[h]
	\centering
	\includegraphics[width=0.4\linewidth]{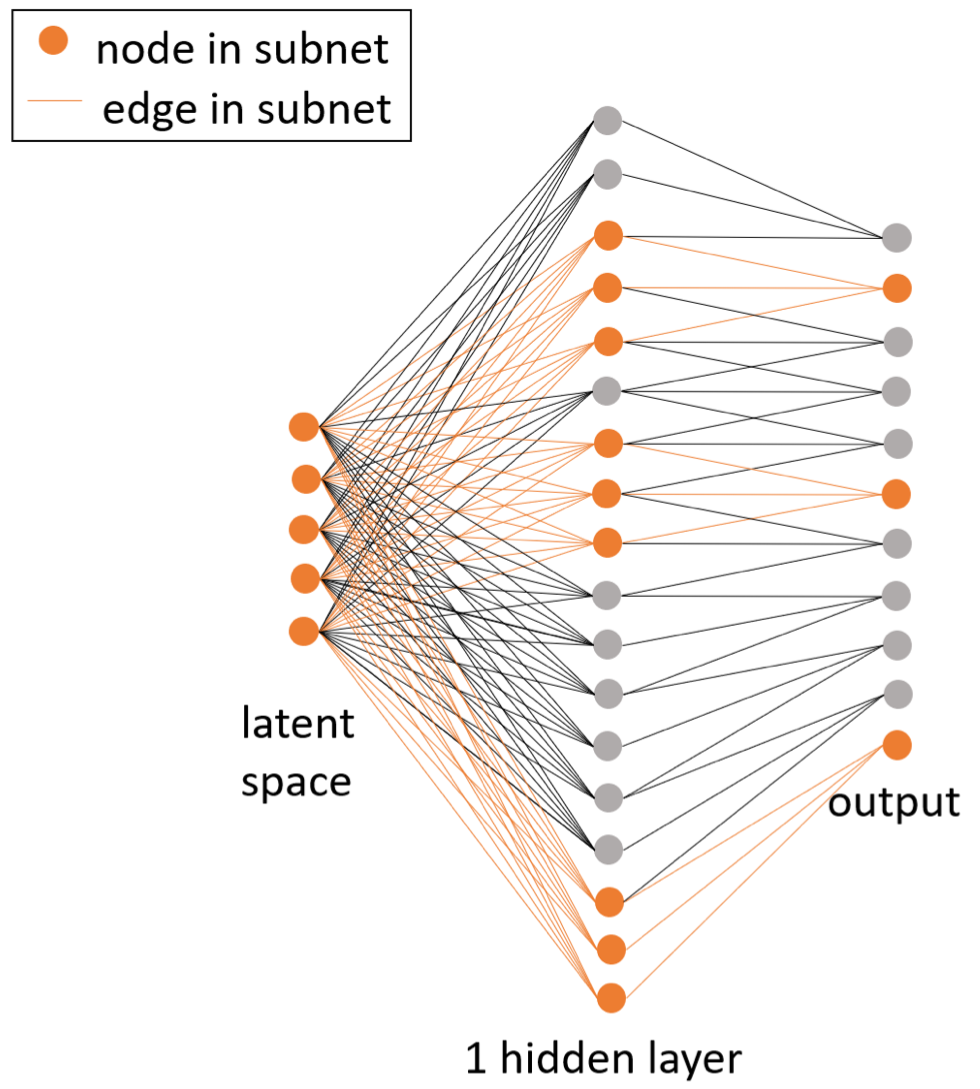}
	\hspace{2em}
	\includegraphics[width=0.4\linewidth]{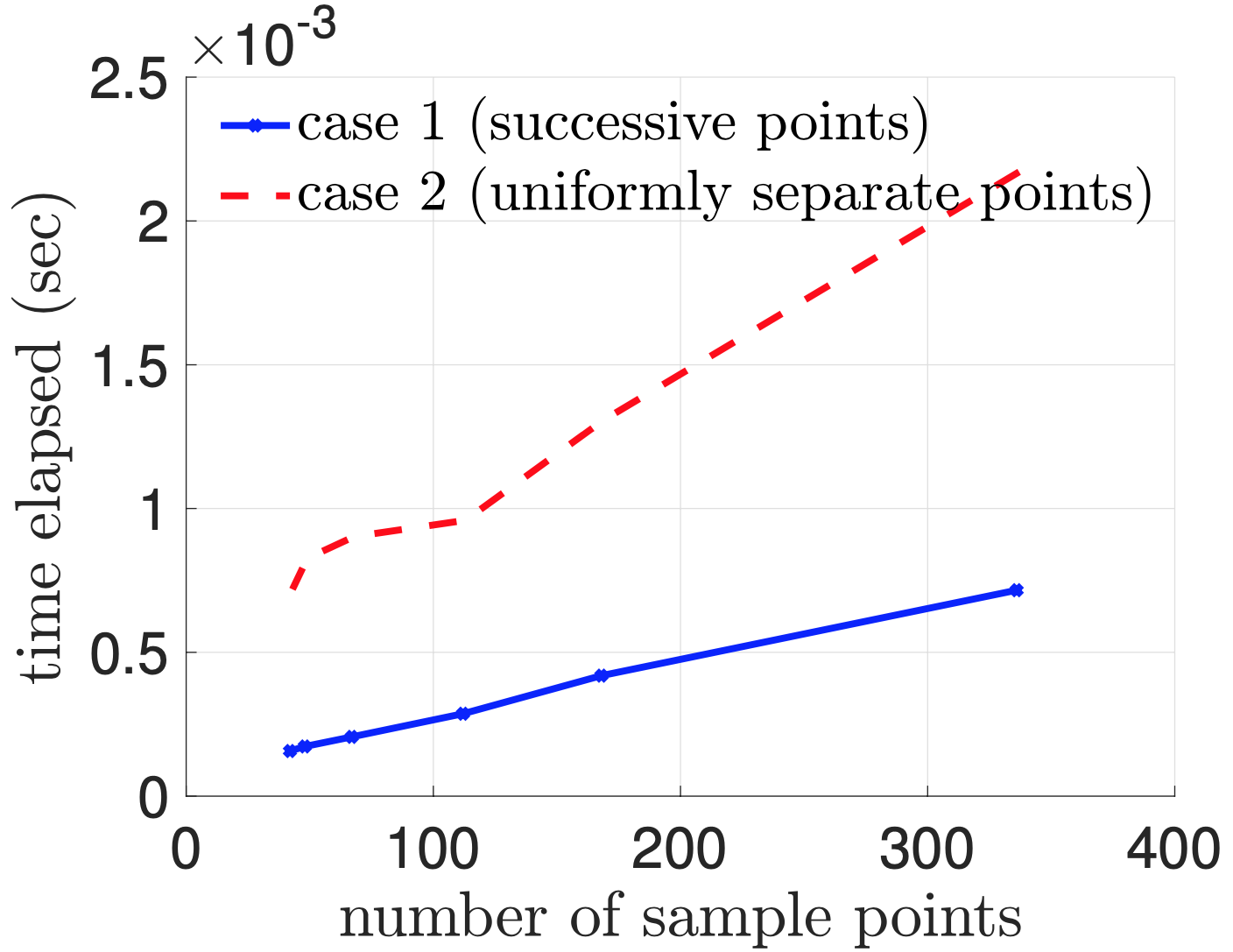}
	\caption{{\it Subnet construction} (Section~\ref{sc:hyper-reduction}).
		To reduce computational cost, a subnet representing the set of active paths, which comprise all neurons and connections needed for the evaluation of selected outputs (highlighted in orange), i.e., the reduced residual $\rvred$, is constructed (left).
		The size of the hidden layer of the subnet depends on which output components of the decoder are needed for the reconstruction of the full-order residual.
		If the full-order residual is reconstructed from from successive outputs of the decoder, the number of neurons in the hidden layer involved in the evaluation becomes minimal due to the specific sparsity patterns proposed. 
		Uniformly distributed components show the least overlap in terms of hidden-layer neurons required, which is why the subnet and therefore the computational cost in the hyper-reduction approach is maximal (right). 
		\footnotesize (Figure reproduced with permission of the authors.)
	}
	\label{fig:Choi-2021-masking-active-path}
\end{figure}

Instead, only those rows of $\Phim_r$, $\Phim$ and $\fv$ which are selected by the sampling matrix $\Zm^T$ need to be extracted or computed when evaluating the above operator. 
In the context of MOR, pre-computations as, e.g., the above operator, but, more importantly, also the computationally demanding collection of full-order solutions and residual snapshots and subsequent SVDs, are attributed to the \emph{offline phase} or \emph{stage}, see, e.g., \cite{Benner.2015}.
Ideally, the \emph{online phase} only requires evaluations of quantities that scale with the dimensionality of the ROM. 

By keeping track of which components of full-order solution $\xvtilde$ are involved in the computation of selected components of the nonlinear term, i.e., $\Zm^T \fv$, the computational cost can be reduced even further. 
In other words, we need not reconstruct all components of the full-order solution $\xvtilde$, but only those components, which are required for the evaluation of $\Zm^T \fv$.
However, the number of components of $\xvtilde$ that is needed for this purpose, which translates into the number of products of rows in $\Phim$ with the reduced-order solution $\xvred$, is typically much larger than the number of sampling indices $p_i$, i.e., the cardinality of the set $\mathcal I$. 

To explain this discrepancy, assume the full-order model to be obtained upon a finite-element discretization. 
Given some particular nodal point, all finite elements sharing the node contribute to the corresponding components of the nonlinear function $\fv$. %and the Jacobian $\Jm_g$.
So we generally must evaluate several element when computing a single component of $\fv$ corresponding to a single sampling index $p_i \in \mathcal I$, which, in turn, involves coordinates of all elements associated with the $p_i$-th degree of freedom.\footnote{
	The ``Unassembled DEIM'' (UDEIM) method proposed in~\cite{Tiso2013} provides a partial remedy for that issue in the context of finite-element problems. 
	In UDEIM, the algorithm is applied to the unassembled residual vector, i.e., the set of element residuals, which restricts the dependency among generalized coordinates to individual elements.
%	{\color{red} NOTE: 2022.08.05 - TODO: mention UDEIM as a partial remedy for that.}
}

For nonlinear manifold methods, we cannot expect much improvement in computational efficiency by the hyper-reduction.
As a matter of fact, the `nonlinearity' becomes twofold if the reduced subspace is a nonlinear manifold:
We do not only have to compute selected components of the nonlinear term $\Zm^T \fv$, we need to evaluate the nonlinear manifold $\gv$. More importantly, from a computational point of view, also relevant rows of the Jacobian of the nonlinear manifold, which are extracted by $\Zm^T \Jm_g (\xvred)$, must be re-computed for every update of the (reduced-order) solution $\xvred$, see Eq.~\eqref{eq:NM-ROM-HR}.

For Petrov-Galerkin-type variants of ROMs, hyper-reduction works in exactly the same way as with their Galerkin counterparts.
The residual in the minimization problem in Eq.~\eqref{eq:NM-LSPG-ROM} is approximated by a gappy reconstruction, i.e.,
\begin{equation}
	\xvred_n = \argmin_{\widetilde \vv \in \realsn} \frac 1 2 \left\Vert (\Zm^T \Phim_r )^\dagger \Zm^T \rvtilde_{BE}^n (\vvtilde; \xvred_{n-1}, \muv) \right\Vert_2^2 .
\end{equation}
From a computational point of view, the same implications apply to Petrov-Galerkin ROMs as for Galerkin-type ROMs, which is why we focus on the latter in our review.

In the approach of 
%\citeauthor{kim.2020a}~\cite{kim.2020a},
%
% CMES style rewriting 
%Kim \emph{et al.}~\cite{kim.2020a},
\cite{kim.2020a},
the nonlinear manifold $\gv$ was represented by a feed-forward neural network, i.e., essentially the decoder of the proposed sparse autoencoder, see Eq.~\eqref{eq:ROM_decoder}.

The computational cost of evaluating the decoder and its Jacobian scales with the number of parameters of the neural network.
Both shallowness and sparsity of the decoder network already account for computational efficiency in regard of the number of parameters. 

Additionally, 
%\citeauthor{kim.2020a}~\cite{kim.2020a}
%
% CMES style rewriting 
%Kim \emph{et al.}~\cite{kim.2020a}
the authors of \cite{kim.2020a}
traced ``active paths'' when evaluating selected components of the decoder and its Jacobian of the hyper-reduced model.
The set of active paths comprises all those connections and neurons of the decoder network which are involved in evaluations of its outputs.
Figure~\ref{fig:Choi-2021-masking-active-path}~(left) highlights the active paths for the computations of the components of the reduced residual $\rvred$, from which the full residual vector $\rvtilde$ is reconstructed within the hyper-reduction method, in orange.

Given all active paths, a subnet of the decoder network is constructed to only evaluate those components of the full-order state which are required to compute the hyper-reduced residual.
The computational costs to compute the residual and its Jacobian depends on the size of the subnet.
As both input and output dimension are given, size translates into width of the (single) hidden layer. 
The size of the hidden layer, in turn, depends on the distribution of the sampling indices $p_i$, i.e., from which components the full residual $\rvred$ is reconstructed. 

For the sparsity patterns assumed, successive output components show the largest overlap in terms of the number of neurons in the hidden layer involved in the evaluation, whereas the overlap is minimal in case of equally spaced outputs.

The cases of successive and equally distributed sampling indices constitute extremal cases, for which the computational time for the evaluation of both the residual and its Jacobian of the 2D-example (Section~\ref{sc:Choi-2d-example})) are illustrated as a function of the dimensionality of the reduced residual (``number of sampling points'') in Figure~\ref{fig:Choi-2021-masking-active-path}~(right). 

%\begin{enumerate}
%	\item Jacobian computation
%%	\item subnet construction
%\end{enumerate}

\begin{figure}[h]
	\centering
	\includegraphics[width=1\linewidth]{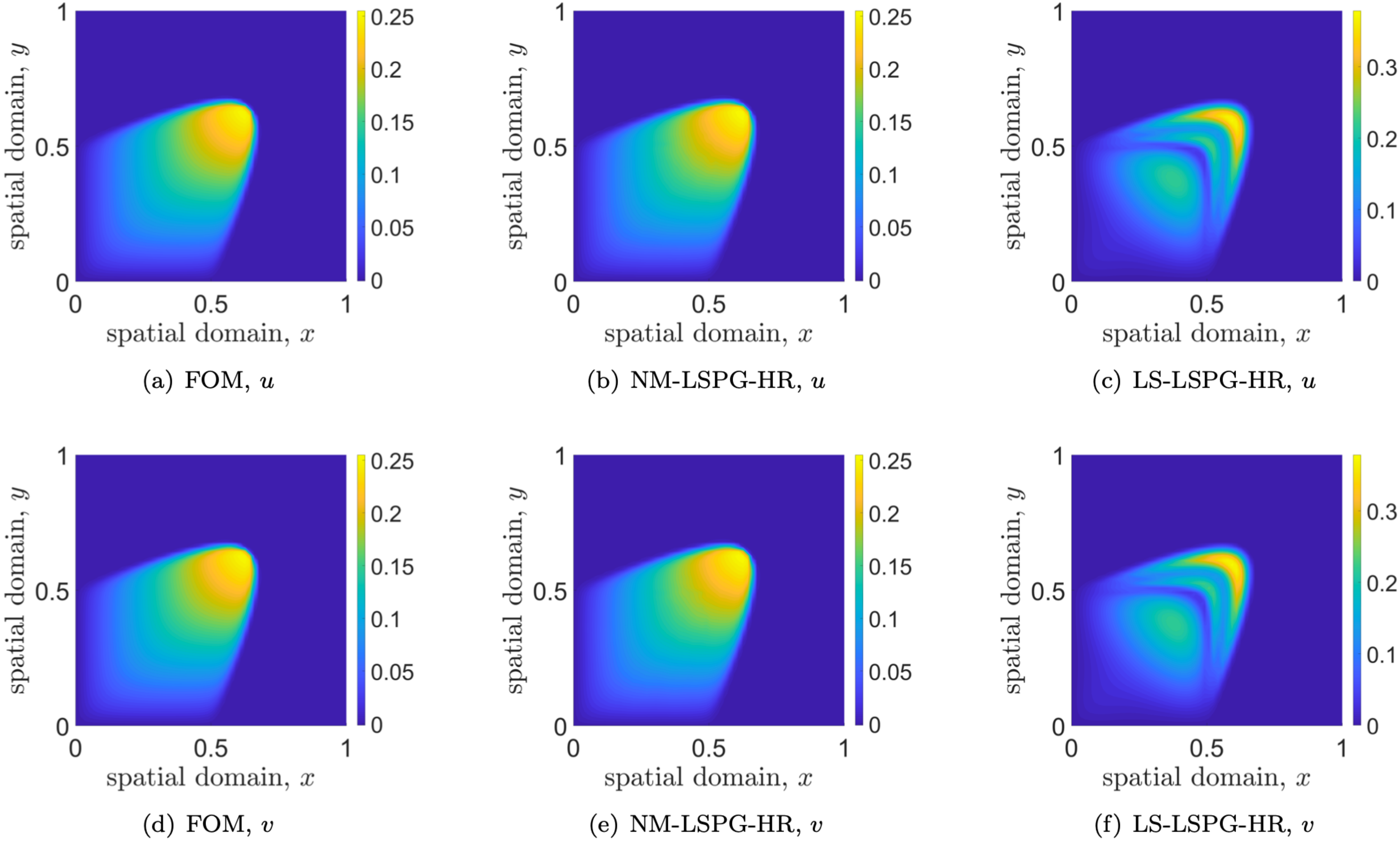}
	\caption{\emph{2-D Burger's equation.  Solution snapshots of full and reduced-order models} (Section~\ref{sc:Choi-2d-example}).
		From left to right, the components $u$ (top row) and $v$ (bottom row) of the velocity field at time $t=2$ are shown for the FOM, the hyper-reduced nonlinear-manifold-based ROM (NM-LSPG-HR)  and the hyper-reduced linear-subspace-based ROM (LS-LSPG-HR).
		Both ROMs have a dimension of $n_s = 5$; with respect to hyper-reduction, the residual basis of the NM-LSPG-HR ROM has a dimension of $n_r = \num{55}$ and $n_z = \num{58}$ components of the full-order residual ("sampling indices") are used in the gappy reconstruction of the reduced residual.
		For the LS-LSPG-HR ROM, both the dimension of the residual basis and the number of sampling indices are $n_r  = n_z= \num{59}$.
		Due to the advection, a steep, shock wave-like gradient develops.
		While FOM and NM-LSPG-HR solutions are visually indistinguishable, the LS-LSPG-HR fails to reproduce the FOM's solution by a large margin (right column).
		Spurious oscillation patterns characteristic of advection-dominated problems (Brooks \& Hughes (1982)~\cite{Brooks.1982}) occur.
		\footnotesize (Figure reproduced with permission of the authors.)
	}
	\label{fig:Choi-2021-results-2d}
\end{figure}

\subsubsection{Numerical example: 2D Burger's equation}
\label{sc:Choi-2d-example}
As a second example, 
%
% CMES style rewriting
%Kim \emph{et al.}~\cite{kim.2020a} extended 
consider now 
Burgers' equation in two (spatial) dimensions \cite{kim.2020a}, instead of in one dimension as in Section~\ref{sc:Choi-1d-example}. 
Additionally, \emph{viscous behavior}, which manifests as the Laplace term on the right-hand side of the following equation, was included as opposed to the one-dimensional problem in Eq.~\eqref{eq:Choi-Burgers-1d}:
\begin{equation}
	\frac{\partial \uv}{\partial t} + \left( \gradient \uv \right) \cdot \uv = \frac{1}{Re} \diverg \left( \gradient \uv \right) , \qquad
	\xv = (x , y)^T \in \Omega = [0, 1] \times [0, 1], \qquad
	t \in [0, 2] .
\end{equation}
The problem was solved on a square (unit) domain $\Omega$, where homogeneous Dirichlet conditions for the velocity field $\uv$ were assumed at all boundaries $\Gamma = \partial \Omega$: 
\begin{equation}
	\uv (\xv, t; \mu) = \mathbf 0  \quad \text{on} \quad  \Gamma = \partial \Omega . %\Gamma = \left\{ \xv \, \vert \, x \in \{0, 1\}, y \in \{ 0, 1 \} \right\} .
\end{equation}
An inhomogeneous flow profile under \SI{45}{\degree} was prescribed as initial conditions ($t=0$),
\begin{equation}
	u(\xv, 0; \mu) = v(\xv, 0; \mu) = \begin{cases}
		\mu \sin \left( 2 \pi x \right)	\sin \left( 2 \pi y \right) 	 &	\text{if} \; \xv \in [0, 0.5] \times [0, 0.5] \\
		0	&	\text{otherwise} 
	\end{cases} ,
\end{equation}
where $u$, $v$ denoted the Cartesian components of the velocity field, and $\mu \in \spaceD = [0.9, 1.1]$ was a parameter describing the magnitude of the initial velocity field.
Viscosity was governed by the Reynolds number $Re$, i.e., the problem was advection-dominated for high Reynolds number, whereas diffusion prevailed for low Reynolds number.

\begin{figure}[h]
	\centering
	\includegraphics[width=1\linewidth]{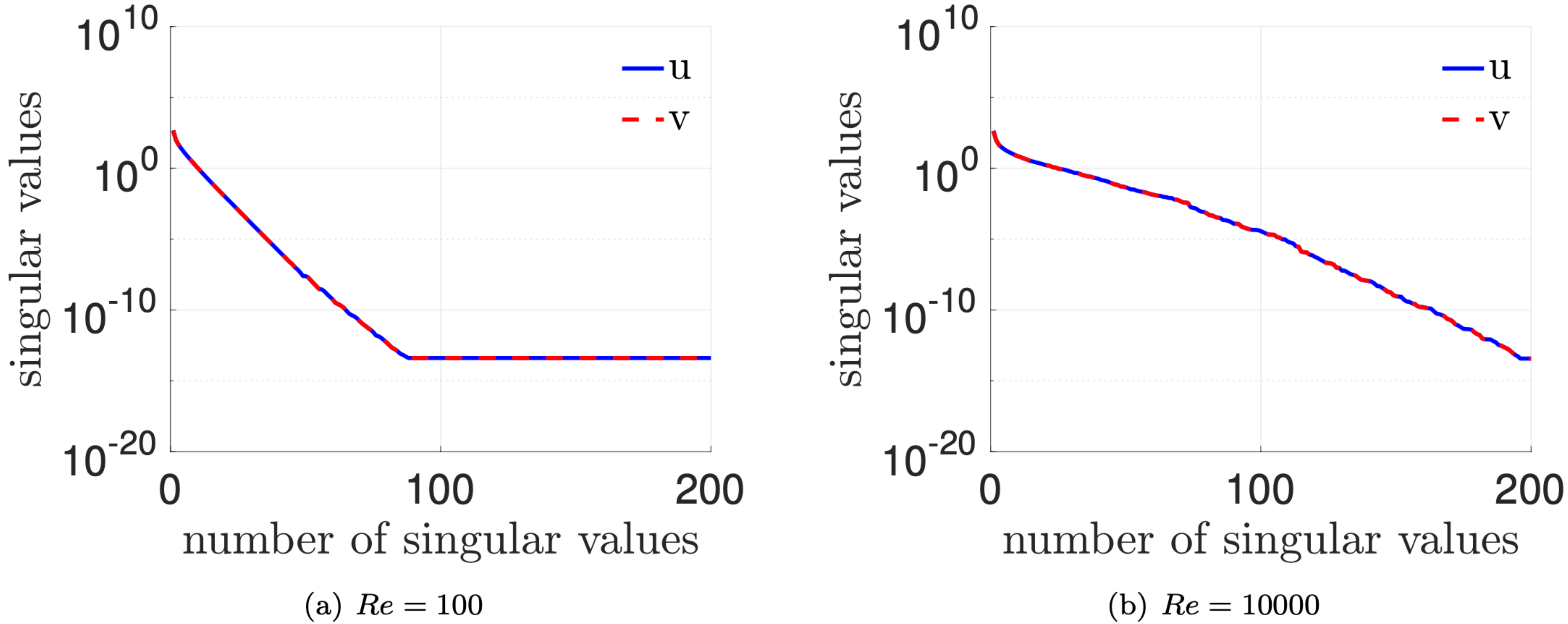}
	\caption{
		\emph{2-D Burger's equation. Reynolds number vs. singular values} (Section~\ref{sc:Choi-2d-example}). Performing SVD on FOM solution snapshots, which were partitioned into $x$ and $y$-components, the influence of the Reynolds number on the singular values is illustrated.
		In diffusion-dominated problems, which are characterized by low Reynolds number, a rapid decay of singular values was observed. Less than \num{100} singular values were non-zero (in terms of double precision accuracy) in the present example for $Re=\num{100}$. 
		In problems with high Reynolds number, in which advection dominates over diffusive processes, the decay of singular values was much slower. As many as \num{200} singular values were different from zero in the case $Re=\num{1e4}$.
		\footnotesize (Figure reproduced with permission of the authors.)		
	}
	\label{fig:Choi-singular-values-UV}
\end{figure}

The semi-discrete FOM was a finite-difference approximation in the spatial dimension on a uniform $60 \times 60$ grid of points $(x_i, y_j)$, where $i \in \left\{ 1, 2, \ldots , 60 \right\}$, $j \in \left\{ 1, 2, \ldots , 60 \right\}$.
First spatial derivatives were approximated by backward differences; central differences were used to approximate second derivatives. 
%Upon 
A
spatial discretization led to 
%we obtain 
a set of ODEs, which was partitioned into two subsets that corresponded to the two spatial directions:
\begin{equation}
	\dt \Uv = \fv_u(\Uv, \Vv) , \qquad
	\dt \Vv = \fv_u(\Uv, \Vv) , \qquad \Uv, \Vv \in \reals^{(n_x-2) \times (n_y - 2)} ,
	\label{eq:Choi-discrete-2d}
\end{equation}
where $\Uv$, $\Vv$ comprised the components of velocity vectors at the grid points in $x$ and $y$-direction, respectively.
For the nonlinear functions $\fv_u, \fv_v :  \reals^{(n_x-2) \times (n_y - 2)} \times \reals^{(n_x-2) \times (n_y - 2)} \to  \reals^{(n_x-2) \times (n_y - 2)}$, which follow from the spatial discretization of the advection and diffusion terms \cite{kim.2020a}.
%readers are referred to the original paper \cite{kim.2020a}.
%
In line with the partitioning the system of equations \eqref{eq:Choi-discrete-2d}, two separate autoencoders were trained for $\Uv$ and $\Vv$, respectively, since less memory was required as opposed to a single autoencoder for the full set of unknowns $(\Uv^T, \Vv^T)^T$.

For time integration, the backward Euler scheme with a constant step size $\Delta t = 2 / n_t$ was applied, where $n_t = \num{1500}$ was used in the example. 
The solutions corresponding to the parameter values $\mu \in \left\{ 0.9, 0.95, 1.05, 1.1 \right\}$ were collected as training data, which amounted to a total of $4 \times (n_t + 1) = \num{6004}$ snapshots.
%\SI{10}{\percent} 
Ten percent (10\%)
of the snapshots were retained as validation set (see Sec.~\ref{sc:training-valication-test}); no test set was used.

Figure \ref{fig:Choi-singular-values-UV} shows the influence of the Reynolds number on the singular values obtained from solution snapshots of the FOM. 
For $Re = \num{100}$, i.e., when diffusion was dominant, the singular values decayed rapidly as compared to an advection-dominated problem with a high Reynolds number of $Re = \num{1e4}$, for which the reduced-order models were constructed in what follows.
In other words, the dimensionality of the tangent space of the FOM's solution was more than twice as large in the advection-dominated case, which limited the feasible reduction in dimensionality by means of linear subspace methods. 
Note that the singular values were the same for both components of the velocity field $u$ and $v$.
The problem was symmetric about the diagonal from the lower-left ($x=0$, $y=0$) to the upper-right ($x=1$, $y=1$) corner of the domain, which was also reflected in the solution snapshots illustrated in Figure~\ref{fig:Choi-2021-results-2d}.
For this reason, we only show the results related to the $x$-component of the velocity field in what follows.

Both autoencoders (for $\Uv$ and $\Vv$, respectively) had the same structure. 
In each encoder, the (single) hidden layer had a width of \num{6728} neurons, which were referred to as {``nodes''} in \cite{kim.2020a}; the  hidden layer of the (sparse) decoder networks was \num{33730} neurons wide.
To train the autoencoders, the Adam algorithm (see Sec. \ref{sc:adam1} and Algorithm \ref{algo:unified-adaptive-learning-rate-2}) was used with an initial learning rate of \num{0.001}. 
The learning rate was decreased by a factor of \num{10} when the training loss did not decrease for \num{10} successive epochs.
The batch size was \num{240}, and training was stopped either after the maximum of \num{10000} epochs or, alternatively, once the validation loss had stopped to decrease for \num{200} epochs. 
The (single) hidden layers of the encoder networks had a width of \SI{6728} neurons; with \SI{33730} neurons, the decoders' hidden layers were almost five times wider.
The parameters of all neural networks were initialized according to the \emph{Kaiming He} initialization~\cite{He.2015b}.

\begin{figure}[h]
	\centering
	\includegraphics[width=0.45\linewidth,trim=0 1cm 500px 0, clip=true]{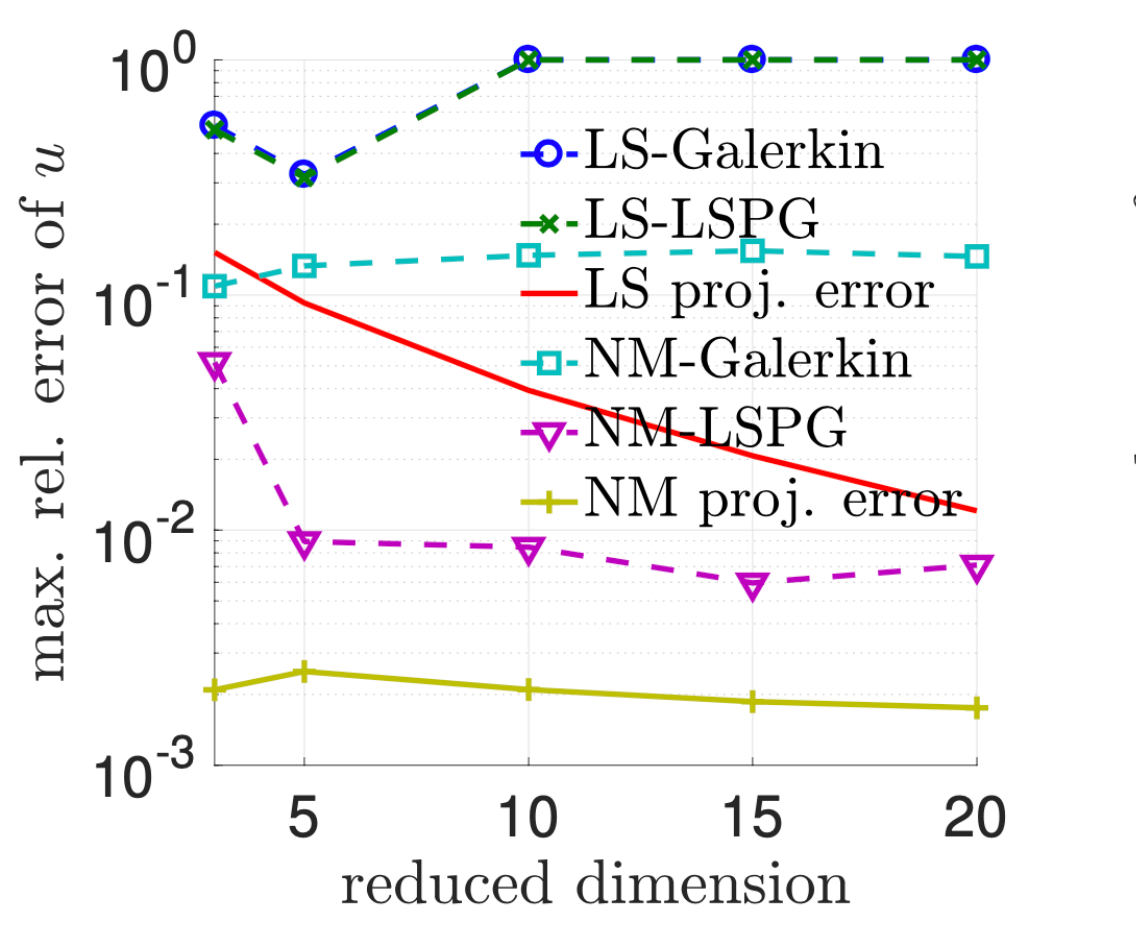}
	\includegraphics[width=0.45\linewidth,trim=0 1cm 500px 0, clip=true]{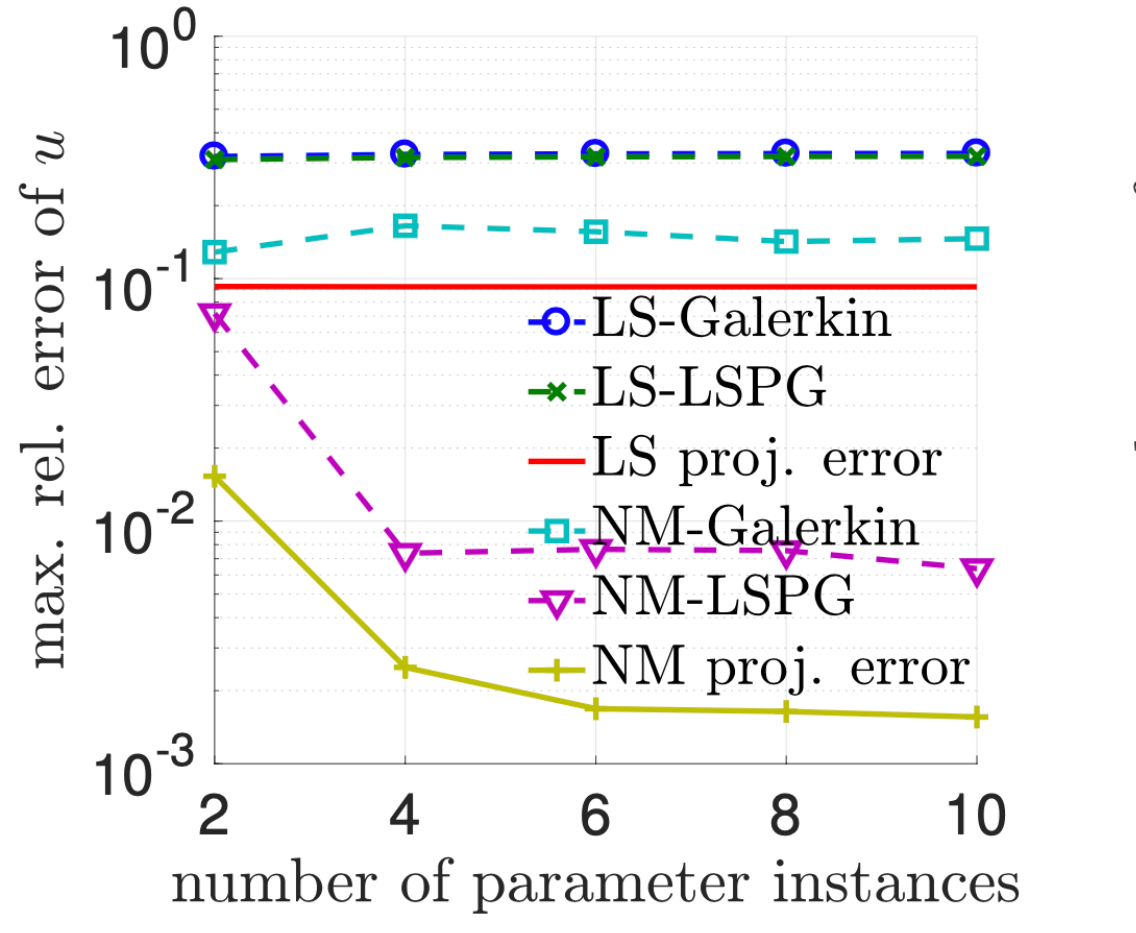}
	\caption{\emph{2D-Burgers' equation: relative errors of nonlinear manifold and linear subspace ROMs} (Section~\ref{sc:Choi-2d-example}).
		\footnotesize (Figure reproduced with permission of the authors.)
	}
	\label{fig:Choi-2021-relerr_vs_dim-2d}
\end{figure}

To evaluate the accuracy of the NM-ROMs proposed  in \cite{kim.2020a}, the Burgers' equation was solved for the target parameter $\mu = 1$, for which no solution snapshots were included in the training data.
Figure~\ref{fig:Choi-2021-relerr_vs_dim-2d} compares the relative errors (Eq.~\eqref{eq:Choi-rel-error}) of the nonlinear-manifold-based and linear-projection-based ROMs as a function of the reduced dimension $n_s$.
Irrespective of whether a Galerkin or Petrov-Galerkin approach was used, nonlinear-manifold based ROMs (NM-Galerkin, NM-LSPG) were superior to their linear-subspace counterparts (LS-Galerkin, LS-LSPG).
The figure also shows the so-called \emph{projections errors}, which are lower bounds for the relative errors of linear-subspace and nonlinear-manifold-based errors, see~\cite{kim.2020a} for their definitions.
Note that the relative error of the NM-LSPG was smaller than the lower error bound of linear-subspace ROMs.
As noted in \cite{kim.2020a}, linear-subspace ROMs performed relatively poorly for the problem at hand, and even failed to converge. 
Both, the LS-Galerkin and the LS-LSPG-ROM showed relative errors of \num{1} if their dimension was \num{10} or more in the present problem.
The NM-Galerkin ROM fell behind the NM-LSPG ROM in terms of accuracy. 
We also note that both NM-based ROMs hardly showed any reduction in error if their dimension was increased beyond five.

\begin{table}[h]
	\centering
	\includegraphics[width=1\linewidth]{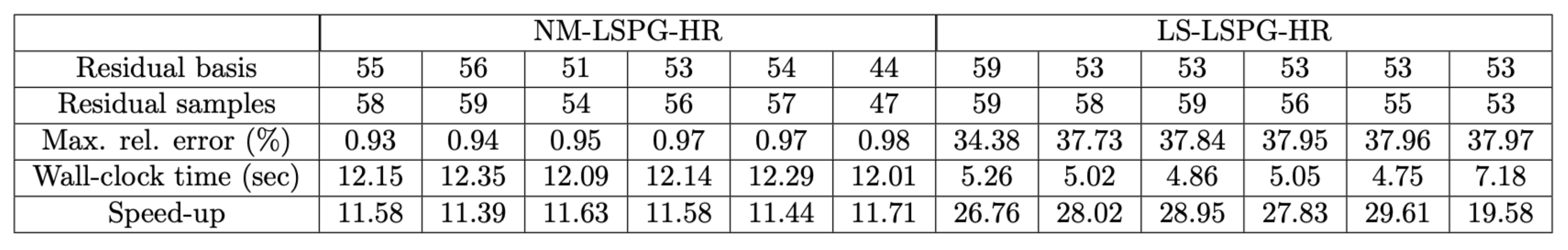}
	\caption{\emph{2-D Burger's equation. Juxtaposition of hyper-reduced ROMs: speed-up and accuracy} (Section~\ref{sc:Choi-2d-example}).
		The 6 respective best least-squares Petrov-Galerkin ROMs built upon nonlinear manifold approximation by means of autoencoders and linear subspaces are compared, where `best' refers to the maximum error relative to the FOM (Eq.~\eqref{eq:Choi-rel-error}).
		The optimal dimension of the basis $\Phim_r$ and the number of sampling indices $p_j$ used in the gappy reconstruction of the nonlinear residual lie in similar ranges for all ROMs listed. 
		The hyper-reduced ROMs achieve speed-up in wall-clock time of factors of approximately \num{11} for the nonlinear-manifold-based approach up to a factor of almost \num{30} in linear-subspace ROMs. 
		While the former show a maximum relative error of below \SI{1}{\percent}, the latter fail to reproduce the FOM's behavior by a large margin.
		\footnotesize (Table reproduced with permission of the authors.)
	}
	\label{tb:Choi-2021-tab-speedup-2d}
\end{table}

The authors of \cite{kim.2020a} also studied the impact how the size of the parameter set $\spaceD_{\rm train}$, which translated into the amount of training data, affected the accuracy of ROMs.
For this purpose, parameter sets with $n_{\rm train} = 2, 4, 6, 8$ values of $\mu \in \spaceD$, which were referred to as ``parameter instances,'' were created, where the target value $\mu = 1$ remained excluded, i.e., 
\begin{equation}
	\spaceD_{\rm train} = \left\{ 0.9 + 0.2 i / n_{\rm train} , i = 0 , \ldots, n_{\rm train}  \right\} \backslash \{1\} ;
\end{equation}
the reduced dimension was set to $n_s = 5$.
Figure~\ref{fig:Choi-2021-relerr_vs_dim-2d} (right) reveals that, for the NM-LSPG ROM, \num{4} ``parameter instances'' were sufficient to reduce the maximum relative error below \SI{1}{\percent} in the present problem.
None of the ROMs benefited from increasing the parameter set, for which the training data were generated.

Hyper-reduction turned out to be crucial with respect to computational efficiency. For a reduced dimension of $n_s = 5$, all ROMs except for the NM-LSPG ROM, which achieved a minor speedup, were less efficient than the FOM in terms of wall-clock time. 
The dimension of the residual basis $n_r$ and the number of sampling indices $n_z$ were both varied in the range from \num{40} to \num{60} to quantify their relation to the maximum relative error.  
For this purpose, the number of training instances was again set to $n_{\rm train} = 4$ and the reduced dimension was fixed to $n_s = 5$.
Table~\ref{tb:Choi-2021-tab-speedup-2d} compares the 6 best---in terms of maximum error relative to the FOM---hyper-reduced least-squares Petrov-Galerkin ROMs based on nonlinear manifolds and linear subspaces, respectively.
The NM-LSPG-HR ROM in \cite{kim.2020a} was able to achieve a speed-up of more than a factor of $\num{11}$ while keeping the maximum relative error below \SI{1}{\percent}. 
Though the speed-up of the linear-subspace counterpart was more than twice as large, relative errors beyond \SI{34}{\percent} rendered these ROMs worthless. 

\begin{figure}[H]
	\centering
	\begin{subfigure}[b]{0.70\textwidth}
		\includegraphics[width=0.95\linewidth]{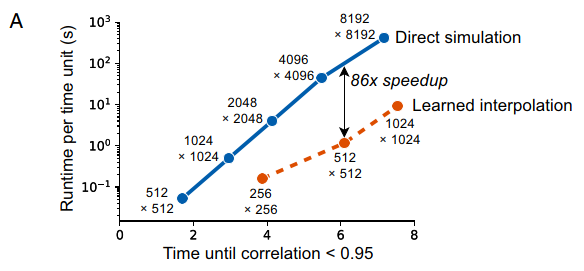}
	\end{subfigure}
	\caption{
		\emph{Machine-learning accelerated CFD} (Section~\ref{sc:Choi-2d-example}).  Speed-up factor, compared to direct integration, was much higher than those obtained from nonlinear model-order reduction in Table~\ref{tb:Choi-2021-tab-speedup-2d}  
		\cite{kochkov2021machine}.
		\footnotesize 
		%		(Figure reproduced with permission of the authors.)
		\href{https://www.pnas.org/page/about/rights-permissions}{Permission of NAS}.
	}
	\label{fig:ConvNet-NS-simulation-1}
\end{figure}

%In the \emph{online phase}, the hyper-reduced linear subspace ROM  
%For the hyper-reduced linear subspace ROM, the computational cost in the \emph{online phase} scales with the dimensionality $n_s$ of the ROM, which is supposed to be much smaller than the dimensionality $N_s$ of the full-order model.

%To compute, selected components of the non-linear term, however, the full-order solution $\xvref + \Phim \xvred$ needs to be computed

%		{\color{red} [NOTE: 2022.10.04 - We need to write a short section or remark on this hybrid method that uses ML to accelerate the integration of NS equations in \cite{kochkov2021machine}.  ENDNOTE]}

\begin{figure}[H]
	\centering
	\begin{subfigure}[b]{0.80\textwidth}
		\includegraphics[width=0.95\linewidth]{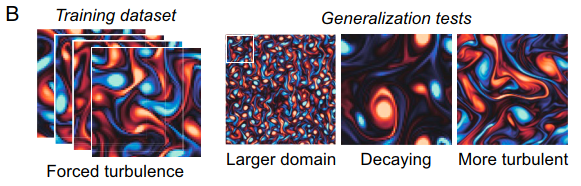}
	\end{subfigure}
	\caption{
		\emph{Machine-learning accelerated CFD} (Section~\ref{sc:Choi-2d-example}).  Good accuracy and good generalization, devoiding of non-physical solutions
		\cite{kochkov2021machine}.
		\footnotesize 
		%		(Figure reproduced with permission of the authors.)
		\href{https://www.pnas.org/page/about/rights-permissions}{Permission of NAS}.
	}
	\label{fig:ConvNet-NS-simulation-2}
\end{figure}

\begin{figure}[H]
	\centering
	\begin{subfigure}[b]{0.95\textwidth}
		\includegraphics[width=0.95\linewidth]{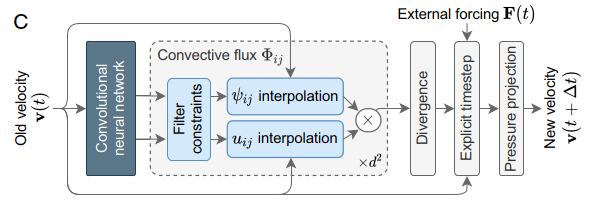}
	\end{subfigure}
	\caption{
		\emph{Machine-learning accelerated CFD} (Section~\ref{sc:Choi-2d-example}).  
%		Block diagram
		The neural network generates interpolation coefficients based on local-flow properties, while ensuring at least first-order accuracy relative to the grid spacing
		\cite{kochkov2021machine}.
		\footnotesize 
%		(Figure reproduced with permission of the authors.)
		\href{https://www.pnas.org/page/about/rights-permissions}{Permission of NAS}.
	}
	\label{fig:ConvNet-NS-simulation-3}
\end{figure}

\begin{rem}
	\label{rm:machine-learning-accelerated-CFD}
	Machine-learning accelerated CFD.
	{\rm  
		A hybrid method between traditional direct integration of the Navier-Stokes equation and machine learning (ML) interpolation was presented in \cite{kochkov2021machine} (Figure~\ref{fig:ConvNet-NS-simulation-3}), where a speed-up factor close to 90, many times higher than those in Table~\ref{tb:Choi-2021-tab-speedup-2d}, was obtained, Figure~\ref{fig:ConvNet-NS-simulation-1}, while generalizing well (Figure~\ref{fig:ConvNet-NS-simulation-2}). Grounded on the traditional direct integration, such hybrid method would avoid non-physical solutions of pure machine-learning methods, such as the physics-inspired machine learning (Section~\ref{sc:PINN-frameworks}, Remark~\ref{rm:PINN-convergence-problem}), maintain higher accuracy as obtained with direct integration, and at the same time benefit from an acceleration from the learned interpolation.  
	}
	$\hfill\blacksquare$
\end{rem}

\begin{rem}
	{\rm In concluding this section, we mention the 2023 review paper
		\cite{bishara2023state}, brought to our attention by a reviewer, on ``A state-of-the-art review on machine learning-based multiscale modeling, simulation, homogenization and design of materials.''  This review paper would nicely complement our present review paper.}
	$\hfill\blacksquare$
\end{rem}

% 2019.12.07, commented out this section 16-other-apps
% all other applications of deep learning
% structural health monitoring
% quantum mechanics
%\input{16-other-apps}
%
% historical perspective

\section{Historical perspective}
\label{sc:history}

% 2019.12.07
% Mark: 20-history.tex, subsection up and down of AI, cybernetics, etc., moved to NOTES

\subsection{Early inspiration from biological neurons}
\label{sc:inspired-from-biology}
\begin{figure}[h]
	\centering
  %
  % 2022.12.17
  % remove ".eps" for arXiv
	% \includegraphics[width=0.7\linewidth]{figures/neuron3.eps}
	\includegraphics[width=0.7\linewidth]{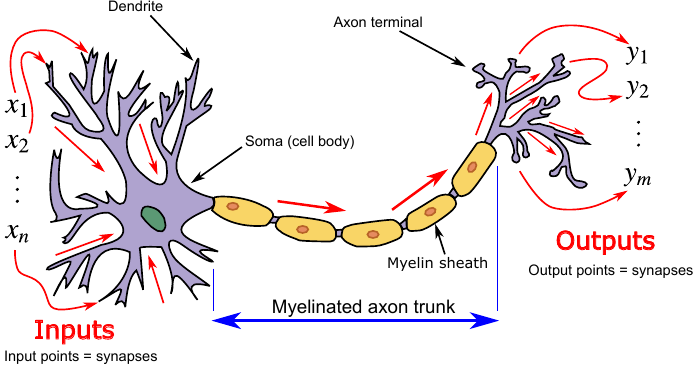}
	\caption{
	 	\emph{Biological Neuron and signal flow} (Sections~\ref{sc:artificial-neuron}, \ref{sc:inspired-from-biology}, \ref{sc:dynamic-volterra-series}) along myelinated axon, with inputs at the synapses (input points) in the dendrites and with outputs at the axon terminals (output points,which are also the synapses for the next neuron).
	 	Each input current $\x_i$ is multiplied by the weight $\weight_{i}$, then all weighted input currents are summed together (linear combination), with $i = 1, \ldots, n$, to form the total synaptic input current $I_s$ into the soma (cell body).  	
	 	The corresponding artificial neuron is in Figure~\ref{fig:neuron5} in Section~\ref{sc:artificial-neuron}.
	 	\footnotesize
	 	(Figure adapted from  
	 	Wikipedia
	 	\href{https://commons.wikimedia.org/w/index.php?title=File:Neuron3.svg&oldid=348383690}{version 14:29, 2 May 2019}).
	}
	%\label{fig:neuron3}
	\label{fig:bio-neuron}
\end{figure}

In the early days, many papers on artificial neural networks, particularly for applications in engineering, started to motivate readers with a figure of a biological neuron as in Figure~\ref{fig:bio-neuron} (see, e.g., \cite{Ghaboussi.1991:rd0001}, Figure~1a), before displaying an artificial neuron (e.g, \cite{Ghaboussi.1991:rd0001}, Figure~1b).  
When artificial neural networks took a foothold in the research community, there was no need to motivate with a biological neuron, e.g., \cite{Oishi.2017:rd9648} \cite{Sze.2017:rd0001}, which began directly with an artificial neuron.

% figure file deleted to have less than 60 files; this figure is not needed.
%\begin{figure}
%  \label{fig:neuronAnatomy}
%  \centering
%  \includegraphics[width=0.7\linewidth]{figures/neuron_anatomy.jpg}
  
%  {\footnotesize From 
%  \href{https://askabiologist.asu.edu/neuron-anatomy}{Neuron Anatomy}. 
%  \href{https://askabiologist.asu.edu/contact/permissions}{@ Arizona Board of Regents / ASU Ask A Biologist}.}
%  \caption{Neuron signal flow.  
%  }
%\end{figure}

% \subsection{Linear combination of inputs, weights, biases}
\subsection{Spatial / temporal combination of inputs, weights, biases}
\label{sc:linear-combo-history}

\begin{figure}[h]
	\centering
	%
	% 2022.12.17
	% add "-eps-converted-to.pdf" for arXiv
	% \includegraphics[width=0.3\linewidth]{figures/perceptron}
	\includegraphics[width=0.3\linewidth]{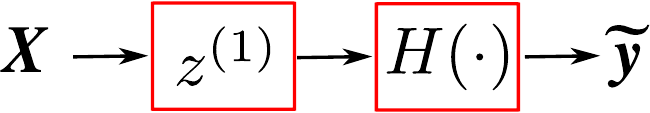}
	\caption{  
		The \emph{perceptron network} (Sections~\ref{sc:XORfunction}, \ref{sc:linear-combo-history})---introduced by 
		%
		% CMES style, rewriting
		Rosenblatt (1958)
		\cite{Rosenblatt.1958}, (1962) \cite{Block1962a}---has a linear combination with weights and bias as expressed in $z^{(1)} (\bx_{i}) = \boldsymbol{w} \bx_{i} + b \in \real$, but differs from the one-layer network in Figure~\ref{fig:XOR-one-layer} in that it is 
		activated by the Heaviside function.
		%
		% CMES style, rewriting
%		\cite{Minsky.1969} showed 
		That the Rosenblatt perceptron cannot represent the XOR function; see Section~\ref{sc:XORfunction}.		
	}
	\label{fig:perceptron}
\end{figure}

\begin{figure}[h]
	\centering
	\includegraphics[width=0.5\linewidth]{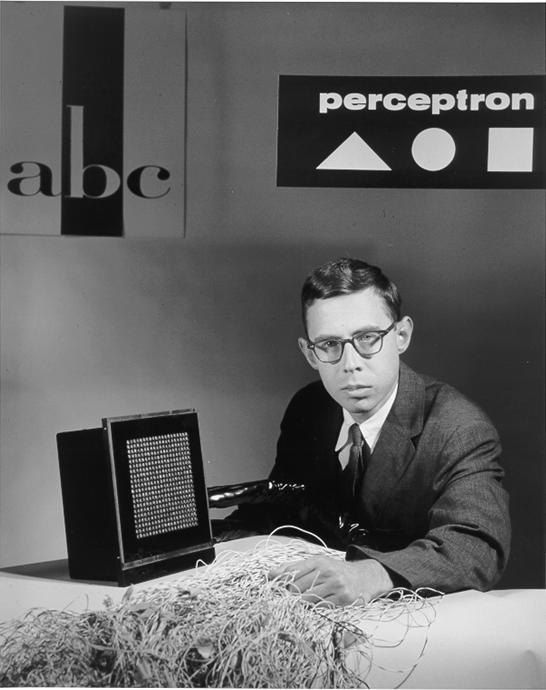}
	\includegraphics[width=0.45\linewidth]{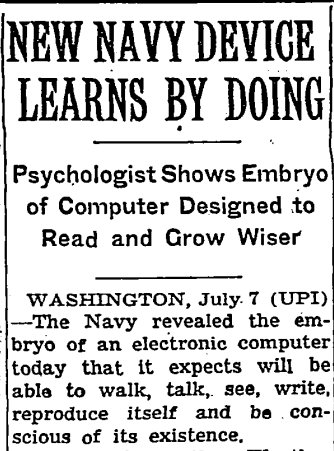}
	\caption{  
		\emph{Rosenblatt and the Mark I computer} (Sections~\ref{sc:depth}, \ref{sc:linear-combo-history}) based on the perceptron, described in the New York Times article titled ``New Navy device learns by doing'' on 1958 July 8 (\href{https://web.archive.org/web/20190912024703/http://jcblackmon.com/wp-content/uploads/2018/01/MBC-Rosenblatt-Perceptron-NYT-article.jpg.pdf}{Internet archive}), as a ``computer designed to read and grow wiser'', and would be able to ``walk, talk, see, write, reproduce itself and be conscious of its existence.  The first perceptron will have about 1,000 electronic ``association cells'' [A-units] receiving electrical impulses from an eye-like scanning device with 400 photo cells''.  
		See also the \href{https://www.youtube.com/watch?v=cNxadbrN_aI}{Youtube video} 	
		``Perceptron Research from the 50's \& 60's, clip''.
		Sometimes, it is incorrectly thought that Rosenblatt's network had only one neuron (A-unit); see Figure~\ref{fig:network-size-time}, Section~\ref{sc:depth}.
	}
	\label{fig:Rosenblatt-Mark-I}
\end{figure}

% recall that Eq.~(\ref{eq:linearComboInputsBias}) represents a linear combination of inputs, with weights and biases.
Both \cite{Nielsen.2015} and \cite{Goodfellow.2016} referred to Rosenblatt (1958) \cite{Rosenblatt.1958}, who first proposed using a linear combination of inputs with weights, and with biases (thresholds).
The authors of 
\cite{Goodfellow.2016}, p.~14, only mentioned the ``linear model'' defined---using the notation convention in Eq.~(\ref{eq:z=wy})---as
\begin{align}
	f(\bx , \boldsymbol{\weight}) = \sum_{i}^n x_i w_i = \boldsymbol{w} \bx
	\label{eq:linear-combo}
\end{align}
without the bias.
%
% CMES style rewriting  
%\cite{Nielsen.2015} on the other hand wrote
On the other hand, it was written in \cite{Nielsen.2015} that ``Rosenblatt proposed a simple rule to compute the output. He introduced weights, real numbers expressing the importance of the respective inputs to the output'' and ``some threshold value,'' and attributed the following equation to Rosenblatt
\begin{align}
	\text{output} = 
	\begin{cases}
		0 \text{ if } \sum_j w_j x_j \le \text{threshold} 
		\\
		1 \text{ if } \sum_j w_j x_j > \text{threshold}
	\end{cases}
	\label{eq:perceptron}
\end{align}
where the threshold is simply the negative of the bias $\biassp{i}{\ell}$ in Eq.~(\ref{eq:linearComboInputsBias})\footnote{
	% the bias $\biassp{i}{\ell}$ in Eq.~(\ref{eq:linearComboInputsBias}) is the negative of a ``threshold $\theta$'' defined in \cite{Rosenblatt.1962}, p.~81; see \cite{Nielsen.2015}.
	%
	% CMES style rewriting
%	\cite{Hardesty.2017:rd0001} wrote: 
	``The Perceptron's design was much like that of the modern neural net, except that it had only one layer with adjustable weights and thresholds, sandwiched between input and output layers'' \cite{Hardesty.2017:rd0001}.
	In the neuroscientific terminology that Rosenblatt (1958) \cite{Rosenblatt.1958} used, the input layer contains the sensory units, the middle (hidden) layer contains the ``association units,'' and the output layer contains the response units.   
	% see also \cite{Rosenblatt.1962}. 
	Due to the difference in notation and due to ``neurodynamics'' as a new field for most readers, we provide here some markers that could help track down where Rosenblatt used linear combination of the inputs.
	Rosenblatt (1962)
	\cite{Rosenblatt.1962}, p.~82, defined the ``transmission function $c^\star_{ij}$'' for the connection between two ``units'' (neurons) $u_i$ and $u_j$, with $c^\star_{ij}$ playing the same role as that of the term $\weight_{ij}^{(\ell)} \ysp{j}{\ell-1}$ in $z_i^{(\ell)}$ in Eq.~(\ref{eq:linearComboInputsBias}).
	Then for an ``elementary perceptron'',  
	%
	% CMES style rewriting
%	\cite{Rosenblatt.1962}, p.~85, 
	the transmission function $c^\star_{ij}$ was defined in \cite{Rosenblatt.1962}, p.~85, to be equal to the output of unit $u_i$ (equivalent to $\ysp{i}{\ell-1}$) multiplied by the ``coupling coefficient'' $v_{ij}$ (between unit $u_i$ and unit $u_j$), with $v_{ij}$ being the equivalent of the weight $\weight_{ij}^{(\ell)}$ in Eq.~(\ref{eq:linearComboInputsBias}), ignoring the time dependence. 
	The word ``weight,'' meaning coefficient, was not used often in \cite{Rosenblatt.1962}, and not at all in \cite{Rosenblatt.1958}.
} 
or $(-b)$ in Figure~\ref{fig:perceptron}, which is a graphical representation of Eq.~(\ref{eq:perceptron}).
%
% CMES style rewriting
The author of
\cite{Nielsen.2015} went on to say, ``That's all there is to how a perceptron works!''  
Such statement could be highly misleading for first-time learners in discounting Rosenblatt's important contributions, which were extensively inspired from neuroscience, and were not limited to the perceptron as a machine-learning algorithm, but also to the development of the Mark I computer, a hardware implementation of the perceptron; see Figure~\ref{fig:Rosenblatt-Mark-I} \cite{Rosenblatt.1960} \cite{Block1962a} \cite{Block1962b}.

Moreover, adding to the confusion for first-time learners, another error and misleading statement about the ``Rosenblatt perceptron'' in connection with Eq.~(\ref{eq:linear-combo}) and Eq.~(\ref{eq:perceptron})---which represent a single neuron---is in \cite{Goodfellow.2016}, p.~13, where it was stated that the ``Rosenblatt perceptron'' involved only a ``single neuron'':
\begin{quote}
	``The first wave started with cybernetics in the 1940s-1960s, with the development of theories of biological learning (McCulloch and Pitts, 1943; Hebb, 1949) and implementations of the first models, such as the perceptron (Rosenblatt, 1958), enabling the training of a single neuron.'' \cite{Goodfellow.2016}, p.~13.
	% pdf p.~36
\end{quote}

The error of considering the Rosenblatt perceptron as having a ``single neuron'' is also reported in Figure~\ref{fig:network-size-time}, which is Figure~1.11 in \cite{Goodfellow.2016}, p.~23.
But the Rosenblatt perceptron as described in the cited reference Rosenblatt (1958) \cite{Rosenblatt.1958} and in Rosenblatt (1960) \cite{Rosenblatt.1960} was a network, called a ``nerve net'':
\begin{quote}
	``Any perceptron, or nerve net, consists of a network of ``cells,'' or signal generating units, and connections between them.''
\end{quote}
Such a ``nerve net'' would surely not just contain a ``single neuron''.  Indeed, the report by Rosenblatt (1957) \cite{Rosenblatt.1957} that appeared a year earlier mentioned a network (with one layer) containing as many a thousand neurons, called ``association cells'' (or A-units):\footnote{
	We were so thoroughly misled in thinking that the ``Rosenblatt perceptron'' was a single neuron that we were surprised to learn that Rosenblatt had built the Mark I computer with many neurons. 
} 
\begin{quote}
	``Thus with 1000 A-units connected to each R-unit [response unit or output], and a system in which 1\% of the A-units respond to stimuli of a given size (i.e., $P_a = .01$), the probability of making a correct discrimination with one unit of training, after $10^6$ stimuli have been associated to each response in the system, is equal to the 2.23 sigma level, or a probability of 0.987 of being correct.'' \cite{Rosenblatt.1957}, p.~16.
\end{quote}
The perceptron with one thousand A-units mentioned in \cite{Rosenblatt.1957} was also reported in the New York Times article ``New Navy device learns by doing'' on 1958 July 8 (\href{https://web.archive.org/web/20190912024703/http://jcblackmon.com/wp-content/uploads/2018/01/MBC-Rosenblatt-Perceptron-NYT-article.jpg.pdf}{Internet archive}); see Figure~\ref{fig:Rosenblatt-Mark-I}. 
Even if the report by Rosenblatt (1957) \cite{Rosenblatt.1957} were not immediately accessible, it was stated in no uncertain terms that the perceptron was a machine with many neurons:
\begin{quote}
	``The organization of a typical  photo-perceptron  (a  perceptron  responding
	To optical patterns as stimuli) is shown
	In  Figure1. ... [Rule] 1. Stimuli  impinge  on  a  retina  of
	Sensory  units  (S-points),  which  are
	Assumed  to  respond  on  an  all-or-nothing basis\footnote{
		Heaviside activation function, see Figure~\ref{fig:perceptron} for the case of one neuron.
	}. 
	[Rule] 2. Impulses are transmitted  to a set
	Of  association  cells  (A-units) [neurons]... If  the  algebraic  sum of excitatory  and
	Inhibitory impulse intensities\footnote{
		Weighted sum / voting (or linear combination) of inputs; see Eq.~(\ref{eq:linear-combo}), Eq.~(\ref{eq:general-perceptron}), Eq.~(\ref{eq:Block1962-linear-combo-2}).
	} is equal
	To  or greater than the  threshold\footnote{
		The negative of the bias $\biassp{i}{\ell}$ in Eq.~(\ref{eq:linearComboInputsBias}).
	} ($\theta$) of
	The A-unit, then the A-unit fires, again
	On an all-or-nothing basis.''
	\cite{Rosenblatt.1958}
\end{quote} 
Figure~1 in \cite{Rosenblatt.1958} described a network (``nerve net'') with many A-units (neurons).   
Does anyone still read the classics anymore?

Rosenblatt's (1962) book \cite{Rosenblatt.1962}, p.~33, provided the following neuroscientific explanation for using a linear combination (weight sum / voting) of the inputs in both time and space:
\begin{quote}
	``The arrival of a single (excitatory) impulse gives rise to a
	partial depolarization of the post-synaptic\footnote{
		Refer to Figure~\ref{fig:bio-neuron}.  A {\em synapse} (meaning ``junction'') is ``a structure that permits a neuron (or nerve cell) to pass an electrical or chemical signal to another neuron'', and consists of three parts: the {\em presynaptic} part (which is an axon terminal of an upstream neuron from which the signal came), the gap called the synaptic cleft, and the {\em postsynaptic} part, located on a dendrite or on the neuron cell body (called the {\em soma}); \cite{Dayan.2001}, p.~6;
		``Synapse'', Wikipedia, 
		\href{https://en.wikipedia.org/w/index.php?title=Synapse&oldid=885986208}{version 16:33, 3 March 2019}.
		A {\em dendrite} is a conduit for transmitting the electrochemical signal received from another neuron, and passing through a synapse located on that dendrite; ``Dendrite'', Wikipedia, 
		\href{https://en.wikipedia.org/w/index.php?title=Dendrite&oldid=892648749}{version 23:39, 15 April 2019}. 
		A synapse is thus an input point to a neuron in a biological neural network. 
		An axon, or nerve fiber, is a ``long, slender projection of a neuron, that conducts electrical impulses known as action potentials'' away from the soma to the axon terminals, which are the presynaptic parts; ``Axon terminal'', Wikipedia, 
		\href{https://en.wikipedia.org/w/index.php?title=Axon_terminal&oldid=885382347}{version 18:13, 27 February 2019}. 
	} membrane surface, which spreads over an appreciable area, and decays exponentially with time. This is called a local excitatory state (l.e.s.). The l.e.s. due to successive impulses is (approximately) additive. Several impulses arriving in sufficiently close succession may thus combine to touch off
	an impulse in the receiving neuron if the local excitatory state at the base of the axon achieves the threshold level. This phenomenon is called \underline{temporal summation}. Similarly, impulses which arrive at different points on the cell body or on the dendrites may combine by \underline{spatial summation} to
	trigger an impulse if the l.e.s. induced at the base of the axon is strong enough.''
\end{quote}

The {\em spatial summation} of the input synaptic currents is also consistent with Kirchhoff's current law of summing the electrical currents at a junction in an electrical network.\footnote{
	See ``Kirchhoff's circuit laws'', Wikipedia, 
	\href{https://en.wikipedia.org/w/index.php?title=Kirchhoff\%27s_circuit_laws&oldid=895023220}{version 14:24, 1 May 2019}.
}
We first look at linear combination in the static case, followed by the dynamic case with Volterra series.

\subsubsection{Static, comparing modern to classic literature}
\label{sc:static-Rosenblatt}
\begin{quote}
	{\it ``A classic is something that everybody wants to have read and nobody wants to read.''}  
	\\ 
	\phantom{a} \hfill Mark Twain
\end{quote}

Readers not interested in reading the classics can skip this section.
Here, we will not review the perceptron algorithm,\footnote{
	See, e.g., ``Perceptron'', Wikipedia, 
	\href{https://en.wikipedia.org/w/index.php?title=Perceptron&oldid=896433053}{version 13:11, 10 May 2019}, and many other references.
} but focus our attention on the historical details not found in many modern references, and connect Eq.~(\ref{eq:linear-combo}) and Eq.~(\ref{eq:perceptron}) to the original paper by Rosenblatt (1958) \cite{Rosenblatt.1958}.
But the problem is that such task is not directly obvious for readers of modern literature, such as 
\cite{Goodfellow.2016} for the following reasons:
\begin{itemize}
	\item
	Rosenblatt's work in \cite{Rosenblatt.1958} was based on neuroscience, which is confusing to those without this background;
	
	\item
	Unfamiliar notations and concepts for readers coming from deep-learning literature, such as \cite{Nielsen.2015}, \cite{Goodfellow.2016};
	
	\item 
	The word ``weight'' was not used at all in \cite{Rosenblatt.1958}, and thus cannot be used to indirectly search for hints of equations similar to Eq.~(\ref{eq:linear-combo}) or Eq.~(\ref{eq:perceptron});
	
	\item 
	The word ``threshold'' was used several times, such as in the sentence ``If the algebraic sum of excitatory and inhibitory impulse intensities is equal to or greater than the threshold ($\theta$)''\footnote{
		Of course, the notation $\theta$ (lightface) here does not designate the set of parameters, denoted by $\bparam$ (boldface) in Eq.~(\ref{eq:theta}). 
	} of the A-unit, then the A-unit fires, again on an all-or-nothing basis''. The threshold $\theta$ is used in Eq.~(2) of \cite{Rosenblatt.1958}: 
	\begin{align}
		e - i  
		- l_e + l_i + g_e - g_i
		\ge 
		\theta
		\ ,
		\label{eq:rosenblatt-linear-combo-bias}
	\end{align}
	where $e$ is the number excitatory stimulus components received by the A-unit (neuron, or associated unit),
	$i$ the number of inhibitory stimulus components,
	$l_e$ the number of lost excitatory components,
	$l_i$ the number of lost inhibitory components,
	$g_e$ the number of gained excitatory components,
	$g_i$ the number of gained inhibitory components. 
	But all these quantities are positive integers, and thus would not be the real-number weights $w_j$ in Eq.~(\ref{eq:perceptron}), of which the inputs $x_j$ also have no clear equivalence in Eq.~(\ref{eq:rosenblatt-linear-combo-bias}). 
\end{itemize}

As will be shown below, it was misleading to refer to \cite{Rosenblatt.1958} for equations such as Eq.~(\ref{eq:linear-combo}) and Eq.~(\ref{eq:perceptron}), even though \cite{Rosenblatt.1958} contained the seed ideas leading to these equations upon refinement as presented in \cite{Block1962a}, which was in turn based on the book by Rosenblatt (1962) \cite{Rosenblatt.1962}.

Instead of a direct reading of \cite{Rosenblatt.1958}, we suggest reading key publications in reverse chronological orders.  We also use the original notations to help readers to identify quickly the relevant equations in the classic literature.

%
% CMES style rewriting
The authors of
\cite{Minsky.1969} introduced a general class of machines, each known under different names, but decided to call all these machines as ``perceptrons'' in honor of the pioneering work of Rosenblatt.
%
% CMES style rewriting
%\cite{Minsky.1969}, p.~10, define 
General perceptrons were defined in \cite{Minsky.1969}, p.~10, as follows.  
Let $\varphi_i$ be the ith image characteristic, called an image predicate, which consists of a verb and an object, such as ``is a circle'', ``is a convex figure'', etc.  
An image predicate is also known as an image feature.\footnote{
	\label{fn:features}
	See \cite{Goodfellow.2016}, p.~3.  Another example of a feature is a piece of information about a patient for medical diagnostics.  ``For many tasks, it is difficult to know which features should be extracted.''  
	For example, to detect cars, we can try to detect the wheels, but ``it is difficult to describe exactly what a wheel looks like in terms of pixel values'', due to shadows, glares, objects obscuring parts of a wheel, etc. 
}
For example, let the ith image characteristic is whether an image ``is a circle'', then $\varphi_i \equiv \varphi_{circle}$.  If an image $X$ is a circle, then $\varphi_{circle} (X) = 1$; if not, $\varphi_{circle} (X) = 0$:
\begin{align}
	\varphi_{circle} (X) = 
	\begin{cases}
	1 \text{ if } X \text{ is a circle}
	\\
	0 \text{ if } X \text{ is not a circle}
	\end{cases}
\end{align}
Let $\Phi$ be a family of simple image predicates: 
\begin{align}
	\Phi = [ \varphi_1 , \ldots , \varphi_n]
\end{align}
A {\em general} perceptron was defined as a more complex predicate, denoted by $\psi$, which was a {\em weighted voting} or {\em linear combination} of the simple predicates in $\Phi$ such that\footnote{
	\cite{Minsky.1969}, p.~10.
}
\begin{align}
	\psi(X) = 1
	% \text{ if and only if }
	\Longleftrightarrow
	% w_1 \varphi_1 + \ldots + w_n \varphi_n > \theta
	\alpha_1 \varphi_1 + \ldots + \alpha_n \varphi_n > \theta
	\label{eq:general-perceptron}
\end{align}
with $\alpha_i$ being the weight associated with the ith image predicate $\varphi_i$, and $\theta$ the threshold or the negative of the bias.
As such ``each predicate of $\Phi$ is supposed to provide some evidence about whether $\psi$ is true for any figure $X$.''\footnote{
	\cite{Minsky.1969}, p.~11.
}  
The expression on the right of the equivalence sign, written with the notations used here, is the general case of Eq.~(\ref{eq:perceptron}).
%
% CMES style rewriting
The authors of
\cite{Minsky.1969}, p.~12, then defined the Rosenblatt perceptron as a special case of Eq.~(\ref{eq:general-perceptron}) in which the image predicates in $\Phi$ were random Boolean functions, generated by a random process according to a probability distribution.

The next paper to read is \cite{Block1962a}, which was based on the book by Rosenblatt (1962) \cite{Rosenblatt.1962}, and from where the following equation\footnote{
	Eq.~(4) in \cite{Block1962a}.
} can be identified as being similar to Eq.~(\ref{eq:general-perceptron}) and Eq.~(\ref{eq:perceptron}), again in its original notation as
\begin{align}
	\sum_{\mu = 1}^{N_a} y_\mu b_{\mu i} > \Theta
	\ , \text{ for } i = 1, \ldots , n
	\ ,
	\label{eq:Block1962-linear-combo-2}
\end{align}
where $y_\mu$ was the weight corresponding to the input 
$b_{\mu i}$ to the ``associated unit'' $a_\mu$ (neuron) from the stimulus pattern $S_i$ (ith example in the dataset), $N_a$ the number of ``associated units'' (neurons), $n$ the number of ``stimulus patterns'' (examples in the dataset), and $\Theta$ the second of two thresholds, which were fixed real non-negative numbers, and which corresponded to the negative of the bias $\biassp{i}{\ell}$ in Eq.~(\ref{eq:linearComboInputsBias}), or $(-b)$ in Figure~\ref{fig:perceptron}.

To discriminate between two classes, the input $b_{\mu i}$ took the value $+1$ or $-1$, when there was excitation coming from the stimulus pattern $S_i$ to the neuron $a_\mu$, and the value $0$ when there was no excitation from $S_i$ to $a_\mu$. 
When the weighted voting or linear combination in Eq.~(\ref{eq:Block1962-linear-combo-2}) surpassed the threshold $\Theta$, then the response was correct (or yields the value +1).

If the algebraic sum $\alpha_\mu^i$ of the connection strengths $C_{\sigma \mu}$ between the neuron (associated unit) $a_\mu$ and the sensory unit $s_\sigma$ inside the pattern (example) $S_i$ surpassed a threshold $\theta$ (which was the first of two thresholds, and which does not correspond to the negative of the bias in modern networks), then the neuron $a_\mu$ was activated:\footnote{
	First equation, unnumbered, in \cite{Block1962a}.  That this equation was unnumbered also indicated that it would not be subsequently referred to (and hence perhaps not considered as important).
} 
%https://stackoverflow.com/questions/3098680/how-to-put-a-symbol-above-another-in-latex
% {\stackrel{\sigma}{s_\sigma \in S_i}}
% {{\sigma} \above {s_\sigma \in S_i}}
\begin{align}
	\alpha_\mu^i 
	:=
	% \sum_{\sigma \text{ s.t. } s_\sigma \in S_i}^{} 
	% \sum_{\overset{\sigma}{s_\sigma \in S_i}}^{}
	% \sum_{\stackrel{\sigma}{s_\sigma \in S_i}}^{}
	\sum_{ \mbox{${\sigma} \atop {s_\sigma \in S_i}$} }
	C_{\sigma \mu} 
	\ge
	\theta
	\label{eq:Block1962-linear-combo-1}
\end{align}
Eq.~(\ref{eq:Block1962-linear-combo-1}) in \cite{Block1962a} would correspond to Eq.~(\ref{eq:rosenblatt-linear-combo-bias}) in \cite{Rosenblatt.1958}, with the connection strengths $C_{\sigma \mu}$ being ``random numbers having the possible values +1, -1, 0''. 

That Eq.~(\ref{eq:Block1962-linear-combo-1}) was not numbered in \cite{Block1962a} indicates that it played a minor role in this paper.  The reason is clear, since 
%
% CMES style rewriting
the author of
\cite{Block1962a} stated\footnote{
	See above Eq.~(1) in \cite{Block1962a}.
} that ``the connections $C_{\sigma \mu}$ do not change'', and thus ``we may disregard the sensory retina altogether'', i.e., Eq.~(\ref{eq:Block1962-linear-combo-1}).

Moreover, 
%
% CMES style rewriting
%\cite{Block1962a} wrote in 
the very first sentence 
in \cite{Block1962a} was
%of the paper 
``The perceptron is a self-organizing and adaptive system proposed by Rosenblatt'', and 
%immediately cited 
the book by \cite{Rosenblatt.1962} was immediately cited as Ref.~1, whereas only much later in the fourth page of \cite{Block1962a} did
the author write ``With the Perceptron, Rosenblatt offered for the first time a model...'', and cited  Rosenblatt's 1958 report first as Ref.~34, followed by the paper \cite{Rosenblatt.1958} as Ref.~35. 

In a major work on AI dedicated to Rosenblatt after his death in a boat accident, 
%
% CMES style rewriting
the authors of
\cite{Minsky.1969}, p.xi, in the Prologue of their book, referred to 
%the book \cite{Rosenblatt.1962}, 
Rosenblatt's (1962) book \cite{Rosenblatt.1962}
and not 
%the paper \cite{Rosenblatt.1958}:
Rosenblatt's (1958) paper \cite{Rosenblatt.1958}:
\begin{quote}
	``{\bf The 1960s: Connectionists and Symbolists}
	\\
	Interest in connectionist networks revived dramatically in 1962 with the publication  of Frank Rosenblatt's book {\em Principles of Neurodynamics} in which he defined the machines he named perceptrons and proved many theories about them.''
\end{quote}

In fact, 
%
% CMES style rewriting
%\cite{Minsky.1969} never referred to the paper \cite{Rosenblatt.1958},
Rosenblatt's (1958) paper \cite{Rosenblatt.1958} was never referred to in \cite{Minsky.1969}, 
except for a brief mention of the influence of ``Rosenblatt's [1958]'' work on p.~19, without 
%ever giving 
the full bibliographic details.  The authors of \cite{Minsky.1969} wrote: 
%of the paper \cite{Rosenblatt.1958}:
\begin{quote}
	``However, it is not our  goal  here  to  evaluate  these  theories [to model brain functioning],  but  only  to  sketch  a picture  of the  intellectual  stage  that  was  set  for  the  perceptron concept. In this setting,  Rosenblatt's  [1958]  schemes quickly  took root, and soon there were perhaps  as many  as  a  hundred  groups, large and small, experimenting  with  the  model  either as a  `learn­ing machine'  or  in  the  guise  of `adaptive'  or  `self-organizing' networks or `automatic control' systems.''
\end{quote}

So why was \cite{Rosenblatt.1958} often referred to for Eq.~(\ref{eq:linear-combo}) or Eq.~(\ref{eq:perceptron}), instead of \cite{Block1962a} or \cite{Rosenblatt.1962},\footnote{
	A search on the Web of Science on 2019.07.04 indicated that \cite{Rosenblatt.1958} received 2,346 citations, whereas \cite{Block1962a} received 168 citations.
	A search on Google Books on the same day indicated that \cite{Rosenblatt.1962} received 21 citations. 
} which would be much better references for these equations?
One reason could be that citing \cite{Block1962a} would not do justice to \cite{Rosenblatt.1958}, which contained the germ of the idea, even though not as refined as four years later in \cite{Block1962a} and \cite{Rosenblatt.1962}.
Another reason could be the herd effect by following other authors who referred to \cite{Rosenblatt.1958}, without actually reading the paper, or without comparing this paper to \cite{Block1962a} or \cite{Rosenblatt.1962}.
A best approach would be to refer to both \cite{Rosenblatt.1958} and \cite{Block1962a}, as papers like these would be more accessible than books like \cite{Rosenblatt.1962}. 

\begin{rem}
	\label{rm:Rosenblatt-hype}
	{The hype on the Rosenblatt perceptron}
	{\rm 
		Mark I computer described in the 1958 New York Times article shown in Figure~\ref{fig:Rosenblatt-Mark-I}, together with the criticism of the Rosenblatt perceptron in \cite{Minsky.1969} for failing to represent the XOR function, led to an early great disappointment on the possibilities of AI when overreached expectations for such device did not pan out, and contributed to the first AI winter that lasted until the 1980s, with a resurgence in interest due to the development of backpropagation and application in psychology as reported in \cite{Rumelhart.1986}.
		But some sixty years since the Mark I computer, AI still cannot even think like human babies yet: 
		%
		% CMES style rewriting
%		, as \cite{Gopnik.2019} recently wrote: 
		``Understanding babies and young children may be one key to ensuring that the current ``AI spring'' continues---despite some chilly autumnal winds in the air'' \cite{Gopnik.2019}.
	}
	$\hfill\blacksquare$
	%\\
	%{\color{red} 2019.08.27.  THE HYPE here. }
\end{rem}

\subsubsection{Dynamic, time dependence, Volterra series}
\label{sc:dynamic-volterra-series}
For time-dependent input $\x (t)$, the {\em continuous temporal summation}, mentioned in \cite{Rosenblatt.1962}, p.~33, is present in all terms other than the constant term in the Volterra series\footnote{
	\cite{Dayan.2001}, p.~46.
	%{\color{red} [NOTE: shorten later ENDNOTE]}
	``The Volterra series is a model for non-linear behavior similar to the Taylor series. It differs from the Taylor series in its ability to capture 'memory' effects. The Taylor series can be used for approximating the response of a nonlinear system to a given input if the output of this system depends strictly on the input at that particular time. In the Volterra series the output of the nonlinear system depends on the input to the system at all other times. This provides the ability to capture the 'memory' effect of devices like capacitors and inductors.
	It has been applied in the fields of medicine (biomedical engineering) and biology, especially {\em neuroscience}.  In mathematics, a Volterra series denotes a functional expansion of a dynamic, nonlinear, time-invariant functional,'' in
	``Volterra series'', Wikipedia, 
	\href{https://en.wikipedia.org/w/index.php?title=Volterra_series&oldid=854737438}{version 12:49, 13 August 2018}.
}
of the estimated output $z(t)$ as a result of the input $x(t)$
\begin{align}
	\begin{array}{llll}
		z(t) 
		&
		=
		& 
		\Ks{0} 
		&
		\displaystyle
		+ 
		\sum_{n=1} 
		\int 
		\cdots 
		\int
		\Ks{n}
		( \tau_1 ,.\,.\,, \tau_n ) 
		\prod_{j=1}^n x (t - \tau_j) 
		d \tau_j
		%\nonumber
		\\[0.5em]
		&
		=
		&
		\Ks{0}
		&
		\displaystyle
		+
		\int
		\Ks{1}
		(\tau_1)
		x (t - \tau_1)
		d \tau_1
		+
		\iint
		\Ks{2}
		(\tau_1 , \tau_2)
		x (t - \tau_1)
		x (t - \tau_2)
		d \tau_1
		d \tau_2
		%\nonumber
		\\[0.5em]
		&
		\phantom{=}
		&
		\phantom{\Ks{0}}	
		&
		\displaystyle
		+
		\iiint
		\Ks{3}
		(\tau_1 , \tau_2 , \tau_3)
		x (t - \tau_1)
		x (t - \tau_2)
		x (t - \tau_3)
		d \tau_1
		d \tau_2
		d \tau_3
		+
		\cdots
	\end{array}
	\label{eq:volterra-series}
\end{align}
where $\Ks{n} (\tau_1 , \ldots , \tau_n)$ is the kernel of the nth order, with all integrals going from $\tau_j = 0$ to the current time $\tau_j = +\infty$, for $j = 1, \ldots, n$.   
The linear-order approximation of the Volterra series in Eq.~(\ref{eq:volterra-series}) is then
\begin{align}
	z(t)
	\approx
	\Ks{0} 
	\displaystyle
	+
	\int\limits_{\tau_1 = 0}^{\tau_1 = +\infty}
	\Ks{1}
	(\tau_1)
	x (t - \tau_1)
	d \tau_1
	=
	\Ks{0} 
	\displaystyle
	+
	\int\limits_{\tau = -\infty}^{\tau = t}
	\Ks{1}
	(t - \tau)
	x (\tau)
	d \tau	
	\label{eq:linear-volterra-series}
\end{align}

%{\color{red} BEGIN capitalize 2020.02.07.}

\noindent
with the continuous linear combination (weighted sum) appearing in the second term.   The convolution integral in Eq.~(\ref{eq:linear-volterra-series}) is the basis for convolutional networks for highly effective and efficient image recognition, inspired by mammalian visual system.  A review of convolutional networks outside the scope here, despite them being the ``greatest success story of biologically inspired artificial intelligence'' \cite{Goodfellow.2016}, p.~353.

For biological neuron models, both 
The input $x(t)$ and the continuous weighted sum $z(t)$ can be either currents, with nA (nano Ampere) as dimension, or firing rates (frequency), with Hz (Hertz) as dimension.

Eq.~(\ref{eq:linear-volterra-series}) is the continuous temporal summation, counterpart of the discrete spatial summation in Eq.~(\ref{eq:linearComboInputsBias}), with the constant term $\Ks{0}$ playing a role similar to that of the bias $\biassp{i}{\ell}$.
The linear kernel $\Ks{1} (\tau_1)$ (also called the Wiener kernel,\footnote{
	Since the first two terms in the Volterra series coincide with the first two terms in the Wiener series; see \cite{Dayan.2001}, p.~46.
} or synaptic kernel in brain modeling)\footnote{
	See \cite{Dayan.2001}, p.~234.
} is the weight on the input $x(t - \tau_1)$, with $\tau_1$ going from $-\infty$ to the current time $t$. 
In other words, the whole history of the input $x(t)$ prior to the current time has an influence on the output $z(t)$, with typically smaller weight for more distant input (fading memory). 
For this reason, the synaptic kernel used in the biological neuron firing-rate models is often chosen to have an exponential decay of the form:\footnote{
	See \cite{Dayan.2001}, p.~234, below Eq.~(7.3).
}
\begin{align}
	\Ks{s} (t) := \Ks{1} (t) = \frac{1}{\tau_s} \exp \left( \frac{-t}{\tau_s} \right)
	\label{eq:synaptic-kernel}
\end{align}
where $\tau_s$ is the synaptic time constant such that the smaller $\tau_s$ is, the less memory of past input values, and
\begin{align}
	\tau_s \rightarrow 0 \Rightarrow \Ks{s} \rightarrow \delta(t) \Rightarrow
	z(t) \rightarrow \Ks{0} + x(t)
	\label{eq:zero-tau-s}
\end{align}
i.e., the continuous weighted sum $z(t)$ would correspond to the instantaneous $x(t)$ (without memory of past input values) as the synaptic time constant $\tau_s$ goes to zero (no memory). 

\begin{rem}
	\label{rm:volterra-exponential-smoothing}
	{\rm 
		The discrete counterpart of the linear part of the Volterra series in Eq.~(\ref{eq:linear-volterra-series}) can be found in the exponential-smoothing time series in Eq.~(\ref{eq:exponential-smoothing-2}), with the kernel $\Ks{1} (t-\tau)$ being the exponential function $\beta^{(t-i)}$; see Section~\ref{sc:exponential-smoothing} on exponential smoothing in forecasting. The similarity is even closer when the synaptic kernel $\Ks{1}$ is of exponential form as in Eq.~(\ref{eq:synaptic-kernel}).
	}  
	$\hfill\blacksquare$
\end{rem}

In firing-rate models of the brain (see Figure~\ref{fig:firing-rate}), the function $x(t)$ represents an input firing-rate at a synapse, $\Ks{0}$ is called the background firing rate, and the weighted sum $z(t)$ has the dimension of firing rate (Hz.

For a neuron with $n$ pre-synaptic inputs $[x_1(t) , \ldots , x_n(t)]$ (either currents or firing rates) as depicted in Figure~\ref{fig:bio-neuron} of a biological neuron, the total input $z(t)$ (current or firing rate, respectively) going into the soma (cell body,  Figure~\ref{fig:bio-neuron}), called total somatic input, is a discrete weighted sum of all post-synaptic continuous weighted sums $z_i(t)$ expressed in Eq.~(\ref{eq:linear-volterra-series}), assuming the same synaptic kernel $\Ks{s}$ at all synapses:
\begin{align}
	z(t)
	=
	\overline{\K}_0
	+
	\sum_{i=1}^{n}
	w_i 
	\int\limits_{\tau = -\infty}^{\tau = t}
		\Ks{s}
		(t - \tau)
		x_i (\tau)
	d \tau	
	\label{eq:total-synaptic-current}
\end{align}
with $\overline{\K}_0$ being the constant bias,\footnote{
	The negative of the bias $\overline{\K}_0$ is the threshold.  The constant bias $\overline{\K}_0$ is called the background firing rate when the inputs $x_i(t)$ are firing rates.
}
and $w_i$ the synaptic weight associated with the synapse $i$.

%{\color{red} END capitalize 2020.02.07.}

Using the synaptic kernel Eq.~(\ref{eq:synaptic-kernel}) in Eq.~(\ref{eq:total-synaptic-current}) for the total somatic input $z(t)$, and differentiate,\footnote{
	In general, for 
	$I(t) = \int_{x=A(t)}^{x=B(t)} f(x,t) dx$, 
	then
	$
	\frac{d I(t)}{dx}
	=
	f(B(t), t) \dt B (t)
	-
	f(A(t), t) \dt A (t)
	+
	\int_{x=A(t)}^{x=B(t)} \frac{\partial f(x,t)}{\partial x}  dx
	$.
}
the following ordinary differential equation is obtained
\begin{align}
	\boxit{
	\tau_s \frac{d z}{dt}
	=
	- z
	+
	\sum_{i}
	w_i x_i
	}
	\label{eq:firing-rate-ODE}
\end{align}

\begin{rem}
	\label{rm:steady-state-RNN-ODE}
	{\rm 
		The second term in Eq.~(\ref{eq:firing-rate-ODE}), with time-independent input $x_i$, is the steady state of the total somatic input $z(t)$:
		\begin{align}
			&
			z(t) = [z_0 - z_\infty] \exp \left( \frac{-t}{\tau} \right) + z_\infty \ , 
			\\
			&
			\text{ with }
			z_0 := z(0) \ , \text{ and }
			z_\infty := \sum_{i} w_i x_i
			\label{eq:steady-state}
			\\
			&
			z(t) \rightarrow z_\infty \text{ as } t \rightarrow \infty
			\ .
		\end{align}
		As a result, the subscript $\infty$ is often used to denote as the steady state solution, such as $R_\infty$ in 
		%
		% CMES style rewriting
%		\cite{Wilson.1999}'s 
		the model of neocortical neurons Eq.~(\ref{eq:Wilson-eq-2}) \cite{Wilson.1999}.
	}
	$\hfill\blacksquare$
\end{rem}

For constant total somatic input $z$, the output firing-rate $y$ is given by an activation function $a(\cdot)$ (e.g, scaled ReLU in Figure~\ref{fig:diode-ReLU} and Figure~\ref{fig:halfwave})  through relation\footnote{
	See \cite{Dayan.2001}, p.~234, in original notation as $v = F(I_s)$, with $v$ being the output firing rate, $F$ an activation function, 
	and $I_s$ the total synaptic current.
}
\begin{align}
\boxit{
	y = a(z) = c \max(0, z) = c [z]_+
}
	\label{eq:firing-rate-constant-current}
\end{align} 
where 
$c$ is a scaling constant to match the slope of the firing rate (F) vs input current (I) relation called the FI curve obtained from experiments; see Figure~\ref{fig:firing-rate} and Figure~\ref{fig:neuron-firing}.
the total somatic input 
$z$ can be thought of as being converted from current (nA) to frequency (Hz) by multiplying with the converting constant $c$.\footnote{
	See \cite{Dayan.2001}, p.~234, subsection ``The Firing Rate''.
}

At this stage, there are two possible firing-rate models.
The first firing-rate model consists of 
(1) Eq.~(\ref{eq:firing-rate-ODE}), the ODE for the total somatic input firing rate $z$, followed by 
(2) the ``static'' relation between output firing-rate $y$ and constant input firing rate $z$, expressed in Eq.~(\ref{eq:firing-rate-constant-current}), but now used for time-dependent total somatic input firing rate $z(t)$.

\begin{rem}
	\label{rm:steady-state-output}
	{\rm
		The steady-state output $y_\infty$ in the first firing-rate model described in Eq.~(\ref{eq:total-synaptic-current}) and Eq.~(\ref{eq:firing-rate-constant-current}) in the case of constant inputs $x_i$ is therefore
		\begin{align}
			y_\infty = a (z_\infty)
			\ ,
		\end{align}
		where $z_\infty$ is given by Eq.~(\ref{eq:steady-state}).
	}
	$\hfill\blacksquare$
\end{rem}

The second firing-rate model consists of using Eq.~(\ref{eq:firing-rate-ODE}) for the total somatic input firing rate $z$, which is then used as input for the following ODE for the output firing-rate $y$:
\begin{align}
\boxit{
	\tau_r \frac{d y}{dt}
	=
	- y
	+
	a(z(t))
	\ ,
}
	\label{eq:firing-rate-model-2}
\end{align}
where the activation function $a(\cdot)$ is applied on the time-dependent total somatic input firing rate $z(t)$, but with a different time constant $\tau_r$, which describes how fast the output firing rate $y$ approaches steady state for {\em constant} input $z$.

Eq.~(\ref{eq:firing-rate-model-2}) is a recurring theme that has been frequently used in papers in neuroscience and artificial neural networks.  Below are a few relevant papers for this review, particularly the \emph{continuous} recurrent neural networks (RNNs)---such as Eq.~(\ref{eq:hahnloser-1}), Eq.~(\ref{eq:dimirosski-RNN-ODE}), and Eqs.~(\ref{eq:space-time-continous-RNN-1})-(\ref{eq:space-time-continous-RNN-2})---which are the counterparts of the \emph{discrete} RNNs in Section~\ref{sc:RNN}.

\begin{figure}[h]
	\centering
	%
	% 2022.12.17
	% add "-eps-converted-to.pdf" for arXiv
	% \includegraphics[width=0.4\linewidth]{figures/Wilson-membrane}
	\includegraphics[width=0.4\linewidth]{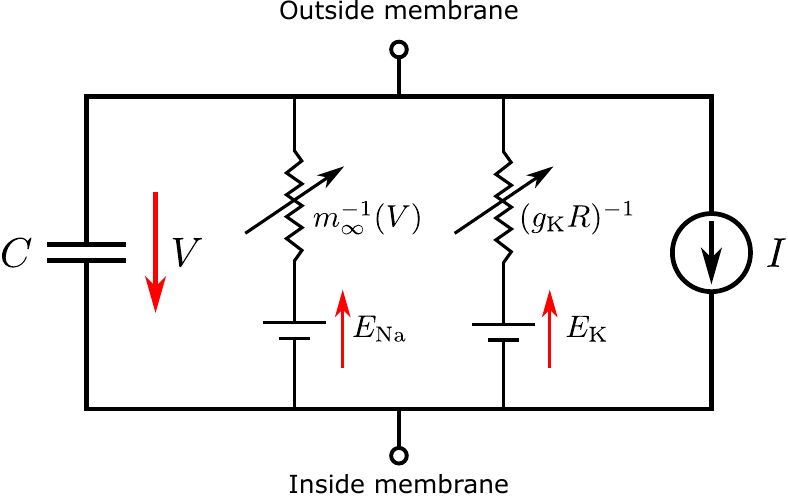}
	\caption{  
		%
		% CMES style rewriting
%		\cite{Wilson.1999}'s 
		\emph{Model of neocortical neurons} in \cite{Wilson.1999} as a simplification of the model in \cite{Hodgkin.1952} (Section~\ref{sc:dynamic-volterra-series}): A capacitor $C$ with a potential $V$ across its plates, in parallel with the equilibrium potentials $E_{\text{Na}}$ (sodium) and $E_{\text{K}}$ (potassium) in opposite direction.  Two variable resistors $m_\infty^{-1} (V)$ and $[g_{\text{K}} R (V)]^{-1}$ are each in series with one of the mentioned two equilibrium potentials.  The capacitor $C$ is also in parallel with a current source $I$.  The notation $R$ is used here for the ``recovery variable'', not a resistor.  See Eqs.~(\ref{eq:Wilson-eq-1})-(\ref{eq:Wilson-eq-2}).	
	}
	\label{fig:Wilson-model}
\end{figure}

%
% CMES style rewriting
%\cite{Wilson.1999}'s 
The
model for neocortical neurons in \cite{Wilson.1999}, a simplification of the model by \cite{Hodgkin.1952}, was employed 
%by 
in
\cite{Steyn-Ross.2016:rd0001} as starting point to develop a formulation that produced SubFigure(b) in Figure~\ref{fig:neuron-firing} consists of two coupled ODE's\footnote{
	Eq.~(1) and Eq.~(2) in \cite{Steyn-Ross.2016:rd0001}.
}
\begin{align}
	C \frac{dV}{dt} & = - m_\infty (V) (V-E_{\text{Na}}) - g_{\text{K}} R (V-E_{\text{K}}) + I
	\label{eq:Wilson-eq-1}
	\\
	\tau \frac{d R}{d t} & = -R + R_\infty (V)
	\label{eq:Wilson-eq-2}
\end{align}
where $m_\infty (V)$ and $R_\infty (V)$ are prescribed quadratic polynomials in the potential $V$, making the right-hand side of Eq.~(\ref{eq:Wilson-eq-1}) of cubic order.
Eq.~(\ref{eq:Wilson-eq-1}) describes the change in the membrane potential $V$ due to the capacitance $C$ in parallel with other circuit elements shown in Figure~\ref{fig:Wilson-model}, with 
(1) $m_\infty (V)$ and $E_{\text{Na}}$ being the activation function and equilibrium potential for the sodium ion (Na$^+$), respectively,
(2) $g_{\text{K}}$, $R$, and $E_{\text{K}}$ being the conductance, recovery variable, and equilibrium potential for the potassium ion (K$^+$), respectively, and 
(3) $I$ the stimulating current.
Eq.~(\ref{eq:Wilson-eq-2}) for the recovery variable $R$ has the same form as Eq.~(\ref{eq:firing-rate-model-2}), with $R_\infty (V)$ being the steady state; see Remark~\ref{rm:steady-state-output}.

To create a continuous recurrent neural network described by ODEs in Eq.~(\ref{eq:hahnloser-1}), the input $x_i (t)$ in Eq.~(\ref{eq:total-synaptic-current}) is replaced by the output $y_j (t)$ (i.e., a feedback), and the bias $\overline{\mathcal K}_0$ becomes an input, now denoted by $x_i$.
Electrical circuits can be designed to approximate the dynamical behavior a spatially-discrete, {\em temporally-continuous} recurrent neural network (RNN) 
described by Eq.~(\ref{eq:hahnloser-1}), which is Eq.~(1) in \cite{Hahnloser.2000:rd0002}:\footnote{
	In original notation, Eq.~(\ref{eq:hahnloser-1}) was written as 
	$	
	\tau_i (x_i)
	\frac{d x_i}{dt}
	+
	x_i
	=
	\left[
	b_i
	+
	\sum_j W_{ij} x_j
	\right]_+
	$
	in \cite{Hahnloser.2000:rd0002}, whose outputs $x_i$ in the previous expression are now rewritten as $y_i$ in Eq.~(\ref{eq:hahnloser-1}), and the biases $b_i (t)$, playing the role of inputs, are rewritten as $x_i (t)$ to be consistent with the notation for inputs and outputs used throughout in the present work; see Section~\ref{sc:matrix} on matrix notation, Eq.~(\ref{eq:matrices_x_y}), and Section~\ref{sc:RNN} on
	{\em discrete} recurrent neural networks, which are discrete in both space and time.
	The paper 
	%
	% CMES style rewriting, just remove "by"
%	by 
	\cite{Hahnloser.2000:rd0002} was cited in both \cite{Dayan.2001} and \cite{Ramachandran.2017:rd0001}, with the latter leading us to it.
}
\begin{align}
	\tau_i
	\frac{d y_i}{dt}
	=
	- 
	y_i
	+
	\left[
	x_i
	+
	\sum_j w_{ij} y_j
	\right]_+
	\ .
	\label{eq:hahnloser-1}
\end{align}
The network is called symmetric if the weight matrix is symmetric, i.e.,
\begin{align}
	w_{ij} = w_{ji}
	\ .
	\label{eq:symmetric-weights}
\end{align}
The difference between Eq.~(\ref{eq:hahnloser-1}) and Eq.~(\ref{eq:firing-rate-model-2}) is that Eq.~(\ref{eq:firing-rate-model-2}) is based on the expression for $z(t)$ in Eq.~(\ref{eq:total-synaptic-current}), and thus has no feedback loop.

\begin{figure}[h]
	\centering
	\includegraphics[width=0.3\linewidth]{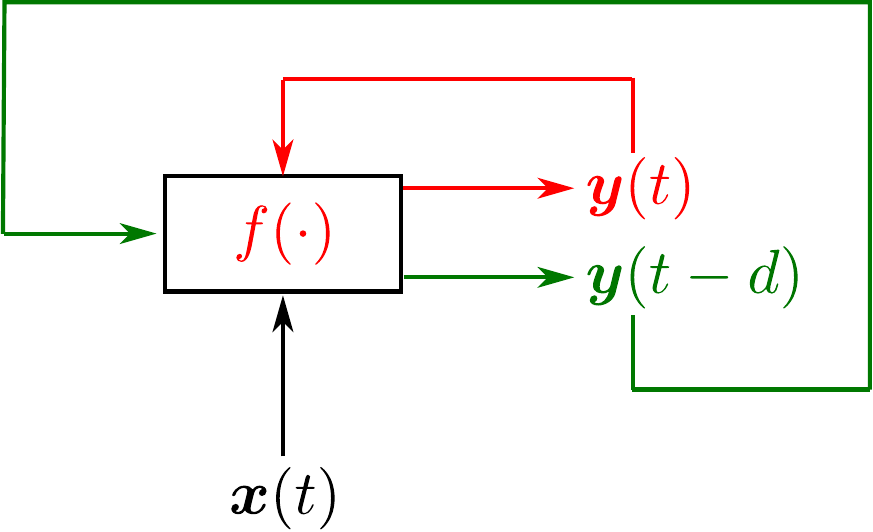}
	\caption{  
		\emph{Continuous recurrent neural network with time-dependent delay} $d(t)$ (green feedback loop, Section~\ref{sc:dynamic-volterra-series}), as expressed in Eq.~(\ref{eq:continuous-RNN-general}), where	
		$f(\cdot)$ is the operator with the first defivative term plus a standard static term---which is an activation function acting on linear combination of input and bias, i.e., $a(z(t))$ as in Eq.~(\ref{eq:activationFunction}) and Eq.~(\ref{eq:expanded-output-1})---$\bx(t)$ the input, $\by (t)$ the output with the red feedback loop, and $\by(t-d)$ the delayed output with the green feedback loop.
		This figure is the more general continuous counterpart of the discrete RNN in Figure~\ref{fig:our-RNN}, represented by Eq.~(\ref{eq:RNN-1}), which is a particular case of  Eq.~(\ref{eq:continuous-RNN-general}).
		We also refer readers to Remark~\ref{rm:hidden-cell} and the notation equivalence $\by(t) \equiv \bh(t)$ as noted in Eq.~(\ref{eq:equiv-y-h-cell}).	
	}
	\label{fig:continuous-RNN-delay}
\end{figure}

A time-dependent time delay $d(t)$ can be introduced into Eq.~(\ref{eq:hahnloser-1}) leading to spatially-discrete, {\em temporally-continuous} RNNs with time delay, which is Eq.~(1) in \cite{Dimirovski.2017}:\footnote{
	In original notation, Eq.~(\ref{eq:dimirosski-RNN-ODE}) was written as $\dt z = - A z + f( W z (t - h(t)) + J)$ in \cite{Dimirovski.2017}, where $\z = \by$,
	$A = \boldsymbol{T}^{-1}$, $h(t) = d(t)$ and $J = \bx$.
}
\begin{align}
	\boldsymbol{T} \frac{d \by}{dt} 
	= 
	- \by + a [ \bx + \bWeight \by (t - d(t)) ]
	\label{eq:dimirosski-RNN-ODE}
\end{align}
where the diagonal matrix $\boldsymbol{T} = \text{Diag} [\tau_1 , ..., \tau_n] \in \real^{n \times n}$ contains the synaptic time constants $\tau_i$ as its diagonal coefficients, the matrix $\by = [\y_{1} , \ldots , \y_n] \in \real^{n \times 1}$ contains the outputs, the bias matrix $\bx \in \real^{n \times 1}$ plays the role of input matrix (thus denoted by $\bx$ instead of $\bbias$), $a(\cdot)$ is the activation function, $\bWeight \in \real^{n \times n}$ the weight matrix, and $d(t)$ the time-dependent delay; see Figure~\ref{fig:continuous-RNN-delay}.

For discrete RNNs, the delay is a constant integer set to one, i.e.,
\begin{align}
	d(n) = 1
	\ ,
	\text{ with }
	n = t \text{ (integer)}
	\ ,
	\label{eq:delay-single-step}
\end{align}
as expressed in Eq.~(\ref{eq:RNN-1}) in Section~\ref{sc:RNN}.

Both Eq.~(\ref{eq:hahnloser-1}) and Eq.~(\ref{eq:dimirosski-RNN-ODE}) can be rewritten in the following form:
\begin{align}
	\by (t)
	=
	- \boldsymbol{T} \frac{d \by (t)}{dt}
	+ a [ \bx (t) + \bWeight \by (t - d(t)) ]
	=
	f( \by (t) , \by(t-d) \bx (t) , d (t) )
	\ .
	\label{eq:continuous-RNN-general}
\end{align}

%pdf p.~251
A densely distributed pre-synaptic input points [see Eq.~(\ref{eq:total-synaptic-current}) and Figure~\ref{fig:bio-neuron} of a biological neurons] can be approximated by a continuous distribution in space, represented by $x(s , t)$, with $s$ being the space variable, and $t$ the time variable.   In this case,  a {\em continuous} RNN in both space and time, called ``continuously labeled RNN'', can be written as follows:\footnote{
	\cite{Dayan.2001}, p.~240, Eq.~(7.14).
}
\begin{align}
	\tau_r \frac{\partial y (s,t)}{\partial t} 
	& =
	- y (s,t)
	+
	a(z(s,t))
	\label{eq:space-time-continous-RNN-1}	
	\\
	z(s,t) 
	& = 
	\rho_s
	\int
	\left[
		W(s , \gamma) x(\gamma , t)
		+
		M(s , \gamma) y(\gamma , t)
	\right]
	d \gamma
	\label{eq:space-time-continous-RNN-2}
\end{align}
where $\rho_s$ is the neuron density, assumed to be constant. Space-time continuous RNNs such as Eqs.~(\ref{eq:space-time-continous-RNN-1})-(\ref{eq:space-time-continous-RNN-2}) have been used to model, e.g., the visually responsive neurons in the premotor cortex \cite{Dayan.2001}, p.~242.

\begin{figure}[h]
	\centering
	\begin{subfigure}[b]{0.40\textwidth}
		\includegraphics[width=\linewidth]{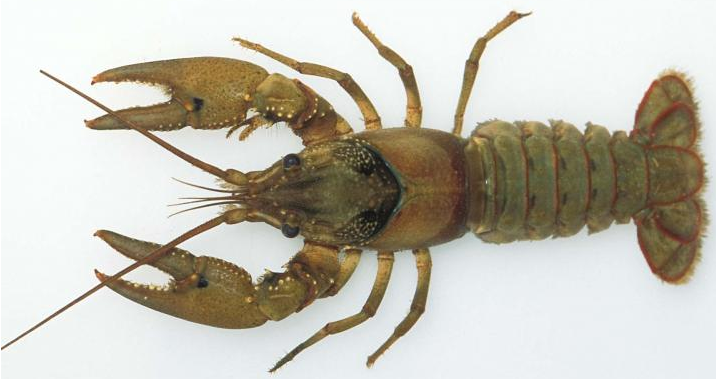}
		\caption{Crayfish. Courtesy of Missouri Dept of Conservation}
	\end{subfigure}
	\ 
	\begin{subfigure}[b]{0.56\textwidth}
		\includegraphics[width=\linewidth]{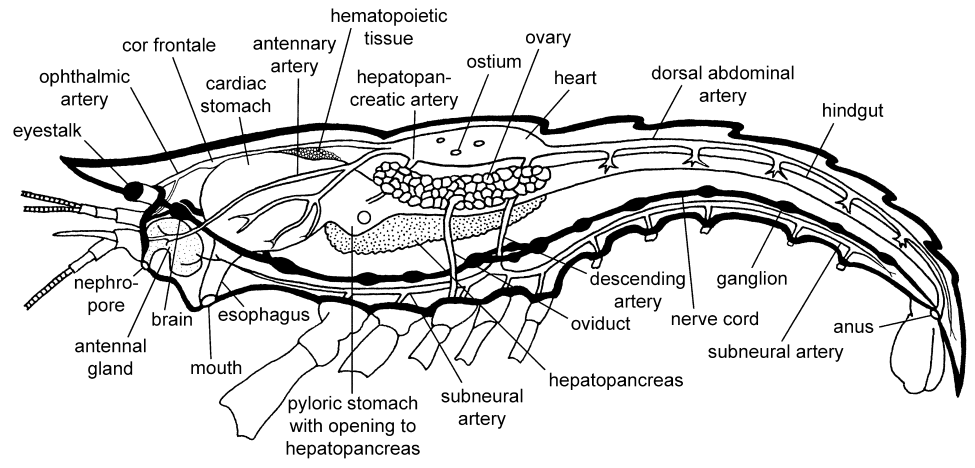}
		\caption{
			Anatomy. Abdominal nerve cord and ganglions (black).
			\cite{Gherardi.2009}.
			A portion of the abdominal nerve cord and a ganglion are shown in Figure~\ref{fig:crayfish-synapse}. 	
			{\footnotesize (Figure reproduced with permission of the authors.)}
		}
	\end{subfigure}
	\caption{
		\emph{Crayfish} (Section~\ref{sc:ReLU-history}), freshwater crustaceans. Anatomy.
	}
	\label{fig:crayfish-anatomy}
\end{figure}

\subsection{Activation functions}
\label{sc:active-function-history}

\subsubsection{Logistic sigmoid}
\label{sc:sigmoid-history}
The use of the logistic sigmoid function (Figure~\ref{fig:sigmoid}) in neuroscience dates back since the seminal work of Nobel Laureates Hodgkin \& Huxley (1952) \cite{Hodgkin.1952} in the form of an electrical circuit (Figure~\ref{fig:Wilson-model}), and since the work reported in \cite{Little.1974:rd0001} in a form closer to today's network; see also \cite{Han.1995} and \cite{Wuraola.2018:rd0001}.

%
% CMES style rewriting
The authors of
\cite{Goodfellow.2016}, p.~219, remarked: ``Despite the early popularity of rectification (see next Section~\ref{sc:ReLU-history}), it was largely replaced by sigmoids in the 1980s, perhaps because sigmoids perform better when neural networks are very small.''

The rectified linear function has, however, made a come back and was a key component responsible for the success of deep learning, and helped inspired a variant that in 2015 surpassed human-level performance in image classification, as ``it expedites convergence of the training procedure [16] and leads to better solutions [21, 8, 20, 34] than conventional sigmoid-like units'' \cite{He.2015b}.  See Section~\ref{sc:parametric-ReLU} on Parametric Rectified Linear Unit.

%{\color{red} HERE, 2020.02.08}

\begin{figure}[h]
	\centering
	%
	% 2022.12.17
	% add "-eps-converted-to.pdf" for arXiv
	% \includegraphics[width=0.8\linewidth]{figures/Furshpan1959-crayfish-synapse-2018-12-08-B}
	\includegraphics[width=0.8\linewidth]{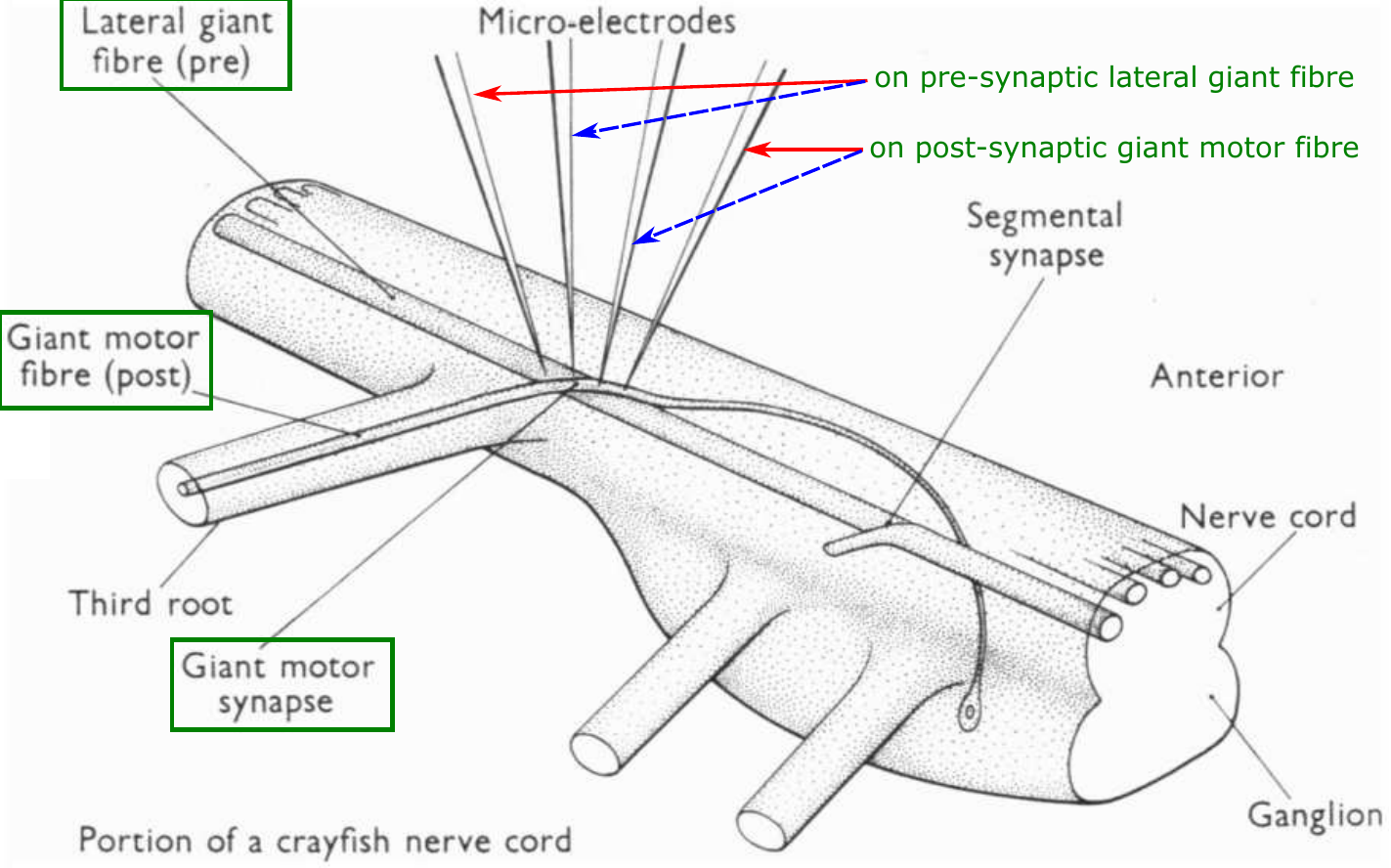}
	\caption{  
		{\em Crayfish giant motor synapse} (Section~\ref{sc:ReLU-history}).  The (pre-synaptic) lateral giant fiber was connected to the (post-synaptic) giant motor fiber through a synapse where the two fibers cross each other at the location  annotated by ``Giant motor synapse'' in the figure.  This synapse was right underneath the giant motor fiber, at the crossing and contact point, and thus could not be seen.  The two left electrodes (including the second electrode from left) were inserted in the lateral giant fiber, with the two right electrodes in the giant motor fiber.  Currents were injected into the two electrodes indicated by solid red arrows, and electrical outputs recorded from the two electrodes indicated by dashed blue arrows   	
		\cite{Furshpan1959a}.
		%{\color{red} ASK PERMISSION}
		{\footnotesize (Figure reproduced with permission of the publisher Wiley.)}	
	}
	\label{fig:crayfish-synapse}
\end{figure}

\begin{figure}[h]
	\centering
	\begin{subfigure}[b]{0.55\textwidth}
		\includegraphics[width=\linewidth]{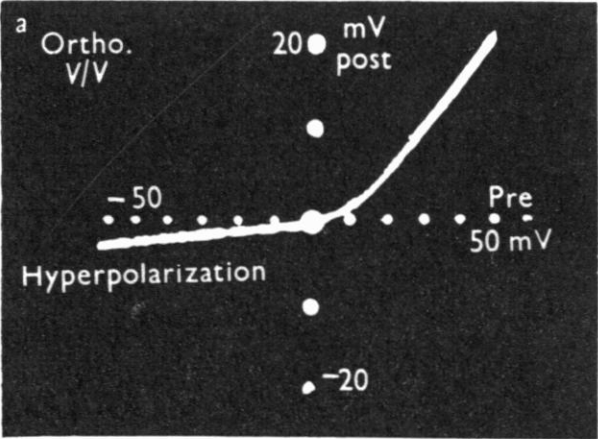}
		\caption{  
			Post-synaptic voltage vs pre-synaptic voltage.		
			Experimental results indicated that currents flow from the pre-synaptic lateral giant fiber, through the synapse, into the post-synaptic giant motor fiber essentially in one direction, with a small leakage when polarity was reversed.		
		}
	\end{subfigure}
	\begin{subfigure}[b]{0.40\textwidth}
		\includegraphics[width=\linewidth]{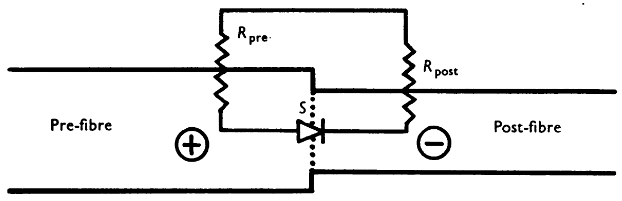}
		\caption{  
			Rectifier circuit model for crayfish giant motor synapse. Conceptual drawing, where the vertical dotted line represents the synapse.  The large tube on the left represents the pre-synaptic lateral giant fiber, the smaller tube on the right the post-synaptic giant motor fiber.	
		}
	\end{subfigure}
	\caption{
		\emph{Crayfish Giant Motor Synapse} (Section~\ref{sc:ReLU-history}).  The response in SubFigure~(a) is similar to that of a rectifier circuit with leaky diode in Figure~\ref{fig:diode-ReLU} and Figure~\ref{fig:I-V-halfwave} [red curve ($-V = V_D + V_R$) in SubFigure~(b)]
		\cite{Furshpan1959a}.
		%{\color{red} ASK PERMISSION}
		{\footnotesize (Figure reproduced with permission of the publisher Wiley.)}
	}
	\label{fig:crayfish-rectifier-experiment}
\end{figure}

\subsubsection{Rectified linear unit (ReLU)}
\label{sc:ReLU-history}
Yoshua Bengio, the senior author of \cite{Goodfellow.2016} and a Turing Award recipient, recounted the rise in popularity of ReLU in deep learning networks in 
%his 
an
interview 
%with
in 
\cite{Ford.2018}:
\begin{quote}
	``The big question was how we could train deeper networks...  Then a few years later, we discovered that we didn't need these approaches [Restricted Boltzmann Machines, autoencoders] to train deep networks, we could just change the nonlinearity. One of my students was working with neuroscientists,
	and we thought that we should try rectified linear units (ReLUs)---we
	called them rectifiers in those days---because they were more
	biologically plausible, and this is an example of actually taking
	inspiration from the brain. 
	We had previously used a sigmoid function to
	train neural nets, but it turned out that by using ReLUs we could
	suddenly train very deep nets much more easily. That was another big
	change that occurred around 2010 or 2011.''
\end{quote}
The student mentioned by Bengio was likely the first author of \cite{Glorot.2011:rd0001}; see also the earlier Section~\ref{sc:activation-functions} on activation functions.

We were aware of Ref.~\cite{Hahnloser.2000:rd0002} appearing in year 2000---in which a spatially-discrete, temperally-continuous recurrent neural network was used with a rectified linear function, as expressed in Eq.~(\ref{eq:hahnloser-1})---through Ref.~\cite{Ramachandran.2017:rd0001}.
On the other hand, prior its introduction in deep neural networks, rectified linear unit had been used in neuroscience since at least 1995, but \cite{Bush.1995} was a book, as cited in \cite{Glorot.2011:rd0001}.  Research results published in papers would appear in book form several years later:
\begin{quote}
	`` The current and third wave, deep learning, started around 2006 (Hinton et al., 2006; Bengio et al., 2007; Ranzato et al., 2007a) and is just now appearing in book form as of 2016. The other two waves [cybernetics and connectionism] similarly appeared in book form much later than the corresponding scientific activity occurred'' \cite{Goodfellow.2016}, p.~13.
\end{quote}

Another clue that the rectified linear function was a well-known, well accepted concept---similar to the relation $\boldsymbol{Kd = F}$ in the finite element method---is that 
%
% CMES style rewriting
the authors of
\cite{Hahnloser.2000:rd0002} did not provide any reference to their own important Eq.~(1), which is reproduced in Eq.~(\ref{eq:hahnloser-1}), as if it was already obvious to anyone in neuroscience. 

Indeed, more than sixty years ago, in a series of papers \cite{Furshpan1957} \cite{Furshpan1959a} \cite{Furshpan1959b}, Furshpan \& Potter established that current flows through a crayfish neuron synapse (Figure~\ref{fig:crayfish-anatomy} and Figure~\ref{fig:crayfish-synapse}) in essentially one direction, thus deducing that the synapse can be modeled as a rectifier, diode in series with resistance, as shown in Figure~\ref{fig:crayfish-rectifier-experiment}.

\subsubsection{New active functions}
\label{sc:new-active-functions}
\begin{figure}[h]
	\centering
	\includegraphics[width=1.0\linewidth]{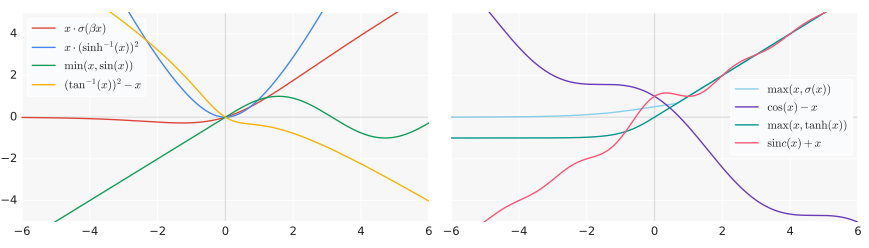}
	\caption{
		\emph{Swish function} (Section~\ref{sc:new-active-functions}) $x \cdot \sigmoid(\beta x)$, with $\sigmoid(\cdot)$ being the logistic sigmoid in Figure~\ref{fig:sigmoid}, and other activation functions 
		\cite{Ramachandran.2017:rd0001}.
		%\\
		{\footnotesize (Figure reproduced with permission of the authors.)} 
	}
	\label{fig:swish}
\end{figure}

``The design of hidden units is an extremely active area of research and does not yet have many definitive guiding theoretical principles,'' \cite{Goodfellow.2016}, p.~186.
Indeed, 
%
% CMES style rewriting
%\cite{Ramachandran.2017:rd0001} found 
the Swish activation function in Figure~\ref{fig:swish} of the form
\begin{align}
a(x) = x \cdot \sigmoid(\beta x) = \frac{x}{1 + \exp (- \beta x )} \ ,
\end{align}
with $\sigmoid(\cdot)$ being the logistic sigmoid given in Figure~\ref{fig:sigmoid} and in
Eq.~\eqref{eq:logistic-sigmoid},
was found in \cite{Ramachandran.2017:rd0001} to outperform the rectified linear unit (ReLU) in a number of benchmark tests.

%new computationally efficient active function \cite{Wuraola.2018:rd0001}

On the other hand, it would be hard to beat the efficiency of the rectified linear function in both evaluating the weighted combination of inputs $\bzp{\ell}$ of layer $(\ell)$ and in computing the gradient with the first derivative of ReLU being the Heaviside function; see Figure~\ref{fig:ReLU}.

A zoo of activation functions is provided in ``Activation function'', Wikipedia,  \href{https://en.wikipedia.org/w/index.php?title=Activation_function&oldid=870454307}{version 22:46, 24 November 2018} and the more recent \href{https://en.wikipedia.org/w/index.php?title=Activation_function&oldid=1099333527}{version 06:30, 20 July 2022}, in which several active functions had been removed, e.g., the ``Square Nonlinearity (SQNL)''\footnote{
	The ``Square Nonlinearity (SQNL)'' activation, having a shape similar to that of the hyperbolic tangent function, appeared in the article ``Activation function'' for the last time in \href{https://en.wikipedia.org/w/index.php?title=Activation_function&oldid=1012711871}{version 22:00, 17 March 2021}, and was
	was removed from the table of activation functions starting from \href{https://en.wikipedia.org/w/index.php?title=Activation_function&oldid=1017248224}{version 18:13, 11 April 2021} with the comment ``Remove SQLU since it has 0 citations; it needs to be broadly adopted to be in this list; Remove SQNL (also from the same author, and this also does not have broad adoption)''; see the article \href{https://en.wikipedia.org/w/index.php?title=Activation_function&offset=&limit=500&action=history}{History}.
} listed in the 2018 version of this zoo.

%{\color{red} HERE, 2020.02.09}

%attention is all you need \cite{Vaswani.2017:rd0001}

\subsection{Back-propagation, automatic differentiation}
\label{sc:backprop-autodiff-history}
%
% CMES style rewriting
%\cite{Werbos.1990} defined:
``At  its core, backpropagation [Section~\ref{sc:backprop}] is  simply an efficient and exact method for calculating all the derivatives of a single target quantity  (such as  pattern classification error) with respect to a large set of input quantities (such as the parameters or weights in a classification rule)'' \cite{Werbos.1990}.

%
% CMES style rewriting
%\cite{Baydin.2018} stated 
In a survey on automatic differentiation in \cite{Baydin.2018}, it was stated that: 
``in simplest terms, backpropagation models learning as gradient descent in neural network weight space, looking for the minima of an objective function.''

Such statement identified back-propagation with an optimization method by  gradient descent.  
But according to 
%
% CMES style rewriting
the authors of
\cite{Goodfellow.2016}, p.~198, ``back-propagation is often misunderstood as meaning the whole learning algorithm for multilayer neural networks'', and clearly distinguish back-propagation {\em only} as a method to compute the gradient of the cost function with respect to the parameters, while another algorithm, such as stochastic gradient descent, is used to perform the learning using this gradient, where performing ``learning'' meant network training, i.e., find the parameters that minimize the cost function, which ``typically includes a performance measure evaluated on the entire training set as well as additional regularization terms.''\footnote{
	See \cite{Goodfellow.2016}, Chap.~8, ``Optimization for training deep models'', p.~267
}

According to \vphantom{\cite{Baydin.2018}}\cite{Baydin.2018}, automatic differentiation, or in short ``autodiff'', is ``a family of techniques similar to but more general than backpropagation for efficiently and accurately evaluating derivatives of numeric functions expressed as computer programs.''

\subsubsection{Back-propagation}
\label{sc:backprop-history}
In an interview published in 2018 \cite{Ford.2018}, Hinton confirmed that backpropagation was independently invented by many people before his own 1986 paper \cite{Rumelhart.1986}.  Here we focus on information that is not found in the review of backpropagation in \cite{Schmidhuber.2015:rd0001}.

%Hinton provided the following story behind the backpropagation paper \cite{Rumelhart.1986} in an interview with \cite{Ford.2018}:
%\begin{quote}
%	``Lots of different people invented different
%	versions of backpropagation before David Rumelhart. They were
%	mainly independent inventions, and it’s something I feel I’ve got too
%	much credit for.''
%\end{quote}
For example, the success reported in \cite{Rumelhart.1986} laid not in backpropagation itself, but in its use in psychology: 
\begin{quote}
	``Back in the mid-1980s, when computers were very slow, I used a
	simple example where you would have a family tree, and I would tell
	you about relationships within that family tree. I would tell you things like Charlotte's mother is Victoria, so I would say Charlotte and mother, and the correct answer is Victoria. I would also say Charlotte and father, and the correct answer is James. Once I've said those two things, because it's a very regular family tree with no divorces, you could use conventional AI to infer using your knowledge of family relations that Victoria must be the spouse of James because Victoria is Charlotte's mother and James is Charlotte's father. The neural net could infer that too, but it didn't do it by using rules of inference, it did it by learning a bunch of features for each person. Victoria and Charlotte would both be a bunch of separate features, and then by using interactions between those vectors of features, that would cause the output to be the features for the correct person. From the features for Charlotte and from the features for mother, it could derive the features for Victoria, and when you trained it, it would learn to do that. The
	most exciting thing was that for these different words, it would learn these feature vectors, and it was learning distributed representations of words.'' \cite{Ford.2018}
\end{quote}
For psychologists, ``a learning algorithm that could learn representations of things was a big breakthrough,'' and Hinton's contribution in \cite{Rumelhart.1986} was to show that ``backpropagation would learn these distributed representations, and that was what was interesting to psychologists, and eventually, to AI people.''
%\begin{quote}
	%``We submitted a paper to Nature in 1986 that had this example of backpropagation learning distributed features of words, and I talked to one of the referees of the paper, and that was what got him really excited about it, that this system was learning these distributed representations. He was a psychologist, and he understood that having a learning algorithm that could learn representations of things was a big breakthrough. My contribution was not discovering the backpropagation algorithm, that was something Rumelhart had pretty much figured out, it was showing that backpropagation would learn these distributed representations, and that was what was interesting to psychologists, and eventually, to AI people.''
%\end{quote}
But backpropagation lost ground to other technologies in machine learning:
\begin{quote}
	``In the early 1990s, ... the support vector machine did better at recognizing handwritten digits than backpropagation, and handwritten digits had been a classic example of backpropagation doing something really well. Because of that, the machine learning community really lost interest in backpropagation'' \cite{Ford.2018}.\footnote{
		See Footnote~\ref{fn:support-vector-machine} on how research on kernel methods (Section~\ref{sc:kernel-machines}) for Support Vector Machines have been recently used in connection with networks with infinite width to understand how deep learning works (Section~\ref{sc:lack-understanding}).
	}
\end{quote}
Despite such setback, psychologists still considered backpropagation as an interesting approach, and continued to work with this method:
\begin{quote}
	There is ``a distinction between AI and machine learning on the one hand, and psychology on the other hand. Once backpropagation became
	popular in 1986, a lot of psychologists got interested in it, and they didn't really lose their interest in it, they kept believing that it was an interesting algorithm, maybe not what the brain did, but an interesting way of developing representations'' \cite{Ford.2018}.
\end{quote}

\noindent
%{\color{red}
%	[NOTE: 2019.07.16, add comment by Hinton about the success for psychology of this 1986 Nature paper.]	
%}
%
% CMES style rewriting
The 2015 review paper 
\cite{Schmidhuber.2015:rd0001} referred to Werbos' 1974 PhD dissertation for a preliminary discussion of backpropagation (BP), 
\begin{quote}
	``Efficient BP was soon explicitly used to minimize cost functions
	by adapting control parameters (weights) (Dreyfus, 1973). Compare some preliminary, NN-specific discussion (Werbos, 1974, Section 5.5.1), a method for multilayer threshold NNs (Bobrowski, 1978), and a computer program for automatically deriving and implementing BP for given differentiable systems (Speelpenning, 1980).''
\end{quote}
and explicitly attributed to Werbos early applications of backpropagation in neural networks (NN): 
\begin{quote}
	``To my knowledge, the first NN-specific application of efficient
	backpropagation was described in 1981 (Werbos, 1981, 2006). Related work was published several years later (LeCun, 1985, 1988; Parker, 1985). A paper of 1986 significantly contributed to the popularization of BP for NNs (Rumelhart, Hinton, \& Williams, 1986), experimentally demonstrating the emergence of useful internal representations in hidden layers.''
\end{quote}
See also  \cite{Werbos.1988} \cite{Werbos.1990} \cite{Werbos.2016}.  The 1986 paper mentioned above was \cite{Rumelhart.1986}.

\subsubsection{Automatic differentiation}
\label{sc:autodiff-history}
%
% CMES style rewriting
The authors of
\cite{Goodfellow.2016}, p.~214, wrote of backprop as a particular case of automatic differentiation (AD):
\begin{quote}
	``The deep learning community has been somewhat isolated from the broader computer science community and has largely developed its own cultural attitudes concerning how to perform differentiation. More generally, the field of {\bf automatic differentiation} is concerned with how to compute derivatives algorithmically.  The back-propagation algorithm described here is only one approach to automatic differentiation.  It is a special case of a broader class of techniques called {\bf reverse mode accumulation}.''
\end{quote}
Let's decode what was said above.  The deep learning community was isolated because it was not in the mainstream of computer science research during the last AI winter, as Hinton described in an interview published in 2018 \cite{Ford.2018}:
\begin{quote}
	``This was at a time when all of us would have been a bit isolated in a fairly hostile environment---the environment for deep learning was fairly hostile until quite recently---it was very helpful to have this funding that allowed us to spend quite a lot of time with each other in small meetings, where we could really share unpublished ideas.''
\end{quote}
If Hinton did not move from Carnegie Mellon University in the US to the University of Toronto in Canada, it would be necessary for him to change research topic to get funding, AI winter would last longer, and he may not get the Turing Award along with LeCun and Bengio \cite{Metz.2019}:
\begin{quote}
	``The Turing Award, which was introduced in 1966, is often called the Nobel Prize of computing, and it includes a \$1 million prize, which the three scientists will share.''
\end{quote}

%{\color{red} HERE 2020.02.19}

A recent review of AD is given in \cite{Baydin.2018}, where backprop was described as a particular case of AD, known as ``reverse mode AD''; see also \cite{Schmidhuber.2015:rd0001}.\footnote{
	We only want to point out the connection between backprop and AD, together with a recent review paper on AD, but will not review AD itself here.
}

\subsection{Resurgence of AI and current state}
\label{sc:resurgence-2}
The success of deep neural networks in the ImageNet competitions since 2012, particularly when they surpassed human-level performance in 2015 (See Figure~\ref{fig:ImageNet-error} and Section~\ref{sc:parametric-ReLU} on Parametric ReLU), was preceded by their success in speech recognition, as recounted by Hinton in a 2018 interview \cite{Ford.2018}: 
\begin{quote}
	``For computer vision, 2012 was the inflection point. For speech, the inflection point was a few years earlier. Two different graduate students at Toronto showed in 2009 that you could make a better speech recognizer using deep learning. They went as interns to IBM and Microsoft, and a third student took their system to Google. The basic system that they had built was developed further, and over the next few years, all these companies' labs converted to doing speech recognition using neural nets. Many of the best people in speech recognition had switched to believing in neural networks before 2012, but the big public impact was in 2012, when the vision community, almost overnight, got turned on its head and this crazy approach turned out to win.''
\end{quote}
The mentioned 2009 breakthrough of applying deep learning to speech recognition did not receive much of the non-technical press as the 2012 breakthrough in computer vision (e.g., \cite{Economist.2016:rd0001} \cite{Economist.2016:rd0002}), and was thus not popularly known, except inside the deep-learning community.

Deep learning is being developed and used to guide consumers in nutrition \cite{Topol.2019:rd0001}:
\begin{quote}
	``Using machine learning, a subtype of artificial intelligence, the billions of data points were analyzed to see what drove the glucose response to specific foods for each individual. In that way, an algorithm was built without the biases of the scientists. 
	
	There are other efforts underway in the field as well. In some continuing nutrition studies, smartphone photos of participants' plates of food are being processed by deep learning, another subtype of A.I., to accurately determine what they are eating. This avoids the hassle of manually logging in the data and the use of unreliable food diaries (as long as participants remember to take the picture).
	
	But that is a single type of data. What we really need to do is pull in multiple types of data---activity, sleep, level of stress, medications, genome, microbiome and glucose---from multiple devices, like skin patches and smartwatches. With advanced algorithms, this is eminently doable. In the next few years, you could have a virtual health coach that is deep learning about your relevant health metrics and providing you with customized dietary recommendations.''
\end{quote}

% 2019.12.07
% commented this whole section out; we may not have time
% \section{Geopolitics of AI}

% \subsection{Race for dominance}

\begin{figure}[h]
	\centering
	\includegraphics[width=0.9\linewidth]{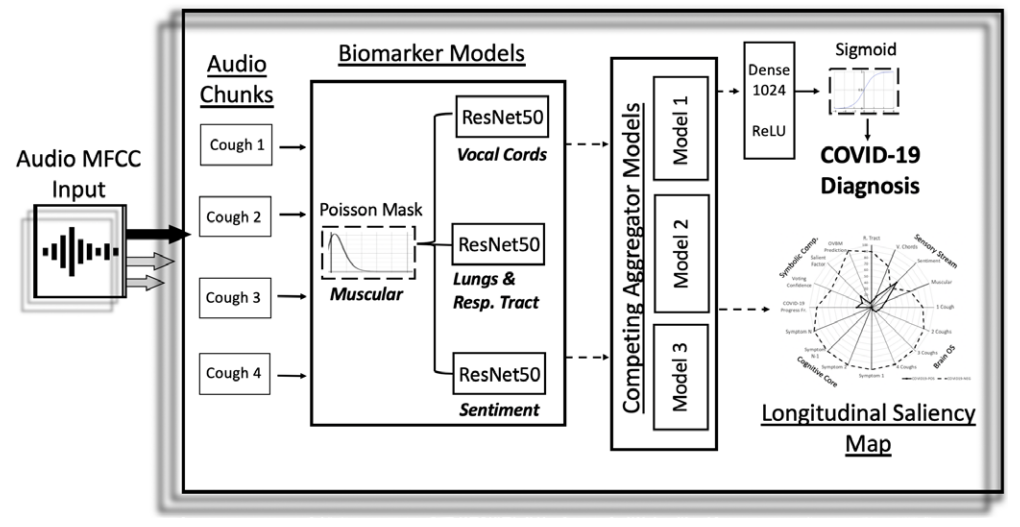}
	\caption{
		\emph{MIT COVID-19 diagnosis by cough recordings}. 
		Machine learning architecture.
		Audio Mel Frequency Cepstrum Coefficients (MFCC) as input.  
		Each cough signal is split into 6 audio chunks, processed by the MFCC package, then passed through the Biomarker 1 to check on muscular degradation.
		The output of Biomarker 1 is input into each of the three Convolutional Neural Networks (CNNs), representing Biomarker 2 (Vocal cords), Biomarker 3 (Lungs \& Respiratory Tract), Biomarker 4 (Sentiment).  The outputs of these CNNs are concatenated and ``pooled'' together to serve as (1) input for ``Competing Aggregator Models'' to produce a ``longitudinal saliency map'', and as (2) input for a deep and dense network with ReLU activation, followed by a ``binary dense layer'' with sigmoid activation to produce Covid-19 diagnosis.
		\cite{laguarta2020covid}.
		%\\
		{\footnotesize (\href{https://creativecommons.org/licenses/by/4.0/}{(CC By 4.0)}} 
	}
	\label{fig:Laguarta-COVID-19}
\end{figure}

%\subsubsection{COVID-19 detection from cough sound recordings}
\subsubsection{COVID-19 machine-learning diagnostics and prognostics}
\label{sc:COVID-19}
While it is not possible to review the vast number of papers on deep learning, it would be an important omission if we did not mention a most urgent issue of our times,\footnote{
	As of 2020.12.18, the COVID-19 pandemic was still raging across the entire United States.
} the COVID-19 (COronaVIrus Disease 2019) pandemic, and how deep learning could help in the diagnostics and prognostics of Covid-19.

{\bf Some reviews of Covid-19 models and software.}
The following sweeping assertion was made in a 2021 MIT Technology Review article titled ``Hundreds of AI tools have been built to catch covid. None of them helped''
\cite{heaven2021hundreds}, based on two 2021 papers that reviewed and appraised the validity and usefulness of Covid-19 models for diagnostics (i.e., detecting Covid-19 infection) and for prognostics (i.e., forecasting the course of Covid-19 in patients) \vphantom{\cite{wynants2021prediction}}\cite{wynants2021prediction} and \vphantom{\cite{roberts2021common}}\cite{roberts2021common}:
\begin{quote}
	``The clear consensus was that AI tools had made little, if any, impact in the fight against covid.''
\end{quote}

A large collection of 37,421 titles (published and preprint reports) on Covid-19 models up to July 2020 were examined in \cite{wynants2021prediction}, where only 169 studies describing 232 prediction models were selected based on CHARMS (CHecklist for critical Appraisal and data extraction for Systematic Reviews of prediction Modeling Studies) \vphantom{\cite{moons2014critical}}\cite{moons2014critical} for detailed analysis, with the risk of bias assessed using PROBAST (Pediction model Risk Of Bias ASsessment Tool)  \vphantom{\cite{wolff2019probast}}\cite{wolff2019probast}.   A follow-up study \cite{roberts2021common} examined 2,215 titles up to Oct 2020, using the same methodology as in \cite{wynants2021prediction} with the added requirement of ``sufficiently documented methodologies'', to narrow down to 62 titles for review ``in most details''.   
In the words of the lead developer of PROBAST, ``unfortunately'' journals outside the medical field were not included since it would be a ``surprise'' that ``the reporting and conduct of AI health models is better outside the medical literature''.\footnote{
	Private communication with Karel (Carl) Moons on 2021 Oct 28.  In other words, only medical journals included in PROBAST would report Covid-19 models that cannot be beaten by models reported in non-medical journals, such as in \cite{laguarta2020covid}, which was indeed not ``fit for clinical use'' to use the same phrase in \cite{heaven2021hundreds}.
}

%The study in \cite{wynants2021prediction} ended in July 2020, with a subsequent follow-up study in \cite{roberts2021common} that ended in Oct 2020.

{\bf Covid-19 diagnosis from cough recordings.}
%\citeauthor{laguarta2020covid} \citeyear{laguarta2020covid}
MIT researchers developed a cough-test smartphone app that diagnoses Covid-19 from cough recordings \vphantom{\cite{laguarta2020covid}}\cite{laguarta2020covid}, and claimed that their app achieved excellent results:\footnote{
	\label{fn:sensitivity-specificity}
	``In medical diagnosis, test sensitivity is the ability of a test to correctly identify those with the disease (true positive rate), whereas test specificity is the ability of the test to correctly identify those without the disease (true negative rate).'' See ``Sensitivity and specificity'',  \href{https://en.wikipedia.org/w/index.php?title=Sensitivity_and_specificity&oldid=1008200627}{Wikipedia version 02:21, 22 February 2021}.
	For the definition of ``AUC'' (Area Under the ROC Curve), with ``ROC'' abbreviating for ``Receiver Operating characteristic Curve'', see ``Classification: ROC Curve and AUC'', in ``Machine Learning Crash Course'', \href{https://developers.google.com/machine-learning/crash-course/classification/roc-and-auc}{Website}. \href{https://web.archive.org/web/20210129220610/https://developers.google.com/machine-learning/crash-course/classification/roc-and-auc}{Internet archive}.
}
\begin{quote}
	``When validated with subjects diagnosed using an official test, the model achieves COVID-19 sensitivity of 98.5\% with a specificity of 94.2\% (AUC: 0.97). For asymptomatic subjects it achieves sensitivity of 100\% with a specificity of 83.2\%.''
	\cite{laguarta2020covid}.
\end{quote}

\noindent
making one wondered why it had not been made available for use by everyone, since ``These inventions could help our coronavirus crisis now. But delays mean they may not be adopted until the worst of the pandemic is behind us'' \cite{matei2020app}.\footnote{
	One author of the present article (LVQ), more than one year after the preprint of \cite{laguarta2020covid}, still spit into a tube for Covid test instead of coughing into a phone.
}

Unfortunately, we suspected that the model in \cite{laguarta2020covid} was also ``not fit for clinical use'' as described in \cite{heaven2021hundreds}, because it has not been put to use in the real world as of 2022 Jan 12 (we were still spitting saliva into a tube instead of coughing into our phone).
In addition, despite being contacted three times regarding the lack of transparency in the description of the model in \cite{laguarta2020covid}, in particular the ``Competing Aggregator Models'' in Figure~\ref{fig:Laguarta-COVID-19}, the authors of \cite{laguarta2020covid} did not respond to our repeated inquiries, confirming the criticism described in Section~\ref{sc:irreproducibility} on  
``{Lack of transparency and irreproducibility of results}'' of AI models.

Our suspicion was confirmed when we found the critical review paper \vphantom{\cite{coppock2021covid}}\cite{coppock2021covid}, in which the pitfalls of the model in \cite{laguarta2020covid}, among other cough audio models, were pointed out, with the single most important question being: Were the audio representations in these machine-learning models, even though correlated with Covid-19 in their respective datasets, the true audio biomarkers originated from Covid-19? The seven grains of salt (pitfalls) listed in \cite{coppock2021covid} were:
\begin{enumerate}
	\item Machine-learning models did not detect Covid-19, but only distinguished between healthy people and sick people, a not so useful task.
	
	\item Surrounding acoustic environment may introduce biases into the cough sound recordings, e.g., Covid-19 positive people tend to stay indoors, and Covid-19 negative people outdoors.
	
	\item Participants providing coughs for the datasets may know their Covid-19 status, and that knowledge would affect their emotion, and hence the machine learning models.
	
	\item The machine-learning models can only be as accurate as the cough recording labels, which may not be valid since participants self reported their Covid-19 status.
	
	\item Most researchers, like the authors of \cite{laguarta2020covid}, don't share codes and datasets, or even information on their method as mentioned above; see also Section~\ref{sc:irreproducibility} ``Lack of transparency''.
%	, prompting the authors of \cite{coppock2021covid} to question: ``How can classification metrics on such a sensitive topic be taken seriously if they cannot be cross checked by other research teams?''
	
	\item The influence of factors such as comorbidity, ethnicity, geography, socio-economics, on Covid-19 is complex and unequal, and could introduce biases in the datasets.
	
	\item Lack of population control (participant identity not recorded) led to non-disjoint training set, development set, and test set. 
	
\end{enumerate}

%\cite{aly2021pay}: more Covid-19 by coughs
%
%\cite{mohammed2021ensemble}: more Covid-19 by coughs
%
%\cite{manshouri2021identifying}: more Covid-19 by coughs
%
%\cite{coppock2021covid}: This article is a critique of the cough recordings approach
%
%{\color{red} HERE 2020.12.28.  add some figures from \cite{laguarta2020covid} with \href{https://creativecommons.org/licenses/by/4.0/}{Creative Commons Attribution 4.0 (CC By 4.0) License}.}

{\bf Other Covid-19 machine-learning models.}
%CMES 2020-2021 papers on Covid-19:
%%
%%\verb*|CMES-2020-21-papers-Covid19.bib|
%%
%
%\verb|CMES-2020-21-papers-Covid19.bib|
%
%
%\cite{WOS:000717043900001}: A Survey on Machine Learning in COVID-19 Diagnosis 
%=============
%First, the procedure of the diagnosis based on machine learning is introduced in detail, which includes medical data collection, image preprocessing, feature extraction, and image classification. Then, we review seven methods in detail: transfer learning, ensemble learning, unsupervised learning and semi-supervised learning, convolutional neural networks, graph neural networks, explainable deep neural networks, and so on. What's more, the advantages and limitations of different diagnosis methods are compared. Although the great achievements in medical images classification in recent years, Corona Virus Disease 2019 images classification based on machine learning still encountered many problems. For example, the highly unbalanced dataset, the difficulty of collecting labeled data, and the poor quality of the data. Aiming at these problems, we propose some solutions and provide a comprehensive presentation for future research.
%=============
A comprehensive review of machine learning for Covid-19 diagnosis based on medical-data collection, preprocessing of medical images, whose features are extracted, and classified is provided in \cite{WOS:000717043900001}, where methods based on cough sound recordings were not included.  Seven methods were reviewed in detail: (1) transfer learning, (2) ensemble learning, (3) unsupervised learning and (4) semi-supervised learning, (5) convolutional neural networks, (6) graph neural networks, (7) explainable deep neural networks.

In \cite{WOS:000688417500004}, deep-learning methods together with transfer learning were reviewed for classification and detection of 
Covid-19 based on chest X-ray, computer-tomography (CT) images, and lung-ultrasound images.  Also reviewed were machine-learning methods for selection of vaccine candidates, natural-language-processing methods to analyze public sentiment during the pandemic.

%\cite{WOS:000684932000005}
%Title = {{Multi-Disease Prediction Based on Deep Learning: A Survey}}
%Abstract = {{In recent years, the development of artificial intelligence (AI) and the gradual beginning of AI's research in the medical field have allowed people to see the excellent prospects of the integration of AI and healthcare. Among them, the hot deep learning field has shown greater potential in applications such as disease prediction and drug response prediction. From the initial logistic regression model to the machine learning model, and then to the deep learning model today, the accuracy of medical disease prediction has been continuously improved, and the performance in all aspects has also been significantly improved. This article introduces some basic deep learning frameworks and some common diseases, and summarizes the deep learning prediction methods corresponding to different diseases. Point out a series of problems in the current disease prediction, and make a prospect for the future development. It aims to clarify the effectiveness of deep learning in disease prediction, and demonstrates the high correlation between deep learning and the medical field in future development. The unique feature extraction methods of deep learning methods can still play an important role in future medical research.}}
For multi-disease (including Covid-19) prediction, methods based on (1) logistic regression, (2) machine learning, and in particular (3) deep learning were reviewed, with difficulties encountered forming a basis for future developments pointed out,  in \cite{WOS:000684932000005}.

%\cite{WOS:000688417500002}: Predicting Genotype Information Related to COVID-19 for Molecular Mechanism Based on Computational Methods
Information on the collection of genes, called genotype, related to Covid-19, was predicted by searching and scoring similarities between the seed genes (obtained from prior knowledge) and candidate genes (obtained from the biomedical literature) with the goal to establish the molecular mechanism of Covid-19 \cite{WOS:000688417500002}. 

In \cite{WOS:000602627600001}, the proteins associated with Covid-19 were predicted using ligand\footnote{
	A ligand is ``usually a molecule which produces a signal by binding to a site on a target protein,' see ``Ligand (biochemistry)'', Wikipedia, \href{https://en.wikipedia.org/w/index.php?title=Ligand_(biochemistry)&oldid=1059255240}{version 11:08, 8 December 2021}. 
} designing and modecular modeling.

%\cite{WOS:000603295200020}: Estimating the Impact of COVID-19 Pandemic on the Research Community in Saudi Arabia. 
In \cite{WOS:000603295200020},  
after evaluating various computer-science techniques using Fuzzy-Analytic Hierarchy Process integrated with the Technique for Order Performance by Similar to Ideal Solution, it was recommended to use Blockchain as the most effective technique to be used by healthcare workers to address Covid-19 problems in Saudi Arabia.

Other Covid-19 machine learning models include
the use of regression algorithms for real-time analysis of Covid-19 pandemic \cite{WOS:000602627600003},
forecasting the number of infected people using the logistic growth curve and the Gompertz growth curve \cite{WOS:000602627600002},
a generalization of the SEIR\footnote{
	SEIR = Susceptible, Exposed, Infectious, Recovered; see ``Compartmental models in epidemiology'', Wikipedia, \href{https://en.wikipedia.org/w/index.php?title=Compartmental_models_in_epidemiology&oldid=1072808123}{version 15:44, 19 February 2022}.
} model and logistic regression for forecasting \cite{WOS:000602627600005}.

\subsubsection{Additional applications of deep learning}
\label{sc:deep-learning-appliations}
%
%CMES 2020-2021 papers on (deep OR machine) learning:
%
%\verb|CMES-2020-21-papers-deep-machine-learning.bib|
%
%{\bf Review paper, deep learning}

The use of deep learning as one of several machine-learning techniques for Covid-19 diagnosis was reviewed in \cite{WOS:000717043900001} \cite{WOS:000688417500004} \cite{WOS:000684932000005}, as mentioned above.

% paper 7
% review paper
% deep learning
% Parkinson's disease diagnosis
% CITED
%
%@article{ WOS:000513774700010,
%Author = {Akyol, Kemal},
%Title = {{Growing and Pruning Based Deep Neural Networks Modeling for Effective Parkinson's Disease Diagnosis}}
%Abstract = {{Parkinson's disease is a serious disease that causes death. Recently, a new dataset has been introduced on this disease. The aim of this study is to improve the predictive performance of the model designed for Parkinson's disease diagnosis. By and large, original DNN models were designed by using specific or random number of neurons and layers. This study analyzed the effects of parameters, i.e., neuron number and activation function on the model performance based on growing and pruning approach. In other words, this study addressed the optimum hidden layer and neuron numbers and ideal activation and optimization functions in order to find out the best Deep Neural Networks model. In this context of this study, several models were designed and evaluated.  The overall results revealed that the Deep Neural Networks were significantly successful with 99.34\% accuracy value on test data. Also, it presents the highest prediction performance reported so far.  Therefore, this study presents a model promising with respect to more accurate Parkinson's disease diagnosis.}}
By growing and pruning deep learning neural networks (DNNs), optimal parameters, such as number of hidden layers, number of neurons, and types of activation functions, were obtained for the diagnosis of Parkinson's disease, with 99.34\% accuracy on test data, compared to previous DNNs using specific or random number of hidden layers and neurons
\cite{WOS:000513774700010}.

In \cite{WOS:000576462800004}, a deep residual network, with gridded interpolation and Swish activation function (see Section~\ref{sc:new-active-functions}), was constructed to generate a single high-resolution image from many low-resolution images obtained from Fundus Fluorescein Angiography (FFA),\footnote{
	Fundus is ``the interior surface of the eye opposite the lens and includes the retina, optic disc, macula, fovea, and posterior pole'' (Wikipedia, \href{https://en.wikipedia.org/w/index.php?title=Fundus_(eye)&oldid=934540306}{version 02:49, 7 January 2020}).
	Fluorescein is an organic compound and fluorescent dye (Wikipedia, \href{https://en.wikipedia.org/w/index.php?title=Fluorescein&oldid=1064139571}{version 19:51, 6 January 2022}).
	Angiography (angio- ``blood vessel'' + graphy ``write, record'', Wikipedia, \href{https://en.wikipedia.org/w/index.php?title=Angiography&oldid=1069444815}{version 10:19, 2 February 2022}) is a medical procedure to visualize the flow of blood (or other biological fluid) by injecting a dye and by using a special camera.  
} resulting in ``superior performance metrics and computational time.''

% paper 9
% model order reduction, reduced order 
% machine learning, finite element
% CITED
%
%@article{ WOS:000696630500001,
%Author = {Lu, Ye and Li, Hengyang and Saha, Sourav and Mojumder, Satyajit and Al Amin, Abdullah and Suarez, Derick and Liu, Yingjian and Qian, Dong and Liu, Wing Kam},
%Title = {{Reduced Order Machine Learning Finite Element Methods: Concept, Implementation, and Future Applications}},
%Abstract = {{This paper presents the concept of reduced order machine learning finite element (FE) method. In particular, we propose an example of such method, the proper generalized decomposition (PGD) reduced hierarchical deeplearning neural networks (HiDeNN), called HiDeNN-PGD. We described first the HiDeNN interface seamlessly with the current commercial and open source FE codes. The proposed reduced order method can reduce significantly the degrees of freedom for machine learning and physics based modeling and is able to deal with high dimensional problems. This method is found more accurate than conventional finite element methods with a small portion of degrees of freedom. Different potential applications of the method, including topology optimization, multi-scale and multi-physics material modeling, and additive manufacturing, will be discussed in the paper.}}
%

%\noindent
%Erroneous CMES BibTeX entry: \cite{cmes.2021.016784.OLD}
%
%\noindent
%Corrected BibTeX entry: \cite{cmes.2021.016784}

%
Going beyond the use of Proper Orthogonal Decomposition and Generalized Falk Method \cite{cmes.2021.016784}, a hierarchichal deep-learning neural network was proposed in \cite{WOS:000696630500001} to be used with the Proper Generalized Decomposition as a model-order-reduction method applied to finite element models.

To develop high precision model to forecast wind speed and wind power, which depend on the conditions of the nearby ``atmospheric pressure, temperature, roughness, and obstacles'',
%
% CMES style rewriting
the authors of
\vphantom{\cite{Deng.2020}}\cite{Deng.2020} applied 
``deep learning, reinforcement learning and transfer learning.''  The challenges in this area are the randomness, the instantaneity, and the seasonal characteristics of wind and the atmosphere.

Self-driving cars must deal with a large variety of real scenarios and of real behaviors, which deep-learning perception-action models should learn to become robust.  But due to a limition of the data, it was proposed in \cite{WOS:000513774700009} to use a new image style transfer method to generate more varieties in data by modifying texture, contrast ratio and image color, and then extended to scenarios that were unobserved before.

Other applications of deep learning include 
a real-time maskless-face detector using deep residual networks \cite{WOS:000642079800003},
topology optimization with embedded physical law and physical constraints
\cite{WOS:000684807600001},
prediction of stress-strain relations in granular materials from triaxial test results
\cite{WOS:000672695700008},
surrogate model for flight-load analysis
\cite{WOS:000684932000010},
% citation no.22
classification of domestic refuse in medical institutions based on transfer learning and convolutional neural network
\cite{WOS:000642079800011},
% citation no.23
convolutional neural network for arrhythmia diagnosis
\cite{WOS:000573973000005},
% citation no.24
e-commerce dynamic pricing by deep reinforcement learning
\cite{WOS:000642088300016},
% citation no.25
network intrusion detection
\cite{WOS:000684807600019},
% citation no.26
road pavement distress detection for smart maintenance
\cite{WOS:000654350400007},
% citation no.27
traffic flow statistics
\cite{WOS:000706701500003},
% citation no.28
multi-view gait recognition using deep CNN and channel attention mechanism
\cite{WOS:000573973000004},
% citation no.29
mortality risk assessment of ICU patients
\cite{WOS:000672695700010},
% citation no.30
stereo matching method based on space-aware network model to reduce the limitation of GPU RAM
\cite{WOS:000642088300010},
%
%2022.06.13 - HERE, 5 more CMES papers - 
% citation no.31
air quality forecasting in Internet of Things
\cite{WOS:000579646600019},
% citation no.32
analysis of cardiac disease abnormal ECG signals
\cite{WOS:000684807600003},
% citation no.33
detection of mechanical parts (nuts, bolts, gaskets, etc.) by machine vision
\cite{WOS:000684932000019},
% citation no.34
asphalt road crack detection 
\cite{WOS:000684932000007},
% citation no.35
steel commondity selection using bidirectional encoder representations from transformers (BERT)
\cite{WOS:000688417500003},
% citation no.36 (actually 34)
short-term traffic flow prediction using LSTM-XGBoost combination model
\cite{WOS:000576462800002},
% citation no.38 (actually 35)
emotion analysis based on multi-channel CNN in social networks
\cite{WOS:000573973000002}.

\begin{figure}[h]
	\centering
	\includegraphics[width=0.49\linewidth]{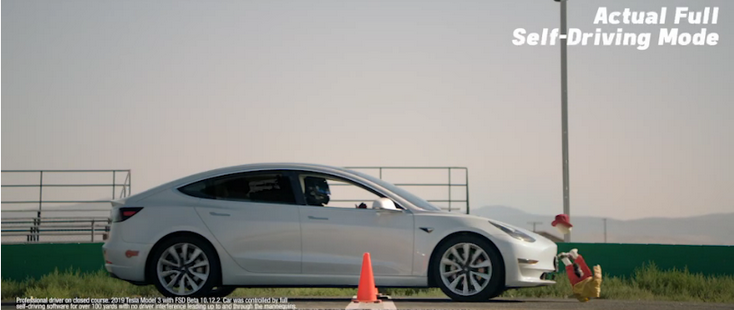}
	\includegraphics[width=0.49\linewidth]{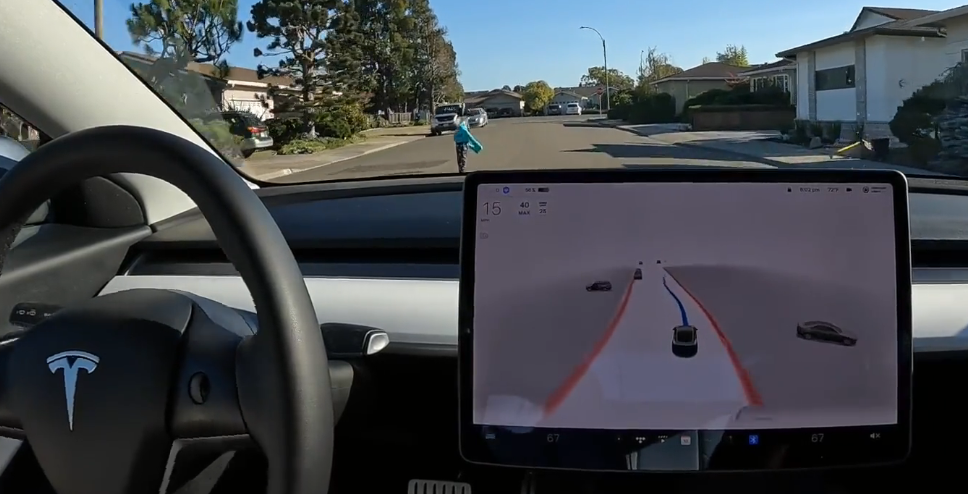}
	\caption{
		\emph{Tesla Full-Self-Driving (FSD) controversy} (Section~\ref{sc:driveless-cars}).    
		\emph{Left:}
		Tesla in FSD mode hit a child-size mannequin, repeatedly in safety tests by The Dawn Project, a software competitor to Tesla, 2022.08.09
		\cite{DawnProjectTeslaFSD} \cite{helmore2022tesla}.
		\emph{Right:} 
		Tesla in FSD mode went around a child-size mannequin at 15 mph in a residential area, 2022.08.14 \cite{TeslaFSDkids} \cite{roth2022youtube}.   
		Would a prudent driver stop completely, waiting for the kid to move out of the road, before proceeding forward?  The driver, a Tesla investor, did not use his own child, indicating that his maneuver was not safe.
		%\\
%		{\footnotesize (Data and video provided by QuantivRisk.)} 
	}
	\label{fig:tesla-hit-kid}
\end{figure}

\begin{figure}[h]
	\centering
	\includegraphics[width=0.90\linewidth]{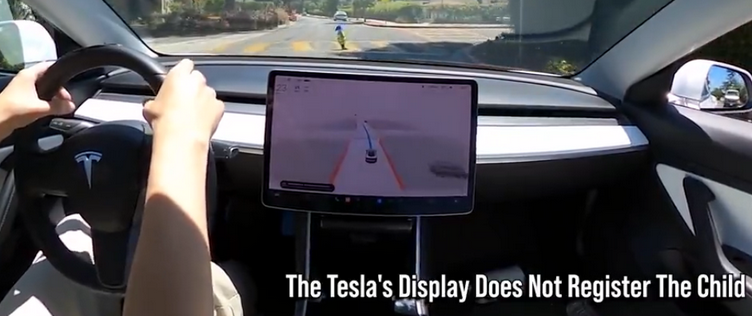}
	\caption{
		\emph{Tesla Full-Self-Driving (FSD) controversy} (Section~\ref{sc:driveless-cars}). 
		The Tesla was about to run down the child-size mannequin at 23 mph, hitting it at 24 mph. The driver did not hold on, but only kept his hands close, to the driving wheel for safety, and did not put his foot on the accelerator.  There were no cones on both sides of the road, and there was room to go around the mannequin.  The weather was clear, sunny. The mannequin wore a bright safety jacket.  Visibility was excellent, 2022.08.15  
%		The driver did not put his foot on the accelerator
		\cite{Tesla-FSD-hit-kid}
		\cite{hawkins2022tesla}.    
		%\\
		%		{\footnotesize (Data and video provided by QuantivRisk.)} 
	}
	\label{fig:tesla-hit-kid-2}
\end{figure}

\begin{figure}[h]
	\centering
	\includegraphics[width=0.49\linewidth]{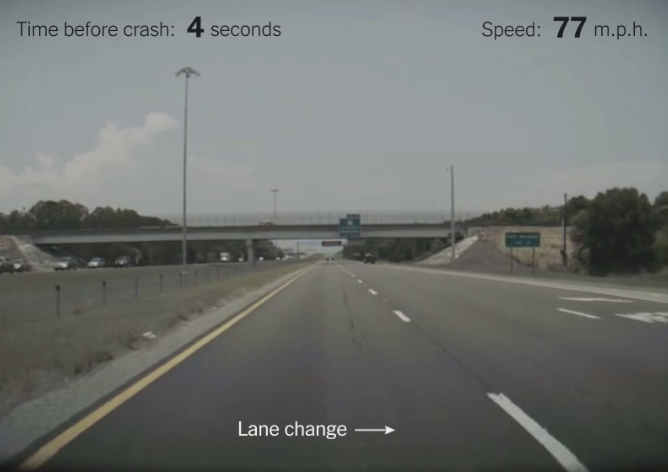}
	\includegraphics[width=0.49\linewidth]{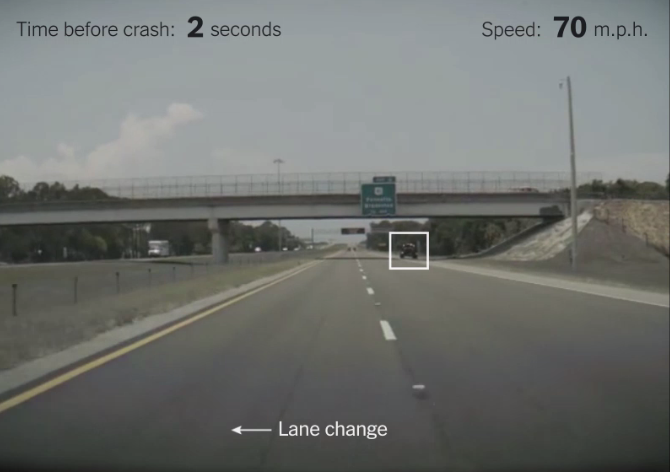}
	\caption{
		\emph{Tesla crash} (Section~\ref{sc:driveless-cars}).  
		July 2020.  
		\emph{Left:}
		``Less than a half-second after [the Tesla driver] flipped on her turn signal, Autopilot started moving the car into the right lane and gradually slowed, video and sensor data showed.'' 
		\emph{Right:} 
		``Halfway through, the Tesla sensed an obstruction---possibly a truck stopped on the side of the road---and paused its lane change. The car then veered left and decelerated rapidly'' 
		\cite{metz2022can}.
		See also Figures~\ref{fig:tesla-crash-2}, \ref{fig:tesla-crash-3}, \ref{fig:tesla-crash-4}.
		%\\
		{\footnotesize (Data and video provided by QuantivRisk.)} 
	}
	\label{fig:tesla-crash-1}
\end{figure}

\begin{figure}[h]
	\centering
	\includegraphics[width=0.49\linewidth]{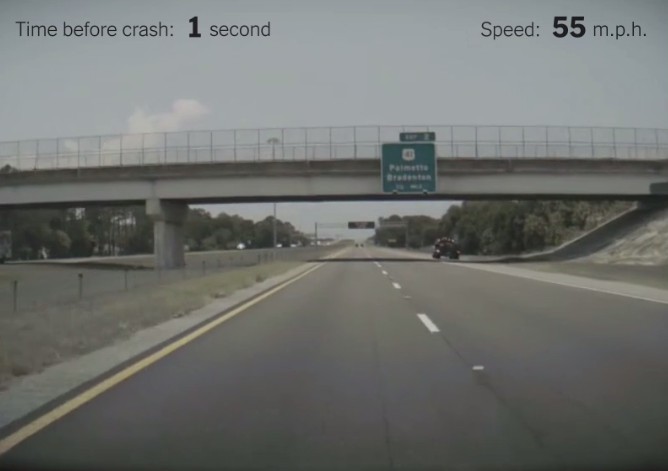}
	\includegraphics[width=0.49\linewidth]{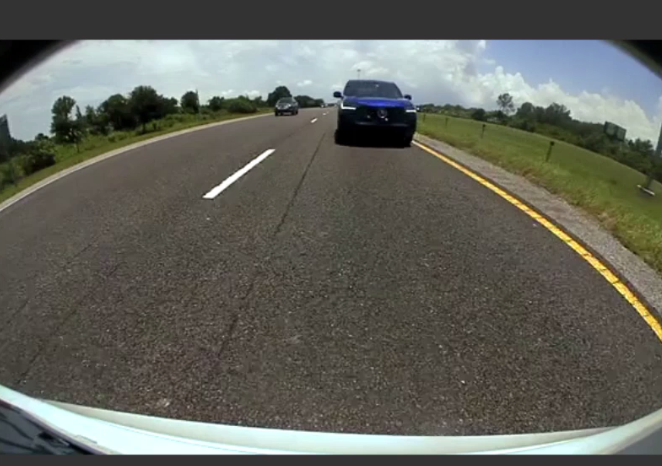}
	\caption{
		\emph{Tesla crash} (Section~\ref{sc:driveless-cars}).
		July 2020.
		``Less than a second after the Tesla has slowed to roughly 55 m.p.h. [\emph{Left}], its rear camera shows a car rapidly approaching [\emph{Right}]''  
		\cite{metz2022can}.
		There were no moving cars on both lanes in front of the Tesla for a long distance ahead (perhaps a quarter of a mile).
		See also Figures~\ref{fig:tesla-crash-1}, \ref{fig:tesla-crash-3}, \ref{fig:tesla-crash-4}.
		%\\
		{\footnotesize (Data and video provided by QuantivRisk.)} 
	}
	\label{fig:tesla-crash-2}
\end{figure}

\begin{figure}[h]
	\centering
	\includegraphics[width=0.49\linewidth]{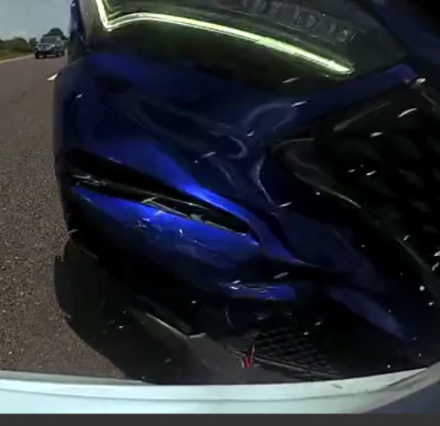}
	\includegraphics[width=0.49\linewidth]{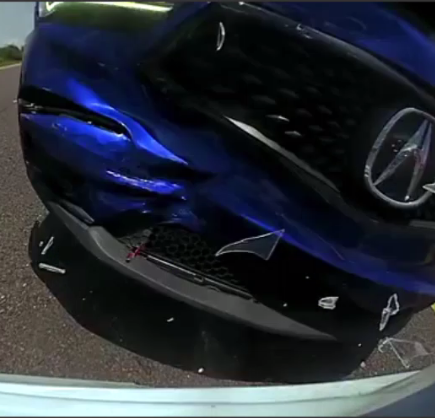}
	\caption{
		\emph{Tesla crash} (Section~\ref{sc:driveless-cars}). 
		July 2020.
		The fast-coming blue car rear-ended the Tesla, indented its own front bumper, with
		flying broken glass (or clear plastic) cover shards captured by the Tesla rear camera  
		\cite{metz2022can}.
		See also Figures~\ref{fig:tesla-crash-1}, \ref{fig:tesla-crash-2}, \ref{fig:tesla-crash-4}.
		%\\
		{\footnotesize (Data and video provided by QuantivRisk.)} 
	}
	\label{fig:tesla-crash-3}
\end{figure}

\begin{figure}[h]
	\centering
	\includegraphics[width=0.49\linewidth]{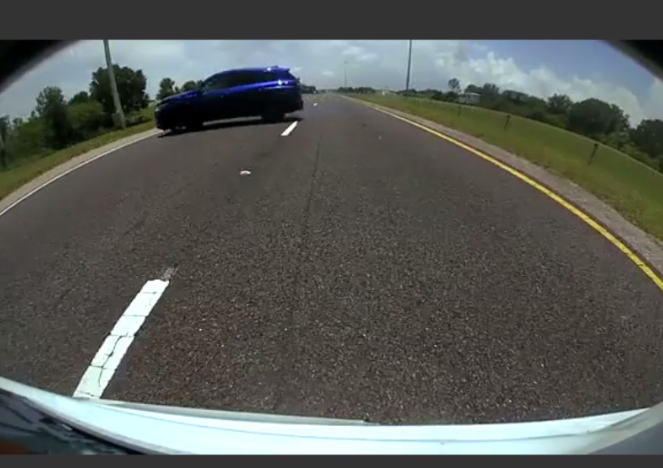}
	\includegraphics[width=0.49\linewidth]{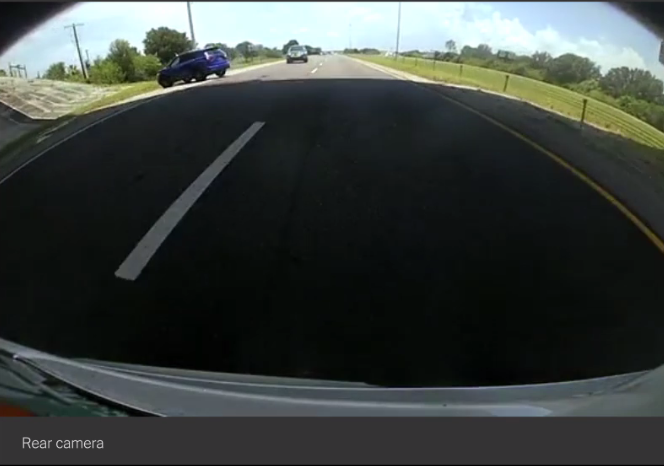}
	\caption{
		\emph{Tesla crash} (Section~\ref{sc:driveless-cars}).  After hitting the Tesla, the blue car ``spun across the highway [\emph{Left}] and onto the far shoulder [\emph{Right}],'' as another car was coming toward on the right lane (left in photo), but still at a safe distance so not to hit it.  
		\cite{metz2022can}.  
		See also Figures~\ref{fig:tesla-crash-1}, \ref{fig:tesla-crash-2}, \ref{fig:tesla-crash-3}.
		%\\
		{\footnotesize (Data and video provided by QuantivRisk.)} 
	}
	\label{fig:tesla-crash-4}
\end{figure}

\section{Closure: Limitations and danger of AI}
\label{sc:closure}
A goal of the present review paper is to bring first-time learners from the beginning level to as close as possible the research frontier in deep learning, with particular connection to, and application in, computational mechanics.

As concluding remarks, we collect here some known limitations and danger of AI in general, and deep learning in particular.  As Hinton pointed out himself a limitation of generalization of deep learning \cite{Metz.2017}:
\begin{quote}
	``If a neural network is trained on images that show a coffee cup only from a side, for example, it is unlikely to recognize a coffee cup turned upside down.''
\end{quote}

\subsection{Driverless cars, crewless ships, ``not any time soon''}
\label{sc:driveless-cars}

In 2016, the former U.S. secretary of transportation, Anthony Foxx, described a rosy future just five years down the road:
``By 2021, we will see autonomous vehicles in operation across the country in ways that we [only] imagine today'' \cite{dujmovic2021you}.

On 2022.06.09, UPI reported that ``Automaker Hyundai and South Korean officials launched a trial service of self-driving taxis in the busy Seoul neighborhood of Gangnam,'' an event described as ``the latest step forward in the country's efforts to make autonomous vehicles an everyday reality.  The new service, called RoboRide, features Hyundai Ioniq 5 electric cars equipped with Level 4 autonomous driving capabilities. The technology allows the taxis to move independently in real-life traffic without the need for human control, although a safety driver will remain in the car'' \cite{maresca2022huyndai}.   According to Huyndai, the safety driver ``only intervenes under limited conditions,'' which were explicitly not specified to the public, whereas the car itself would ``perceive, make decisions and control its own driving status.'' 

But what is ``Level 4 autonomous driving''?  Let's look at the startup autononous-driving company Waymo.  Their Level 4 consists of ``mapping the territory in a granular fashion (including lane markers, traffic signs and lights, curbs, and crosswalks). The solution incorporates both GPS signals and real-time sensor data to always determine the vehicle's exact location. Further, the system relies on more than 20 million miles of real-world driving and more than 20 billion miles in simulation, to allow the Waymo Driver to anticipate what other road users, pedestrians, or other objects might do''  \cite{kirkpatrick2022still}.

Yet Level 4 is still far from Level 5, for which ``vehicles are fully automated with no need for the driver to do anything but set the destination and ride along. They can drive themselves anywhere under any conditions, safely'' \cite{bogna2022is}, and would still be many years later away \cite{kirkpatrick2022still}. 

Indeed, exactly two months after Huyndai's announcement of their Level 4 test pilot program, 
on 2022.08.09, \emph{The Guardian} reported that in a series of safety tests, a
``professional test driver using Tesla's Full Self-Driving mode repeatedly hit a child-sized mannequin in its path'' \cite{helmore2022tesla}; Figure~\ref{fig:tesla-hit-kid}, left.  ``It's a lethal threat to all Americans,  putting children at great risk in communities across the country,'' warned The Dawn Project's founder, Dan O'Dowd, who described the test results as ``deeply disturbing,'' as the vehicle tended to ``mow down children at crossroads,'' and who argued for prohibiting Tesla vehicles from running in the street until Tesla self driving software could be proven safe.  

The Dawn Project test results were contested by a Tesla investor, who posted a video on 2022.08.14 to prove that the Tesla Full-Self-Driving (FSD) system worked as advertized (Figure~\ref{fig:tesla-hit-kid}, right).  The next day, 2022.08.15, Dan O'Dowd posted a video proving that the Tesla under FSD mode ran over a child-size mannequin at 24 mph in clear weather, with excellent visibility, no cones on either side of the Tesla, and without the driver pressing his foot on the accelerator (Figure~\ref{fig:tesla-hit-kid-2}).
%See Figures~\ref{fig:tesla-hit-kid}, \ref{fig:tesla-hit-kid-2} and the references in the captions regarding a controversy surrounding these test results.

``In June [2022], the National Highway Traffic Safety Administration (NHTSA), said it was expanding an investigation into 830,000 Tesla cars across all four current model lines. The expansion came after analysis of a number of accidents revealed patterns in the car's performance and driver behavior'' \cite{helmore2022tesla}. ``Since 2016, the agency has investigated 30 crashes involving Teslas equipped with automated driving systems, 19 of them fatal. NHTSA's Office of Defects Investigation is also looking at the company's autopilot technology in at least 11 crashes where Teslas hit emergency vehicles.''

In 2019, it was reported that several car executives thought that driveless cars were still several years in the future because of the difficulty in anticipating human behavior \cite{Boudette.2019}.
%\begin{quote}
%	``Ford and other companies say the industry overestimated the arrival of autonomous vehicles, which still struggle to anticipate what other drivers and pedestrians will do.
%	
%	\hspace{5mm}
%	A year ago, Detroit and Silicon Valley had visions of putting thousands of self-driving taxis on the road in 2019, ushering in an age of driverless cars.
%	
%	\hspace{5mm}
%	Most of those cars have yet to arrive---and it is likely to be years before they do. Several carmakers and technology companies have concluded that making autonomous vehicles is going to be harder, slower and costlier than they thought.
%	
%	\hspace{5mm}
%	`We overestimated the arrival of autonomous vehicles,' Ford's chief executive, Jim Hackett, said at the Detroit Economic Club in April [2019].
%	
%	\hspace{5mm}
%	The two automakers plan to use autonomous-vehicle technology from a Pittsburgh start-up, Argo AI, in ride-sharing services in a few urban zones as early as 2021. But Argo's chief executive, Bryan Salesky, said the industry's bigger promise of creating driverless cars that could go anywhere was ``way in the future.''
%	
%	\hspace{5mm}
%	He and others attribute the delay to something as obvious as it is stubborn: human behavior.''
%\end{quote}
The progress of Huyndai's driveless taxis has not solved the challenge of dealing with human behavior, as there was still a need for a ``safety driver.''  
%But the goal of having a fully driveless car is getting closer to reality.

``On [2022] May 6, Lyft, the ride-sharing service that competes with Uber  sold its Level 5 division, an autonomous-vehicle unit, to Woven Planet, a Toyota subsidiary. After four years of research and development, the company seems to realize that autonomous driving is a tough nut to crack---much tougher than the team had anticipated.

``Uber came to the same conclusion, but even earlier, in December. The company sold Advanced Technologies Group, its self-driving unit, to Aurora Innovation, citing high costs and more than 30 crashes, culminating in a fatality as the reason for cutting its losses.

``Finally, several smaller companies, including Zoox, a robo-taxi company; Ike, an autonomous-trucking startup; and Voyage, a self-driving startup; have also passed the torch to companies with bigger budgets'' \cite{dujmovic2021you}.

``Those startups, like many in the industry, have underestimated the sheer difficulty of ``leveling up'' vehicle autonomy to the fabled Level 5 (full driving automation, no human required)'' \cite{dujmovic2021you}.

On top of the difficulty in addressing human behavior, there were other problems, perhaps in principle less challenging, so we thought, as reported in \cite{kirkpatrick2022still}:
``widespread adoption of autonomous driving is still years away from becoming a reality, largely due to the challenges involved with the development of accurate sensors and cameras, as well as the refinement of algorithms that act upon the data captured by these sensors.

``This process is extremely data-intensive, given the large variety of potential objects that could be encountered, as well as the near-infinite ways objects can move or react to stimuli (for example, road signs may not be accurately identified due to lighting conditions, glare, or shadows, and animals and people do not all respond the same way when a car is hurtling toward them).

``Algorithms in use still have difficulty identifying objects in real-world scenarios; in one accident involving a Tesla Model X, the vehicle's sensing cameras failed to identify a truck's white side against a brightly lit sky.''

In addition to Figure~\ref{fig:tesla-hit-kid}, another example was a Tesla crash in July 2020 in clear, sunny weather, with little clouds, as shown in Figures~\ref{fig:tesla-crash-1}, \ref{fig:tesla-crash-2}, \ref{fig:tesla-crash-3}, \ref{fig:tesla-crash-4}.  The self-driving system could not detect that a static truck was parked on the side of a highway, and due to the foward and changing-lane motion of the Tesla, the software could have thought that it was running into the truck, and veered left while rapidly decelerating to avoid collision with the truck.  As a result, the Tesla was rear-ended by another fast coming car from behind on its left side \cite{metz2022can}.

``Pony.ai is the latest autonomous car company to make headlines for the wrong reasons. It has just lost its permit to test its fleet of autonomous vehicles in California over concerns about the driving record of the safety drivers it employs.  It's a big blow for the company, and highlights the interesting spot the autonomous car industry is in right now. After a few years of very bad publicity, a number of companies have made real progress in getting self-driving cars on the road'' \cite{guinness2022what}.

The 2022 article ``I'm the Operator': The Aftermath of a Self-Driving Tragedy'' \cite{smiley2022Im} described these ``few years of very bad publicity'' in stunning, tragic details about an Uber autonomous-vehicle operator, Rafela Vasquez, who did not take over the control of the vehicle in time, and killed a jaywalking pedestrian.

The classification software of the Uber autonomous driving system could not recognize the pedestrian, but vacillated between a ``vehicle'', then ``other'', then a ``bicycle'' \cite{smiley2022Im}.

``At 2.6 seconds from the object, the system identified it as `bicycle.' At 1.5 seconds, it switched back to considering it `other.' Then back to `bicycle' again. The system generated a plan to try to steer around whatever it was, but decided it couldn't. Then, at 0.2 seconds to impact, the car let out a sound to alert Vasquez that the vehicle was going to slow down. At two-hundredths of a second before impact, traveling at 39 mph, Vasquez grabbed the steering wheel, which wrested the car out of autonomy and into manual mode.  It was too late. The smashed bike scraped a 25-foot wake on the pavement. A person lay crumpled in the road'' \cite{smiley2022Im}.

The operator training program manager said ``I felt shame when I heard of a lone frontline employee has been singled out to be charged of negligent homicide with a dangerous instrument.  We owed Rafaela better oversight and support. We also put her in a tough position.'' Another program manager said ``You can't put the blame on just that one person. I mean, it's absurd. Uber had to know this would happen. We get distracted in \emph{regular} driving.  It's not like somebody got into their car and decided to run into someone.  They were working within a framework. And that framework created the conditions that allowed that to happen.'' \cite{smiley2022Im}.

After the above-mentioned fatality caused by an Uber autonomous car with a single operator in it, ``many companies temporarily took their cars off the road, and after it was revealed that only one technician was inside the Uber car, most companies resolved to keep two people in their test vehicles at all times'' \cite{metz2022this}.  Having two operators in a car would help to avoid accidents, but the pandemic social-distancing rule often prevented such arrangement from happening.

``Many self-driving car companies have no revenue, and the operating costs are unusually high. Autonomous vehicle start-ups spend \$1.6 million a month on average---four times the rate at financial tech or health care companies'' \cite{metz2022this}. 

``Companies like Uber and Lyft, worried about blowing through their cash in pursuit of autonomous technology, have tapped out. Only the deepest-pocketed outfits like Waymo, which is a subsidiary of Google's parent company, Alphabet; auto giants; and a handful of start-ups are managing to stay in the game.

``Late last month, Lyft sold its autonomous vehicle unit to a Toyota subsidiary, Woven Planet, in a deal valued at \$550 million. Uber offloaded its autonomous vehicle unit to another competitor in December. And three prominent self-driving start-ups have sold themselves to companies with much bigger budgets over the past year'' \cite{metz2022the}.

%\cite{bogna2022is}
%\cite{metz2022this}
%\cite{metz2022the}

Similar problems exist with building autonomous boats to ply the oceans without a need for a crew on board \cite{nims2020robot}:
\begin{quote}
	``When compared with autonomous cars, ships have the advantage of not having to make split-second decisions in order to avoid catastrophe. The open ocean is also free of jaywalking pedestrians, stoplights and lane boundaries. That said, robot ships share some of the problems that have bedeviled autonomous vehicles on land, namely, that they're bad at anticipating what humans will do, and have limited ability to communicate with them.  
%	Technically, it's not possible yet to make an autonomous ship that operates safely and efficiently in crowded areas and in port areas.''
\end{quote}

Shipping is a dangerous profession, as there were some 41 large ships lost at sea due to fires, rogue waves, or other accidents, in 2019 alone.  But before an autonomous ship can reach the ocean, it must get out of port, and that remains a technical hurdle not yet overcome:
\begin{quote}
	`` 'Technically, it's not possible yet to make an autonomous ship that operates safely and efficiently in crowded areas and in port areas,' says Rudy Negenborn, a professor at TU Delft who researches and designs systems for autonomous shipping.
	
	Makers of autonomous ships handle these problems by giving humans remote control. But what happens when the connection is lost? Satisfactory solutions to these problems have yet to arrive, adds Dr. Negenborn.''
\end{quote}

The onboard deep-learning computer vision system was trained to recognize ``kayaks, canoes, Sea-Doos'', but a person standing on a paddle board would look like someone walking on water to the system \cite{nims2020robot}.  See also Figures~\ref{fig:tesla-crash-1}, \ref{fig:tesla-crash-2}, \ref{fig:tesla-crash-3}, \ref{fig:tesla-crash-4} on the failure of the Tesla computer vision system in detecting a parked truck on the side of a highway. 
 
\begin{figure}[h]
	\centering
	\includegraphics[width=0.80\linewidth]{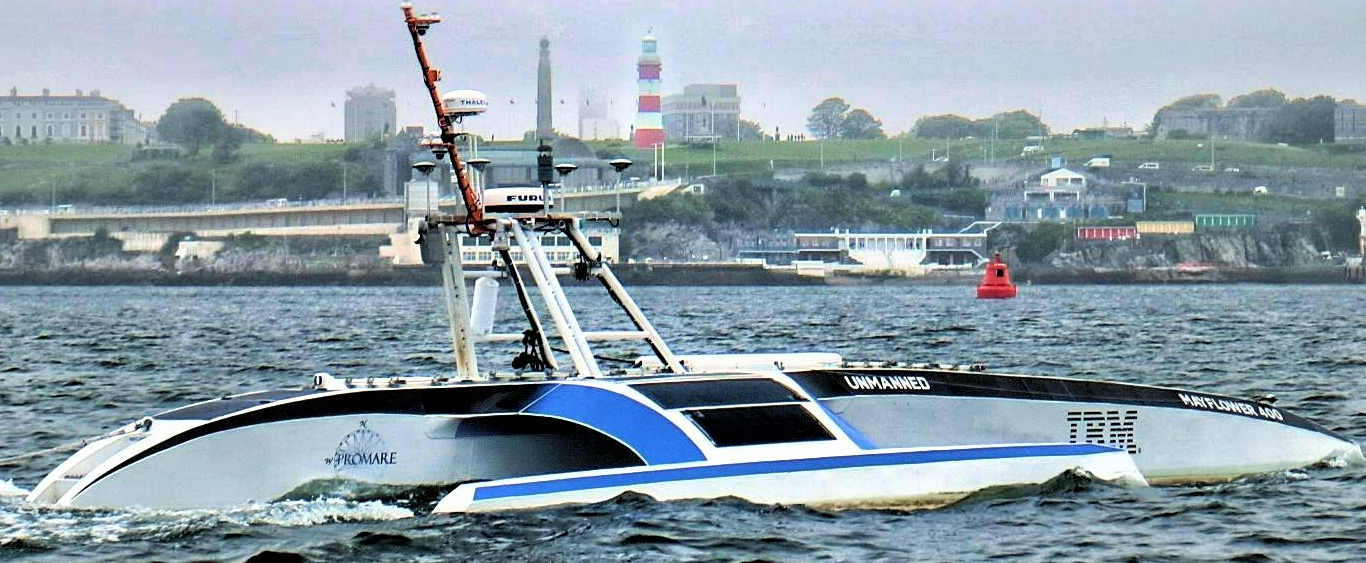}
	\caption{
		\emph{Mayflower autonomous ship} (Section~\ref{sc:driveless-cars}) sailing from Plymouth, UK, planning to arrive at Plymouth, MA, U.S., like the original Mayflower 400 years ago, but instead
		arriving at Halifax, Nova Scotia, Canada, on 2022 Jun 05, due to mechanical problems
		\cite{obrien2022autonomous}.
		%		\cite{nims2020robot}.  
		%\\
		{\footnotesize (\href{https://creativecommons.org/licenses/by-sa/4.0/deed.en}{CC BY-SA 4.0}, Wikipedia, \href{https://commons.wikimedia.org/w/index.php?title=File:Mayflower_Autonomous_Ship_inside_Plymouth_Sound.jpg&oldid=675176688}{version 16:43, 17 July 2022}.)} 
	}
	\label{fig:mayflower-auto-ship}
\end{figure}

Beyond the possible lost of connection in a human remote-control ship, mechanical failure did occur, such as that happened for the Mayflower autonomous ship shown in Figure~\ref{fig:mayflower-auto-ship} \cite{obrien2022autonomous}.  Measures would have to be taken when mechanical failure happens to a crewless ship in the middle of a vast ocean. 

See also the interview of S.J. Russell 
%by 
in
\cite{Ford.2018} on the need to develop hybrid systems that have classical AI along side with deep learning, which has limitations, even though it is good at classification and perception,\footnote{
	S.J. Russell also appeared in the \hyperref[para:AI-making-easier-to-kill]{video} ``AI is making it easier...'' mentioned at the end of this closure section.
}  and Section~\ref{sc:barrier-of-meaning} on the barrier of meaning in AI.

\subsection{Lack of understanding on why deep learning worked}
\label{sc:lack-understanding}

Such lack of understanding is described in {\em The Guardian}'s  Editorial on the 2019 New Year Day \cite{Guardian.2019:rd0001} as follows:
\begin{quote}
	``Compared with conventional computer programs, [AI that teaches itself] acts for reasons incomprehensible to the outside world. It can be trained, as a parrot can, by rewarding the desired behaviour; in fact, this describes the whole of its learning process. But it can't be consciously designed in all its details, in the way that a passenger jet can be. If an airliner crashes, it is in theory possible to reconstruct all the little steps that led to the catastrophe and to understand why each one happened, and how each led to the next. Conventional computer programs can be debugged that way. This is true even when they interact in baroquely complicated ways. But neural networks, the kind of software used in almost everything we call AI, can't even in principle be debugged that way. We know they work, and can by training encourage them to work better. But in their natural state it is quite impossible to reconstruct the process by which they reach their (largely correct) conclusions.''
\end{quote}

\begin{figure}[h]
	\centering
	\includegraphics[width=0.80\linewidth]{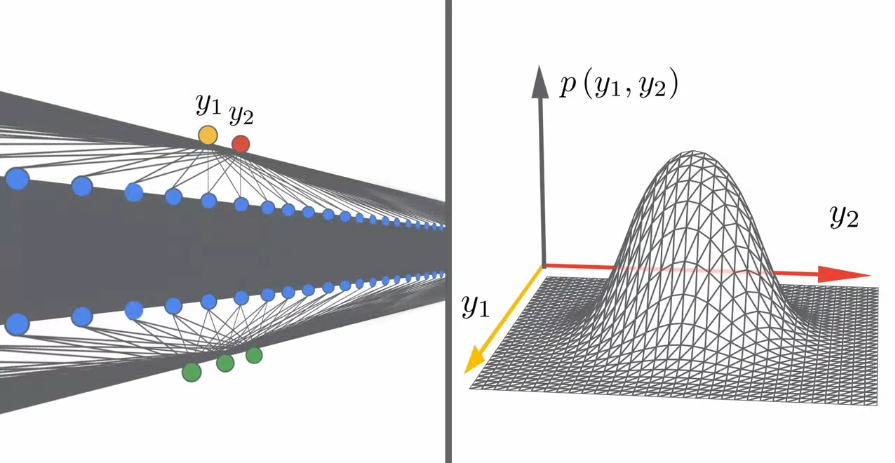}
	\caption{
		\emph{Network with infinite width} (left) and Gaussian distribution (Right) (Section~\ref{sc:training-valication-test}, \ref{sc:lack-understanding}). 
		``A number of recent results have shown that DNNs that are allowed to become infinitely wide converge to another, simpler, class of models called Gaussian processes. In this limit, complicated phenomena (like Bayesian inference or gradient descent dynamics of a convolutional neural network) boil down to simple linear algebra equations. Insights from these infinitely wide networks frequently carry over to their finite counterparts. As such, infinite-width networks can be used as a lens to study deep learning, but also as useful models in their own right''
		\cite{novak2020fast} \cite{lee2018deep}.
		See Figures~\ref{fig:double-descent-risk} and \ref{fig:test-error-large-params} for the motivation for networks with infinite width.
		%		\cite{nims2020robot}.  
		%\\
		{\footnotesize (\href{https://creativecommons.org/licenses/by-sa/4.0/deed.en}{CC BY-SA 4.0}, Wikipedia, \href{https://commons.wikimedia.org/w/index.php?title=File:Infinitely_wide_neural_network.webm&oldid=665988221}{version 03:51, 18 June 2022}.)}  
	}
	\label{fig:network-infinite-width}
\end{figure}

The 2021 breakthough in computer science, as declared by the Quanta Magazine \cite{quanta2021breakthroughs}, was the discovery of the connection between shallow networks with infinite width (Figure~\ref{fig:network-infinite-width}) and kernel machines (or methods) as a first step in trying to understand how deep-learning networks work; see Section~\ref{sc:kernel-machines} on ``Kernel machines'' and Footnote~\ref{fn:support-vector-machine}.

%\cite{https://doi.org/10.48550/arxiv.1711.00165}
%\cite{lee2018deep}

\subsection{Barrier of meaning}
\label{sc:barrier-of-meaning}
Deep learning could not think like humans do, and could be easily fooled as 
%
% CMES style rewriting
%\cite{Mitchell.2018:rd0001} wrote on this subject:
reported in \cite{Mitchell.2018:rd0001}:
\begin{quote}
	``
	Machine learning algorithms don't yet understand things the way humans do---with sometimes disastrous consequences.
	
	\hspace{5mm}
	Even more worrisome are recent demonstrations of the vulnerability of A.I. systems to so-called adversarial examples. In these, a malevolent hacker can make specific changes to images, sound waves or text documents that while imperceptible or irrelevant to humans will cause a program to make potentially catastrophic errors.
	
	\hspace{5mm}
	The possibility of such attacks has been demonstrated in nearly every application domain of A.I., including computer vision, medical image processing, speech recognition and language processing. Numerous studies have demonstrated the ease with which hackers could, in principle, fool face- and object-recognition systems with specific minuscule changes to images, put inconspicuous stickers on a stop sign to make a self-driving car's vision system mistake it for a yield sign or modify an audio signal so that it sounds like background music to a human but instructs a Siri or Alexa system to perform a silent command.
	
	\hspace{5mm}
	These potential vulnerabilities illustrate the ways in which current progress in A.I. is stymied by the barrier of meaning. Anyone who works with A.I. systems knows that behind the facade of humanlike visual abilities, linguistic fluency and game-playing prowess, these programs do not---in any humanlike way---understand the inputs they process or the outputs they produce. The lack of such understanding renders these programs susceptible to unexpected errors and undetectable attacks.
	
	\hspace{5mm}
	As the A.I. researcher Pedro Domingos noted in his book \emph{The Master Algorithm}, `People worry that computers will get too smart and take over the world, but the real problem is that they're too stupid and they've already taken over the world.'
	''
\end{quote}
Such barrier of meaning is also a barrier for AI to tackle human controversies; see Section~\ref{sc:human-problems}.
See also Section~\ref{sc:driveless-cars} on driverless cars not coming any time soon, which is related to the above barrier of meaning.

\subsection{Threat to democracy and privacy}
On the 2019 new-year day, {\em The Guardian} \cite{Guardian.2019:rd0001} not only reported the most recent breakthrough in AI research on the development of AlphaZero, a software possessing superhuman performance in several ``immensely complex'' games such as Go (see Section~\ref{sc:resurgence-2} on resurgence of AI and current state), they also reported another breakthrough as a more ominous warning on a ``Power struggle'' to preserve liberal democracies against authoritarian governments and criminals:

\begin{quote}
	``The second great development of the last year makes bad outcomes much more likely. This is the much wider availability of powerful software and hardware. Although vast quantities of data and computing power are needed to train most neural nets, once trained a net can run on very cheap and simple hardware. This is often called the democratisation of technology but it is really the anarchisation of it. Democracies have means of enforcing decisions; anarchies have no means even of making them. The spread of these powers to authoritarian governments on the one hand and criminal networks on the other poses a double challenge to liberal democracies. Technology grants us new and almost unimaginable powers but at the same time it takes away some powers, and perhaps some understanding too, that we thought we would always possess.''
\end{quote} 

Nearly three years later, a report of a national poll of 2,200 adults in the U.S., released on 2021.11.15, indicated that three in four adults were concerned about the loss of privacy, ``loss of trust in elections (57\%), in threats to democracy (52\%), and in loss of trust in institutions (56\%). Additionally, 58\% of respondents say it has contributed to the spread of misinformation''
\cite{stevens2021new}.

\begin{figure}[h]
	\centering
	\includegraphics[height=4cm]{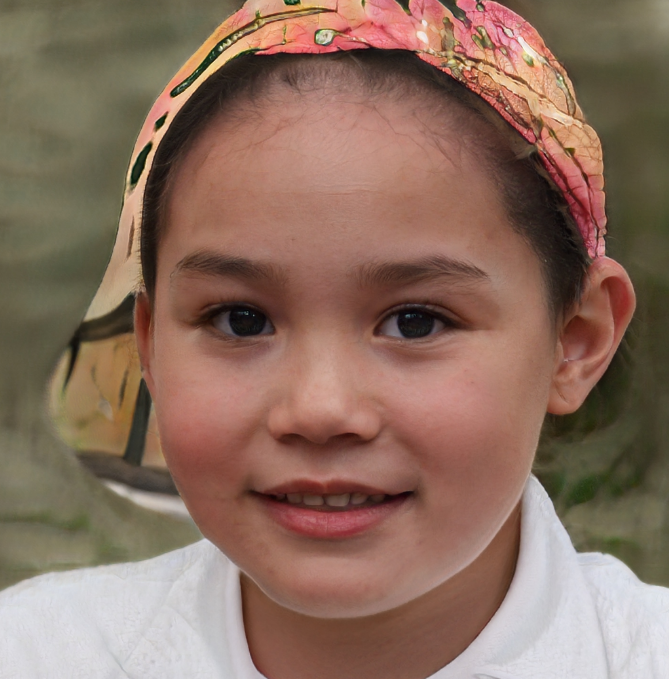}
	\includegraphics[height=4cm]{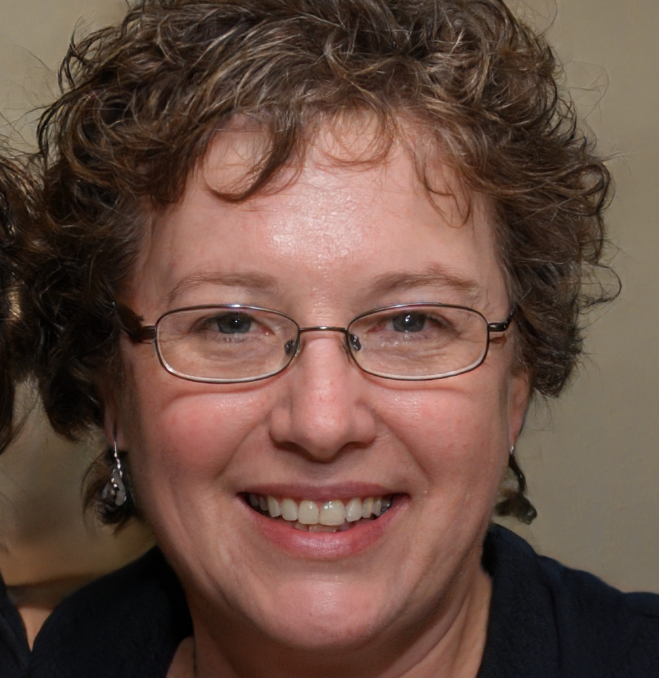}
	\includegraphics[height=4cm]{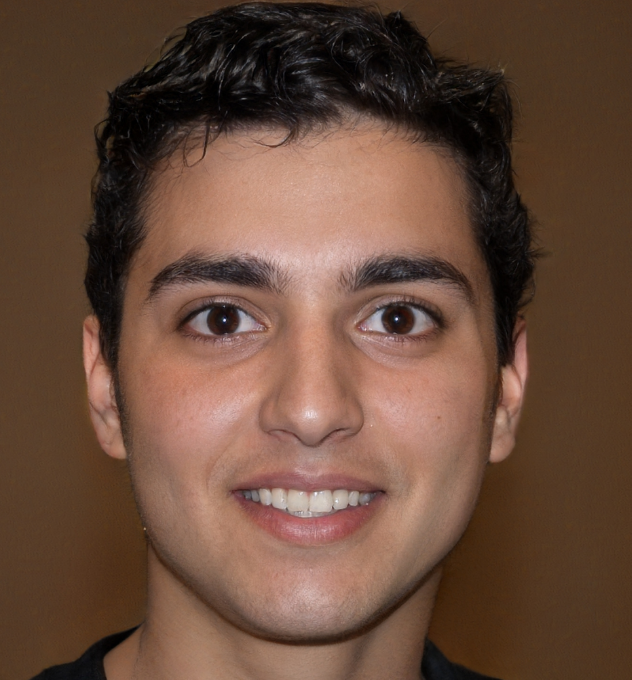}
	\caption{
		\emph{Deepfake images} (Section~\ref{sc:deepfakes}).  AI-generated portraits using Generative Adversarial Network (GAN) models.
		See also \cite{hu2020computer} \cite{sencar2022multimedia}, Chap.~8, ``GAN Fingerprints in Face Image Synthesis.''
		%\\
		{\footnotesize (Images from \href{https://thispersondoesnotexist.com/}{`This Person Does Not Exist'} site.)} 
	}
	\label{fig:deepfakes}
\end{figure}

\subsubsection{Deepfakes}
\label{sc:deepfakes}
AI software available online helping to create videos that show someone said or did things that the person did not say or do represent a clear danger to democracy, as these deepfake videos could affect the outcome of an election, among other misdeeds, with risk to national security.
Advances in machine learning have made deepfakes ``ever more realistic and increasingly resistant to detection'' \cite{Chesney.2019}; see Figure~\ref{fig:deepfakes}. 
%
% CMES style rewriting 
%Concurred \cite{Ingram.2019}, who wrote:
The authors of \cite{Ingram.2019} concurred:
\begin{quote}
	``Deepfake videos made with artificial intelligence can be a powerful force because they make it appear that someone did or said something that they never did, altering how the viewers see politicians, corporate executives, celebrities and other public figures.  The tools necessary to make these videos are available online, with some people making celebrity mashups and one app offering to insert users' faces into famous movie scenes.''
\end{quote}

To be sure, deepfakes do have benefits in education, arts, and individual autonomy \cite{Chesney.2019}.  In education, deepfakes could be used to provide information to students in a more interesting manner.  For example, deepfakes make it possible to ``manufacture videos of historical figures speaking directly to students, giving an otherwise unappealing lecture a new lease on life''.  In the arts, deepfake technology allowed to resurrect long dead actors for fresh roles in new movies.  An example is a recent \emph{Star Wars} movie with the deceased actress Carrie Fisher.  In helping to maintain some personal autonomy, deepfake audio technology could help restore the ability to speak for a person suffered from some form of paralysis that prevents normal speaking. 

On the other hand, 
%
% CMES style rewriting
the authors of
\cite{Chesney.2019} 
cited a long list of harmful uses of deepfakes, from harm to individuals or organizations (e.g., exploitation, sabotage), to harm to society (e.g., distortion of democratic discourse, manipulation of elections, eroding trust in institutions, exacerbating social division, undermining public safety, undermining diplomacy, jeopardizing national security, undermining journalism, crying deepfake news as liar's dividend).\footnote{
	Watch also Danielle Citron's 2019 TED talk ``How deepfakes undermine truth and threaten democracy'' \cite{citron2019deepfakes}.
}  See also \cite{manheim2019artificial} \cite{hao2019why} \cite{feldstein2019how} \cite{pearce2021beware}. 

Researchers have been in a race to develop methods to detect deepfakes, a difficult technological challenge \cite{Harwell.2019}.  One method is to spot the subtle characteristics of how someone spoke to provide a basis to determine whether a video was true or fake \cite{Ingram.2019}.  But that method was not a top-five winner of the \emph{DeepFake Detection Challenge} (DFDC) \cite{DFDC2019-2020} organized in the period 2019-2020 by ``The Partnership for AI, in collaboration with large companies including Facebook, Microsoft, and Amazon,'' with a total prize money of one million dollars, divided among the top five winners, out of more than two thousand teams \cite{groh2022deepfake}.

\begin{figure}[h]
	\centering
	\includegraphics[width=0.95\linewidth]{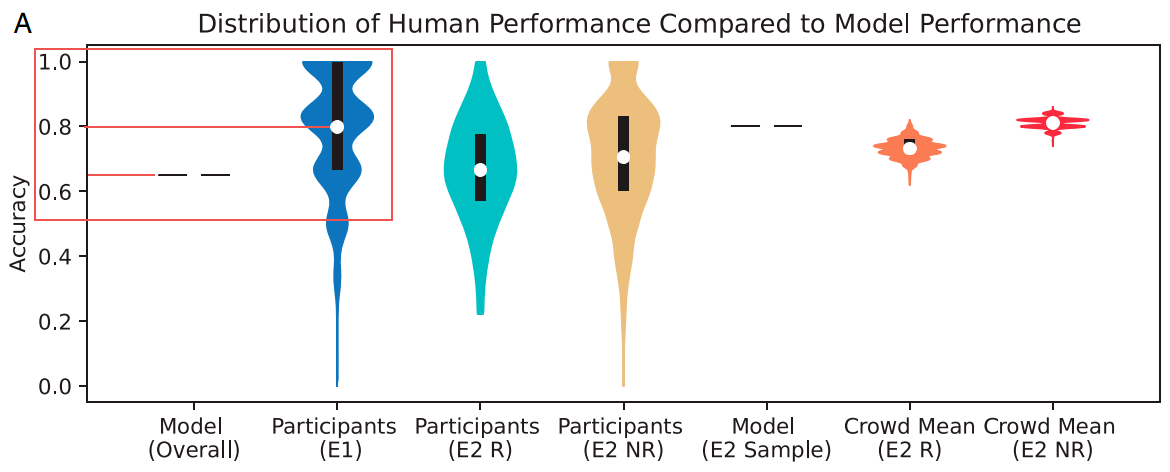}
	\caption{
		\emph{DeepFake detection} (Section~\ref{sc:deepfakes}).  Violin plots. 
		{\large $\bullet$}
		\emph{Individual vs machine}.
		The leading model had an accuracy of 65\% on 4,000 videos (Col.~1).
		In Experiment 1 (E1), 5,524 participants were asked to identify a deepfake from each of 56 pairs of videos.
%		select which of the two videos was a deepfake, and this for 56 pairs of videos.  
		The participants had a mean accuracy of 80\% (white dot in Col.~2), with 82\% of the participants having an accuracy better than that of the leading model (65\%). 
		In Experiment 2 (E2), using a subset of randomly sampled videos, the recruited (R) participants had mean accuracy at 66\% (Col.~3),  the non-recruited (NR) participants at 69\% (Col.~4), and leading model at 80\%. 
		{\large $\bullet$}
		\emph{Crowd wisdom vs machine}.
		Crowd mean is the average accuracy by participants for each video.  R participants had a crowd-mean average accuracy at 74\%, NR participants at 80\%, which was the same for the leading model, and NR participants who viewed at least 10 videos at 86\%
		\cite{groh2022deepfake}.
		%		\cite{nims2020robot}.  
		%\\
		{\footnotesize \href{https://creativecommons.org/licenses/by-nc-nd/4.0/}{(CC BY-NC-ND 4.0)}
		} 
	}
	\label{fig:deepfake-1}
\end{figure}

Human's ability to detect of deepfakes compared well with the ``leading model,'' i.e., the DFCD top winner \cite{groh2022deepfake}. The results were ``at odds with the commonly held view in media forensics that ordinary people have extremely limited ability to detect media manipulations'' \cite{groh2022deepfake}; see Figure~\ref{fig:deepfake-1}, where the width of a violin plot,\footnote{
	See the classic original paper \cite{hintze1998violin}, which was cited 1,554 times on Google Scholar as of 2022.08.24.  See also \cite{lewinson2019violin} with Python code and resulting images on \href{https://github.com/erykml/medium_articles/blob/master/Statistics/violin_plots.ipynb}{GitHub}.
} at a given accuracy, represents the number of participants.  In Col.~2 of Figure~\ref{fig:deepfake-1}, the area of the blue violin \emph{above} the leading model accuracy of 65\% represents 82\% of the participants, represented by the area of the whole violin.
A crowd does have a collective accuracy comparable to (or for those who viewed at least 10 videos, better than) the leading model; see Cols.~5, 6, 7 in Figure~\ref{fig:deepfake-1}.

While it is difficult to detect AI deepfakes, the MIT Media Lab DeepFake detection project advised to pay attention to the following eight facial features \cite{mitDeepFakes}:

\begin{enumerate}
	\item
	``Face. High-end DeepFake manipulations are almost always facial transformations.
	
	\item
	``Cheeks and forehead. Does the skin appear too smooth or too wrinkly? Is the agedness of the skin similar to the agedness of the hair and eyes? DeepFakes are often incongruent on some dimensions.
	
	\item
	``Eyes and eyebrows. Do shadows appear in places that you would expect? DeepFakes often fail to fully represent the natural physics of a scene. 
	
	\item
	``Glasses. Is there any glare? Is there too much glare? Does the angle of the glare change when the person moves? Once again, DeepFakes often fail to fully represent the natural physics of lighting.
	
	\item
	``Facial hair or lack thereof. Does this facial hair look real? DeepFakes might add or remove a mustache, sideburns, or beard. But, DeepFakes often fail to make facial hair transformations fully natural.
	
	\item 
	``Facial moles.  Does the mole look real?
	
	\item 
	Eye ``blinking. Does the person blink enough or too much?
	
	\item 
	``Size and color of the lips. Does the size and color match the rest of the person's face?''
	
\end{enumerate}

\subsubsection{Facial recognition nightmare}
\label{sc:facial-recognition}
``We're all screwed'' as a Clearview AI, startup company, uses deep learning to identify faces against a large database involving more than three billions photos collected from ``Facebook, Youtube, Venmo and millions of other websites'' \cite{Hill.2020}.   Their software ``could end your ability to walk down the street anonymously, and provided it to hundreds of law enforcement agencies''.   More than 600 law enforcement agencies have started to use Clearview AI software to ``help solve shoplifting, identity theft, credit card fraud, murder and child sexual exploitation cases''.  On the other hand, the tool could be abused, such as identifying ``activists at a protest or an attractive stranger on the subway, revealing not just their names but where they lived, what they did and whom they knew''.  Some large cities such as San Francisco has banned to use of facial recognition by the police.

A breach of Clearview AI database occurred just a few weeks after the article by \cite{Hill.2020}, an unforeseen, but not surprising, event \cite{Morrison.2020}: 
\begin{quote}
	``Clearview AI, the controversial and secretive facial recognition company, recently experienced its first major data breach---a scary prospect considering the sheer amount and scope of personal information in its database, as well as the fact that access to it is supposed to be restricted to law enforcement agencies.''
\end{quote}
The leaked documents showed that Clearview AI had a large range of customers, ranging from law-enfor\-cement agencies (both domestic and internatinal), to large retail stores (Macy's, Best Buy, Walmart).  Experts describe Clearview AI's plan to produce a publicly available face recognition app as ``dangerous''.   
So we got screwed again.

There was a documented wrongful arrest by face-recognition algorithm that demonstrated racism, i.e., a bias toward people of color \cite{hill2020wrong}.  A detective showed the wrongful-arrest victim a photo that was clearly not the victim, and asked ``Is this you?'' to which the victim replied ``You think all black men look alike?''

%This racial bias was well-known:
It is well known that AI has 
``propensity to replicate, reinforce or amplify harmful existing social biases'' \cite{raji2020closing}, such as racial bias \cite{metz2021who} among others:
``An early example arose in 2015, when a software engineer pointed out that Google's image-recognition system had labeled his Black friends as `gorillas.' Another example arose when Joy Buolamwini, an algorithmic fairness researcher at MIT, tried facial recognition on herself---and found that it wouldn't recognize her, a Black woman, until she put a white mask over her face. These examples highlighted facial recognition's failure to achieve another type of fairness: representational fairness'' \cite{samuel2022why}.\footnote{
	See also \cite{heilinger2022ethics} on a number of relevant AI ethical issues such as: ``Who bears responsibility in the event of harm resulting from the use of an AI system; How can AI systems be prevented from reflecting existing discrimination, biases 
	and social injustices based on their training data, thereby exacerbating them; How can the privacy of people be protected, given that personal data can be collected and analysed so easily by many.''  Perhaps the toughest question is ``Who should get to decide which moral intuitions, which values, should be embedded in algorithms?'' \cite{samuel2022why}.
}  

%But the cavalry was already coming to rescue.  
A legal measure has been taken against gathering data for  facial-recognition software.
In May 2022, Clearview AI was slapped with a ``\$10 million for scraping UK faces from the web. That might not be the end of it''; in addition, ``the firm was also ordered to delete all of the data it holds on UK citizens''
\cite{heikkila2022the}.   

There were more of such measures: ``Earlier this year, Italian data protection authorities fined Clearview AI \texteuro 20 million (\$21 million) for breaching data protection rules. Authorities in Australia, Canada, France, and Germany have reached similar conclusions. 

Even in the US, which does not have a federal data protection law, Clearview AI is facing increasing scrutiny. Earlier this month the ACLU won a major settlement that restricts Clearview from selling its database across the US to most businesses. In the state of Illinois, which has a law on biometric data, Clearview AI cannot sell access to its database to anyone, even the police, for five years''
\cite{heikkila2022the}.
 
%{\color{red} HERE, 2020.02.05}

\subsection{AI cannot tackle controversial human problems}
\label{sc:human-problems}
If there was a barrier of meaning as described in Section~\ref{sc:barrier-of-meaning}, it is clear that there are many problems that AI could not be trained to solve since even humans do not agree on how to classify certain activities as offending or acceptable.  
%
% CMES style rewriting
%\cite{Metz.2019:rd0001} wrote: 
It was written in \cite{Metz.2019:rd0001} the following:
\begin{quote}
	``Mr. Schroepfer---or Schrep, as he is known internally---is the person at Facebook leading the efforts to build the automated tools to sort through and erase the millions of [hate-speech] posts. But the task is Sisyphean, he acknowledged over the course of three interviews recently.
	
	\hspace{5mm}
	That's because every time Mr. Schroepfer [Facebook's Chief Technology Officer] and his more than 150 engineering specialists create A.I. solutions that flag and squelch noxious material, new and dubious posts that the A.I. systems have never seen before pop up---and are thus not caught. The task is made more difficult because ``bad activity'' is often in the eye of the beholder and humans, let alone machines, cannot agree on what that is.
	
	\hspace{5mm}
	``I don't think I'm speaking out of turn to say that I've seen Schrep cry at work,'' said Jocelyn Goldfein, a venture capitalist at Zetta Venture Partners who worked with him at Facebook.'' 
\end{quote}

\subsection{So what's new? Learning to think like babies}
\label{sc:whats-new}
Because of AI's inability to understand (barrier of meaning) and to solve controversial human issues, a idea to tackle such problems is to start with baby steps in trying to teach AI to think like babies, as recounted by \cite{Gopnik.2019}:
\begin{quote}
	``The problem is that these new algorithms are beginning to bump up against significant limitations. They need enormous amounts of data, only some kinds of data will do, and they're not very good at generalizing from that data. Babies seem to learn much more general and powerful kinds of knowledge than AIs do, from much less and much messier data. In fact, human babies are the best learners in the universe. How do they do it? And could we get an AI to do the same?
	
	\hspace{5mm}
	First, there's the issue of data. AIs need enormous amounts of it; they have to be trained on hundreds of millions of images or games. 
	
	\hspace{5mm}
	Children, on the other hand, can learn new categories from just a small number of examples. A few storybook pictures can teach them not only about cats and dogs but jaguars and rhinos and unicorns.
	
	\hspace{5mm}
	AIs also need what computer scientists call ``supervision.'' In order to learn, they must be given a label for each image they ``see'' or a score for each move in a game. Baby data, by contrast, is largely unsupervised. 
	
	\hspace{5mm}
	Even with a lot of supervised data, AIs can't make the same kinds of generalizations that human children can. Their knowledge is much narrower and more limited, and they are easily fooled by what are called ``adversarial examples.'' For instance, an AI image recognition system will confidently say that a mixed-up jumble of pixels is a dog if the jumble happens to fit the right statistical pattern---a mistake a baby would never make.''
\end{quote}
Regarding early stopping and generalization error in network training, see Remark~\ref{rm:generalization-capability} in Section~\ref{sc:training-valication-test}.
To make AIs into more robust and resilient learners, researchers are developing methods to build curiosity into AIs, instead of focusing on immediate rewards.

\begin{figure}[h]
	\centering
	\includegraphics[width=0.58\linewidth]{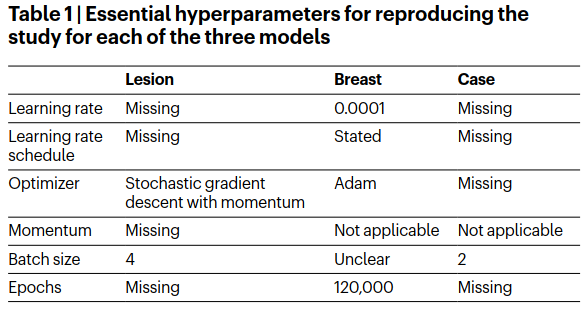}
	\caption{
		\emph{Lack of transparency and irreproducibility} (Section~\ref{sc:irreproducibility}). 
		The table shows many missing pieces of information for the three networks---Lesion, Breast, and Case models---used to detect breast cancer.
		Learning rate, Section~\ref{sc:learning-rate}.
		Learning-rate schedule, Section~\ref{sc:generic-SGD}, Figure~\ref{fig:minibatch-increase-training-schedules} in Section~\ref{sc:minibatch-size-increase}.
		SGD with momentum, Section~\ref{sc:SGD-momentum} and Remark~\ref{rm:choice-momentum-param}.
		Adam algorithm, Section~\ref{sc:adam1}. 
		Batch size, Sections~\ref{sc:generic-SGD} and \ref{sc:minibatch-size-increase}.
		Epoch, Footnote~\ref{fn:epoch}.
		\cite{haibe-kains2020transparency}.
%		\\
		{\footnotesize (Figure reproduced with permission of the authors)} 
	}
	\label{fig:Haibe-Kains-irreproducibility}
\end{figure}

\subsection{Lack of transparency and irreproducibility of results}
\label{sc:irreproducibility}
For ``multiple years now'', there have been articles on deep learning that looked more like a promotion/advertisement for newly developed AI technologies, rather than scientific papers in the traditional sense that published results should be replicable and verifiable \cite{heaven2020ai}.  But it was only on 2020 Oct 14 that many scientists \vphantom{\cite{haibe-kains2020transparency}}\cite{haibe-kains2020transparency} had enough and protested 
the lack of transparency in AI research
%this practice 
in a ``damning'' article in \emph{Nature}, a major scientific journal.
\begin{quote}
	``We couldn't take it anymore,'' says Benjamin Haibe-Kains, the lead author of the response, who studies computational genomics at the University of Toronto. ``It's not about this study in particular---it's a trend we've been witnessing for multiple years now that has started to really bother us.''
	\cite{heaven2020ai}
\end{quote}
The particular contentious study was published by the Google-Health authors of \vphantom{\cite{mckinney2020international}}\cite{mckinney2020international} on the use of AI in medical imaging to detect breast cancer.  But these authors of \cite{mckinney2020international} provided so little information about their code and how it was tested that their article read more like a ``promotion of proprietary tech'' than a scientific paper.  Figure~\ref{fig:Haibe-Kains-irreproducibility} shows the missing pieces of crucial information to reproduce the results.
A question would immediately come to mind: Why would a reputable journal like \emph{Nature} accept such a paper?  Was the review rigorous enough?
\begin{quote}
	``When we saw that paper from Google, we realized that it was yet another example of a very high-profile journal publishing a very exciting study that has nothing to do with science,'' Haibe-Kains says. ``It's more an advertisement for cool technology. We can't really do anything with it.''
	\cite{heaven2020ai}
\end{quote}
According to \cite{haibe-kains2020transparency}, even though the Google-Health authors of \cite{mckinney2020international} stated that ``all experiments and implementation details were described in sufficient detail in the supplementary methods section of their Article to `support replication with non-proprietary libraries','' that was a subjective statement, and replicating their results would be a difficult task, since such textual description can hide a high level of complexity of the code, and  nuances in the computer code can have large effects in the training and evaluation results.

\begin{quote}
	``AI is feeling the heat for several reasons. For a start, it is a newcomer. It has only really become an experimental science in the past decade, says Joelle Pineau, a computer scientist at Facebook AI Research and McGill University, who coauthored the complaint. `It used to be theoretical, but more and more we are running experiments,' she says. `And our dedication to sound methodology is lagging behind the ambition of our experiments.' ''
	\cite{heaven2020ai}
\end{quote}

No progress in science could be made if results were not verifiable and replicable by independent researchers.

\subsection{Killing you!}
\label{para:AI-making-easier-to-kill}
Oh, one more thing: ``A.I. Is Making it Easier to Kill (You). Here's How,'' New York Times Documentaries, 2019.12.13 (\href{https://www.nytimes.com/video/technology/100000006082083/lethal-autonomous-weapons.html}{Original website}) (\href{https://www.youtube.com/watch?v=GFD_Cgr2zho}{Youtube}).

And getting better at it every day, e.g., by using ``a suite of artificial intelligence-driven systems that will be able to control networked `loyal wingman' type drones and fully autonomous unmanned combat air vehicles'' \cite{trevithick2020general}; see also ``Collaborative Operations in Denied Environment (CODE) Phase 2 Concept Video''  (\href{https://www.youtube.com/watch?v=2cWa7hCAwkk}{Youtube}).

%{\color{red} HERE 2020.12.28.  add a ref and a photo and a reference to a youtube video on new AI application in military drones that can avoid zones without internet coverage, fly in formation autonomously.}

%\section{References, all}
% all references in the bib file are listed below.

% \nocite{*}

% \bibliographystyle{acm}
% \bibliographystyle{plain}

% \bibliography{%
%	data-driven-computation-mechanic,%
%	data-driven-computing-or-computation,%
%	deep-learning_alex,%
%	deep-learning-comput-mechanic,%
%	deep-learning-misc,%
%	ghaboussi,%
%	mechanics,%
%	miscellaneous,%
%	neural-network-mechanics,%
%	vuquoc-DL,%
%}

% 2020.03.28
%\addcontentsline{toc}{section}{Acknowledgement}
%\section*{Acknowledgement}
%\label{sc:acknowledgement}
%The support of the COMET-K2 Center of the Linz Center of Mechatronics (LCM) funded by the Austrian federal government and the federal state of Upper Austria is acknowledged.

% 2020.02.17
% add References to table of contents
%https://tex.stackexchange.com/questions/8458/making-the-bibliography-appear-in-the-table-of-contents
\addcontentsline{toc}{section}{References}
% 2019.02.10, need to put all bib files on one line for TexStudio to automatically 
% list the bib entry labels to select the desired label, without having to remember
% the desired label or having to look up the desired label in the bib file.  
% splitting the bib files into several lines took away this useful feature.
\bibliography{data-driven-computation-mechanic,data-driven-computing-or-computation,deep-learning_alex,deep-learning-comput-mechanic,deep-learning-misc,ghaboussi,mechanics,miscellaneous,neural-network-mechanics,vuquoc-DL,CMES-2020-21-papers-Covid19,CMES-2020-21-papers-deep-machine-learning}
\label{sc:references}

% 2020.02.17
% https://texfaq.org/FAQ-appendix
% 2022.06.08 - Do NOT use the command \appendix for CMES style
%\appendix
%\addcontentsline{toc}{section}{Appendices}

\newpage
% 2022.06.08 - USE the appendices environment for CMES style
\noindent
{\Large \bf Appendices}
\addcontentsline{toc}{section}{Appendices}
\begin{appendices}
		
	% compare backprop pseudocode to that in Goodfellow 2016
	%\newpage
\section{Backprop pseudocodes, notation comparison}
\label{app:backprop-pseudocode}
%\addcontentsline{toc}{section}{Appendices}

To connect the backpropagation Algorithm~\ref{algo:backprop} in Section~\ref{sc:backprop} to Algorithm~6.4 in \cite{Goodfellow.2016}, p.206, Section 6.5.4 on ``Back-Propagation Computation in Fully Connected MLP'',\footnote{
	MLP = MultiLayer Perceptron.
}  
a different form of Algorithm~\ref{algo:backprop}, where the ``while'' loop is used, is provided in Algorithm~\ref{algo:backprop-2}, where the ``for'' loop is used. This information would be especially useful for first-time learners.  See also Remark~\ref{rm:backprop}.

\begin{algorithm}[h]
	{\bf Backpropagation pseudocode 2}
	\\
	\KwData{
		% Layer outputs $\byp{\ell}$, for $\ell=L, \cdots , 1$
		\\
		$\bullet$ Input into network $\bx = \byp{0} \in \real^{\widths{0} \times 1}$
		\\
		$\bullet$ Learning rate $\epsilon$; see Section~\ref{sc:learning-rate} on deterministic optimization  
		\\
		$\bullet$ Results from \emph{any} forward propagation: 
		\\
		\hspace{5mm} $\star$ Network parameters $\bparam = \{\bparamp{1} , \cdots , \bparamp{L} \}$ (all layers)
		\\
		\hspace{5mm} $\star$ Layer weighted inputs and biases $\bzp{\ell}$, for $\ell = 1, \cdots, L$
		\\
		\hspace{5mm} $\star$ Layer outputs $\byp{\ell}$, for $\ell = 1, \cdots, L$
	}
	\KwResult{
		Updated network parameters $\bparam$ to reduce cost function $J$.
	}
	\vphantom{Blank line}
	{\bf Initialize:} 
	\\
	$\bullet$ Gradient $\boldsymbol{r} = \partial J / \partial \byp{L} \in \real^{1 \times \widths{L}}$ (row) on predicted output $\bout = \byp{L}$,  Eq.(\ref{eq:d-cost-output-L}) 
	%\\
	%$\bullet$ Set layer counter $\ell$ to last layer $L$, i.e., $\ell \leftarrow L$
	\;
	
	\vphantom{Blank line}
	\For{$\ell = L, \ldots , 1$}{
		
		$\blacktriangleright$ Compute gradient on weighted inputs (pre-nonlinear activation) $\bzp{\ell}$:
		\\
		$\boldsymbol{r} \leftarrow \partial J / \partial \bzp{\ell} = \boldsymbol{r} \odot \g^\prime (\bzp{\ell})^T \in \real^{1 \times \widths{\ell}}$ (row), Eq.(\ref{eq:gradient-common})
		\label{lst:line:gradient-on-weighted-inputs}
		\;
		
		$\blacktriangleright$ Compute gradient on layer parameters $\bparamp{\ell}$:
		\\
		$\partial J / \partial \bparamp{\ell} = \boldsymbol{r}^{T}
		\expand{\by}^{(\ell-1) T}
		\in
		\real^{\widths{\ell} \times [\widths{\ell-1}  + 1]}$ (row), Eq.(\ref{eq:gradient-7})
		\;
		
		$\blacktriangleright$ Compute gradient on layer outputs $\byp{\ell-1}$: 
		%of lower-level layer $(\ell-l)$:
		\\
		$\boldsymbol{r} \leftarrow \partial J / \partial \byp{\ell-1} = \boldsymbol{r}\bWeight^{(\ell)}
		\in
		\real^{1 \times \widths{\ell -1}}$ (row), Eq.(\ref{eq:gradient-y(l-1)-4})
		\;
		
	}
	
	\vphantom{Blank line}
	\caption{
		\emph{Backpropagation pseudocode.} Alternative presentation of Algorithm~\ref{algo:backprop}, where the ``while'' loop is used, to compare to Algorithm 6.4 in \cite{Goodfellow.2016}, p.206, where the ``for'' loop was used, and where there was no layer-parameter update step (Line~\ref{lst:line:update-params-backprop} of Algorithm~\ref{algo:backprop}).
		See also Table~\ref{tb:backprop-comparison} for the correspondence between the notations employed here and those in \cite{Goodfellow.2016}, and the block diagrams in Figure~\ref{fig:backprop-1} and Figure~\ref{fig:backprop-2}.
	}
	\label{algo:backprop-2}
\end{algorithm}

In Algorithm~\ref{algo:backprop-2}, the regularization of the cost function $J$ is not considered, i.e., we omit the penalty (or regularization) term $\lambda \Omega(\bparam)$ used in Algorithm~6.4 in \cite{Goodfellow.2016}, p.206, where the regularized cost was $J(\bparam) + \lambda \Omega(\bparam)$.   As pointed out in Section~\ref{sc:weight-decay}, weight decay is more general than $L_2$ regularization, and would be the preferred method to avoid overfitting.

In Table~\ref{tb:backprop-comparison}, the correspondence between the notations employed here and those in \cite{Goodfellow.2016}, p.206, is provided. 

\begin{table}
	\caption{
		Equivalence of backprop Algorithm~\ref{algo:backprop-2} and Algorithm~6.4 in \cite{Goodfellow.2016}, p.206.  Comparison of notations.  The mathematical expressions in Algorithm~6.4 are reproduced here in their original notations, except for the matrix dimensions, which were not given in Algorithm~6.4 of \cite{Goodfellow.2016}.
	}
	\centering
	% fixed column widths
	%https://tex.stackexchange.com/questions/286950/how-to-create-a-table-with-fixed-column-widths
	%
	% padding
	%https://tex.stackexchange.com/questions/31672/column-and-row-padding-in-tables
	{\renewcommand{\arraystretch}{1.5}
		\begin{tabular}{|p{3.0in}|p{3.0in}|}
			\hline
			Algorithm~\ref{algo:backprop-2}, current notation 
			& 
			Goodfellow Algorithm 6.4, original notation
			\\
			\hline
			\hline
			$\bullet$ 
			Layer index = $\ell$
			&
			$\bullet$ 
			Layer index = $k$
			\\
			\hline
			$\bullet$ 
			Gradient = $\boldsymbol{r} \in \real^{1 \times \widths{\ell}}$ (row)
			&
			$\bullet$ 
			Gradient = $\boldsymbol{g} \in \real^{\widths{k} \times 1}$ (column)
			\\
			\hline
			$\bullet$ 
			Output of layer $(\ell)$ = $\byp{\ell} \in \real^{\widths{\ell} \times 1}$ (column)
			&
			$\bullet$ 
			Output of layer $(k)$ = $\boldsymbol{h}^{(k)} \in \real^{\widths{k} \times 1}$ (column)
			\\
			\hline
			$\bullet$ 
			Predicted output (last layer $(L)$): $\bout = \byp{L}$
			&
			$\bullet$ 
			Predicted output (last layer $(l)$): $\hat{\by} = \boldsymbol{h}^{(l)}$
			\\
			\hline
			$\bullet$ 
			Weighted inputs to layer $(\ell)$: $\bzp{\ell}$
			&
			$\bullet$ 
			Pre-nonlinear activation for layer $(k)$: $\boldsymbol{a}^{(k)}$
			\\
			\hline
			$\bullet$ 
			Activation function: $\g(\cdot)$
			&
			$\bullet$ 
			Activation function: $f(\cdot)$
			\\
			\hline
			$\bullet$ 
			Gradient on output of layer $(\ell)$:
			&
			$\bullet$ 
			Gradient on output of layer $(k)$:
			\\
			$\partial J / \partial \byp{\ell} \in \real^{1 \times \widths{\ell}}$  (row)
			& 
			$\nabla_{\boldsymbol{h}^{(k)}} J \in \real^{\widths{k} \times 1}$ (column)
			\\
			\hline
			$\bullet$ Gradient on weighted inputs:
			&
			$\bullet$ Gradient on prenonlinear activation:
			\\
			$\displaystyle \boldsymbol r^{(\ell)}
			:=
			\frac{\partial J}{\partial \bzp{\ell}}
			=
			\frac{\partial J}{\partial \byp{\ell}}
			\odot 
			\g^\prime ( \bz^{(\ell) T} )
			\in 
			\real^{1 \times \widths{\ell}}
			\text{ (row)}$ 
			& 
			$\boldsymbol{g} 
			=
			\nabla_{\boldsymbol{a}^{(k)}} J
			= 
			\nabla_{\boldsymbol{h}^{(k)}} J \odot f^\prime (\boldsymbol{a}^{(k)})
			\in 
			\real^{\widths{k} \times 1}
			\text{ (col)}$
			\\
			Code line~\ref{lst:line:gradient-on-weighted-inputs} in Algorithm~\ref{algo:backprop-2}
			&
			Code in Algorithm 6.4, \cite{Goodfellow.2016}
			\\
			$\boldsymbol{r} \leftarrow \partial J / \partial \bzp{\ell} = \boldsymbol{r} \odot \g^\prime (\bzp{\ell})^T \in \real^{1 \times \widths{\ell}}$ (row)
			&
			$\boldsymbol{g} \leftarrow \nabla_{\boldsymbol{a}^{(k)}} J = \boldsymbol{g} \odot f^\prime (\boldsymbol{a}^{(k)}) \in \real^{\widths{k} \times 1}$ (column)
			\\
			\hline
			$\bullet$
			Layer parameters
			&
			$\bullet$
			Layer parameters 
			\\
			$\bparamp{\ell} = \left[ \bWeight^{(\ell)} , \bbiasp{\ell} \right] \in
			\real^{\widths{\ell} \times [\widths{\ell-1}  + 1]}$
			&
			$\left[ \bWeight^{(k)} , \bbiasp{k} \right] \in
			\real^{\widths{k} \times [\widths{k-1}  + 1]}$
			\\
			\hline
			$\bullet$
			Gradient on layer parameters $\bparamp{\ell}$
			&
			$\bullet$
			Gradient on layer parameters $\left[ \bWeight^{(k)} , \bbiasp{k} \right]$
			\\
			$
			\displaystyle
			\frac{\partial J}{\partial \bparamp{\ell}} = \left[ \frac{\partial J}{\partial \bWeight^{(\ell)}}  \, | \,  \frac{\partial J}{\partial \bbiasp{\ell}}  \right]
			\in
			\real^{\widths{\ell} \times [\widths{\ell-1}  + 1]}
			$ 
			&
			$\left[ \nabla_{\bWeight^{(k)}} J \, | \, \nabla_{\boldsymbol{b}^{(k)}} J  \right] \in
			\real^{\widths{k} \times [\widths{k-1}  + 1]}$ 
			\\
			$\partial J / \partial \bparamp{\ell} = \boldsymbol{r}^{T}
			\expand{\by}^{(\ell-1) T}
			=
			\left[ \boldsymbol{r}^{T} \by^{(\ell-1) T} \, | \, \boldsymbol{r}^{T} \right]
			$ 
			&
			$\nabla_{\bWeight^{(k)}} J = \boldsymbol{g} \boldsymbol{h}^{(k-1) T}$ and
			$\nabla_{\boldsymbol{b}^{(k)}} J = \boldsymbol{g}$  
			\\
			$\partial J / \partial \bWeight^{(\ell)} = \boldsymbol{r}^{T} \by^{(\ell-1) T}$ and
			$\partial J / \partial \bbiasp{\ell} = \boldsymbol{r}^{T}$
			&
			(no regularization in cost function, $\boldsymbol{g} = \boldsymbol{r}^T$)
			\\
			\hline
			$\bullet$
			Gradient on layer outputs $\byp{\ell-1}$
			&
			$\bullet$
			Gradient on layer outputs $\boldsymbol{h}^{(k-1)}$
			\\
			$\boldsymbol{r} \leftarrow \partial J / \partial \byp{\ell-1} = \boldsymbol{r}\bWeight^{(\ell)}
			\in
			\real^{1 \times \widths{\ell -1}}$ (row)
			&
			$\boldsymbol{g} \leftarrow \nabla_{\boldsymbol{h}^{(k-1)}} J =  \bWeight^{(k) T} \boldsymbol{g}
			\in
			\real^{\widths{k -1} \times 1}$ (column)
			\\
			\hline
		\end{tabular}
	}
	\label{tb:backprop-comparison}
\end{table}

	% correct figure on LSTM in Goodfellow 2016
	\newpage
\section{Another LSTM block diagram}
\label{app:another-lstm-block-diagram}
An alternative block diagram for the folded RNN with LSTM cell corresponding to Figure~\ref{fig:our-lstm_cell} is shown Figure~\ref{fig:lstm-cell-goodfellow} below:
\begin{figure}[H]
	\centering
	%
	% 2022.12.17
	% add "-eps-converted-to.pdf" for arXiv
	% \includegraphics[width=1.0\textwidth]{figures/RNN-LSTM-correct-Goodfellow-Fig10-16-p405-B}
	\includegraphics[width=1.0\textwidth]{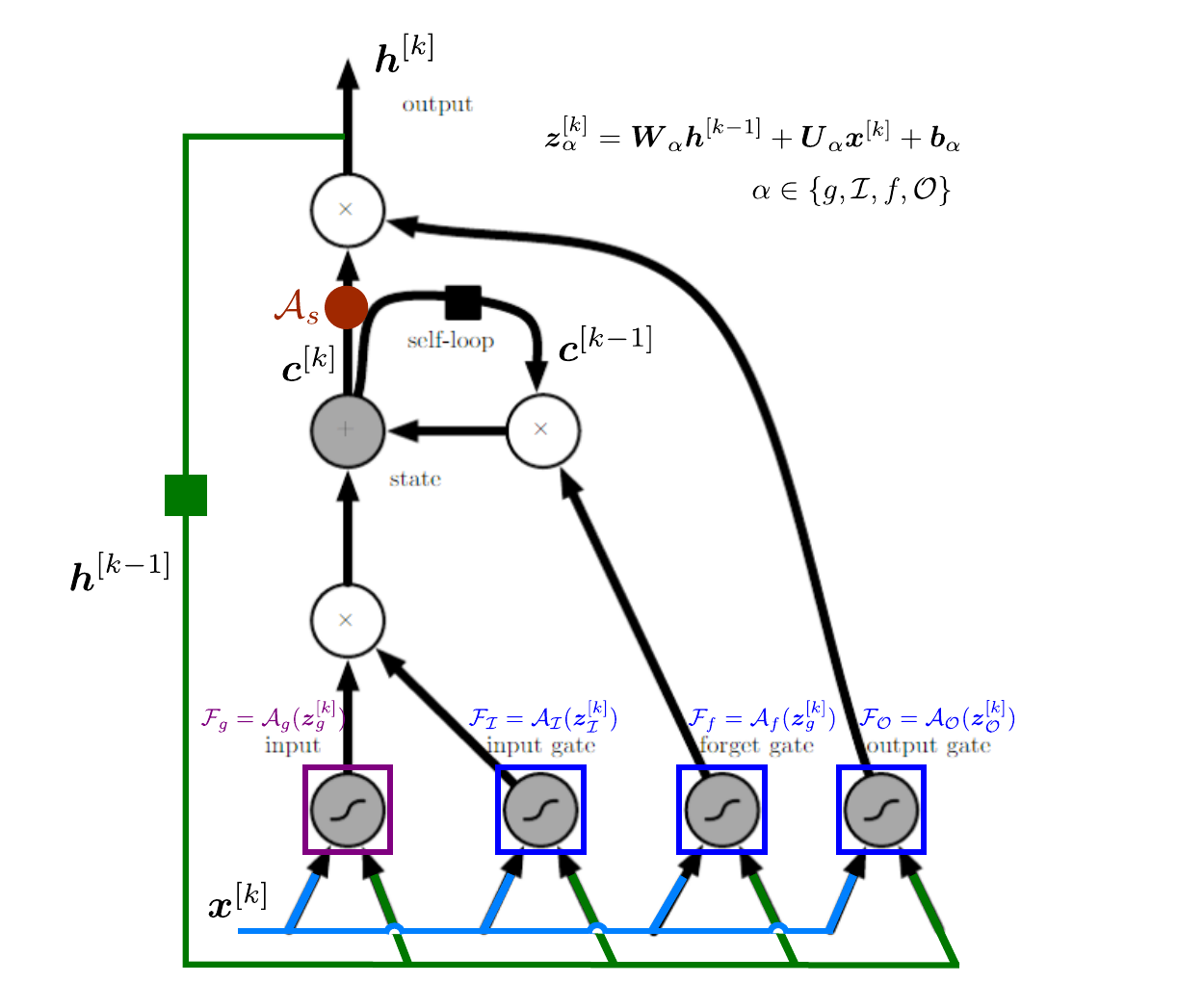}
	\caption{
		\emph{Folded RNN and LSTM cell, two feedback loops with delay, block diagram}. Typical state $[k]$. 
		Corrections for the figures in (1) Figure~10.16 in the online book \href{https://www.deeplearningbook.org}{Deep Learning} by Goodfellow et al 2016, \href{https://www.deeplearningbook.org/contents/rnn.html}{Chap.10}, p.405 (referred to here as ``DL-A'', or ``Deep Learning, version A''), and (2) \cite{Goodfellow.2016}, p.398 (referred to as ``DL-B'').
		{\footnotesize (The above figure is adapted from a figure reproduced with permission of the authors.)}
	}
	\label{fig:lstm-cell-goodfellow}
\end{figure}

Figure~10.16 in the online book \href{https://www.deeplearningbook.org}{Deep Learning} by Goodfellow et al 2016, \href{https://www.deeplearningbook.org/contents/rnn.html}{Chap.10}, p.405 (referred to here as ``DL-A'', or ``Deep Learning, version A''), was either incomplete with missing important details.  Even the \emph{updated} Figure~10.16 in \cite{Goodfellow.2016}, p.398 (referred to as ``DL-B''), was still incomplete (or incorrect).  

The corrected arrows, added annotations, and colors correspond to those in the equivalent Figure~\ref{fig:our-lstm_cell}.  The corrections are described below.

{\bf \emph{Error 1:}}
The cell state $\boldsymbol{c}^{[k]}$ should be squashed by the state sigmoidal activation function $\mathcal A_s$ with range $(-1, +1)$ (brown dot, e.g., tanh function) before being multiplied by the scaling factor $\mathcal F_{\mathcal O}$ coming from the output gate to produce the hidden state $\boldsymbol{h}^{[k]}$. 
This correction is for both DL-A and DL-B.

{\bf \emph{Error 2:}}
The hidden-state feedback loop (green) should start from the hidden state $\boldsymbol{h}^{[k]}$, delayed by one step, i.e., $\boldsymbol{h}^{[k-1]}$, which is fed to all four gates:
(1) The \emph{externally-input} gate $g$ (purple box) with activation function $\mathcal A_g$ having the range $(-1, +1)$, 
(2) the \emph{input} gate $\mathcal I$,
(3) the \emph{forget} gate $f$,
(4) the \emph{output} gate $\mathcal O$.
The activations $\mathcal A_\alpha$, with $\alpha \in \{ \mathcal I, f, \mathcal O \}$ (3 blue boxes) all have the interval $(0,1)$ as their range.
This hidden-state feedback loop was missing in DL-A, whereas the hidden-state feedback loop in DL-B incorrectly started from the summation operation (grey circle, just below $\boldsymbol{c}^{[k]}$) in the cell-state feedback loop, and did not feed into the input gate $g$.

{\bf \emph{Error 3:}}
Four pairs of arrows pointing into the four gates $\{ g, \mathcal I, f, \mathcal O \}$, with one pair per gate, were intended as inputs to these gates, but were without annotation, and thus unclear / confusing.  Here, for each gate, one arrow is used for the hidden state $\boldsymbol{h}^{[k-1]}$, and the other arrow for the input $\boldsymbol{x}^{[k]}$.
This correction is for both DL-A and DL-B.

%{\color{red} NOTE FOR US: 2019.08.20. the above corrected figure can be stored on Wikepedia, where we would refer back to our paper, as shown in the caption, which is intended for Wikipedia.   here, we can then link to the figure on Wikipedia; we don't have to keep it here.} 

	% derivation of conditional Gaussian distribution
	% derivation of conditional Gaussian distribution
% Started 2022.12.01
\section{Conditional Gaussian distribution}
\label{app:Gaussian-distribution-conditional}
The derivation of Eqs.~\eqref{eq:Gaussian-posterior-1}-\eqref{eq:Gaussian-posterior-3} is provided here helps develop a better feel for the Gaussian distribution, and facilitates the understanding of the conditional Gaussian process posterior described in Section~\ref{sc:Gaussian-posterior-sampling}.

If two sets of variables have a joint Gaussian distribution, i.e., these two sets are jointly Gaussian, then the conditional probability distribution of one set given the other set is also Gaussian.\footnote{
	See, e.g., \cite{bishop2006pattern}, p.~85.
}  The two sets of variables considered here is the observed values in $\by$ and the test values in $\byt$ in Eq.~\eqref{eq:Gaussian-posterior-0}.  Define
\begin{align}
	&
	\byh
	:=
%	\begin{Bmatrix}
%		\f (\bx)
%		\\
%		\f (\bxt)
%	\end{Bmatrix}
%	=
	\begin{Bmatrix}
		\by
		\\
		%		\bfstar
		\byt
	\end{Bmatrix}
	\ , \quad
	\bmuh
	:=
	\begin{Bmatrix}
%		\bkar{\mu} (\bx)
		\bmu
		\\
%		\bkar{\mu} (\bxt)
		\bmut
	\end{Bmatrix}
	\ , \quad
	\bkars{C}{\byh \byh}
	:=
	\begin{bmatrix}
		\bkars{C}{\by \by , \nu} & \bkars{C}{\by \byt}
		\\
		\bkars{C}{\byt \by} & \bkars{C}{\byt \byt}
	\end{bmatrix}
	=
	\begin{bmatrix}
		\bkar{K} (\bx , \bx) + \nu^2 \bkar{I} & \bkar{K} (\bx , \bxt)
		\\
		\bkar{K}^T (\bx , \bxt) & \bkar{K} (\bxt , \bxt)
	\end{bmatrix}
	\ , 
	\label{eq:Gaussian-posterior-4}
	\\
	&
	\bkarsp{C}{\byh \byh}{-1}
	:=
	\begin{bmatrix}
		\bkars{C}{\by \by , \nu} & \bkars{C}{\by \byt}
		\\
		\bkars{C}{\byt \by} & \bkars{C}{\byt \byt}
	\end{bmatrix}^{-1}
	=
	\begin{bmatrix}
		\bkars{D}{\by \by} & \bkars{D}{\by \byt}
		\\
		\bkarsp{D}{\by \byt}{T} & \bkars{D}{\byt \byt}
	\end{bmatrix}
	\ , \text{ with }
	\bkarsp{D}{\by \byt}{T} = \bkars{D}{\byt \by}
	\ ,
	\label{eq:Gaussian-posterior-5}
\end{align} 
%$\bkarh{y} := \left[\by , \byt \right]^T$ and $\bkarh{\mu} := \left[\bkar{\mu} , \bkart{\mu} \right]$, 
then expand the exponent in the Gaussian joint probability Eq.~\eqref{eq:multivariate-gaussian-distribution-1}, with $(\bkarh{y} - \bkarh{\mu})$ instead of $(\by - \bkar{\mu})$, to have the Mahalanobis distance\footnote{
	See, e.g., \cite{bishop2006pattern}, p.~80.
} $\Delta$ squared, a quadratic form in terms of $(\byh - \bmuh)$, written as:
\begin{align} 
	\Delta^2
	:=
	\left(\byh - \bkarh{\mu} \right)^T \bkarsp{C}{\byh \byh}{-1}  \left(\byh - \bkarh{\mu} \right)
	&
	=
	(\by - \bmu)^T \bkars{D}{\by \by} (\by - \bmu)
	+ 
	(\by - \bmu)^T \bkars{D}{\by \byt}  (\byt - \bmut)
	\nonumber
	\\
	&
	+ 
	(\byt - \bmut)^T \bkars{D}{\byt \by}  (\by - \bmu)
	+ 
	(\byt - \bmut)^T \bkars{D}{\byt \byt}  (\byt - \bmut)
	\ ,
	\label{eq:Mahalanobis-expanded-1}
\end{align}
which is also a quadratic form in terms of $(\byt - \bmut)$, based on the symmetry of $\bkarsp{C}{\byh \byh}{-1}$, the inverse of the covariance matrix, in Eq.~\eqref{eq:Gaussian-posterior-5}, implying that the distribution of the test values $\byt$ is Gaussian.  The covariance matrix and the mean of the Gaussian distribution over $\byt$ are determined by identifying the quadratic term and the linear term in $\byt$, compared to the expansion of the general Gaussian distribution $\mathcal{N} (\bz | \bkar{m}, \bkar{C})$ over the variable $\bz$ with mean $\bkar{m}$ and covariance matrix $\bkar{C}$:
\begin{align}
	\left(\bz - \bkar{m} \right)^T \bkarp{C}{-1}  \left(\bz - \bkar{m} \right)
	=
	\bz^T \bkarp{C}{-1} \bz - 2 \bz^T \bkarp{C}{-1} \bkar{m}
	+
	\text{ constant}
	\ ,
	\label{eq:Mahalanobis-expanded-2}
\end{align}
where the constant is independent of $\bz$.   Expand Eq.~\eqref{eq:Mahalanobis-expanded-1} to have
\begin{align}
	\Delta^2 = \byt^T \bkars{D}{\byt \byt} \byt
	- 2 \byt^T 
	\left[ 
	\bkars{D}{\byt \byt} \bmut - \bkars{D}{\byt \by}(\by - \bmu)
	\right]
	+ \text{ constant}
	\ ,
\end{align}
and compare to Eq.~\eqref{eq:Mahalanobis-expanded-2}, then  
for the conditional distribution $p(\byt | \by)$ of the test values $\byt$ at $\bxt$ given the data $(\bx , \by)$,
the covariance matrix $\bkars{C}{\byt | \by}$ and the mean $\bkars{\mu}{\byt | \by}$ 
are
\begin{align}
	&
	\bkarsp{C}{\byt | \by}{-1}
	=
	\bkars{D}{\byt \byt}
	\Rightarrow
	\bkars{C}{\byt | \by}
	=
	\bkarsp{D}{\byt \byt}{-1}
%	&
	\label{eq:conditional-distribution-covariance}
	\\
	&
%	\ , \quad
	\bkarsp{C}{\byt | \by}{-1} \bkars{\mu}{\byt | \by}
	=
	\left[ 
	\bkars{D}{\byt \byt} \bmut - \bkars{D}{\byt \by}(\by - \bmu)
	\right]
	\Rightarrow
%	&
	\bkars{\mu}{\byt | \by} 
	=
	\bkars{C}{\byt | \by}
	\left[ 
	\bkars{D}{\byt \byt} \bmut - \bkars{D}{\byt \by}(\by - \bmu)
	\right]
	\ ,
	\\
	&
	\phantom{\bkarsp{C}{\byt | \by}{-1} \bkars{\mu}{\byt | \by}
	=
	\left[ 
	\bkars{D}{\byt \byt} \bmut - \bkars{D}{\byt \by}(\by - \bmu)
	\right]}
	\Rightarrow
	\bkars{\mu}{\byt | \by} 
	= 
	\bmut - \bkarsp{D}{\byt \byt}{-1} \bkars{D}{\byt \by}(\by - \bmu)
	\ ,
	\label{eq:conditional-distribution-mean}
\end{align}
in which Eq.~\eqref{eq:conditional-distribution-covariance}$_2$ had been used.

At this point, the submatrices $\bkars{D}{\byt \byt}$ and $\bkars{D}{\byt \by}$ can be expressed in terms of the submatrices of the partitioned matrix $\bkars{C}{\byh \byh}$ in Eq.~\eqref{eq:Gaussian-posterior-4} as follows.  From the definition of the matrix $\bkars{D}{\byh \byh}$, the inverse of the covariance matrix $\bkars{C}{\byh \byh}$:
\begin{align}
	\bkars{D}{\byh \byh}
	\bkars{C}{\byh \byh}
	=
	\begin{bmatrix}
		\bkars{D}{\by \by} & \bkars{D}{\by \byt}
		\\
		\bkars{D}{\byt \by} & \bkars{D}{\byt \byt}
	\end{bmatrix}
	\begin{bmatrix}
		\bkars{C}{\by \by} & \bkars{C}{\by \byt}
		\\
		\bkars{C}{\byt \by} & \bkars{C}{\byt \byt}
	\end{bmatrix}
	=
	\begin{bmatrix}
		\bkar{I} & \bkar{0}
		\\
		\bkar{0} & \bkar{I}
	\end{bmatrix}
	\label{eq:covariance-inverse-1}
\end{align}
the 2nd row gives rise to a system of two equations for two unknowns $\bkars{D}{\byh \by}$ and $\bkars{D}{\byh \byh}$:
\begin{align}
	\begin{bmatrix}
		\bkars{D}{\byt \by} & \bkars{D}{\byt \byt}
	\end{bmatrix}
	\begin{bmatrix}
		\bkars{C}{\by \by} & \bkars{C}{\by \byt}
		\\
		\bkars{C}{\byt \by} & \bkars{C}{\byt \byt}
	\end{bmatrix}
	=
	\begin{bmatrix}
		\bkar{0} & \bkar{I}
	\end{bmatrix}
	\ ,
	\label{eq:covariance-inverse-2}
\end{align}
in which the covariance matrix $\bkars{C}{\byh \byh}$ is symmetric, and is a particular case of the non-symmetric problem of expressing $(\bkar{F} , \bkar{G})$ in terms of $(\bkar{P} , \bkar{Q} , \bkar{R} , \bkar{S})$:
\begin{align}
	\begin{bmatrix}
		\bkar{F} & \bkar{G}
	\end{bmatrix}
	\begin{bmatrix}
		\bkar{P} & \bkar{Q}
		\\
		\bkar{R} & \bkar{S}
	\end{bmatrix}
	=
	\begin{bmatrix}
		\bkar{0} & \bkar{I}
	\end{bmatrix}
	\Rightarrow
%	\bkar{F} \bkar{P} + \bkar{G} \bkar{R} = \bkar{0}
	\bkar{G} = - \bkar{F} \bkar{P} \bkarp{R}{-1}
	\Rightarrow
	\bkarp{G}{-1} \bkar{F} = - \bkar{R} \bkarp{P}{-1}
	\ ,
	\label{eq:covariance-inverse-submatrix-21-22-a}
\end{align}
from the first equation, and leads to
\begin{align}
	&
	\bkar{F} \bkar{Q} + \bkar{G} \bkar{S} = \bkar{I}
%	&
	\Rightarrow
	\bkar{F} \left[ \bkar{Q} - \bkar{P} \bkarp{R}{-1} \bkar{S} \right] = \bkar{I}
	\ ,
	\\
	&
	\Rightarrow
	\bkar{F} = \left[ \bkar{Q} - \bkar{P} \bkarp{R}{-1} \bkar{S} \right]^{-1}
	\ , \quad
	\bkar{G} = - \left[ \bkar{Q} - \bkar{P} \bkarp{R}{-1} \bkar{S} \right]^{-1} \bkar{P} \bkarp{R}{-1}
	\ , \quad
	\bkarp{G}{-1} = \bkar{S} - \bkar{R} \bkarp{P}{-1} \bkar{Q}
	\ ,
	\label{eq:covariance-inverse-submatrix-21-22-b}
\end{align}
which, after using Eq.~\eqref{eq:covariance-inverse-2} to identify $(\bkar{F}, \bkar{G})  = (\bkars{D}{\byt \by} , \bkars{D}{\byt \byt})$ and $(\bkar{P}, \bkar{Q}, \bkar{R}, \bkar{S}) = (\bkars{C}{\by \by}, \bkars{C}{\by \byt}, \bkarsp{C}{\by \byt}{T}$, $\bkars{C}{\byt \byt})$, 
and then when replaced in Eq.~\eqref{eq:conditional-distribution-covariance} for the conditional covariance $\bkars{C}{\byt | \by}$ and Eq.~\eqref{eq:conditional-distribution-mean} for the conditional mean $\bkars{\mu}{\byt | \by}$, yields Eq.~\eqref{eq:Gaussian-posterior-3} and Eq.~\eqref{eq:Gaussian-posterior-2}, respectively.

\begin{rem}
	{\rm
	Another way to obtain indirectly Eq.~\eqref{eq:Gaussian-posterior-3} and Eq.~\eqref{eq:Gaussian-posterior-2}, without derivation, is to use the identity of the inverse of a partitioned matrix in Eq.~\eqref{eq:inverse-partition-matrix}, as done in \cite{bishop2006pattern}, p.~87:
	\begin{align}
		\begin{bmatrix}
			\bkar{P} & \bkar{Q}
			\\
			\bkar{R} & \bkar{S}
		\end{bmatrix}^{-1}
		=
	%	\begin{bmatrix}
	%		\bkar{M} & - \bkar{M} \bkar{Q} \bkarp{S}{-1}
	%		\\
	%		- \bkarp{S}{-1} \bkar{R} \bkar{M} & \bkarp{S}{-1} + \bkarp{S}{-1} \bkar{R} \bkar{M} \bkar{Q} \bkarp{S}{-1}
	%	\end{bmatrix}
		\left[
			\begin{array}{ll}
				\bkar{M} & - \bkar{M} \bkar{Q} \bkarp{S}{-1}
				\\
				- \bkarp{S}{-1} \bkar{R} \bkar{M} & \bkarp{S}{-1} + \bkarp{S}{-1} \bkar{R} \bkar{M} \bkar{Q} \bkarp{S}{-1}
			\end{array}
		\right]
		\ , \quad
		\bkar{M} := \left[ \bkar{P} - \bkar{Q} \bkarp{S}{-1} \bkar{R} \right]^{-1}
		\ .
		\label{eq:inverse-partition-matrix}
	\end{align}
	This method is less satisfactory since without derivation, there is no feel of where the matrix elements in Eq.~\eqref{eq:inverse-partition-matrix} came from.
	In fact, the derivation of the 1st row of Eq.~\eqref{eq:inverse-partition-matrix} follows exactly the same line as for Eqs.\eqref{eq:covariance-inverse-submatrix-21-22-a}-\eqref{eq:covariance-inverse-submatrix-21-22-b}.  The 2nd row in Eq.~\eqref{eq:inverse-partition-matrix} looks complex, 
%	and its derivation is less obvious.  
	but before getting into its derivation, we note that exactly the same line of derivation for the 1st row can be straightforwardly followed to arrive at different, and simpler, expressions of the 2nd-row matrix elements $(2,1)$ and $(2,2)$, which are similar to those in the 1st row, and which were already derived in Eq.~\eqref{eq:covariance-inverse-submatrix-21-22-b}. 
	
	It can be easily verified that
%	\begin{align}
%		&
%		\left[
%			\bkar{M} \quad - \bkar{M} \bkar{Q} \bkarp{S}{-1}
%		\right]
%		\begin{bmatrix}
%			\bkar{P}
%			\\
%			\bkar{R}
%		\end{bmatrix}
%		=
%		\bkar{M} \bkar{P} - \bkar{M} \bkar{Q} \bkarp{S}{-1} \bkar{R}
%		=
%		\bkar{M} \bkarp{M}{-1}
%		=
%		\bkar{I}
%		\ ,
%		\\
%		&
%		\left[
%		\bkar{M} \quad - \bkar{M} \bkar{Q} \bkarp{S}{-1}
%		\right]
%		\begin{bmatrix}
%			\bkar{Q}
%			\\
%			\bkar{S}
%		\end{bmatrix}
%		=
%		\bkar{M} \bkar{Q} - \bkar{M} \bkar{Q} \bkarp{S}{-1} \bkar{S}
%		=
%		\bkar{0}
%		\ ,
%	\end{align}
%	and similarly for the second row.
	\begin{align}
		\left[
		\begin{array}{ll}
			\bkar{M} & - \bkar{M} \bkar{Q} \bkarp{S}{-1}
			\\
			- \bkarp{S}{-1} \bkar{R} \bkar{M} & \bkarp{S}{-1} + \bkarp{S}{-1} \bkar{R} \bkar{M} \bkar{Q} \bkarp{S}{-1}
		\end{array}
		\right]
		\begin{bmatrix}
			\bkar{P} & \bkar{Q}
			\\
			\bkar{R} & \bkar{S}
		\end{bmatrix}
		=
		\begin{bmatrix}
			\bkar{I} & \bkar{0}
			\\
			\bkar{0} & \bkar{I}
		\end{bmatrix}
		\ .
		\label{eq:inverse-partition-matrix-verified}
	\end{align}
	To derive the 2nd row of Eq.~\eqref{eq:inverse-partition-matrix}, premultiply the 1st row (which had been derived as mentioned above) of Eq.~\eqref{eq:inverse-partition-matrix-verified} by $(-\bkarp{S}{-1} \bkar{R})$ to have
	\begin{align}
		(-\bkarp{S}{-1} \bkar{R})
		\begin{bmatrix}
			\bkar{M} & -\bkar{M} \bkar{Q} \bkarp{S}{-1}
		\end{bmatrix}
		\begin{bmatrix}
			\bkar{P} & \bkar{Q}
			\\
			\bkar{R} & \bkar{S}
		\end{bmatrix}
		=
		(-\bkarp{S}{-1} \bkar{R})
		\begin{bmatrix}
			\bkar{I} & \bkar{0}
		\end{bmatrix}
		=
		\begin{bmatrix}
			(-\bkarp{S}{-1} \bkar{R}) & \bkar{0}
		\end{bmatrix}
		\ .
		\label{eq:inverse-partition-matrix-2nd-row}
	\end{align}
	To make the right-hand side become 
	$\left[\bkar{0} \quad \bkar{I} \right]$, add to both sides of Eq.~\eqref{eq:inverse-partition-matrix-2nd-row} the matrix 
	$\left[ (\bkarp{S}{-1} \bkar{R})  \quad \bkar{I} \right]$ to obtain
%	the 2nd row of Eq.~\eqref{eq:inverse-partition-matrix-verified}.
	Eq.~\eqref{eq:inverse-partition-matrix-verified}'s 2nd row, whose complex expressions did not contribute to the derivation of the conditional Gaussian posterior mean Eq.~\eqref{eq:Gaussian-posterior-2} and covariance Eq.~\eqref{eq:Gaussian-posterior-3}.

	Yet another way to derive Eqs.~\eqref{eq:Gaussian-posterior-2}-\eqref{eq:Gaussian-posterior-3} is to use the more complex proof in \cite{mises1964mathematical}, p.~429, which was referred to in \cite{rasmussen2006gaussian}, p.~200 (see also Footnote~\ref{fn:Gaussian-distribution-conditional}).
	}
	$\hfill\blacksquare$
\end{rem}

In summary, the above derivation is simpler and more direct than in \cite{bishop2006pattern}, p.~87, and in \cite{mises1964mathematical}, p.~429.

%{\color{red} NOTE: 2022.12.01 - I am HERE.}

	% cybernetics
	
\begin{figure}[h]
	\centering
	\includegraphics[width=0.7\linewidth]{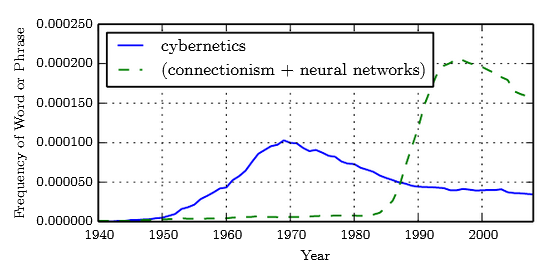}
	\caption{
		\emph{The first two waves of AI}, according to \cite{Goodfellow.2016}, p.13, showing the ``cybernetics'' wave (blue line) started in the 1940s peaked before 1970, then gradually declined toward 2006 and beyond.  The results were based on a search for frequency of words in Google Books.  It was mentioned, incorrectly, that the work of Rosenblatt
		%		 (\citeyear{Rosenblatt.1957}-\citeyear{Rosenblatt.1962}) 
		(1957-1962) \cite{Rosenblatt.1957}-\cite{Rosenblatt.1962}
		was limited to one neuron; see Figure~\ref{fig:network-size-time} and Figure~\ref{fig:Rosenblatt-Mark-I}.
		{\footnotesize (Figure reproduced with permission of the authors.)}
	}
	\label{fig:waves}
\end{figure}

%\newpage
%\subsection{Three waves of AI}
\section{The ups and downs of AI, cybernetics}
\label{sc:3wavesAI}
\label{app:cybernetics}

%{\color{red} NOTE: 2022.08.31 - I am HERE to correct, to rewrite for CMES style, and to link section numbers to figure / table captions.}

%
% CMES style rewriting
The authors of
\cite{Goodfellow.2016}, p.13, divided the wax-and-wane fate of AI into three waves, with the first wave called the ``cybernetics'' that started in the 1940s, peaked before 1970, then began a gradual descent toward 1986, when the second wave picked up with the publication of \cite{Rumelhart.1986} on an application of backpropagation to psychology; see Section~\ref{sc:backprop-history} on a history of backpropagation.
Since Goodfellow (the first author of \cite{Goodfellow.2016}) worked at Google at the time, and would have access to the scanned books in the Google Books collection to do the search.   For a concise historical account of ``cybernetics'', see \cite{Sepulchre.2020}.

\begin{figure}[h]
	\centering
	\includegraphics[width=1.0\linewidth]{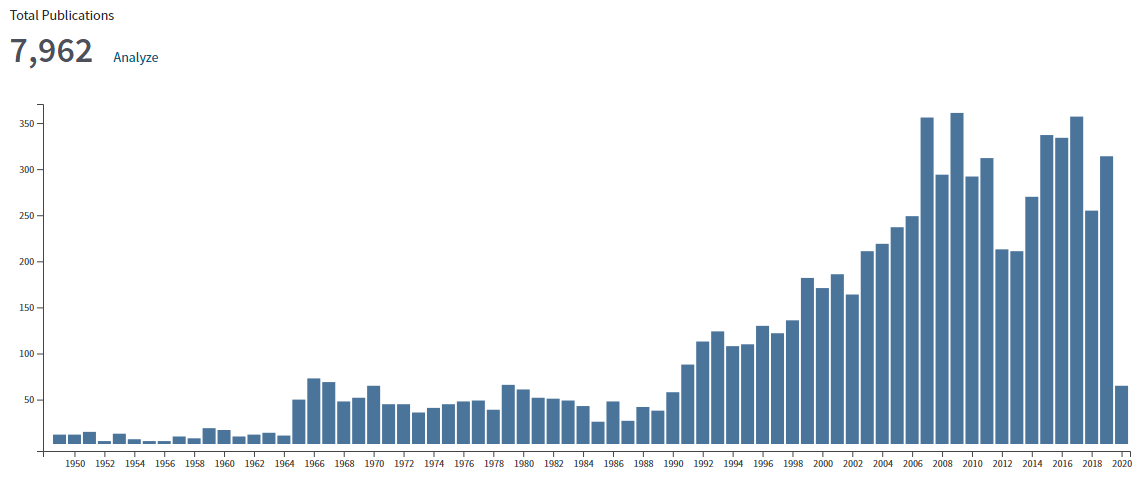}
	\caption{
		\emph{Cybernetics papers}, (Appendix~\ref{sc:3wavesAI}).  Web of Science search on 2020.04.15, having more than 100 Web of Science categories.  The first paper was \cite{Copeland.1949}.
		There was no clear wave that crested before 1970, but actually the number of papers in Cybernetics continue to increase over the years.
	}
	\label{fig:W0S-cybernetics-1}
\end{figure}

We had to rely on Web of Science to do the ``topic'' search for the keyword ``cyberneti*'', i.e., using the query {\tt ts=(cyberneti*)}, with ``*'' being the search wildcard, which can stand for any character that follows.
Figure~\ref{fig:W0S-cybernetics-1} is the result,\footnote{
	\label{fn:WoS-search-cybernetics-1}
	The total number of papers on the topic ``cyberneti*''  was 7,962 on 2020.04.15---as shown in Figure~\ref{fig:W0S-cybernetics-1} obtained upon clicking on the ``Citation Report'' button in the Web of Science---and 8,991 on 2022.08.08.  Since the distribution in Figure~\ref{fig:W0S-cybernetics-1}, the points made in the figure caption and in this section remain the same, there was no need to update the figure to its 2022.08.08 version. 
}  spanning an astoundingly vast and diverse number of more than 100 categories,\footnote{
	\label{fn:WoS-search-cybernetics-2}
	The number of categories has increased to 244 in the Web of Sciecne search on 2022.08.08, mentioned in Footnote~\ref{fn:WoS-search-cybernetics-1}, with the number of papers in Computer Science Cybernetics at 2,952, representing 32\% of the 8,991 papers in this topic.
} listed in descending order of number of papers in parentheses: Computer Science Cybernetics (2,665 papers), Computer Science Artificial Intelligence (601), Engineering Electrical Electronic (459), ..., Philosophy (229), ..., Social Sciences Interdisciplinary (225), ..., Business (132), ..., Psychology Multidisciplinary (128), ..., Psychiatry (90), ..., Art (66), ..., Business Finance (43), ...,  Music (31), ..., Religion (27), ..., Cell biology (21), ..., Law (21), ...  

The first 
%
% CMES style rewriting
%paper by \cite{Copeland.1949} 
paper in 1949 \cite{Copeland.1949}
was categorized as Mathematics.   More recent papers include Biological Science, e.g., \cite{Chavalarias.2020}, Building Construction, e.g., \cite{Togashi.2020}, Accounting, e.g., \cite{Jube.2020}.

It is interesting to note that McCulloch who co-authored the well-known paper \cite{McCulloch.1943} was part of the original cybernetics movement that started in the 1940s, as noted 
%
% CMES style rewriting
%by
in 
\cite{Kline.2015}:
\begin{quote}
	``Warren McCulloch, the ``chronic chairman'' and founder of the cybernetics conferences.\footnote{
		These cybernetics conferences were called the Macy conferences, held during a short period from 1946 to 1953, and involved researchers from diverse fields: not just mathematics, physics, engineering, but also anthropology and physiology, \cite{Kline.2015}, pp.2-3.
	} An eccentric physiologist, McCulloch had coauthored a foundational article of cybernetics on the 
	brain's neural network.''
\end{quote}
But 
McCulloch \& Pitt's 1943 
paper \cite{McCulloch.1943}---often cited in artificial-neural-network papers (e.g., \cite{Ghaboussi.1991:rd0001}, \cite{Schmidhuber.2015:rd0001}) and books (e.g., \cite{Goodfellow.2016}), and dated six years before \cite{Copeland.1949}---was placed in the Web of Science category ``Biology; Mathematical \& Computational Biology,'' and thus did not show up in the search with keyword ``cyberneti*'' shown in Figure~\ref{fig:W0S-cybernetics-1}.
A reason is \cite{McCulloch.1943} did not contain the word ``cybernetics,'' which was not invented until 1948 with the famous book by Wiener, and which was part of the title of \cite{Copeland.1949}.  Cybernetics was a ``new science'' with a ``mysterious name and universal aspirations'' \cite{Kline.2015}, p.5.

\begin{quote}
	``What exactly is (or was) cybernetics? This has been a perennial ongoing topic of debate within the American Society for Cybernetics throughout its 50-year history. ... the word has a much older history reaching back to Plato, Amp\`ere (``Cybern\'etique = the art of growing''), and others. ``Cybernetics'' comes from the Greek word for governance, \emph{kybernetike}, and the related word, \emph{kybernetes}, steersman or captain''
	\cite{Cariani.2017}.
\end{quote}

\begin{figure}[h]
	\centering
	% 5 computer science categories
	%\includegraphics[width=1.0\linewidth]{figures/WoS-cybernetics-computer-science-ALL-2020-04-17}
	% 6 computer science categories
	\includegraphics[width=1.0\linewidth]{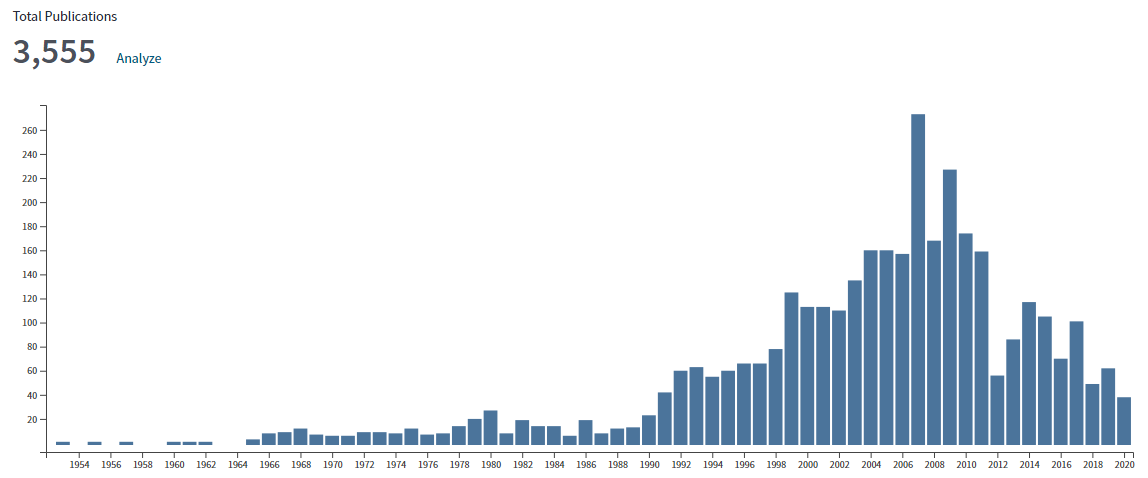}
	\caption{
		\emph{Cybernetics papers}, (Appendix~\ref{sc:3wavesAI}).  Web of Science search on 2020.04.17, ALL Computer-Science categories (3,555 papers): Cybernetics (2,666), Artificial Intelligence (602), Information Systems (432), Theory Methods (300), Interdisciplinary Applications (293), Software Engineering (163).  The wave crest was in 2007, with a tiny bump in 1980.
	}
	\label{fig:W0S-cybernetics-2a}
\end{figure}
Steering a ship is controlling its direction.
\cite{Eisenhart.1949} defined cybernetics as
\begin{quote}
	``... (feedback) control and communication theory pertinent
	to the description, analysis, or construction of systems that involve 
	(1) mechanisms (receptors) for the reception of messages or stimuli, 
	(2) means (circuits) for communication of these to 
	(3) a central control unit that responds by feeding back through the system
	(4) instructions that (will or tend to) produce specific
	actions on the part of (5) particular elements (effectors) of the system. ...   
	The central concept in cybernetics is a feedback
	mechanism that, in response to information (stimuli,
	messages) received through the system, feeds back to
	the system instructions that modify or otherwise alter
	the performance of the system.''
\end{quote}
Even though \cite{Cariani.2017} did not use the word ``control'', the definition is similar:
\begin{quote}
	``The core concepts involved natural and artificial systems organized to attain internal stability (homeostasis), to adjust internal structure and behavior in light of experience (adaptive, self-organizing systems), and to pursue autonomous goal-directed (purposeful, purposive) behavior.''
	\cite{Cariani.2017}
\end{quote}
and is succinctly summarized by \cite{WEH.1949}:
\begin{quote}
	``If ``cybernetics'' means ``control and communication,'' what  does it not mean?  It would  be difficult to think of any process in which nothing is either controlled or communicated.''
\end{quote} 
which is the reason why cybernetics is found in a large number of different fields.  \cite{Kline.2015}, p.4, offered a similar, more detailed explanation of cybernetics as encompassing all fields of knowledge:
\begin{quote}
	``Wiener and Shannon defined the amount of information transmitted in communications systems with a formula mathematically equivalent to entropy (a measure of the degradation of energy). Defining information in terms of one of the pillars of physics convinced many re searchers that information theory could bridge the physical, biological, and 
	social sciences. The allure of cybernetics rested on its promise to model mathematically the purposeful behavior of all organisms, as well as inanimate systems. Because cybernetics included information theory in its purview, its proponents thought it was more universal than Shannon's theory, that it 
	applied to all fields of knowledge.''
\end{quote}

\begin{figure}[h]
	\centering
	\includegraphics[width=0.9\linewidth]{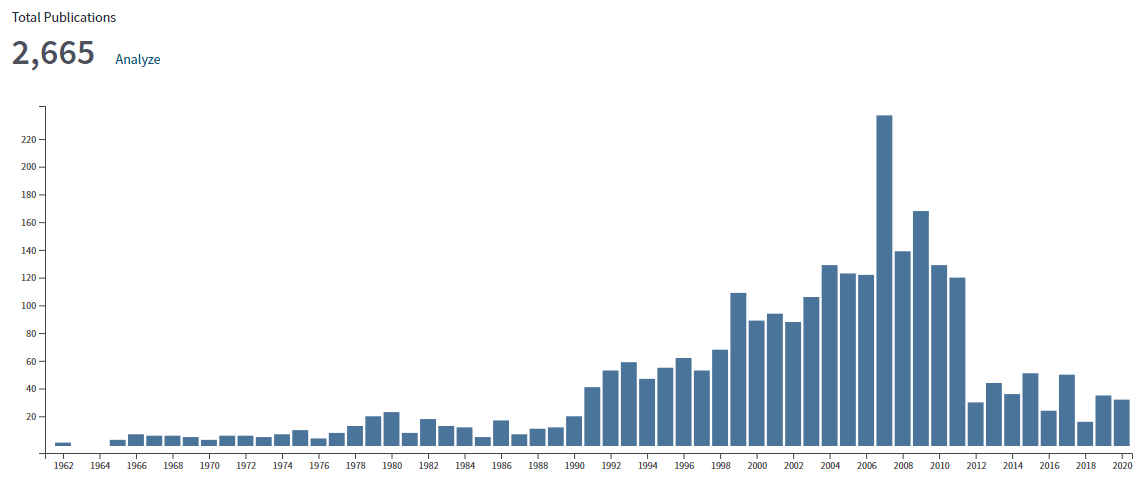}
	\caption{
		\emph{Cybernetics papers}, (Appendix~\ref{sc:3wavesAI}).  Web of Science search on 2020.04.15 (two days before Figure~\ref{fig:W0S-cybernetics-2a}), category Computer Science Cybernetics (2,665 papers).
		The wave crest was in 2007, with a tiny bump in 1980.
	}
	\label{fig:W0S-cybernetics-2b}
\end{figure}

\begin{figure}[h]
	\centering
	\includegraphics[width=0.9\linewidth]{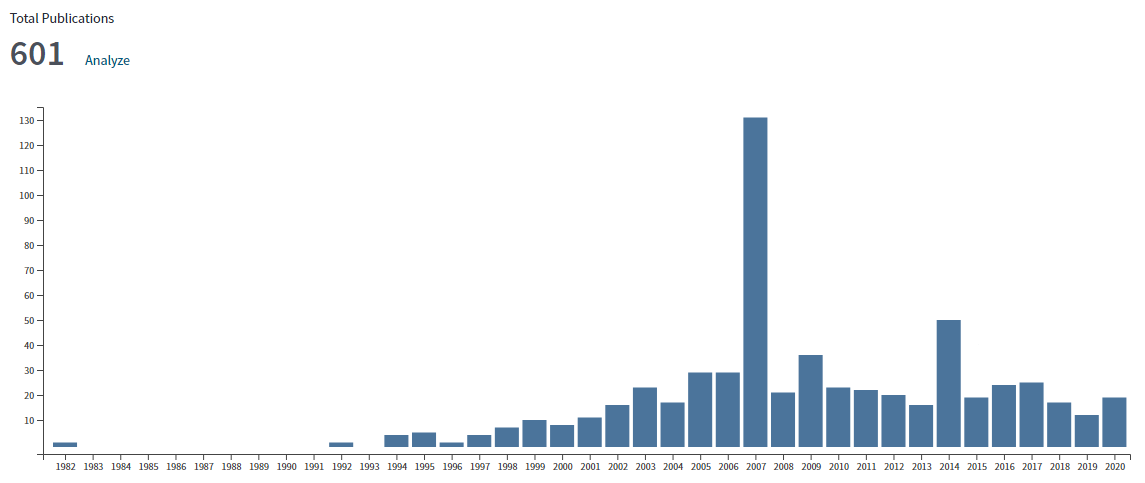}
	\caption{
		\emph{Cybernetics papers}, (Appendix~\ref{sc:3wavesAI}).  Web of Science search on 2020.04.15  (two days before Figure~\ref{fig:W0S-cybernetics-2a}), category Computer Science Artificial Intelligence (601 papers).
		Similar to Figure~\ref{fig:W0S-cybernetics-2b}, the wave crest was in 2007, but with no tiny bump in 1980, since the first paper was in 1982.    
	}
	\label{fig:W0S-cybernetics-3}
\end{figure}

% 2022.12.07
% the figure below was a screenshot of the figure further below, which 
% was drawn with tikz and resized with \resizebox.
% both methods yielded the same quality to naked eyes.
%
%\begin{figure}[h]
%	\centering
%	%	\includegraphics[width=1.0\linewidth]{figures/WoS-cybernetics-computer-AI-2020-04-16}
%	\includegraphics[width=0.9\linewidth]{figures/AI-cybernetics.png}
%	\caption{{\em Artificial Intelligence} (AI), {\em Machine Learning} (ML), and {\em Deep Learning} (DL). {\em Cybernetics} is broad and encompasses many fields, including AI.  
%		See also Figure~\ref{fig:AI.ML.DL}.
%		% Section~\ref{sc:history} on Historical perspective.
%	}
%	\label{fig:cybernetics}
%\end{figure}

% 2022.12.07
% the figure below was drawn with tikz and resized with \resizebox
% and is the same as the figure above using a screenshot of the figure
% below.
% both methods yielded the same quality to naked eyes.
%
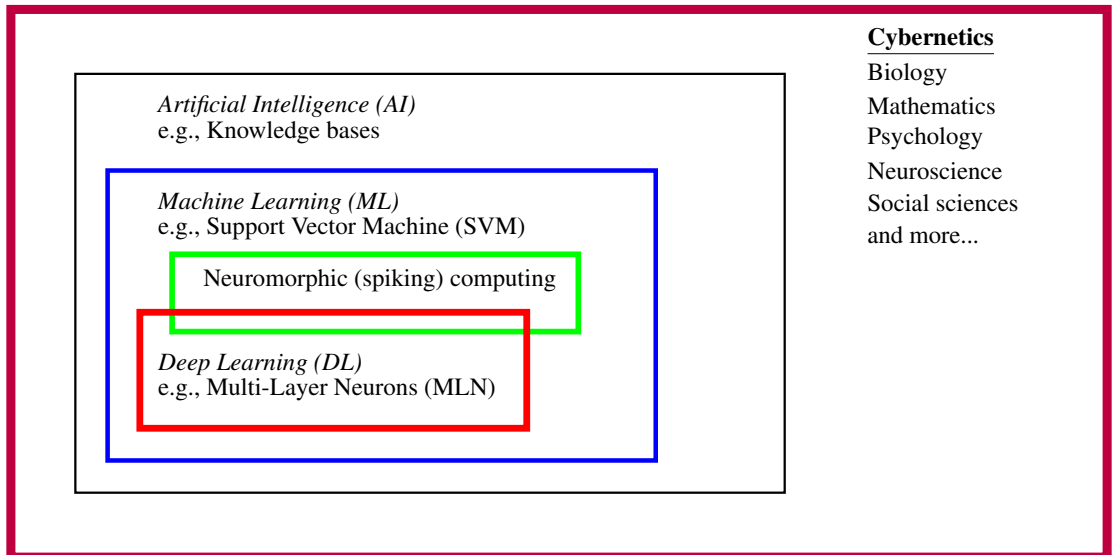
\begin{figure}[h]
	\centering
	% tikz tanh
	% https://tex.stackexchange.com/questions/176101/plotting-the-graph-of-hyperbolic-tangent
	% tikz plot function axes same scale
	% https://tex.stackexchange.com/questions/411465/same-scale-on-the-axes-but-different-lengths-of-the-axes/411472
	% tikz plot function scale figure
	% https://tex.stackexchange.com/questions/105570/how-to-plot-functions-like-x-fy-using-tikz
	% added "scale = 1.5" in the preamble of the axis environment to scale up the figure
	% added "smooth" so that the curve is smooth instead of piecewise linear, and having kinks
	
	% sigmoidal logistic function
	% need to use the following tikz libraries (add calc)
	% \usetikzlibrary{arrows,intersections,calc}
	%
	% start \resizebox
	\resizebox{0.90\linewidth}{!}{
	\begin{tikzpicture}[
	thick,
	>=stealth',
	dot/.style = {
		draw,
		fill = white,
		circle,
		inner sep = 0pt,
		minimum size = 4pt
	}
	]
	
	% artificial intelligence
	\draw[black, line width = 1 pt] (0,0) rectangle (11,6.5);
	\draw (1,6.0) node[label = {right:{\em Artificial Intelligence (AI)}}] {};
	\draw (1,5.6) node[label = {right:e.g., Knowledge bases}] {};
	
	% machine learning
	\draw[blue, line width = 2 pt] (0.5,0.5) rectangle (9,5.0);
	\draw (1,4.5) node[label = {right:{\em Machine Learning (ML)}}] {};
	% \draw (1,4.1) node[label = {right:e.g., Logistic regression}] {};
	\draw (1,4.1) node[label = {right:e.g., Support Vector Machine (SVM)}] {};
	
	% in between machine learning and deep learning
	\draw[green, line width = 2.5 pt] (1.5,2.5) rectangle (7.8,3.7);
	\draw (1.7,3.3) node[label = {right:Neuromorphic (spiking) computing}] {};
	
	% deep learning
	\draw[red, line width = 3 pt] (1.0,1.0) rectangle (7,2.8);
	\draw (1,2.0) node[label = {right:{\em Deep Learning (DL)}}] {};
	\draw (1,1.6) node[label = {right:e.g., Multi-Layer Neurons (MLN)}] {};
	
	% deep learning
	%\draw[red, line width = 3 pt] (1.0,1.0) rectangle (7,3.5);
	%\draw (1,2.5) node[label = {right:{\em Deep Learning (DL)}}] {};
	%\draw (1,2.1) node[label = {right:e.g., Multi-Layer Neurons (MLN)}] {};
	
	% cybernetics
	%\draw[green, line width = 4 pt] (-1.0,-1.0) rectangle (16,7.5);
	\draw[purple, line width = 4 pt] (-1.0,-1.0) rectangle (16,7.5);
	\draw (12.0,7.0) node[label = {right:\underline{\bf Cybernetics}}] {};
	\draw (12.0,6.5) node[label = {right:{Biology}}] {};
	%\draw (12.0,6.0) node[label = {right:{Engineering}}] {};
	\draw (12.0,6.0) node[label = {right:{Mathematics}}] {};
	\draw (12.0,5.0) node[label = {right:{Neuroscience}}] {};
	\draw (12.0,5.5) node[label = {right:{Psychology}}] {};
	\draw (12.0,4.5) node[label = {right:{Social sciences}}] {};
	\draw (12.0,4.0) node[label = {right:{and more...}}] {};
	% \draw (1,2.1) node[label = {right:e.g., Multi-Layer Neurons (MLN)}] {};
	
	\end{tikzpicture}
	} % end \resizebox

	\caption{{\em Artificial Intelligence} (AI), {\em Machine Learning} (ML), and {\em Deep Learning} (DL). {\em Cybernetics} is broad and encompasses many fields, including AI.  
	See also Figure~\ref{fig:AI.ML.DL}.
		% Section~\ref{sc:history} on Historical perspective.
	}
	\label{fig:cybernetics}
\end{figure}

In 1969, the then president of the International Association of Cybernetics asked ``But after all what is cybernetics? Or rather what is it not, for paradoxically the more people talk about cybernetics the less they seem to agree on a definition,'' then identified several meanings: A mathematical control theory, automation, computerization, communication theory, study of human-machine analogies, philosophy explaining the mysteries of life! \cite{Kline.2015}, p.5. 

So was there a first wave in AI called ``cybernetics'' ?
Back in Oct 2018, we conveyed our search result at that time---which was similar to Figure~\ref{fig:W0S-cybernetics-1}, but clearly did not support the existence of the cybernetics wave shown in Figure~\ref{fig:waves}---to Y. Bengio of \cite{Goodfellow.2016}, who then replied: 
\begin{quote}
	``Ian [Goodfellow] did those figures, but my take on your observations is that the later surge in 'new cybernetics' does not have much more to do with artificial neural networks. I'm not sure why the Google Books search did not catch that usage, though.''
\end{quote}  
We then selected only the categories that had the words ``Computer Science'' in their names; there were only six such categories among more than 100 categories, as shown in Figure~\ref{fig:W0S-cybernetics-2a}.  A similar figure obtained in Oct 2018 was also shared with Bengio, who had no further comment.  The wave crest in Figure~\ref{fig:W0S-cybernetics-2a} occurred in 2007, with a tiny bump in 1980, but not before 1970 as in Figure~\ref{fig:waves}.

Figure~\ref{fig:W0S-cybernetics-2b} is the histogram for the largest single category Computer Science Cybernetics with 2,665 papers.  In this figure, similar to Figure~\ref{fig:W0S-cybernetics-2a}, the wave crest here also occurred in 2007, with a tiny bump in 1980.

Figure~\ref{fig:W0S-cybernetics-3} is the histogram for the category Computer Science Artificial Intelligence with 601 papers.  Again here, similar to Figure~\ref{fig:W0S-cybernetics-2a} and Figure~\ref{fig:W0S-cybernetics-2b}, the wave crest here also occurred in 2007, with no bump in 1980.
The first document, a 5-year plan report of Latvia, appeared in 1982.  There is a large ``impulse'' of number of papers in 2007, and a smaller ``impulse'' in 2014, but no smooth bump. 
There were no papers for 9 years between 1982 and 1992, in which a single paper appeared in the series ``Lecture Notes in Artificial Intelligence'' on cooperative agents.

Cybernetics, including the original cybernetics moment, as described in \cite{Kline.2015}, encompassed many fields and involved many researchers not working on neural nets, such as Wiener, John von Neumann, Margaret Mead (anthropologist), etc., whereas the physiologist McCulloch co-authored the first ``foundational article of cybernetics on the brain's neural network''.  So it is not easy to attribute even the original cybernetic moment to research on neural nets alone.  Moreover, many topics of interest to researchers at the time involve natural systems (including \cite{McCulloch.1943}), and thus natural intelligence, instead of artificial intelligence.

\end{appendices}

\end{document}